\RequirePackage{fix-cm}
\documentclass[twocolumn]{svjour3}          
\smartqed  
\usepackage{graphicx}
\usepackage{amsmath,amssymb} 
\usepackage{color}
\usepackage{dsfont}
\usepackage[algo2e]{algorithm2e}
\usepackage{algorithmic}
\usepackage{algorithm}
\usepackage{subfigure}
\usepackage{booktabs}
\usepackage{multirow}
\usepackage[pagebackref=true,breaklinks=true,letterpaper=true,colorlinks,bookmarks=false,citecolor=blue]{hyperref}
\usepackage{cite}
\usepackage{natbib}

\begin{document}

\title{You Only Look Yourself: Unsupervised and Untrained Single Image Dehazing Neural Network
}


\author{Boyun Li         \and
        Yuanbiao Gou \and
        Shuhang  Gu \and
        Jerry Zitao Liu \and
        Joey Tianyi Zhou \and
        Xi Peng
}


\institute{B. Li, Y. Gou, and X. Peng \at
               College of Computer Science, Sichuan University. 
              Chengdu, 610065, China.\\
              \email{\{liboyun.gm, gouyuanbiao, pengx.gm\}@gmail.com}           
			\and
			S. Gu \at
			School of Electrical and Information Engineering, University of Sydney.\\
			\email{shuhanggu@gmail.com}
           \and
           J. Liu \at
           TAL AI Lab, Beijing, China\\
           \email{liuzitao@100tal.com}
           \and
           J. Zhou \at
           Institute of High Performance Computing, ASTAR, Singapore\\
           \email{joey.tianyi.zhou@gmail.com}
}

\date{Received: date / Accepted: date}

\maketitle

\begin{abstract}
In this paper, we study two challenging and less-touched problems in single image dehazing, namely, how to make deep learning achieve image dehazing without training on the ground-truth clean image (unsupervised) and a image collection (untrained). An unsupervised neural network will avoid the intensive labor collection of hazy-clean image pairs, and an untrained model is a ``real'' single image dehazing approach which could remove haze based on only the observed hazy image itself and no extra images is used. Motivated by the layer disentanglement idea, we propose a novel method, called you only look yourself (\textbf{YOLY}) which could be one of the first unsupervised and untrained neural networks for image dehazing. In brief, YOLY employs three jointly subnetworks to separate the observed hazy image into several latent layers, \textit{i.e.}, scene radiance layer, transmission map layer, and atmospheric light layer. After that, these three layers are further composed to the hazy image in a self-supervised manner. Thanks to the unsupervised and untrained characteristics of YOLY, our method bypasses the conventional training paradigm of deep models on hazy-clean pairs or a large scale dataset, thus avoids the labor-intensive data collection and the domain shift issue. Besides, our method also provides an effective learning-based haze transfer solution thanks to its layer disentanglement mechanism. Extensive experiments show the promising performance of our method in image dehazing compared with 14 methods on four databases.
\end{abstract}

\section{Introduction}
\label{sec:1}
Haze is a typical atmospheric phenomenon which occurs when the dust, smoke and other particles accumulate in relatively dry air. These particles absorb and scatter the light greatly, thus attenuating the scene radiance reflected from scene point and confuse it with the scattering light. Haze will lead to a decrease in visibility of the scene point and the images captured under this weather condition will become poor in contrast and lose the visual details. Many vision tasks such as object detection would suffer from performance degradation due to these terrible hazy images. Therefore, image dehazing, as a preprocessing step and visual enhancement technology, has been extensively researched and achieves remarkable performance~\citep{Tan, DehazeGAN, DehazeNet, DCPDN, AOD-Net, SFSU, CMAda, NHR}. 

In recent, many researchers have shifted their focus to remove haze from a single image which is more promising but more challenging in practice since without any extra information beyond the observed image itself. A variety of methods have been proposed~\citep{DCP,CAP,GRM,FVR} by employing a widely recognized atmospheric scattering physical model~\citep{Physical-Model}, which could be roughly divided into two categorizes: prior- and learning-based methods.

\begin{figure}[!t]
	\label{Figure:Introduction}
	\begin{center}
	\subfigure[]{
		\label{Figure:First:Haze}
		\includegraphics[scale=0.1]{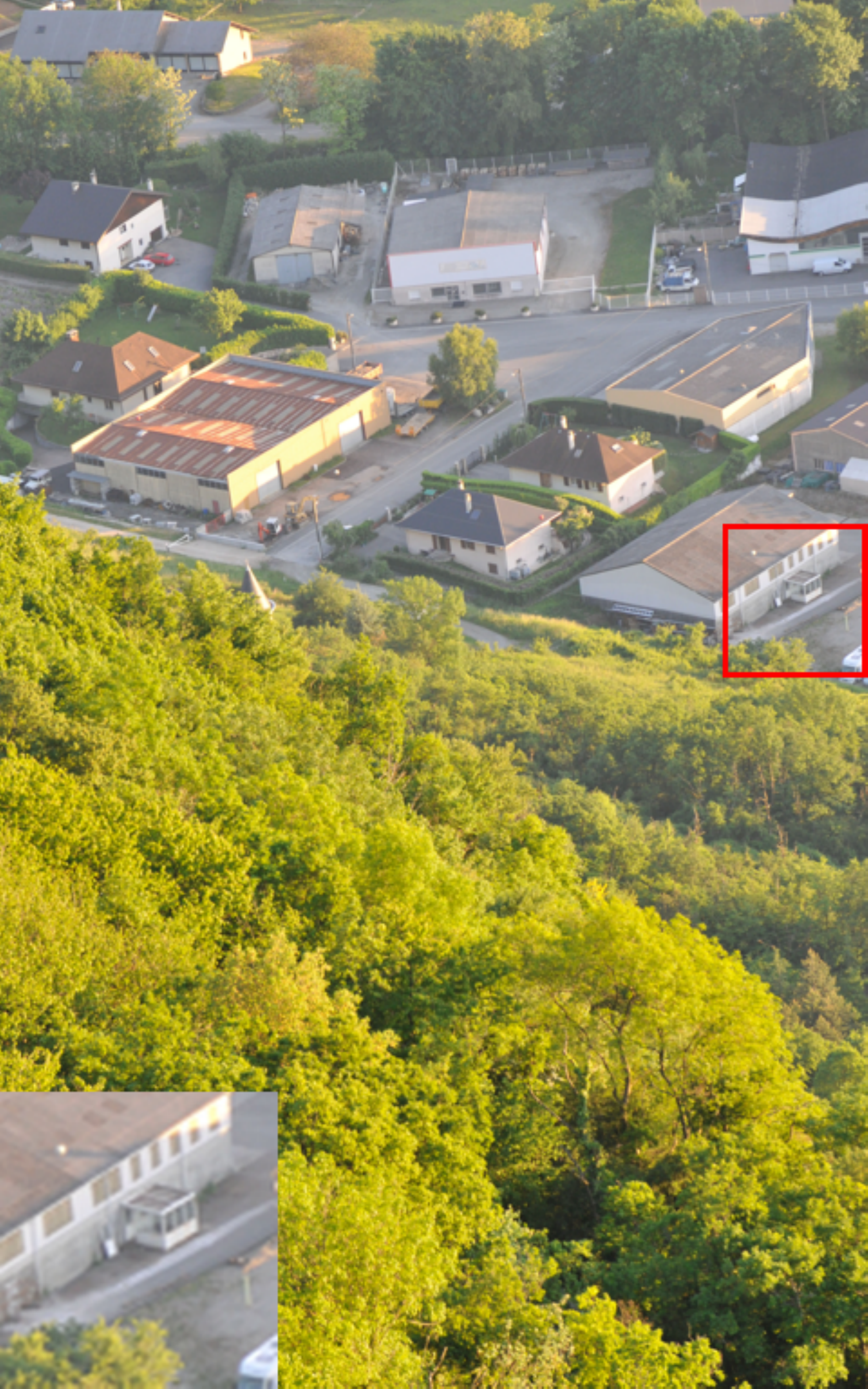}
	}
	\subfigure[]{
		\label{Figure:First:DehazeNet}
		\includegraphics[scale=0.1]{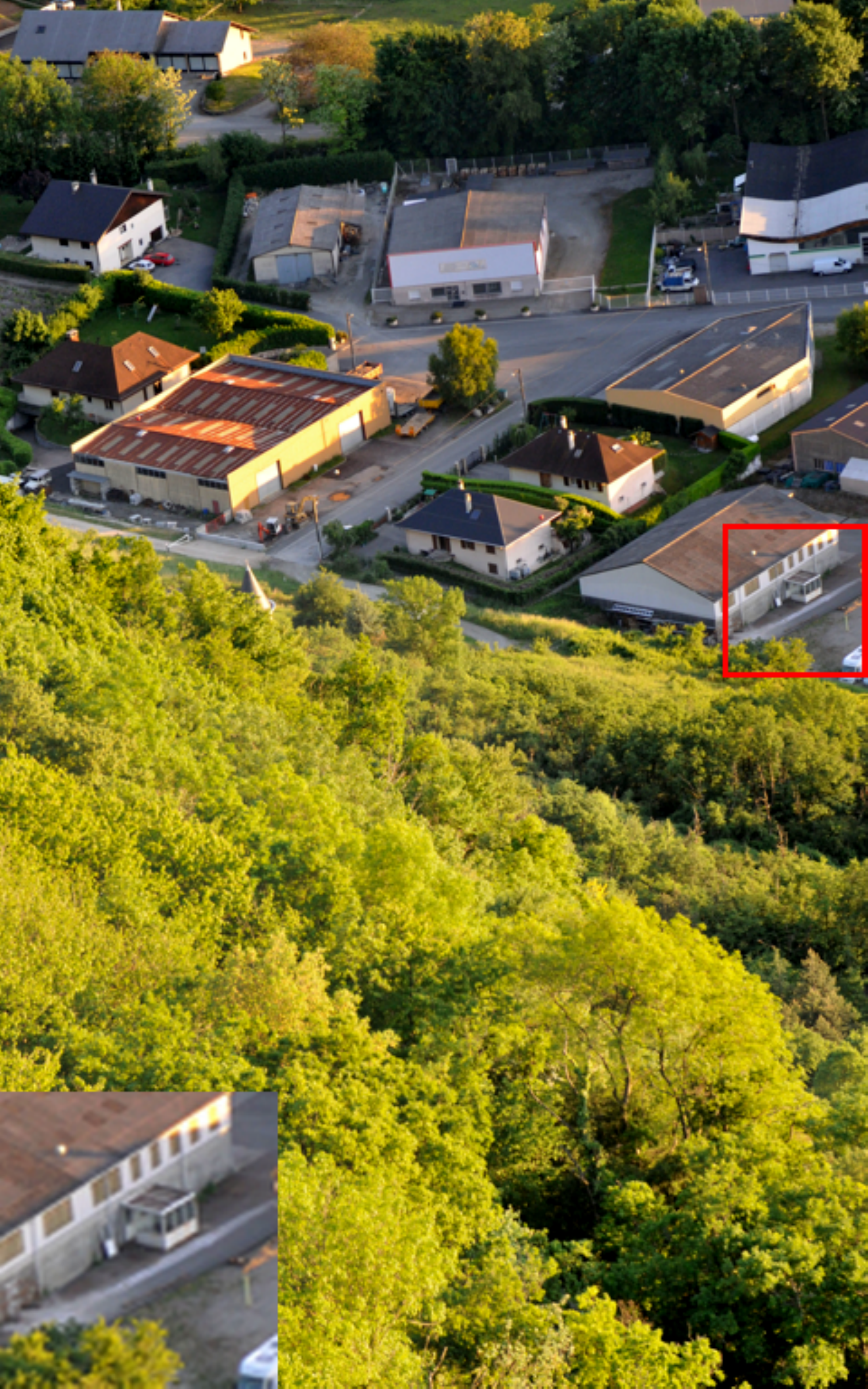}
	}
	\subfigure[]{
		\label{Figure:First:YOLY}
		\includegraphics[scale=0.1]{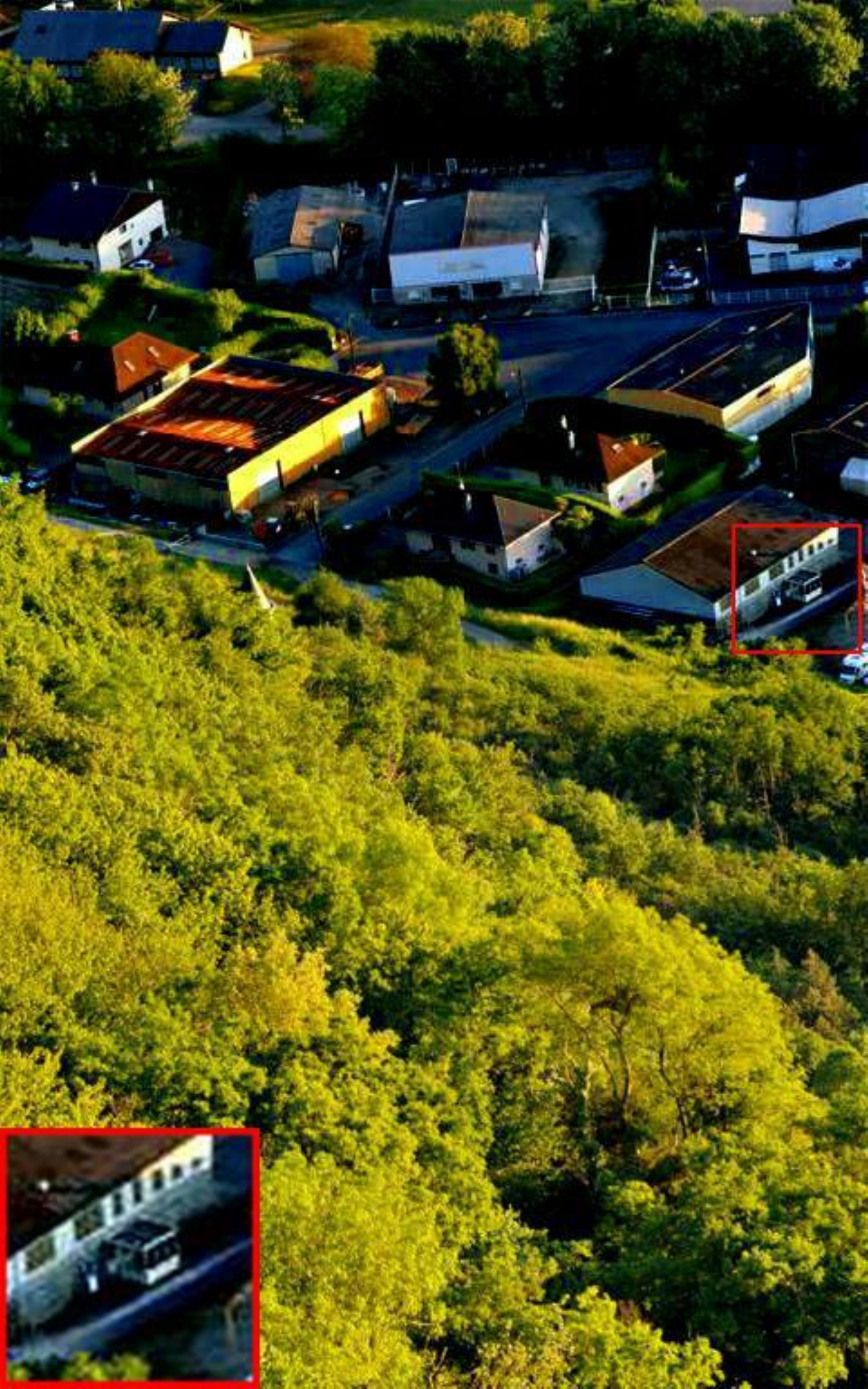}
	}
	\subfigure[]{
		\label{Figure:First:T}
		\includegraphics[scale=0.1]{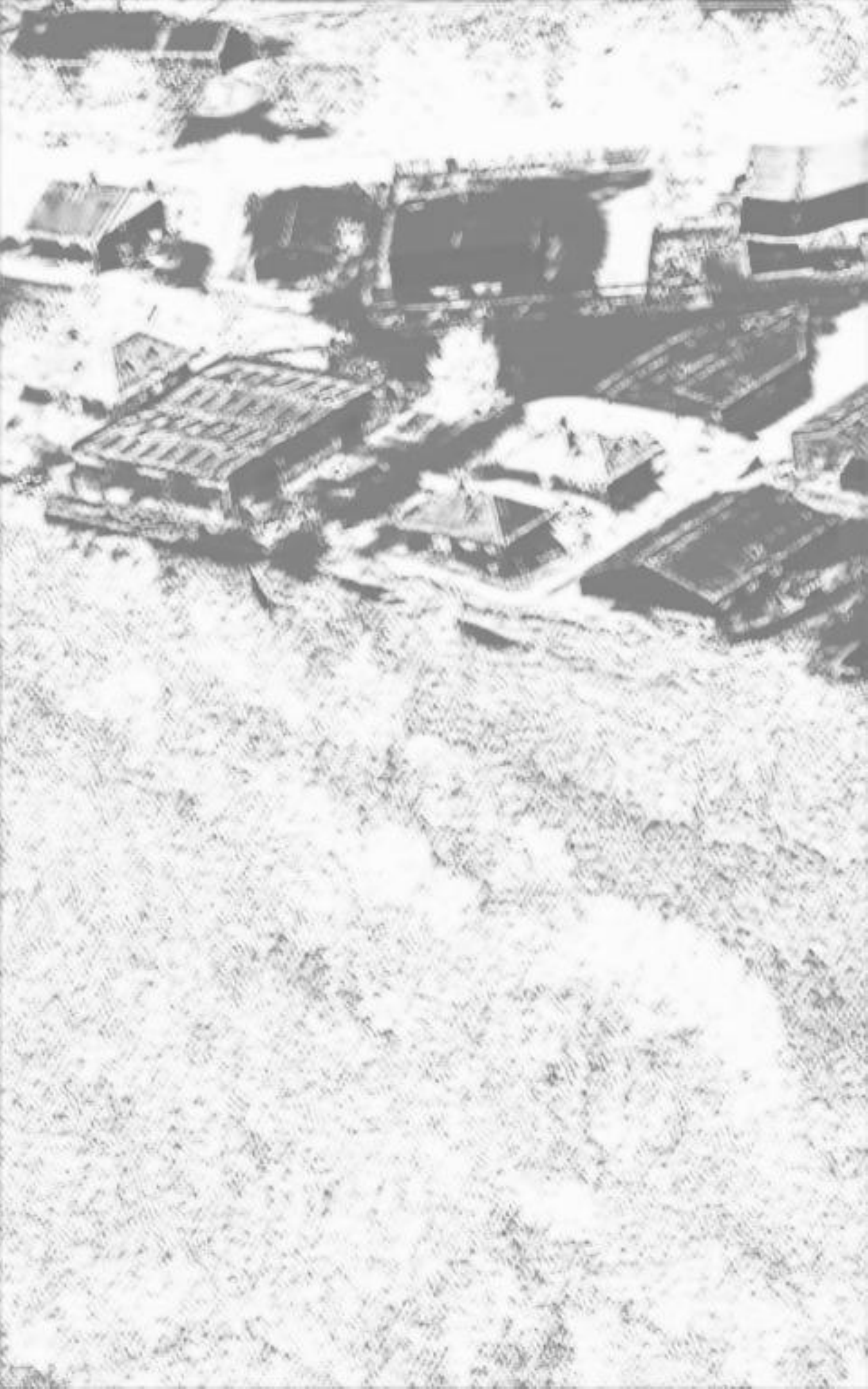}
	}
	\end{center}
	\caption{A visual illustration of the proposed YOLY. a) a real world hazy image; b) the clean image recovered by DehazeNet; c) the clean image recovered by YOLY; (d) the transmission map estimated by YOLY. Note that DehazeNet is pretrained using a collection of hazy-clean image pairs as did in~\citep{DehazeGAN}, whereas our method obtains the result only using the observed hazy image itself. From the results, one could find that our method qualitatively performs better than DehazeNet. Zoom-in is recommended to see the detailed comparisons.}
	\end{figure}

To be specific, prior-based methods employ some handcrafted priors derived from the intrinsic properties of image, such as texture, contrast, and chromatic aberration. For instance, \cite{DCP} observe there exists the dark channel in the local patches of the outdoor haze-free images and accordingly propose using such a dark channel prior (DCP) to estimate the transmission map and atmospheric light for reconstructing the clean image. With the assumption of the image depth is positively correlated to the difference between brightness and saturation, \cite{CAP} propose color attenuation prior (CAP) to estimate the transmission map. Although these methods have achieved remarkable performance, the dehazing quality heavily depends on the consistency between the adopted prior and the actual image properties.

To alleviate the dependence of priors, learning-based methods especially deep learning based models have recently attracted much attention and a lot of efforts have been devoted in recent~\citep{DehazeNet, AOD-Net, GridDehazeNet, DCPDN, EPDN, SUFSPSD, JTME, MSPPDN}. Different from prior-based methods, the parameters of atmospheric scattering model are learned from a large scale training dataset. For example, DehazeNet~\citep{DehazeNet} estimates the transmission map by utilizing a deep neural network trained on a large scale dataset with the supervision of the ground-truth transmission map. Although these learning methods have achieved state-of-the-art performance in image dehazing, almost all of them work in a supervised and trained manner. Namely, they usually require a large scale training dataset which is with some kind of ground truths (\textit{e.g.}, hazy-clean image pair). Once the conditions are unsatisfied, these learning-based methods would be failed. 

In practice, however, it is daunting even impossible to collect a large scale dataset with the desirable ground truth due to the variation in scene and the other factors such as illumination. Therefore, most of methods resort to collect some clean images first and then synthesize the corresponding hazy images via the atmospheric scattering model with the handcrafted parameters. However, the synthesized hazy images are probably less informative and inconsistent with the real hazy images, which would lead to the domain shift issue when the model trained on the synthetic dataset is applied to the real-world hazy images. To address the above issues, it is highly expected to develop a novel deep neural network which could work in an unsupervised and untrained manner simultaneously, while achieving the promising performance. However, to the best of our knowledge, such a challenging task is less touched heretofore.

In this paper, we propose a novel neural network called you only look yourself (\textbf{YOLY}) which employs three joint subnetworks (\textit{i.e.}, J-Net, T-Net, and A-Net) to disentangle a given hazy image into three component layers, \textit{i.e.}, scene radiance layer, transmission map layer, and atmospheric light layer. After that, these three layers are further used to reconstruct the observed input. Thanks to the proposed novel objective function and network structure, YOLY performs image dehazing by only using the information contained in the observed single hazy image. In other words, the proposed network performs like a ``real'' single image dehazing method, which does not follow the conventional paradigm of training the neural network on an image set with some ground truths. The major advantages of our YOLY is that it could avoid the labor-intensive data collection and domain shift issue like existing deep learning based methods, while achieving promising results. Besides, benefiting from the disentanglement of scene radiance, transmission map, and atmospheric light, our method provides an effective way to synthesize new hazy images in a learning-based rather than handcrafted fashion, which could also be the first study afak. 

To summarize, the contributions are given as follows: 
\begin{itemize}
    \item To the best of our knowledge, this work could be one of the first unsupervised and untrained neural networks for single image dehazing, which removes haze without training on a image collection and the ground-truth clean image pairs. To the best of knowledge, the closest work is DDIP~\citep{Double-DIP} which is with significantly difference with the proposed YOLY and the discussion on their distinction is presented in Section~\ref{sec:2}.
    \item A new neural network (\textit{i.e.}, YOLY) is proposed, which consists of three joint disentanglement subnetworks. In brief, two non-degenerate convolutional neural networks (J-Net and T-Net) are used to obtain the clean image and the transmission map, and a variational auto-encoder (A-Net) is used to obtain the atmospheric light. The whole network is learned in a self-supervised manner, which only explore and exploit the information rooted in the observed hazy image as the supervision. 
    \item Our method could also be used to synthesize new hazy images by transferring the haze from one image to another clean image, thus avoiding handcrafting the parameters of the physical model. 
\end{itemize}

\section{Related Work}
\label{sec:2}
In this section, we briefly introduce some existing image dehazing methods from the perspective of prior- and learning-based category. Besides, we will also introduce recent development in unsupervised neural networks for image enhancement. 

\subsection{Prior-based methods} 

Almost all existing prior-based methods are shallow models, which aim to explore various of handcrafted priors from the hazy-free images, such as texture, contrast and chromatic aberration. With the prior, the transmission map and atmospheric light could be estimated, and thus the under-constrained dehazing problem could be well posed. Following this paradigm, dark channel prior~\citep{DCP} was proposed based on the following observation. Namely, most local patches in haze-free outdoor images contain some pixels with very low intensities in at least one color channel. By using such a prior, DCP successfully estimate the transmission map and atmospheric light of hazy image. Besides, a variety of priors are employed based on different observations/assumptions, such as color attenuation prior (CAP)~\citep{CAP}, non-local color prior (NCP)~\citep{NLD} and haze-line prior (HLP)~\citep{HLP}.

Although promising performance has been achieved by these prior-based methods, the dehazing performance is not always desirable due to the inconsistency between the adopted prior and changeable environment. Moreover, these methods are shallow models which might be with limited capacity of handling complex data. 

\subsection{Learning-based methods}
Different from the prior-based methods, learning-based methods adopt a data-driven manner to learn the transmission map and/or atmospheric light. In recent, motivated by the success of neural networks, some studies~\citep{DehazeNet, MSCNN, DehazeGAN, AOD-Net, GridDehazeNet, EPDN} have been conducted to apply neural networks to image dehazing, which have achieved the state-of-the-art performance. For instance, \cite{DehazeNet} proposed DehazeNet which utilizes a trainable convolutional neural network to estimate the transmission map under the supervision of the ground-truth transmission maps. \cite{MSCNN} proposed a multi-scale convolutional neural network wherein a coarse- and fine-scale network are combined to estimate the transmission map for dehazing. \cite{DehazeGAN} proposed simultaneously estimating the transmission map and the atmospheric light by using a generative adversarial network. In summary, like the neural network in other tasks, the success of these deep deep image dehazing methods also rely on a large scale training dataset which is with some truthful supervisors such as the hazy-clean image pairs.

The differences of our method with these existing deep learning based methods~\citep{DehazeNet, MSCNN, DehazeGAN, AOD-Net, GridDehazeNet, EPDN} are given in the following aspects: 1) the proposed YOLY works in an unsupervised rather than supervised manner. In other words, our method does not need the hazy-clean pair images; 2) YOLY is an ``untrained'' instead of ``trained'' model. In other words, YOLY does not require training on a dataset, which could directly handle each single hazy image when it is observed. These two advantages make our method avoid the labor-intensive data collection and the domain-shift issues of using the synthetic hazy images to address the real-world images; 3) our method could disentangle the clean image, the transmission map, and the atmospheric light from the hazy image. This makes transferring haze from the real scenes to another image possible. Namely, our method provides an effective solution to synthesize new hazy images in a data-driven rather than human-specific way. 

\subsection{Unsupervised Deep Image Enhancement Methods}
Although there are only few efforts~\citep{Double-DIP} in developing unsupervised approach for single image dehazing, some methods have been proposed for other image enhancement tasks in  recent~\citep{N2N,N2V,DIP,Double-DIP,DD}. For example, Noise2Noise (N2N)~\citep{N2N} shows that simple statistical arguments lead to new capabilities in data denoising using neural networks. However, it has to use a corrupted image set with the same noise distribution to train the neural network. In other words, it cannot handle the single image case like our method does despite the difference in tasks. Deep image prior (DIP)~\citep{DIP} is another recently proposed unsupervised method which fits a corrupted image using a random noise vector and early-stoping strategy to recover the clean image. However, it is a daunting task to determine the training epoch at which the desirable result is obtained. 

Although both the aforementioned methods and our YOLY are unsupervised methods, there are largely different. First, most of the aforementioned methods are not specifically designed for single image dehazing. Will be shown in our experiments, they cannot achieve encouraging performance in such a challenging task (haze is a kind of signal-dependent noises). Second, the methods such as N2N and its variants require using a data collection for training, whereas our method will only use the observed hazy image itself. Third, our method is based on the layer disentanglement idea which is different from these methods in the methodology. It should be pointed out that our method is also remarkably different from the recently proposed DDIP~\citep{Double-DIP} in the following two aspects. On one hand, the loss and the network structure are totally different. To be specific, our method employs variational inference to model the atmospheric light, whereas DDIP adopts a U-Net-like structure to fit the image same with DIP. Moreover, our YOLY utilizes the color attenuation as a supervisor to estimate the clean image, whereas DDIP mainly employs the early-stopping fitting strategy~\citep{DIP}. On the other hand, the input and the working mechanism are different. In brief, DDIP takes three random noises as inputs and feeds them into three generator networks to fit the hazy image, which utilizes the properties of DIP. In contrast, our method directly feeds the hazy image as the conditional input into three subnetworks so that different layers are disentangled. In other words, DDIP performs layer composition in a bottom-to-top fashion, whereas our YOLY performs layer disentanglement in a top-to-bottom fashion.

\begin{figure*}[!th]
	\begin{center}
		\includegraphics[scale=0.6]{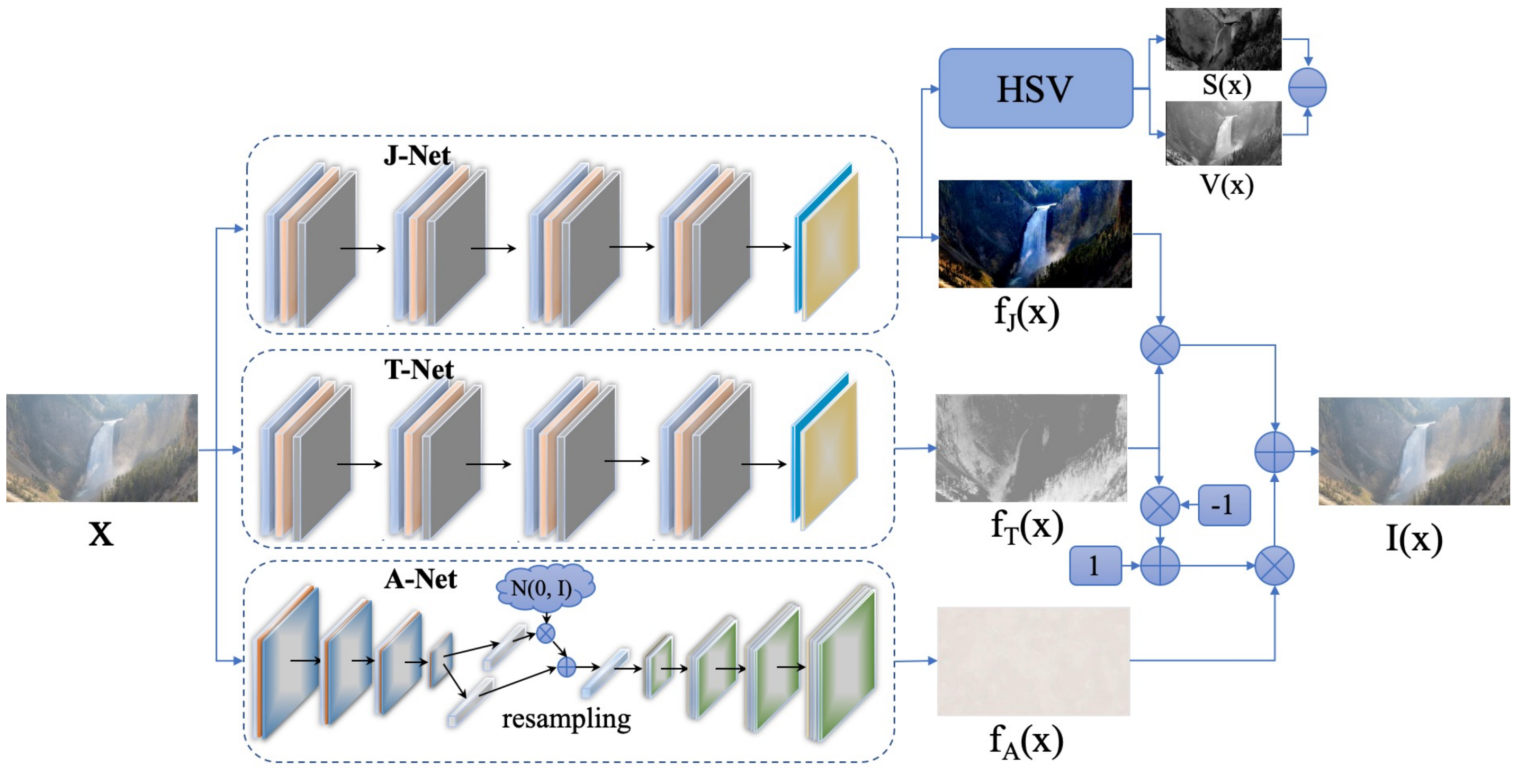}
	\end{center}
	\caption{\label{Figure:Net}The architecture of YOLY. In brief, YOLY consists of three joint learning subnetworks which are the clean image estimation network (J-Net), the transmission map estimation network (T-Net), and the atmospheric light estimation network (A-Net). Taking a single hazy image as the input, these three subnetworks disentangle the input into three different layers which are then utilized to reconstruct the input hazy image in a decomposition-composition fashion. Note that, the atmospheric light $f_{A}(x)$ would be approximately even exact homogeneous if the transmission map $f_{T}(x)$ could be well recovered as shown in the figure. HSV denotes the operation of transforming the recovered image to its HSV version. }
\end{figure*}

\section{Proposed Method}
\label{sec:3}

Given a single hazy image $x$ as the input, we aim to recover the clean image $J(x)$ without using the information beyond the image content itself. The basic idea of our methods is to disentangle $x$ into three layer components using three joint subnetworks as shown in Fig~\ref{Figure:Net}. More specifically, YOLY simultaneously feeds $x$ into a clean image estimation network (J-Net), a  transmission map estimation network (T-Net), and an atmospheric light estimation network (A-Net). After that, the outputs of them are further combined to reconstruct $x$ at the top of YOLY through the atmospheric scattering physical model. In such a way, the whole model is learned in an unsupervised fashion and these subnetworks are optimized in an end-to-end manner. Formally, at the top layer of YOLY, we aim to minimize the following loss:
\begin{equation}
	\label{eq:L_Rec}
	\mathcal{L}_{Rec} = \|I(x)-x\|_{p},
\end{equation}
where $\|\cdot\|_{p}$ denotes $p$-norm of a given  matrix. In this paper, we simply adopt Frobenius norm. $I(x)$ is computed by composing the outputs of the three subnetworks via
\begin{equation}
	\label{physical model}
	I(x)=J(x)T(x)+A(1-T(x))
\end{equation}
where $J(x)$ denotes the clean image predicted by J-Net, $T(x)$ is the medium transmission map predicted by T-Net, and $A$ is the global atmospheric light on each pixel coordinates predicted by A-Net. It should be pointed out that some algorithms~\citep{DehazeGAN} have been proposed by learning these factors based on the above physical model, however, to the best of our knowledge, there is few efforts have been devoted to developing unsupervised methods so far. 

The loss $\mathcal{L}_{Rec}$ is designed to constrain the entire network including the subnetworks to well reconstruct the hazy image $x$ after layer disentanglement. In other words, it guides the layer disentanglement and composition through incorporating the haze generation process. In the following, we will illustrate how these three networks could obtained desired layer components, besides directly utilizing the above self-supervision.   

\subsection{J-Net}
J-Net aims to predict the clean image $J(x)$ from the hazy image $x$. As demonstrated in~Fig.(\ref{Figure:Net}), J-Net takes a non-degenerate architecture by following~\citep{DCGAN}. Namely, our J-Net does not implement the down-sampling operation, thus preventing the loss of detail in $J(x)$. More specifically, J-Net only consists of the convolutional layer, batch normalization layer~\citep{BN}, and LeakyReLU activation. In the last layer, we choose sigmoid function to normalize the output into $[0, 1]$. More details about the network implementation could refer to the experimental setting and supplementary materials. 

To supervise J-Net, we propose the following loss function 
\begin{equation}
	\label{eq:LJ}
	\mathcal{L}_{J} = \|V(f_{J}(x)) - S(f_{J}(x))\|_{p},
\end{equation}
where $V(f_{J}(x))$ denotes the brightness of $f_{J}(x)$, and $S(f_{J}(x))$ denotes the saturation of $f_{J}(x)$.

The loss term $\mathcal{L}_{J}$ is designed based on the observation in~\citep{CAP}. Namely, the depth of clean image is positively correlated to the difference between brightness and saturation of a clean image. To utilize this prior in an unsupervised manner, we recast the prior as the above formulation, \textit{i.e.}, the difference between the value and saturation should be as small as possible in the predicted $J(x)$. Eq.~(\ref{eq:LJ}) has two advantages. On one hand, the formulation has sub-gradients and pluggable into our model to enjoy the joint optimization through back-propagation.  On the other hand, it makes possibility to  recover the clean image without using the ground-truth clean image.

\subsection{T-Net}
As the clean background and the transmission map are dependent of the input $x$, we adopt the similar network structure for J-Net and T-Net. There are only two differences between them. To be specific, the output layer of J-Net is with three channels, whereas the output layer of T-Net is with only one channel for computational efficiency. On the other hand, T-Net does not employ an explicit loss, which only utilizes the self-supervision back-propagated from the top layer of YOLY to guide the optimization.

\subsection{A-Net}

A-Net aims to estimate the global atmospheric light from the observed  images. As the global atmospheric light $A$ is independent of the image content and owns the global property, it is reasonable to assume that $A$ is sampled from a latent Gaussian distribution. Accordingly, we recast the learning of $A$ as a variational inference problem~\citep{VAE}. To be specific, A-Net consists of an encoder, a symmetric decoder, and an intermediate block. Both the encoder and the decoder consist of four blocks. In the encoder, the blocks consist of a convolutional layer, a ReLU activation function, and a max pooling layer in sequence. In the decoder, the blocks sequentially perform upsampling, convolution, batch normalization, and ReLU activation. To learn the latent Gaussian model, the intermediate block is used to transform the output (\textit{i.e.}, $z$) of the encoder to the latent Gaussian distribution $\mathcal{N}(\mu_{z}, \sigma_{z}^{2})$, \textit{i.e.}, $z\rightarrow \{\mu_{z}, \sigma_{z}^{2}\}$, where $\mu_{z}$ and $\sigma_{z}^{2}$ are mean and variance of the learned Gaussian model. Through resampling from the Gaussian model, the reconstruction of the latent code could be generated, \textit{i.e.}, $\mathcal{N}(\mu_{z}, \sigma_{z}^{2})$ $\rightarrow\hat{z}$. After that, $\hat{z}$ is fed into the decoder to obtain the reconstruction of the disentangled atmospheric light $f_{A}(x)$. 

The loss function for A-Net is formulated as below:
\begin{equation}
\label{eq:LA}
	\mathcal{L}_{A}=\mathcal{L}_{H}+\mathcal{L}_{KL}+\lambda\mathcal{L}_{Reg},
\end{equation}
where $\mathcal{L}_{H}$ is the loss between the disentangled atmospheric light $f_{A}(x)$ and the initial hint $A(x)$, where $A(x)$ is estimated from $x$. $\mathcal{L}_{KL}$ and $\mathcal{L}_{Reg}$ denote the loss for variational inference and regularization term, respectively. $\lambda$ is a nonnegative parameter to balance the regularization.   

To be exact, $\mathcal{L}_{H}$ is defined as
\begin{equation}
\label{eq:LH}
	\mathcal{L}_{H}=\|f_{A}(x)-A(x)\|_{F}.
\end{equation}

$\mathcal{L}_{KL}$ aims to minimize the difference between the latent code $z$ and the corresponding reconstruction $\hat{z}$ re-sampled from the Gaussian model.  To enjoy the end-to-end optimization using the standard stochastic gradient methods, the reparameterization trick could be used to yield a lower bound estimator~\citep{VAE}. Mathematically, 
\begin{align}
\label{eq:KL}
	\mathcal{L}_{KL}&=KL(\mathcal{N}(\mu_z, \sigma^2_z) \arrowvert\arrowvert \mathcal{N}(0, I))\notag\\
	&=\frac{1}{2}\sum_{i}\left((\mu_{z_{i}})^{2}+(\sigma_{z_{i}})^{2}-1-\log(\sigma_{z_{i}})^{2}\right)
\end{align}
where $KL(\cdot)$ denotes the Kullback-Leibler divergence between two distributions, $z_{i}$ denotes  the $i$-th dimension of $z$. 

To avoid overfitting, we enforce $\mathcal{L}_{Reg}$ on the outputs of A-Net, \textit{i.e.}, $f_{A}(x)$. Formally, 
\begin{equation}
	\label{eq:LReg}
	\mathcal{L}_{Reg}(x)=\frac{1}{2m}\sum_{i=1}^{m}(x_{i}-\frac{1}{|N(x_{i})|}\sum_{y_{i}\in N(x_{i})}y_{i})^{2},
\end{equation}
where $\mathcal{N}(x_{i})$ is the second order neighborhood of $x_{i}$, $|\mathcal{N}(x_{i})|$ is the neighborhood size, and $m$ denotes the pixel number of $x$. Clearly, the regularizations play a role of mean filtering, which enforce $A(x)$ to be smooth. Note that, the high-frequency details of the recovered haze-free image might lose if the above regularization is enforced on the output of J-Net.

In summary, the total loss of our YOLY is as below:
\begin{align}
	\label{eq:finalloss}
	\mathcal{L}
	=&\mathcal{L}_{Rec}+\mathcal{L}_{J}+\mathcal{L}_{H}+\mathcal{L}_{KL}+\lambda\mathcal{L}_{Reg}\notag\\
	=&\|I(x)-x\|_{F}^{2}+\|V(f_{J}(x))-S(f_{J}(x))\|_{F}^{2}\notag\\
	&+\|f_{A}(x)-A(x)\|_{F}^{2}\notag\\
	&+\sum_{i}\left((\mu_{z_{i}})^{2}+(\sigma_{z_{i}})^{2}-1-\log(\sigma_{z_{i}})^{2}\right)\notag\\
	&+\frac{\lambda}{m}\sum_{i=1}^{m}(f_{A}(x)_{i}-\frac{1}{|N(f_{A}(x)_{i})|}\sum_{y_{i}\in N(f_{A}(x)_{i})}y_{i})^{2}
\end{align}

\section{Experiments}

In this section, we evaluate our method on two synthetic datasets and two real-world datasets, comparing with 14 baseline methods in terms of PSNR and SSIM. In the following, we will first demonstrate the experimental setting, and then show the qualitative and quantitative results on the datasets, as well as the time cost of YOLY. After that, we will show the effectiveness of our method for haze transfer. Finally, the ablation study is presented to verify the effectiveness of the proposed method.  

\subsection{Experimental Settings}
In this part, we elaborate on the used datasets, baselines, the evaluation metrics, and the implementation details. 

\textbf{Datasets:} 
RESIDE~\citep{RESIDE} is a new large scale haze image dataset, of which the testing subsets consist of Synthetic Objective Testing Set (SOTS) and Hybrid Subjective Testing Set (HSTS). To be specific, SOTS consists of 500 indoor hazy images, which are synthesized by the physical model with handcrafted parameters. HSTS contains 10 synthetic haze images and 10 real-world hazy images. In our experiments, we take SOTS and HSTS for evaluations. Besides, we also manually collect 10 hazy real-world images from Internet for a more comprehensive investigation.  

\textbf{Baselines:} 
For comprehensive comparisons, we compare the proposed YOLY with 14 methods which are divided into three groups, namely, four learning-based dehazing methods, five prior-based dehazing methods and five unsupervised deep image enhancement methods. It should be pointed out that, both the prior-based and unsupervised deep methods remove haze from image without using the ground-truth clean image, and their major difference is that the former is the shallow model whereas the latter is based on deep neural networks.

To be specific, the learning-based dehazing methods are  DehazeNet~\citep{DehazeNet}, MSCNN~\citep{MSCNN}, AOD-Net~\citep{AOD-Net} and CAP~\citep{CAP}. Here, although CAP use prior information, we still classify it into the learning-based category because it employs the ground-truth transmission  as the supervisor. The prior-based dehazing approaches are DCP~\citep{DCP}, FVR~\citep{FVR}, BCCR~\citep{BCCR}, GRM~\citep{GRM},  NLD~\citep{NLD}.  The unsupervised deep image enhancement methods contain N2N~\citep{N2N},  N2V~\citep{N2V}, DIP~\citep{DIP},  DD~\citep{DD} and DDIP~\citep{Double-DIP}. Noticed that, N2N, N2V, DD, and DIP are specifically designed for other image enhancement tasks rather than image dehazing, which are compared in our experiments for two reasons. On one hand, there is no unsupervised deep model for singe image dehazing excepted DDIP so far. Thus, the comparisons with them could provide a more extensive study. On the other hand, the inferior performance achieved by them shows that it would achieve undesirable result if these methods are simply applied to the dehazing task. Furthermore, among all the tested methods, only N2V, DD, DIP, DDIP, and our YOLY are untrained neural networks, which means that they do not require training data and only use the given hazy image. 

\textbf{Evaluation Metrics:} Like~\citep{MSCNN,AOD-Net,Double-DIP,EPDN, DCPDN}, two popular metrics are used in the quantitative comparisons, \textit{i.e.}, PSNR and SSIM. Higher value of these metrics indicates better dehazing performance. 

\textbf{Experimental Configurations:} We conduct experiments on an NVIDIA Titan RTX GPU in PyTorch. To optimize YOLY, we employ the ADAM optimizer~\citep{Adam} with the default learning rate and the maximal epoch of 500. For better reproducibility, we do not exhaustively tune parameters for our method and instead fix $\lambda = 0.1$ for all the evaluations. To initialize the hint, we use the method similar to \citep{DCP}. Regarding some of baselines, we directly refer to the best results reported in the original papers. For the baselines without the corresponding results, we carry out them by using the source codes provided by the authors and adopting their parameter settings. The source code of YOLY will be released on Github. 

\begin{table*}
\centering
\caption{Results on the synthetic indoor database (SOTS). The bold number indicates the best method of each category of methods.}
\label{Table:SOTS}
\begin{tabular}{c| cccc ccccc}
\toprule
\multicolumn{1}{c|}{\multirow{2}{*}{Metrics}} & \multicolumn{4}{c|}{Learning-based Dehazing Methods}  & \multicolumn{5}{c}{Prior-based Dehazing Methods} \\
\cline{2-10}
 & DehazeNet & MSCNN & AOD-Net & \multicolumn{1}{c|}{CAP} & DCP & FVR & BCCR & GRM & NLD  \\ 
\hline
PSNR & \textbf{21.14} & 17.57 & 19.06 & \multicolumn{1}{c|}{19.05} & 16.62 & 15.72 & 16.88 & \textbf{18.86} & 17.29\\ 
SSIM & 0.8472 & 0.8102 & \textbf{0.8504} & \multicolumn{1}{c|}{0.8364}  & 0.8179 & 0.7483 & 0.7913 & \textbf{0.8553} & 0.7489\\ 
\midrule
\multicolumn{1}{c|}{\multirow{2}{*}{Metrics}} & \multicolumn{6}{c|}{Unsupervised Neural Networks}  \\
\cline{2-7}
 & N2N & N2V & DIP & DD & \multicolumn{1}{c|}{DDIP} & \multicolumn{1}{c|}{Ours} \\ 
\cline{1-7}
PSNR & 14.49  & 10.67  & 12.28  & 11.92  & \multicolumn{1}{c|}{16.97}  & \multicolumn{1}{c|}{\textbf{19.41}}
\\ 
SSIM & 0.7078  & 0.5397  & 0.5782  & 0.6404  & \multicolumn{1}{c|}{0.7147}  & \multicolumn{1}{c|}{\textbf{0.8327}}  \\ 
\bottomrule
\end{tabular}
\end{table*}

\begin{figure*}[!t]
	\def \m_width{0.075}
	\def \a_height{1cm}
	\def \b_height{1cm}
	\begin{center}
		\subfigure{\includegraphics[width=\m_width\textwidth, height=\a_height]{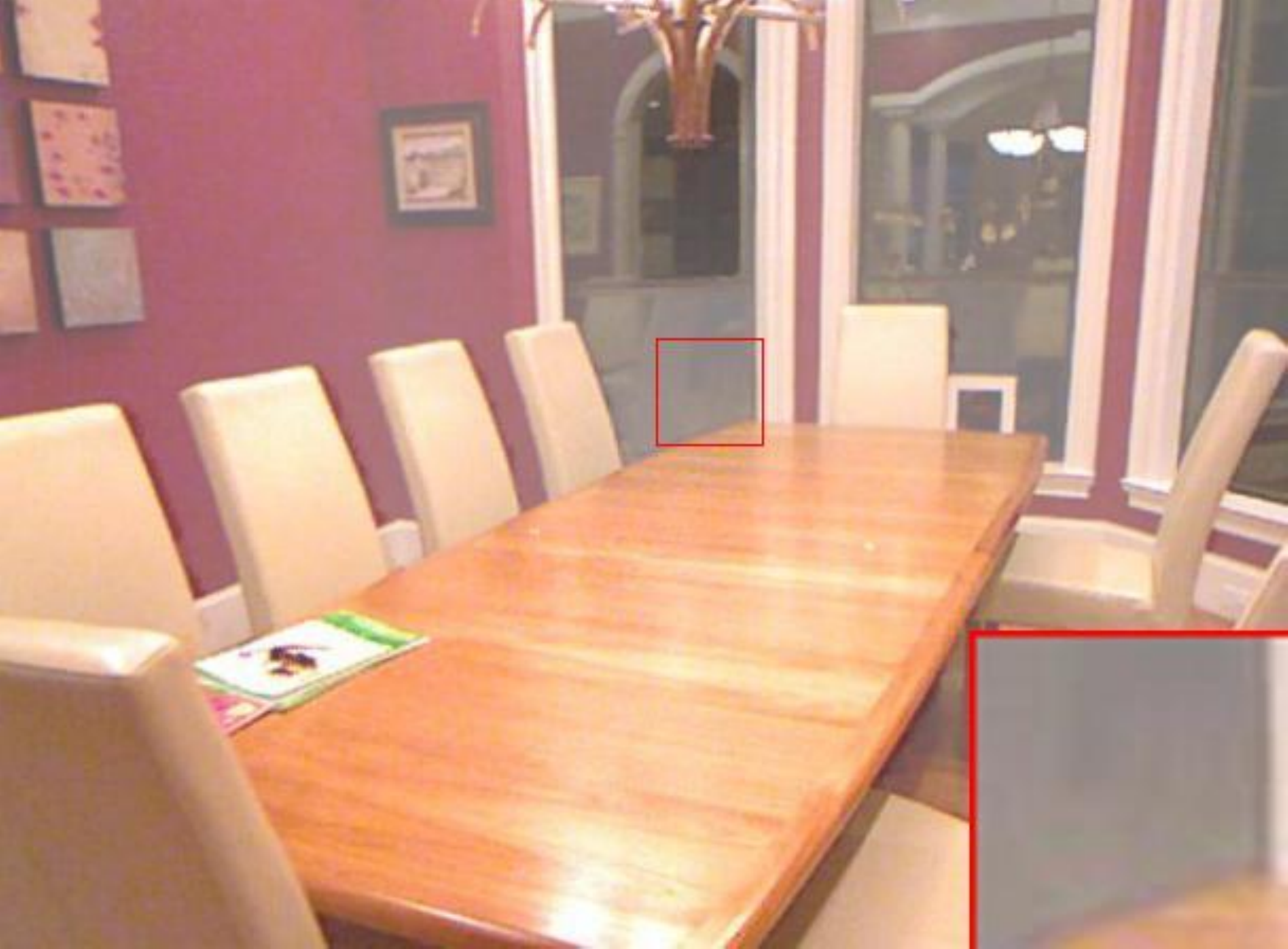}}
		\subfigure{\includegraphics[width=\m_width\textwidth, height=\a_height]{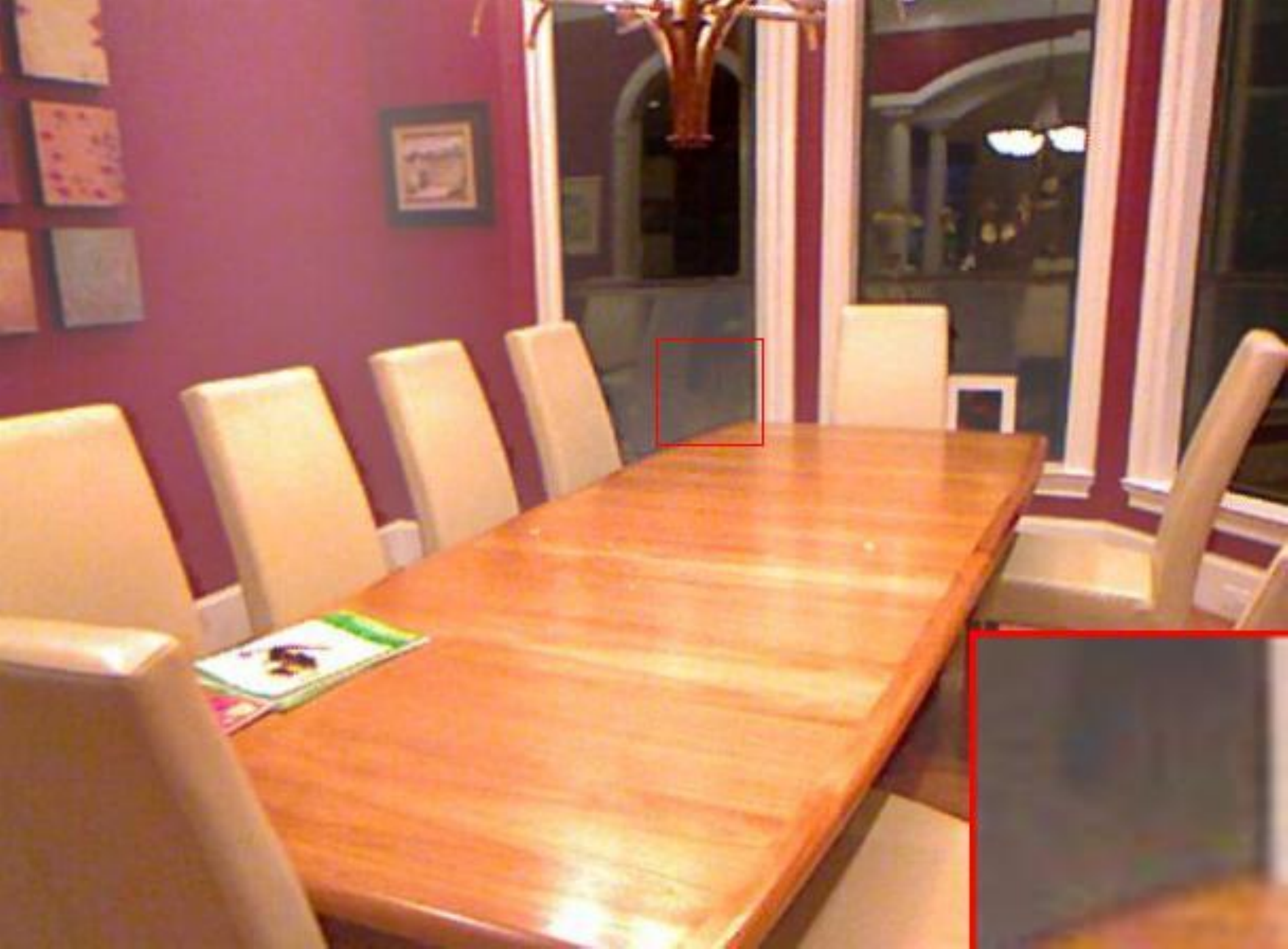}}
		\subfigure{\includegraphics[width=\m_width\textwidth, height=\a_height]{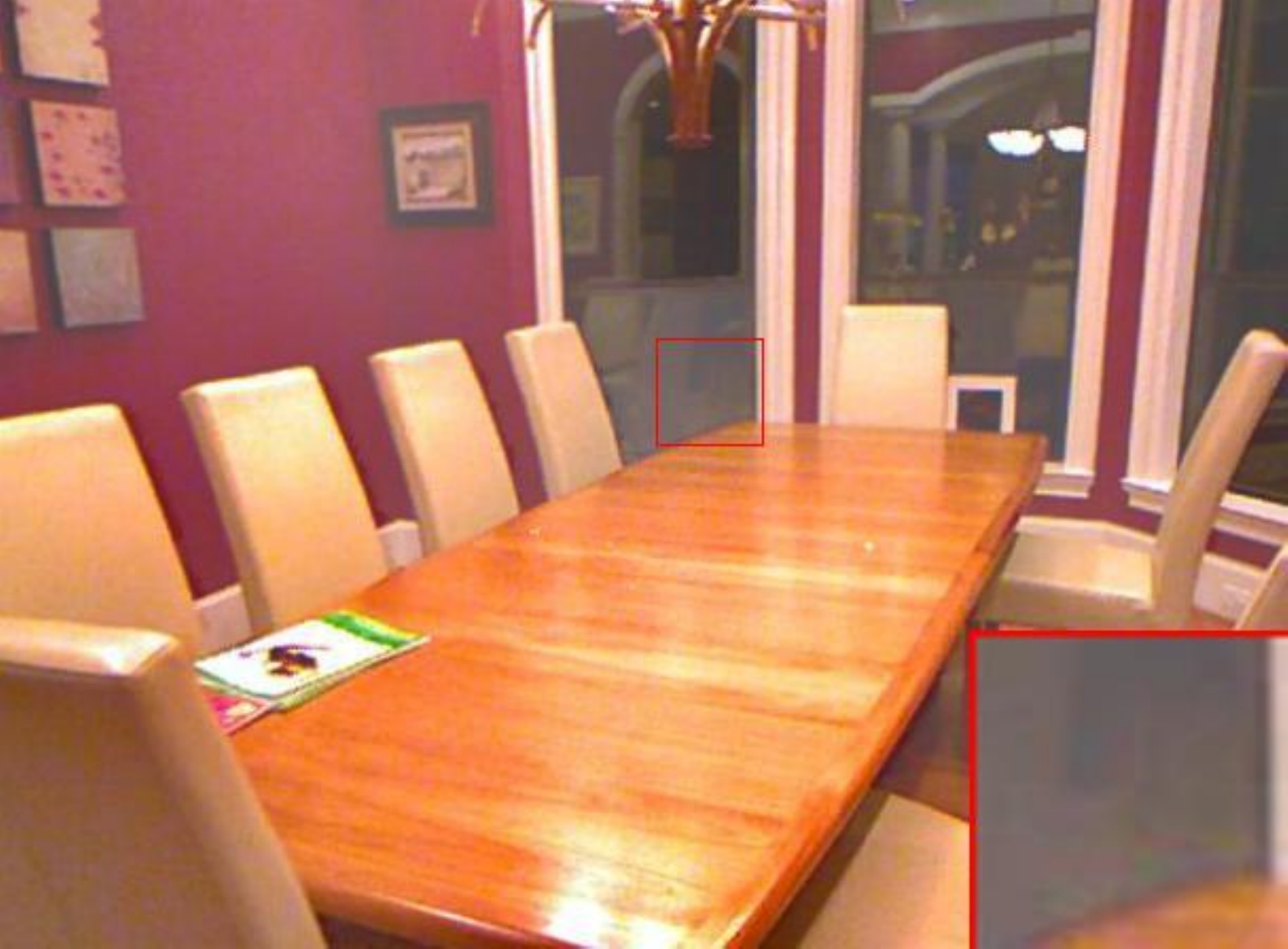}}
		\subfigure{\includegraphics[width=\m_width\textwidth, height=\a_height]{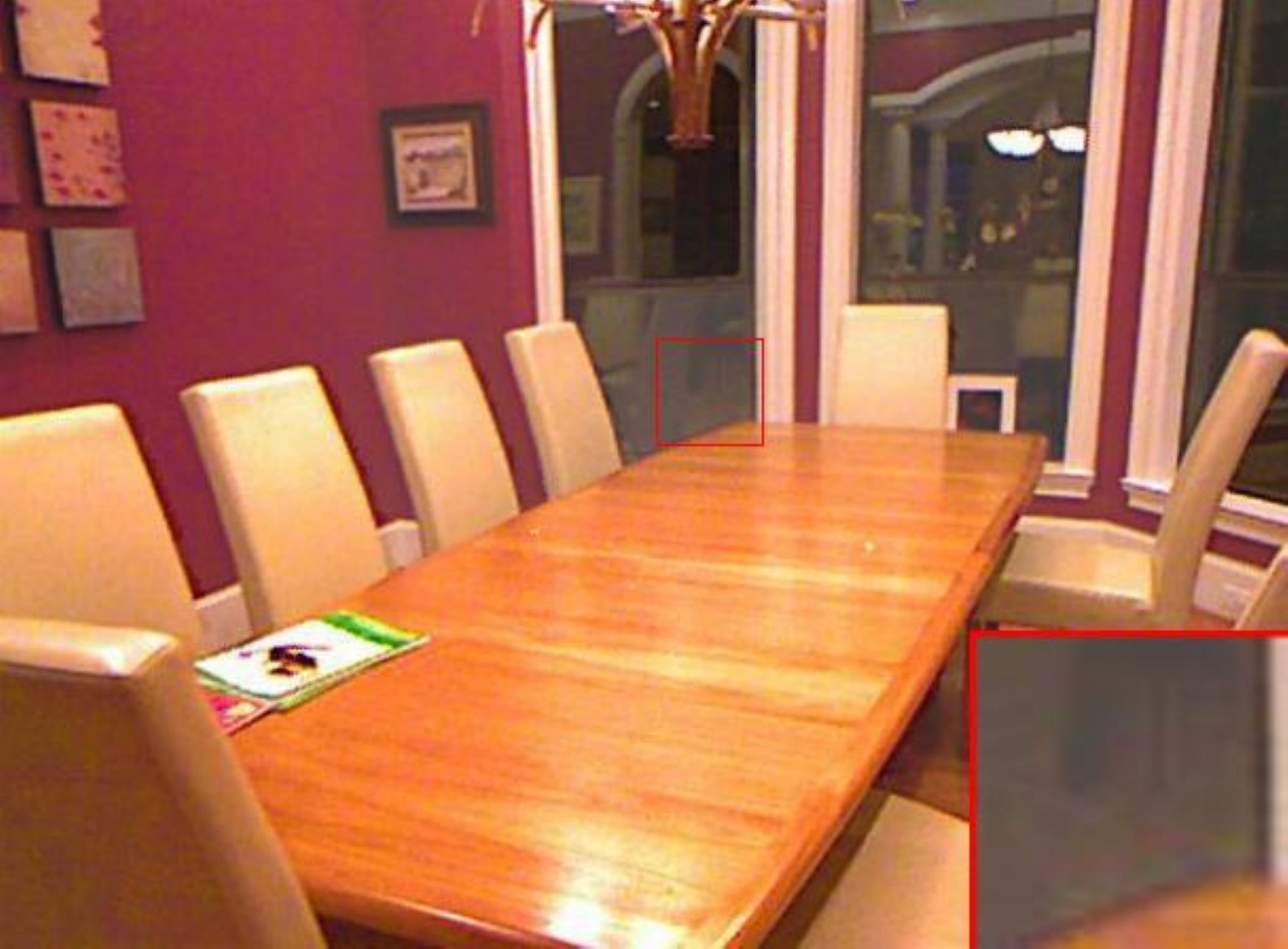}}
		\subfigure{\includegraphics[width=\m_width\textwidth, height=\a_height]{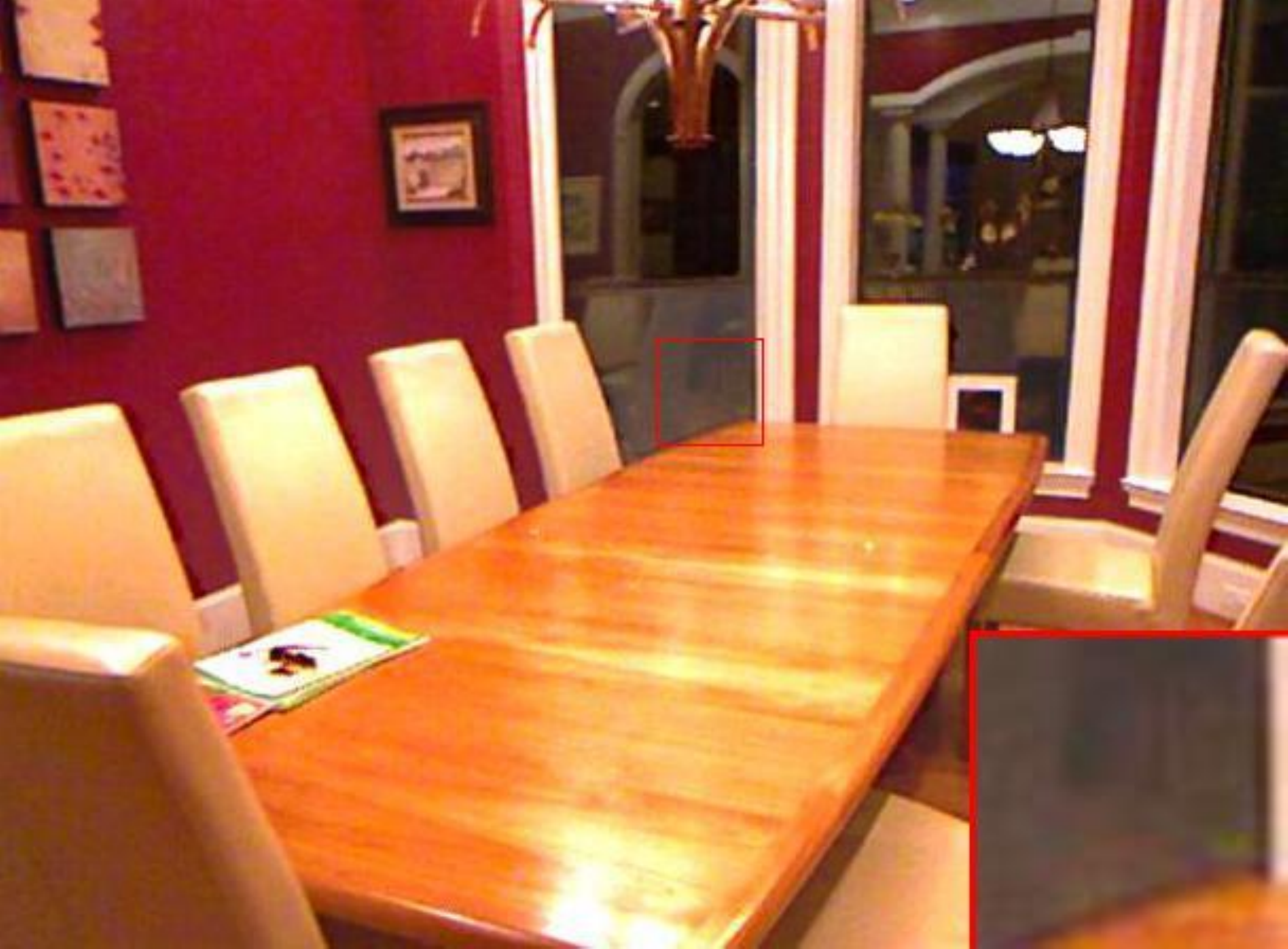}}
		\subfigure{\includegraphics[width=\m_width\textwidth, height=\a_height]{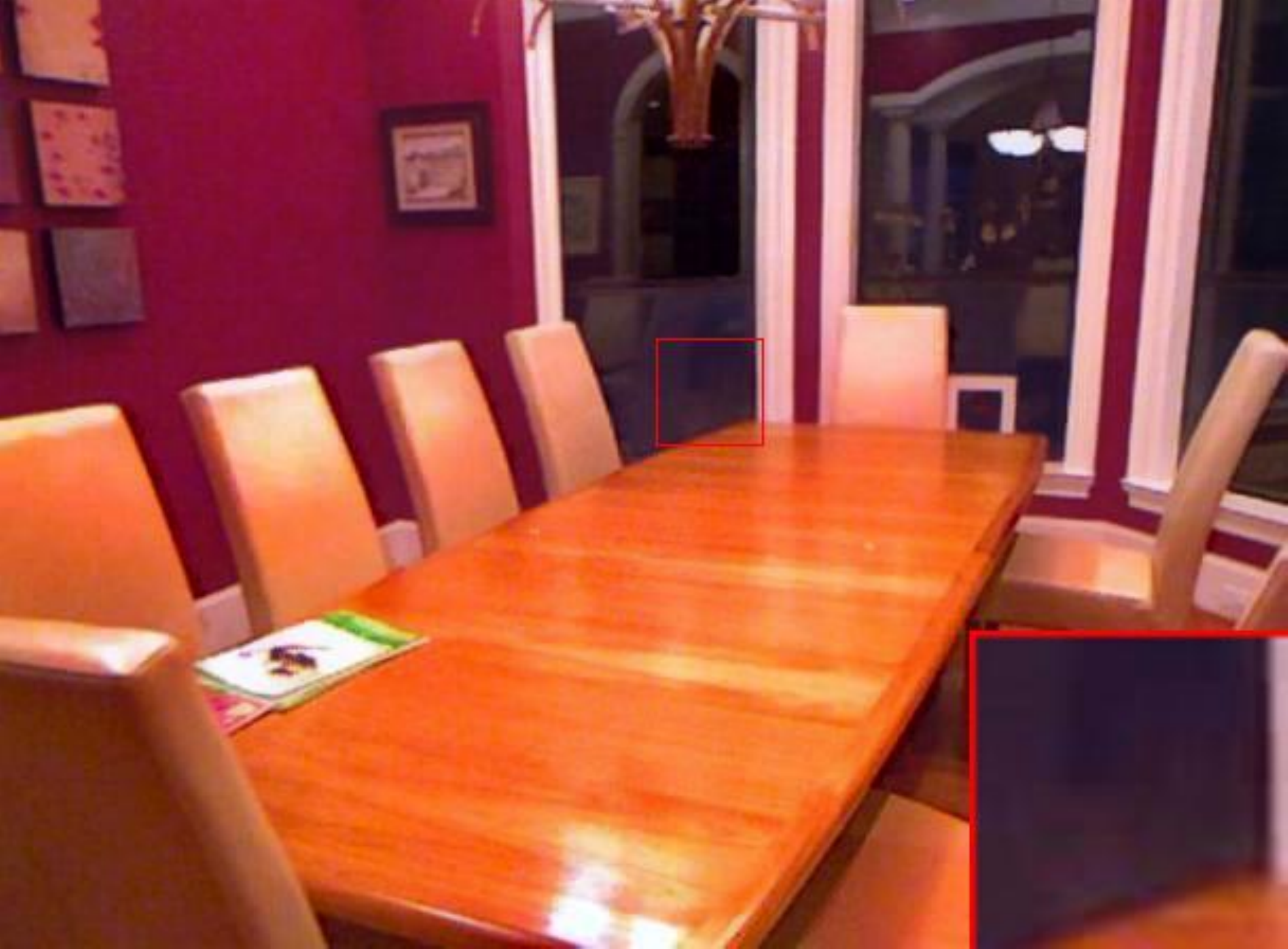}}
		\subfigure{\includegraphics[width=\m_width\textwidth, height=\a_height]{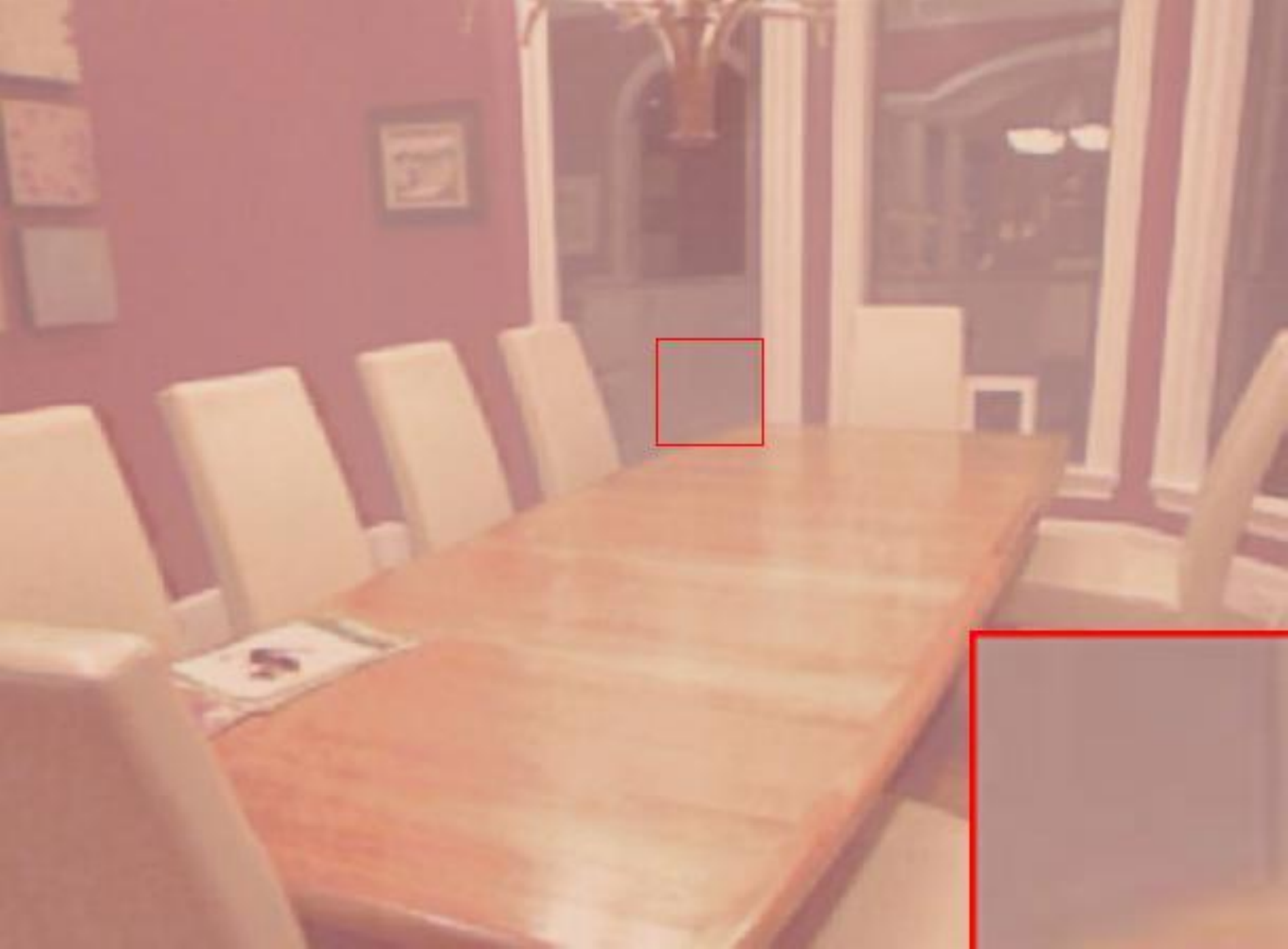}}
		\subfigure{\includegraphics[width=\m_width\textwidth, height=\a_height]{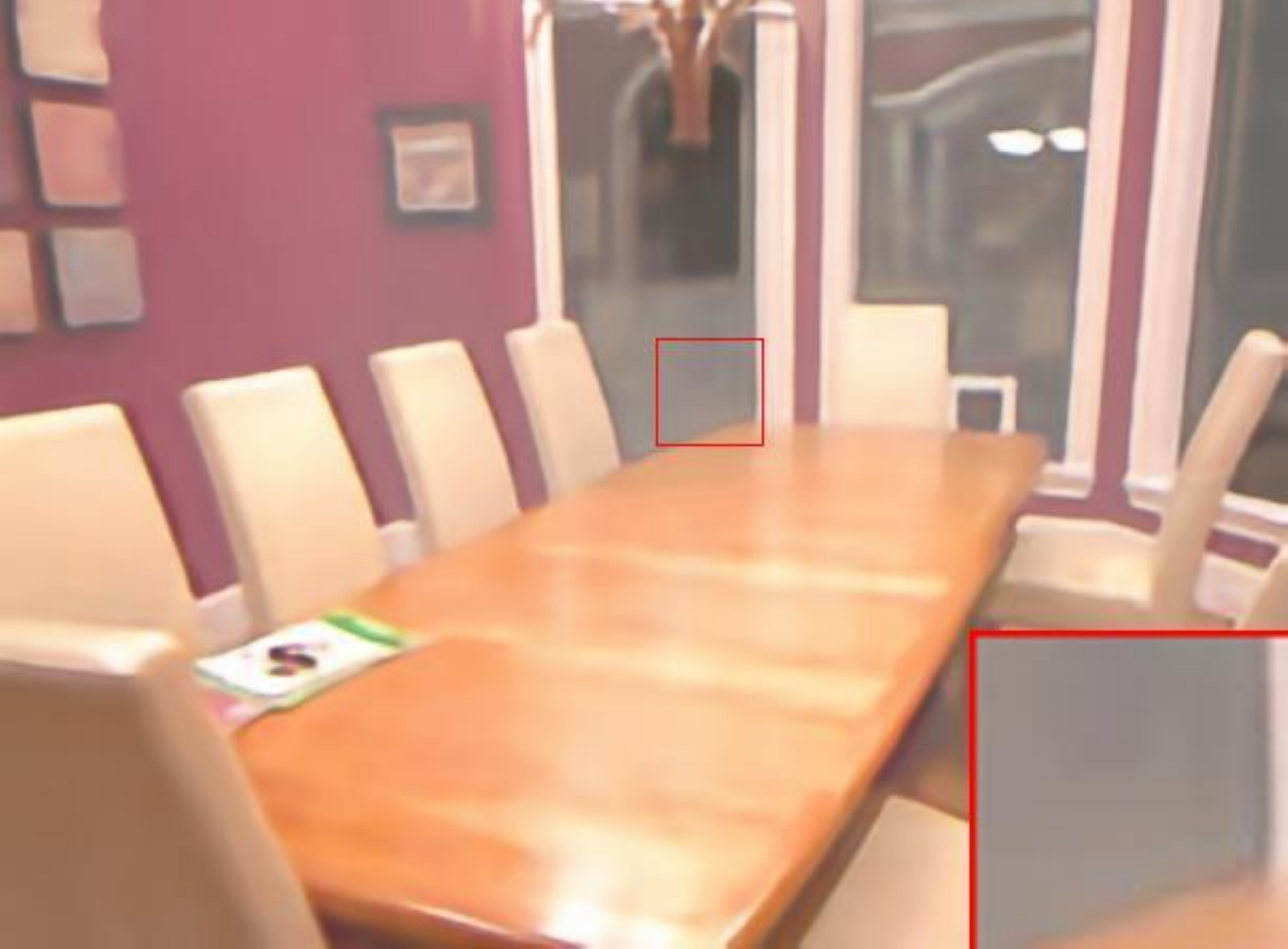}}
		\subfigure{\includegraphics[width=\m_width\textwidth, height=\a_height]{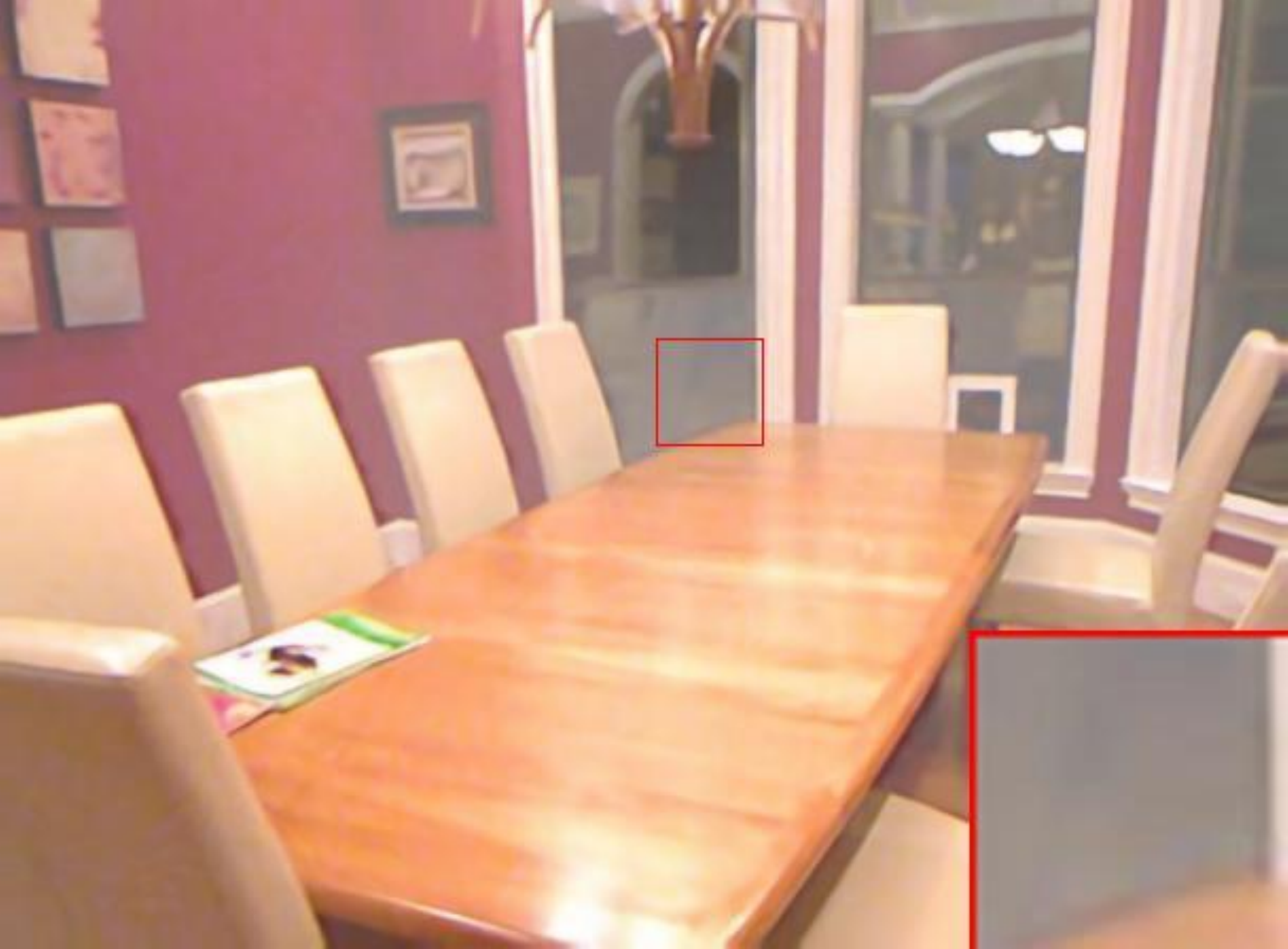}}
		\subfigure{\includegraphics[width=\m_width\textwidth, height=\a_height]{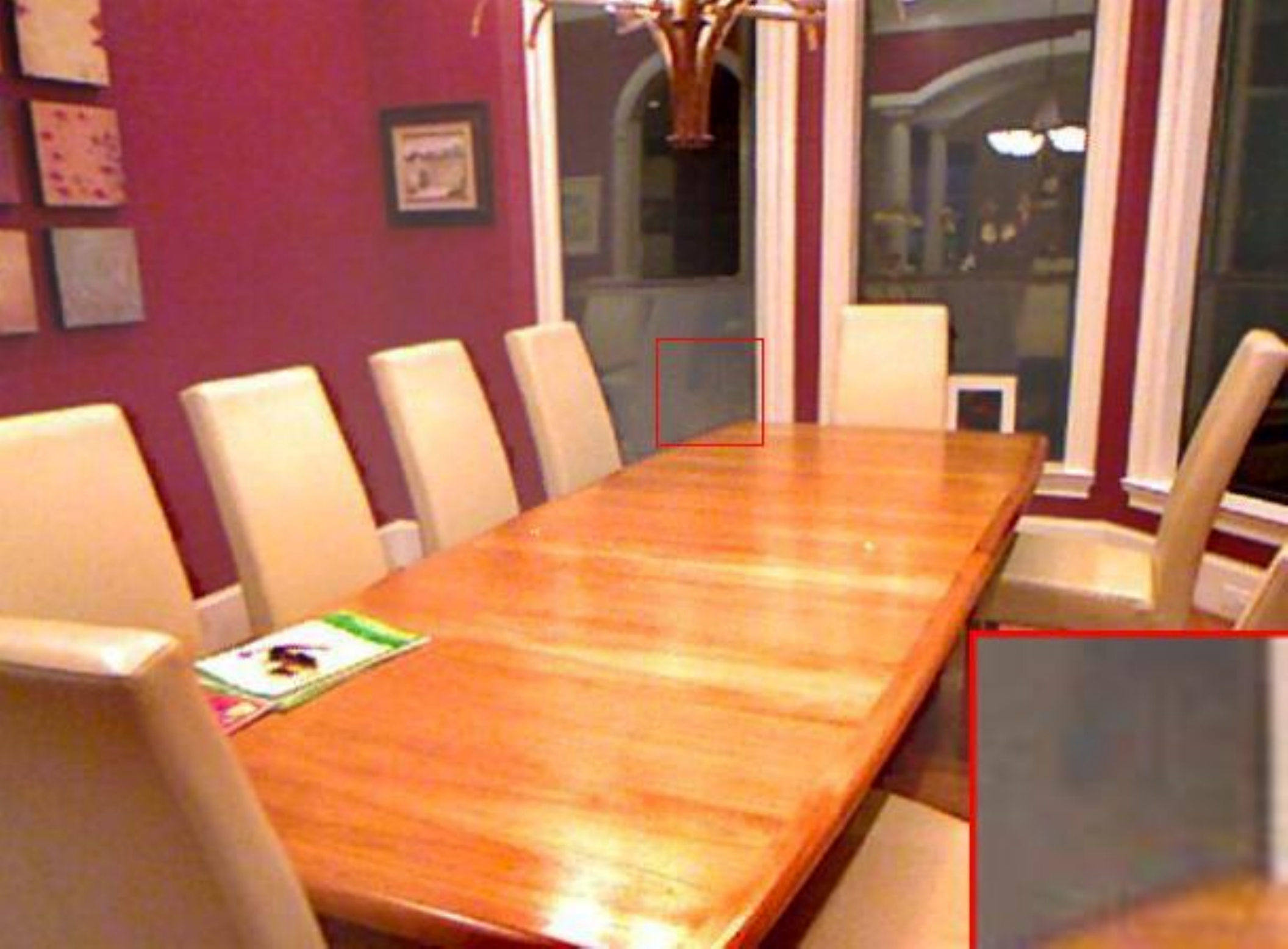}}
		\subfigure{\includegraphics[width=\m_width\textwidth, height=\a_height]{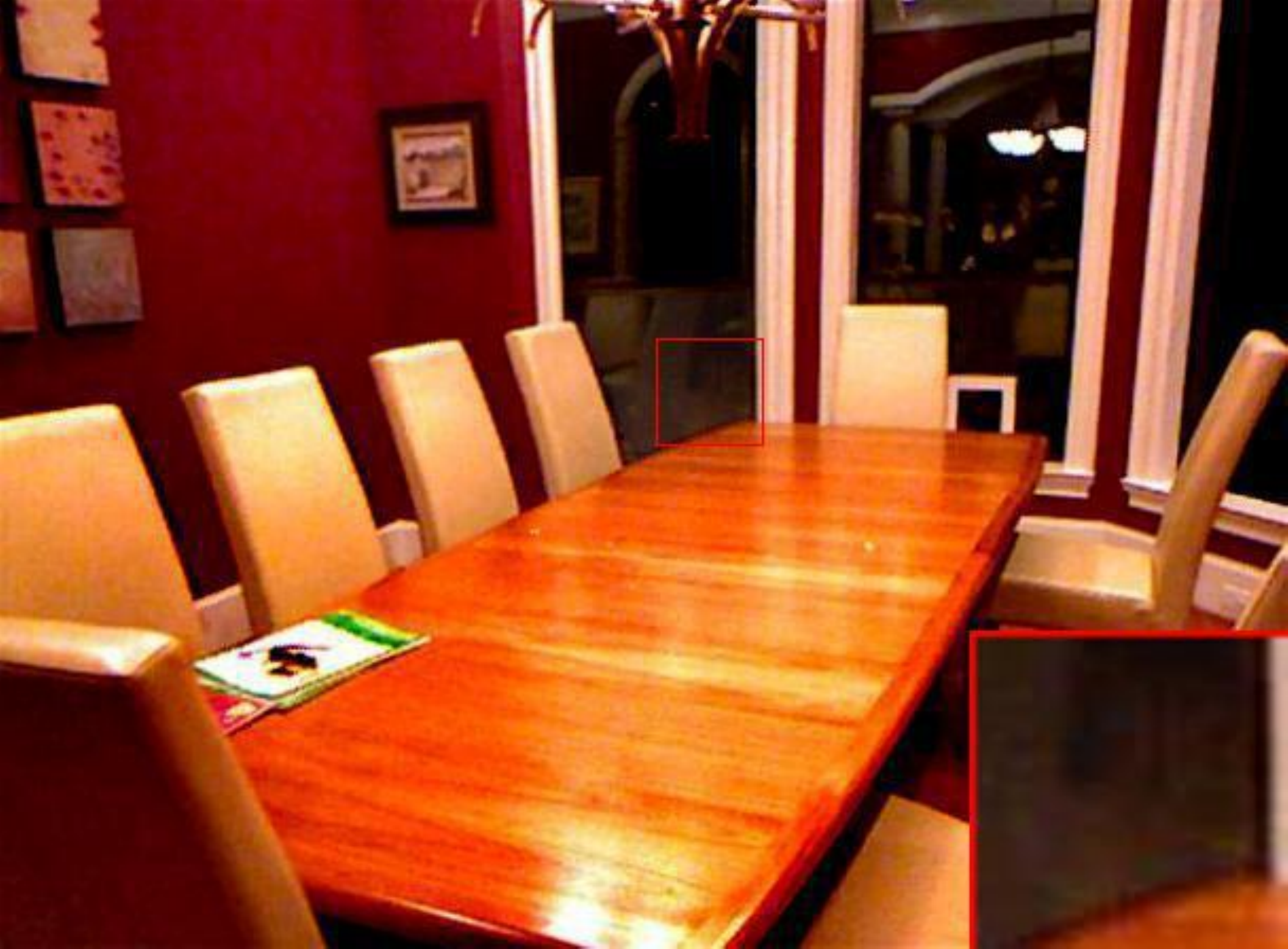}}
		\subfigure{\includegraphics[width=\m_width\textwidth,height=\a_height]{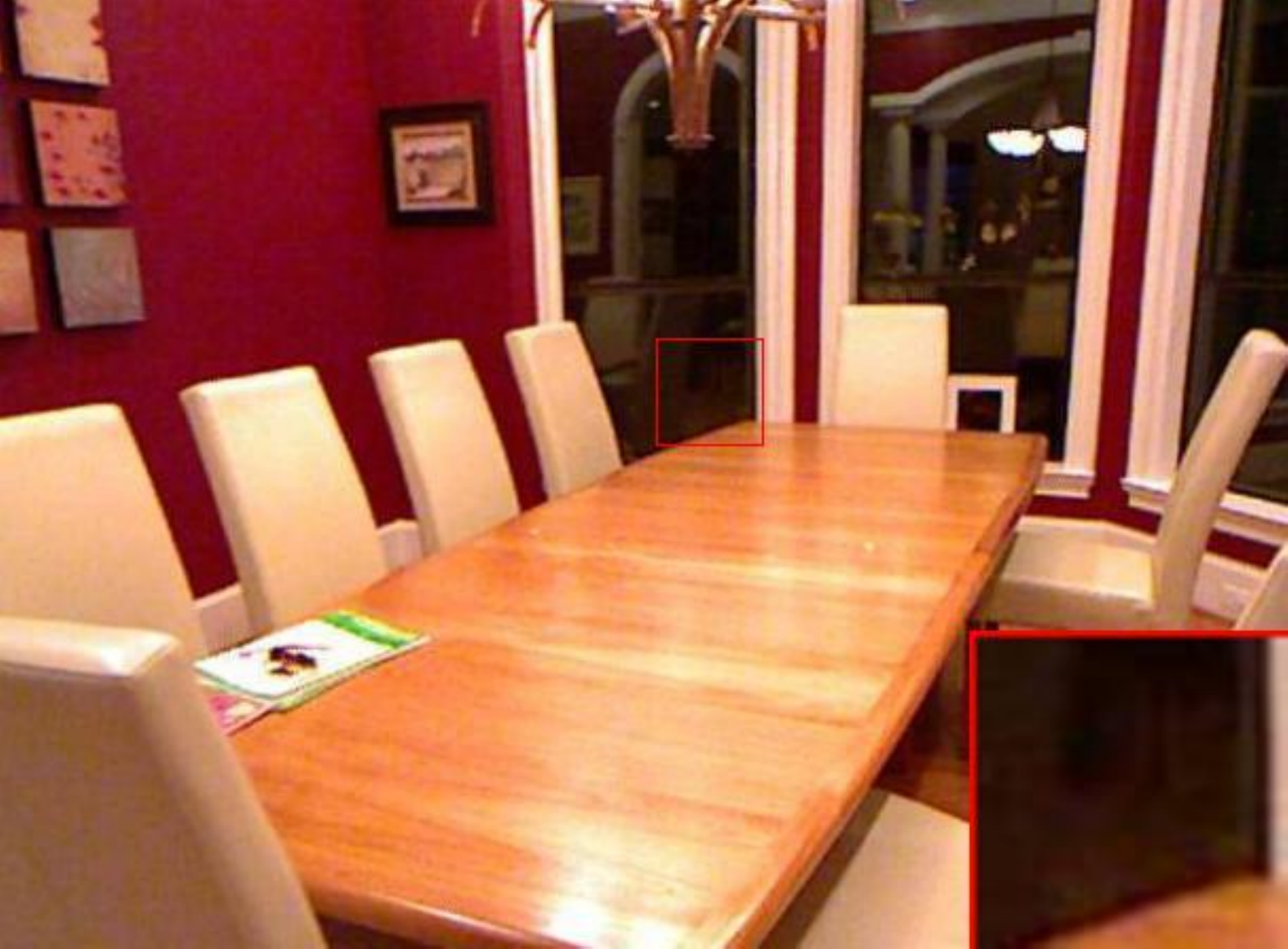}}
	\end{center}

	\vspace{-0.7cm}
		\begin{center}
		\subfigure{\includegraphics[width=\m_width\textwidth, height=\a_height]{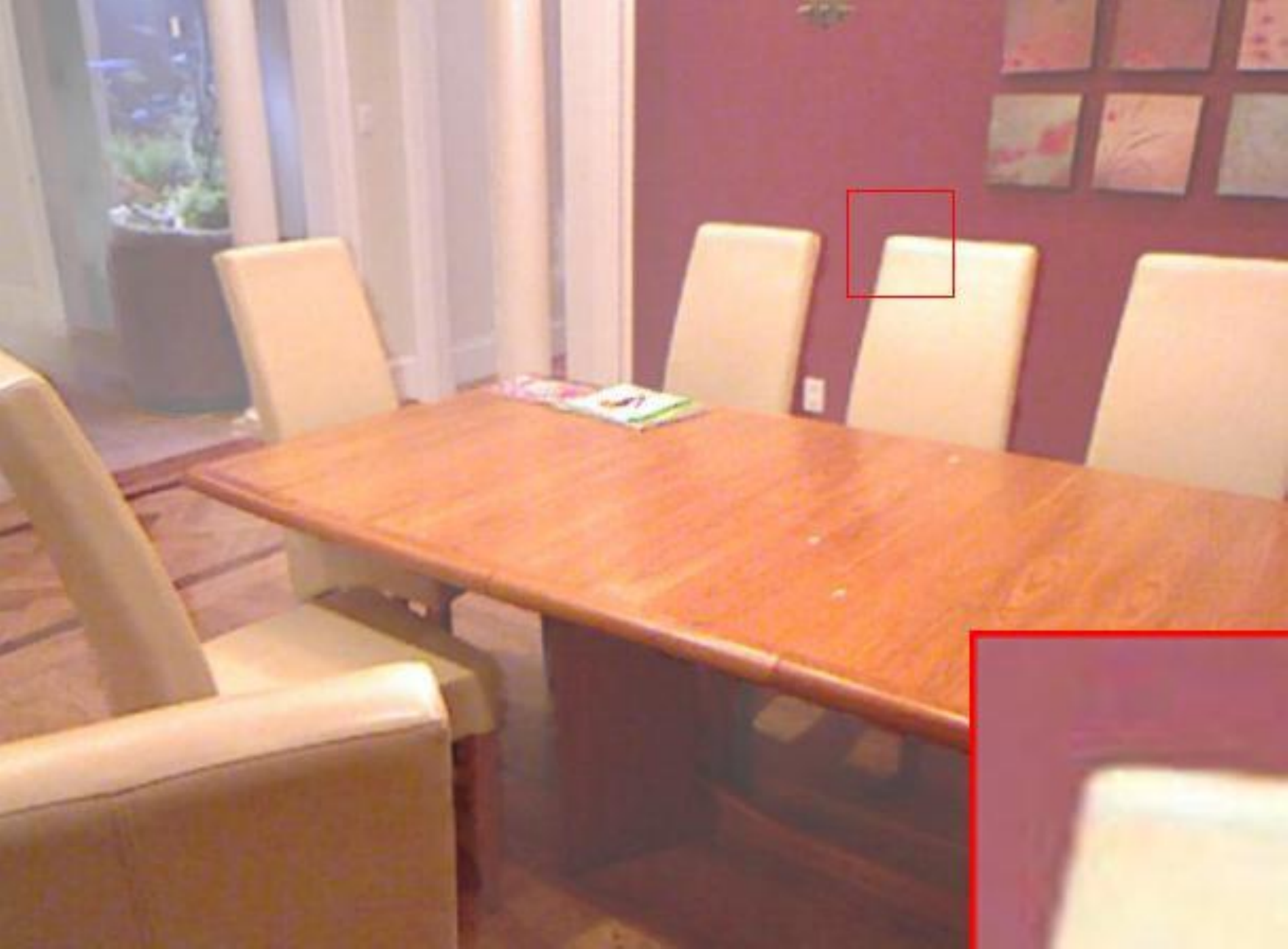}}
		\subfigure{\includegraphics[width=\m_width\textwidth, height=\a_height]{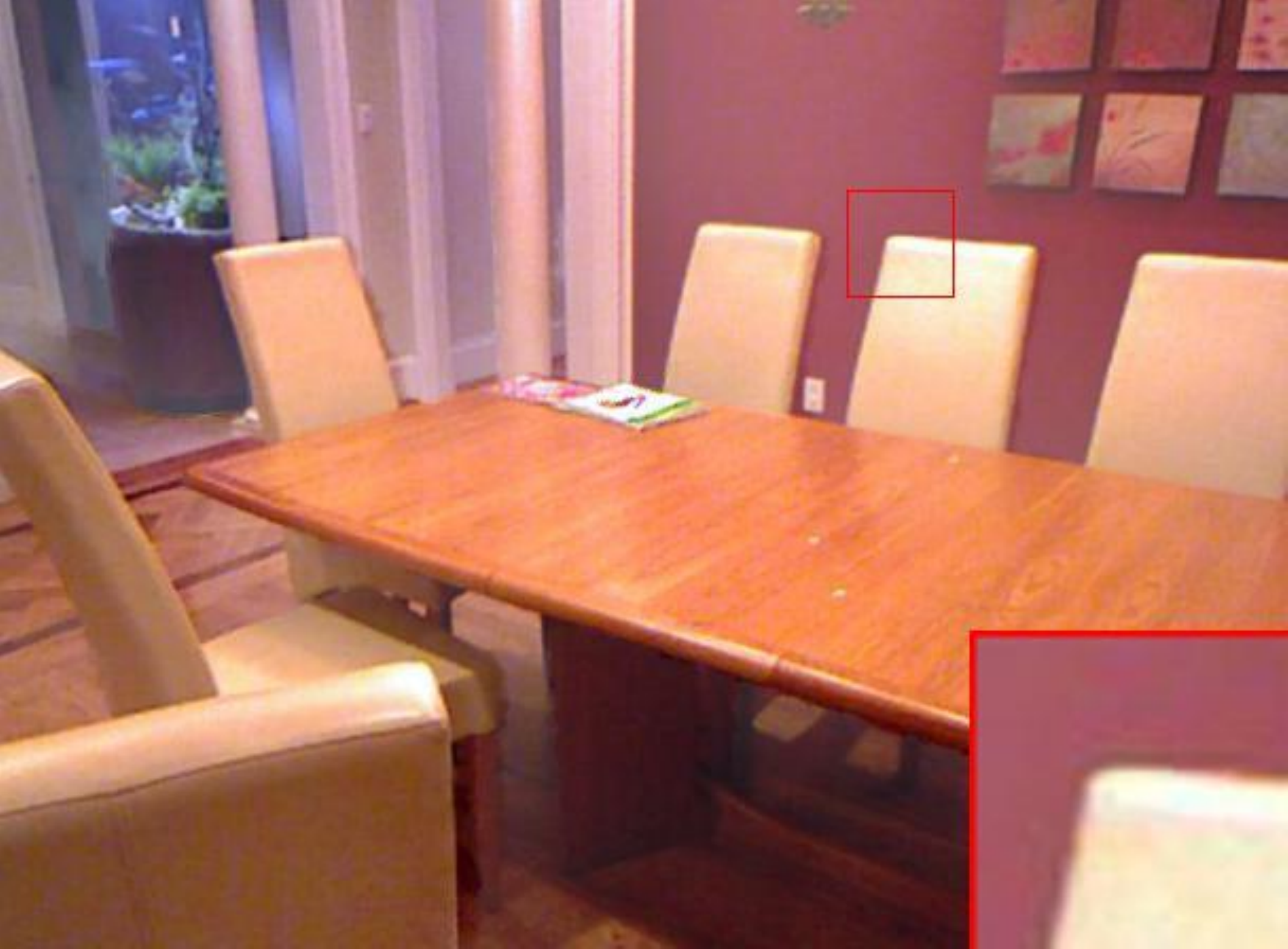}}
		\subfigure{\includegraphics[width=\m_width\textwidth, height=\a_height]{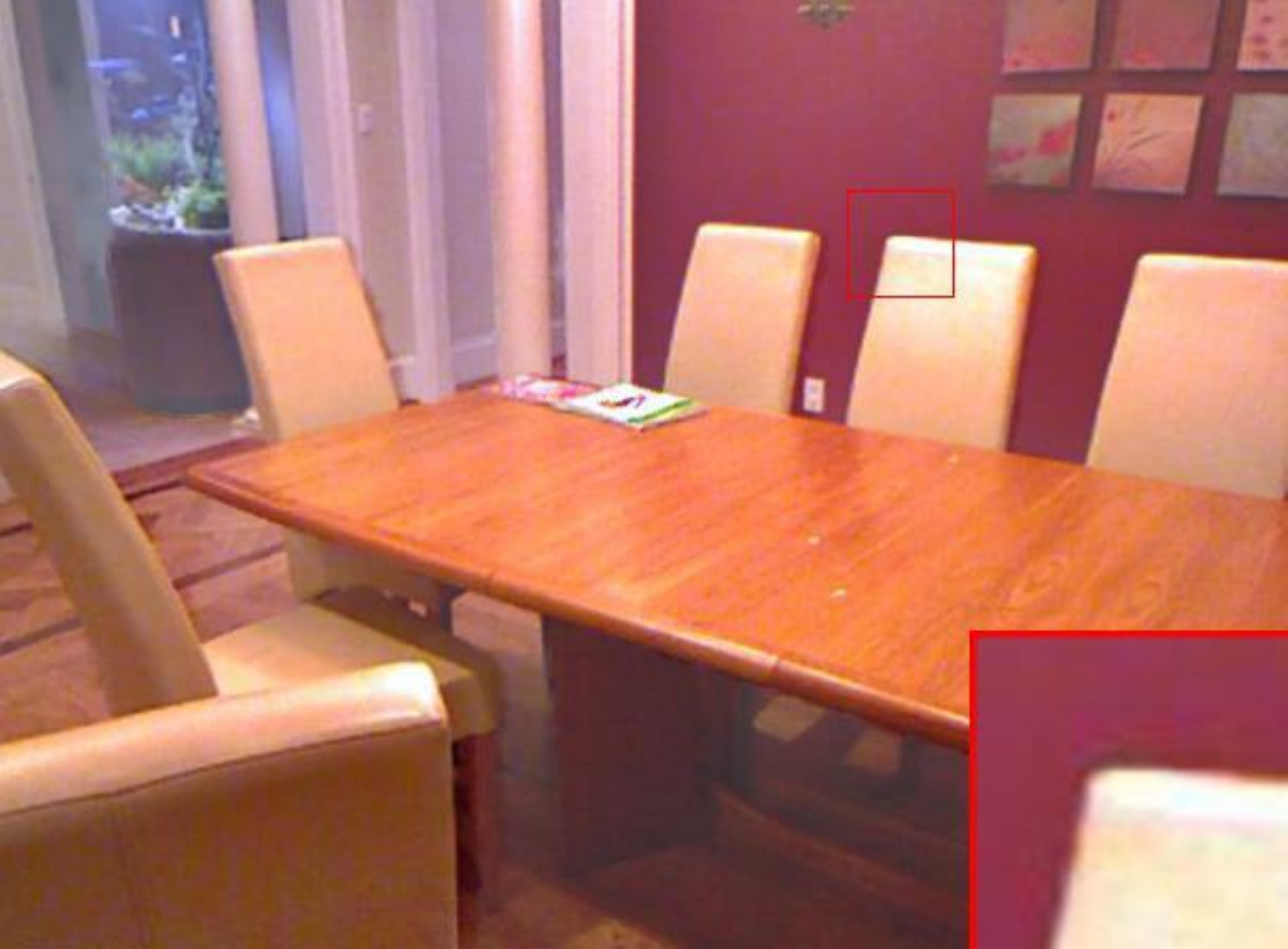}}
		\subfigure{\includegraphics[width=\m_width\textwidth, height=\a_height]{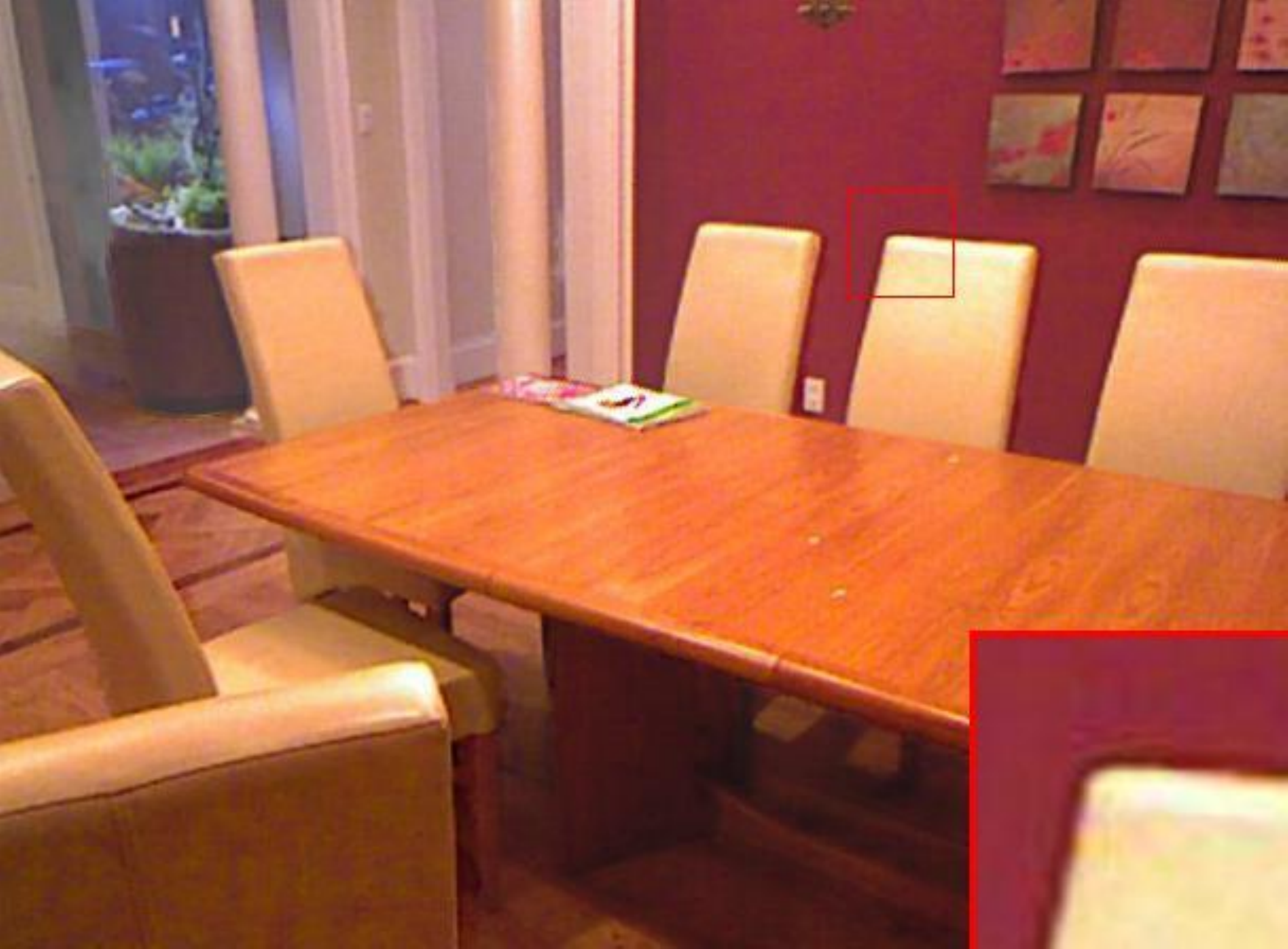}}
		\subfigure{\includegraphics[width=\m_width\textwidth, height=\a_height]{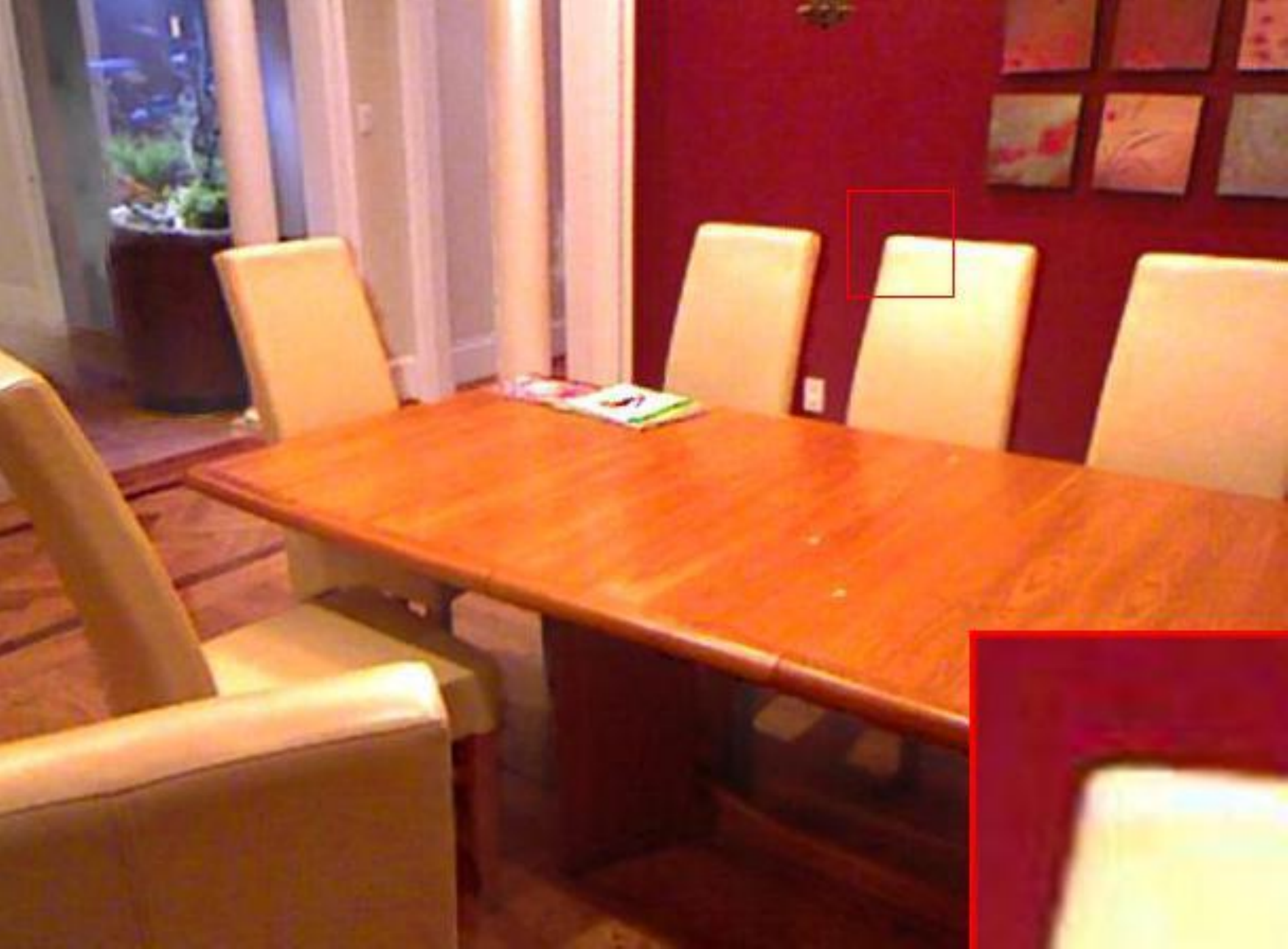}}
		\subfigure{\includegraphics[width=\m_width\textwidth, height=\a_height]{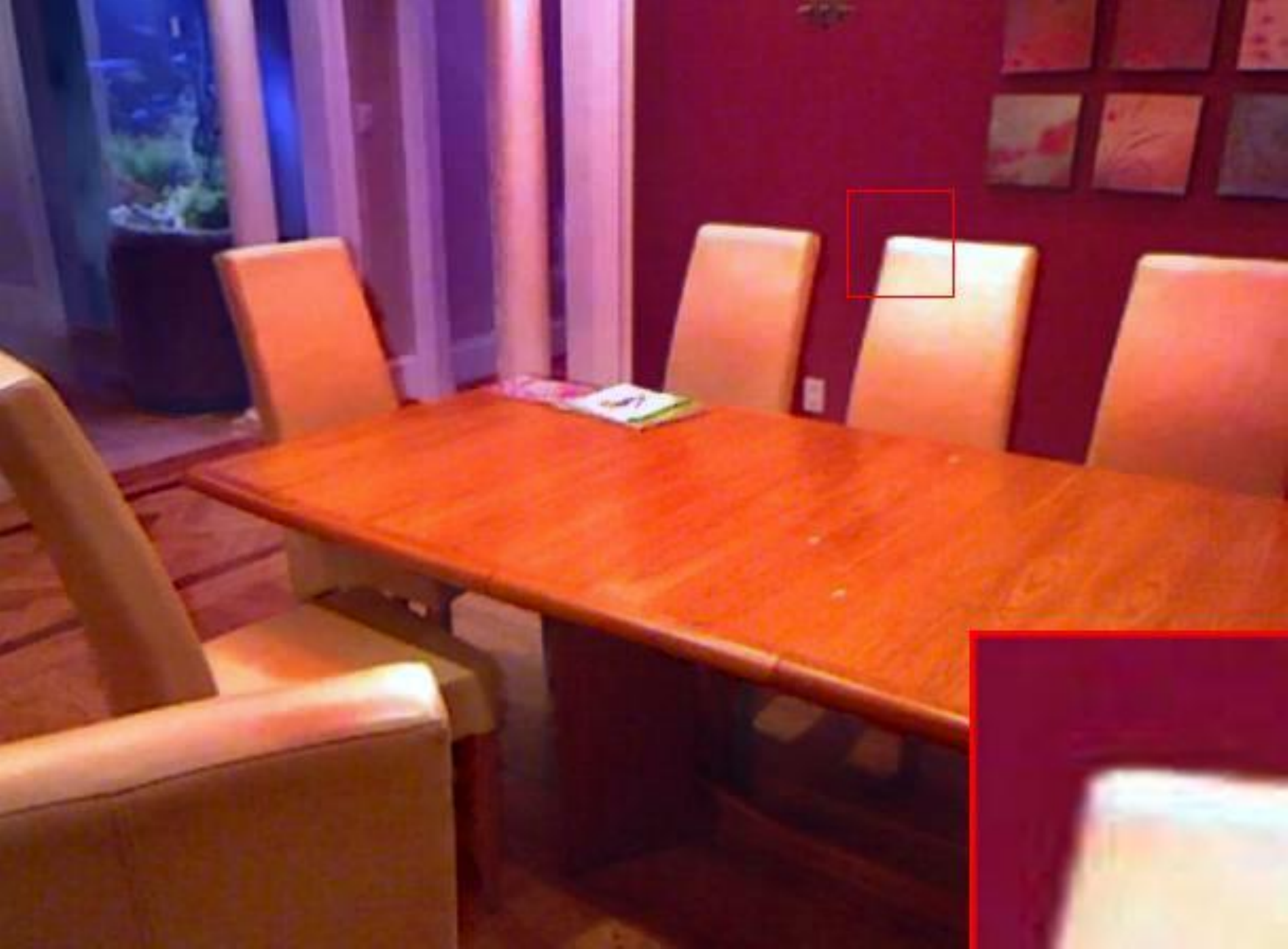}}
		\subfigure{\includegraphics[width=\m_width\textwidth, height=\a_height]{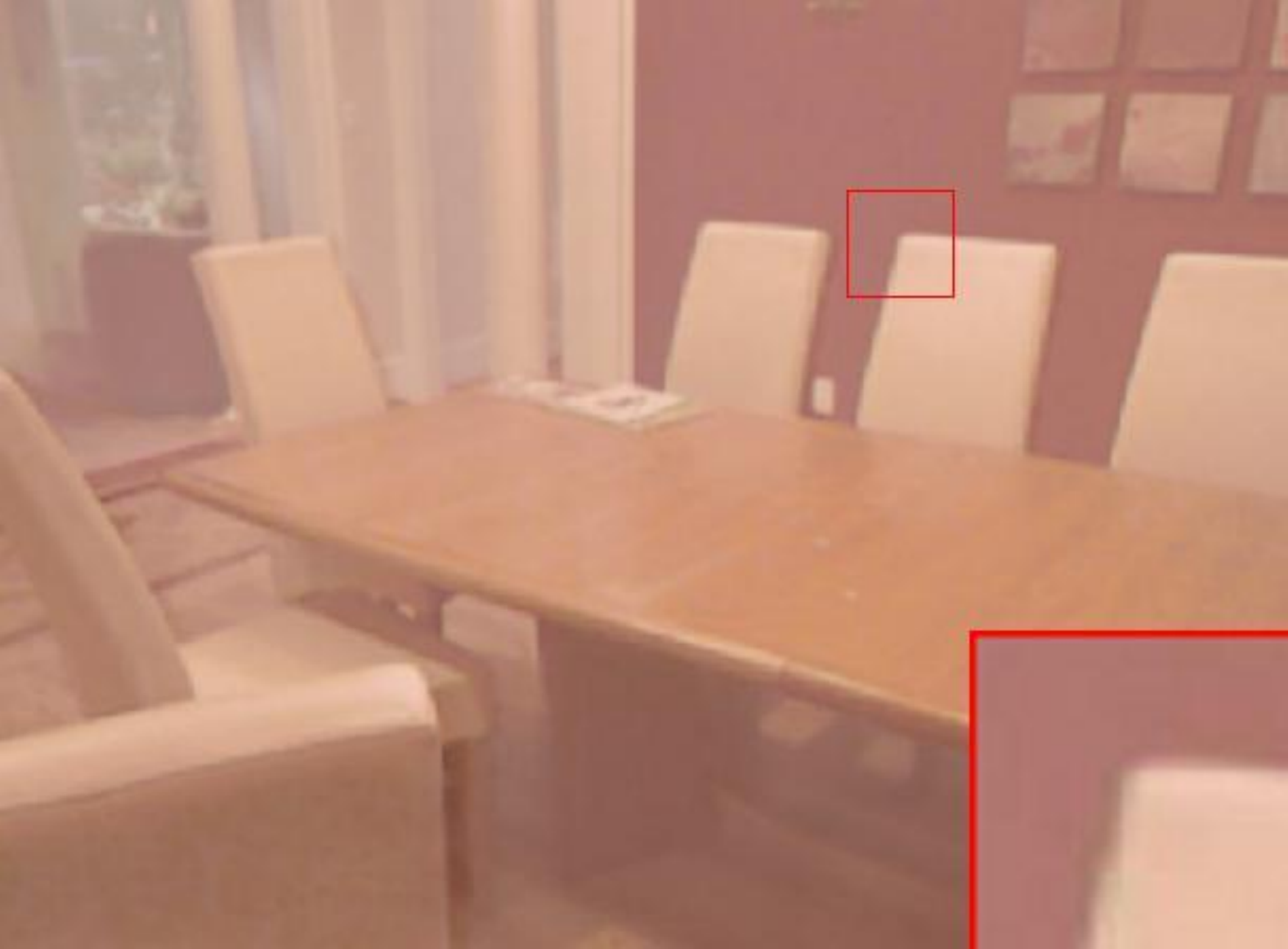}}
		\subfigure{\includegraphics[width=\m_width\textwidth, height=\a_height]{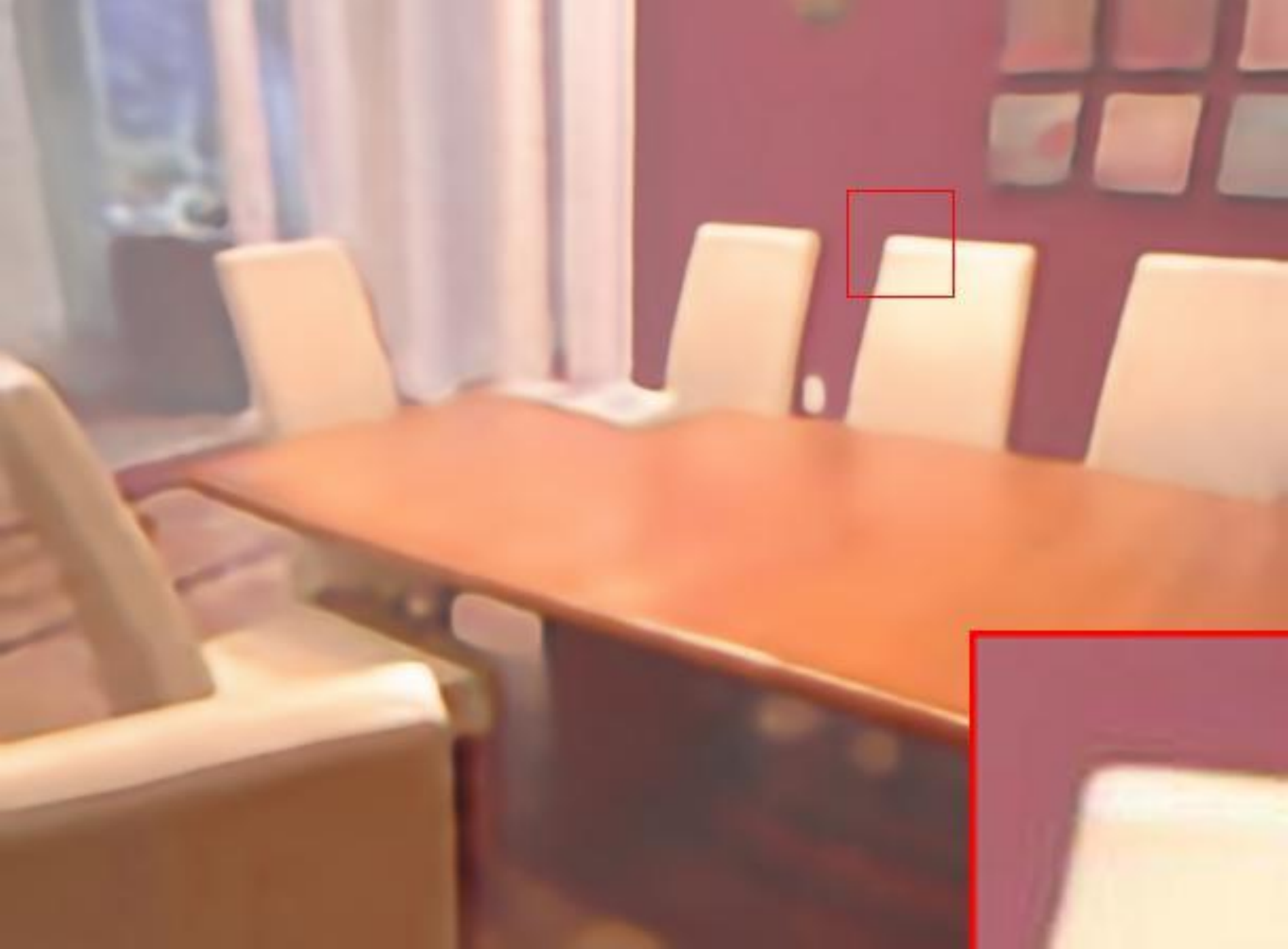}}
		\subfigure{\includegraphics[width=\m_width\textwidth, height=\a_height]{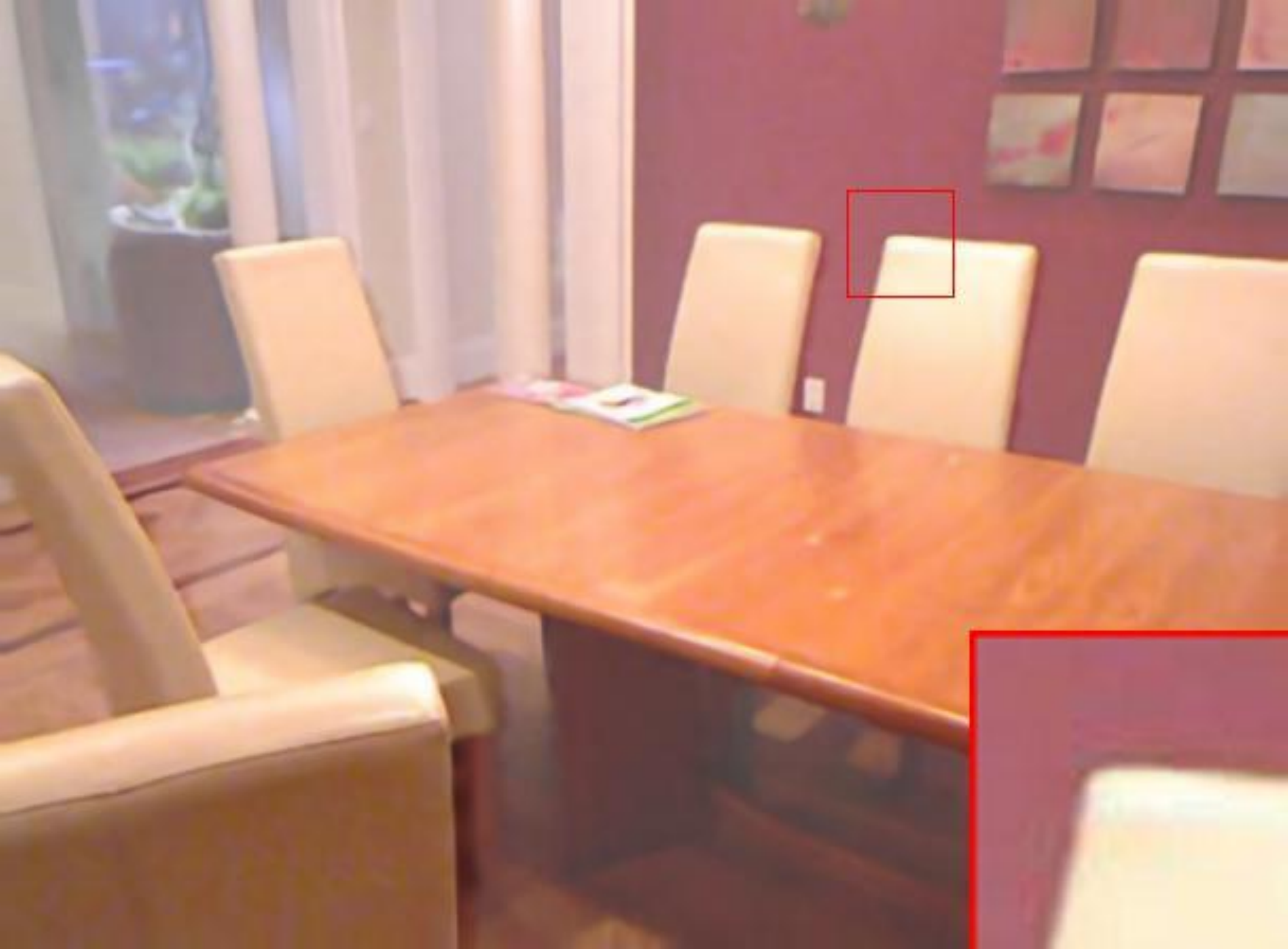}}
		\subfigure{\includegraphics[width=\m_width\textwidth, height=\a_height]{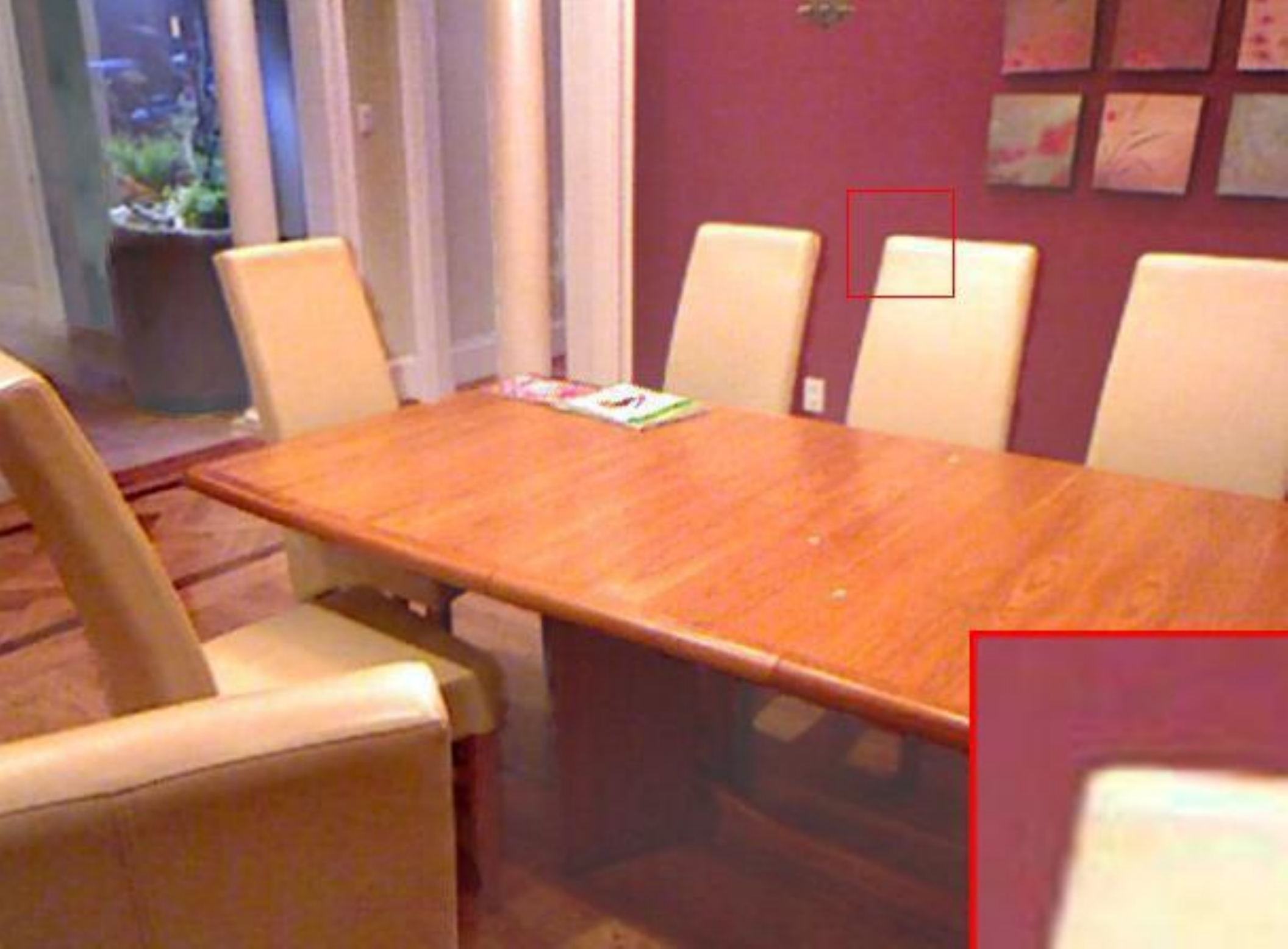}}
		\subfigure{\includegraphics[width=\m_width\textwidth, height=\a_height]{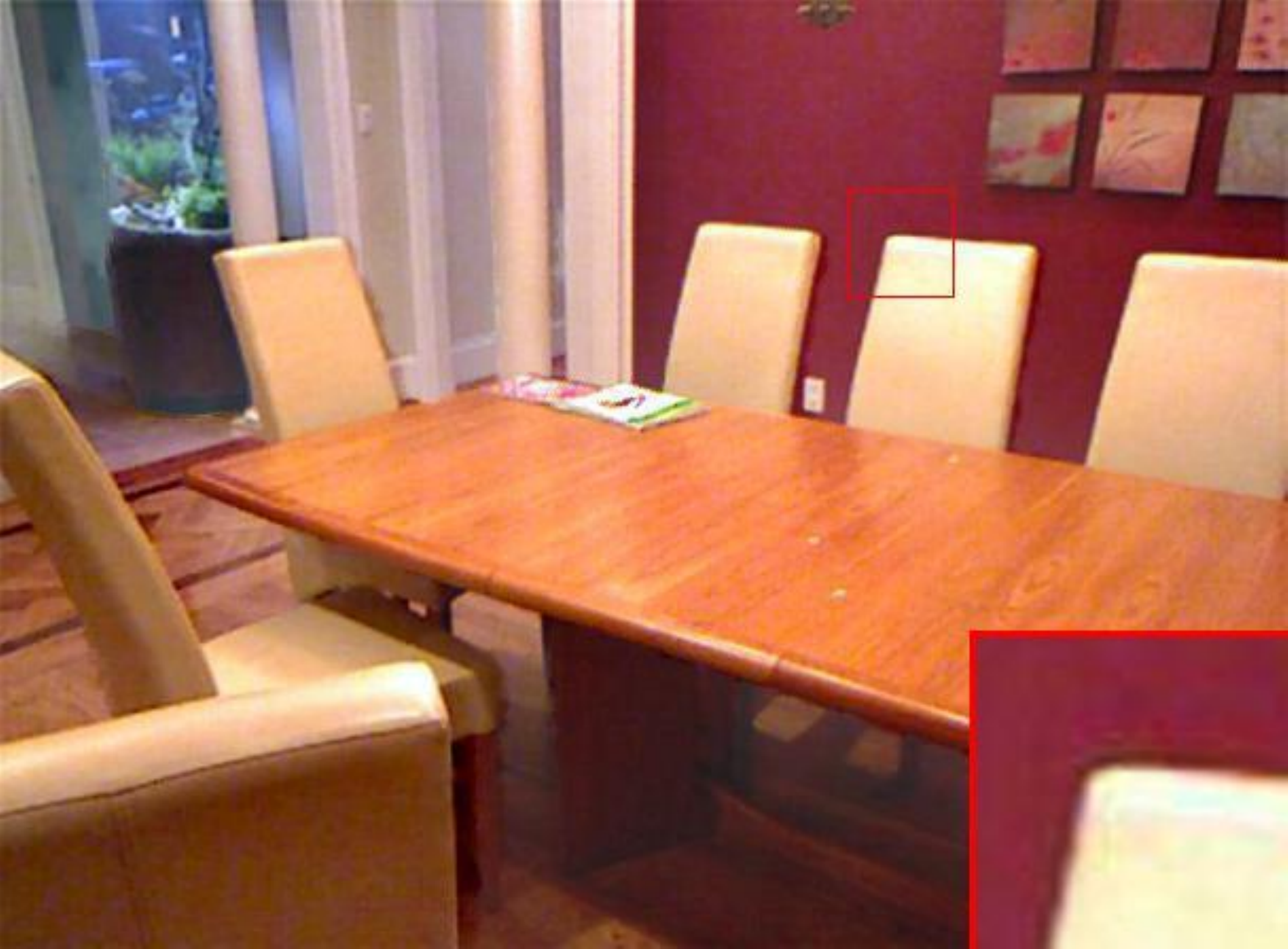}}
		\subfigure{\includegraphics[width=\m_width\textwidth,height=\a_height]{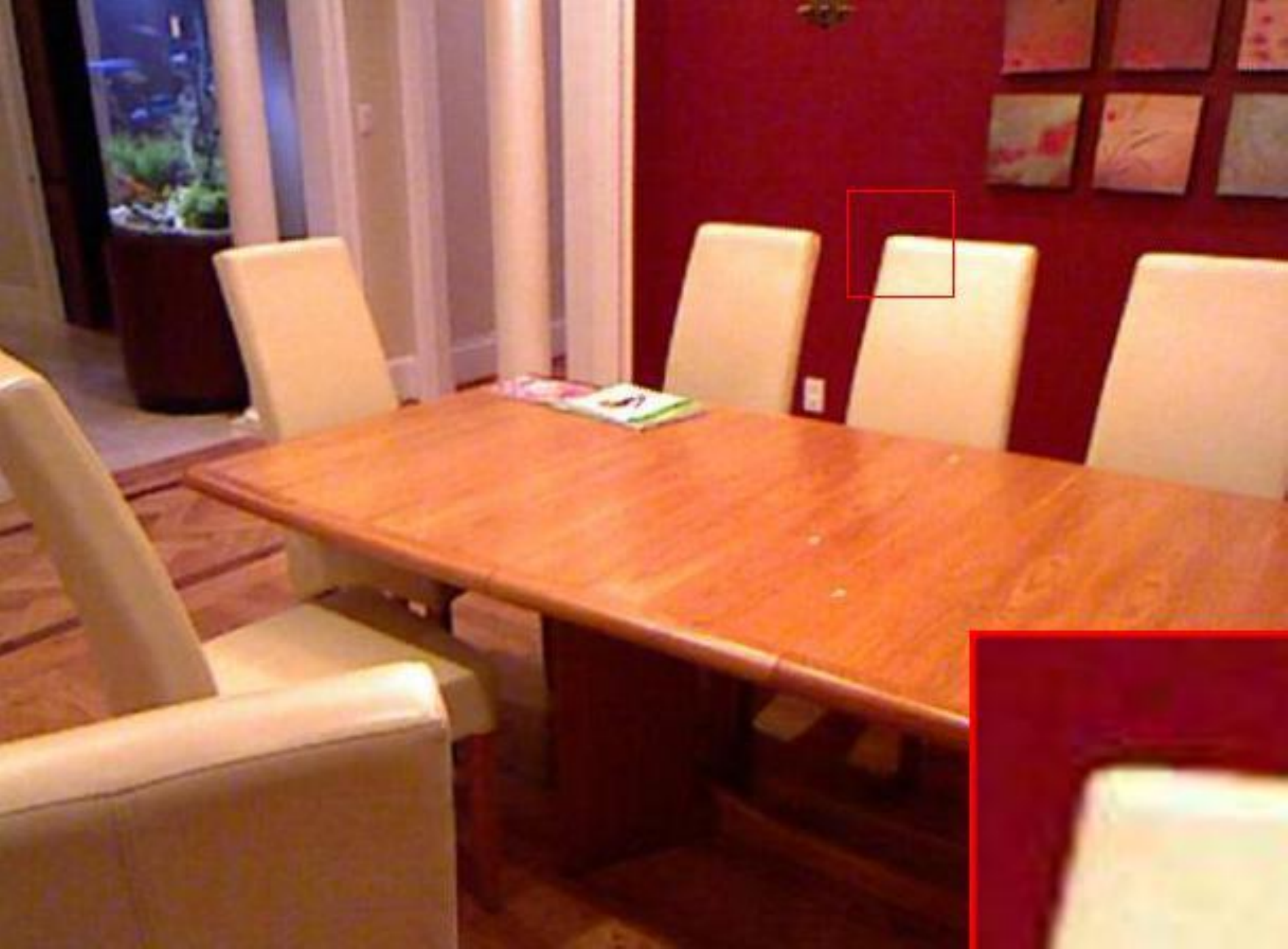}}
	\end{center}
	
	\vspace{-0.7cm}
		\begin{center}
		\subfigure{\includegraphics[width=\m_width\textwidth, height=\a_height]{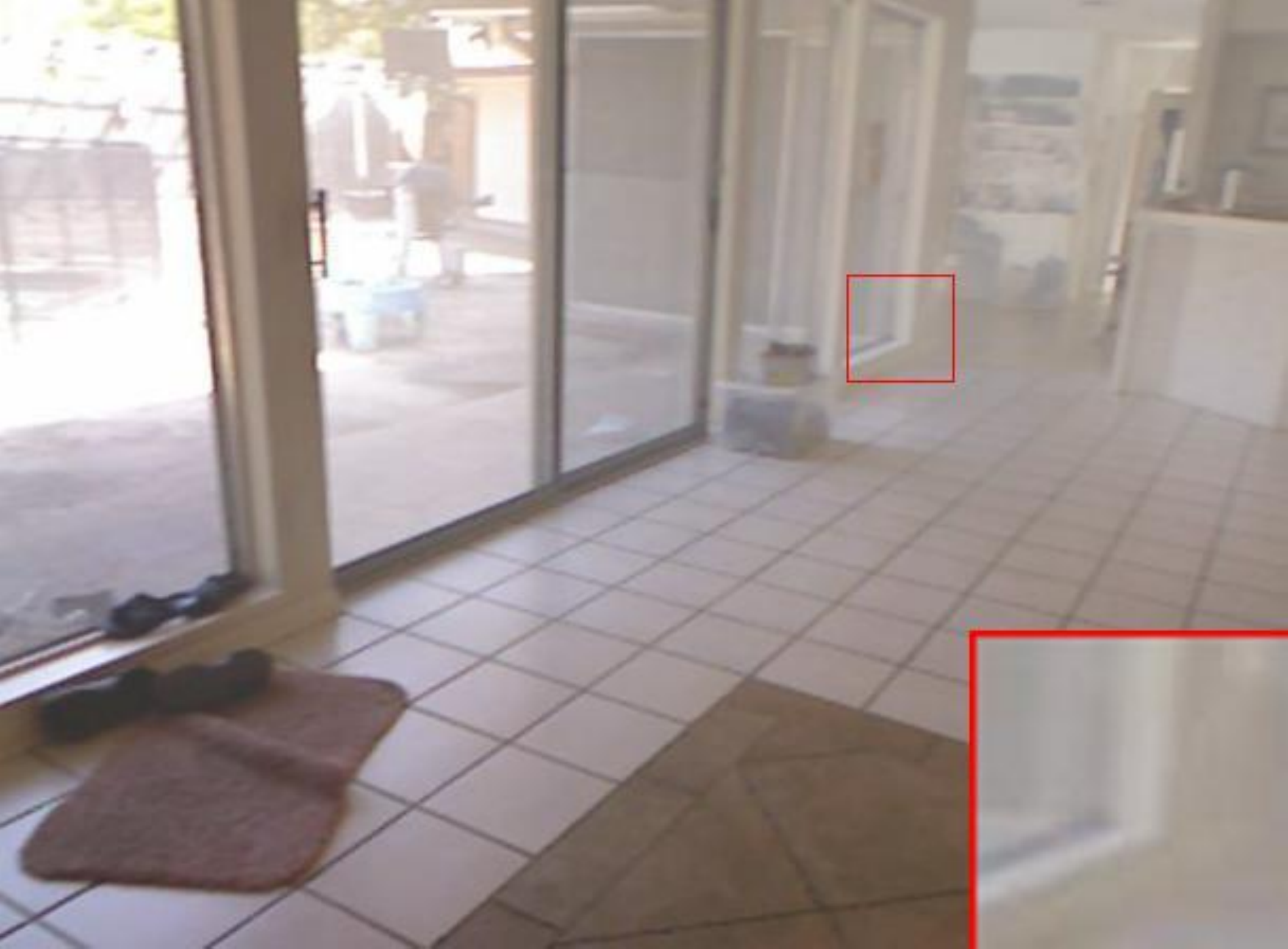}}
		\subfigure{\includegraphics[width=\m_width\textwidth, height=\a_height]{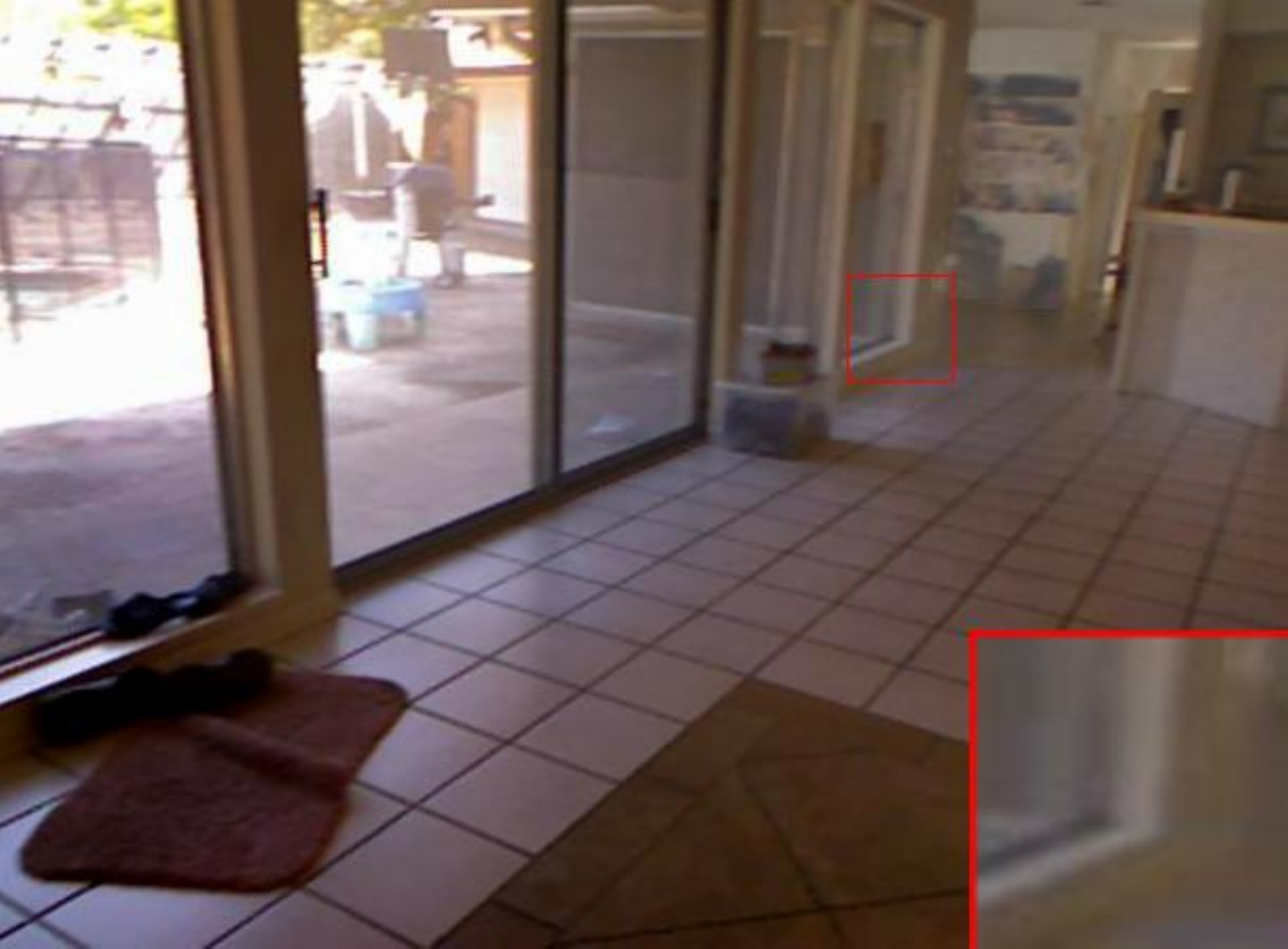}}
		\subfigure{\includegraphics[width=\m_width\textwidth, height=\a_height]{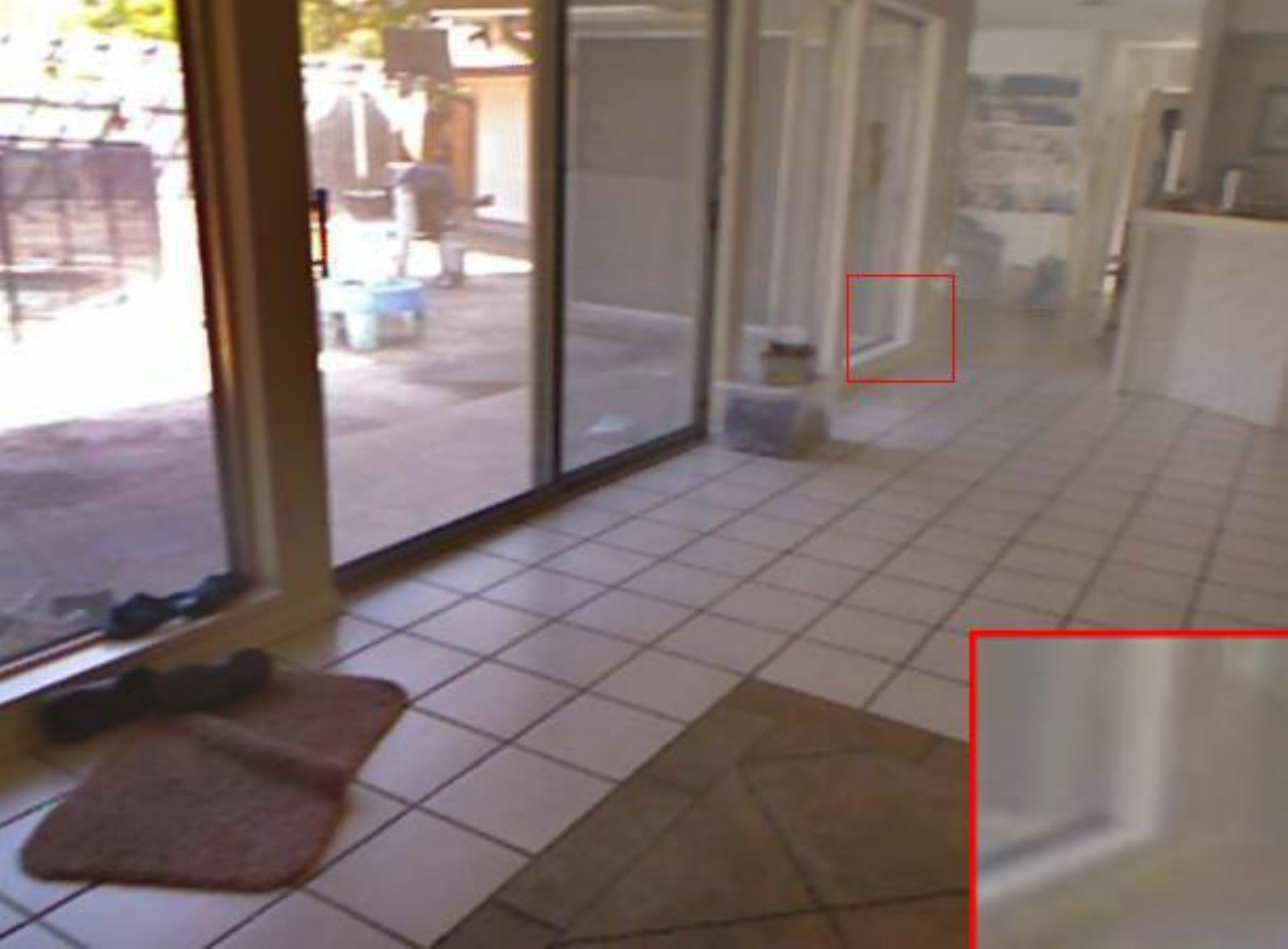}}
		\subfigure{\includegraphics[width=\m_width\textwidth, height=\a_height]{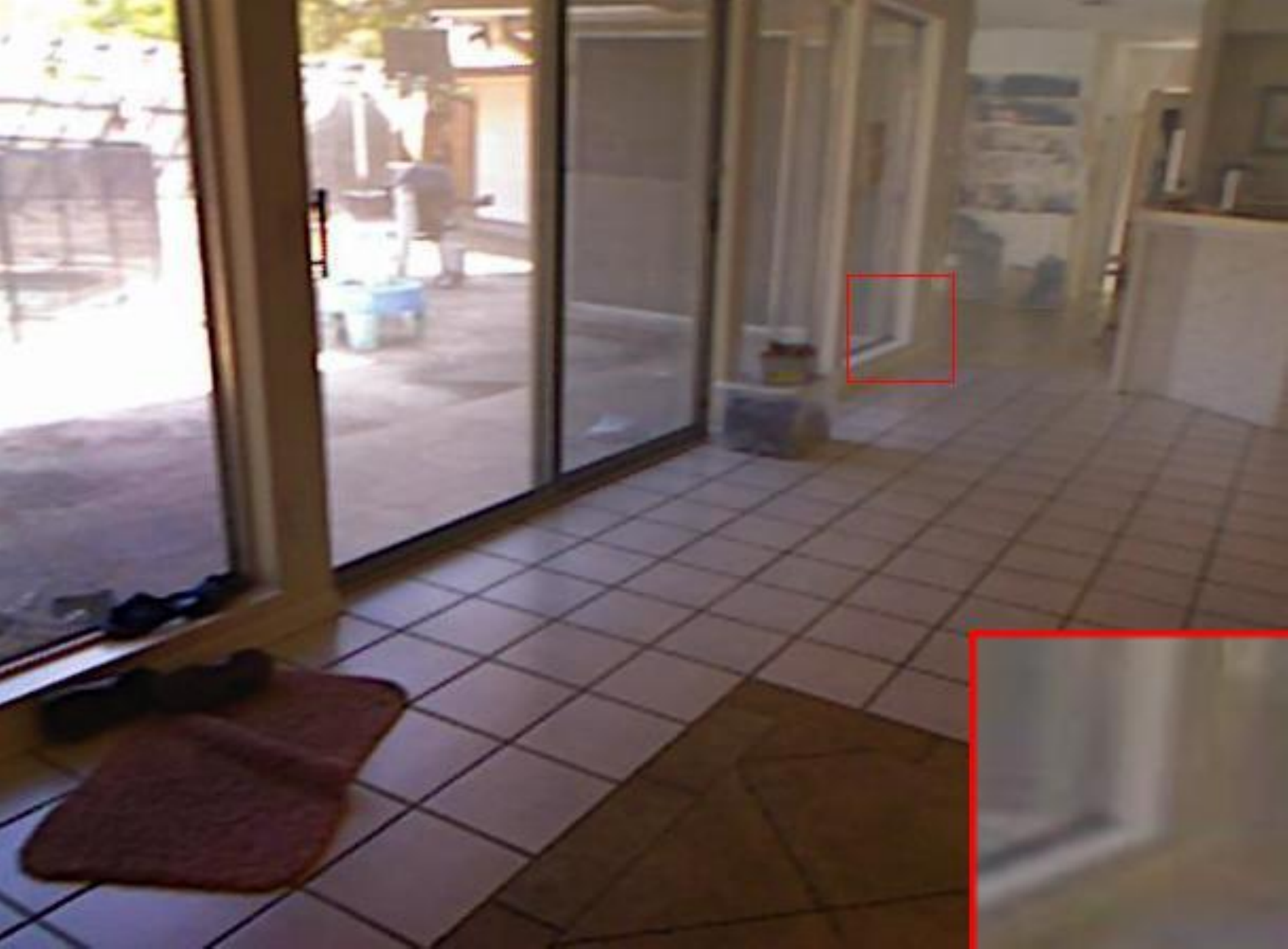}}
		\subfigure{\includegraphics[width=\m_width\textwidth, height=\a_height]{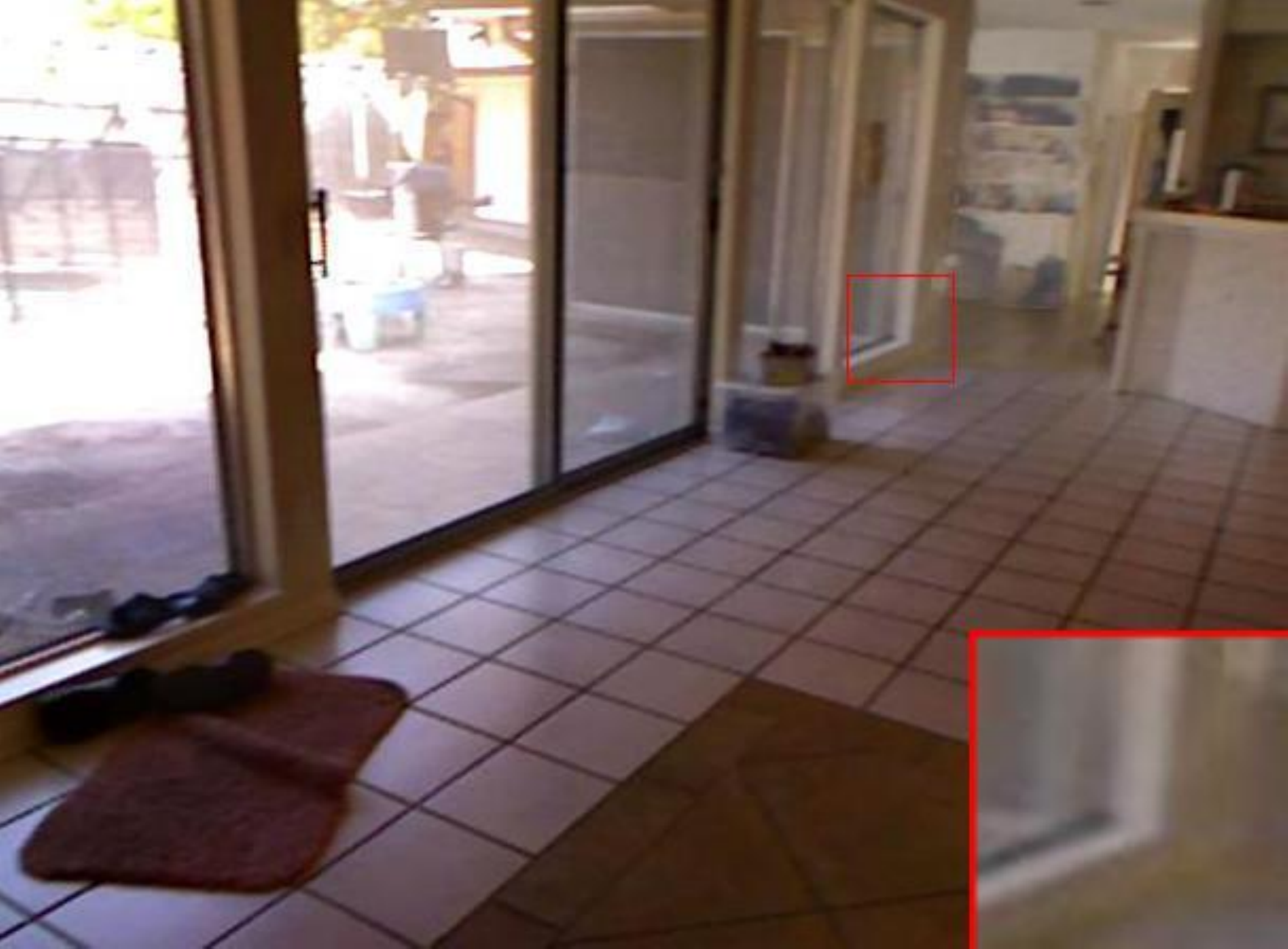}}
		\subfigure{\includegraphics[width=\m_width\textwidth, height=\a_height]{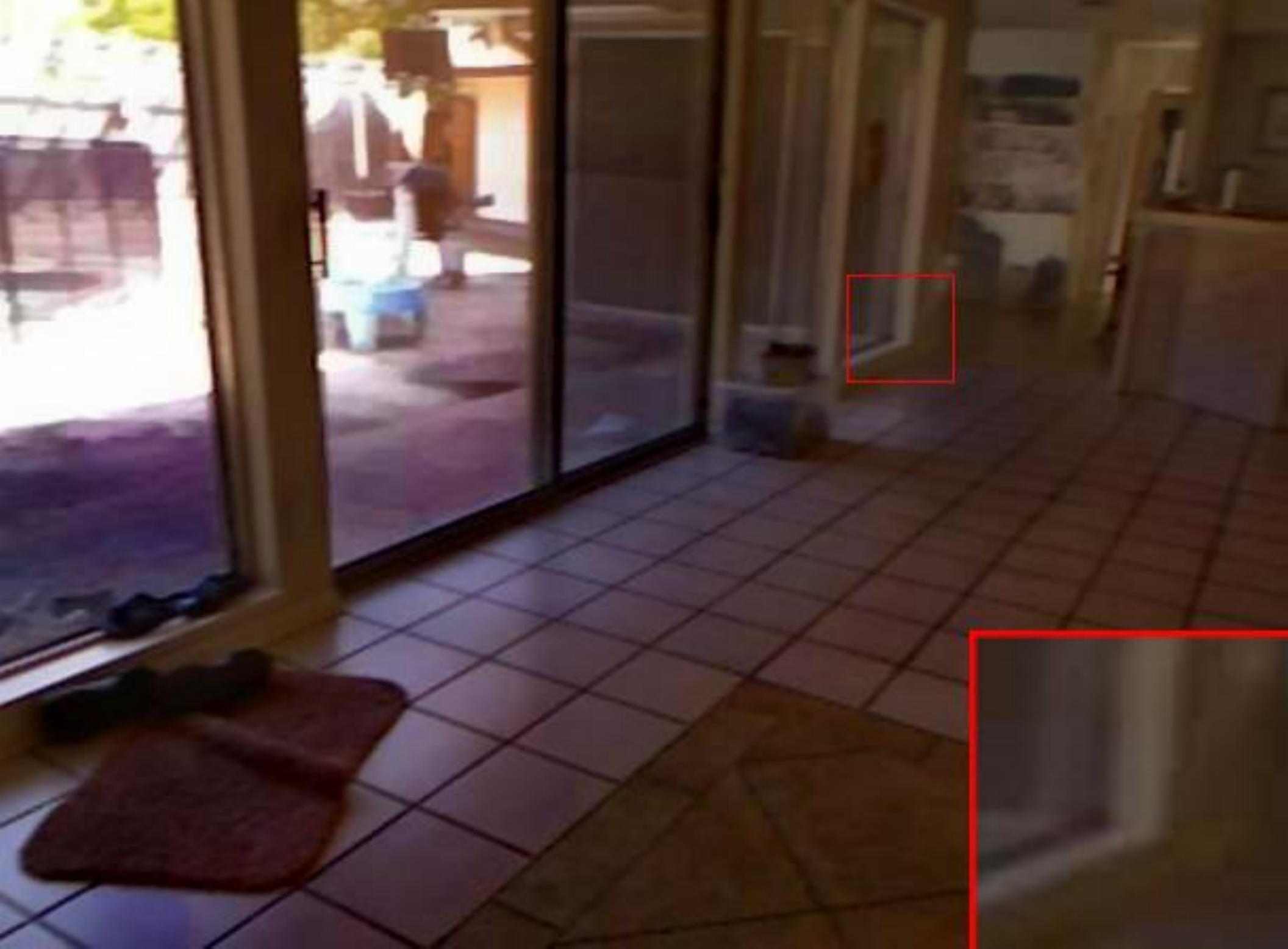}}
		\subfigure{\includegraphics[width=\m_width\textwidth, height=\a_height]{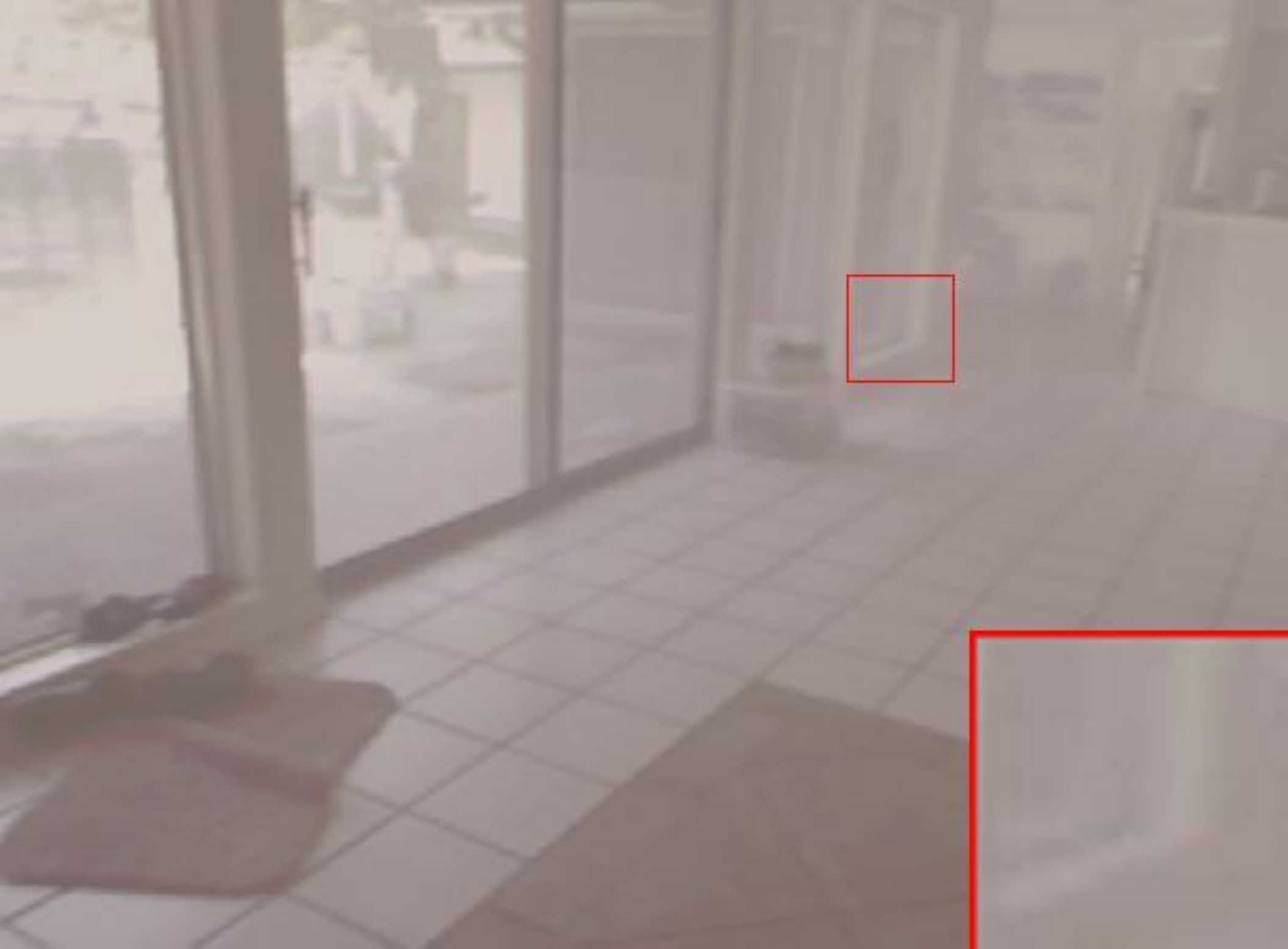}}
		\subfigure{\includegraphics[width=\m_width\textwidth, height=\a_height]{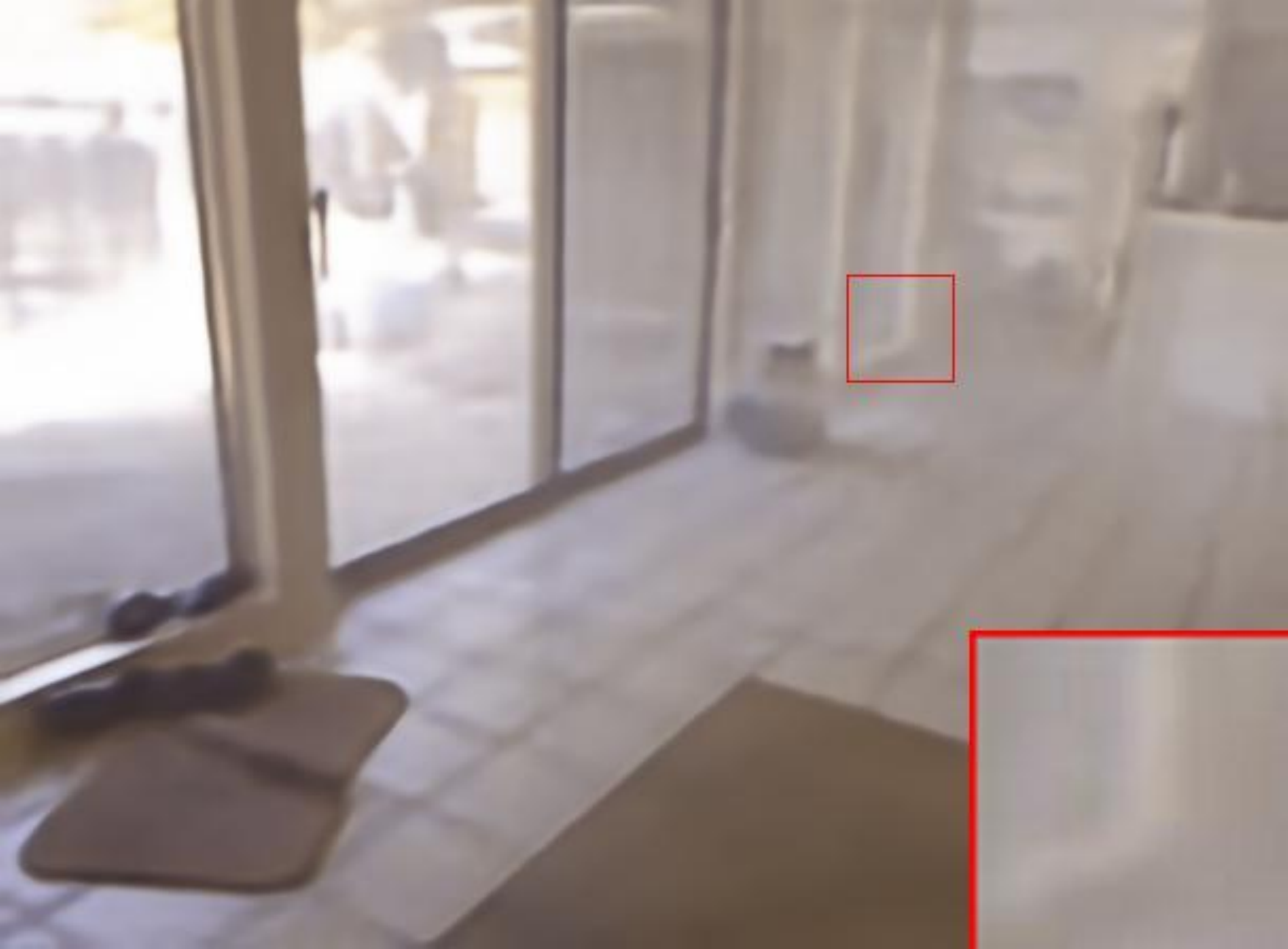}}
		\subfigure{\includegraphics[width=\m_width\textwidth, height=\a_height]{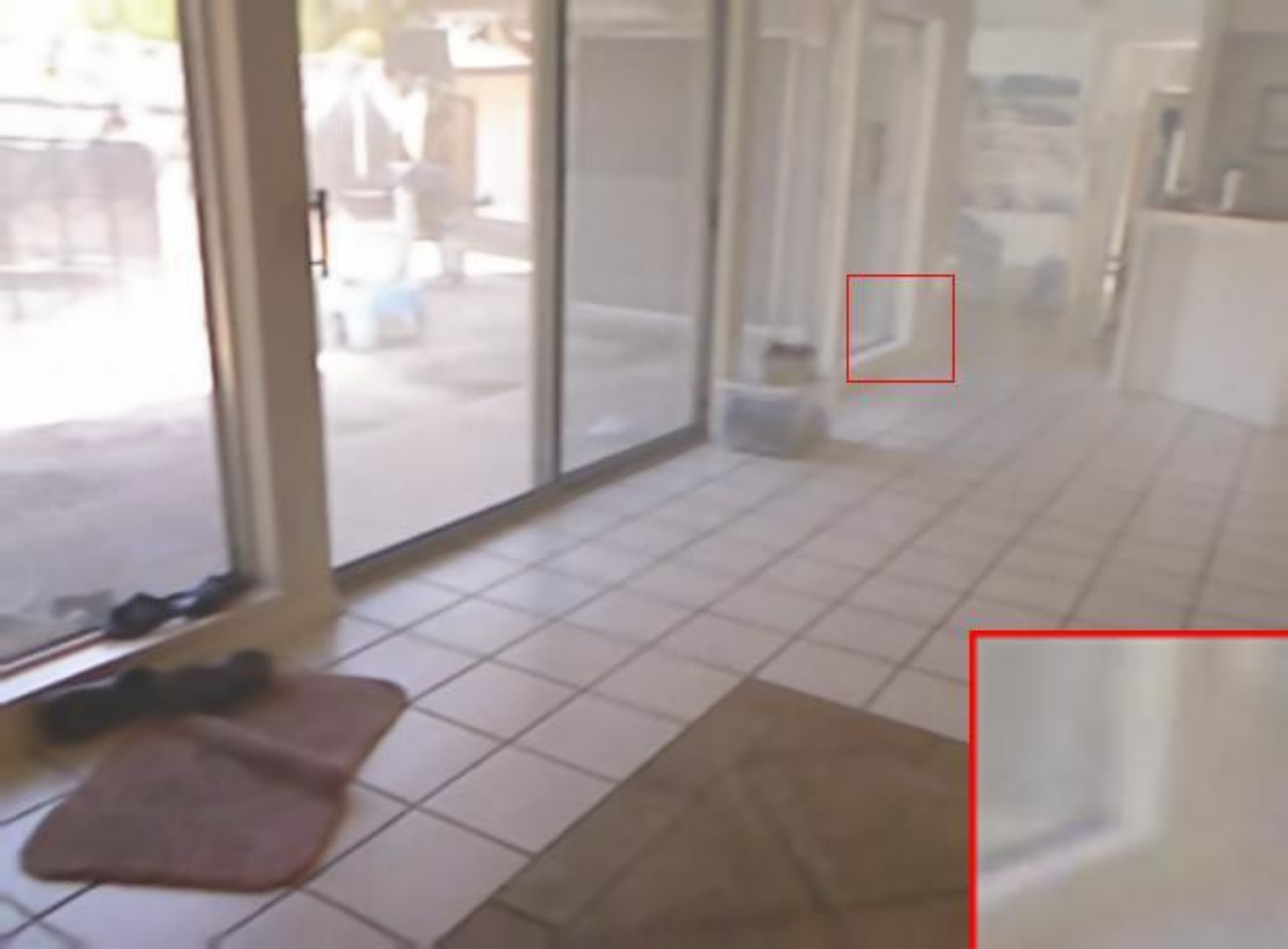}}
		\subfigure{\includegraphics[width=\m_width\textwidth, height=\a_height]{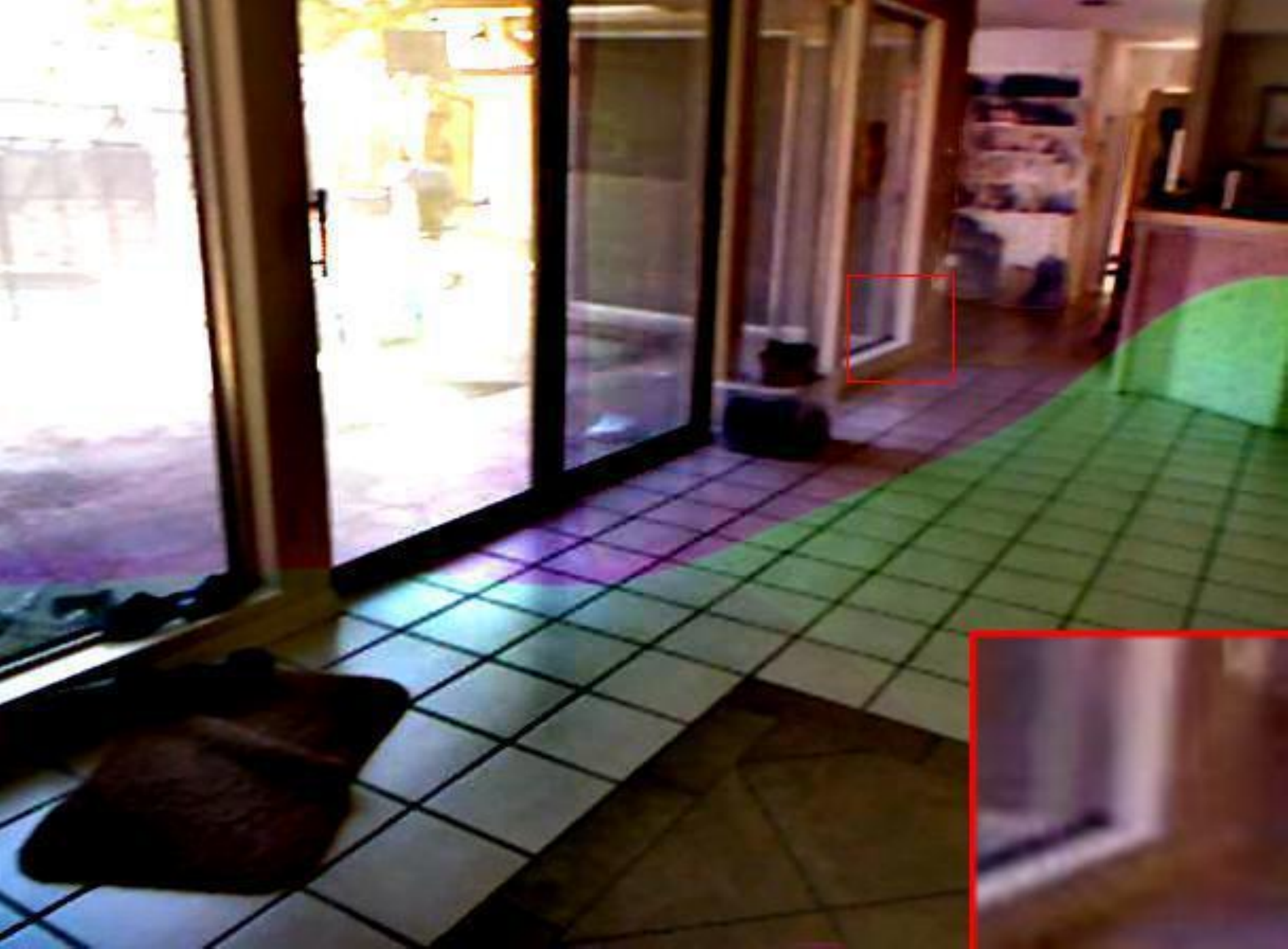}}
		\subfigure{\includegraphics[width=\m_width\textwidth, height=\a_height]{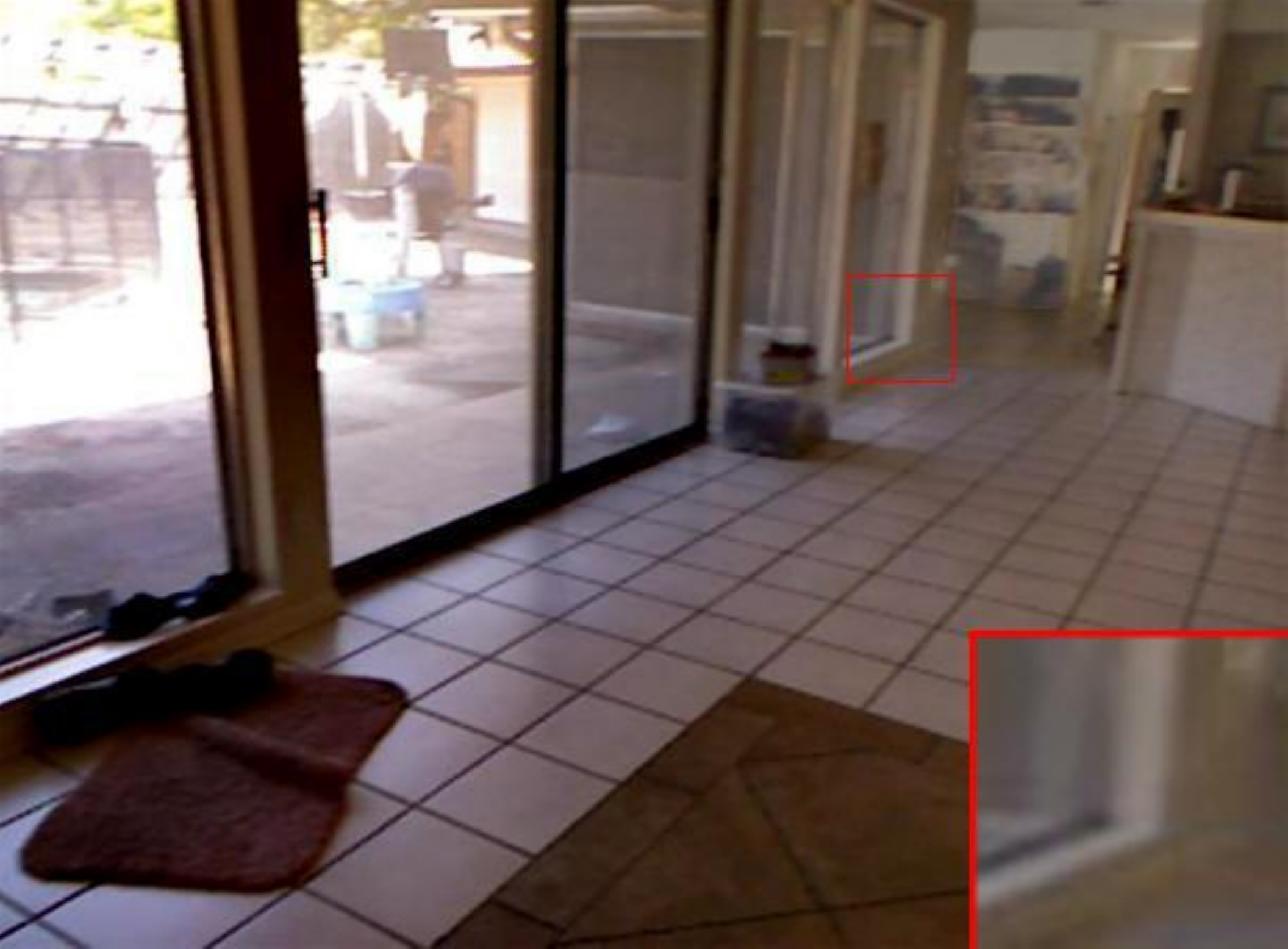}}
	\subfigure{\includegraphics[width=\m_width\textwidth,height=\a_height]{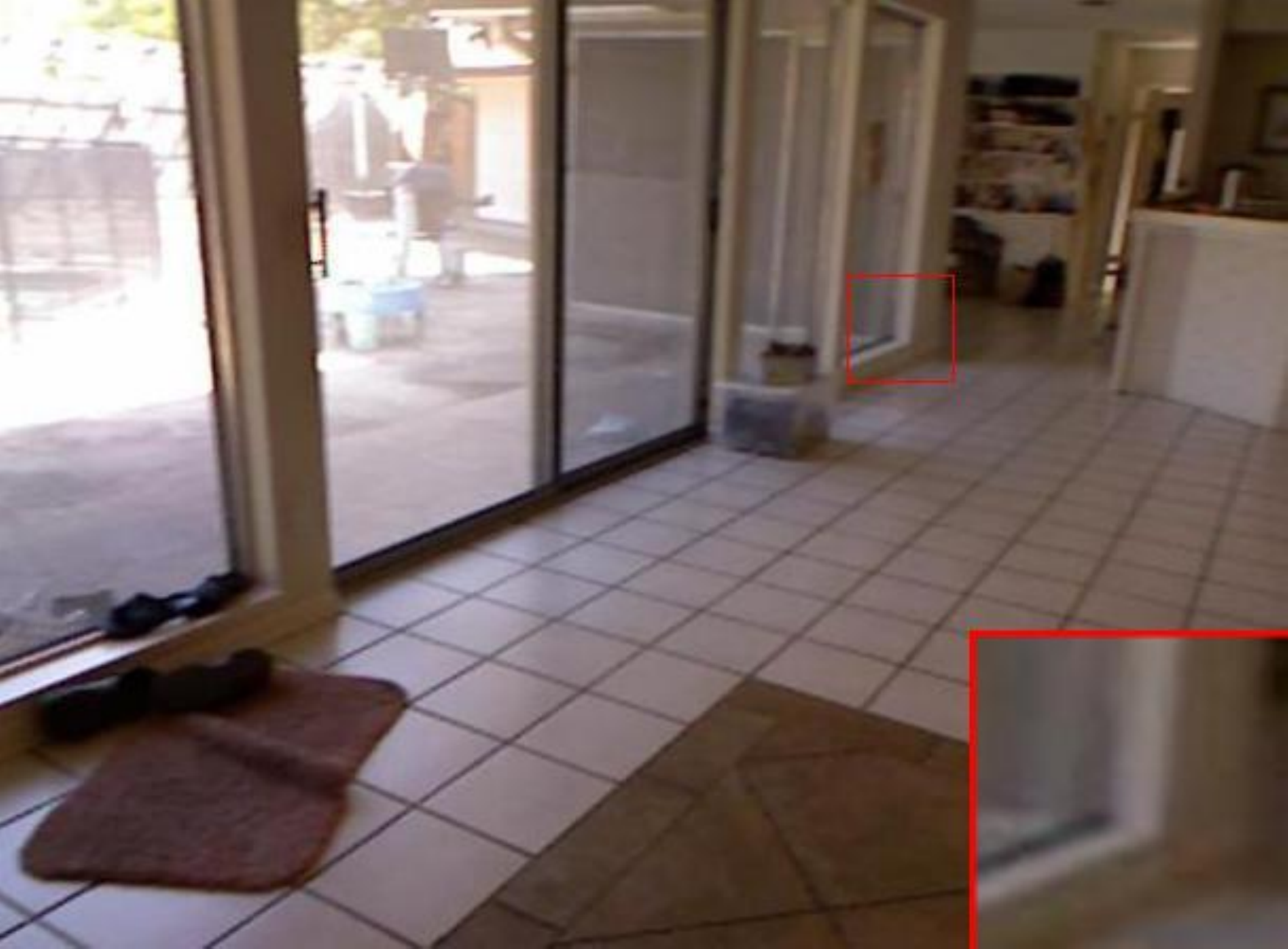}}
	\end{center}
	
	\vspace{-0.7cm}
		\begin{center}
		\subfigure{\includegraphics[width=\m_width\textwidth, height=\a_height]{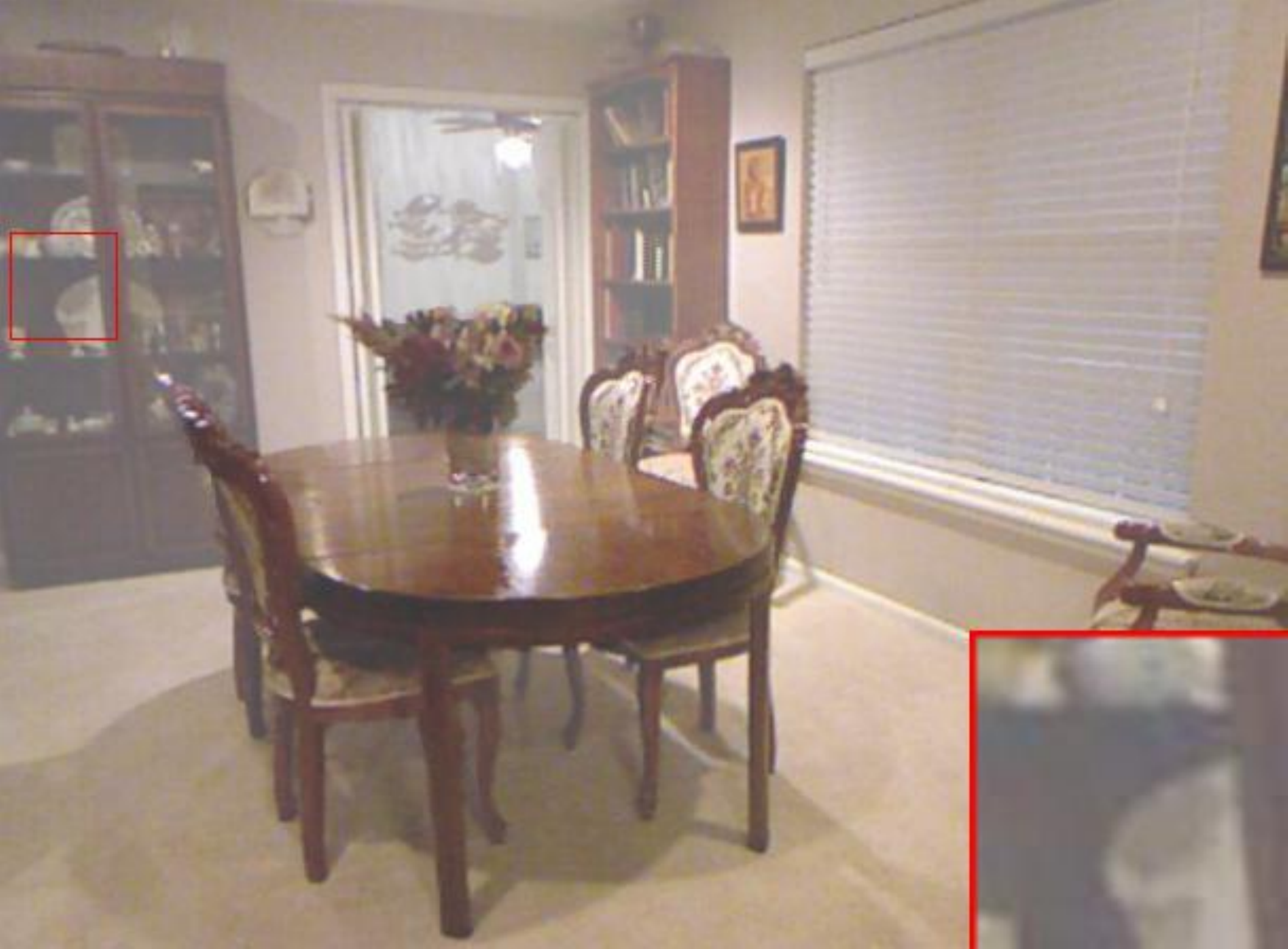}}
		\subfigure{\includegraphics[width=\m_width\textwidth, height=\a_height]{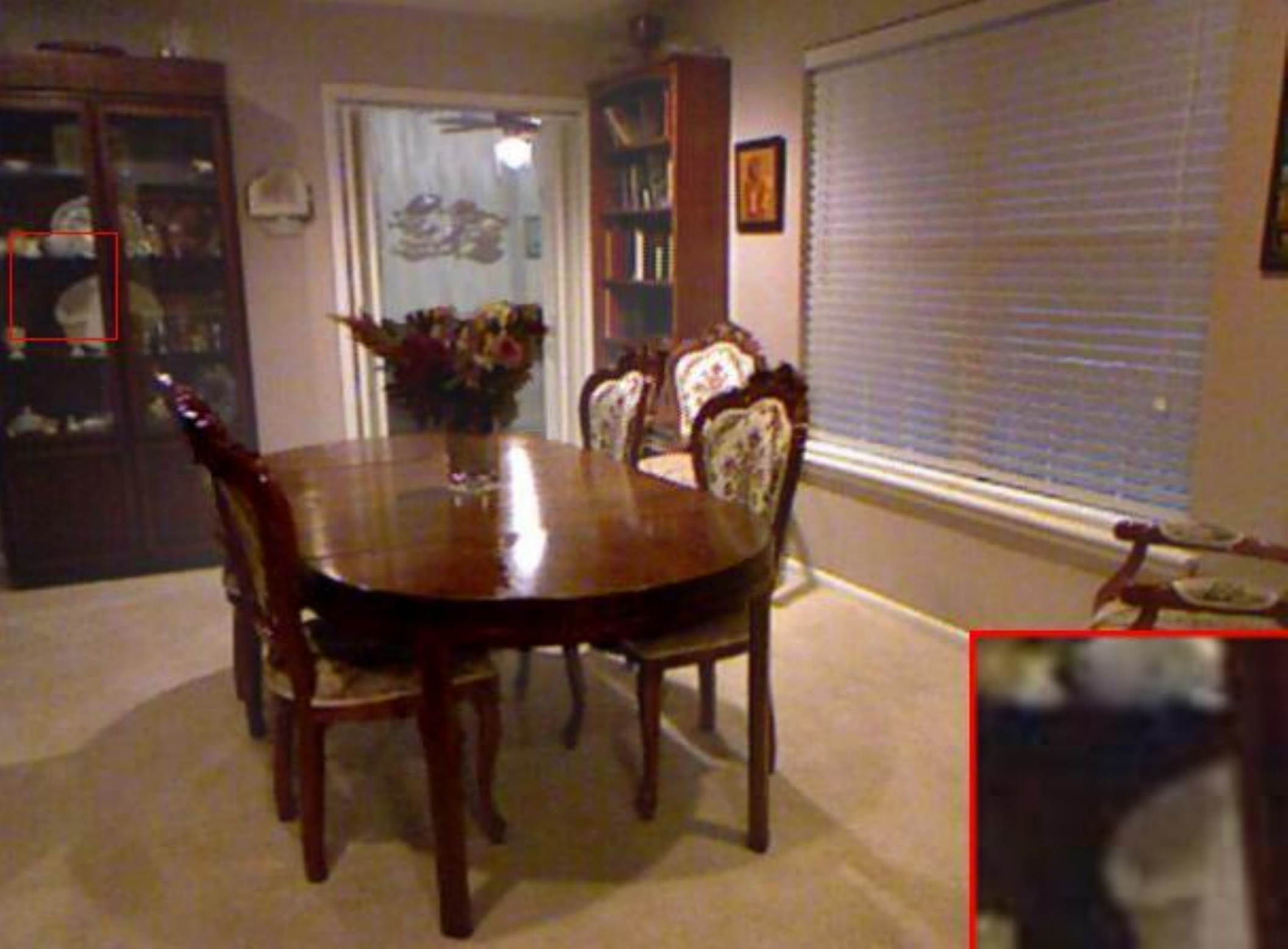}}
		\subfigure{\includegraphics[width=\m_width\textwidth, height=\a_height]{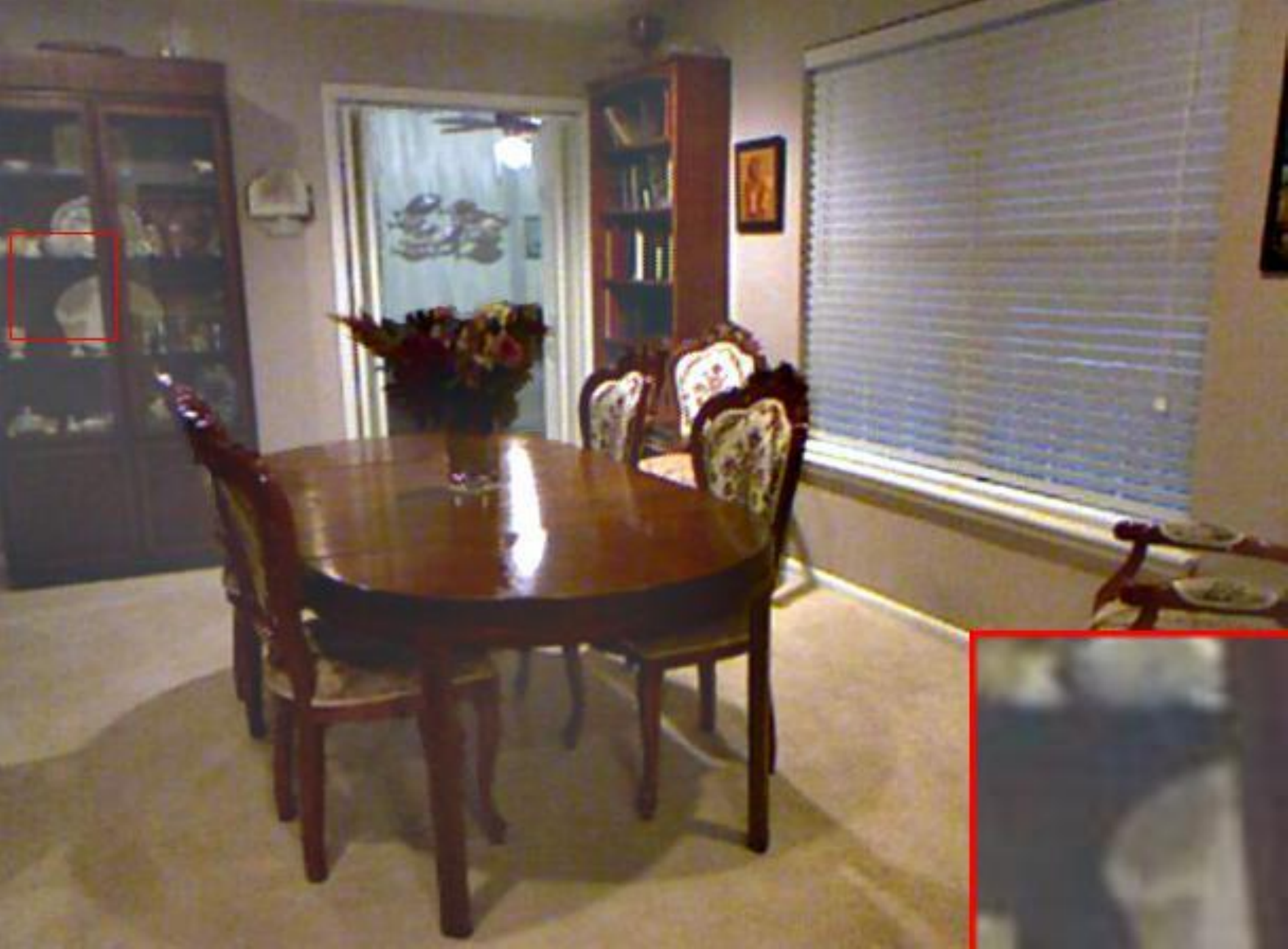}}
		\subfigure{\includegraphics[width=\m_width\textwidth, height=\a_height]{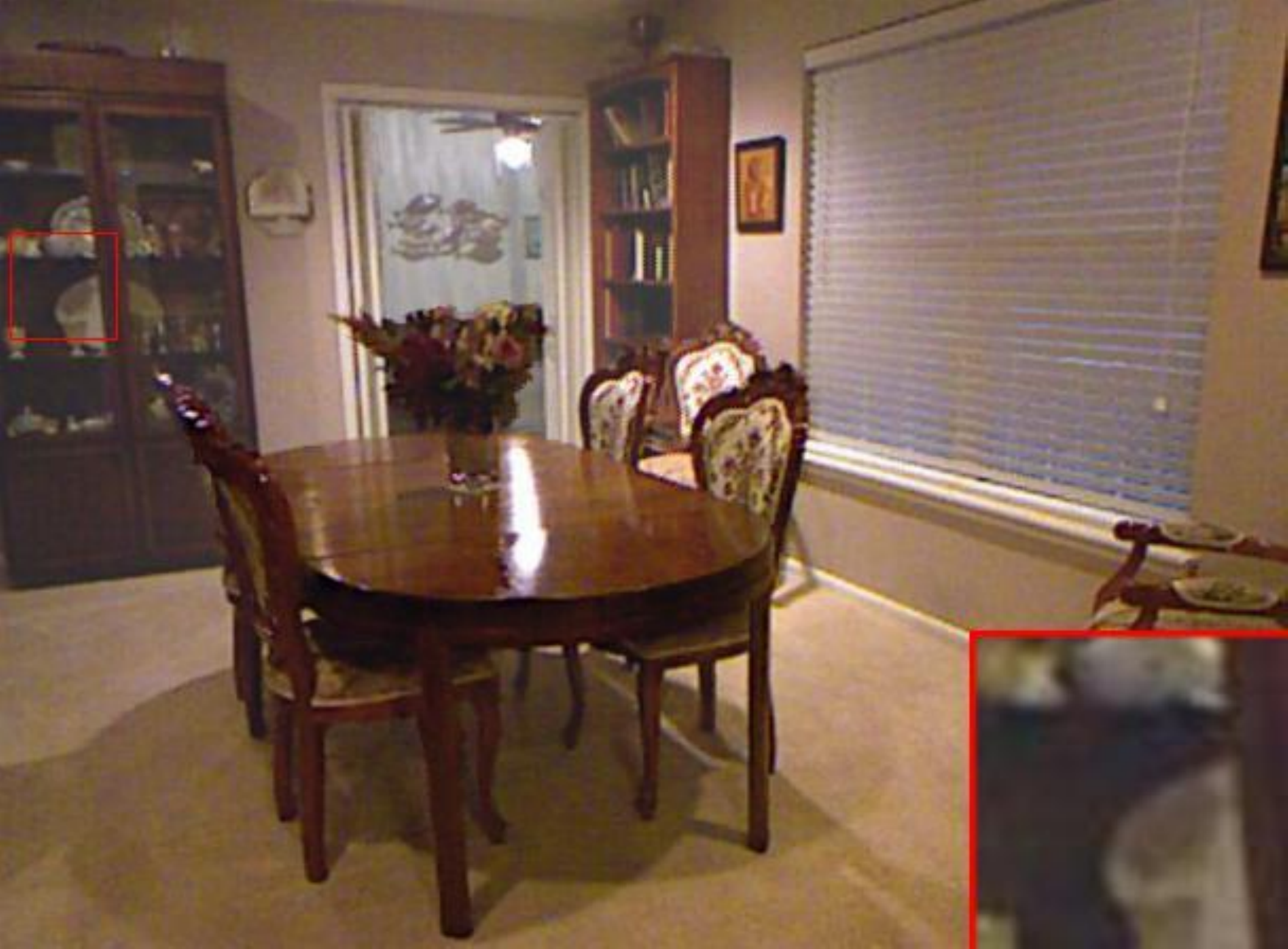}}
		\subfigure{\includegraphics[width=\m_width\textwidth, height=\a_height]{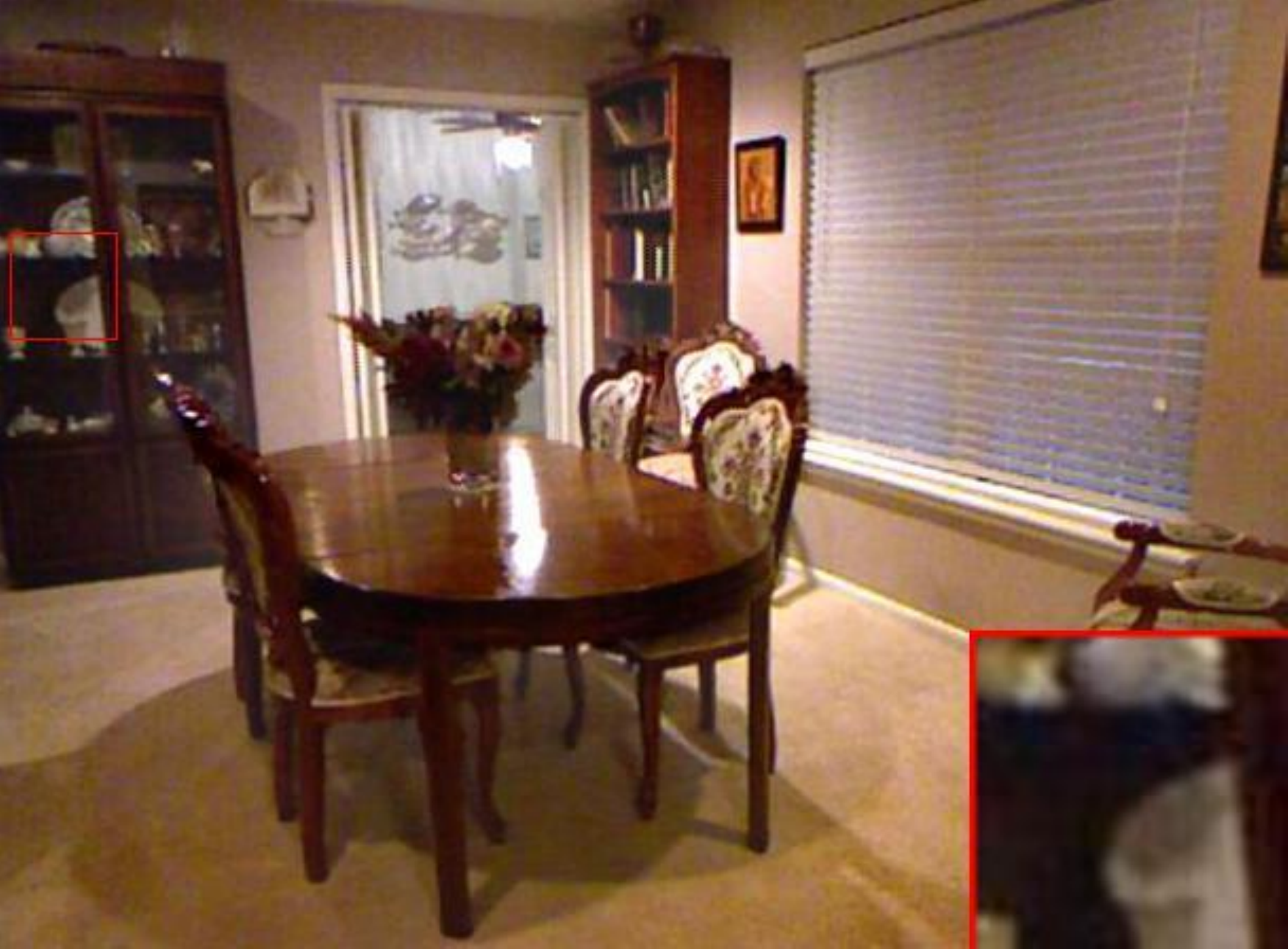}}
		\subfigure{\includegraphics[width=\m_width\textwidth, height=\a_height]{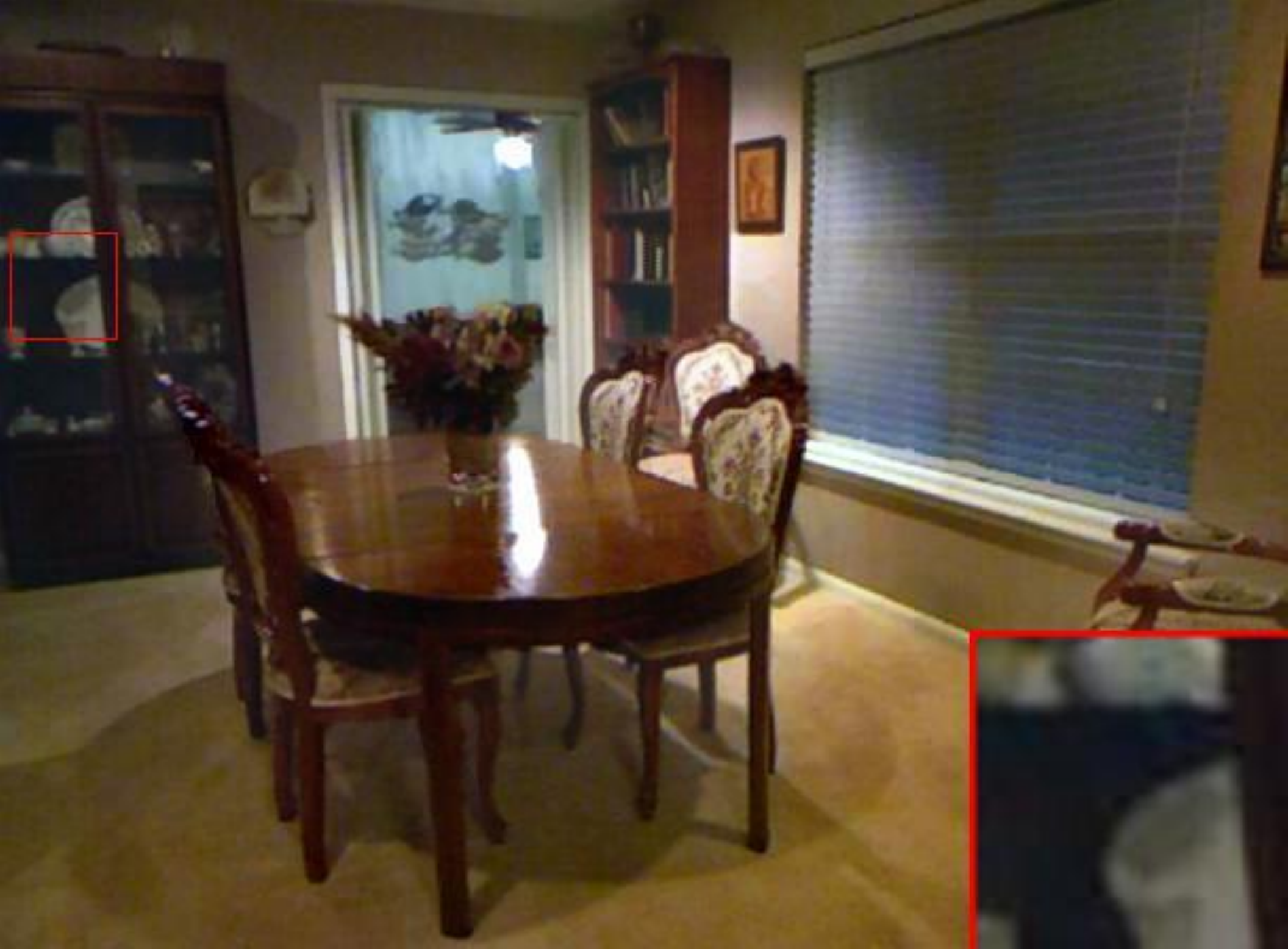}}
		\subfigure{\includegraphics[width=\m_width\textwidth, height=\a_height]{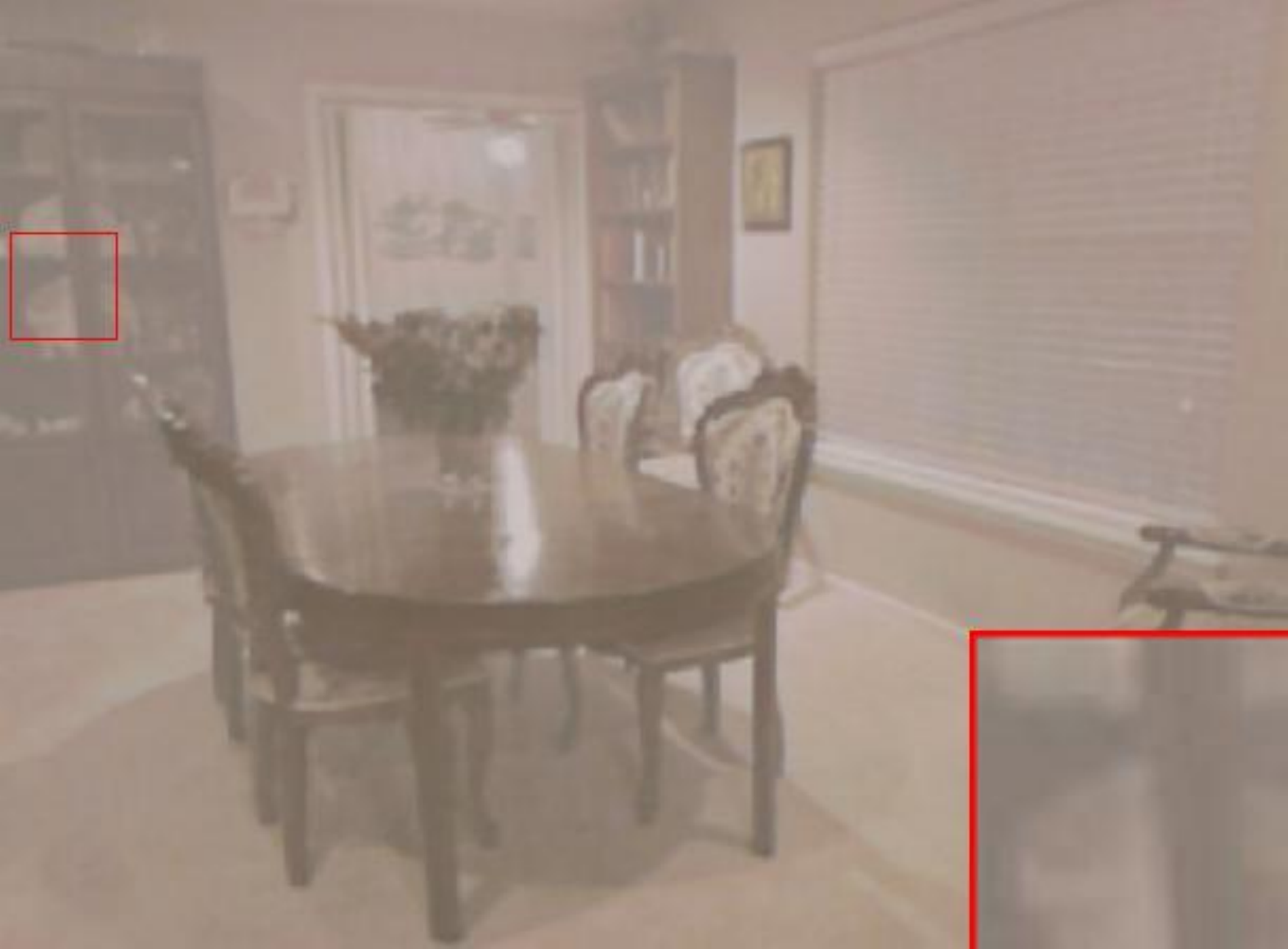}}
		\subfigure{\includegraphics[width=\m_width\textwidth, height=\a_height]{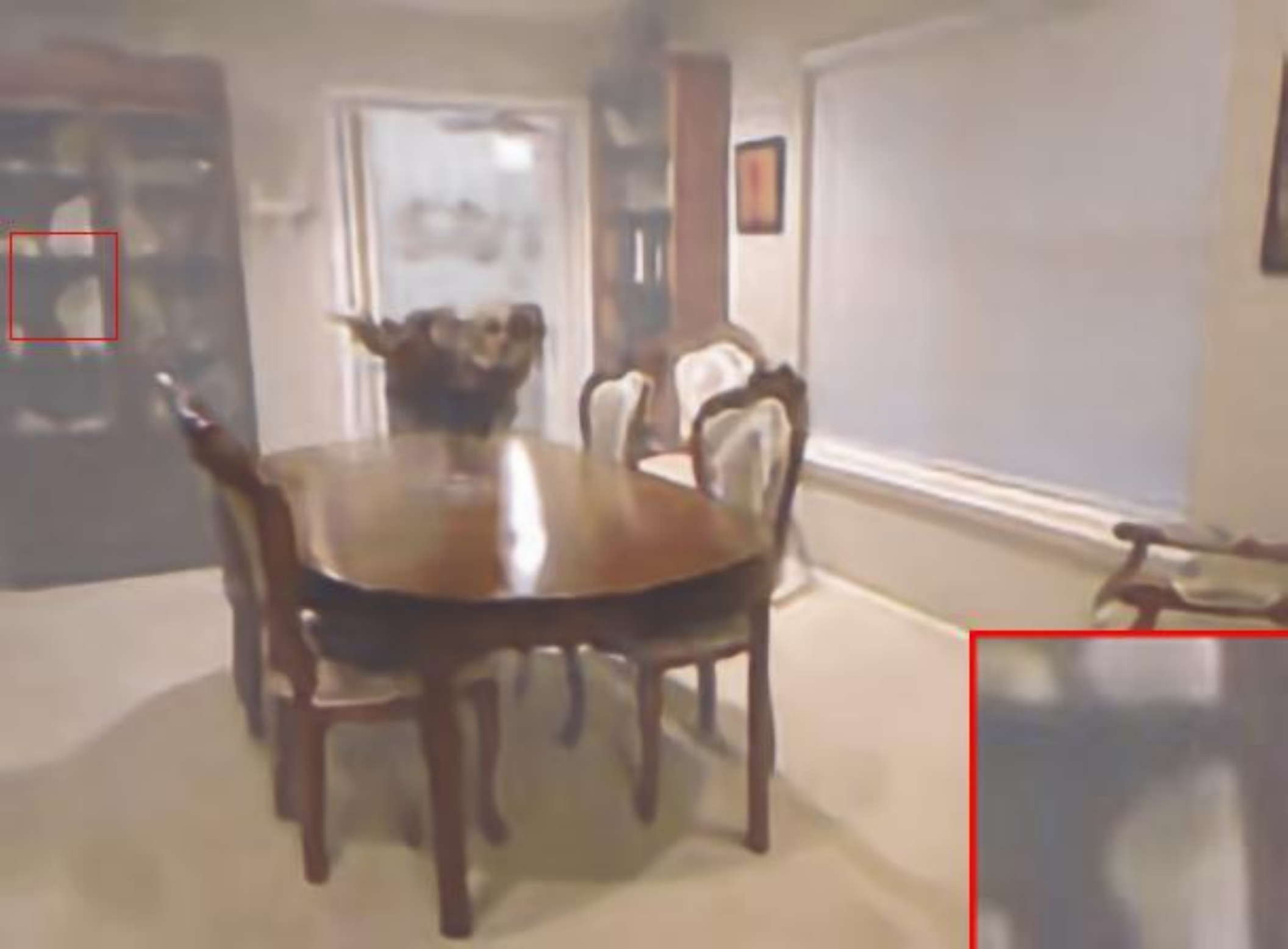}}
		\subfigure{\includegraphics[width=\m_width\textwidth, height=\a_height]{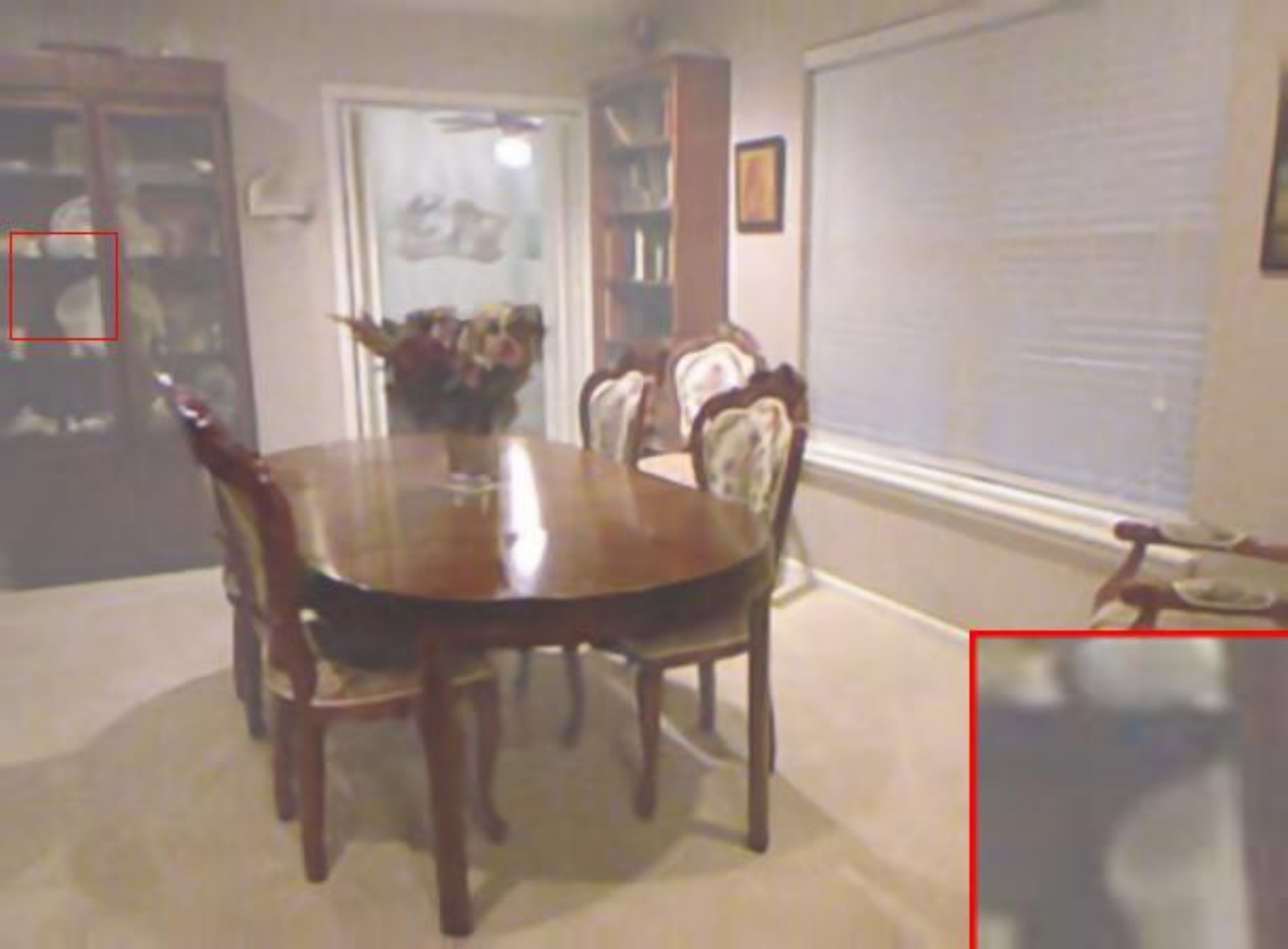}}
		\subfigure{\includegraphics[width=\m_width\textwidth, height=\a_height]{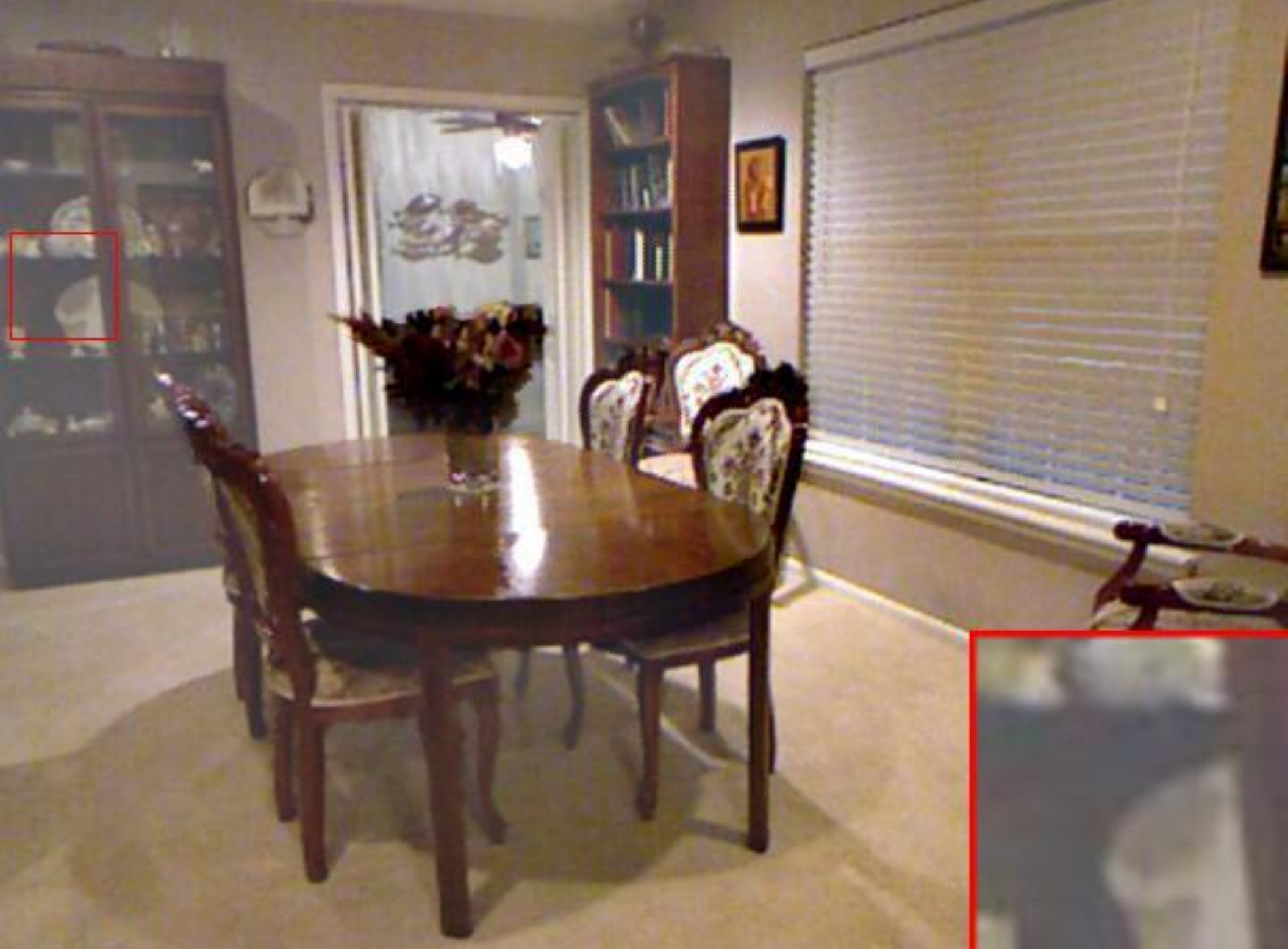}}
		\subfigure{\includegraphics[width=\m_width\textwidth, height=\a_height]{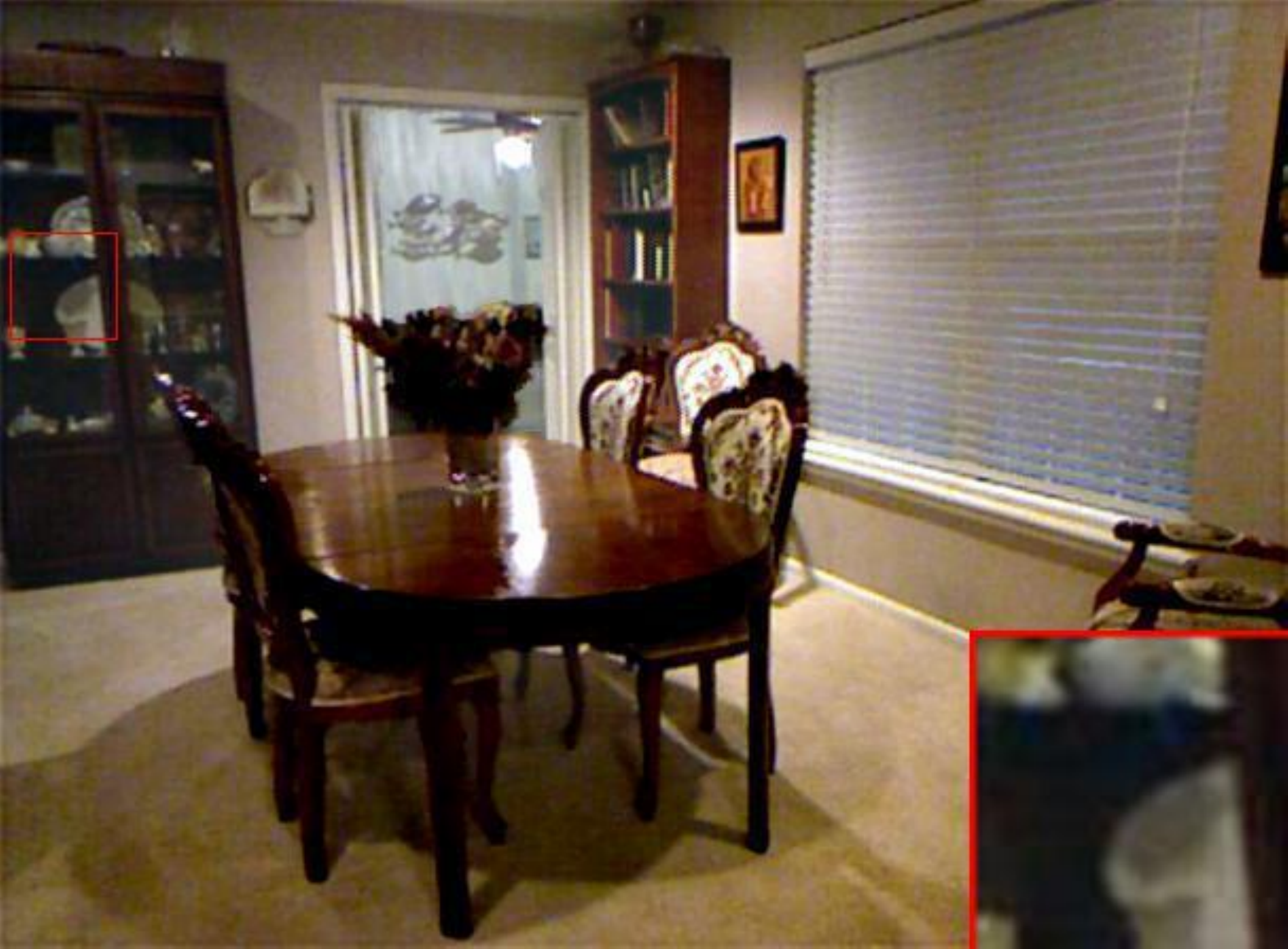}}
	\subfigure{\includegraphics[width=\m_width\textwidth,height=\a_height]{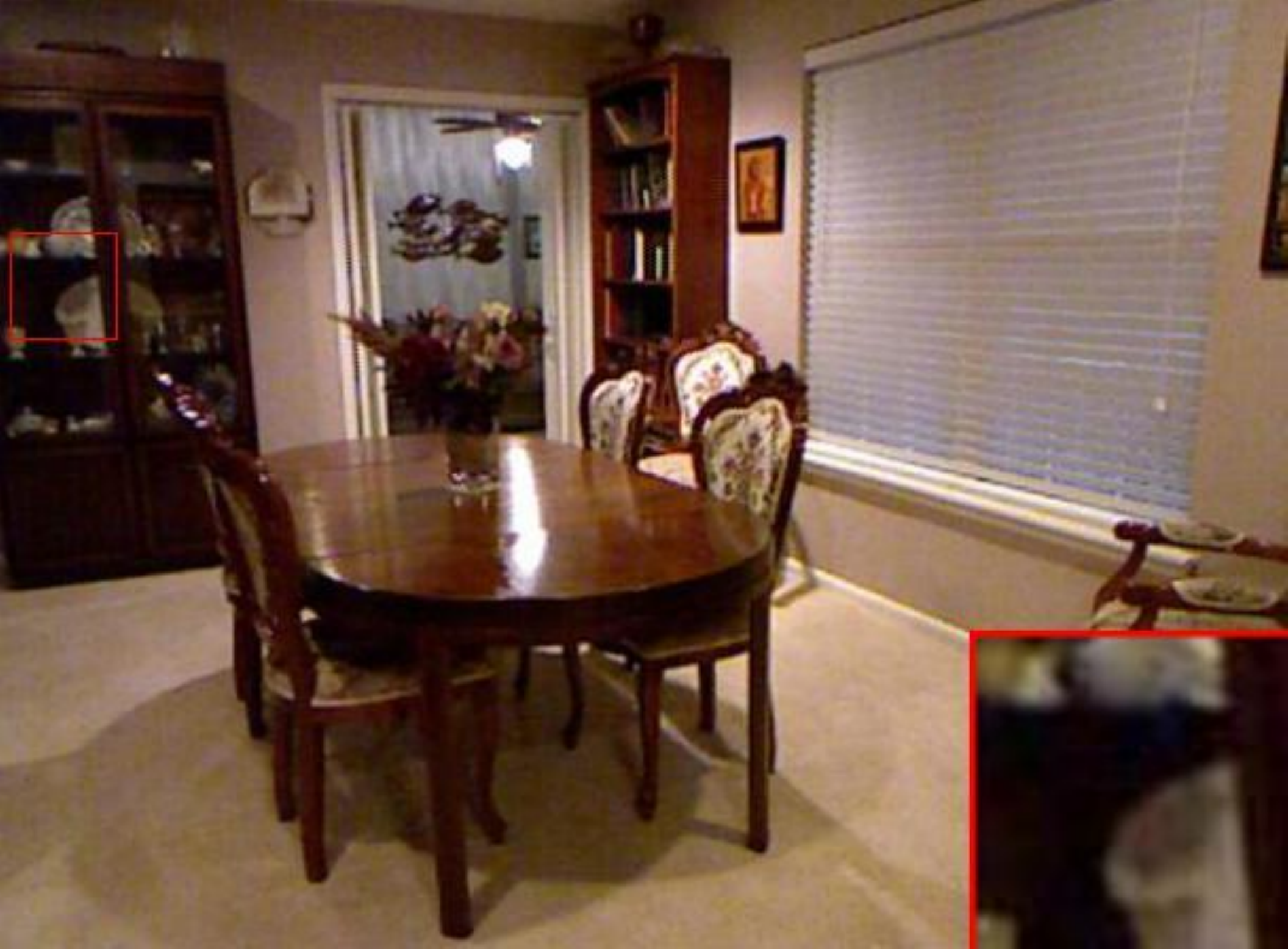}}
	\end{center}
	
	\vspace{-0.7cm}
	
	\begin{center}
	\setcounter{subfigure}{0}
		\subfigure[]{\label{Figure:SOTS:Input}\includegraphics[width=\m_width\textwidth, height=\a_height]{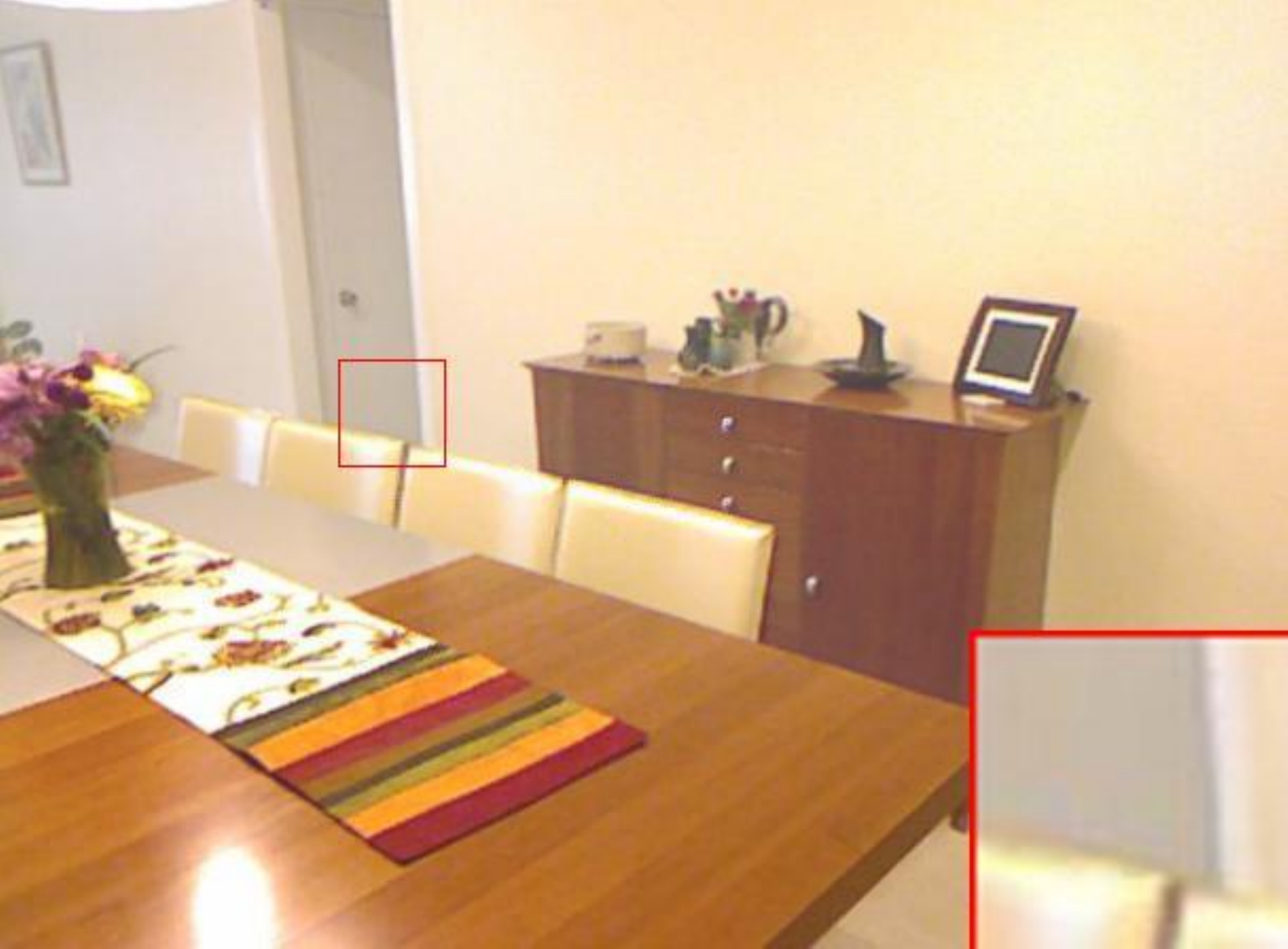}}
		\subfigure[]{\includegraphics[width=\m_width\textwidth, height=\a_height]{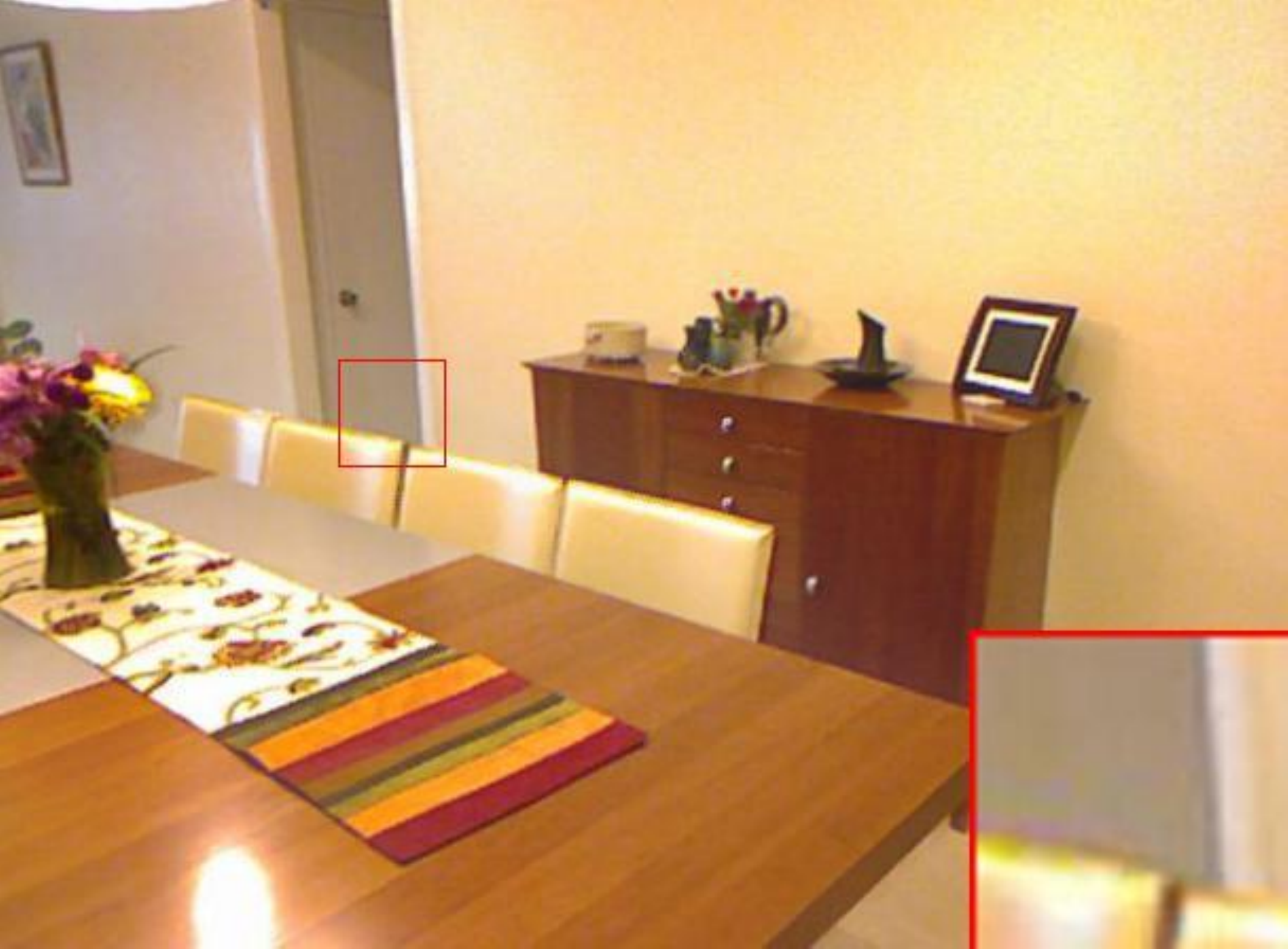}}
		\subfigure[]{\includegraphics[width=\m_width\textwidth, height=\a_height]{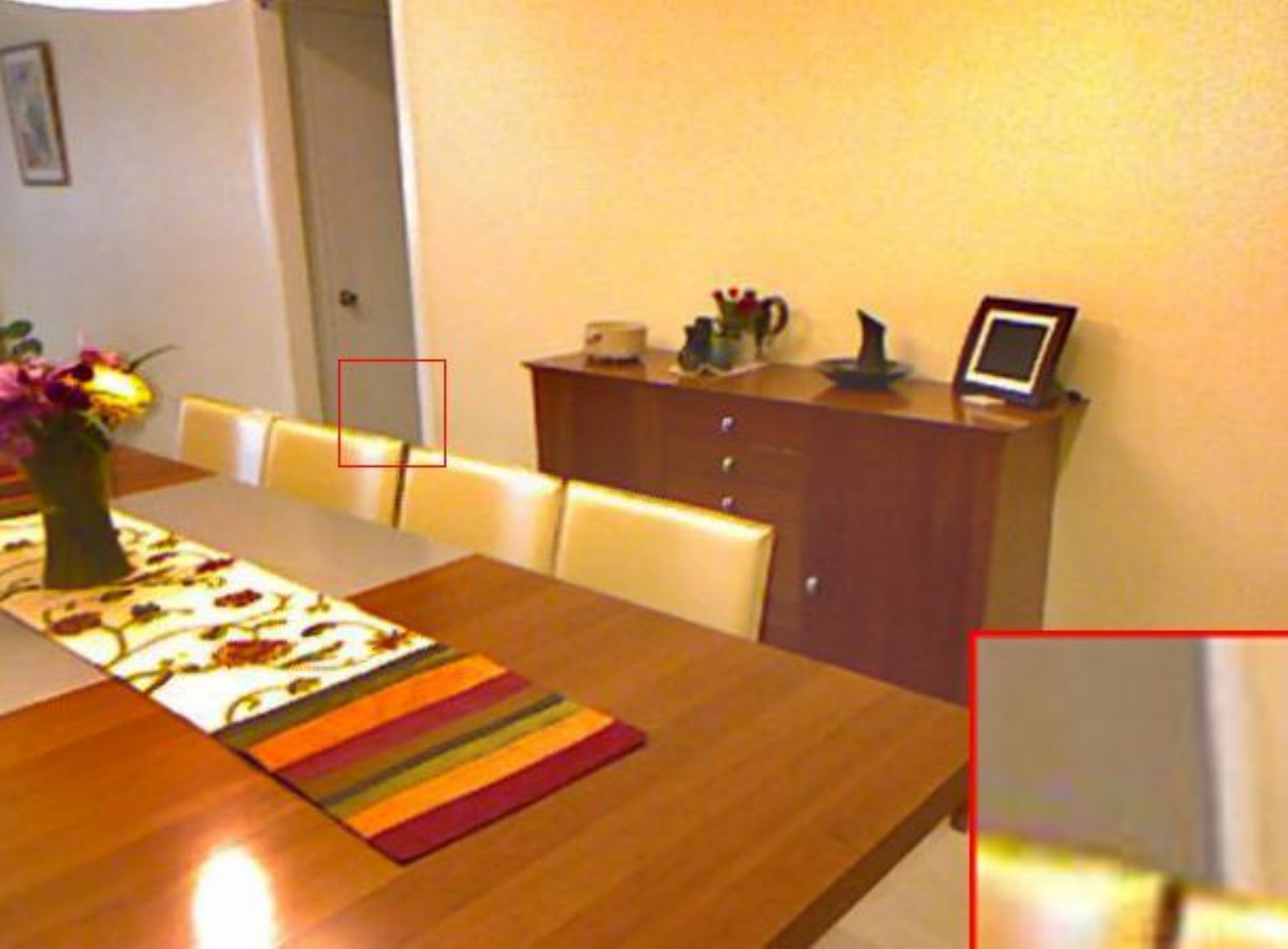}}
		\subfigure[]{\includegraphics[width=\m_width\textwidth, height=\a_height]{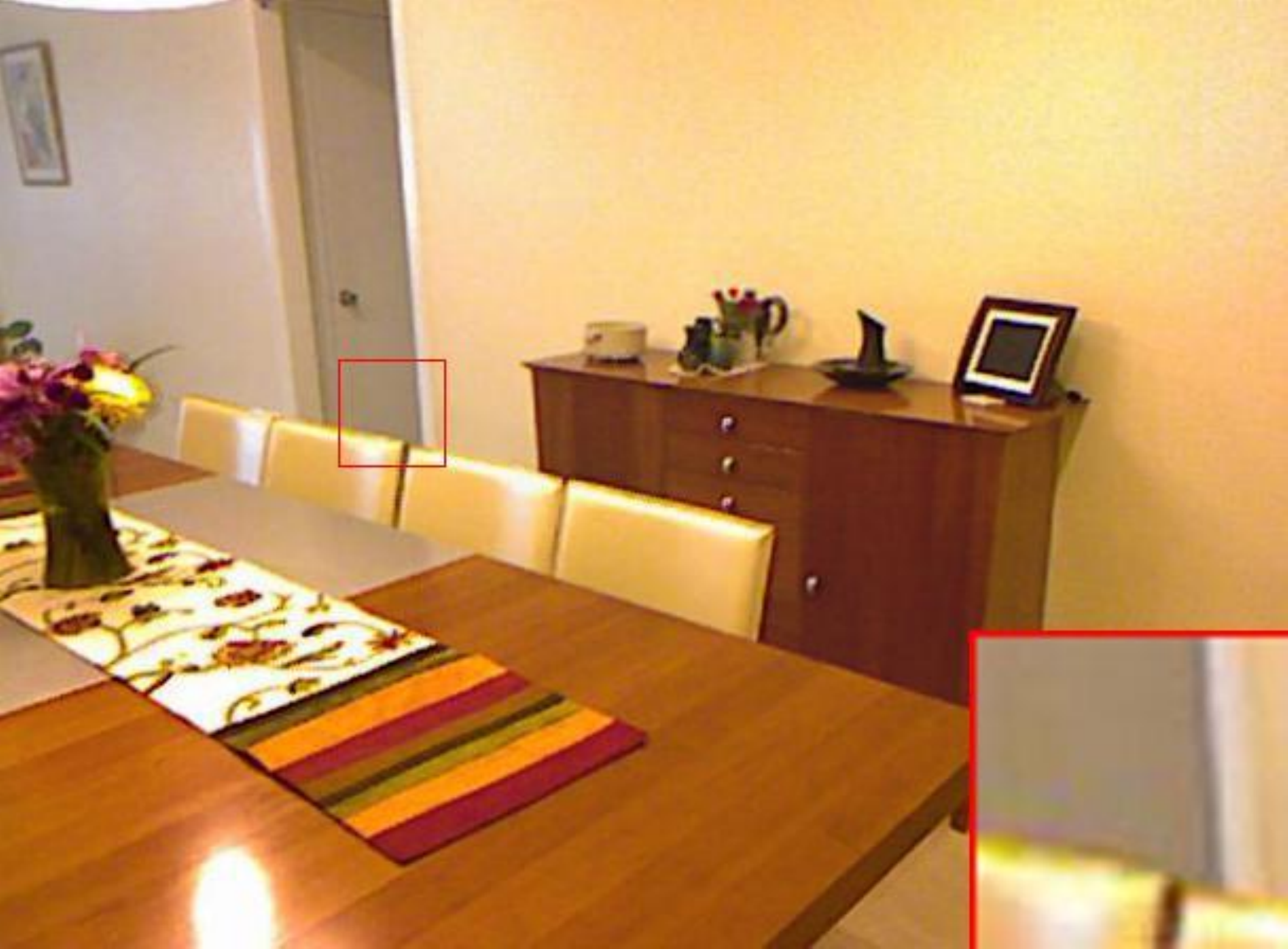}}
		\subfigure[]{\includegraphics[width=\m_width\textwidth, height=\a_height]{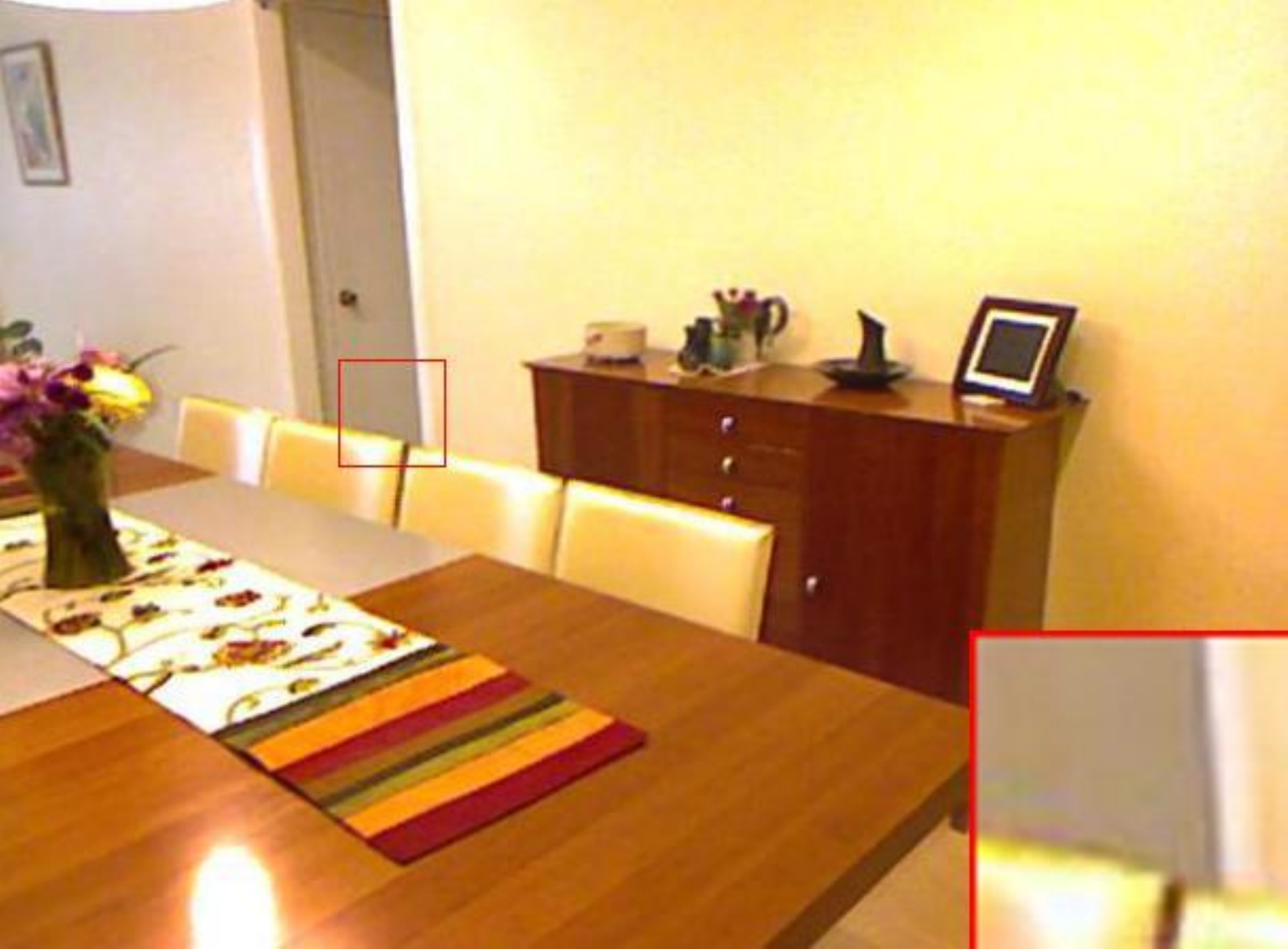}}
		\subfigure[]{\includegraphics[width=\m_width\textwidth, height=\a_height]{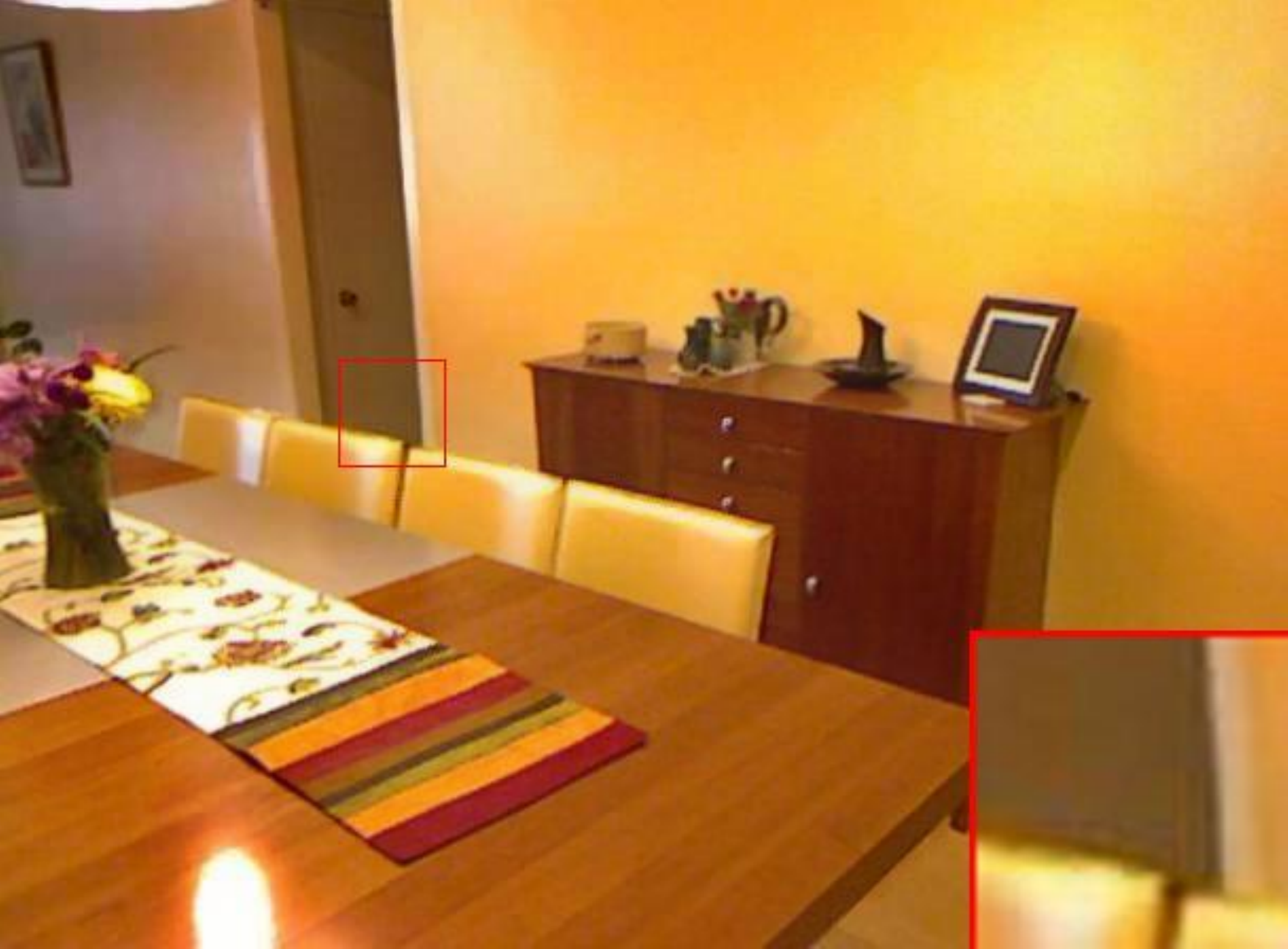}}
		\subfigure[]{\includegraphics[width=\m_width\textwidth, height=\a_height]{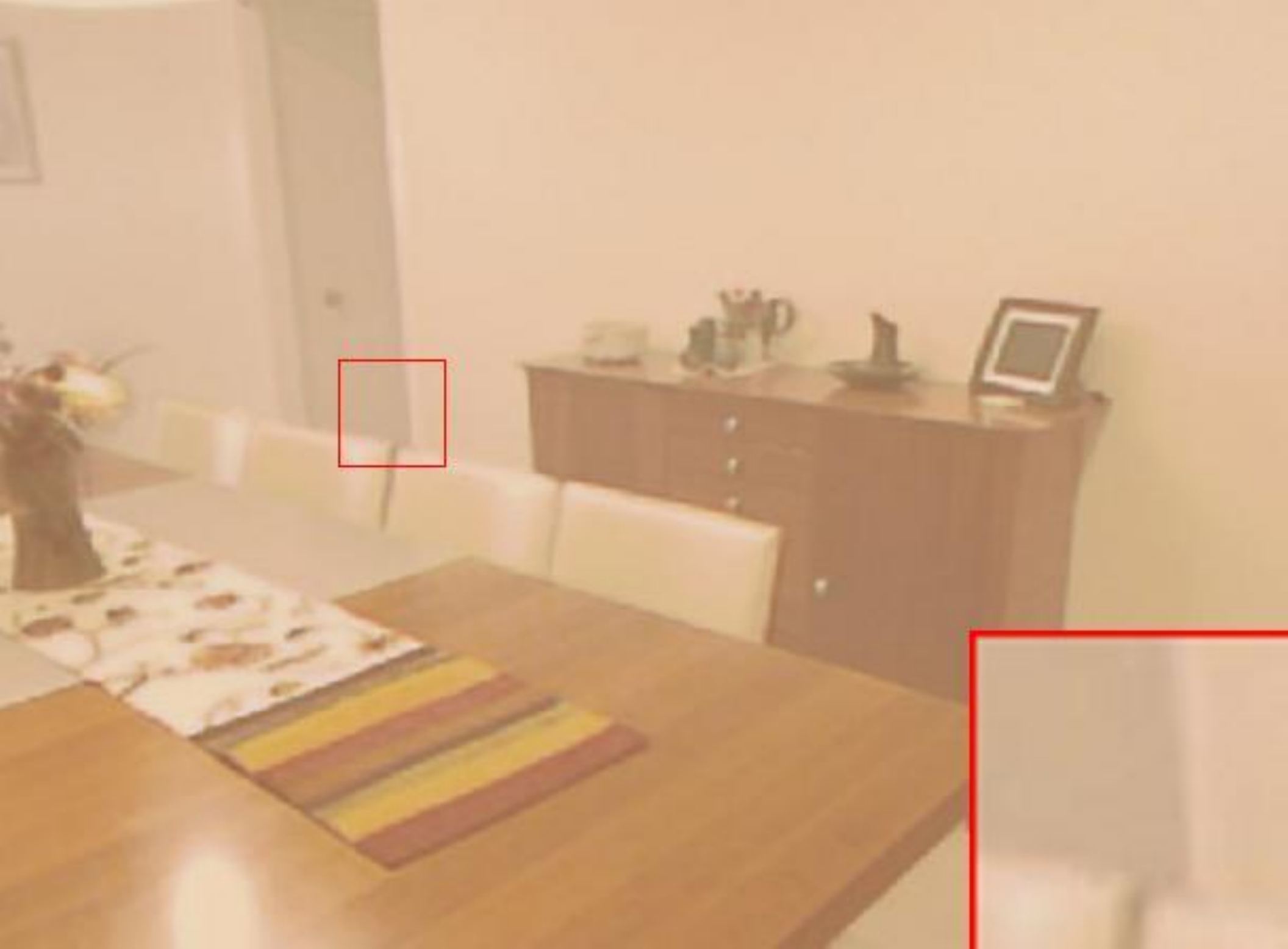}}
		\subfigure[]{\includegraphics[width=\m_width\textwidth, height=\a_height]{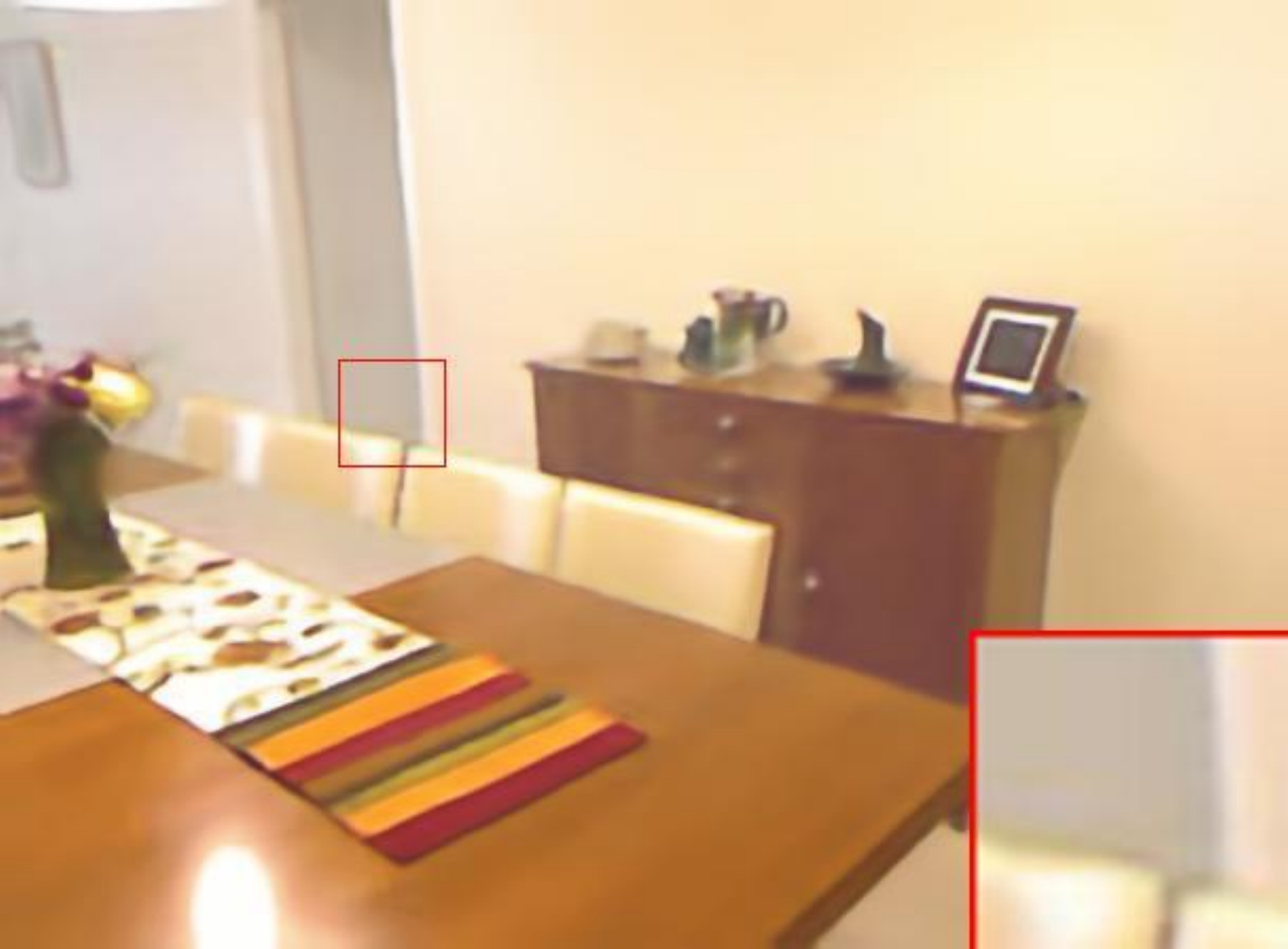}}
		\subfigure[]{\includegraphics[width=\m_width\textwidth, height=\a_height]{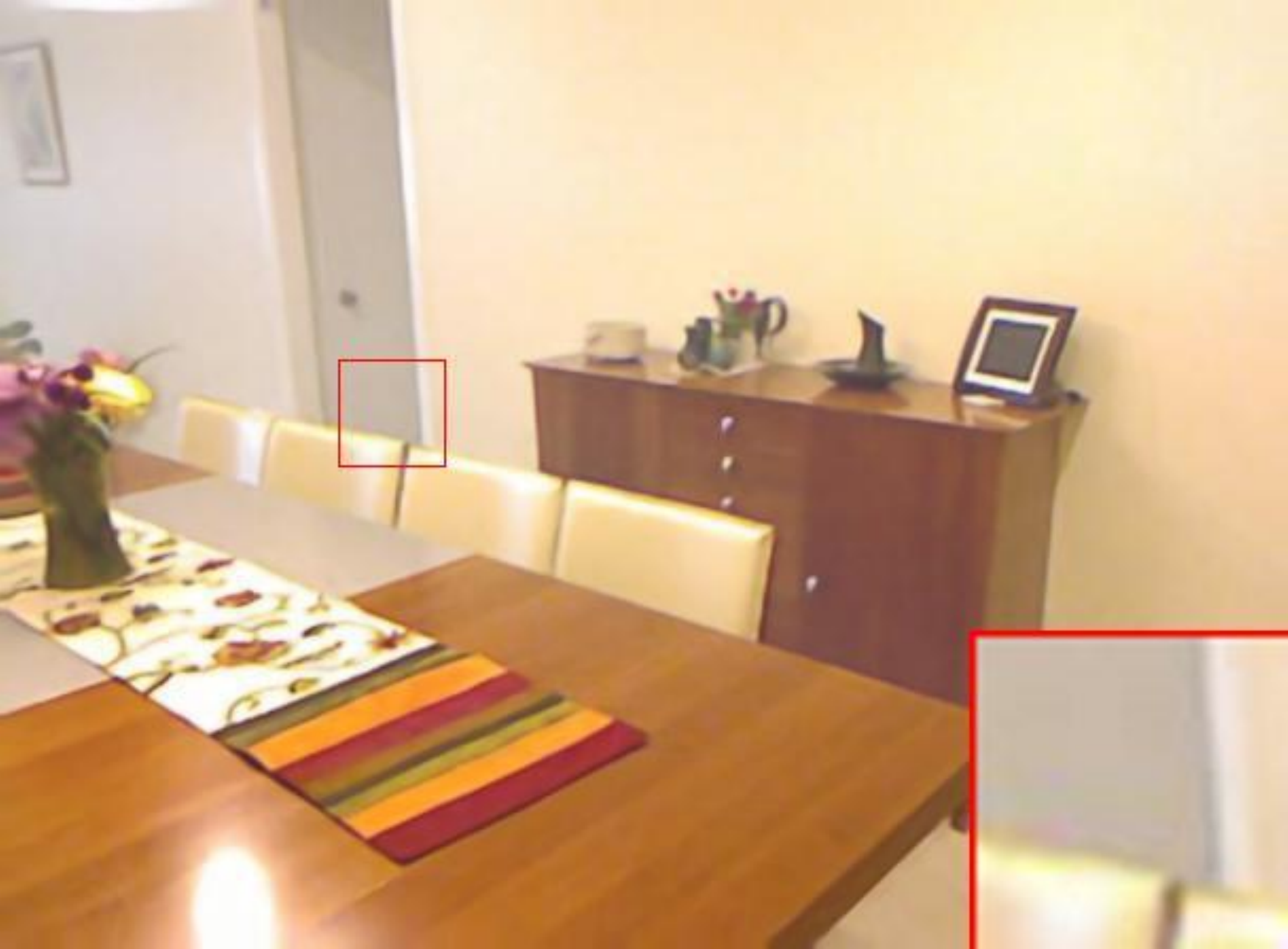}}
		\subfigure[]{\includegraphics[width=\m_width\textwidth, height=\a_height]{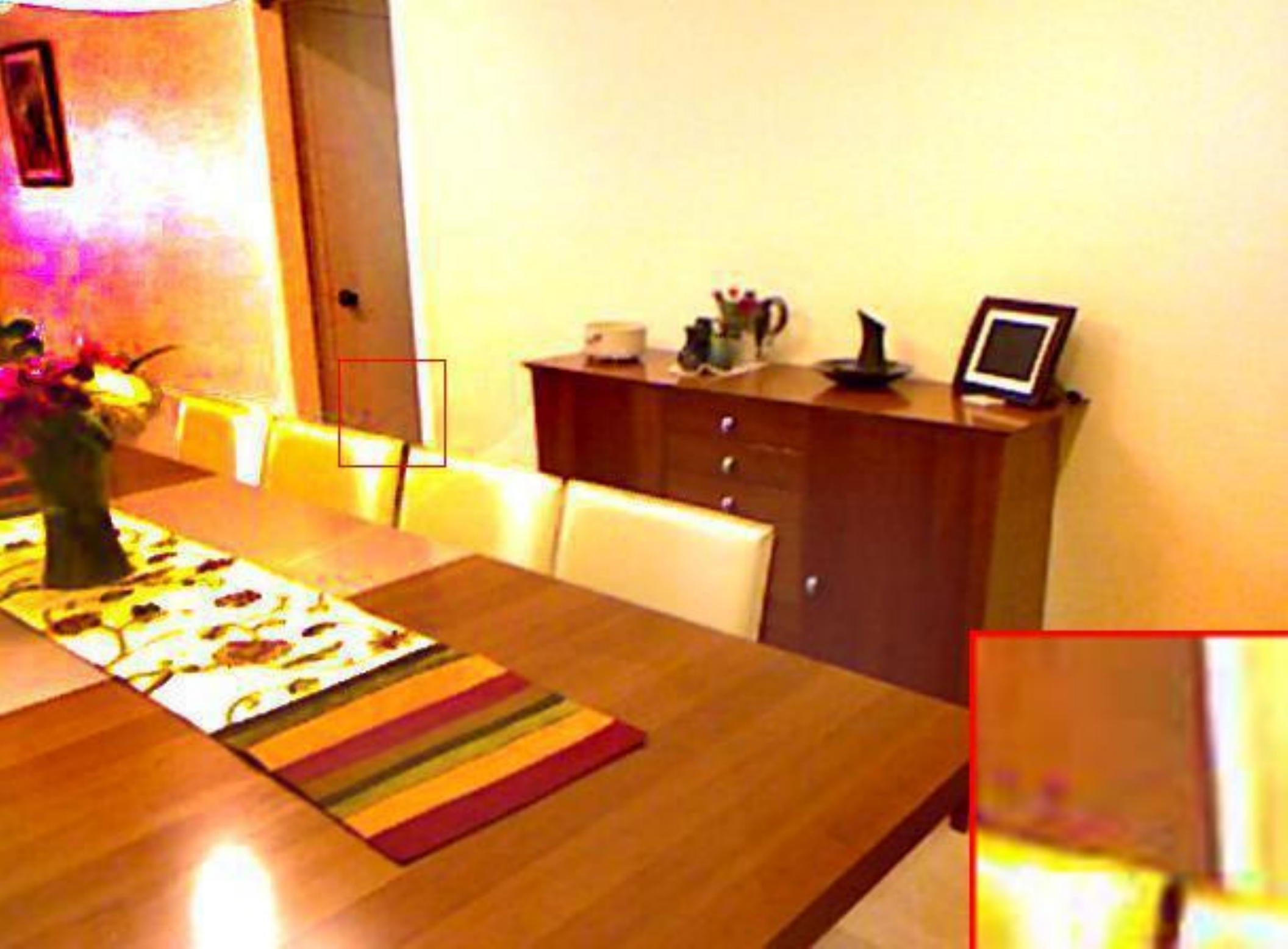}}
		\subfigure[]{\includegraphics[width=\m_width\textwidth, height=\a_height]{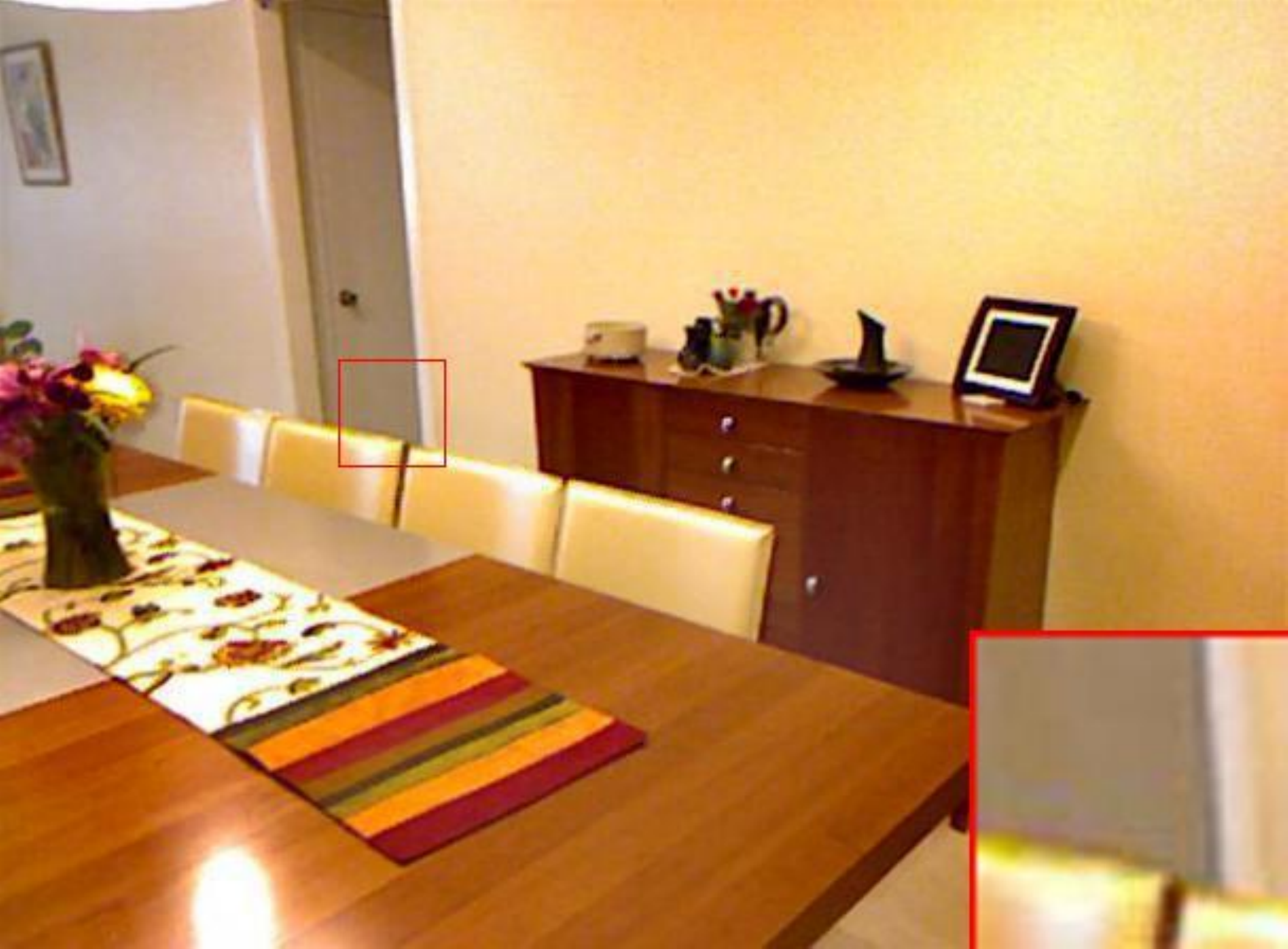}}
		\subfigure[]{\label{Figure:SOTS:GT}\includegraphics[width=\m_width\textwidth,height=\a_height]{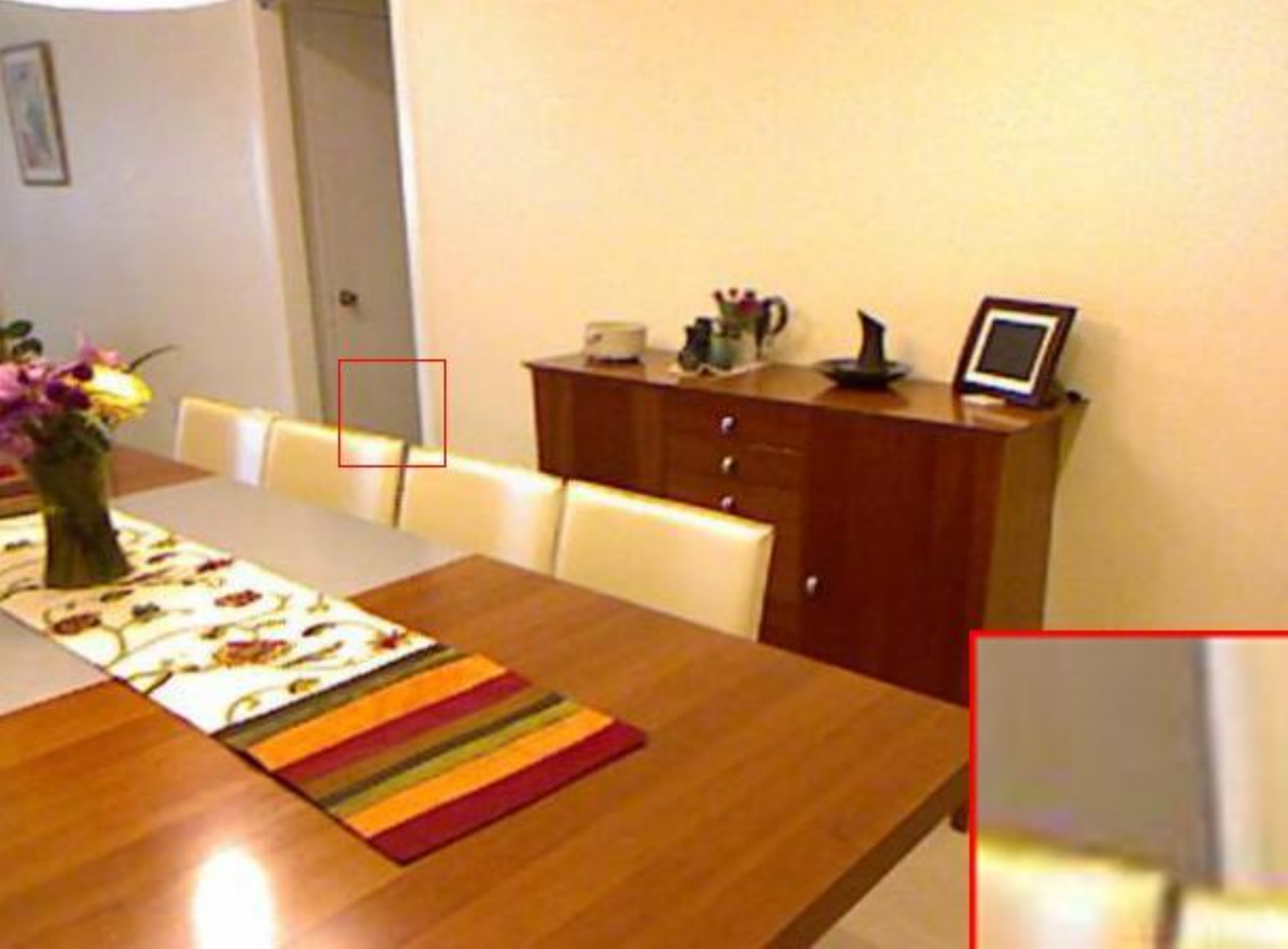}}
	\end{center}
\caption{\label{Figure:SOTS} Visual Results on SOTS. From the left to the right column (\textit{i.e.}, Figs.~(\ref{Figure:SOTS:Input}--\ref{Figure:SOTS:GT})), the input hazy image, DehazeNet~\citep{DehazeNet}, MSCNN~\citep{MSCNN}, AOD-Net~\citep{AOD-Net}, DCP~\citep{DCP},  GRM~\citep{GRM}, N2V~\citep{N2V}, DIP~\citep{DIP}, DD~\citep{DD}, DDIP~\citep{Double-DIP}, our YOLY and the ground truth are presented in column-wise.  Some areas are highlighted by red rectangles and zooming-in is recommended for a better visualization and comparison. }
\end{figure*}

\begin{table*}[!t]
\centering
\caption{Results on the synthetic outdoor database (HSTS). The bold number indicates the best method of each category of methods.}
\label{Table:HSTS}
\begin{tabular}{c| cccc ccccc}
\toprule
\multicolumn{1}{c|}{\multirow{2}{*}{Metrics}} & \multicolumn{4}{c|}{Learning-based Dehazing Methods} & \multicolumn{5}{c}{Prior-based Dehazing Methods} \\
\cline{2-10}
 & DehazeNet & MSCNN & AOD-Net & \multicolumn{1}{c|}{CAP} & DCP & FVR & BCCR & GRM & NLD\\ 
\hline
PSNR & \textbf{24.48} & 18.64 & 20.55 & \multicolumn{1}{c|}{21.53} & 14.84 & 14.48 & 15.08 & 18.54 & \textbf{18.92}\\ 
SSIM & \textbf{0.9153} & 0.8168 & 0.8973 & \multicolumn{1}{c|}{0.8726} & 0.7609 & 0.7624 & 0.7382 & \textbf{0.8184} & 0.7411 \\ 
\midrule
\multicolumn{1}{c|}{\multirow{2}{*}{Metrics}} & \multicolumn{6}{c|}{Unsupervised Neural Networks}  \\
\cline{2-7}
 & N2N & N2V & DIP & DD & \multicolumn{1}{c|}{DDIP} & \multicolumn{1}{c|}{Ours} \\ 
\cline{1-7}
PSNR & - & 11.79 & 14.55 & 14.66 & \multicolumn{1}{c|}{20.91} & \multicolumn{1}{c|}{\textbf{23.82}} \\ 
SSIM & - & 0.5450 & 0.5573 & 0.6409 & \multicolumn{1}{c|}{0.8842} & \multicolumn{1}{c|}{\textbf{0.9125}} \\ 
\bottomrule
\end{tabular}
\end{table*}

\begin{figure*}[!t]
	\def \m_width{0.075}
	\def \a_height{2cm}
	\def \b_height{1cm}
	\begin{center}
		\subfigure{\includegraphics[width=\m_width\textwidth, height=\a_height]{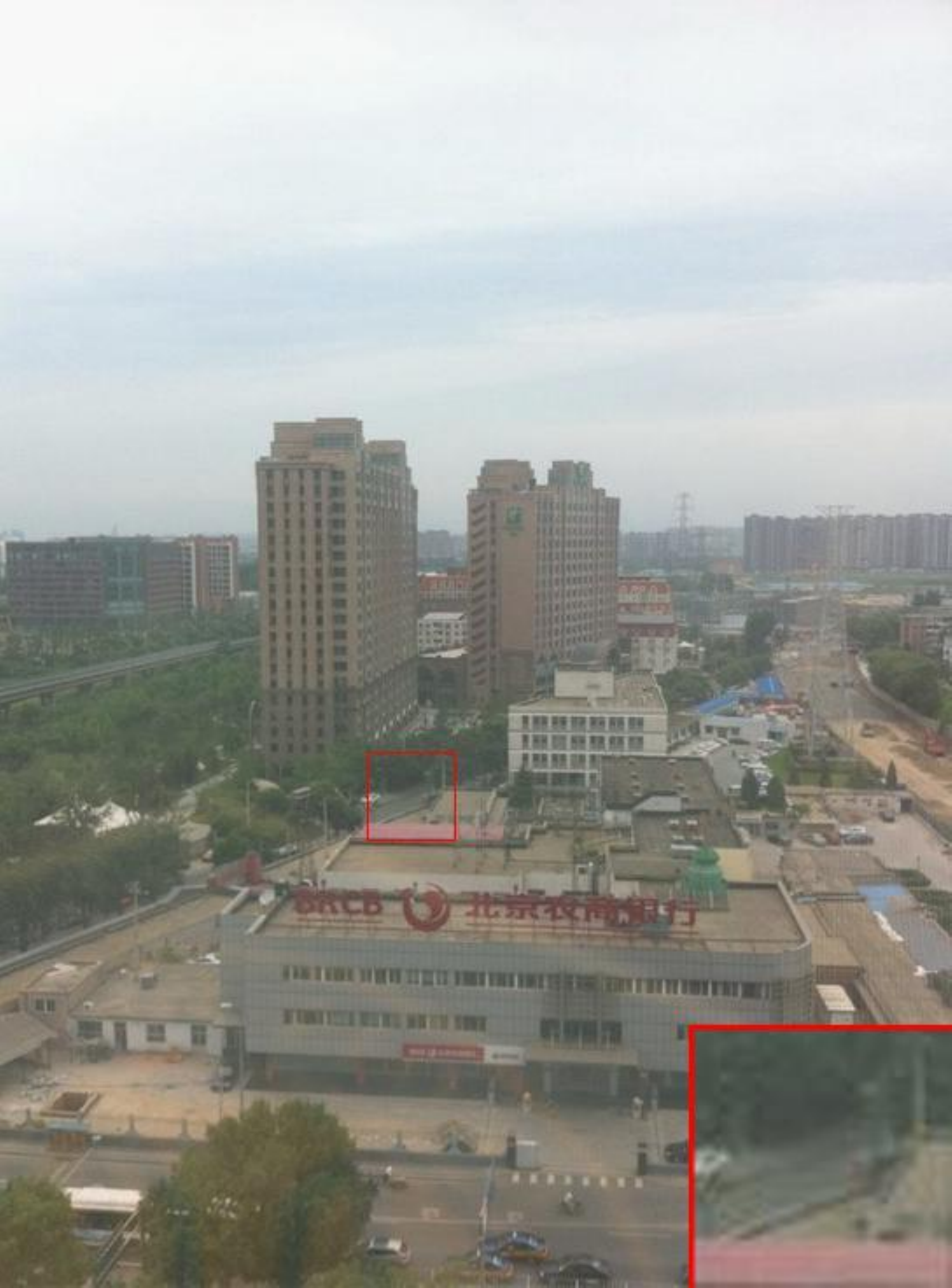}}
		\subfigure{\includegraphics[width=\m_width\textwidth, height=\a_height]{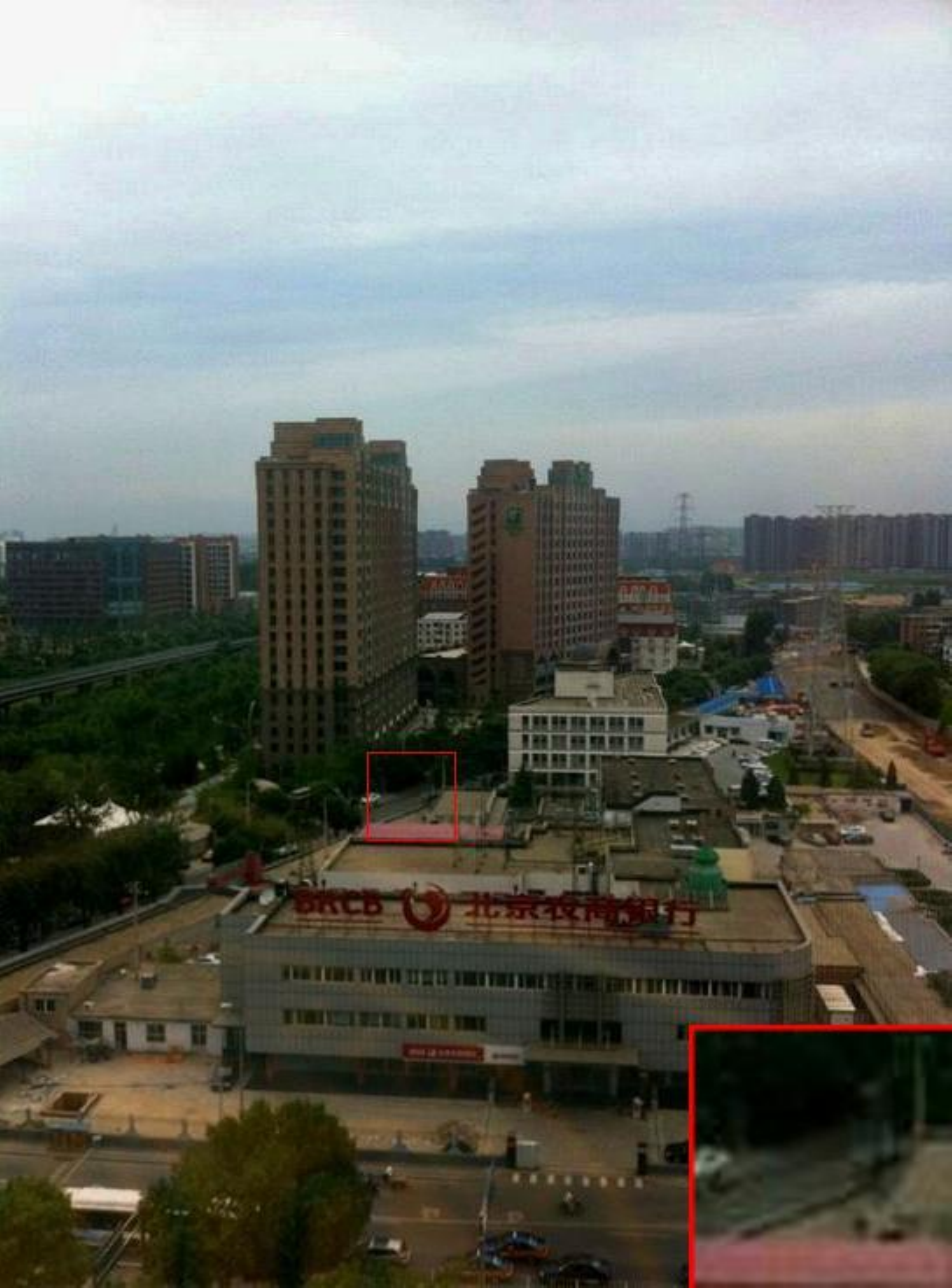}}
		\subfigure{\includegraphics[width=\m_width\textwidth, height=\a_height]{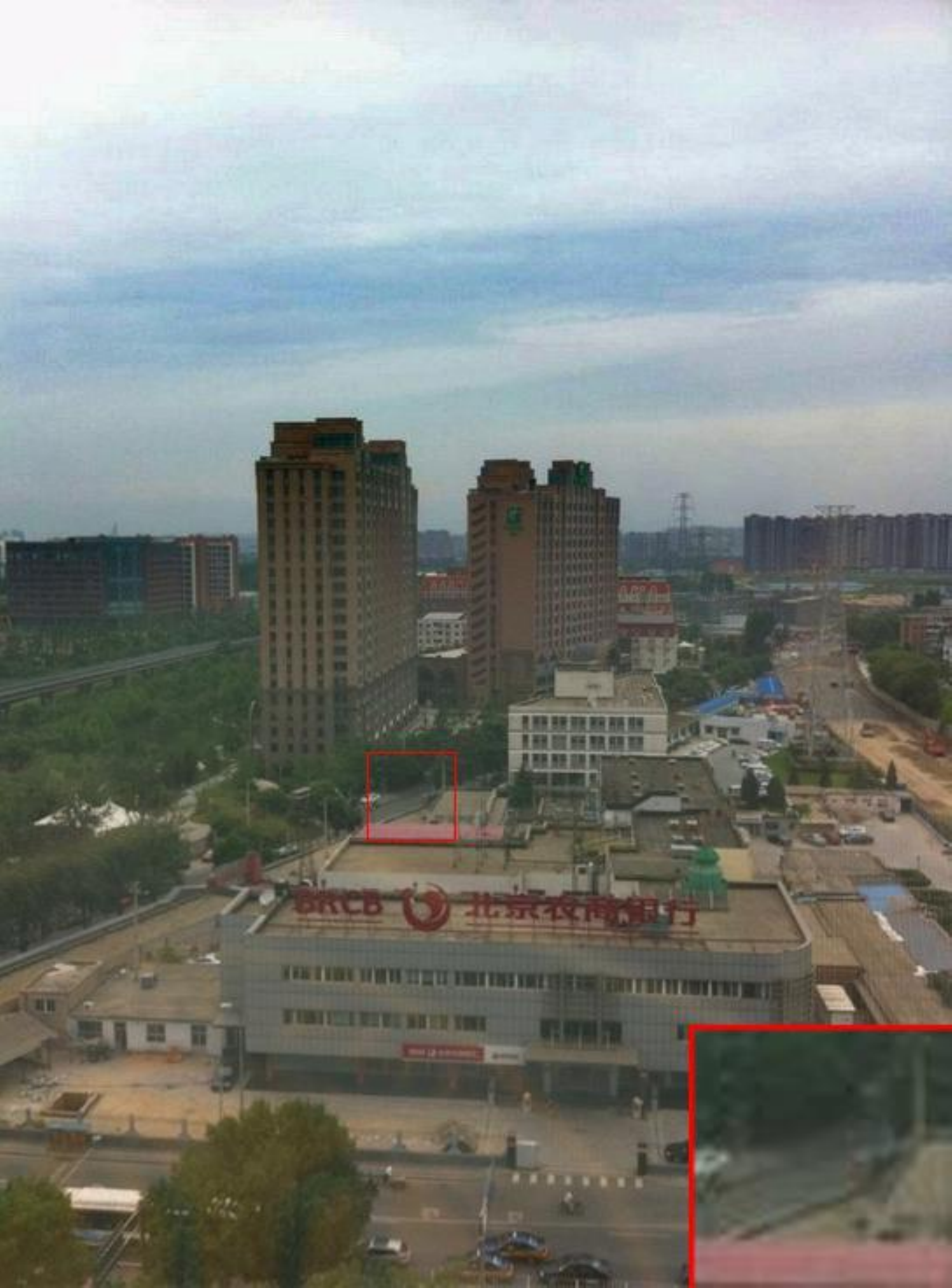}}
		\subfigure{\includegraphics[width=\m_width\textwidth, height=\a_height]{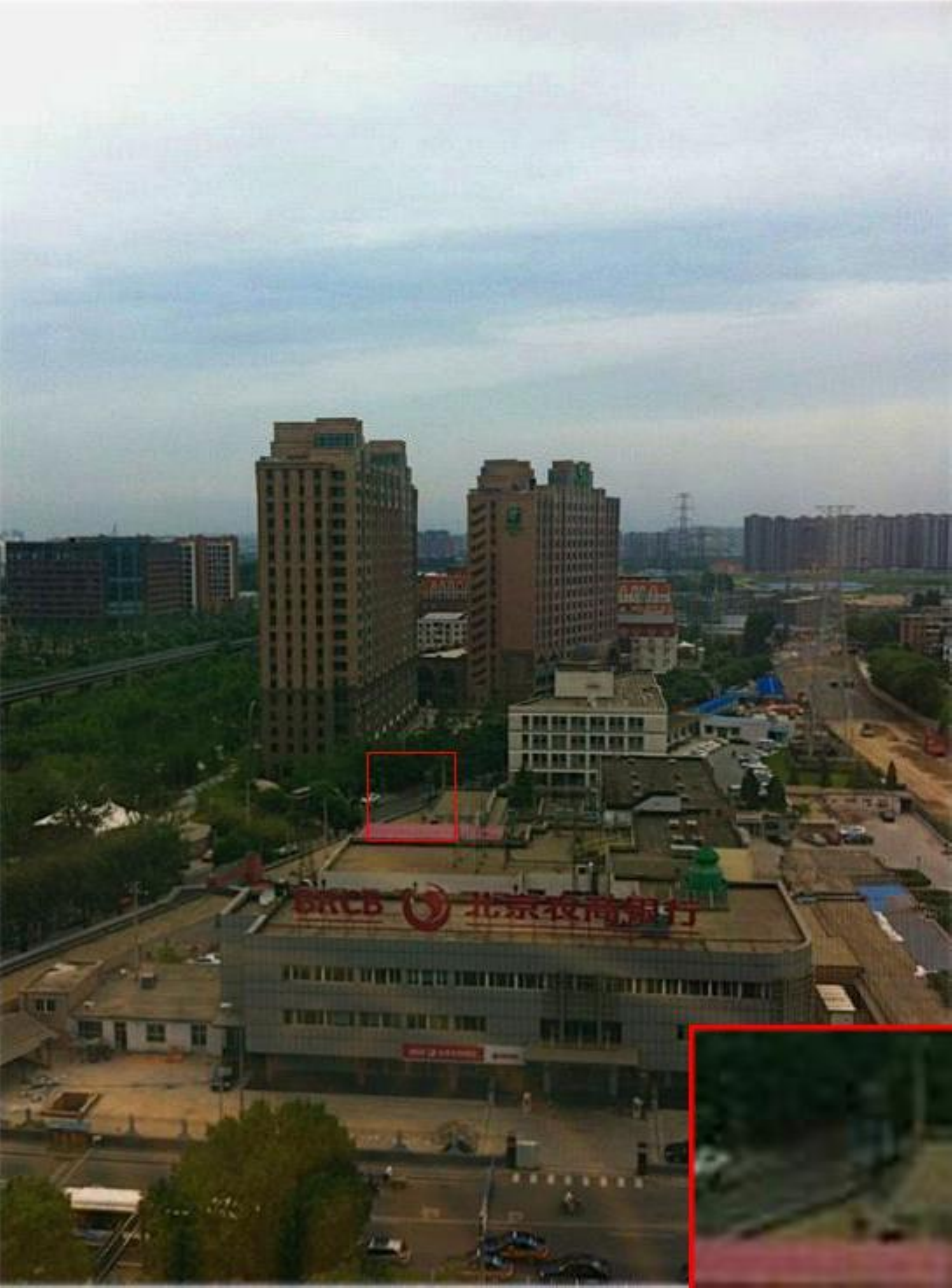}}
		\subfigure{\includegraphics[width=\m_width\textwidth, height=\a_height]{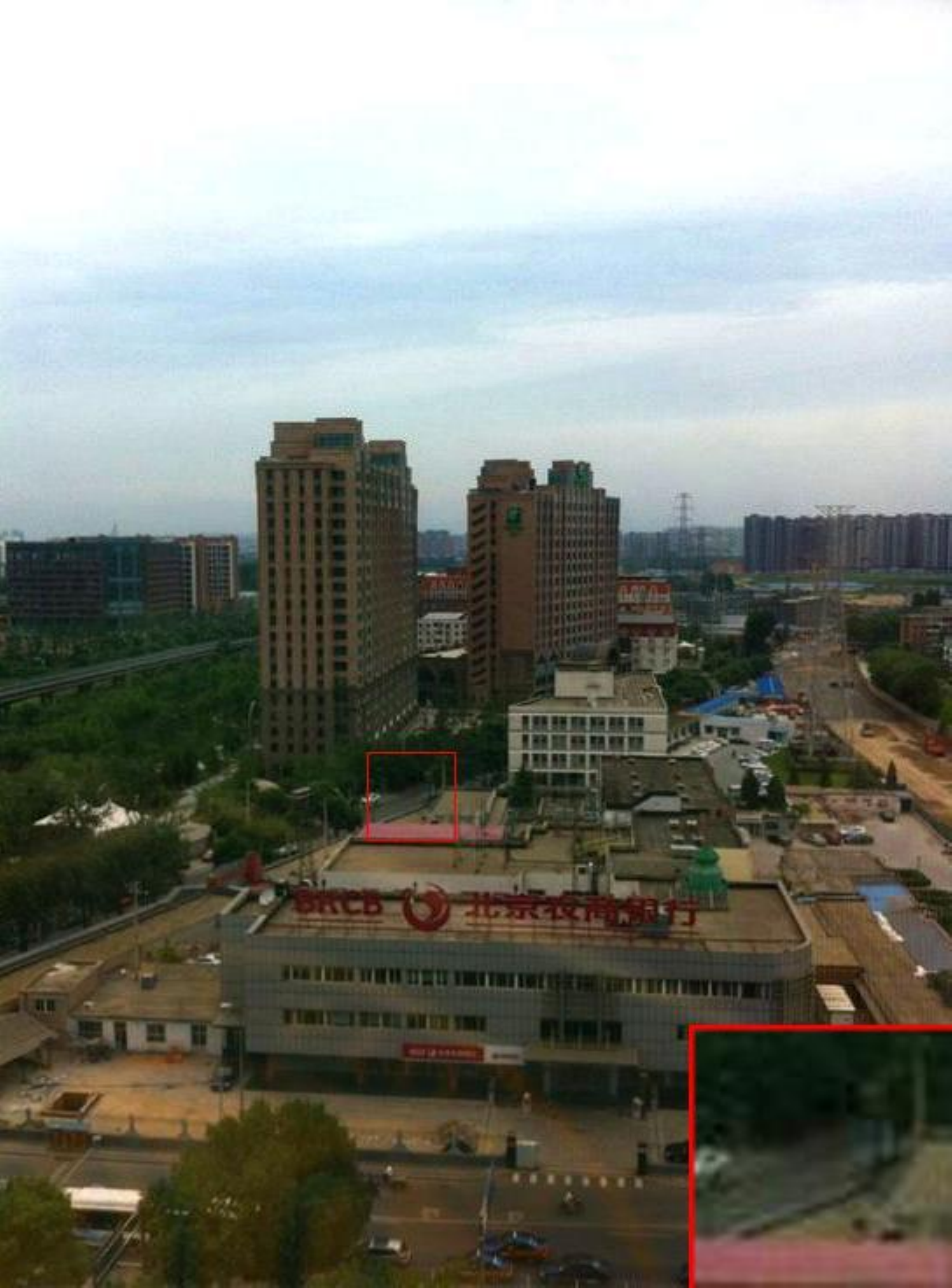}}
		\subfigure{\includegraphics[width=\m_width\textwidth, height=\a_height]{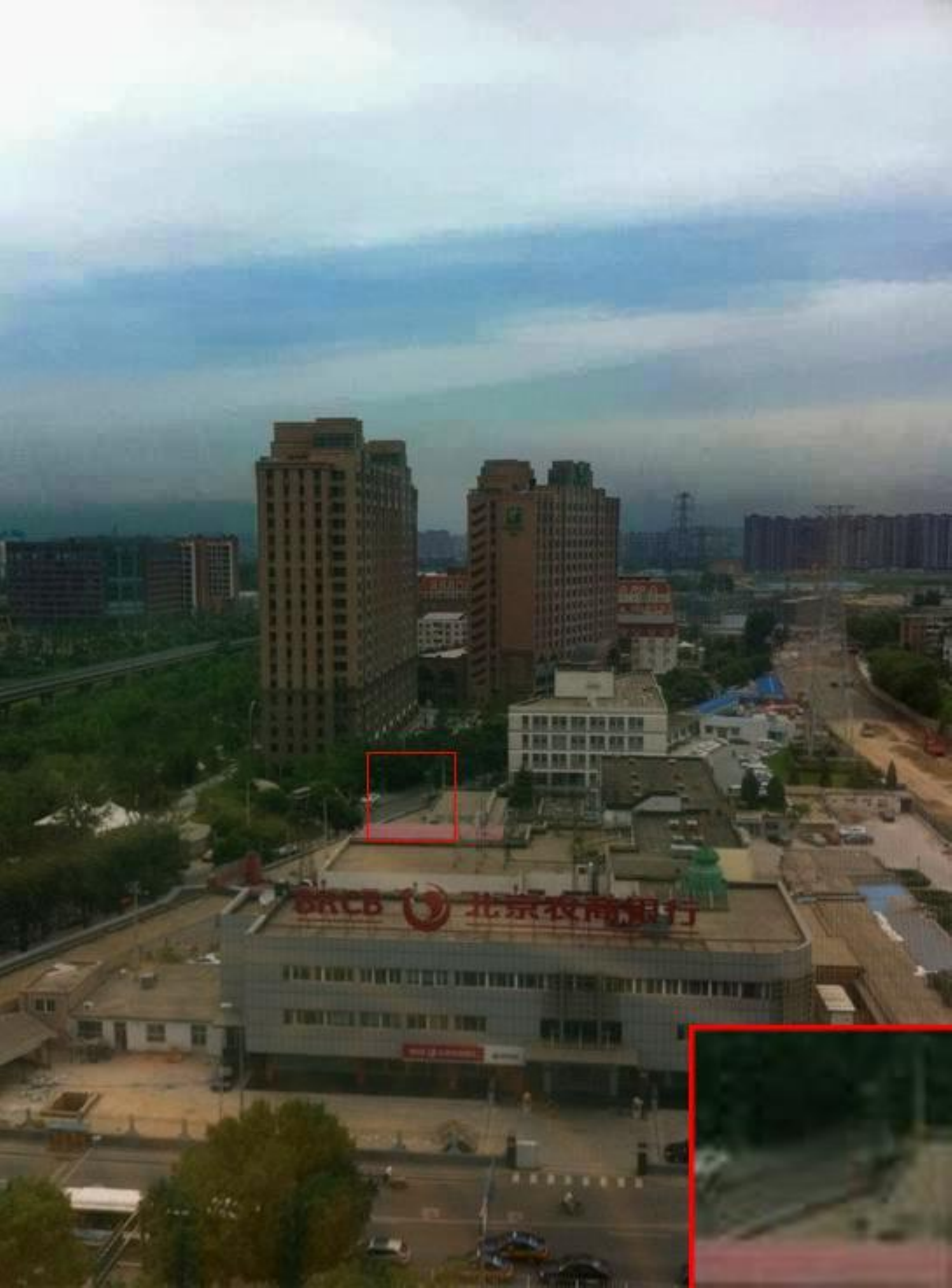}}
		\subfigure{\includegraphics[width=\m_width\textwidth, height=\a_height]{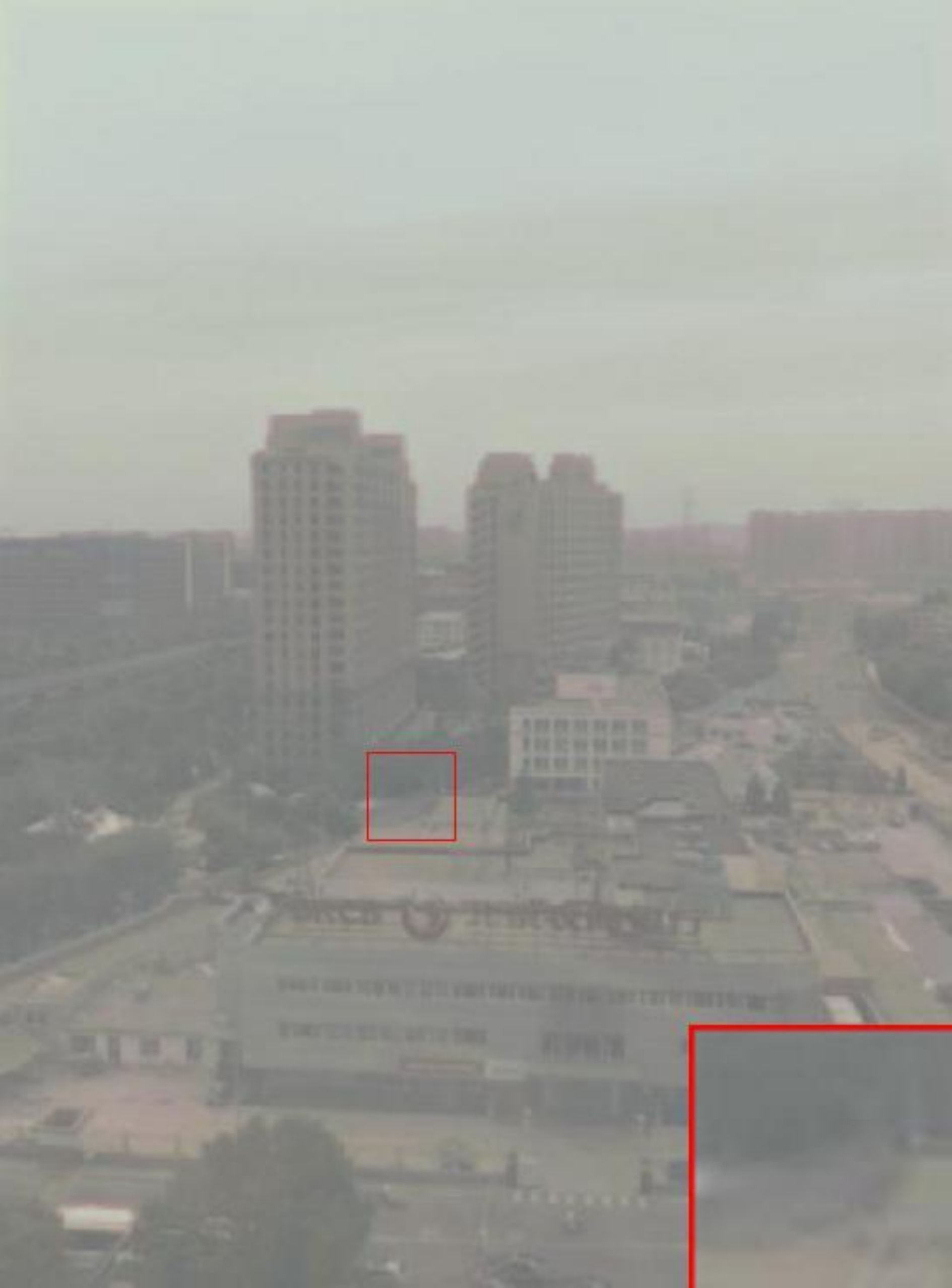}}
		\subfigure{\includegraphics[width=\m_width\textwidth, height=\a_height]{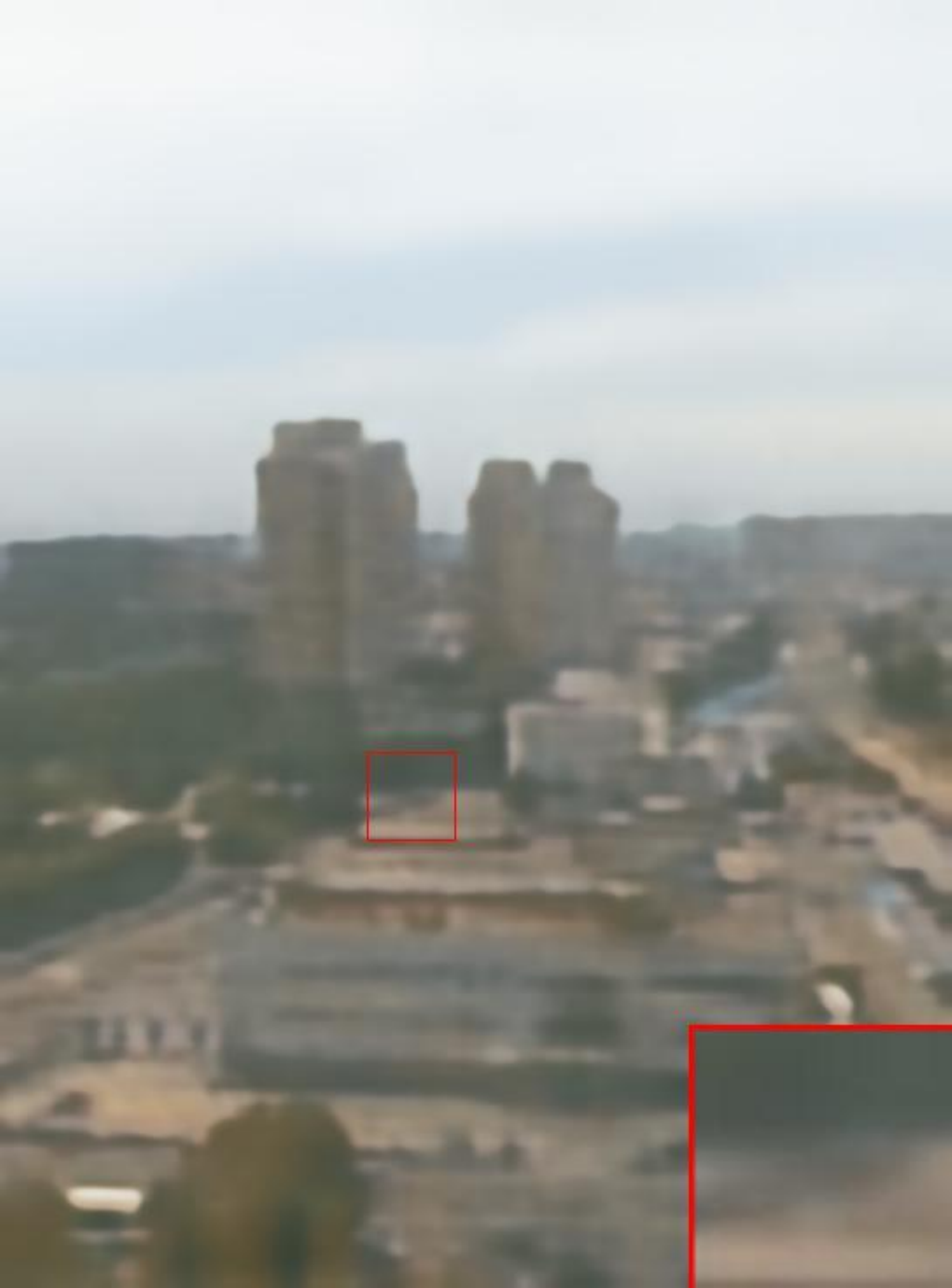}}
		\subfigure{\includegraphics[width=\m_width\textwidth, height=\a_height]{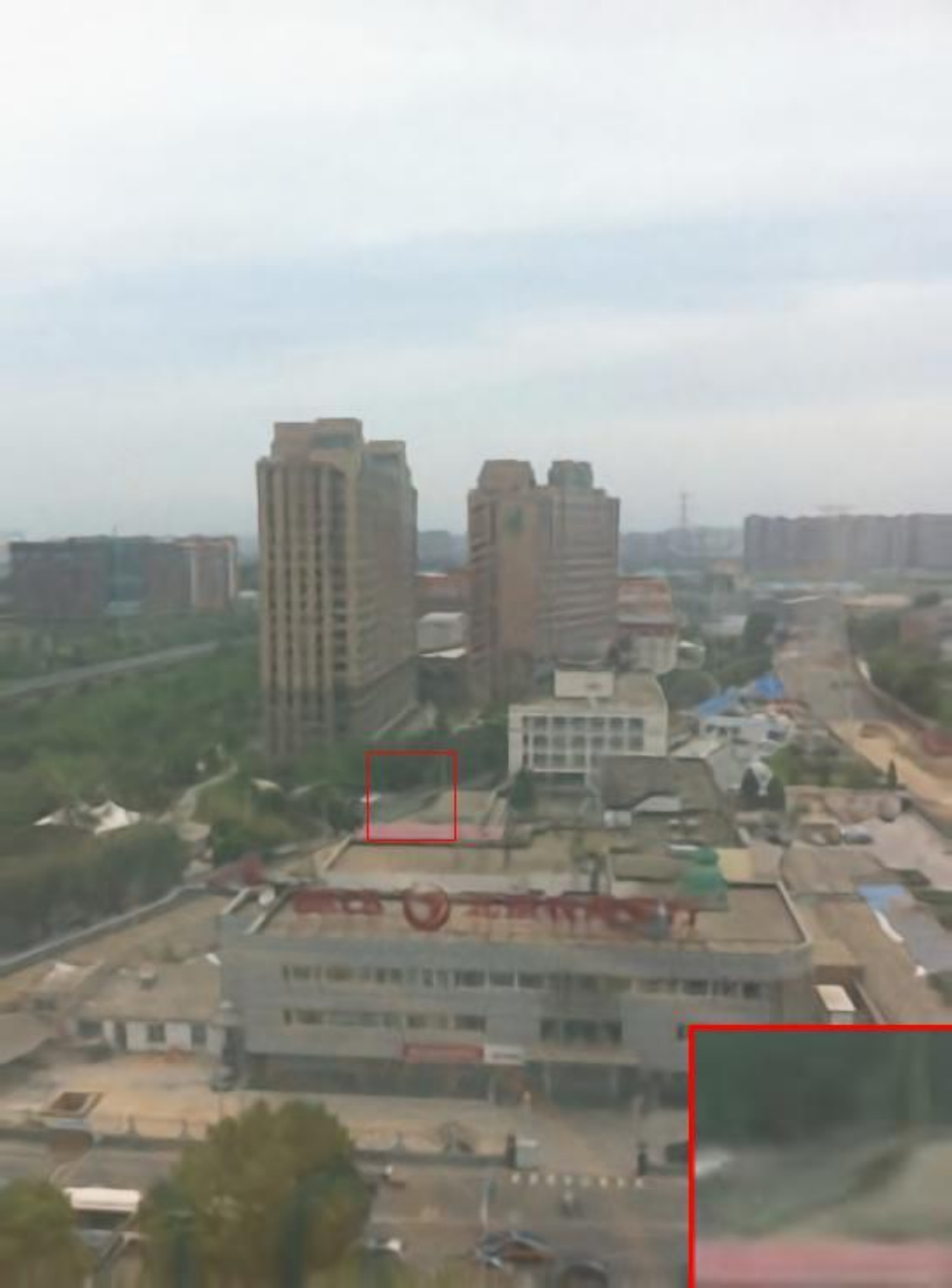}}
		\subfigure{\includegraphics[width=\m_width\textwidth, height=\a_height]{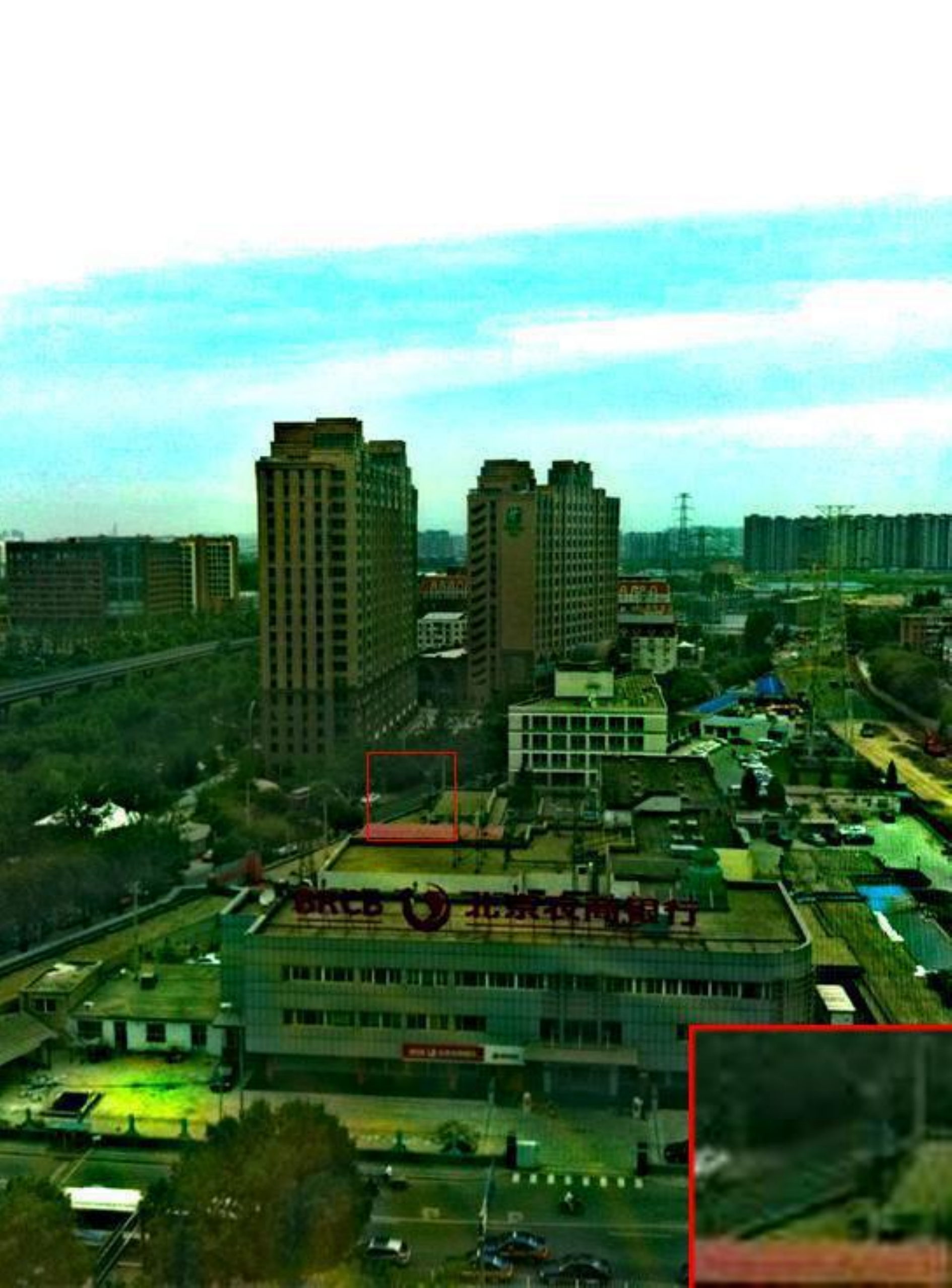}}
		\subfigure{\includegraphics[width=\m_width\textwidth, height=\a_height]{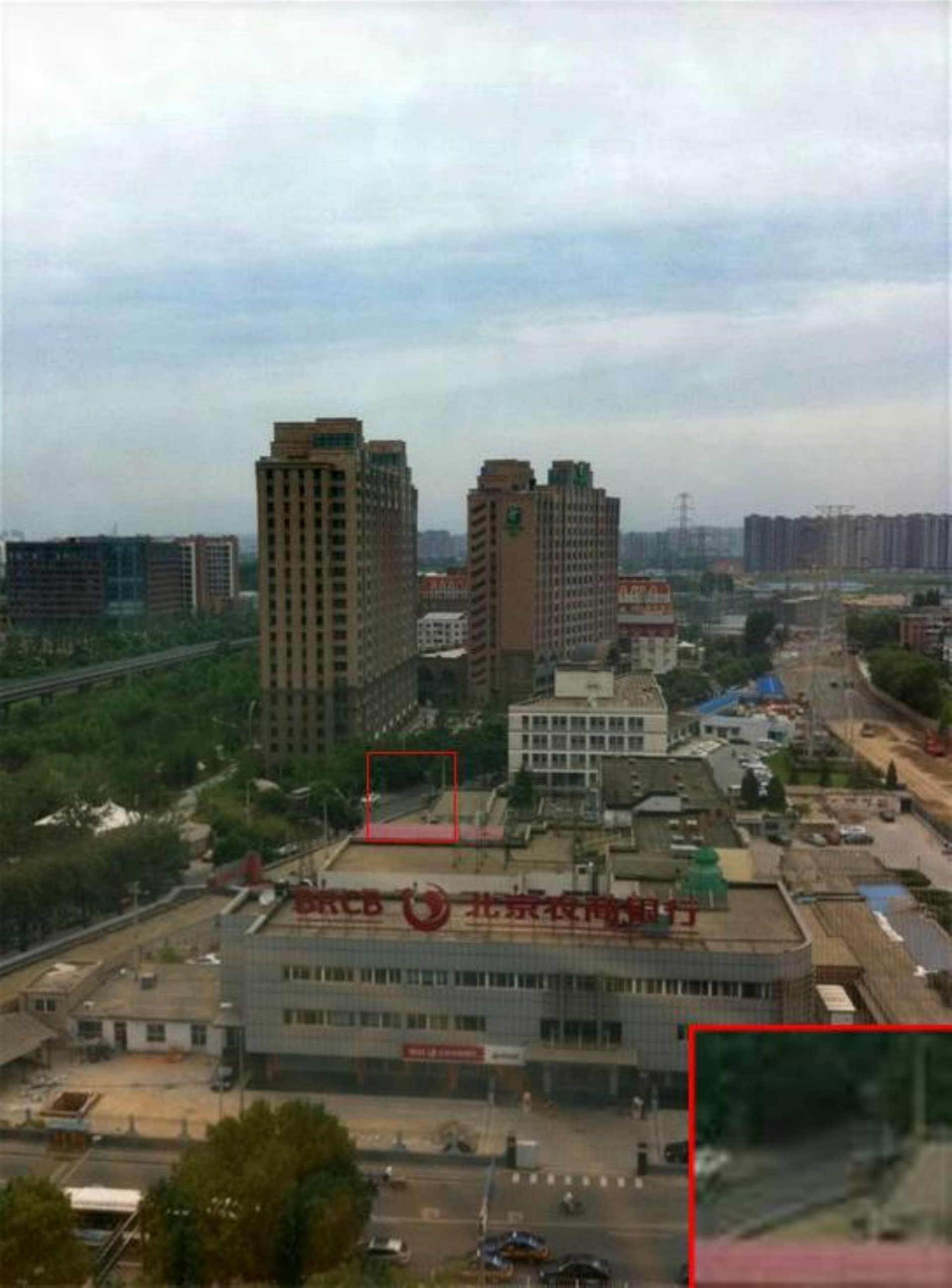}}
		\subfigure{\includegraphics[width=\m_width\textwidth,height=\a_height]{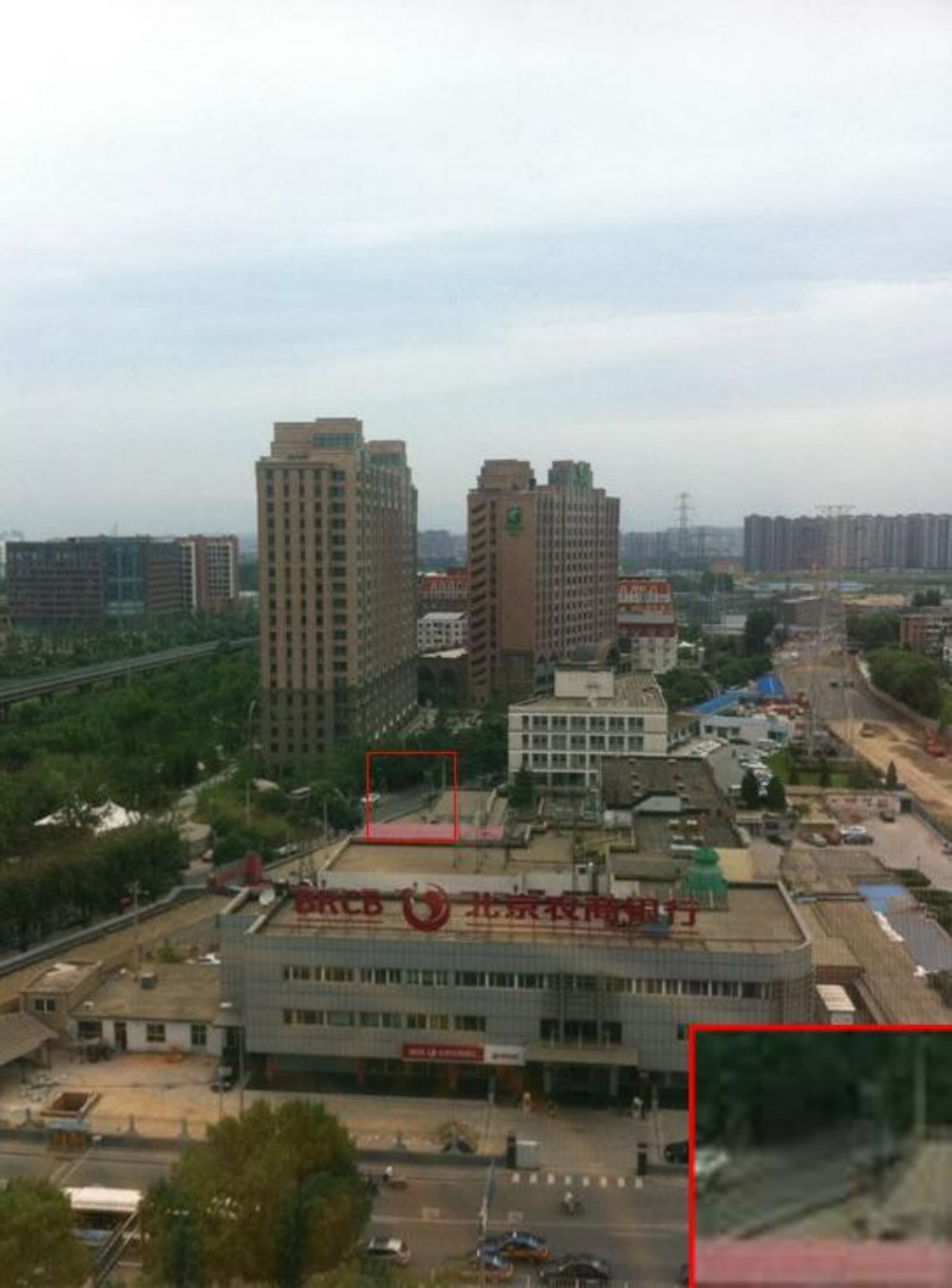}}
	\end{center}

	\vspace{-0.7cm}
	\begin{center}
		\subfigure{\includegraphics[width=\m_width\textwidth, height=\b_height]{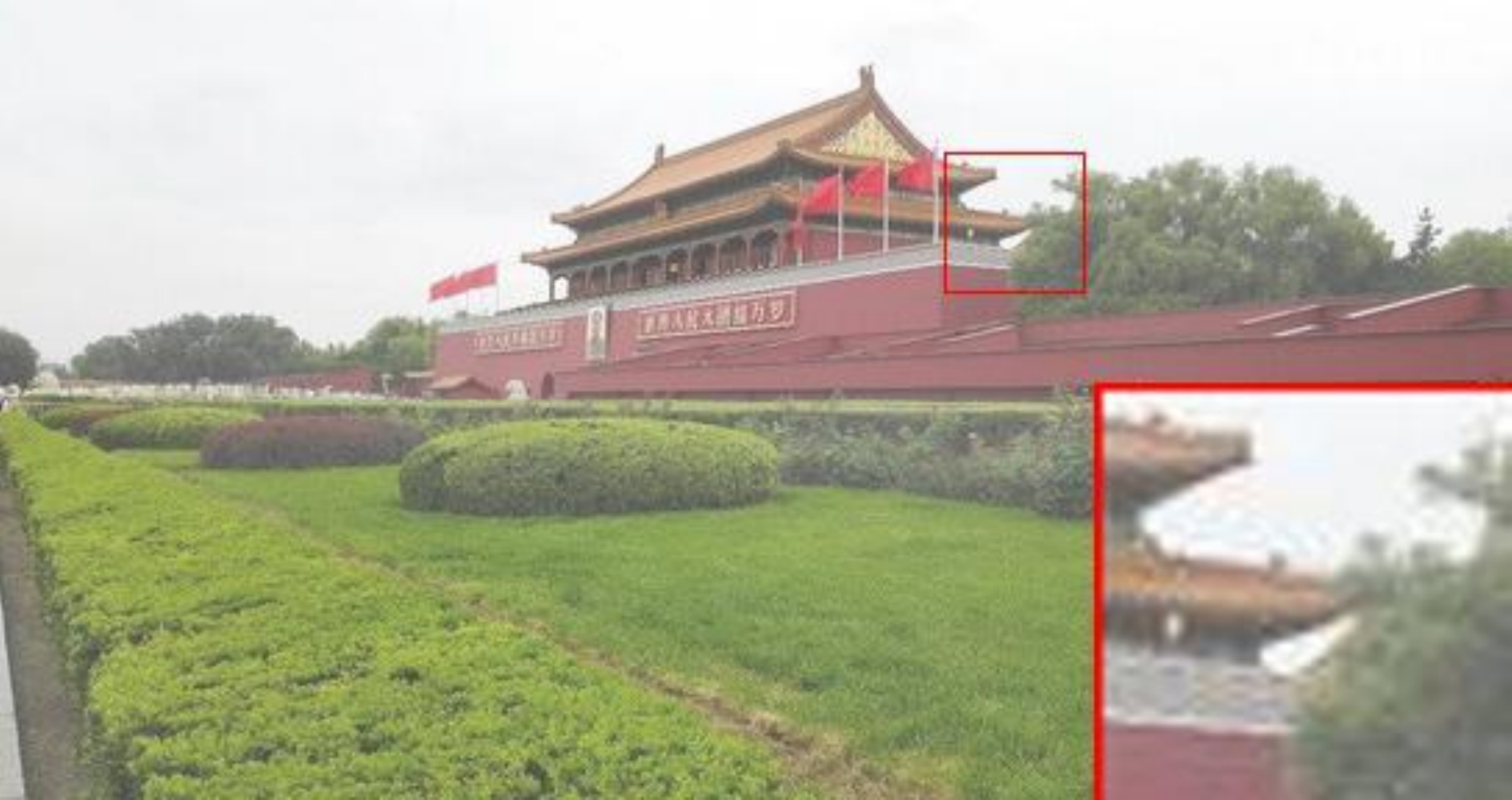}}
		\subfigure{\includegraphics[width=\m_width\textwidth, height=\b_height]{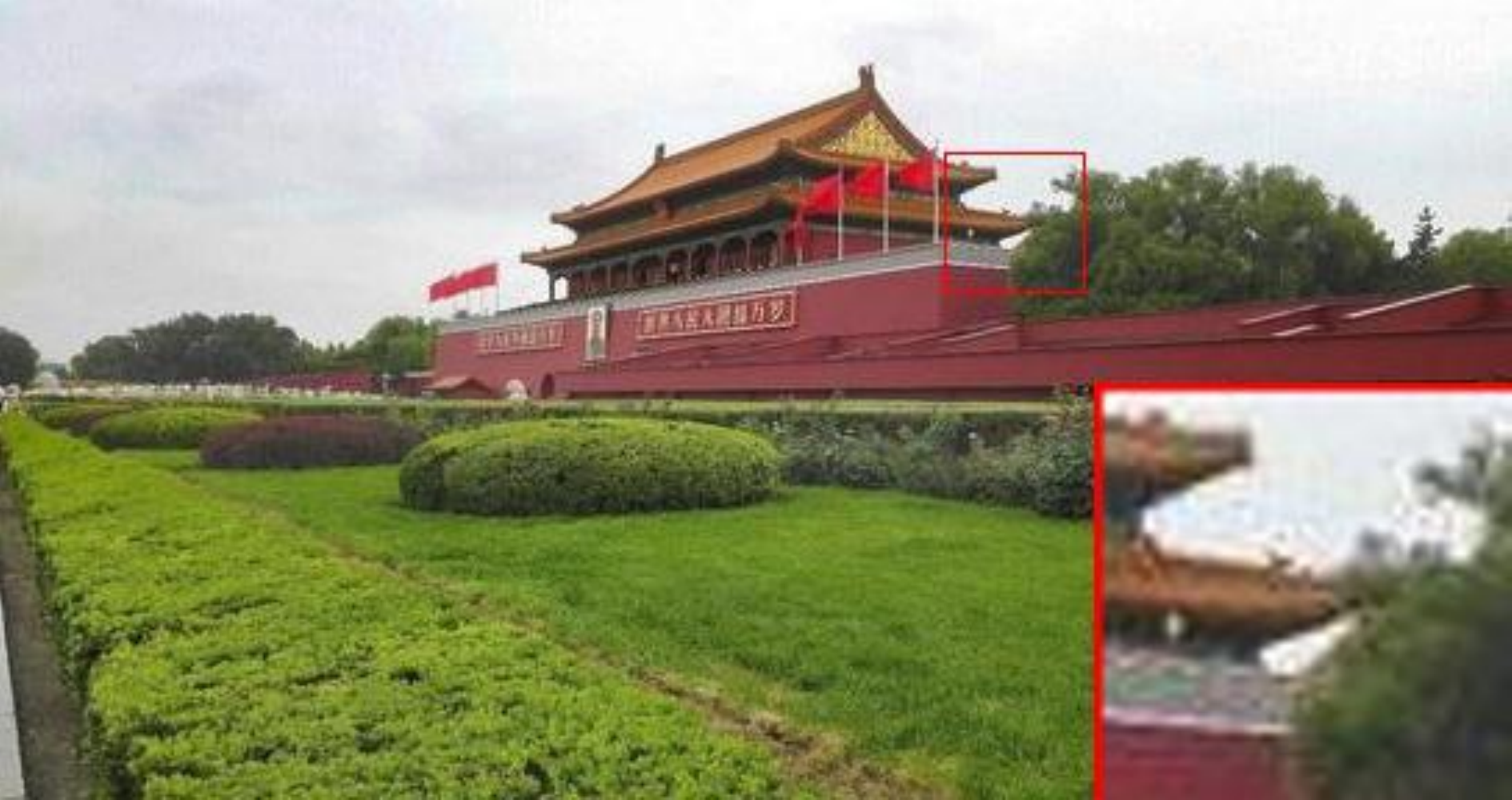}}
		\subfigure{\includegraphics[width=\m_width\textwidth, height=\b_height]{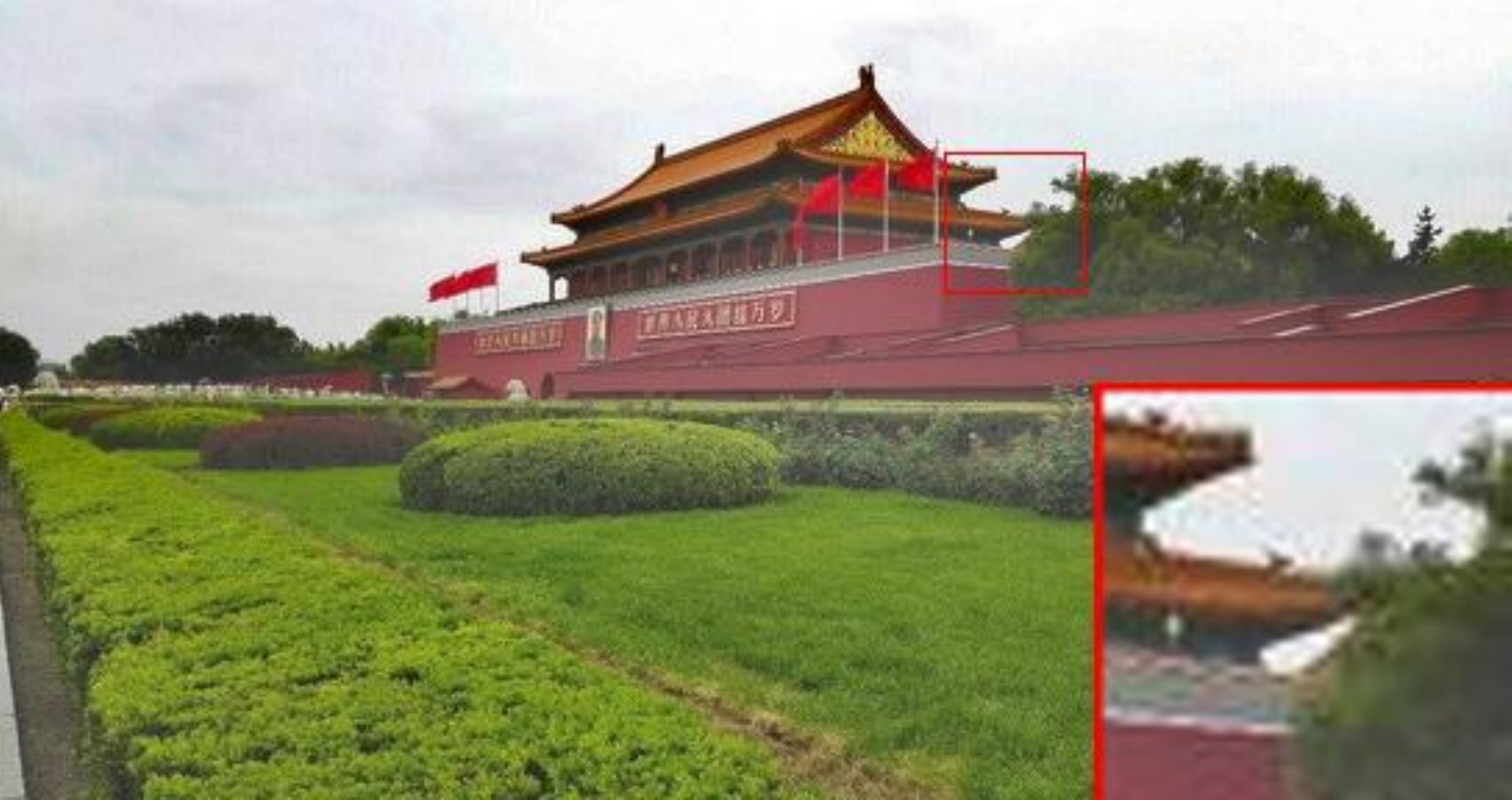}}
		\subfigure{\includegraphics[width=\m_width\textwidth, height=\b_height]{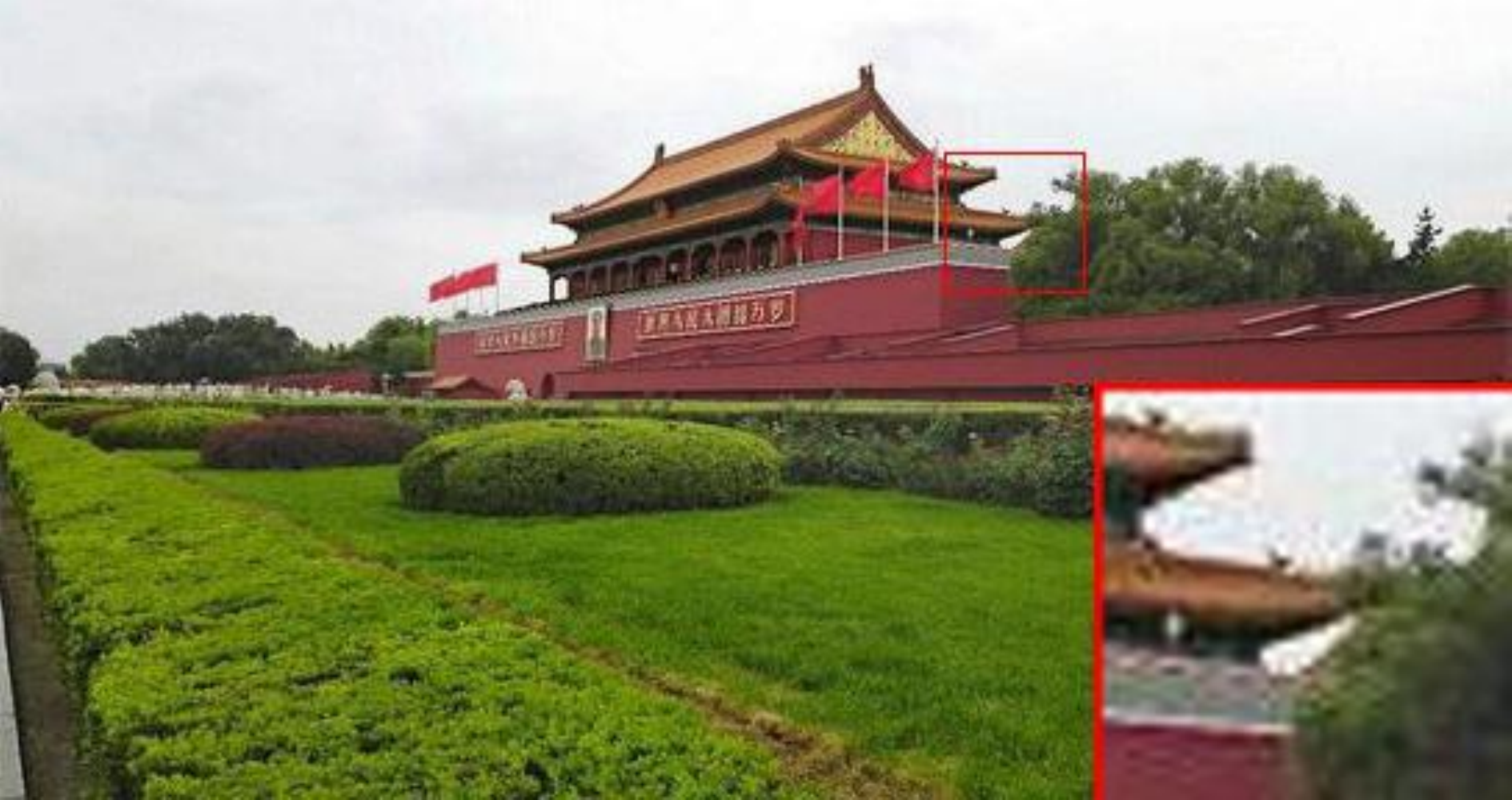}}
		\subfigure{\includegraphics[width=\m_width\textwidth, height=\b_height]{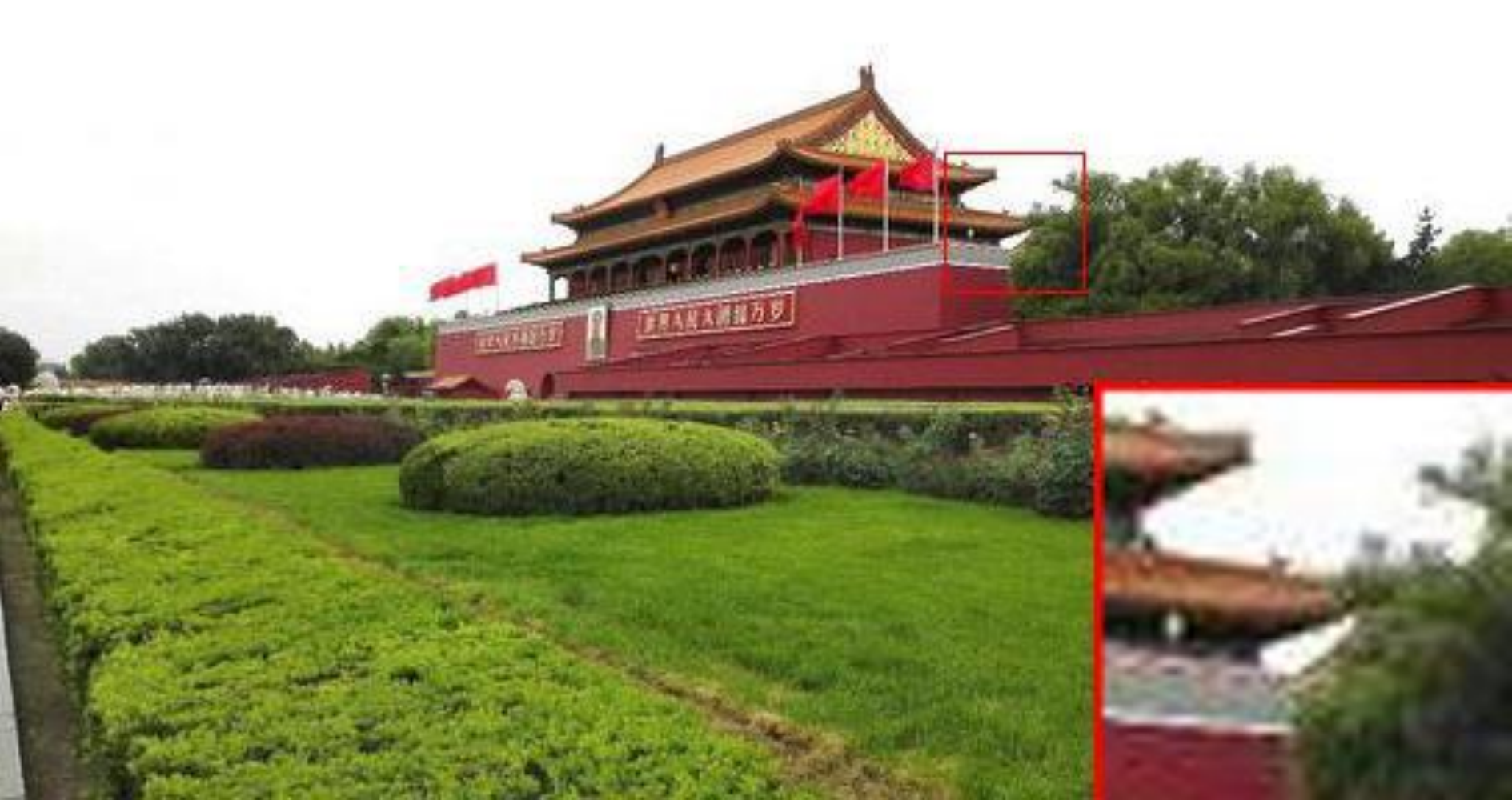}}	
		\subfigure{\includegraphics[width=\m_width\textwidth, height=\b_height]{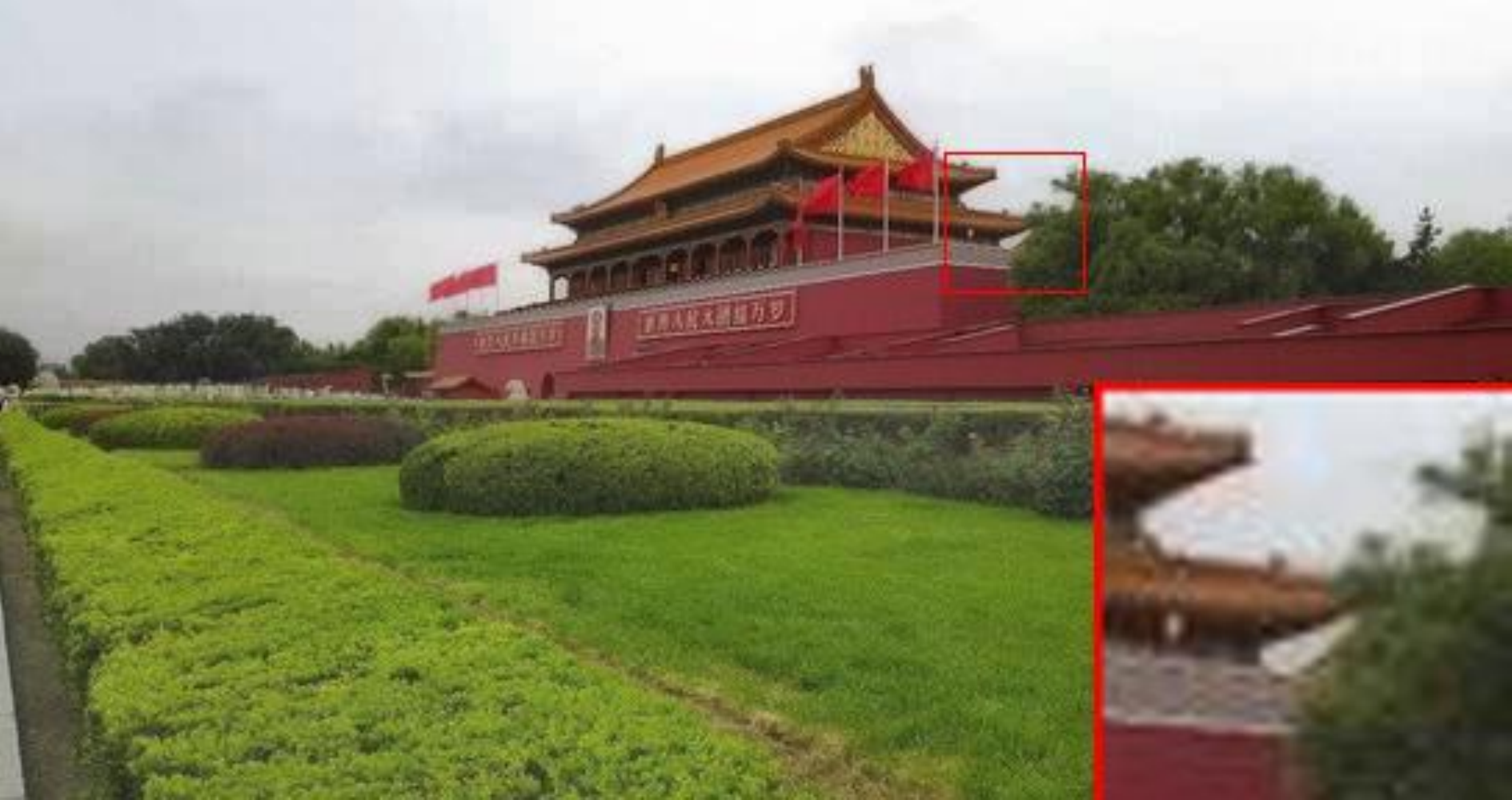}}
		\subfigure{\includegraphics[width=\m_width\textwidth, height=\b_height]{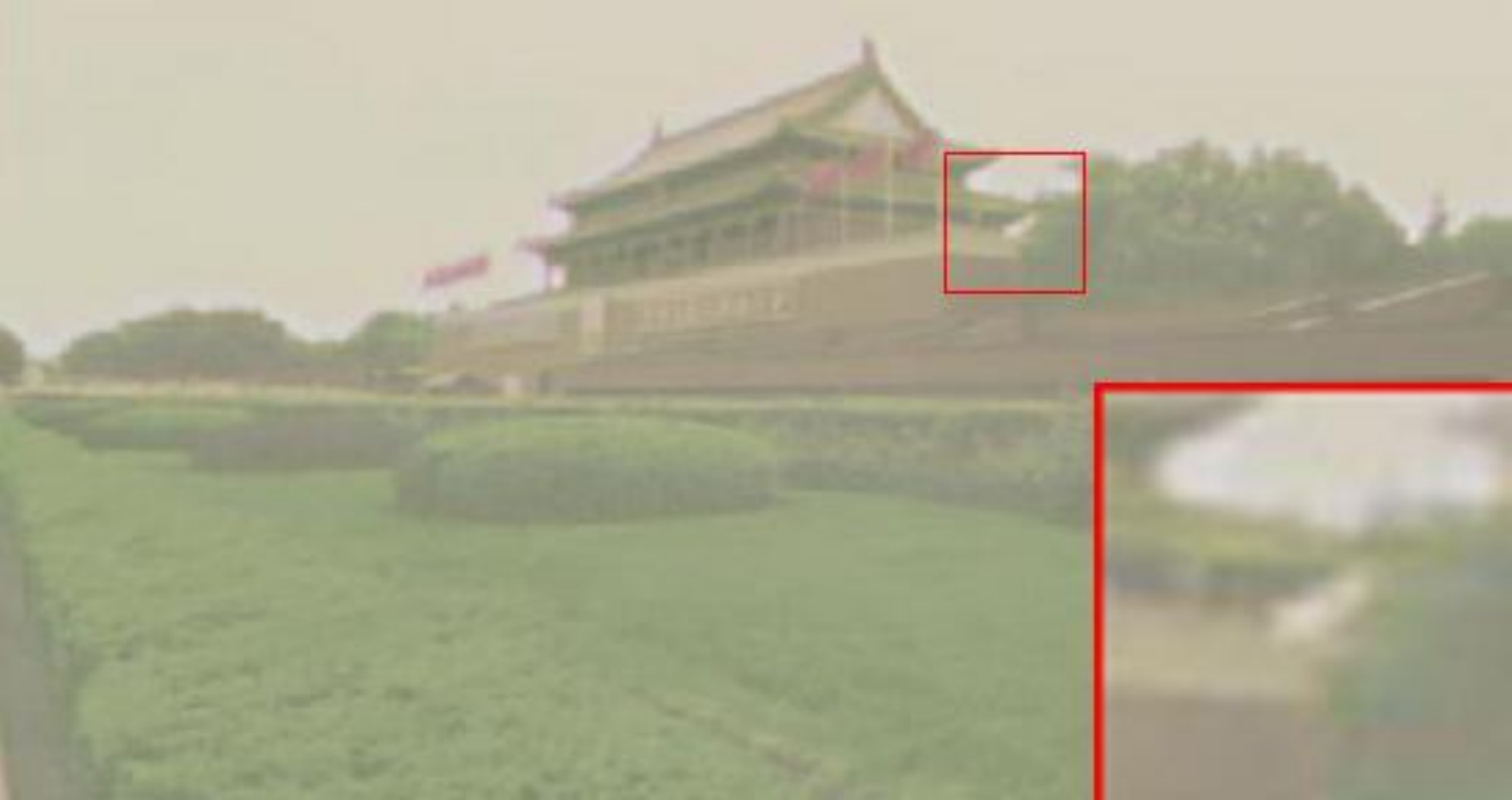}}
		\subfigure{\includegraphics[width=\m_width\textwidth, height=\b_height]{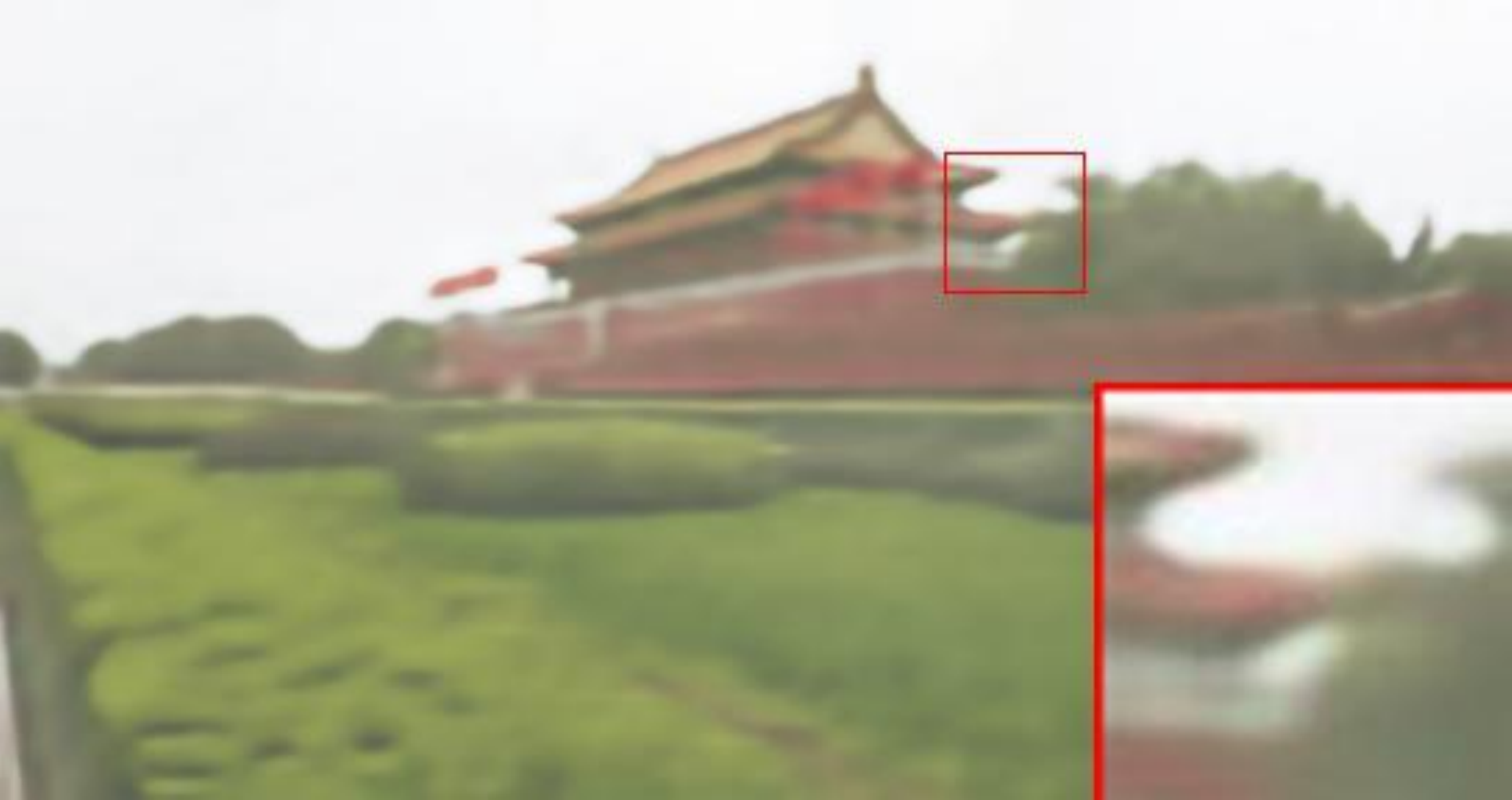}}
		\subfigure{\includegraphics[width=\m_width\textwidth, height=\b_height]{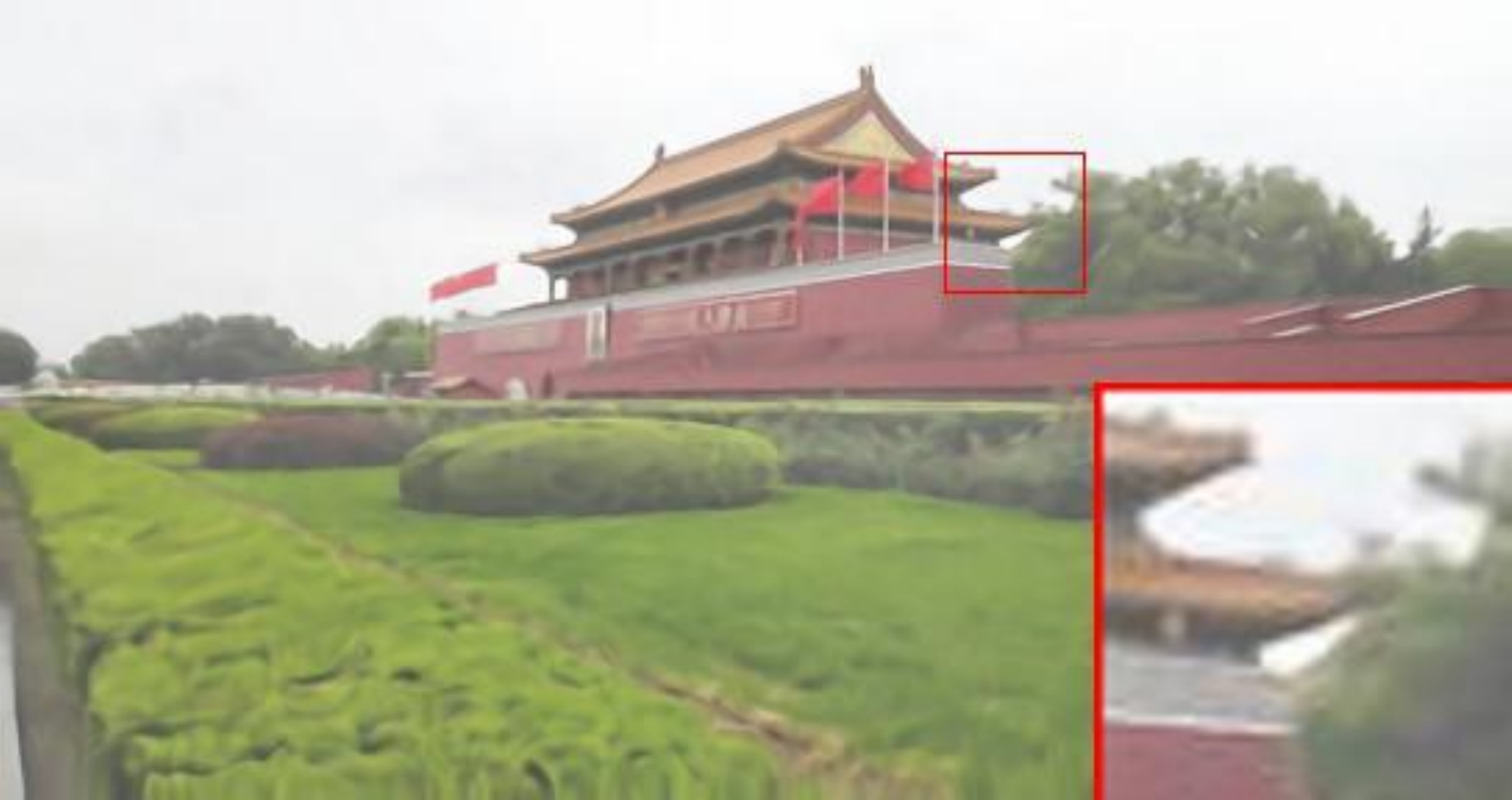}}
		\subfigure{\includegraphics[width=\m_width\textwidth, height=\b_height]{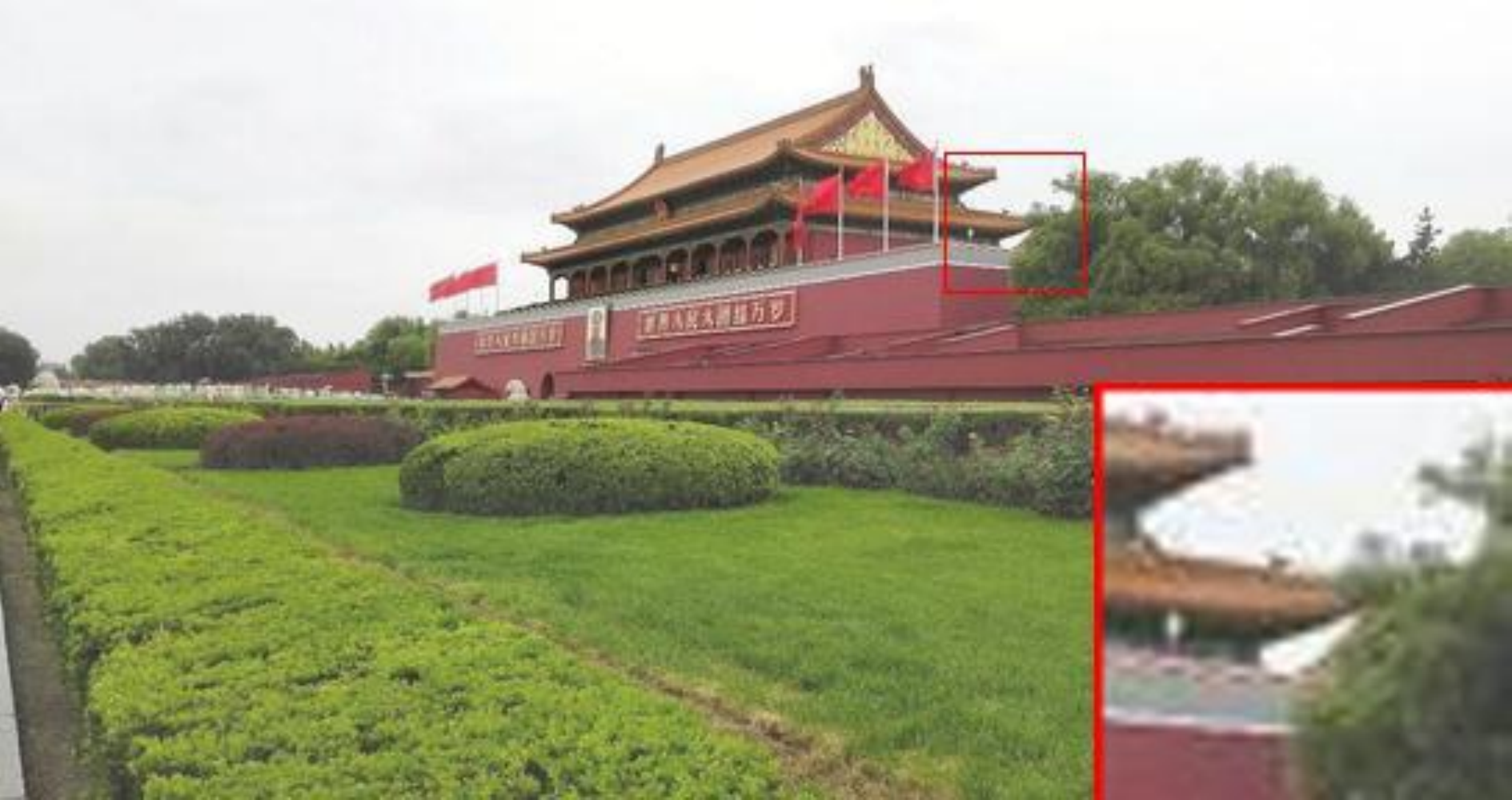}}
		\subfigure{\includegraphics[width=\m_width\textwidth, height=\b_height]{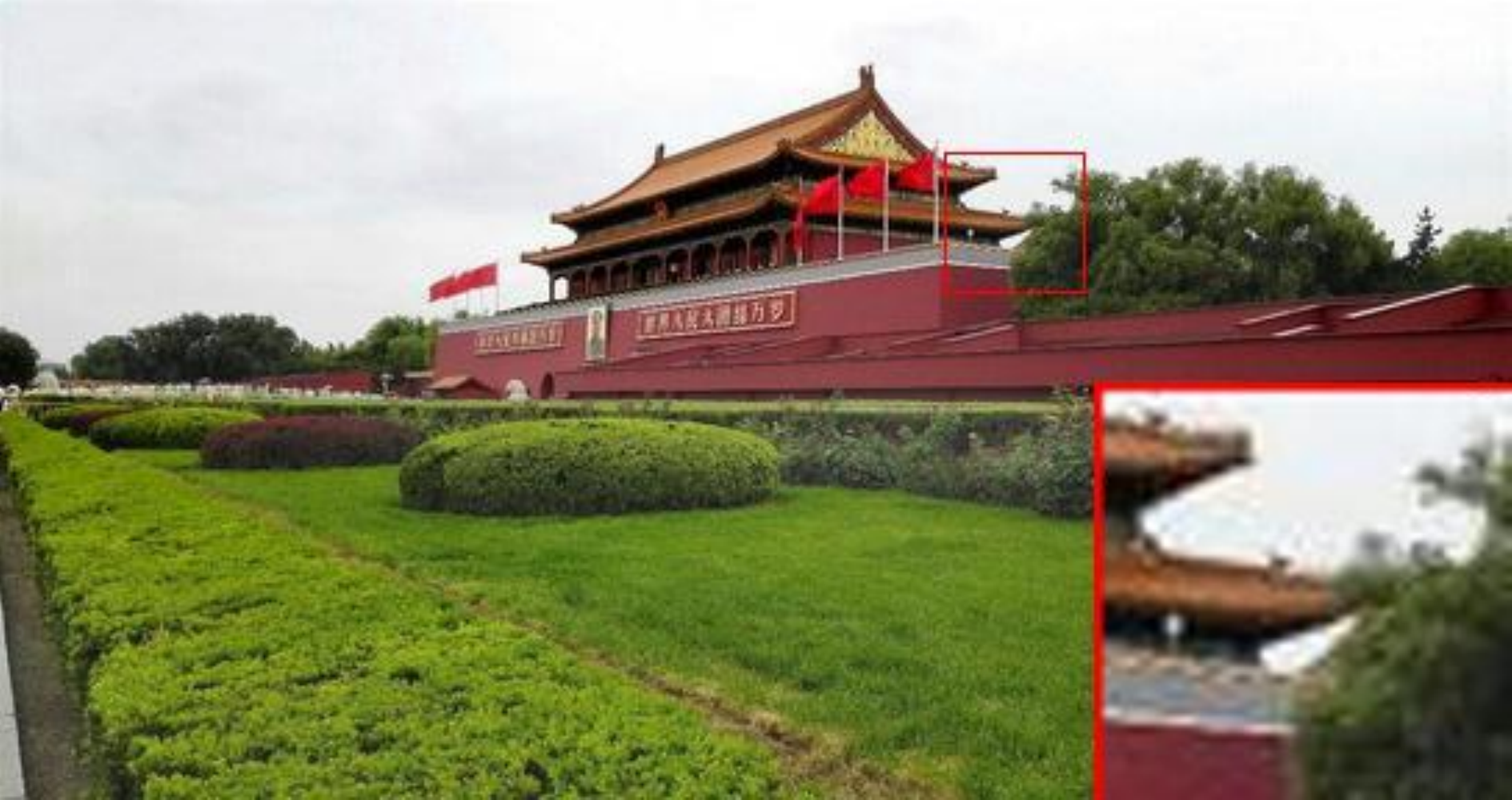}}
		\subfigure{\includegraphics[width=\m_width\textwidth,height=\b_height]{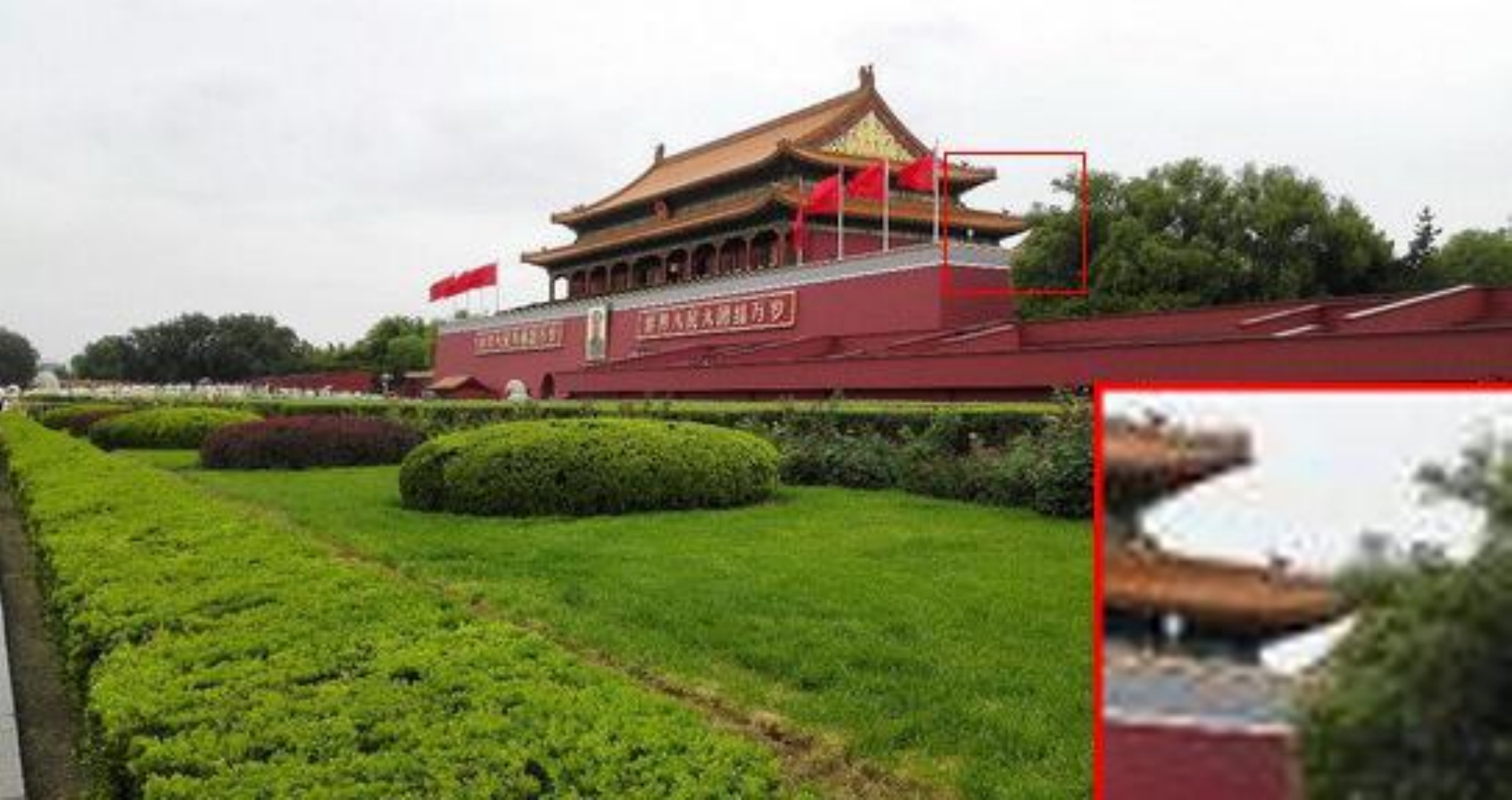}}
	\end{center}
	
	\vspace{-0.7cm}
	\setcounter{subfigure}{0}
	\begin{center}
		\subfigure[]{\label{Figure:HSTS:Input}\includegraphics[width=\m_width\textwidth, height=\b_height]{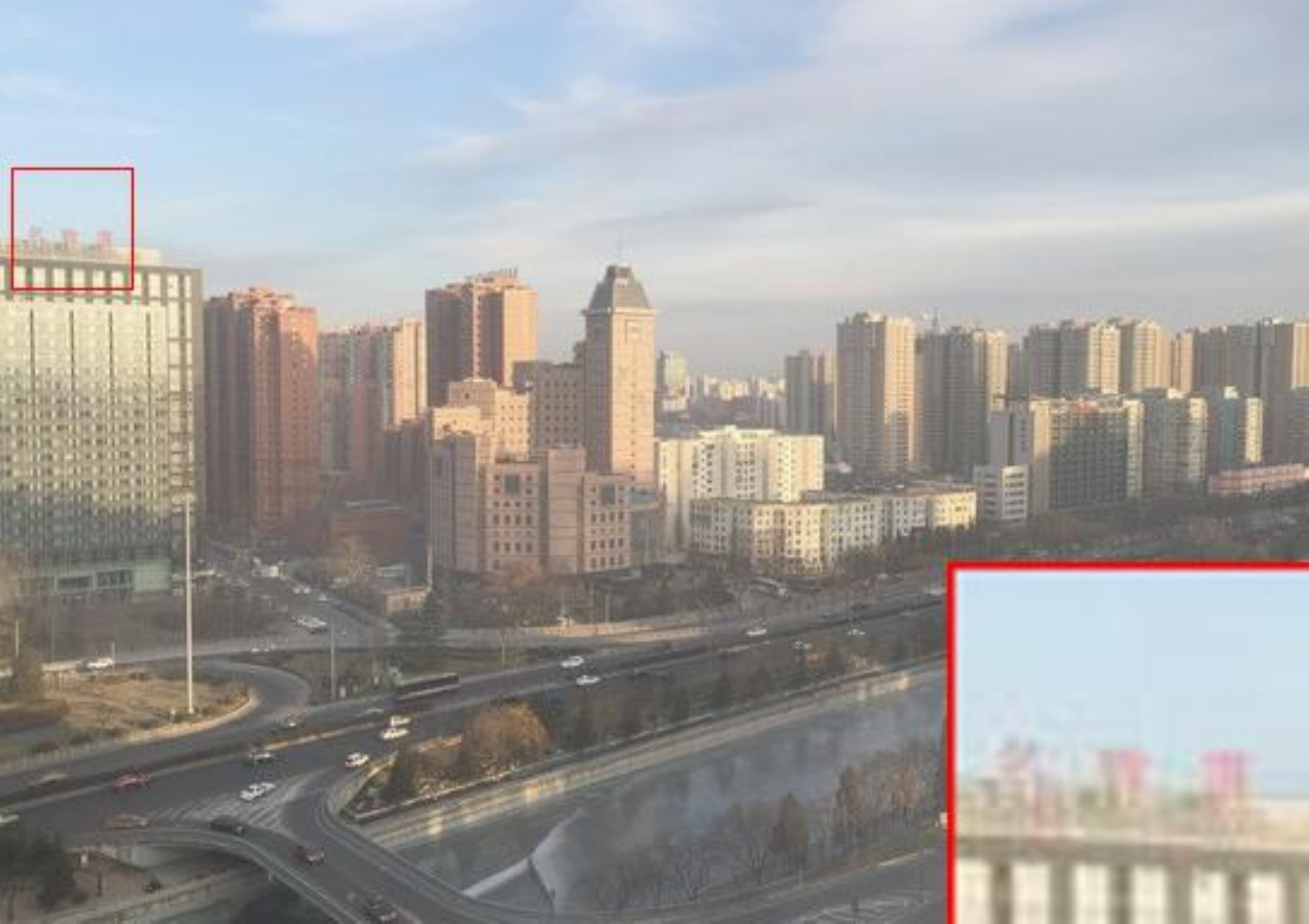}}
		\subfigure[]{\label{Figure:HSTS:DehazeNet}\includegraphics[width=\m_width\textwidth,height=\b_height]{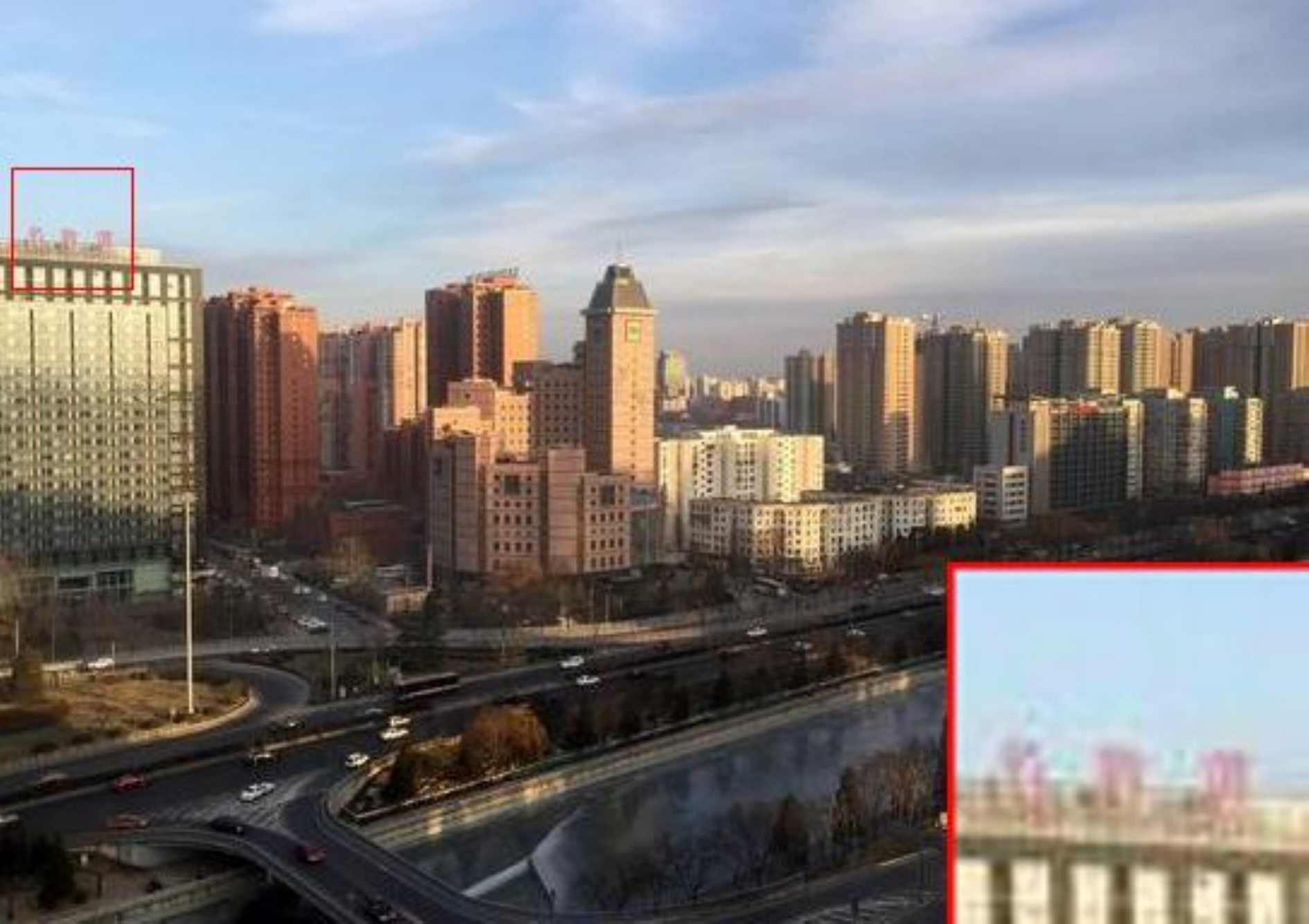}}
		\subfigure[]{\includegraphics[width=\m_width\textwidth,height=\b_height]{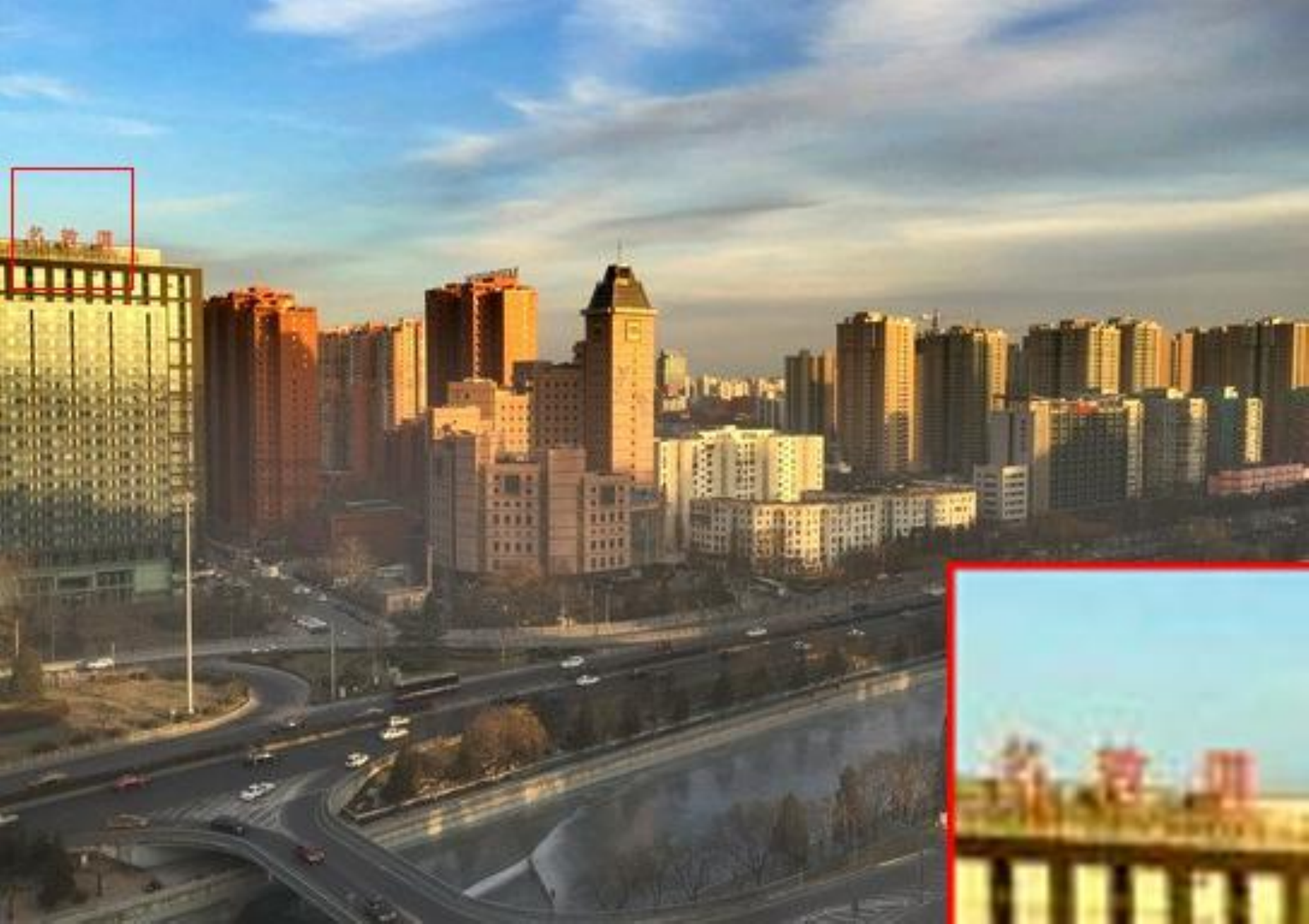}}
		\subfigure[]{\includegraphics[width=\m_width\textwidth,height=\b_height]{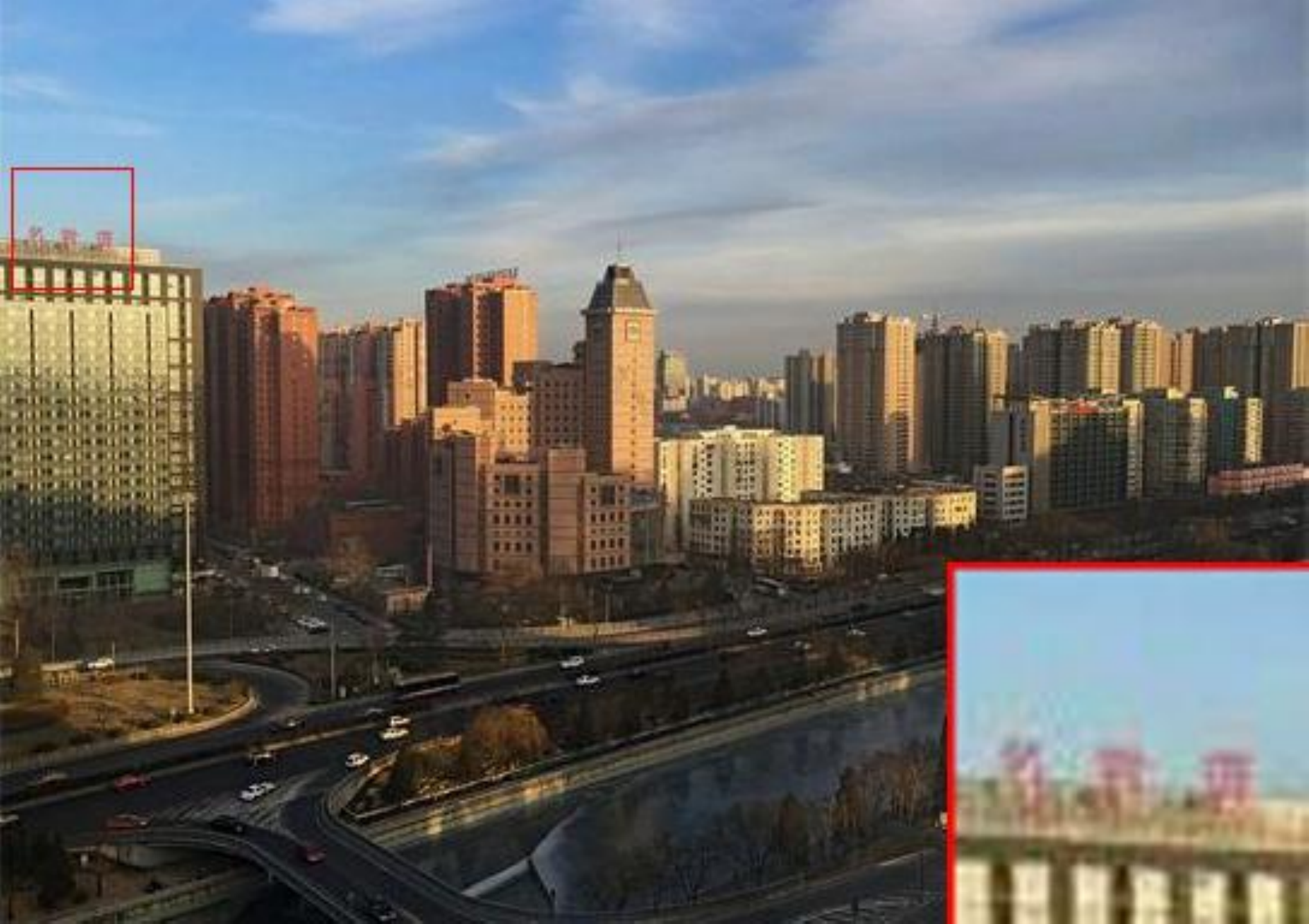}}
		\subfigure[]{\includegraphics[width=\m_width\textwidth,height=\b_height]{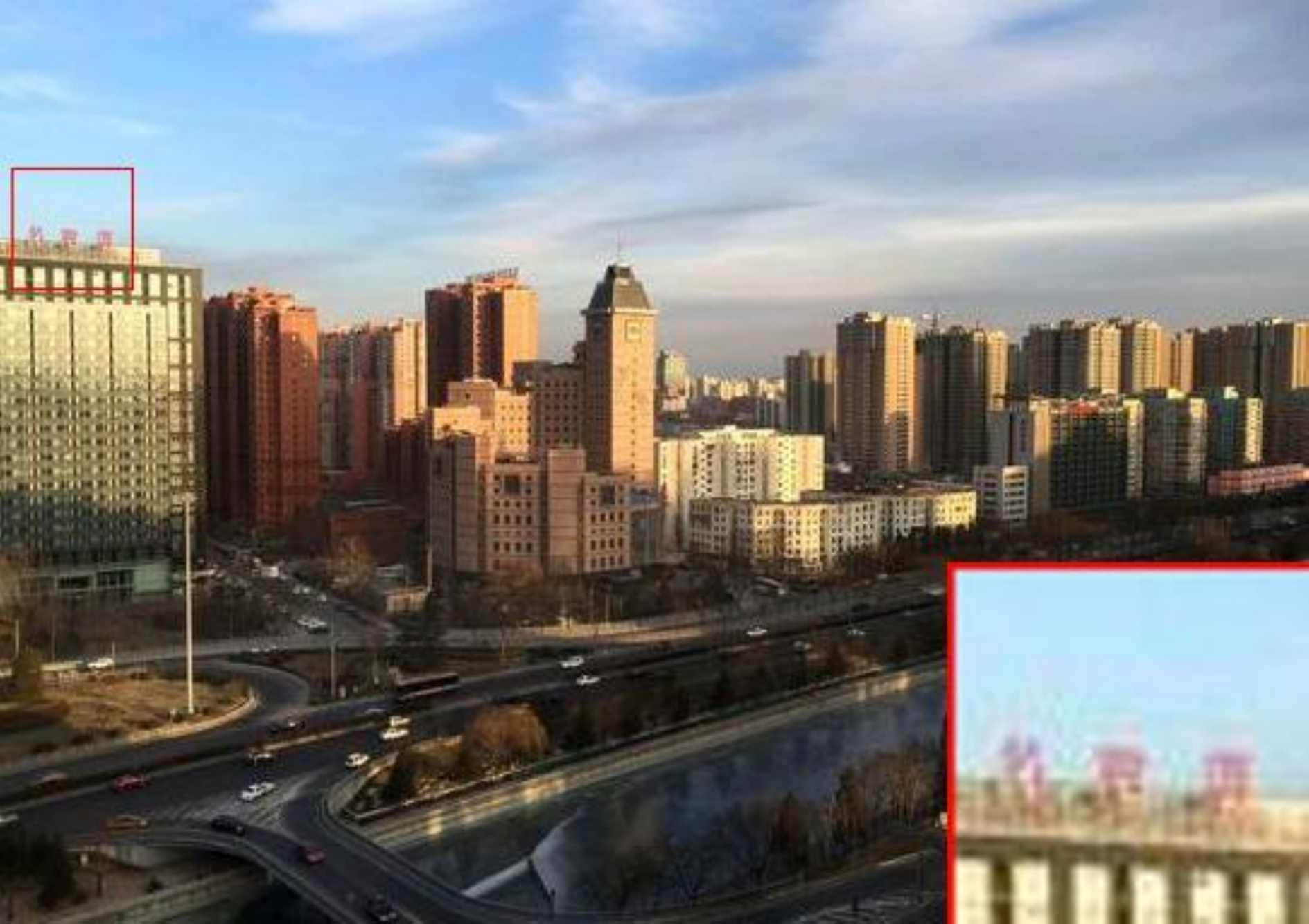}}
		\subfigure[]{\includegraphics[width=\m_width\textwidth,height=\b_height]{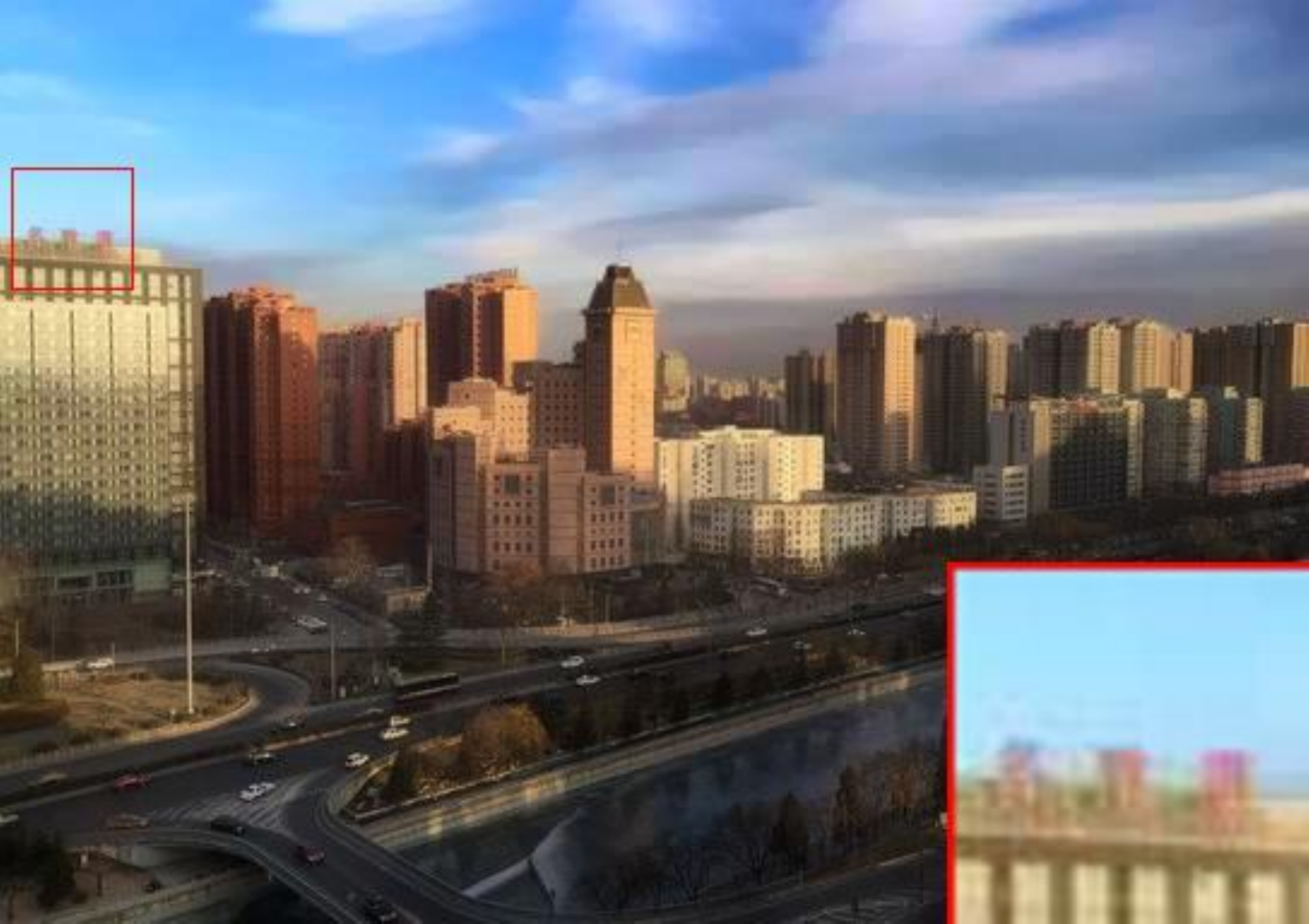}}
		\subfigure[]{\includegraphics[width=\m_width\textwidth,height=\b_height]{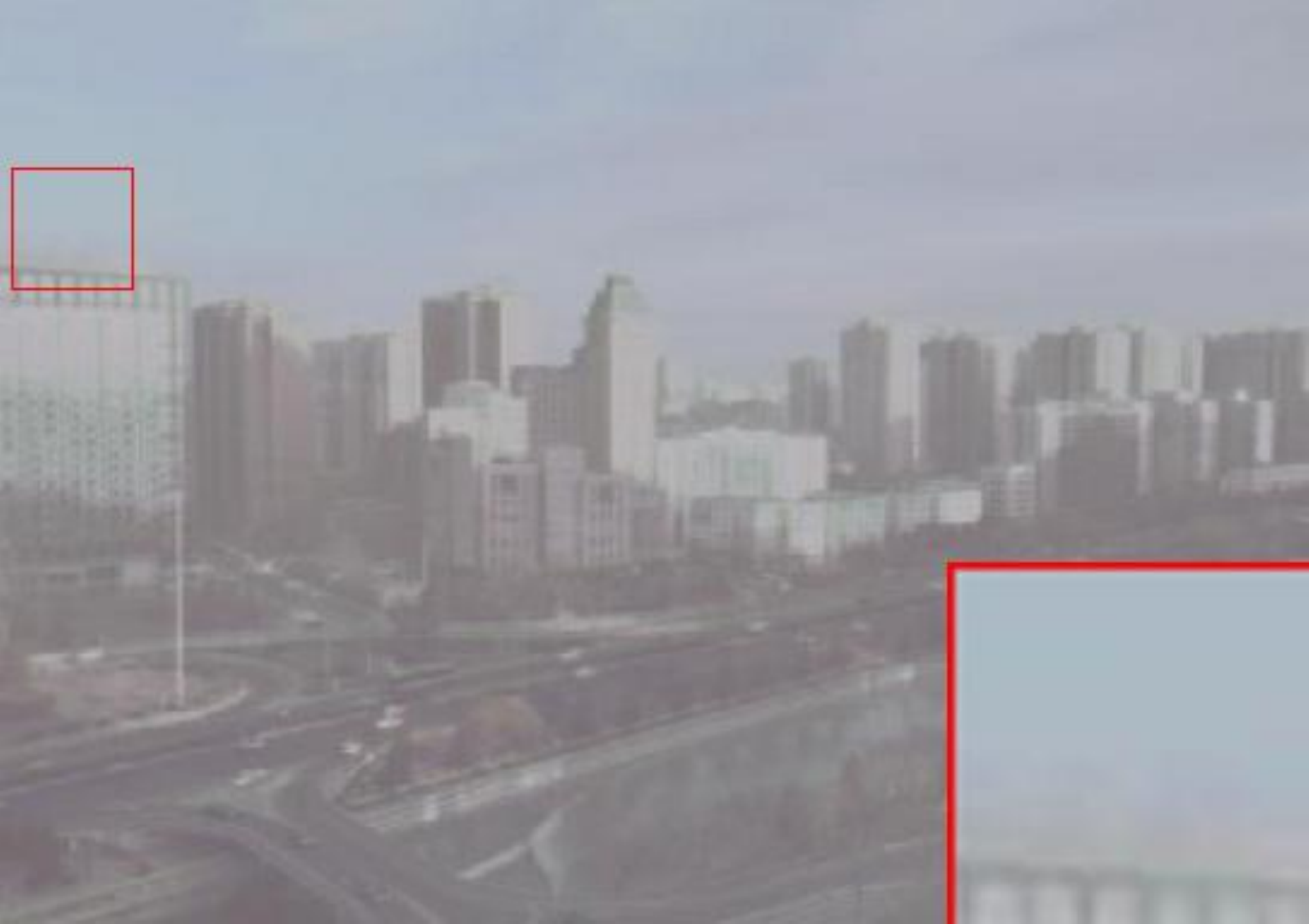}}
		\subfigure[]{\includegraphics[width=\m_width\textwidth,height=\b_height]{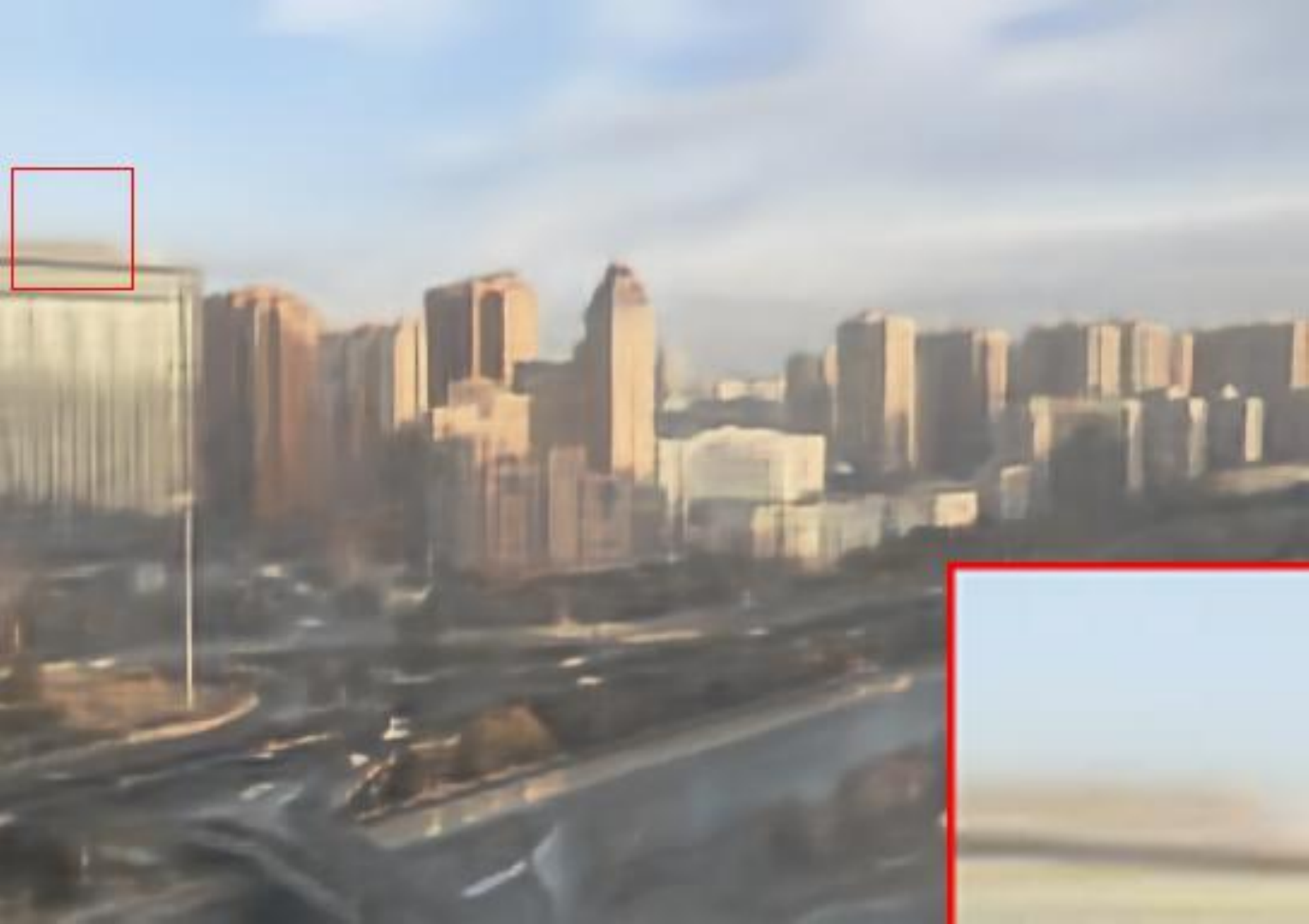}}
		\subfigure[]{\includegraphics[width=\m_width\textwidth,height=\b_height]{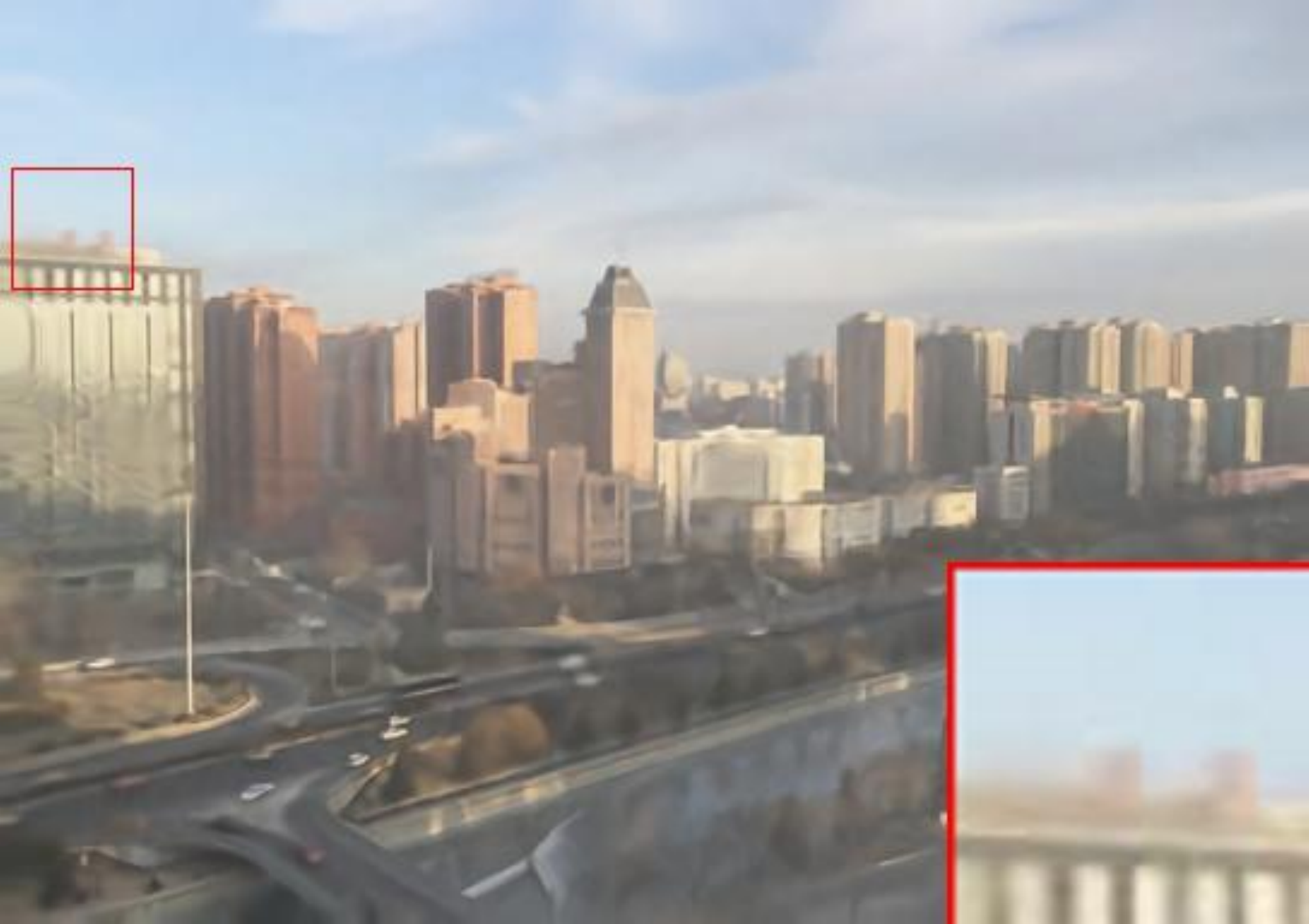}}
		\subfigure[]{\includegraphics[width=\m_width\textwidth,height=\b_height]{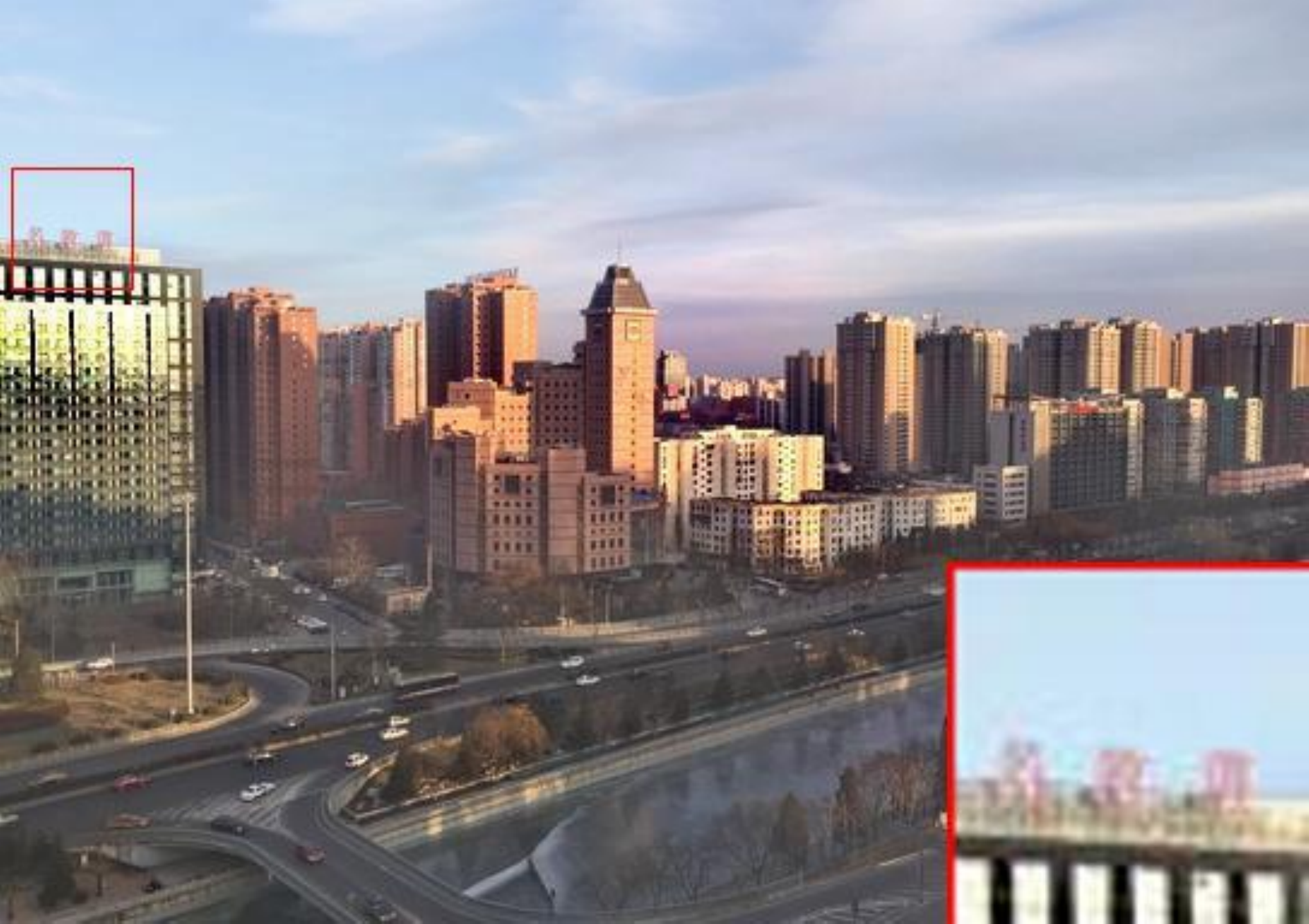}}
		\subfigure[our]{\label{Figure:HSTS:SL}\includegraphics[width=\m_width\textwidth,height=\b_height]{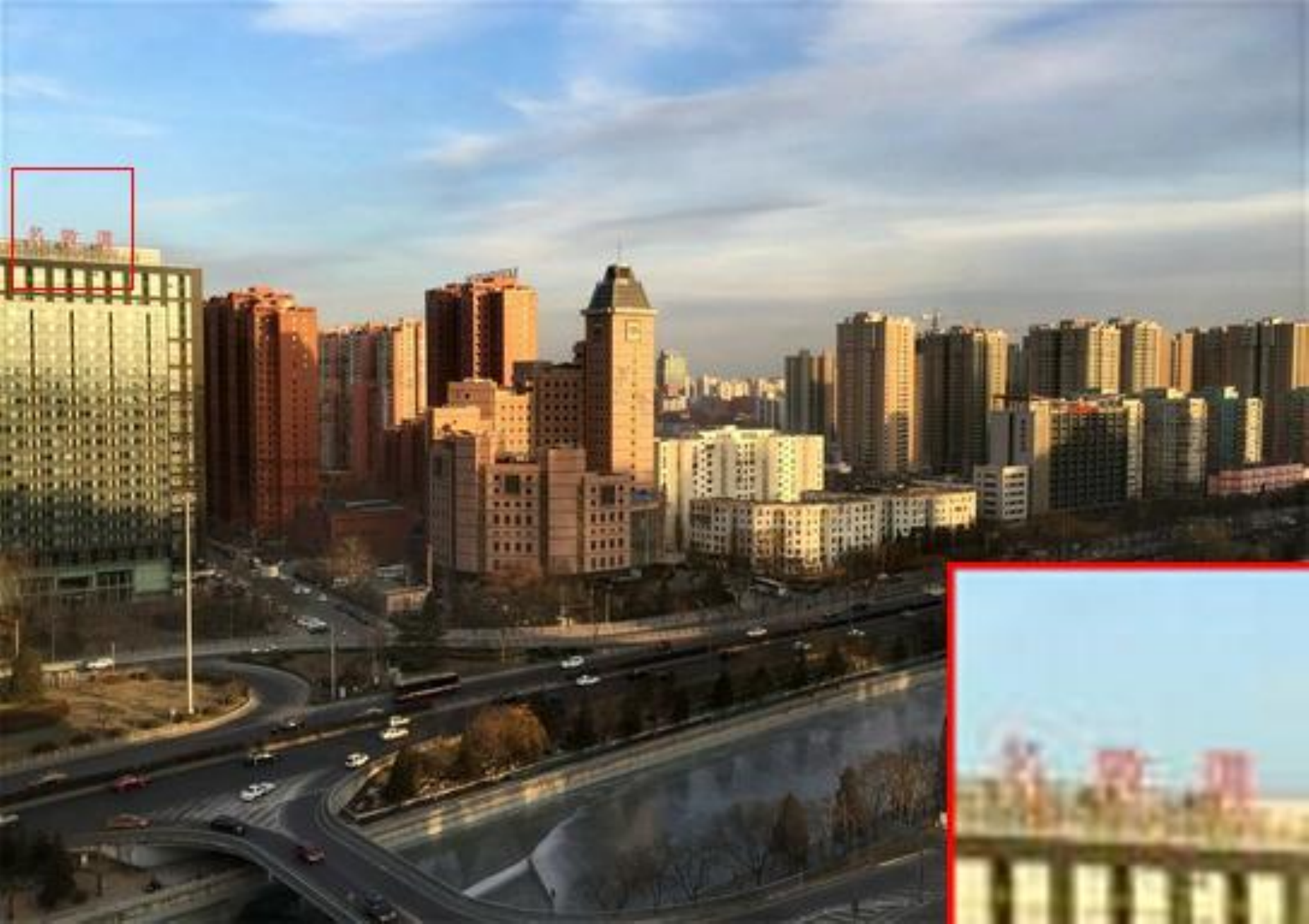}}
		\subfigure[]{\label{Figure:HSTS:GT}\includegraphics[width=\m_width\textwidth,height=\b_height]{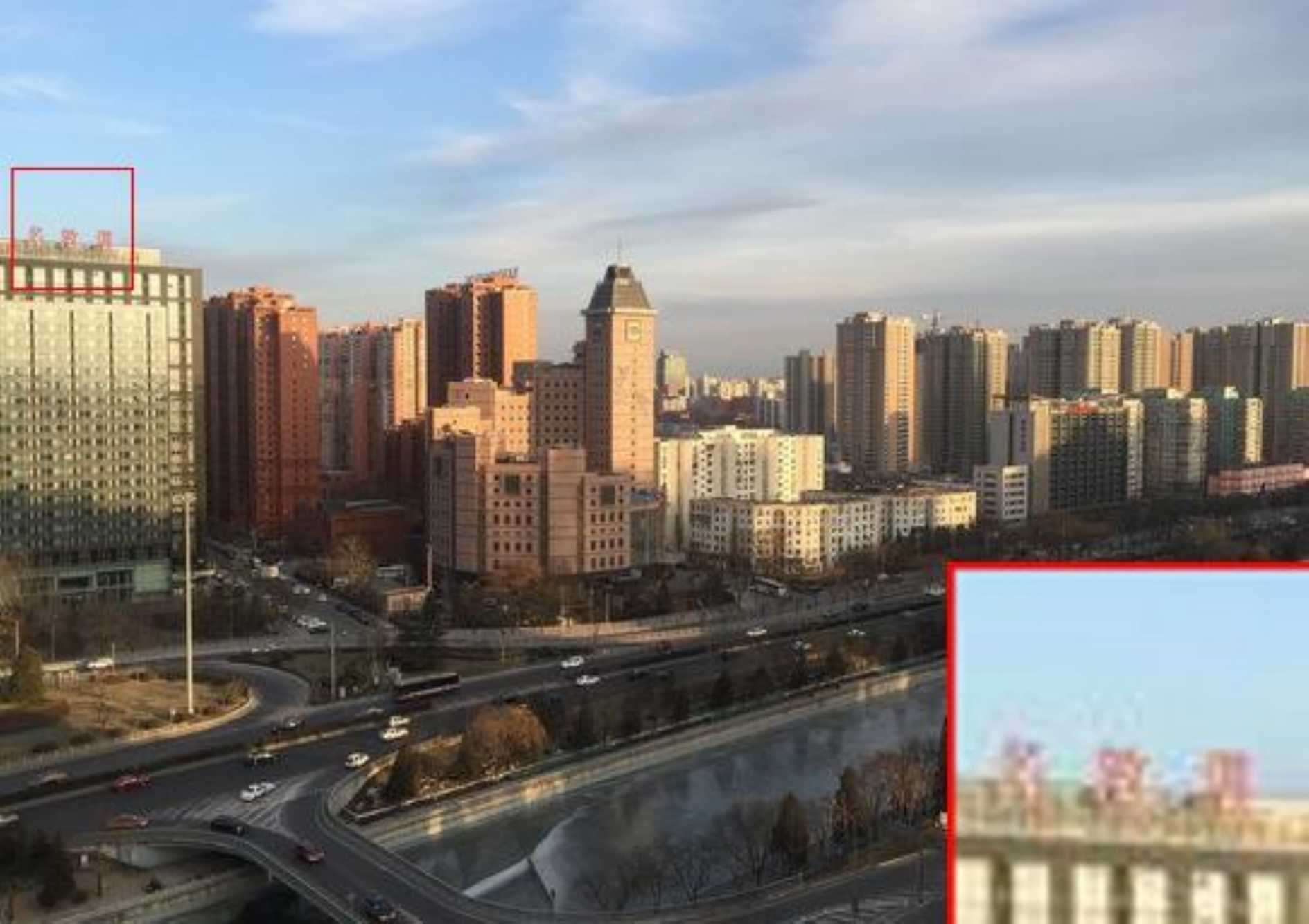}}
	\end{center}
		\caption{\label{Figure:HSTS} Visual Results on HSTS. From the left to the right column (\textit{i.e.}, Figs.(\ref{Figure:HSTS:Input}--\ref{Figure:HSTS:GT})), the input hazy image, DehazeNet~\citep{DehazeNet}, MSCNN~\citep{MSCNN}, AOD-Net~\citep{AOD-Net}, DCP~\citep{DCP}, GRM~\citep{GRM}, N2V~\citep{N2V}, DIP~\citep{DIP}, DD~\citep{DD}, DDIP~\citep{Double-DIP}, our YOLY and the ground truth are presented in column-wise. Zooming-in is recommended for a better visualization.}
\end{figure*}

\subsection{Comparisons on Synthetic Datasets}

Table~\ref{Table:SOTS} and Fig.~\ref{Figure:SOTS} report the  quantitative and qualitative results on the indoor testing dataset (\textit{i.e.}, SOTS).  It is worthy to note that, we do not illustrate the visualization results of FVR, BCCR, and NLD due to the space limitation and their relatively inferior performance. From Table~\ref{Table:SOTS} and Fig.~\ref{Figure:SOTS}, one could observe that: 
\begin{itemize}
	\item Compared with the unsupervised deep methods, YOLY is 2.44 and 0.1180 higher than the best of them in terms of PSNR and SSIM, respectively. Compared with the unsupervised shallow method (\textit{i.e.}, prior-based), YOLY achieves a gain of 0.55 over the best method in SSIM. 
	\item YOLY surpasses all supervised methods excepted DehazeNet in the quantitative comparisons. In the qualitative comparisons, YOLY shows the best visual results as shown in Fig.~\ref{Figure:SOTS}, especially in the border areas of the figures. For example, DehazeNet fails to obtain a desirable result from the first image and remove the haze from the area close to the chair in the second image.  
	\item On the dataset, YOLY takes about 34.08s to obtain the final result for each image, \textit{i.e.}, it costs 68.15ms to run each epoch. Note that, this is the whole time cost of YOLY and no more training is required. 
\end{itemize}

Different from the indoor scenes, the outdoor scenes are usually much more complex in depth, illustration, and background. As shown in Table~\ref{Table:HSTS} and Fig.~\ref{Figure:HSTS}, our method achieves promising dehazing performance in the outdoor scenes (HSTS). More specifically, 
\begin{itemize}
	\item Table~\ref{Table:HSTS} shows that our method is remarkably superior to all the unsupervised baselines  including deep and prior-based shallow methods. More specifically, it outperforms the best unsupervised deep method by 2.91 and 0.0283 in terms of PSNR and SSIM, respectively. It is also 4.9 and 0.0941 higher than the best prior-based dehazing methods. Note that, N2N cannot achieve result on the HSTS dataset because it requires multiple hazy samples from the same scene whereas this dataset only includes a single sample for each scene. 
	\item Although YOLY is quantitively worse than DehazeNet, it shows better  recovery performance in the visual comparison (see Fig.~\ref{Figure:HSTS:DehazeNet} of DehazeNet and Fig.~\ref{Figure:HSTS:SL} of YOLY for example). In fact, the haze-free images recovered by our YOLY seem more favorite than the ground truth clean image (Fig.~\ref{Figure:HSTS:GT}) which might be corrupted during data collection. 
	\item The visualization results partially show the limitations of the traditional prior-based methods in handling the complex scenes which may violate the adopted prior. More specifically, DCP shows some distortions in the sky area. The main reason is that the sky and the bright areas do not satisfy the assumption of DCP, thus resulting in inaccurate estimation of the transmission map. 
	\item  It is worth noting that all unsupervised deep models excepted DDIP and YOLY cannot perform well in the qualitative and quantitative comparisons. The reason is that they are not specifically designed for dehazing. This shows the difficulty and necessity in developing new deep unsupervised single image dehazing methods. 
	\item On the dataset, YOLY takes about 28.58s for handling each image averagely and each epoch costs 57.17ms. 
	\end{itemize}

\begin{figure*}[!t]
	\def \m_width{0.095}
	\def \a_height{1.2cm}
	\def \b_height{1.2cm}
	\begin{center}
		\subfigure{\includegraphics[width=\m_width\textwidth, height=\a_height]{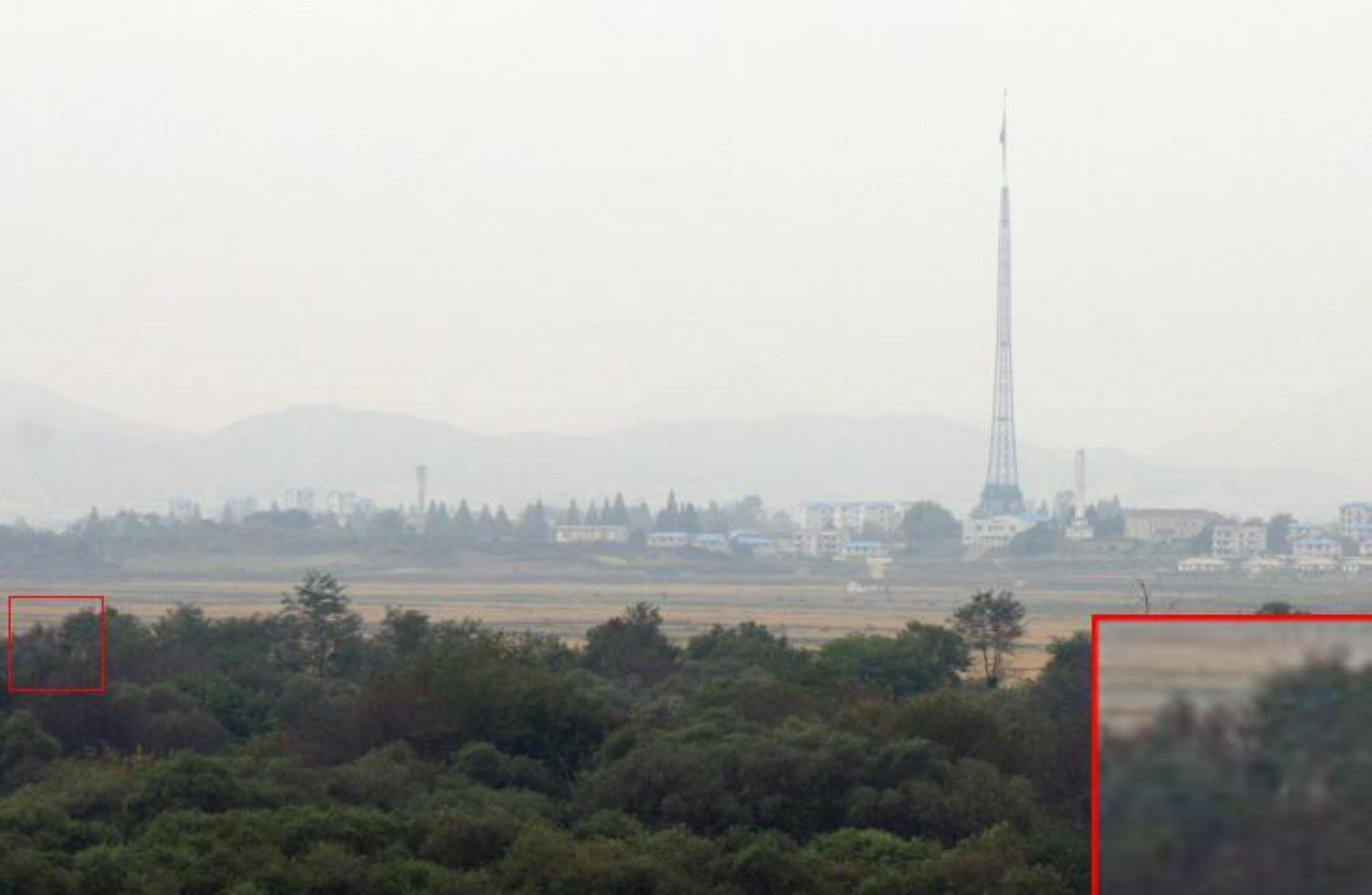}}
		\subfigure{\includegraphics[width=\m_width\textwidth, height=\a_height]{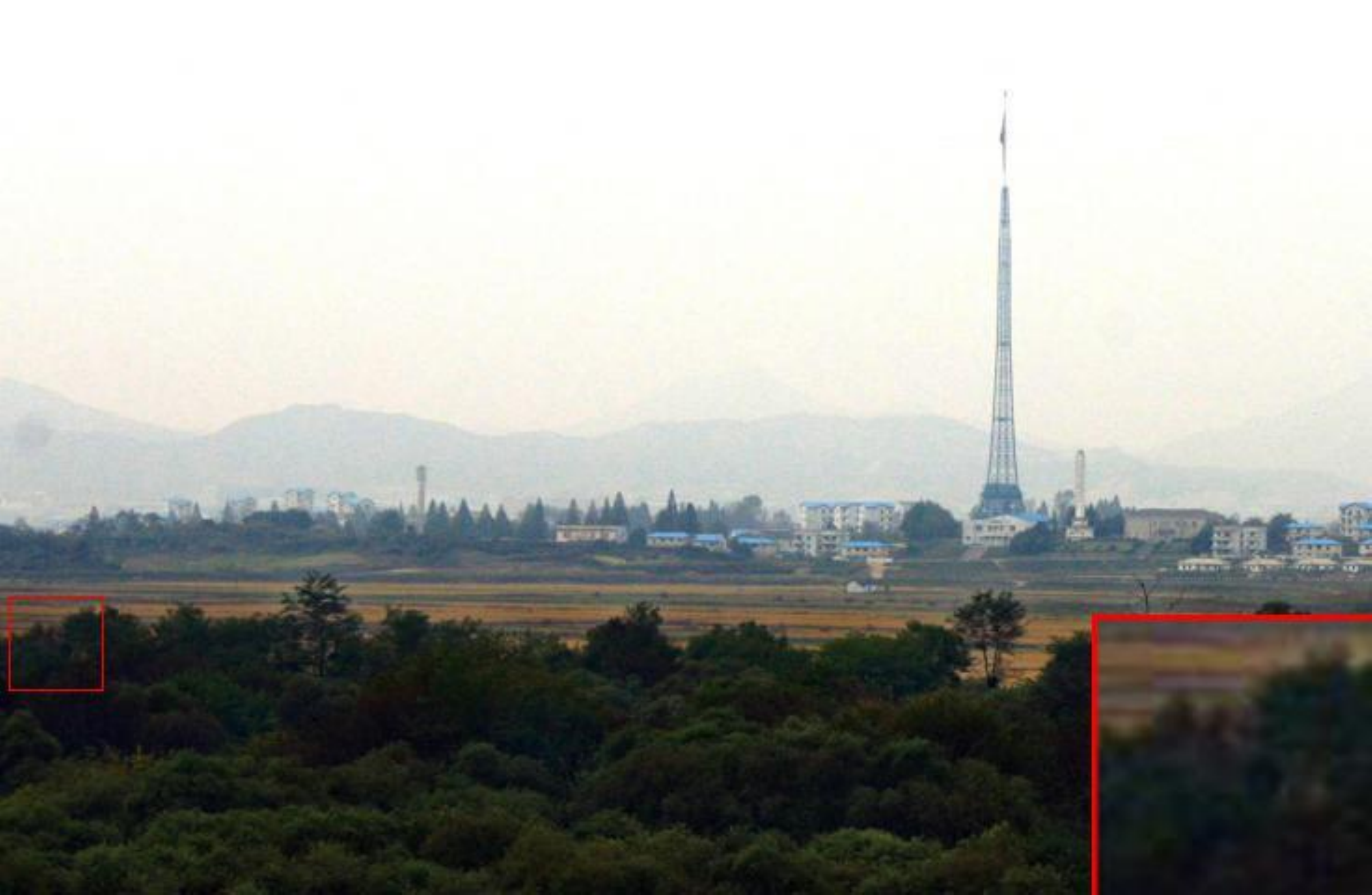}}
		\subfigure{\includegraphics[width=\m_width\textwidth, height=\a_height]{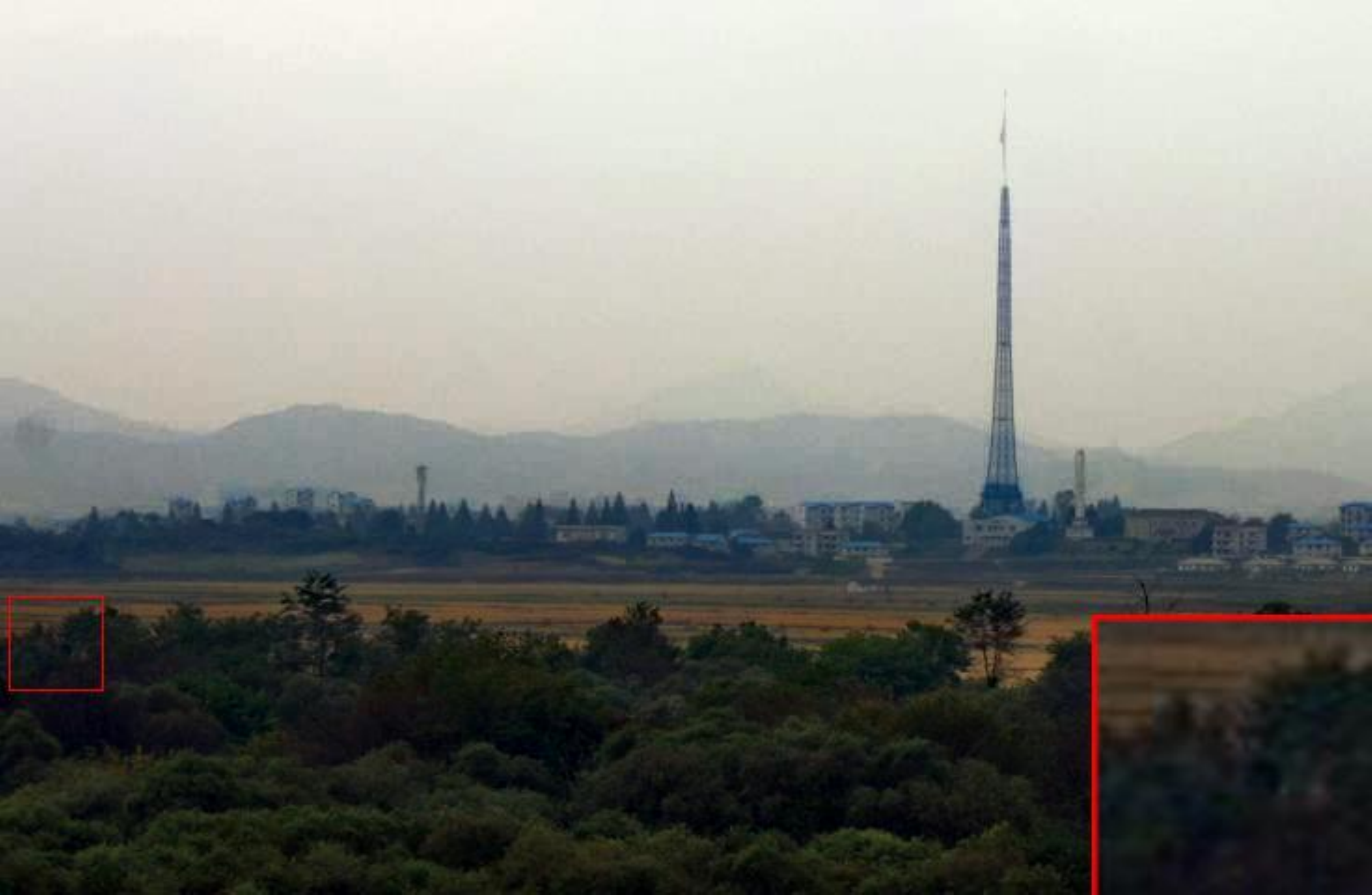}}
		\subfigure{\includegraphics[width=\m_width\textwidth, height=\a_height]{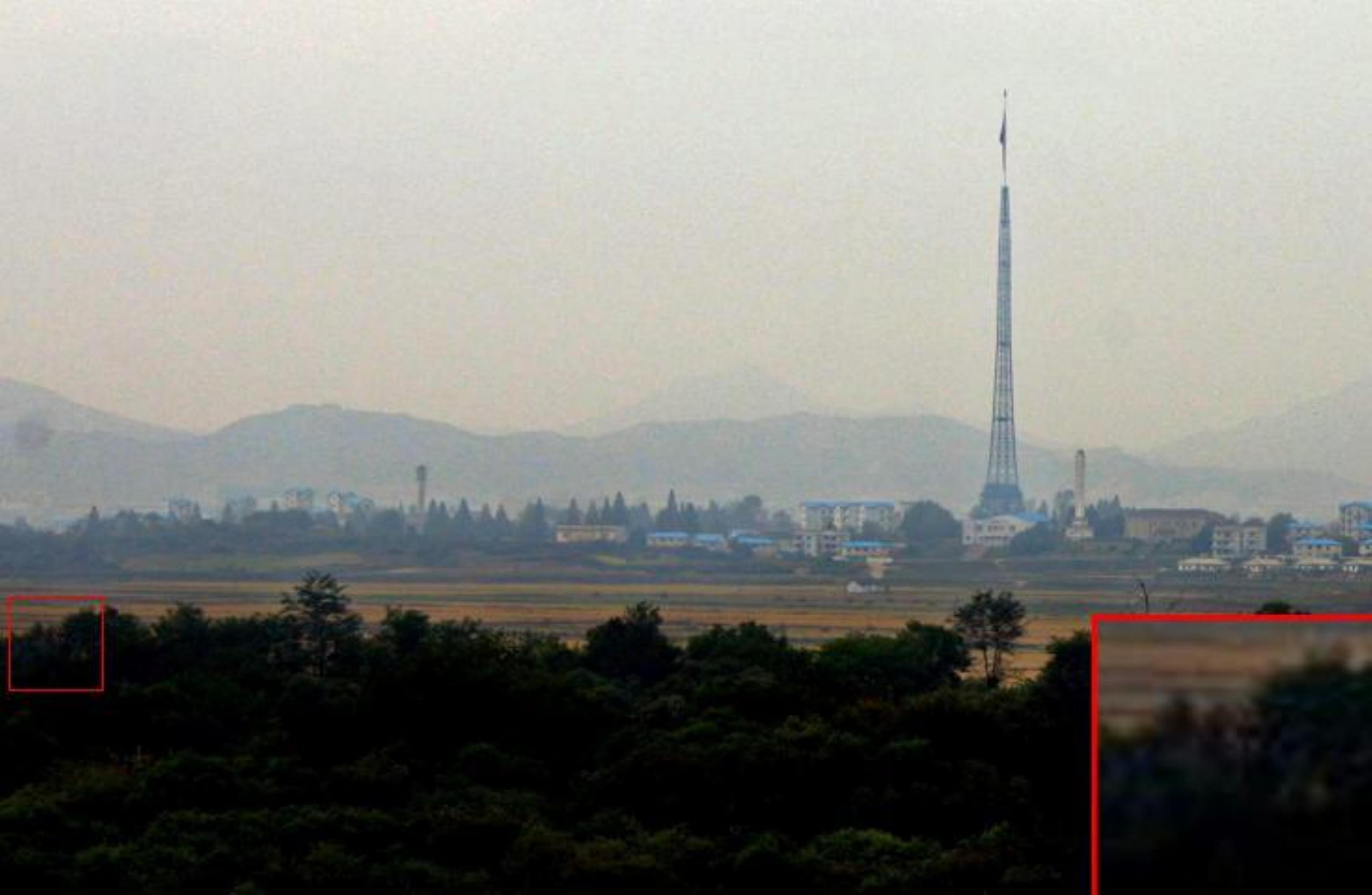}}
		\subfigure{\includegraphics[width=\m_width\textwidth, height=\a_height]{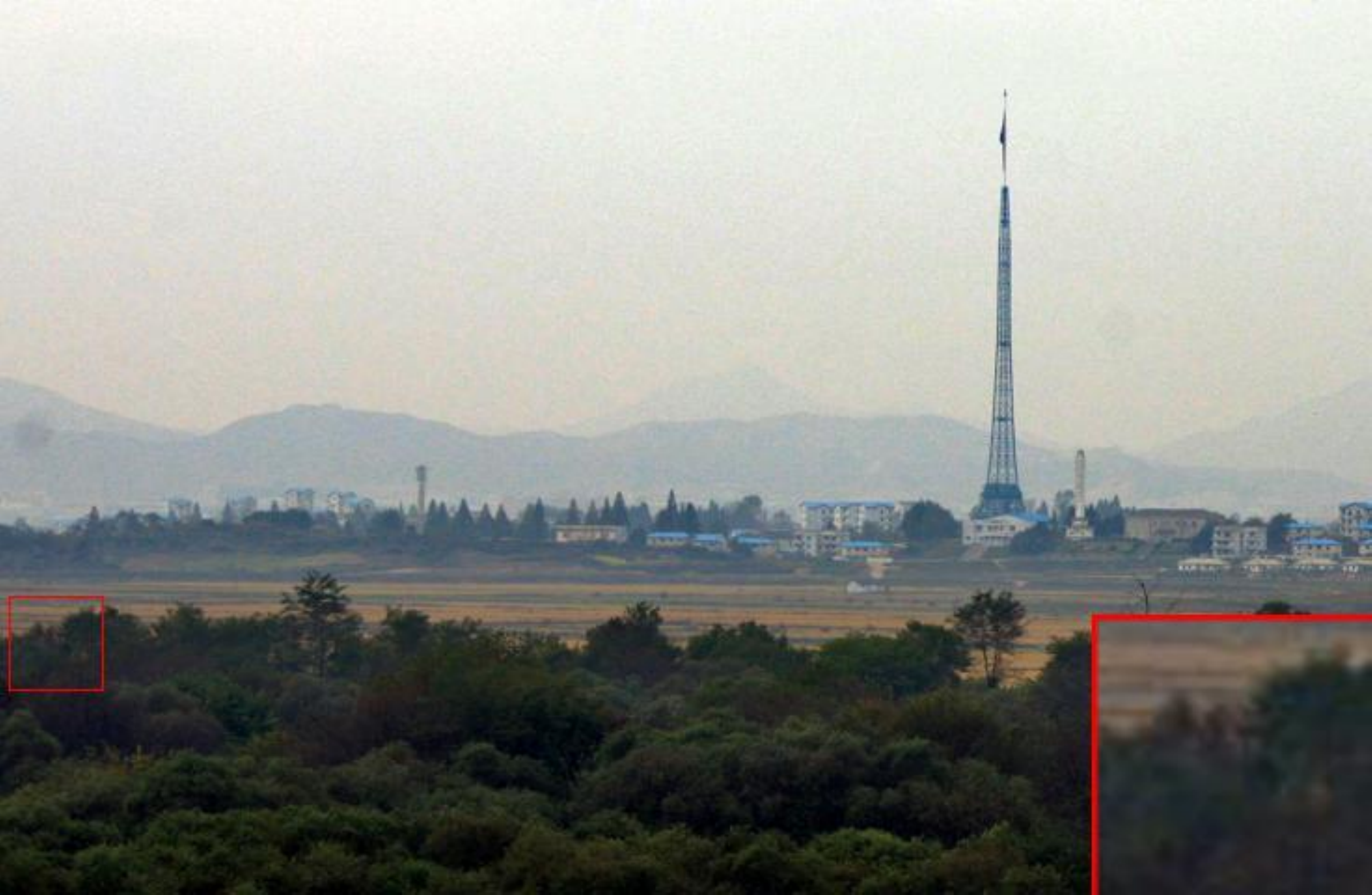}}
		\subfigure{\includegraphics[width=\m_width\textwidth, height=\a_height]{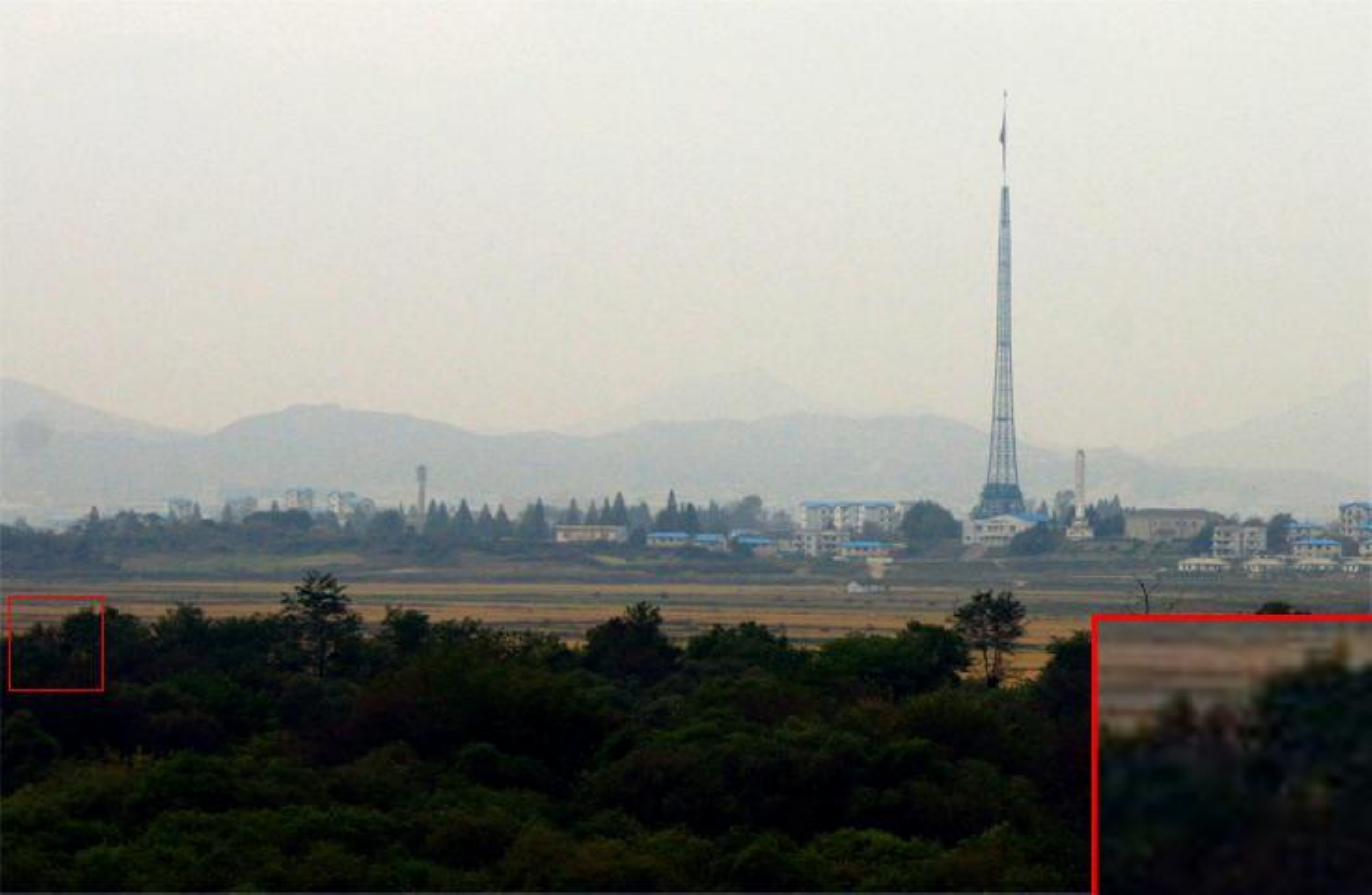}}
		\subfigure{\includegraphics[width=\m_width\textwidth, height=\a_height]{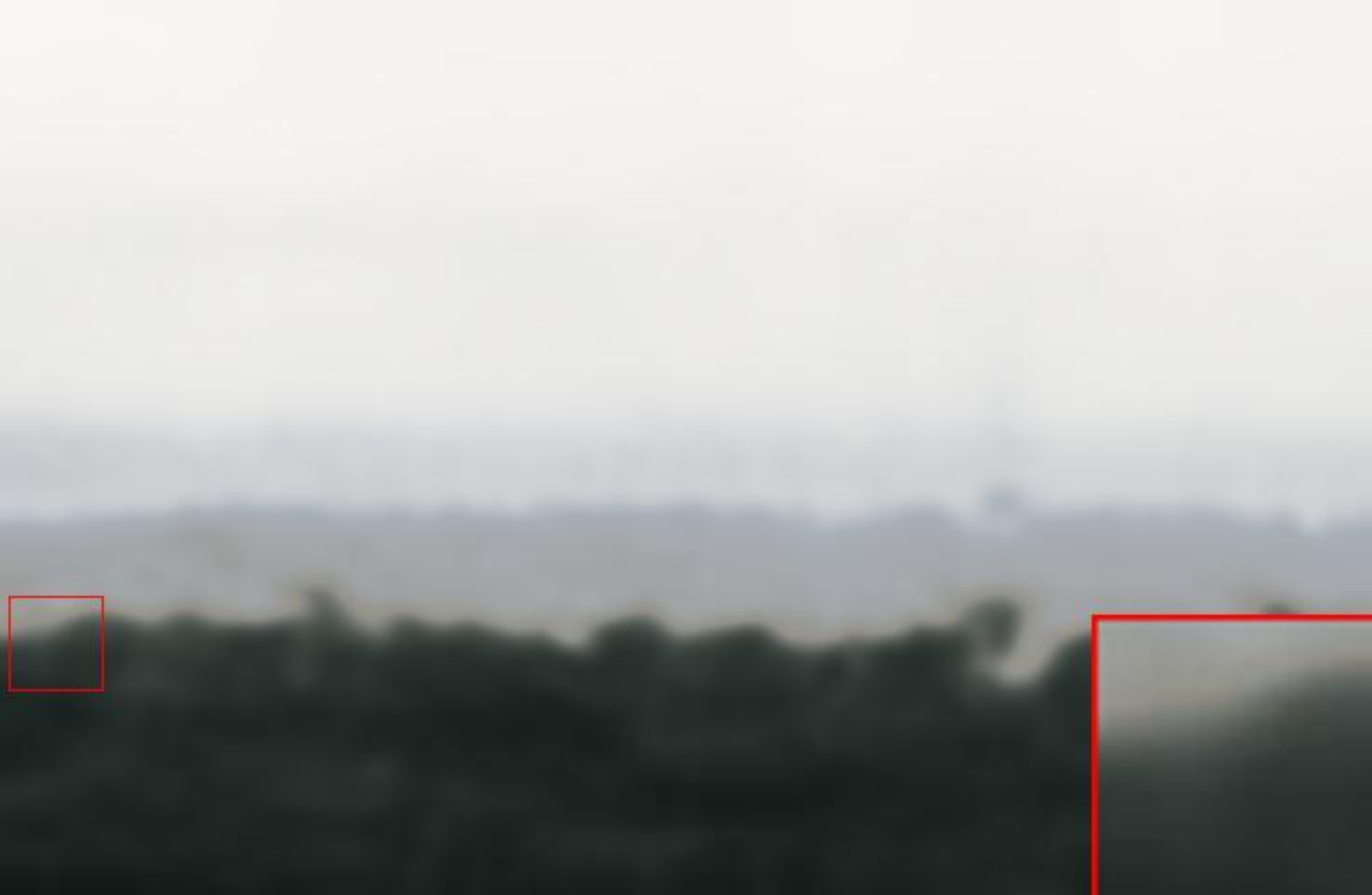}}
		\subfigure{\includegraphics[width=\m_width\textwidth, height=\a_height]{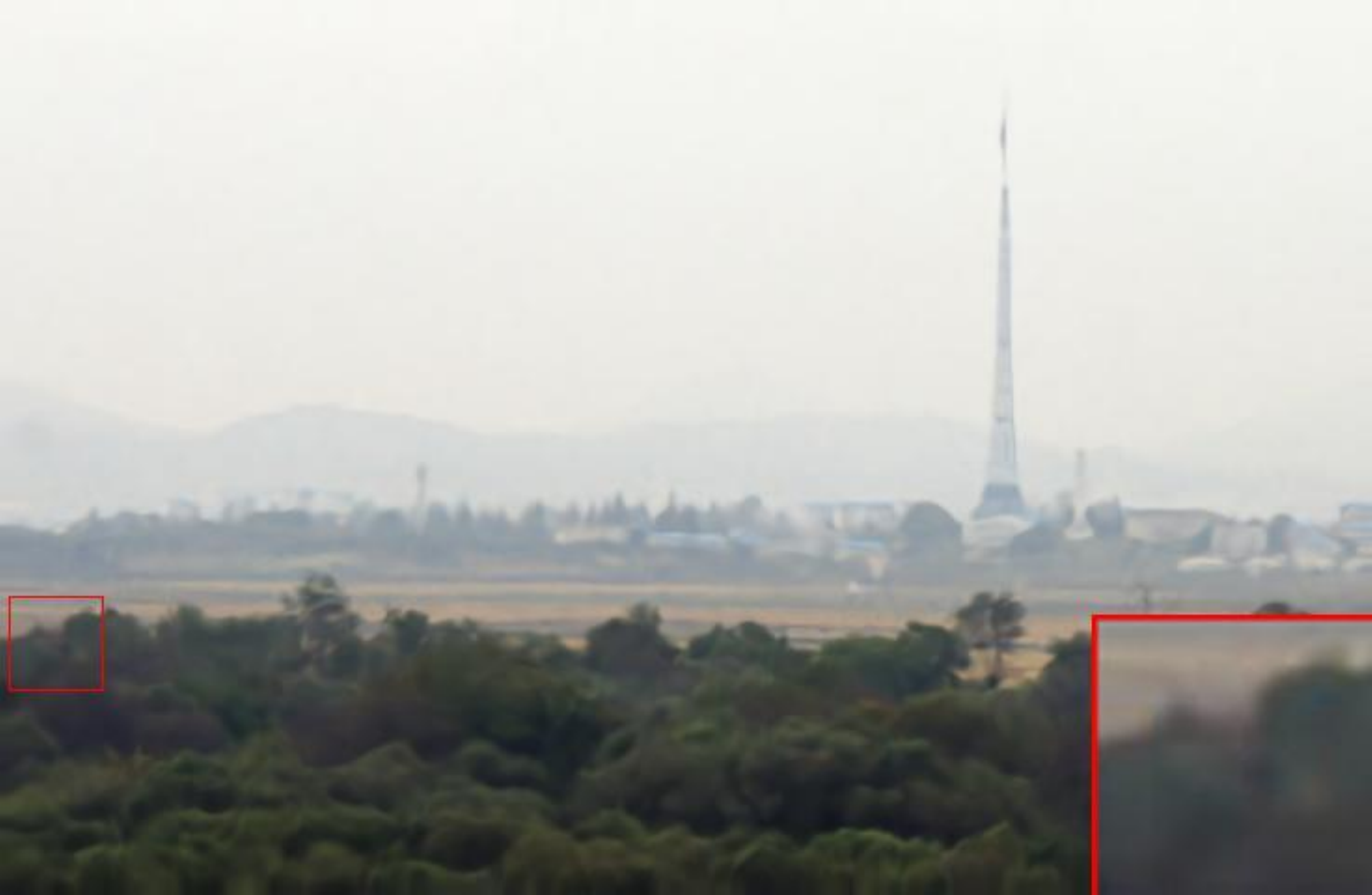}}
		\subfigure{\includegraphics[width=\m_width\textwidth, height=\a_height]{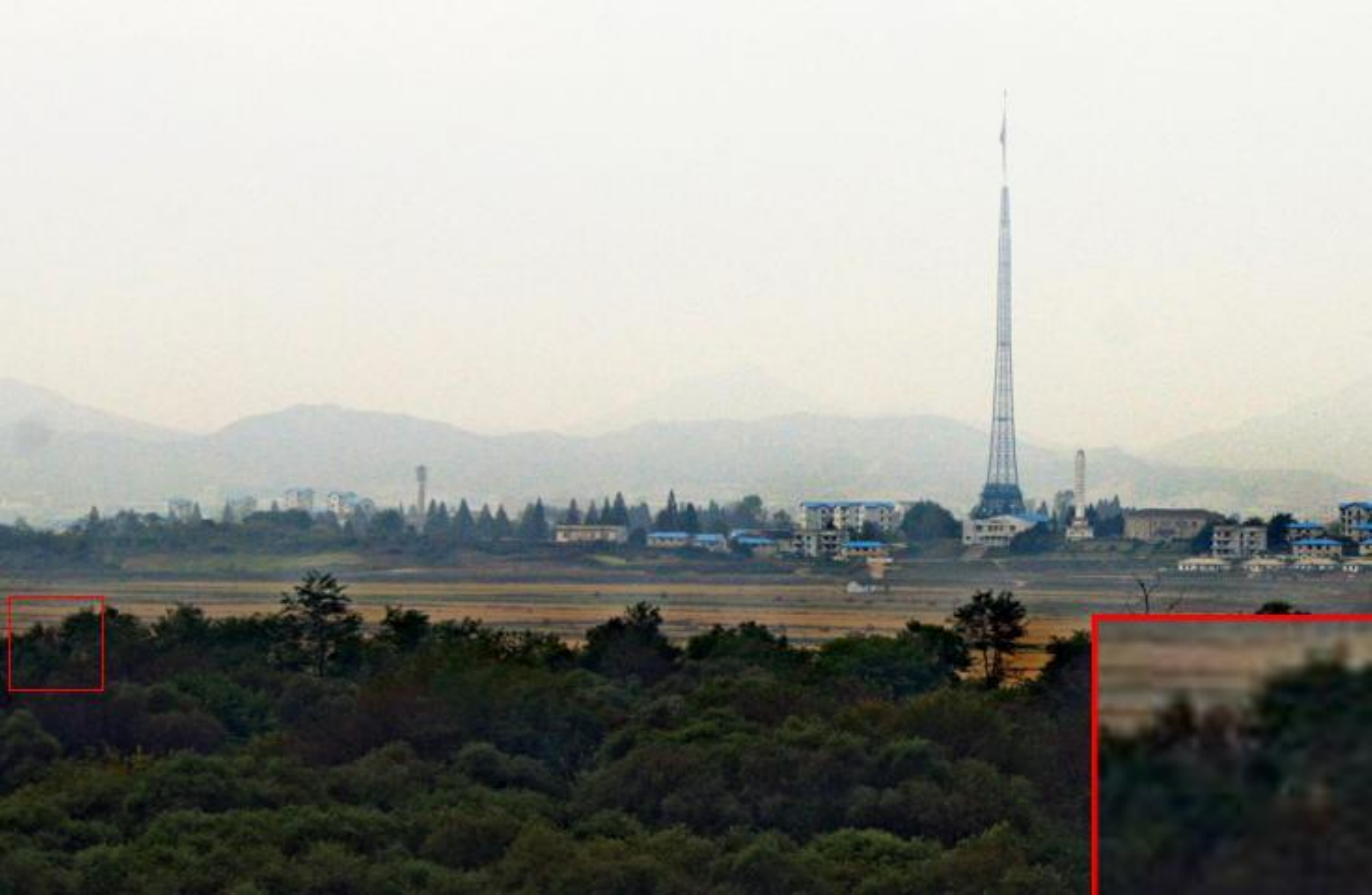}}
		\subfigure{\includegraphics[width=\m_width\textwidth, height=\a_height]{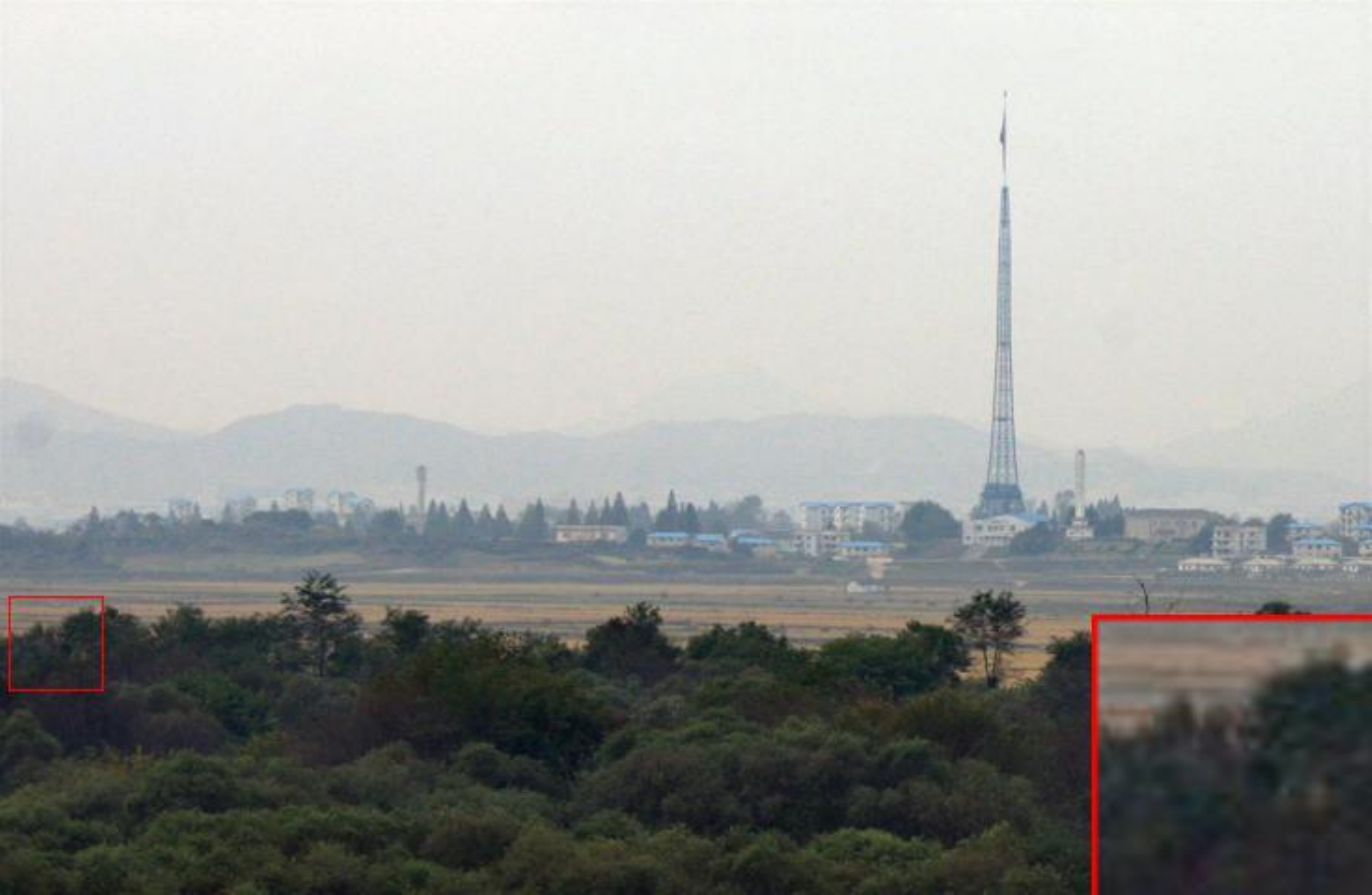}}
	\end{center}
	\vspace{-0.7cm}


	\begin{center}
	\setcounter{subfigure}{0}
		\subfigure[]{\label{Figure:RW:Input}\includegraphics[width=\m_width\textwidth, height=\b_height]{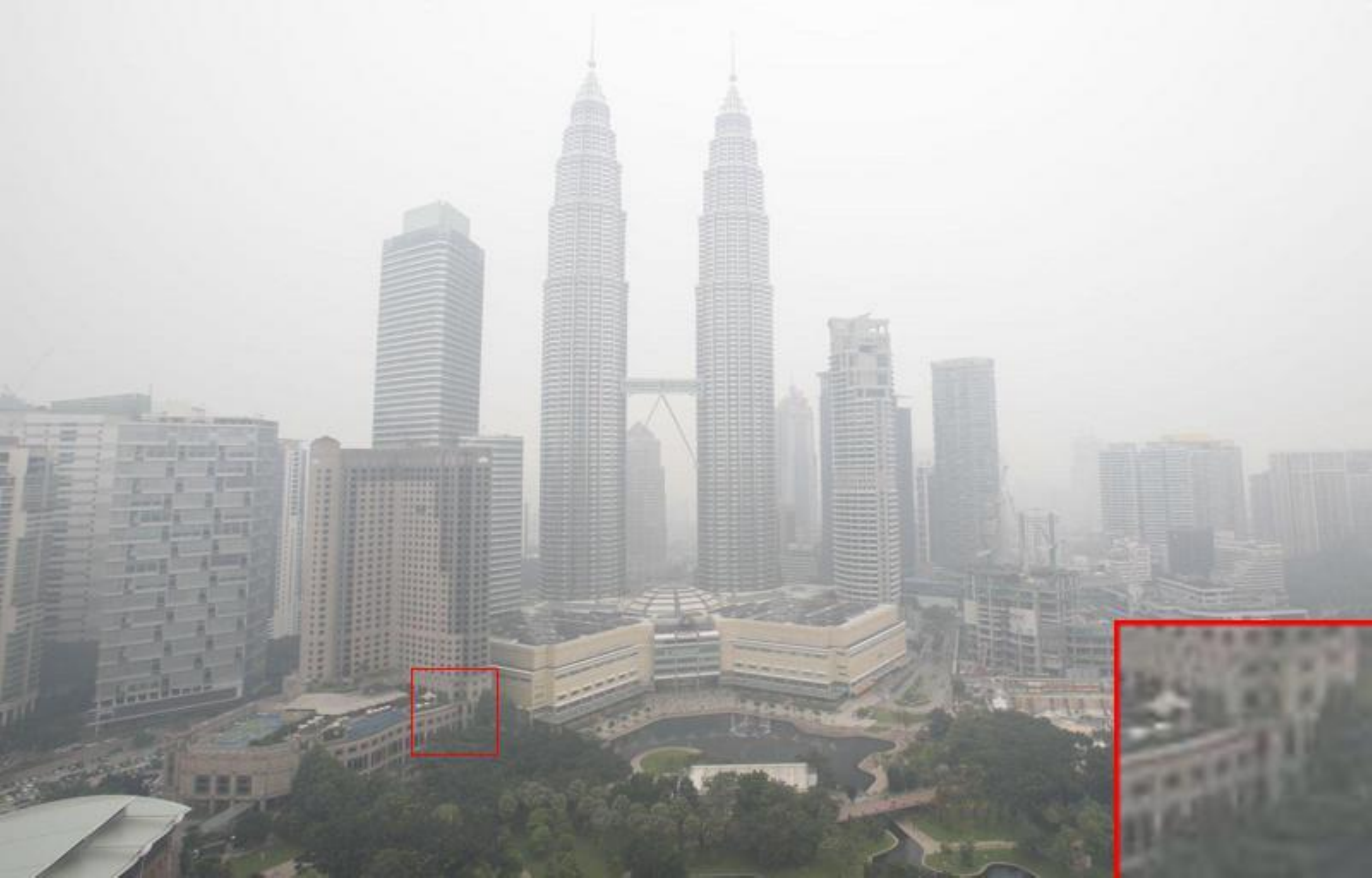}}
		\subfigure[]{\label{Figure:RW:DehazeNet}\includegraphics[width=\m_width\textwidth,height=\b_height]{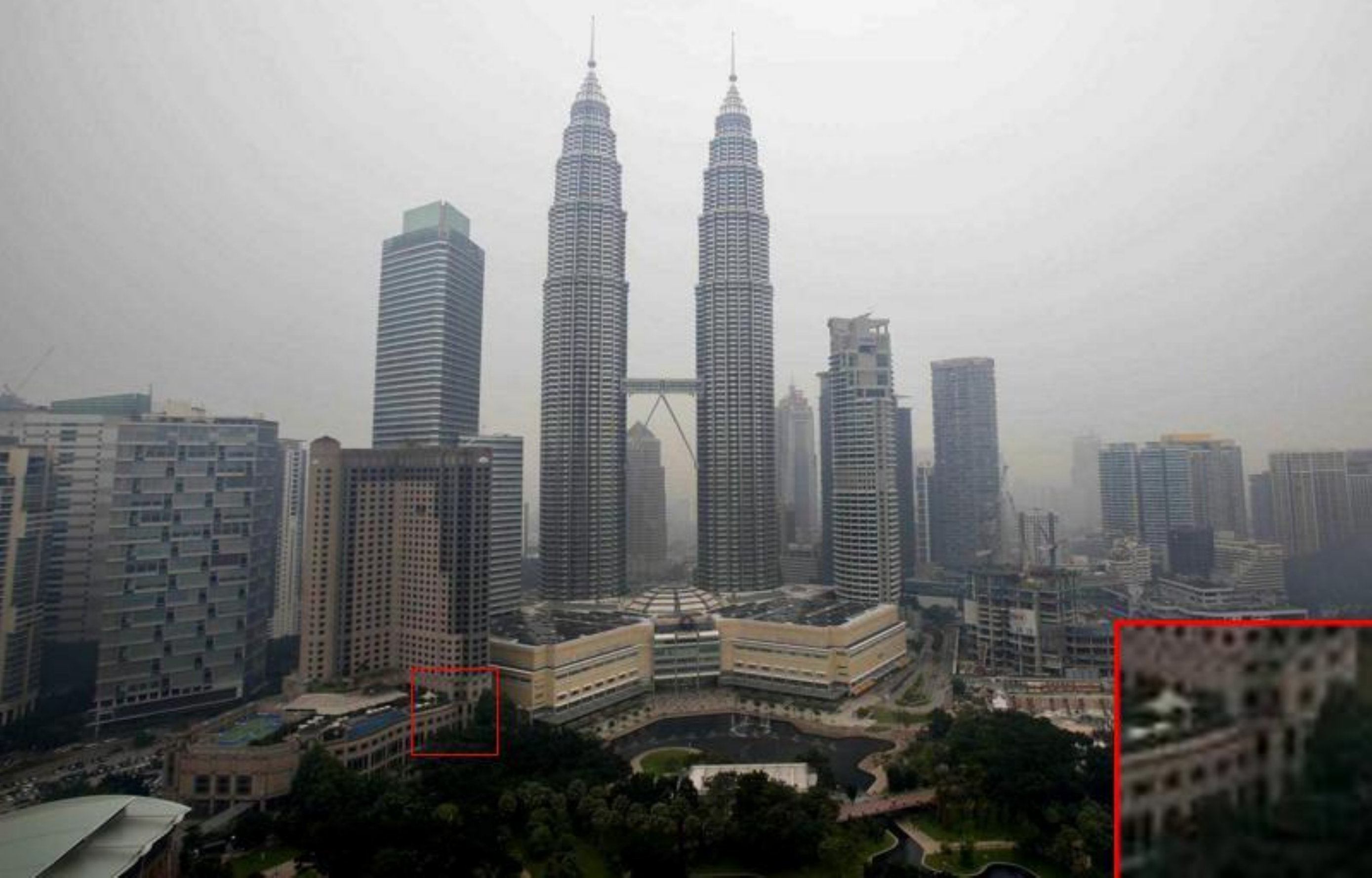}}
		\subfigure[]{\includegraphics[width=\m_width\textwidth,height=\b_height]{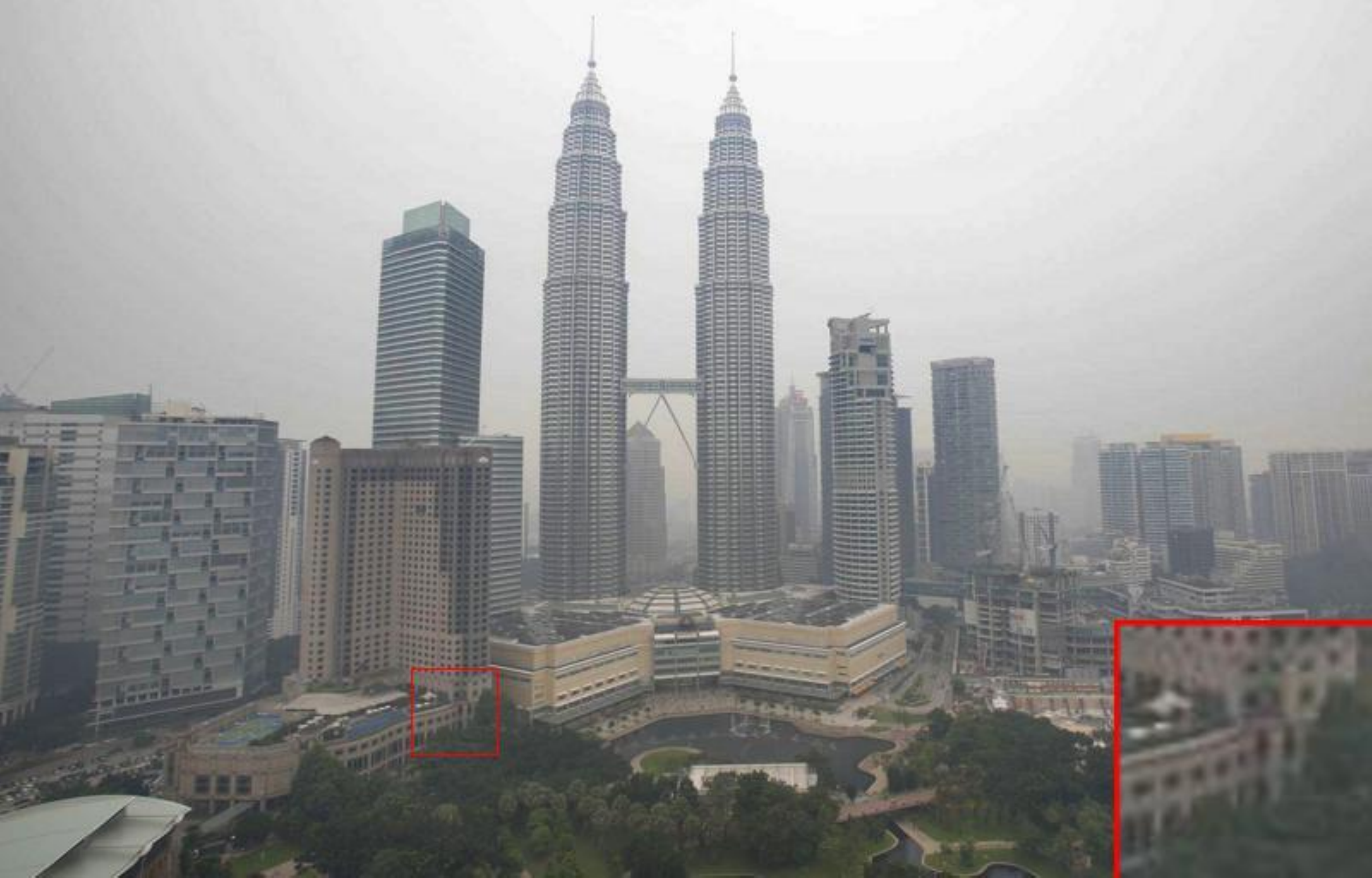}}
		\subfigure[]{\includegraphics[width=\m_width\textwidth,height=\b_height]{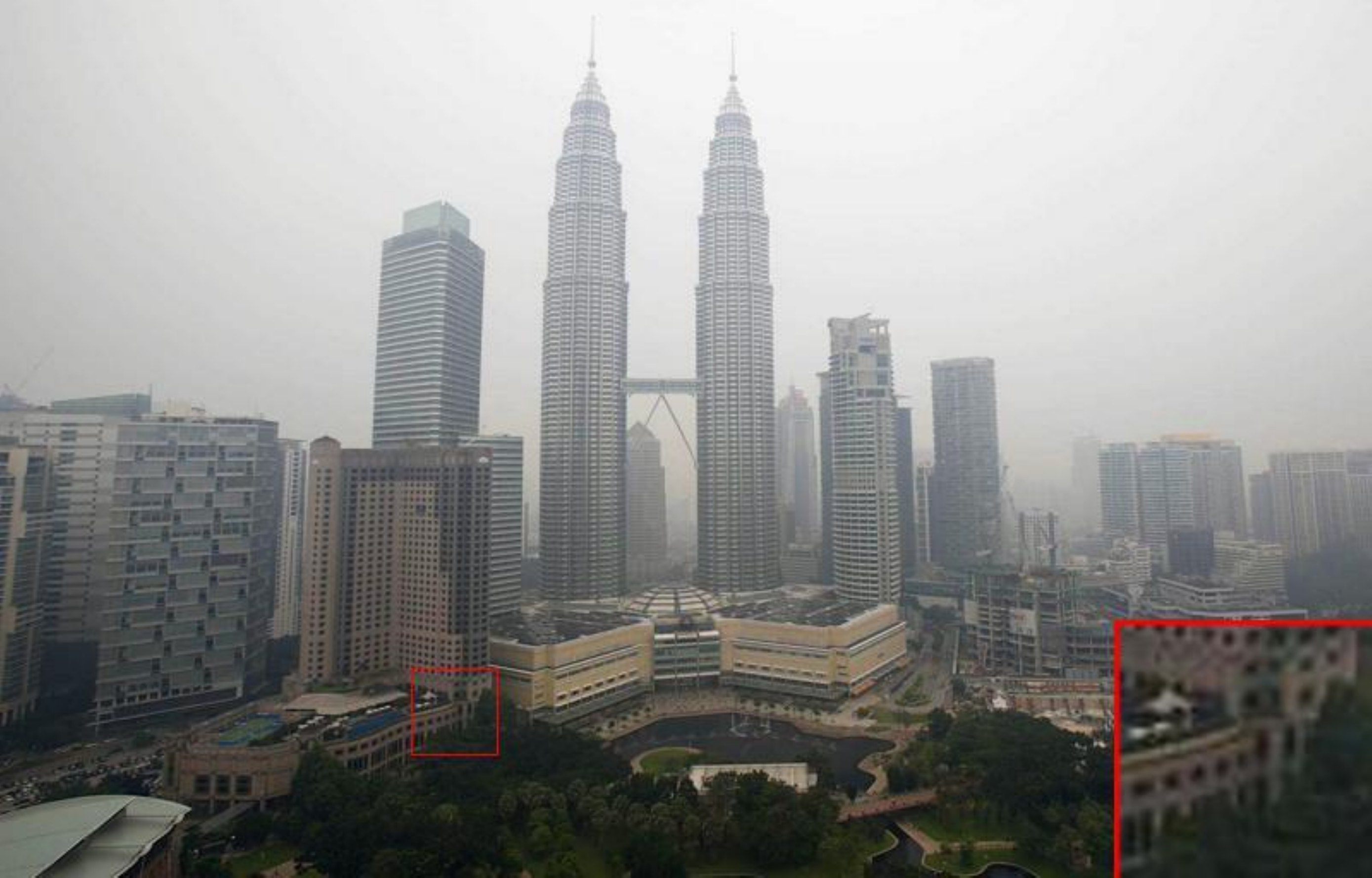}}
		\subfigure[]{\includegraphics[width=\m_width\textwidth,height=\b_height]{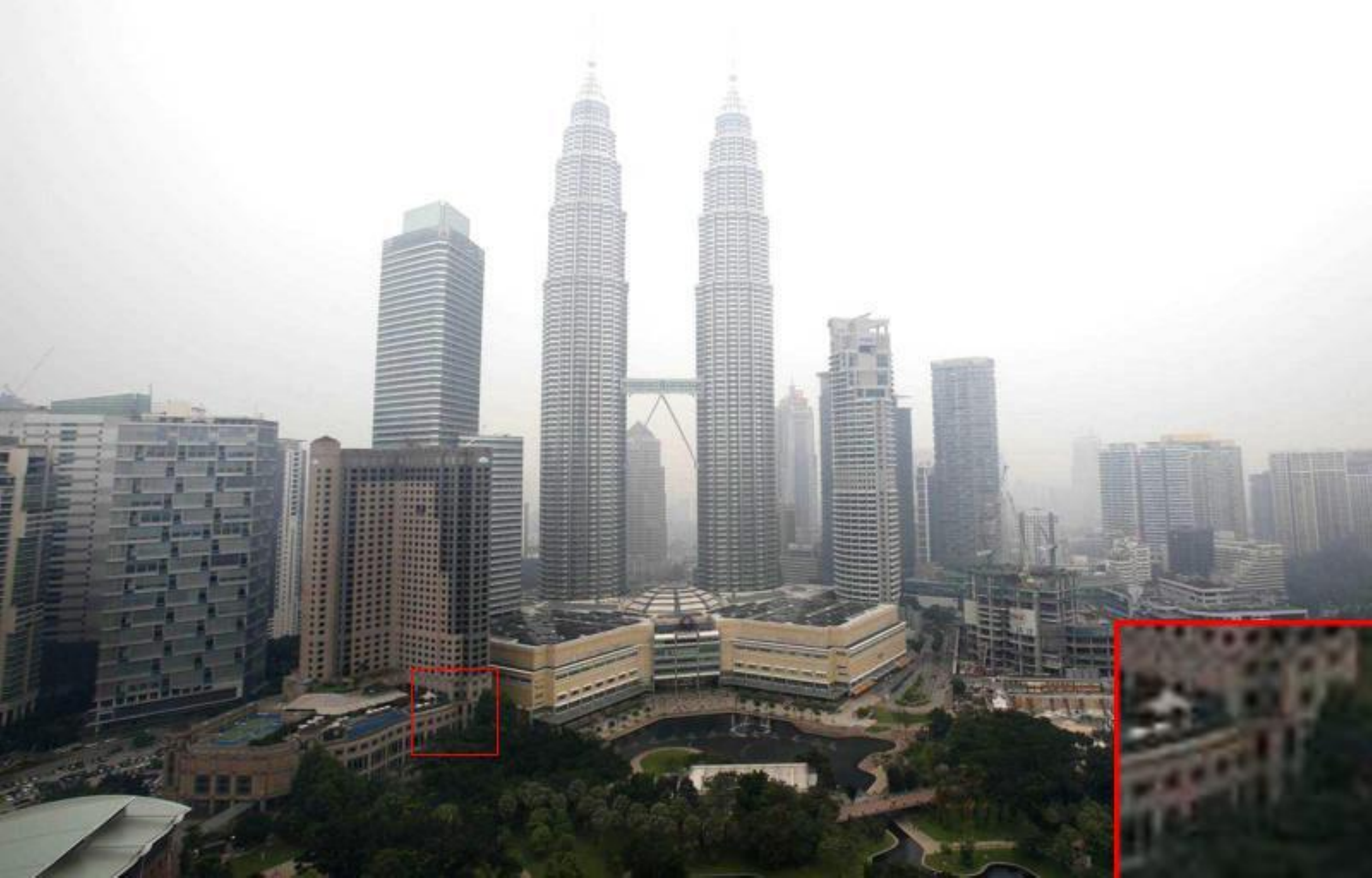}}
		\subfigure[]{\includegraphics[width=\m_width\textwidth,height=\b_height]{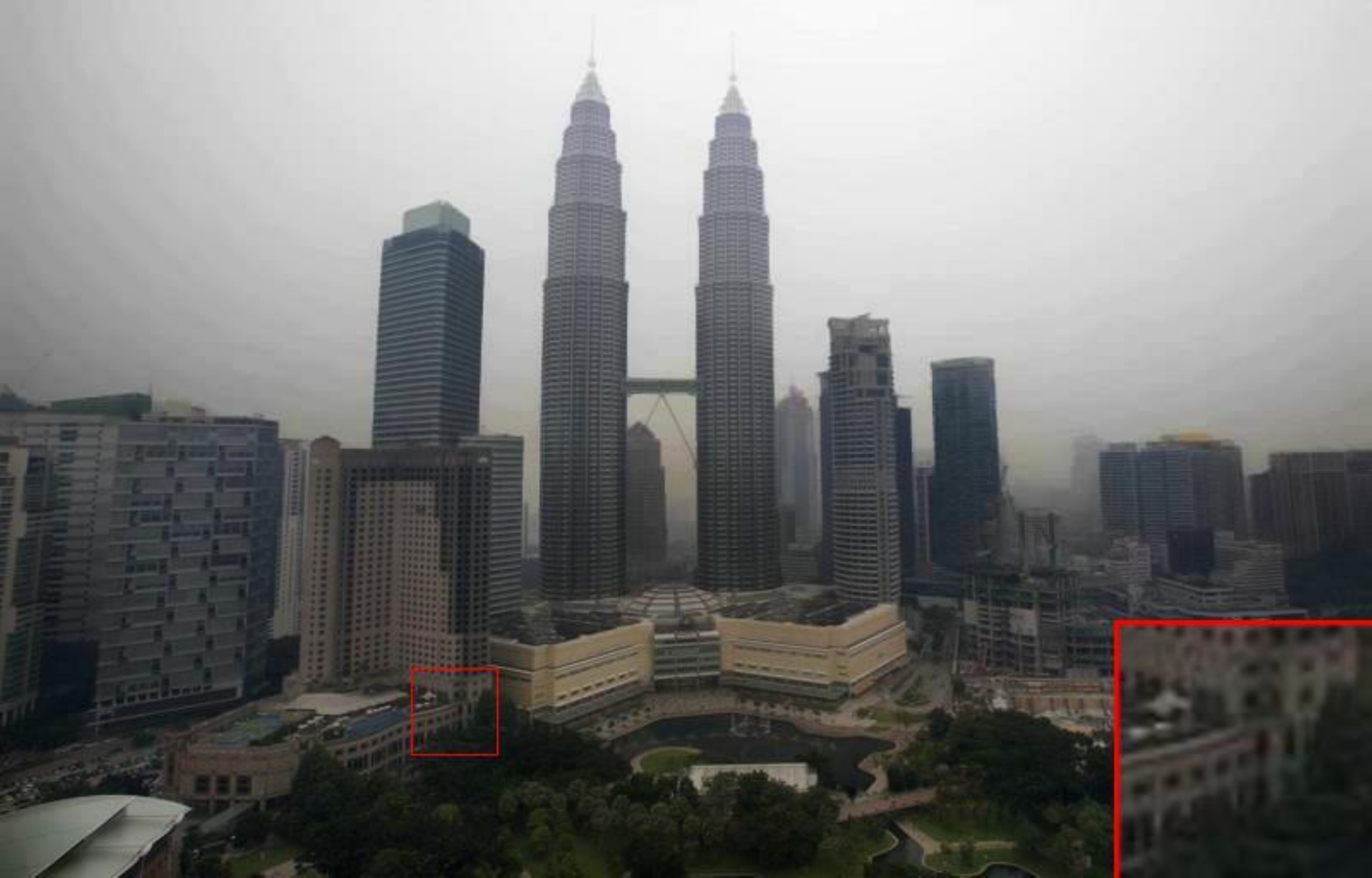}}
		\subfigure[]{\includegraphics[width=\m_width\textwidth,height=\b_height]{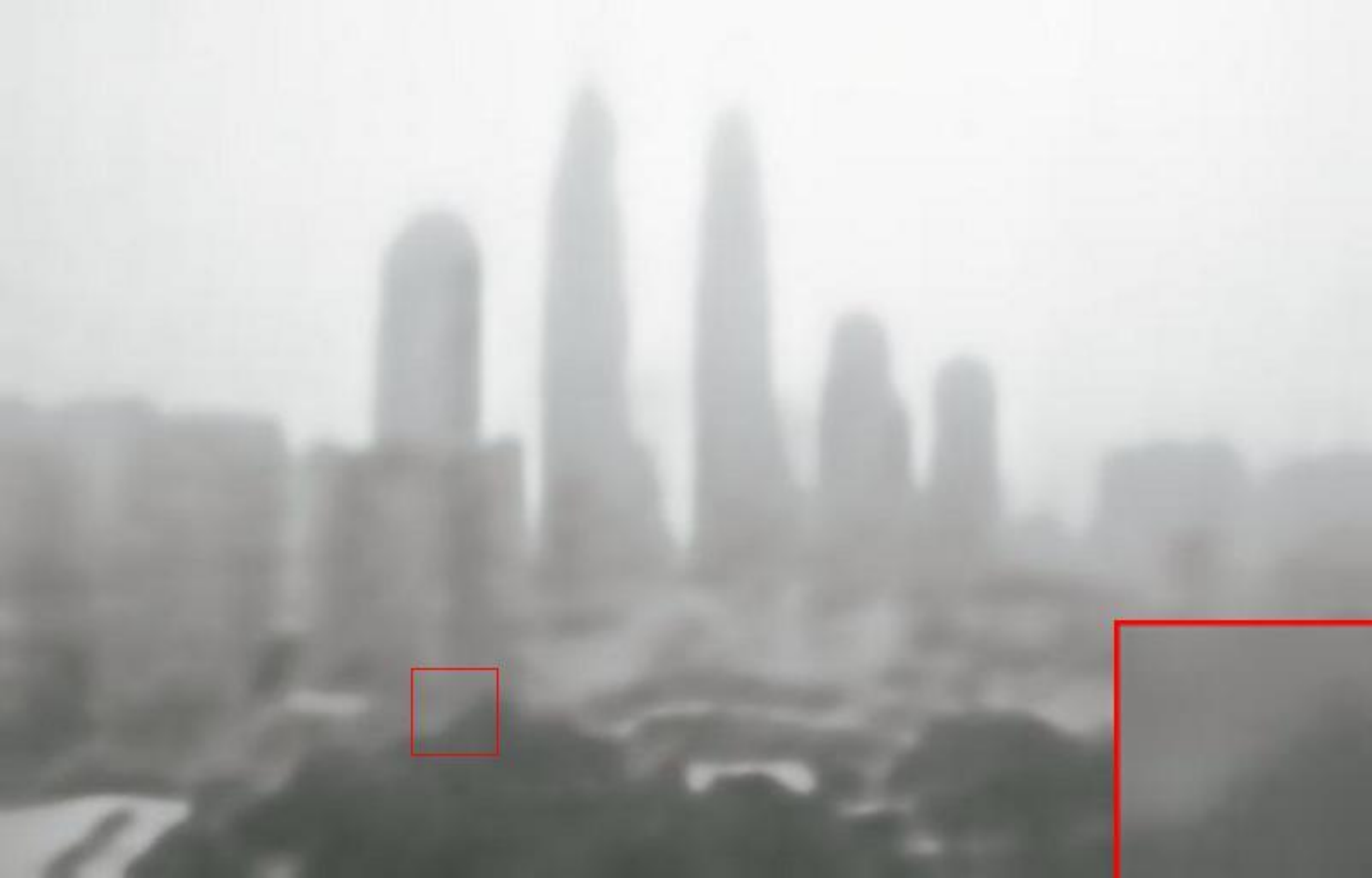}}
		\subfigure[]{\includegraphics[width=\m_width\textwidth,height=\b_height]{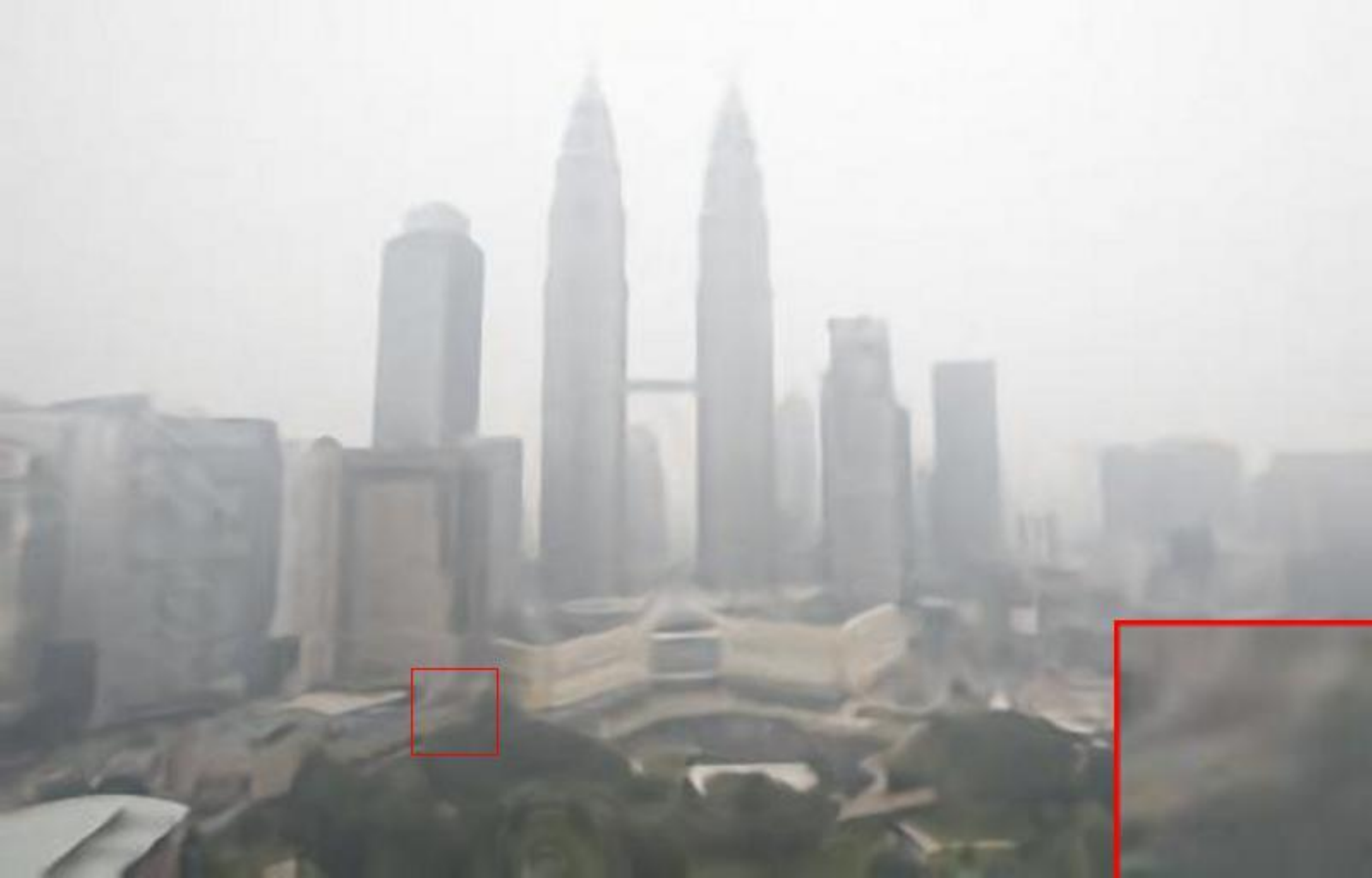}}
		\subfigure[]{\includegraphics[width=\m_width\textwidth,height=\b_height]{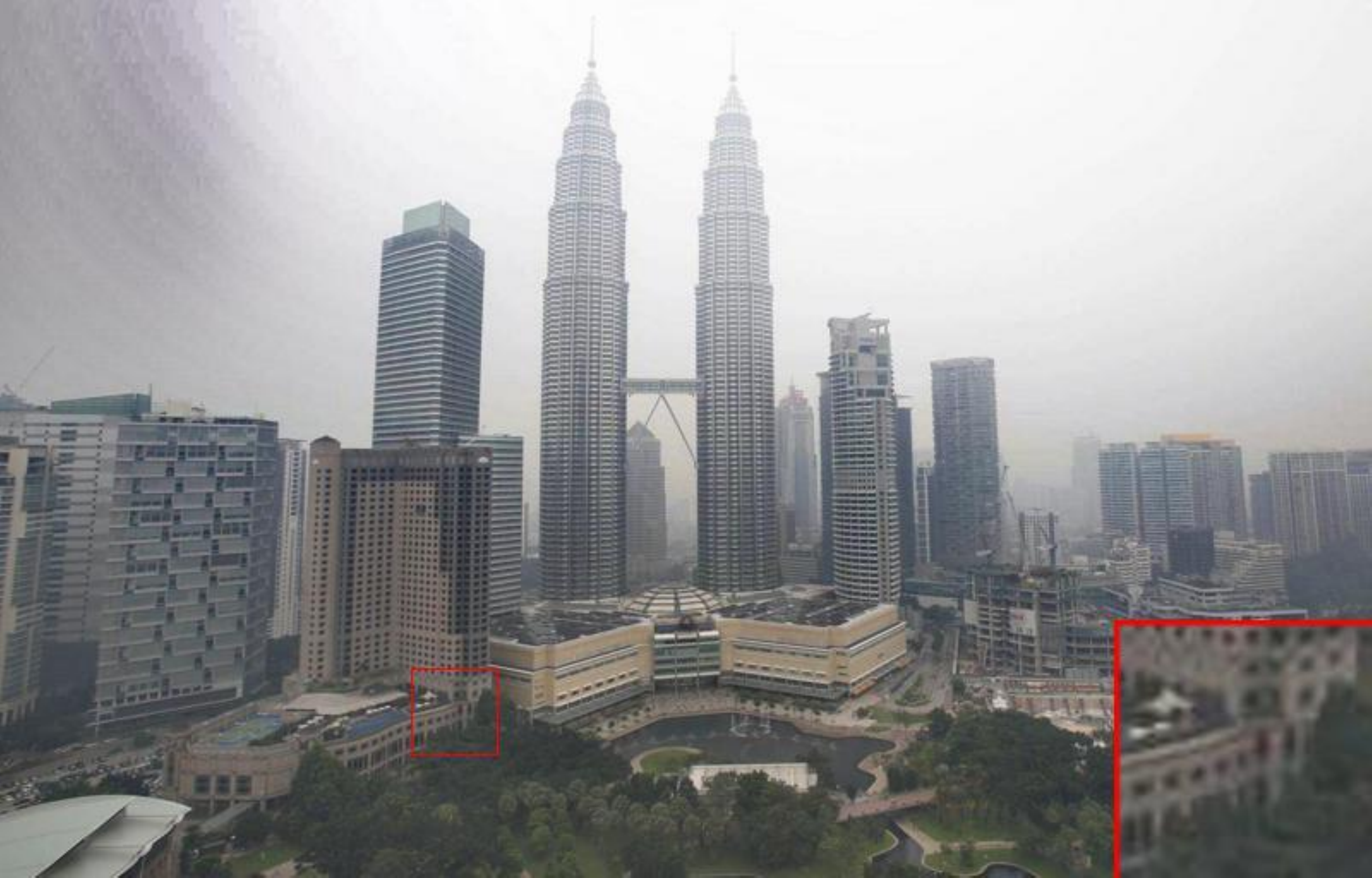}}
		\subfigure[]{\label{Figure:RW:SL}\includegraphics[width=\m_width\textwidth,height=\b_height]{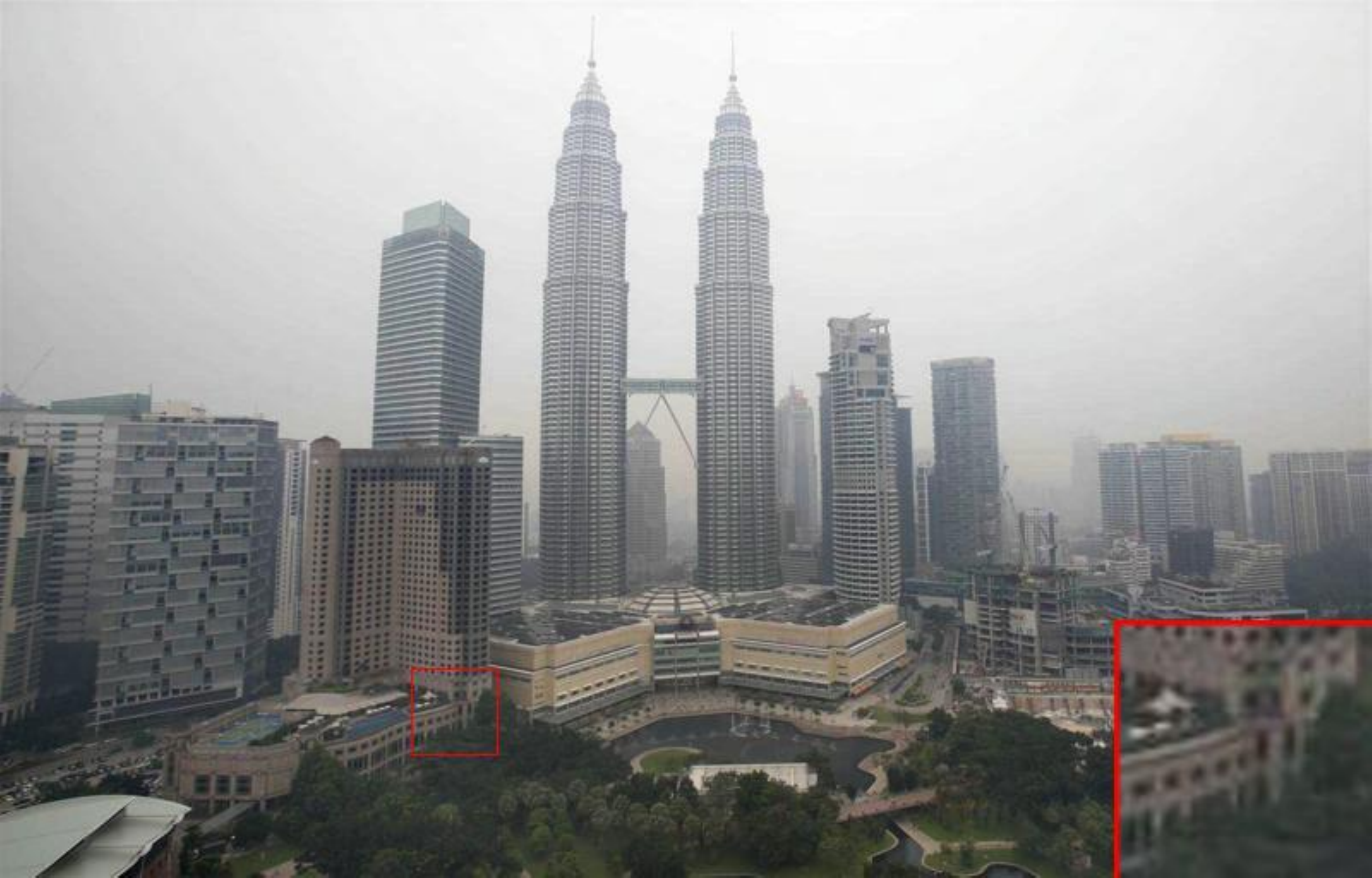}}
	\end{center}
	\caption{\label{Figure:RW} Visual Results on the Real-World Dataset. From the left to the right column (\textit{i.e.}, Figs.(\ref{Figure:RW:Input}--\ref{Figure:RW:SL})), the input hazy image, DehazeNet~\citep{DehazeNet}, MSCNN~\citep{MSCNN}, AOD-Net~\citep{AOD-Net}, DCP~\citep{DCP},  GRM~\citep{GRM}, DIP~\citep{DCP}, DD~\citep{DD}, DDIP~\citep{Double-DIP} and our method are presented in column-wise. Some areas are highlighted by red rectangles and zooming-in is recommended for a better visualization and comparison. }
\end{figure*}

\begin{figure*}[!t]
	\begin{center}
	\def \m_wid{0.0743}
		\subfigure{\includegraphics[scale=\m_wid]{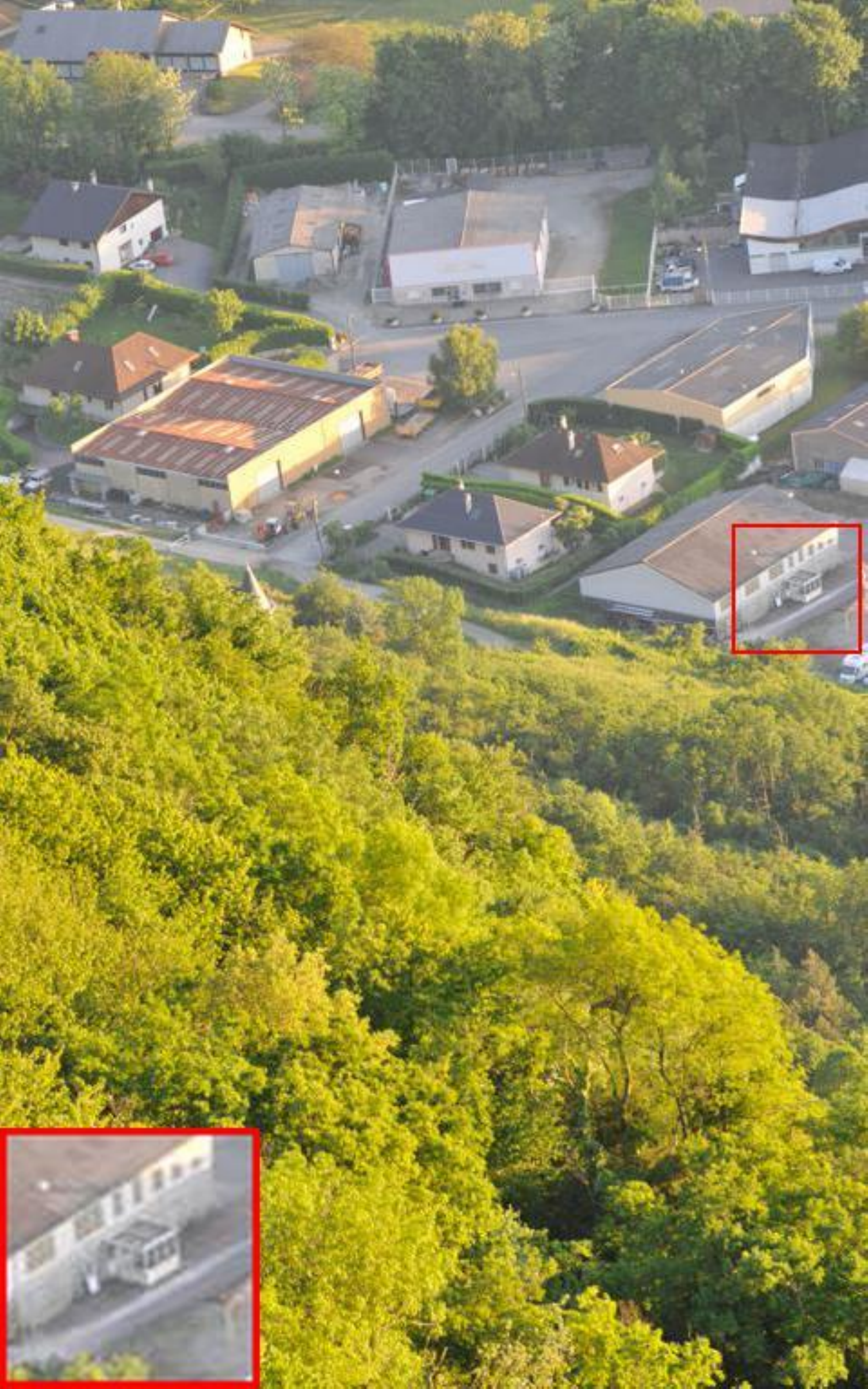}}
		\subfigure{\includegraphics[scale=\m_wid]{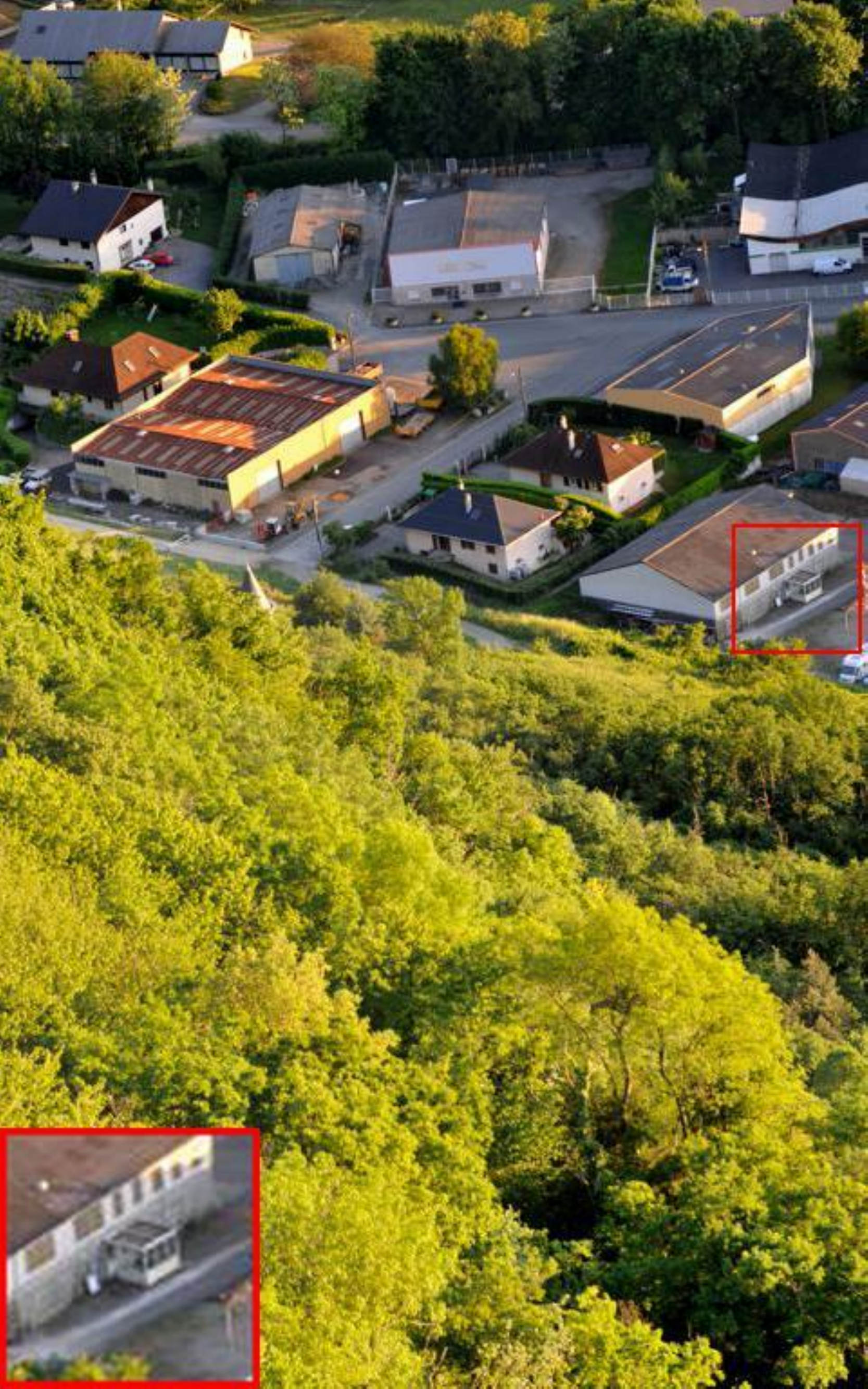}}
		\subfigure{\includegraphics[scale=\m_wid]{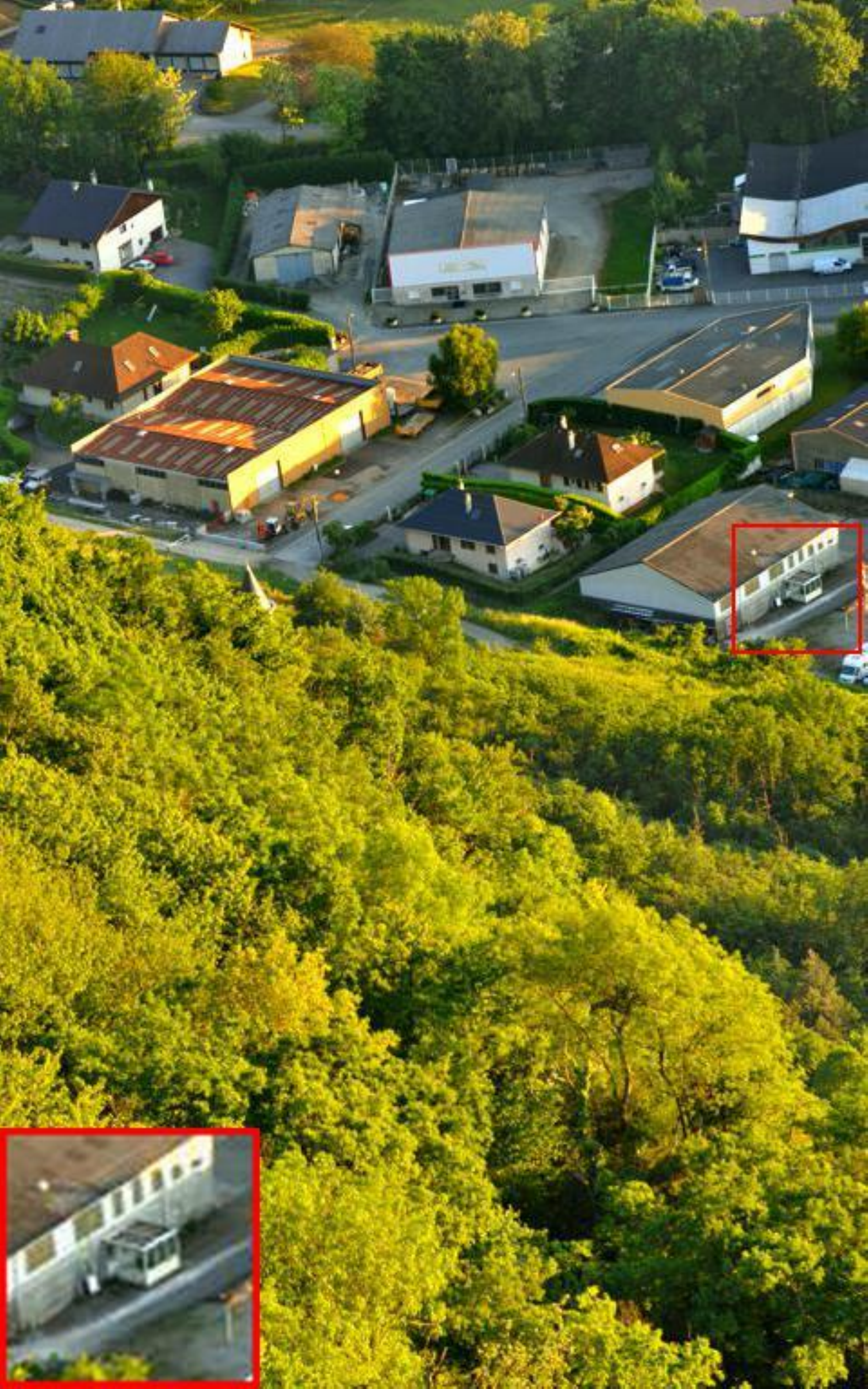}}
		\subfigure{\includegraphics[scale=\m_wid]{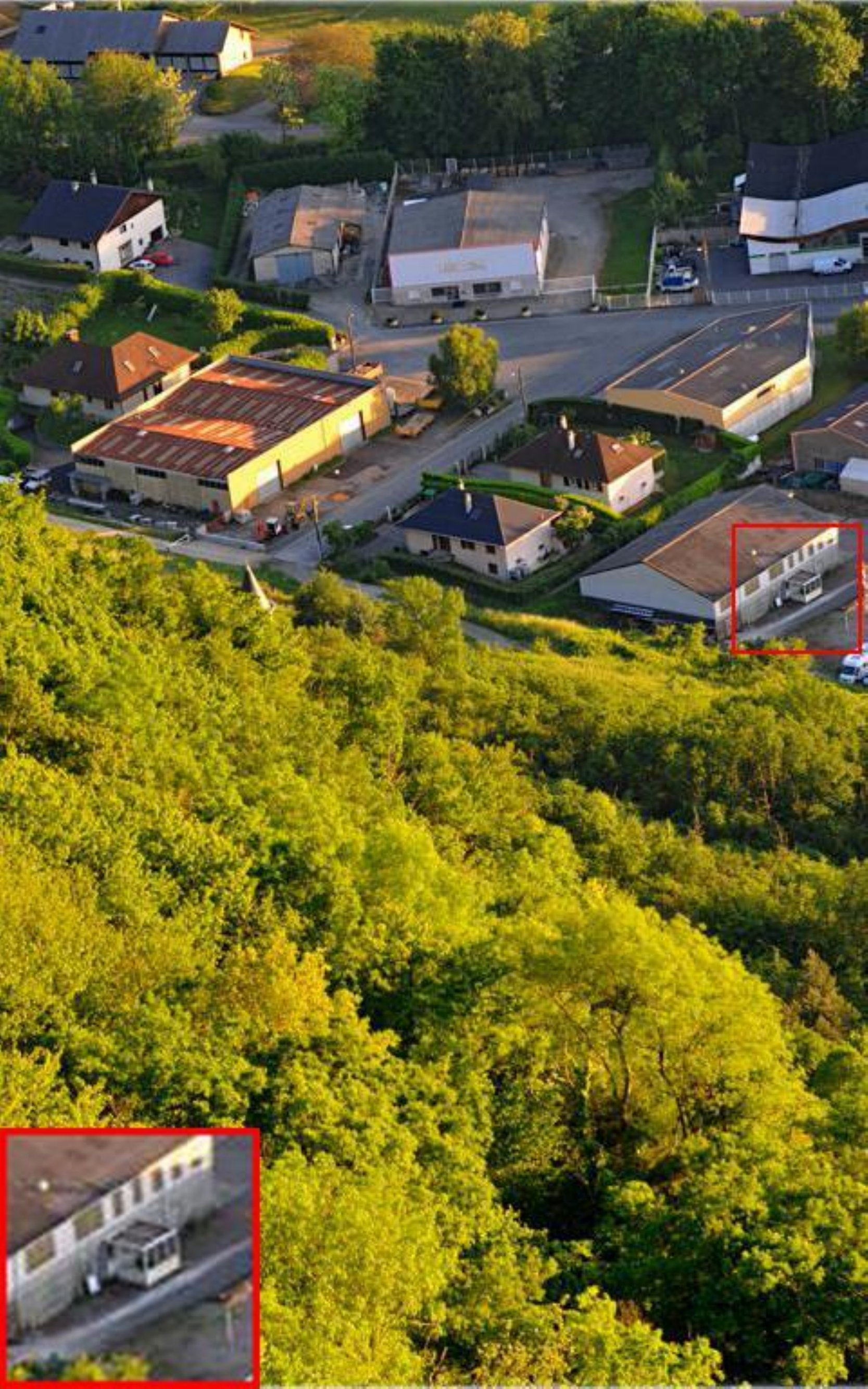}}
		\subfigure{\includegraphics[scale=\m_wid]{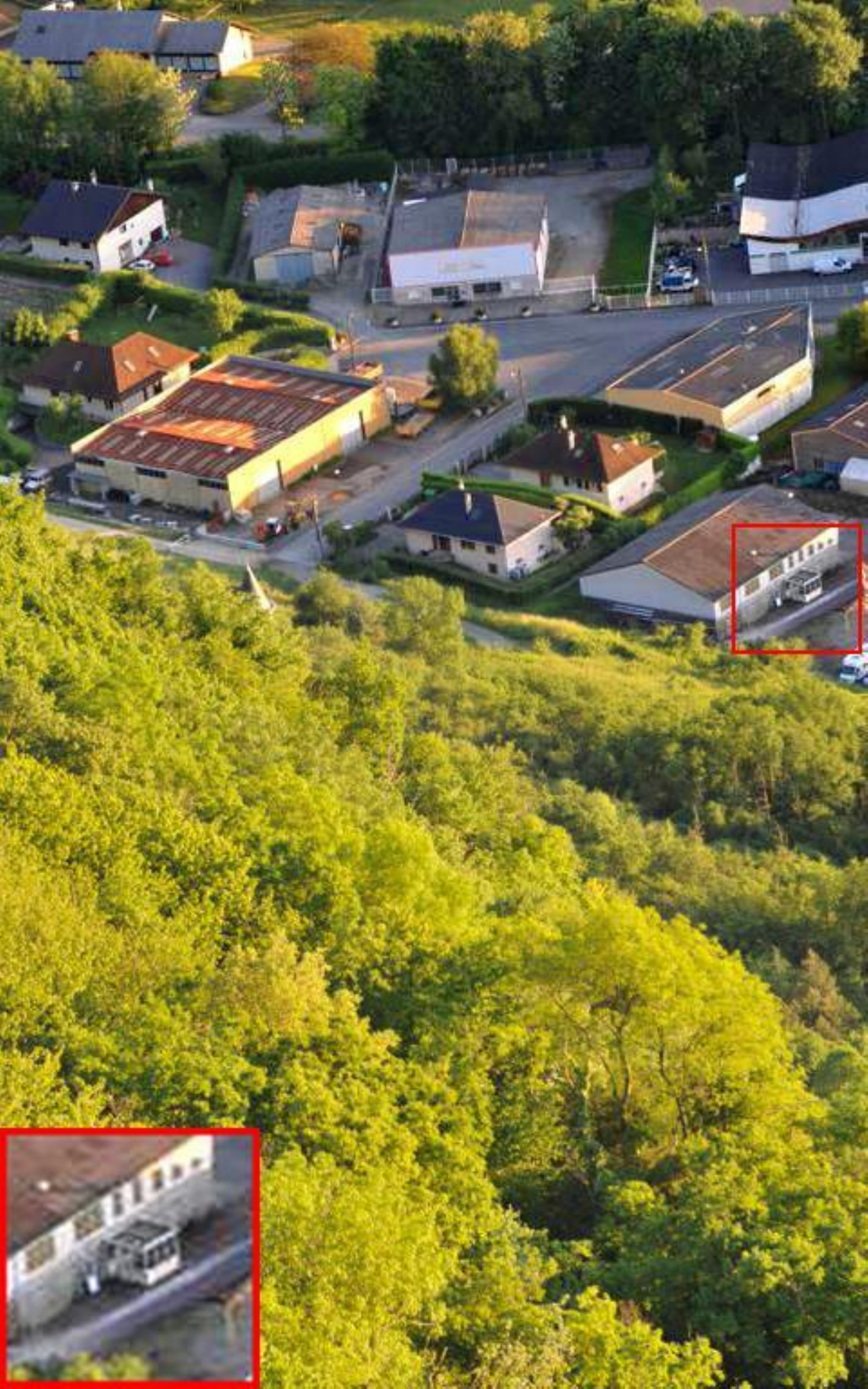}}
	    \subfigure{\includegraphics[scale=\m_wid]{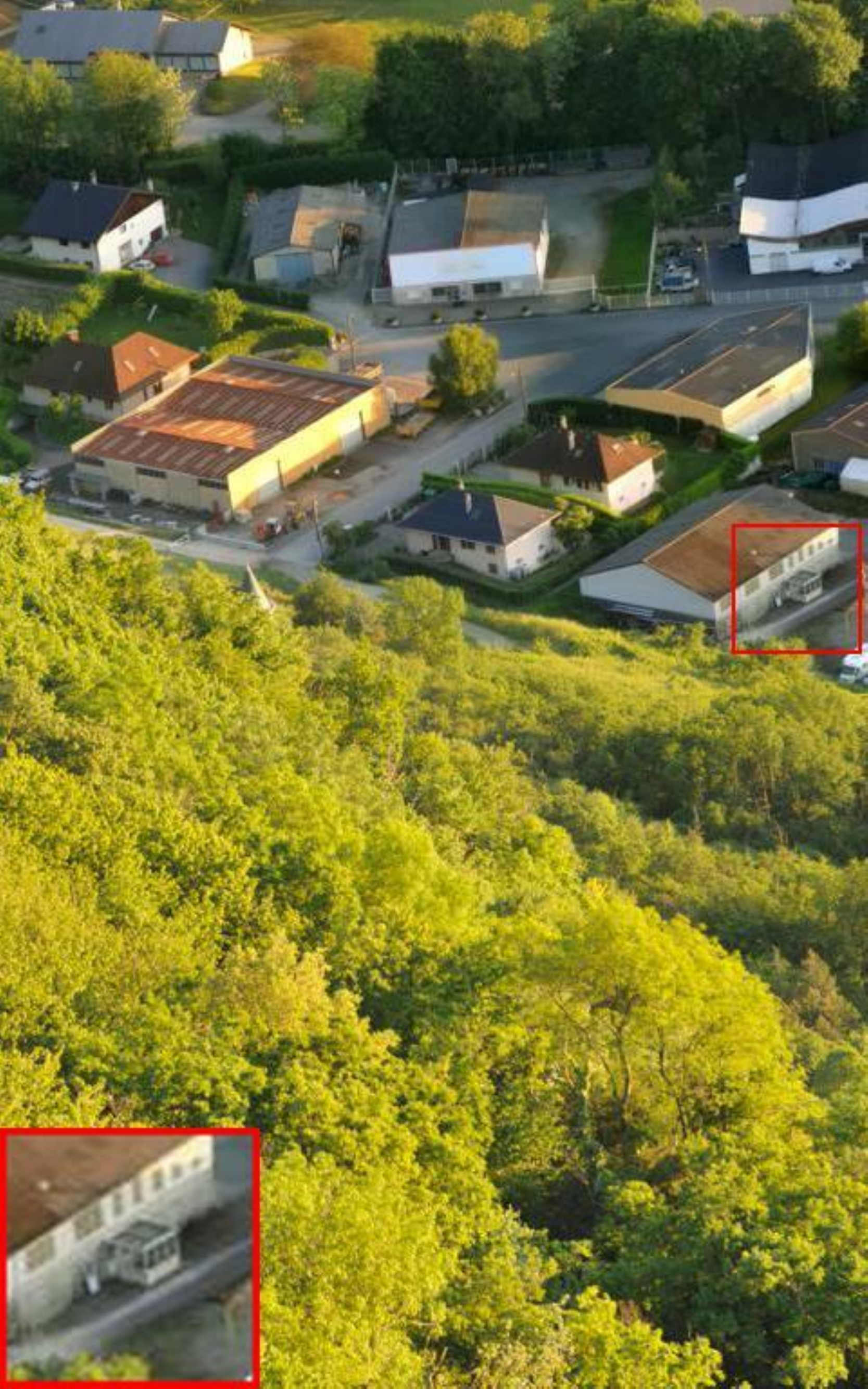}}
		\subfigure{\includegraphics[scale=\m_wid]{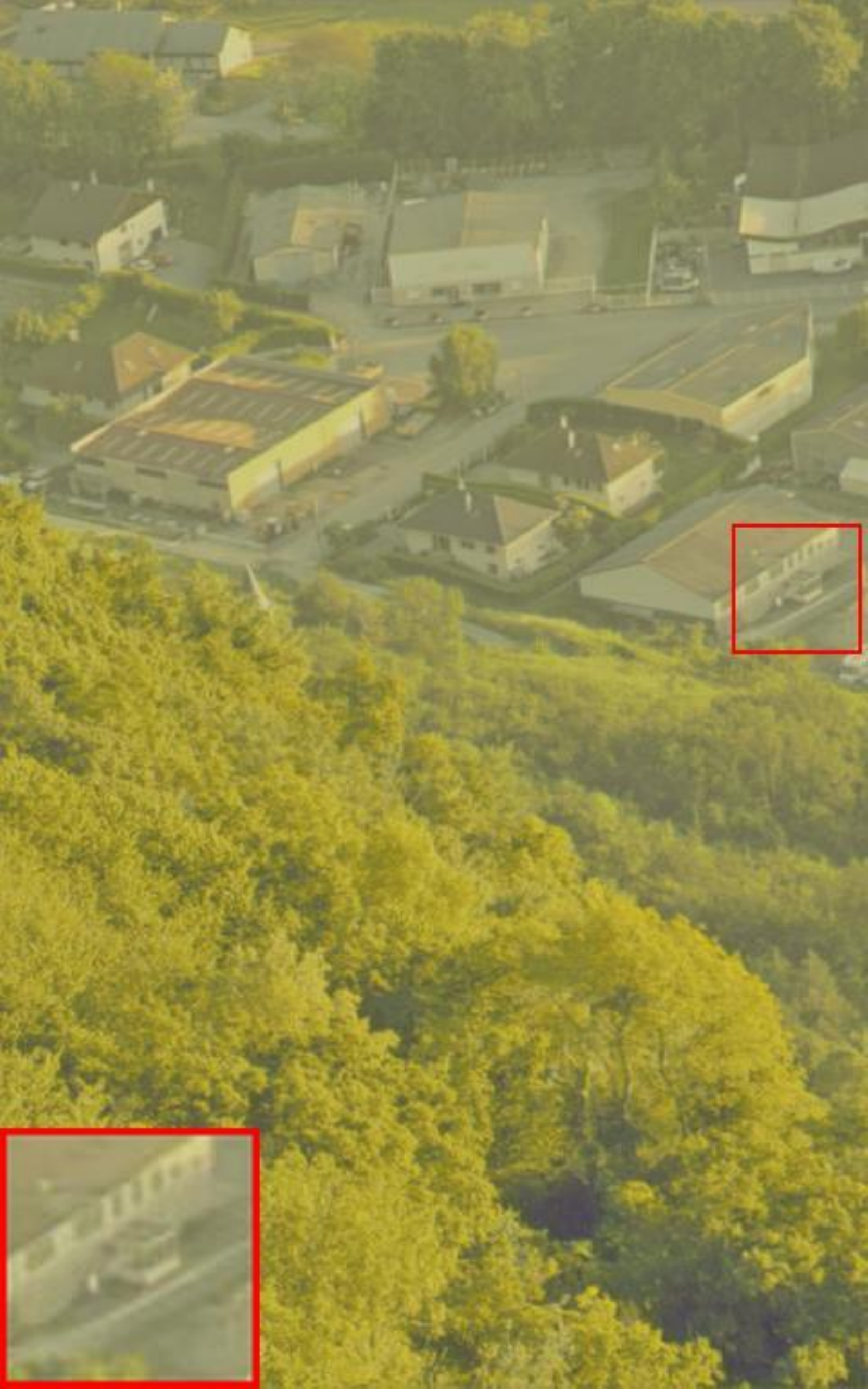}}
		\subfigure{\includegraphics[scale=\m_wid]{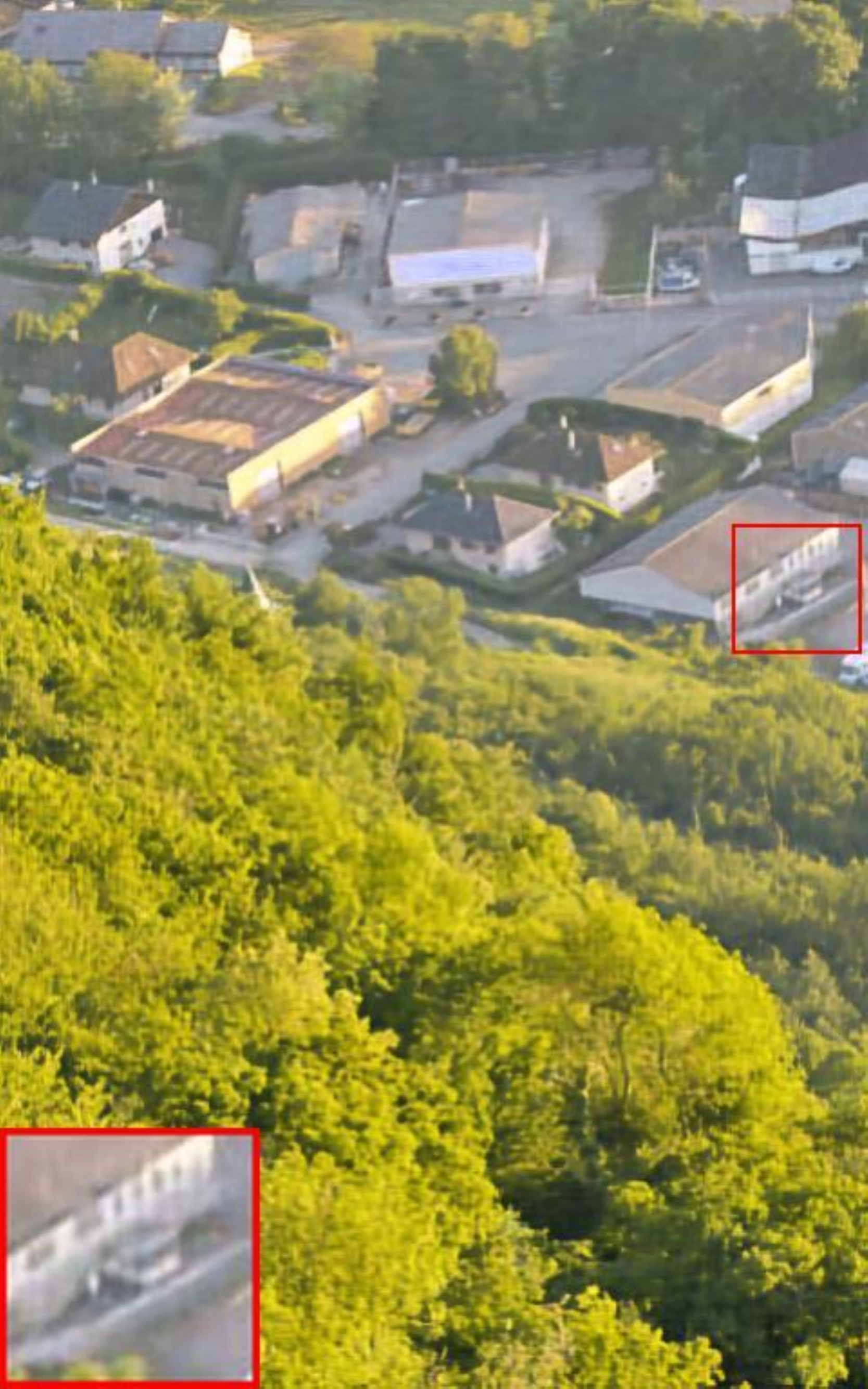}}
		\subfigure{\includegraphics[scale=\m_wid]{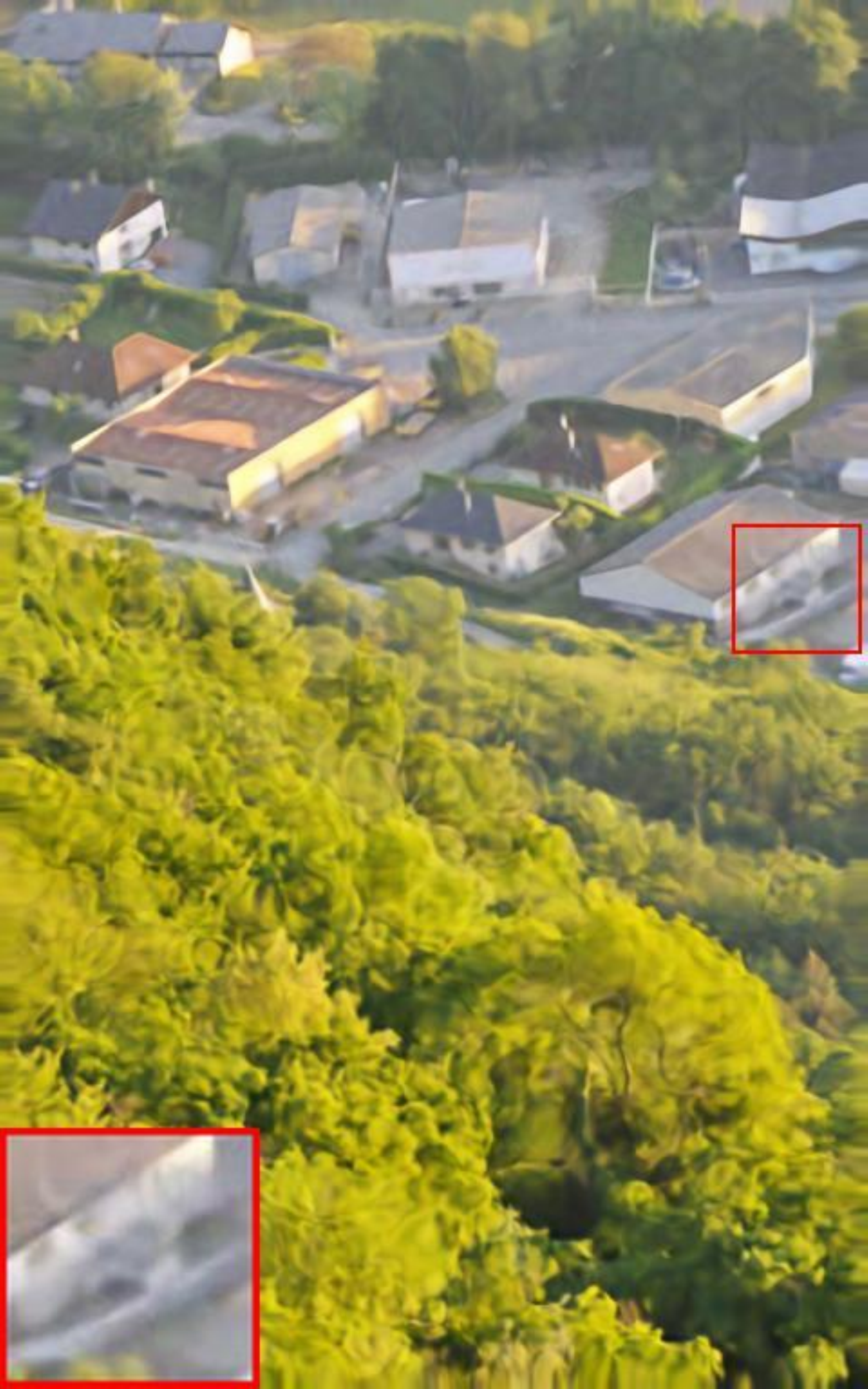}}
		\subfigure{\includegraphics[scale=\m_wid]{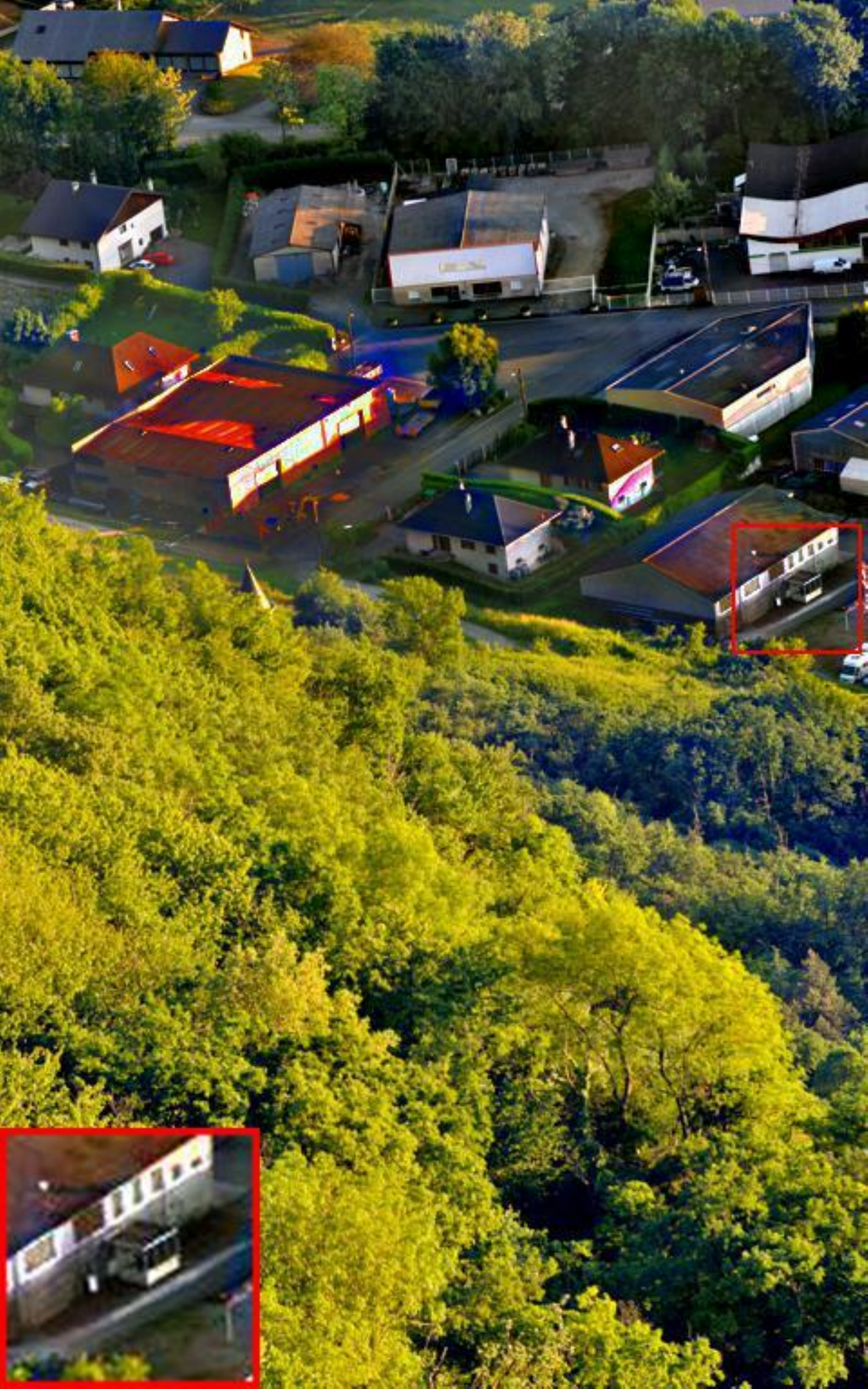}}
		\subfigure{\includegraphics[scale=\m_wid]{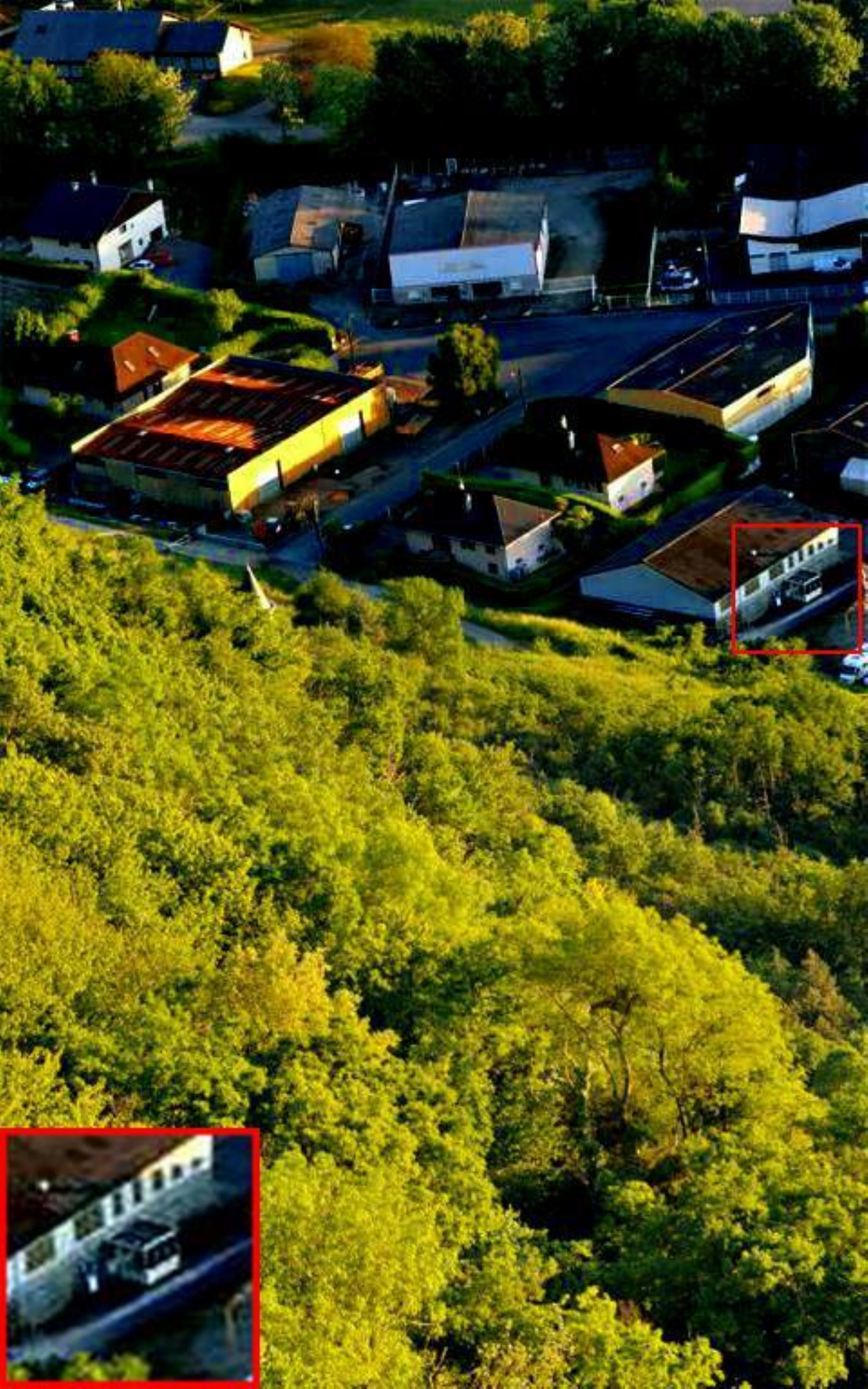}}
		\subfigure{\includegraphics[scale=\m_wid]{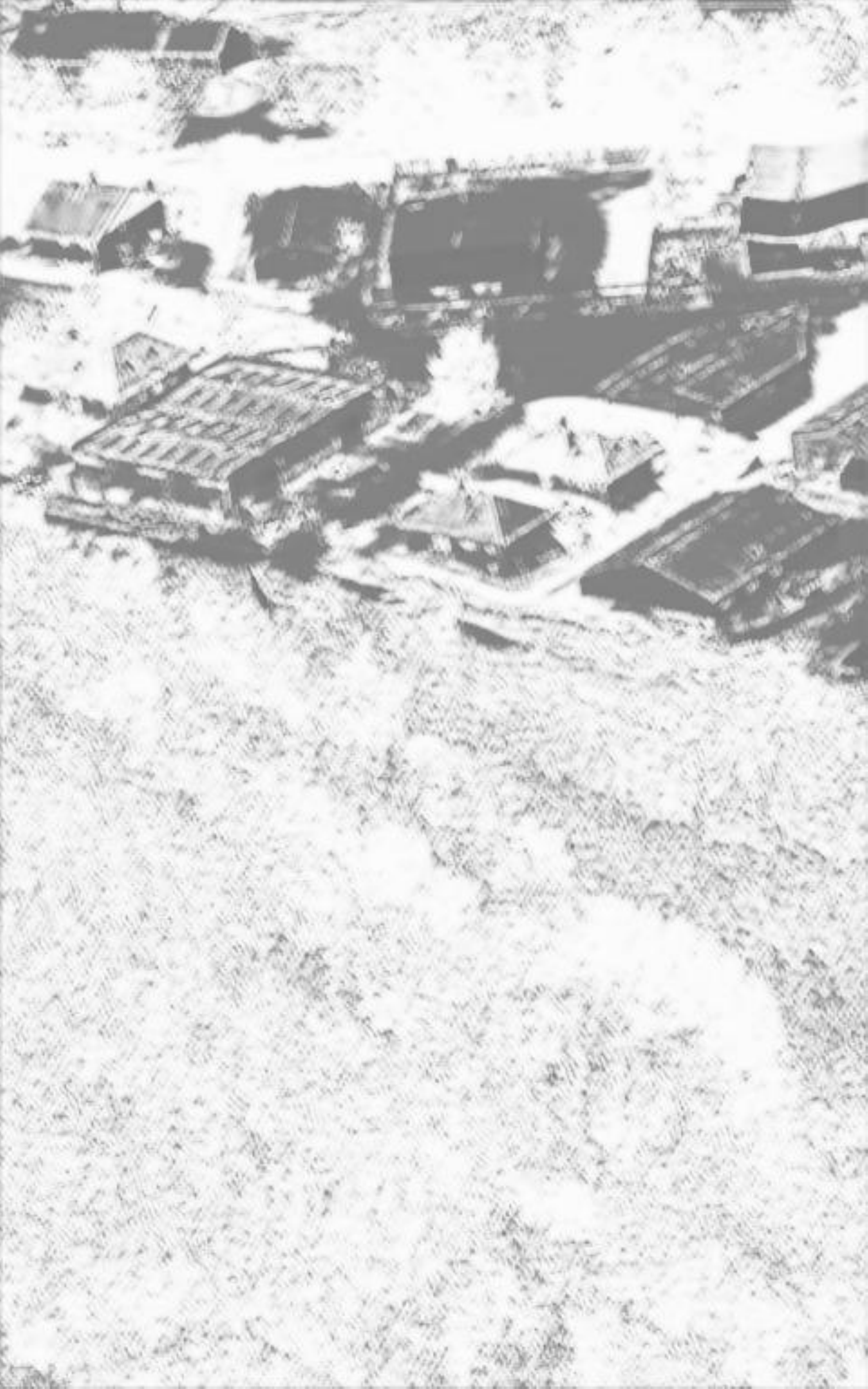}}
	\end{center}
	\vspace{-0.8cm}
	
	\begin{center}
	\def \m_wid{0.0254}
	\def \two_wid{0.0508}
		\subfigure{\includegraphics[scale=\m_wid]{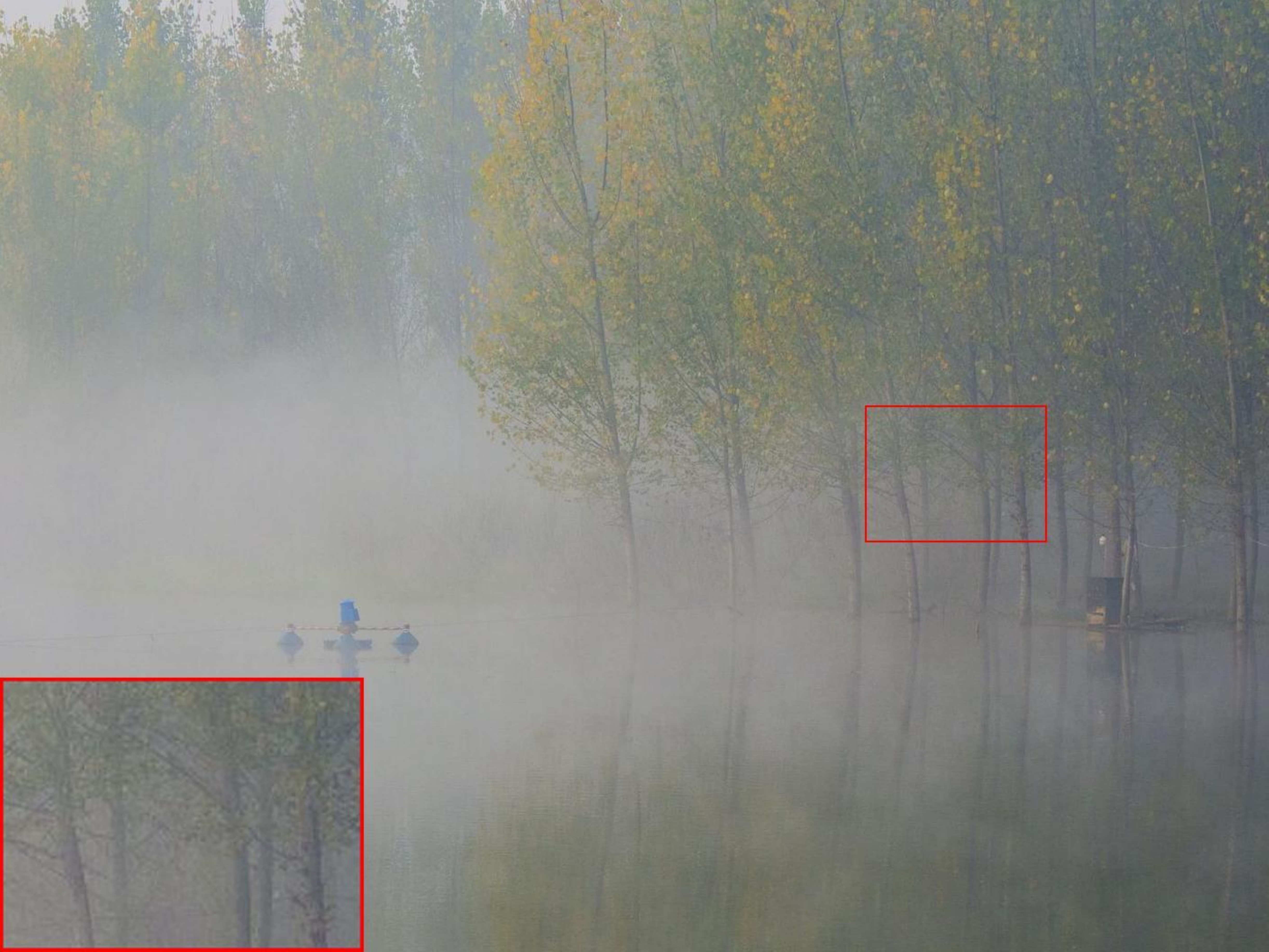}}
		\subfigure{\includegraphics[scale=\m_wid]{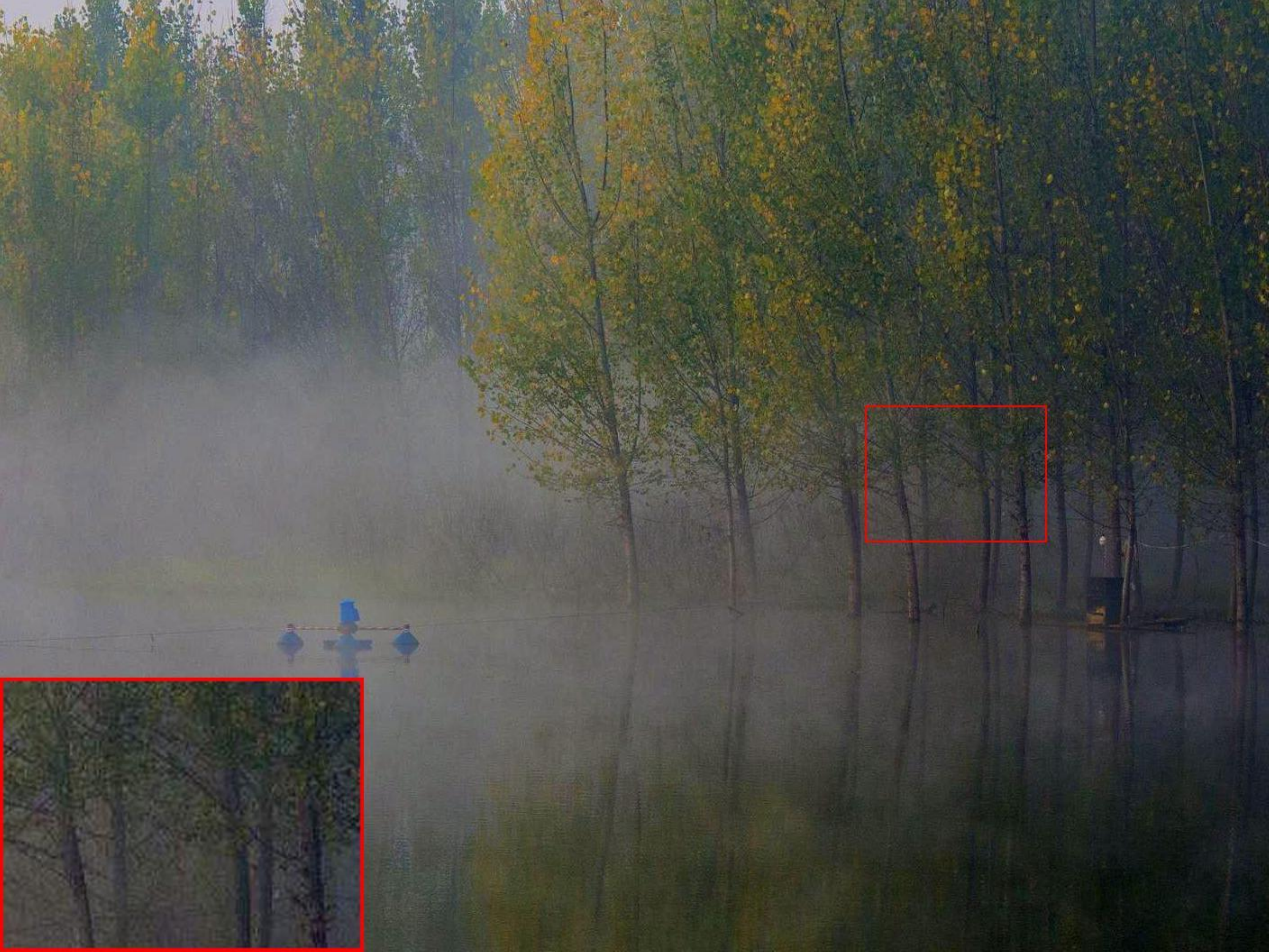}}
		\subfigure{\includegraphics[scale=\m_wid]{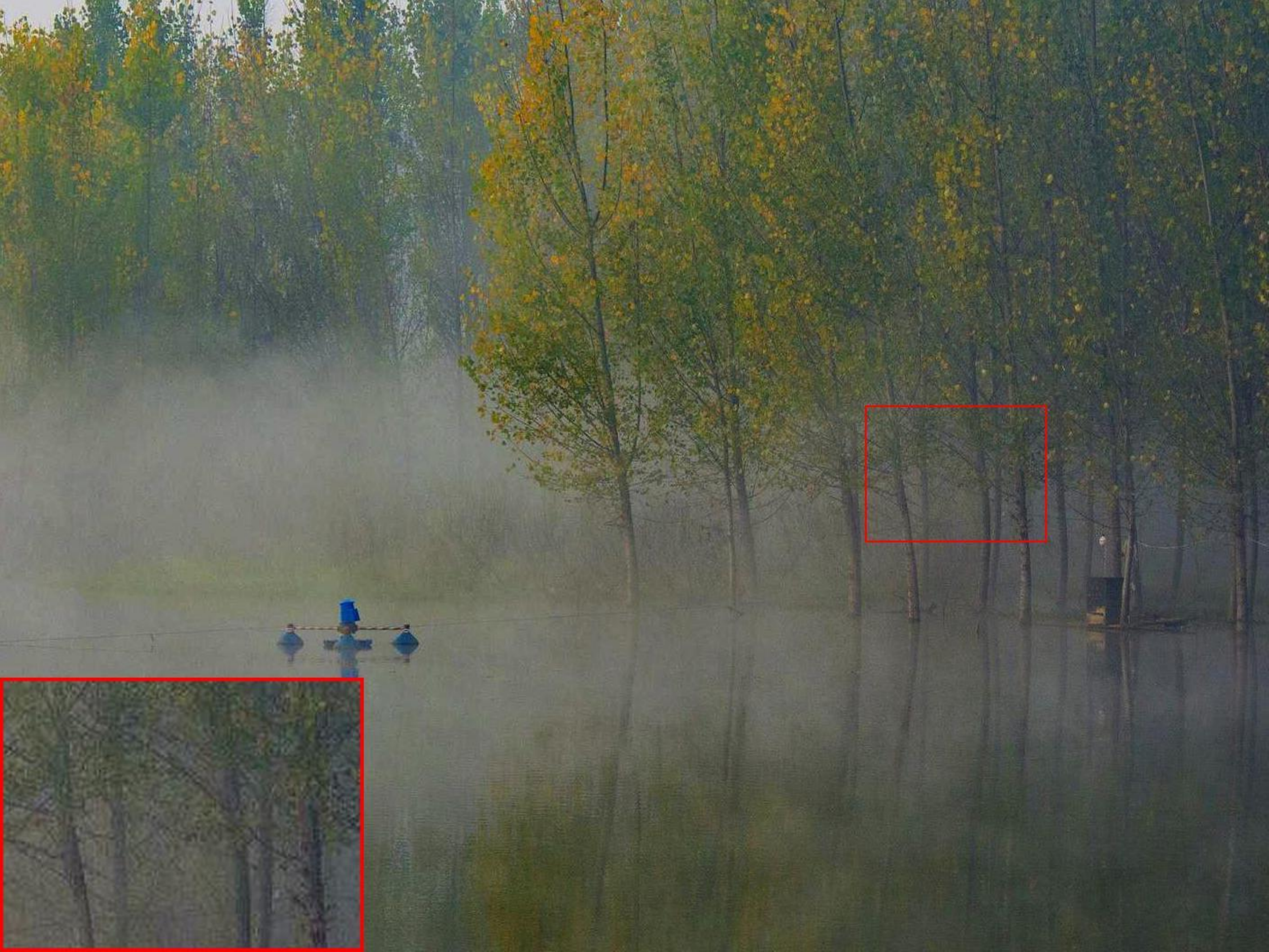}}
		\subfigure{\includegraphics[scale=\m_wid]{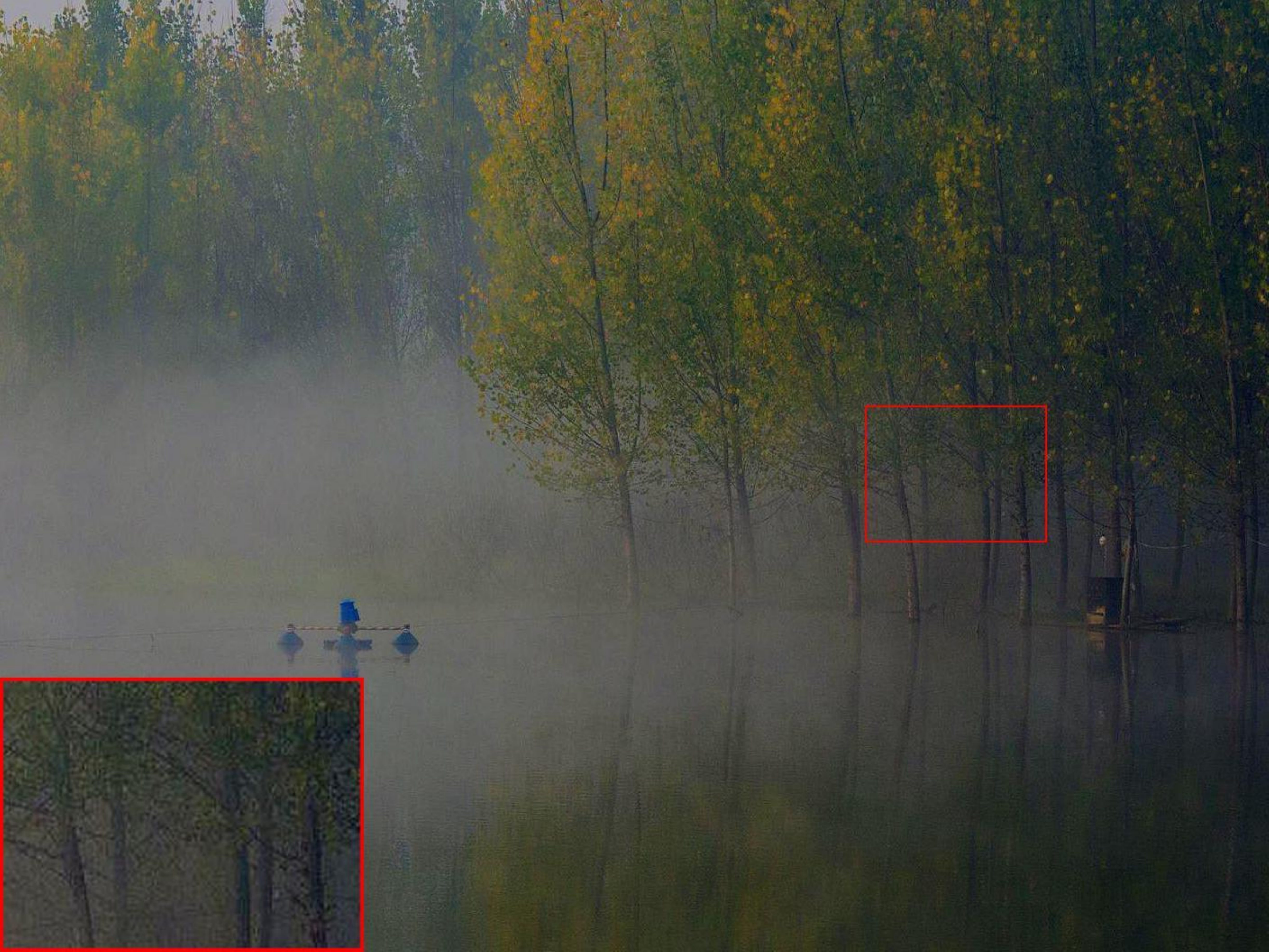}}
		\subfigure{\includegraphics[scale=\m_wid]{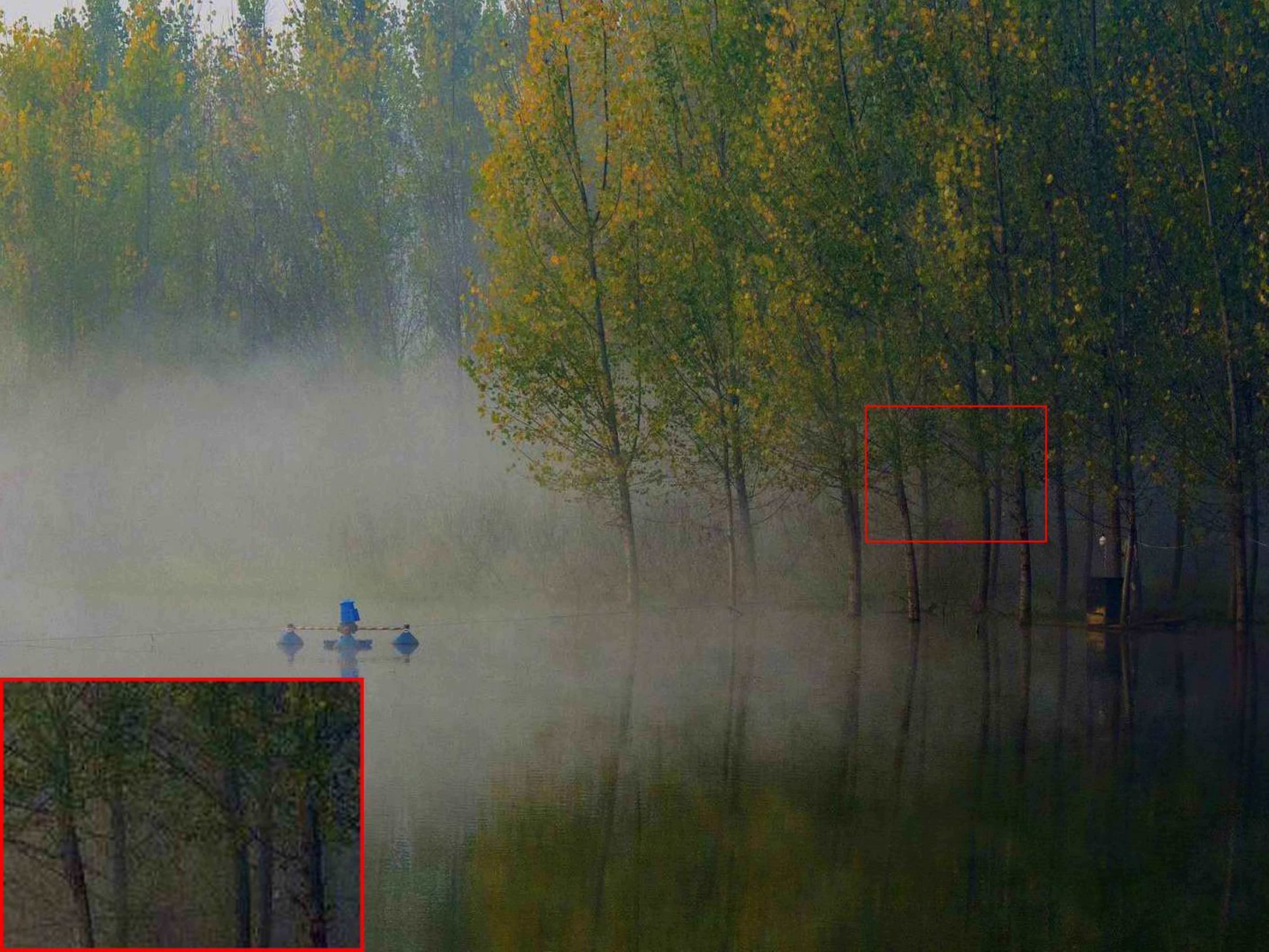}}
		\subfigure{\includegraphics[scale=\m_wid]{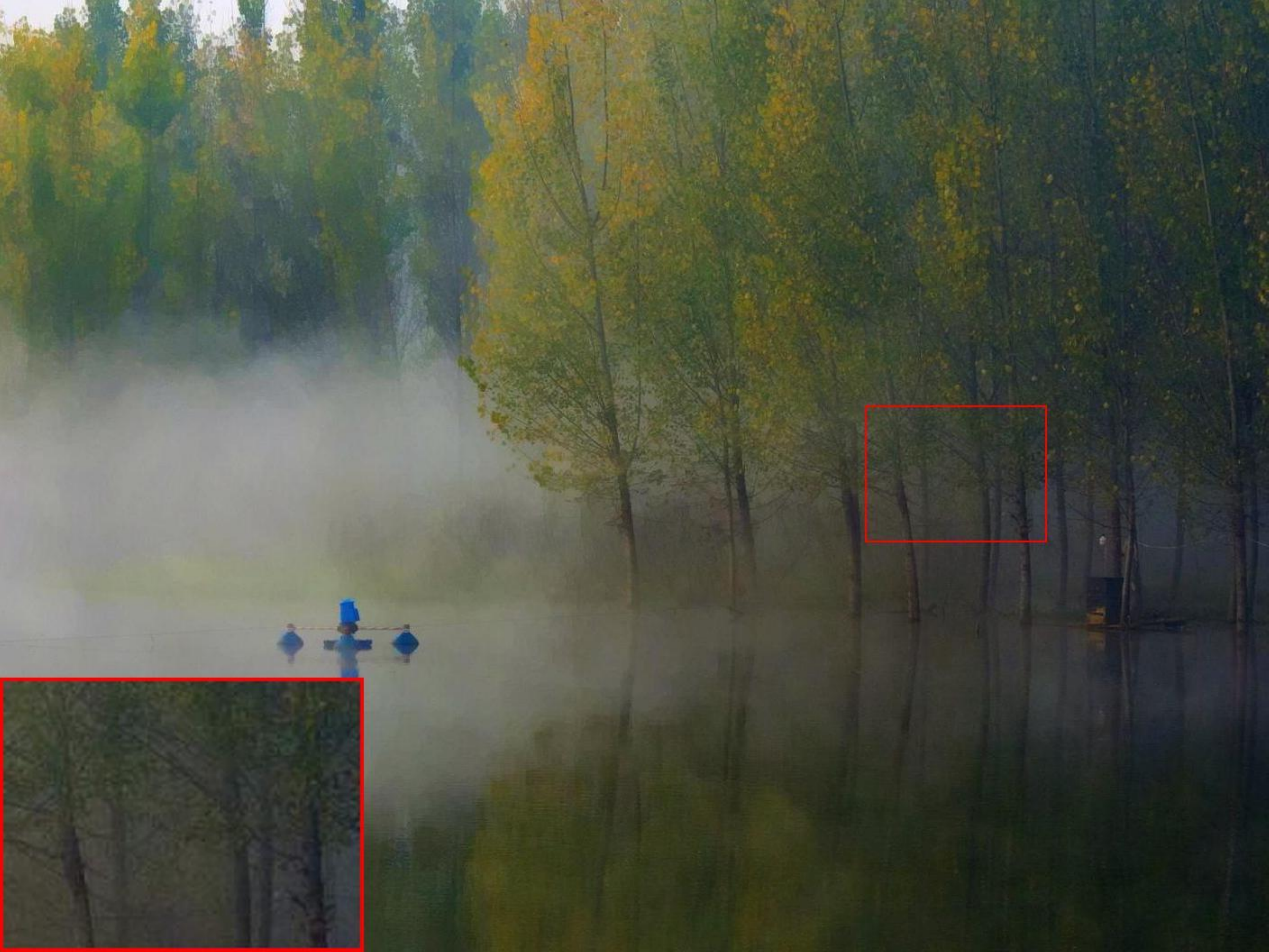}}
		\subfigure{\includegraphics[scale=\m_wid]{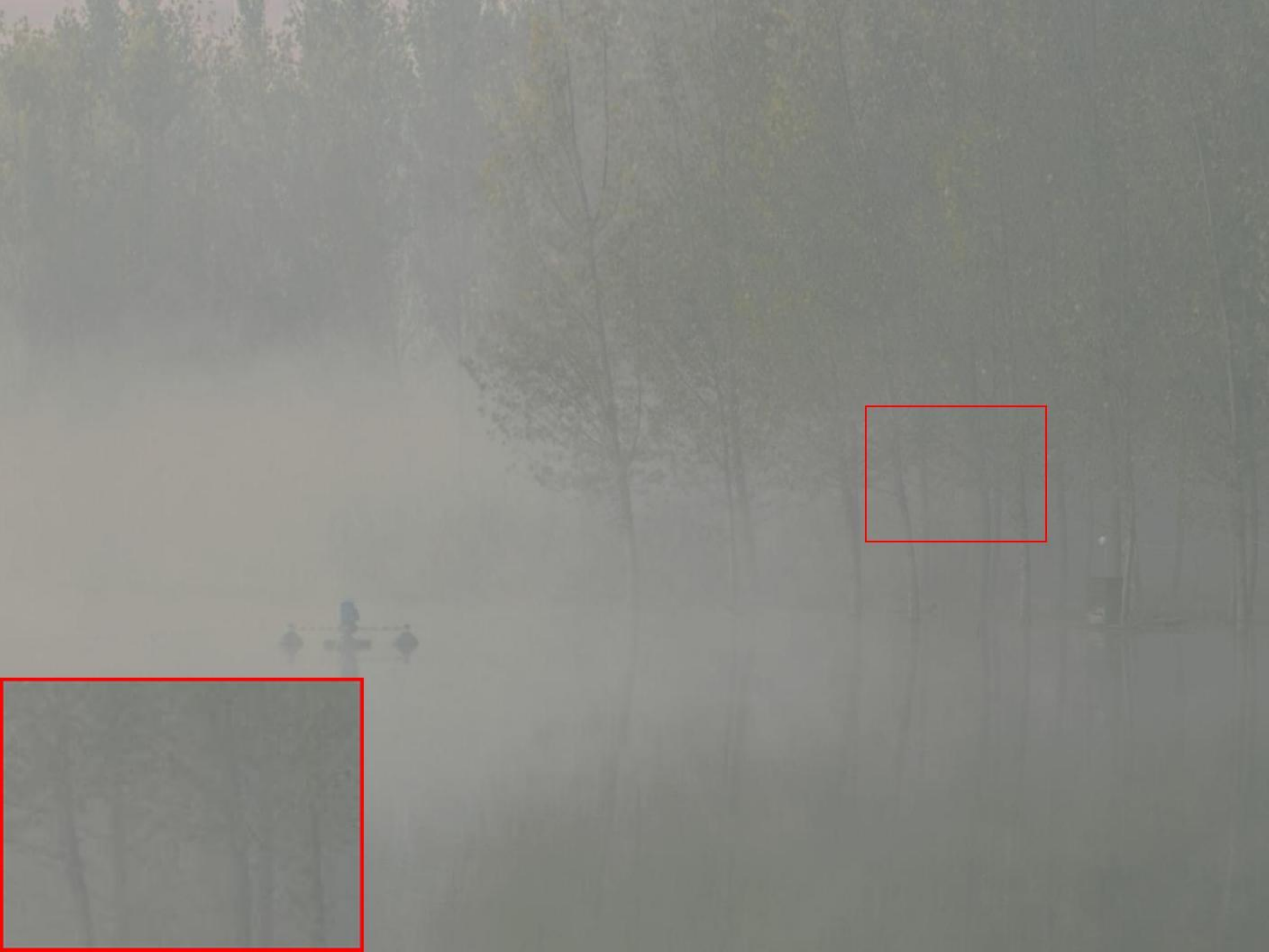}}
		\subfigure{\includegraphics[scale=\m_wid]{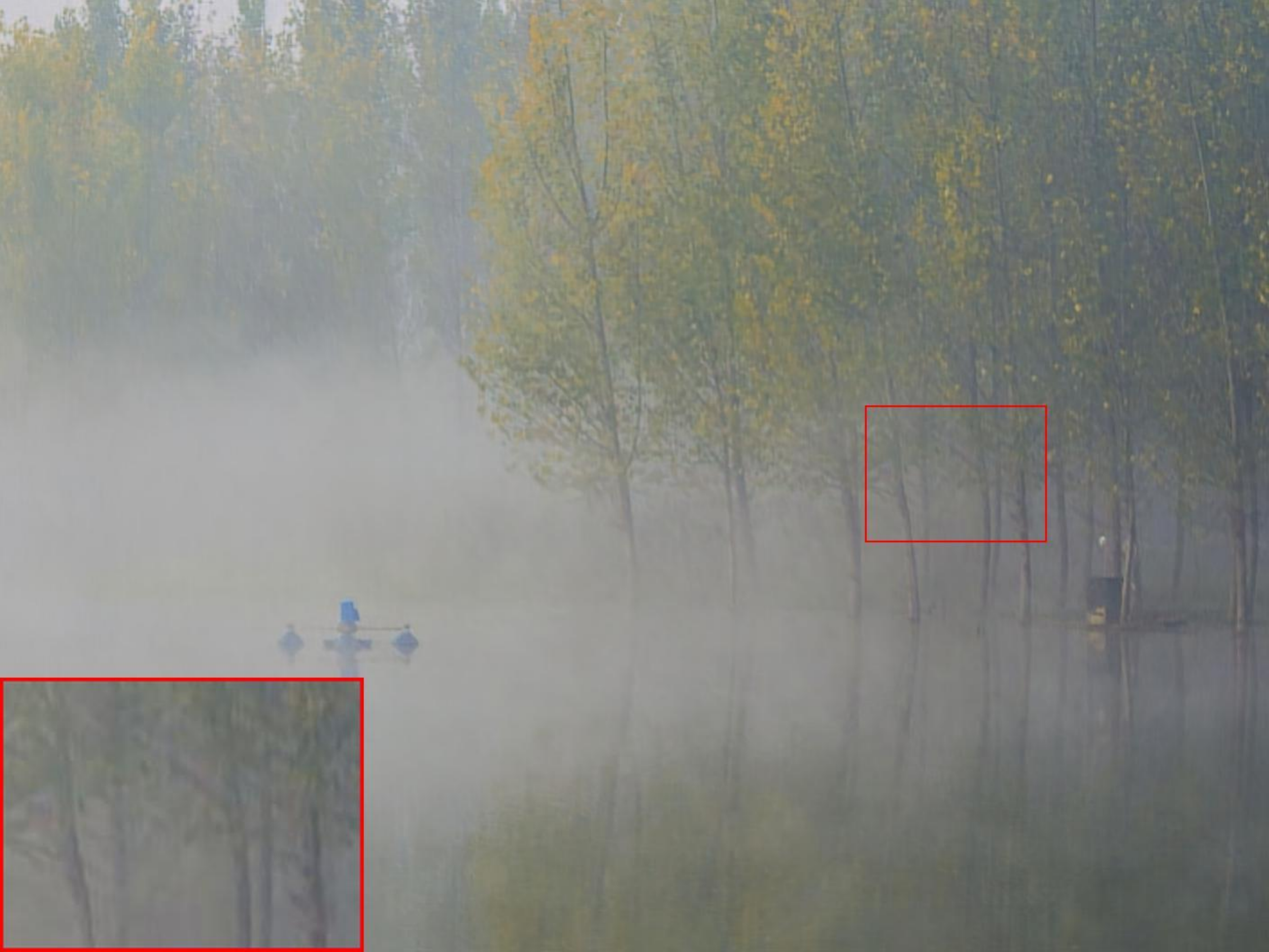}}
		\subfigure{\includegraphics[scale=\two_wid]{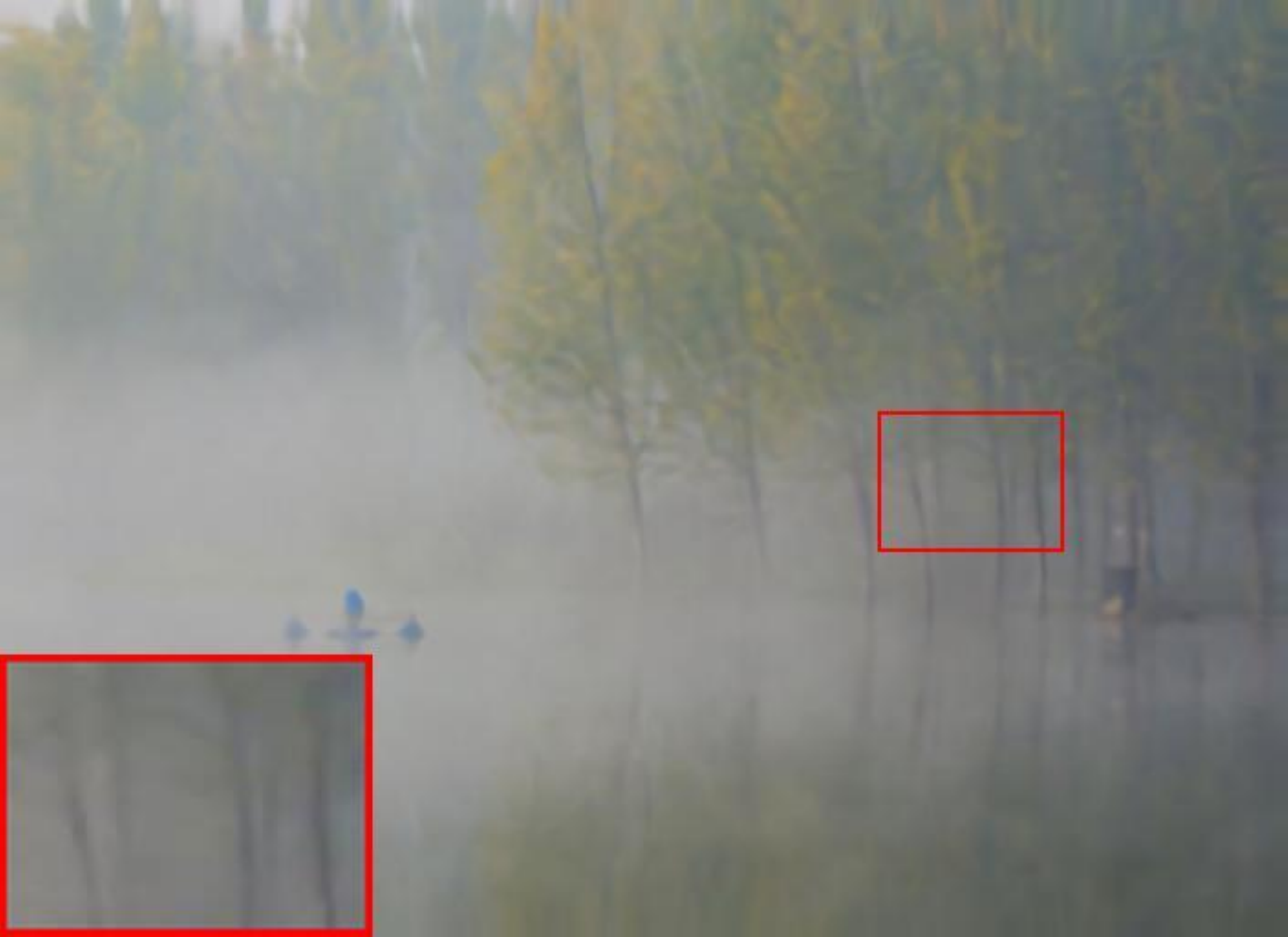}}
		\subfigure{\includegraphics[scale=\m_wid]{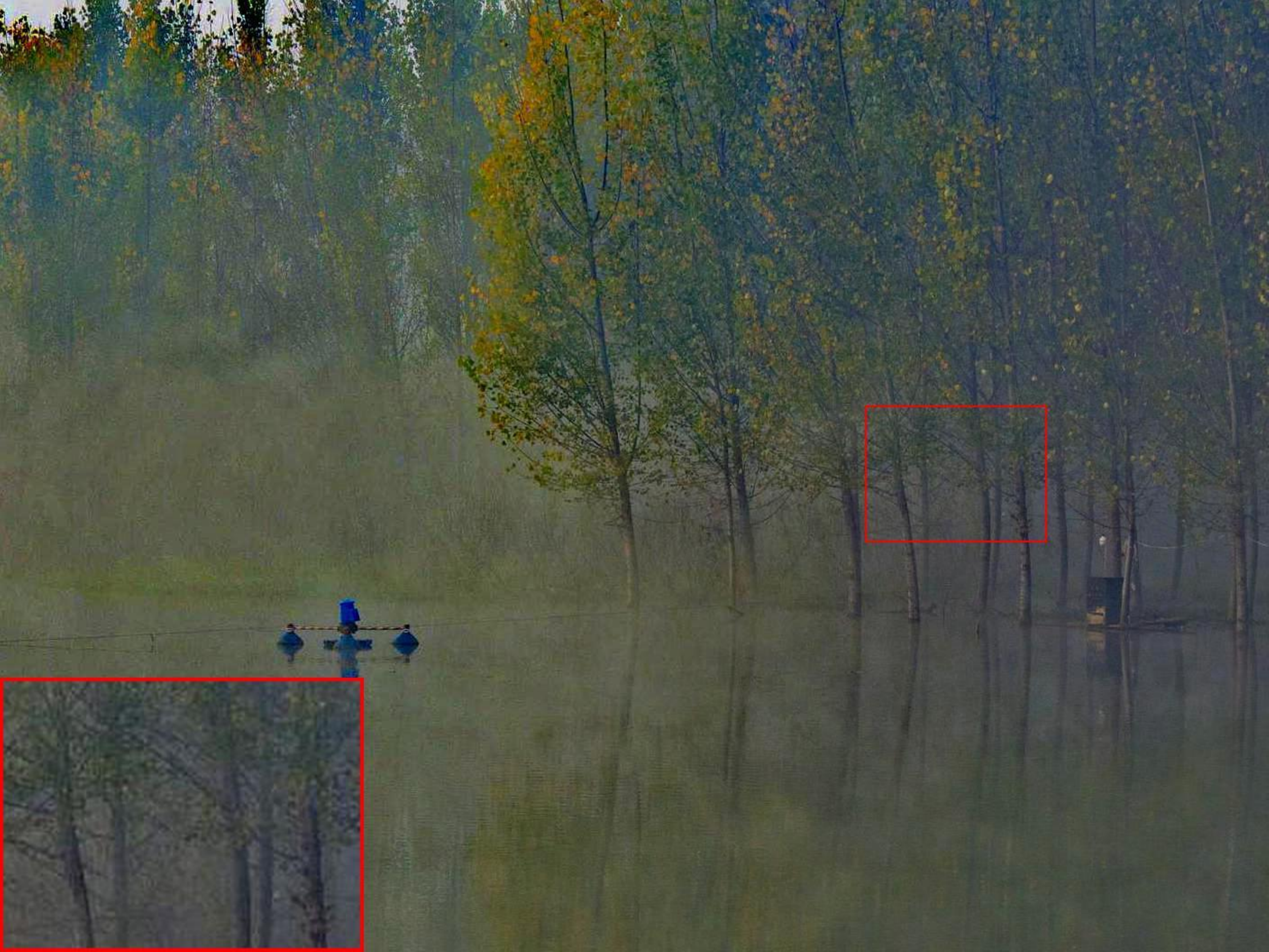}}
		\subfigure{\includegraphics[scale=\two_wid]{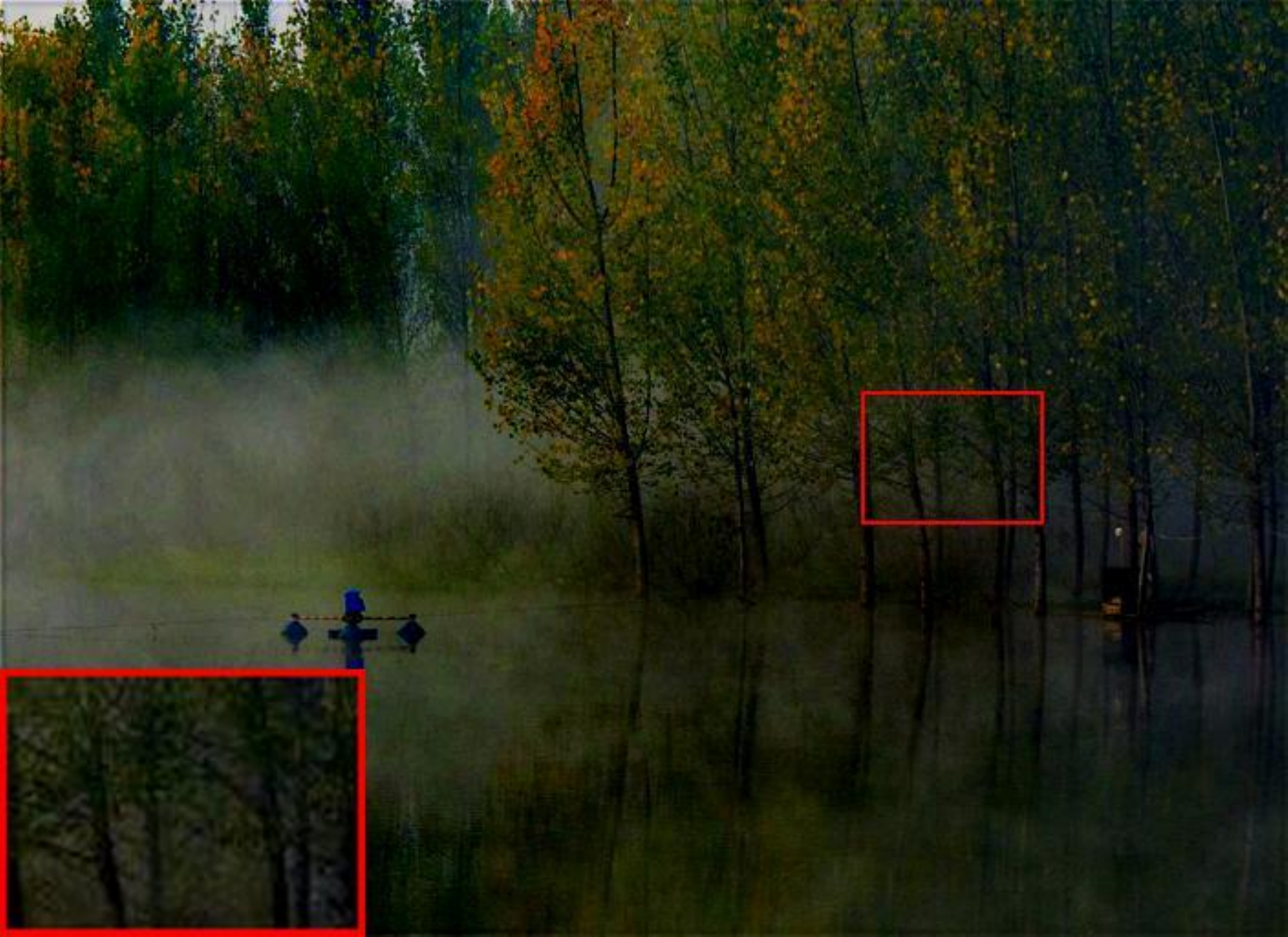}}
		\subfigure{\includegraphics[scale=\two_wid]{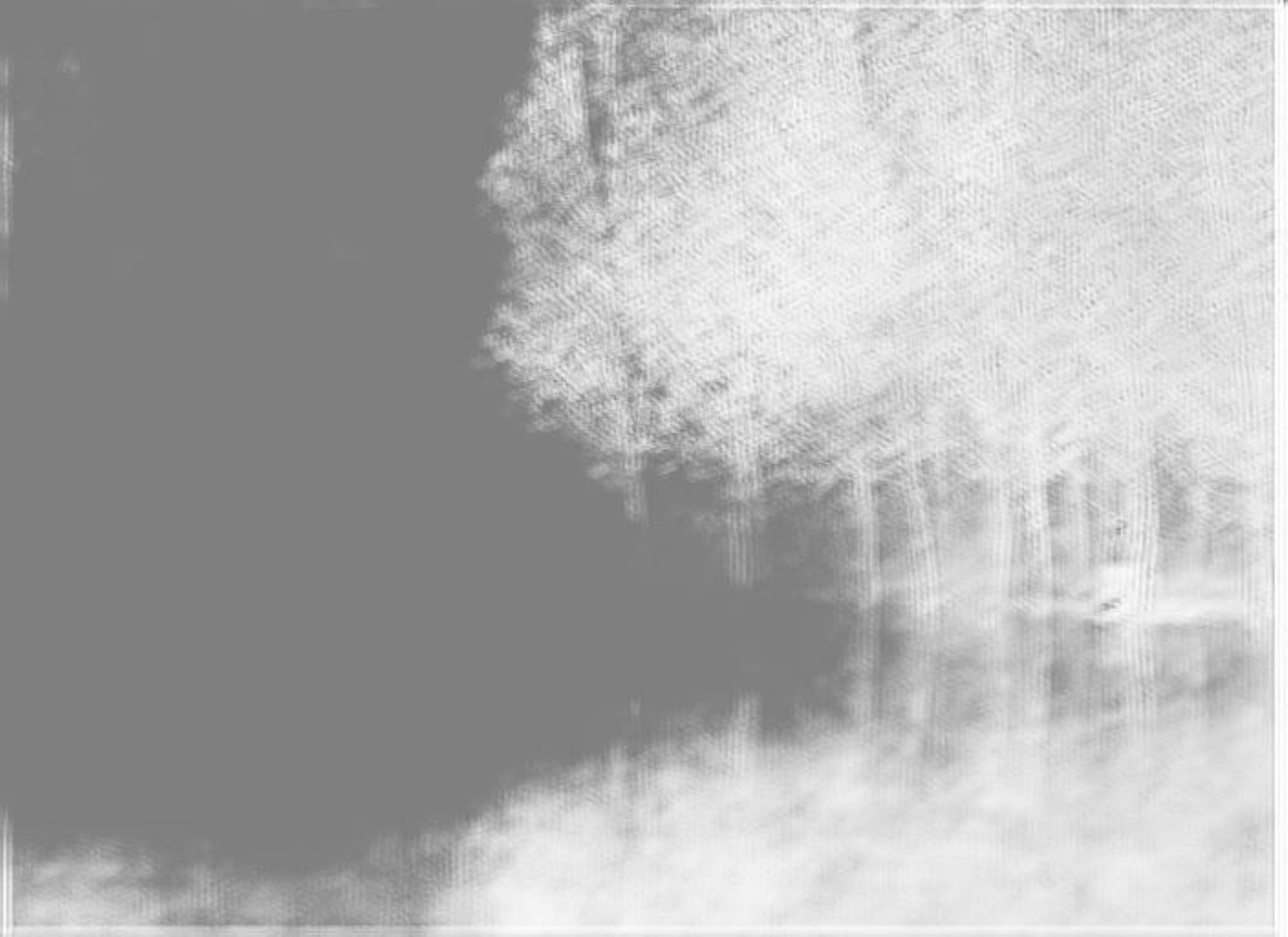}}
	\end{center}
	\vspace{-0.8cm}
	
	\begin{center}
	\def \m_wid{0.0558}
		\subfigure{\includegraphics[scale=\m_wid]{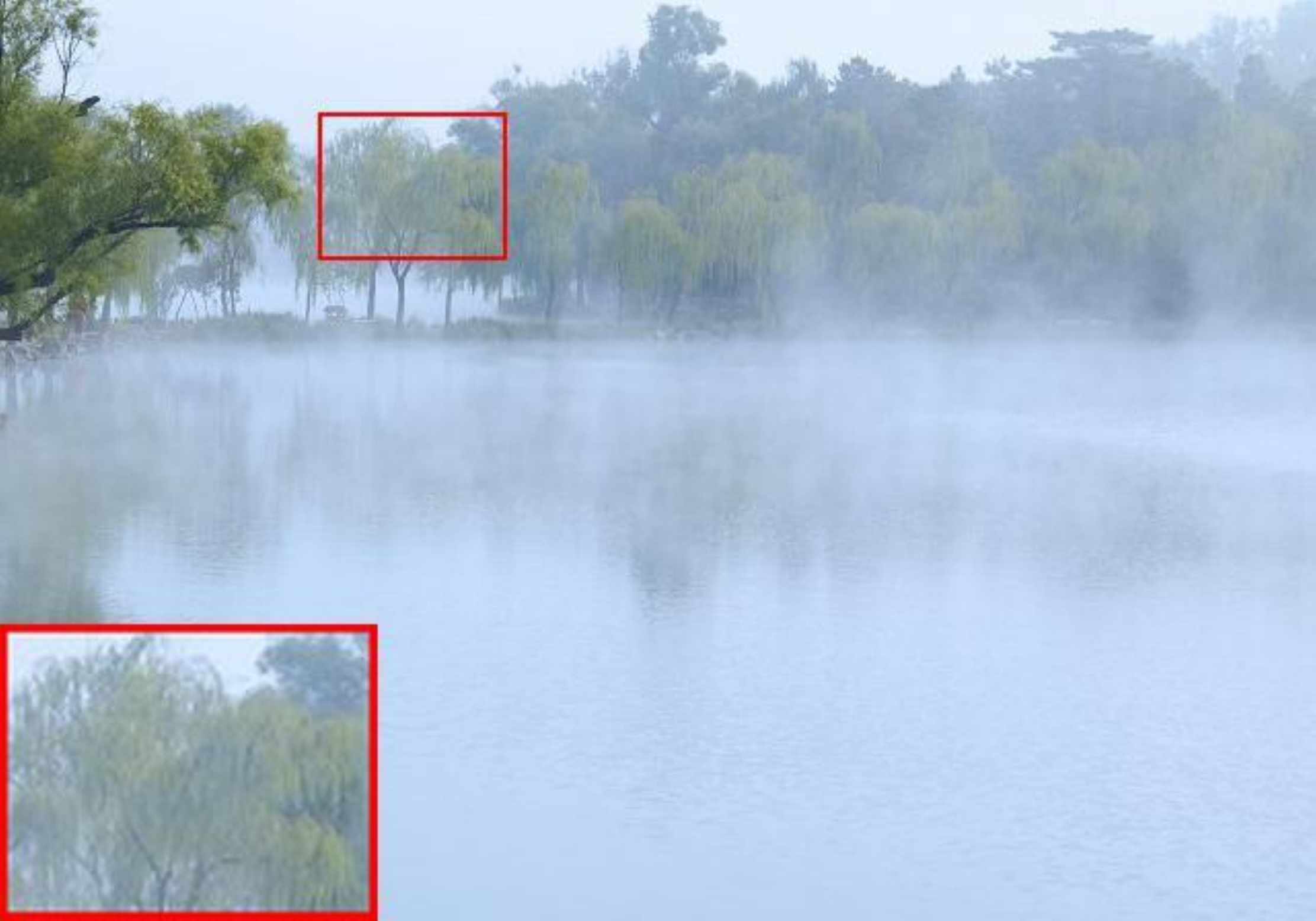}}
		\subfigure{\includegraphics[scale=\m_wid]{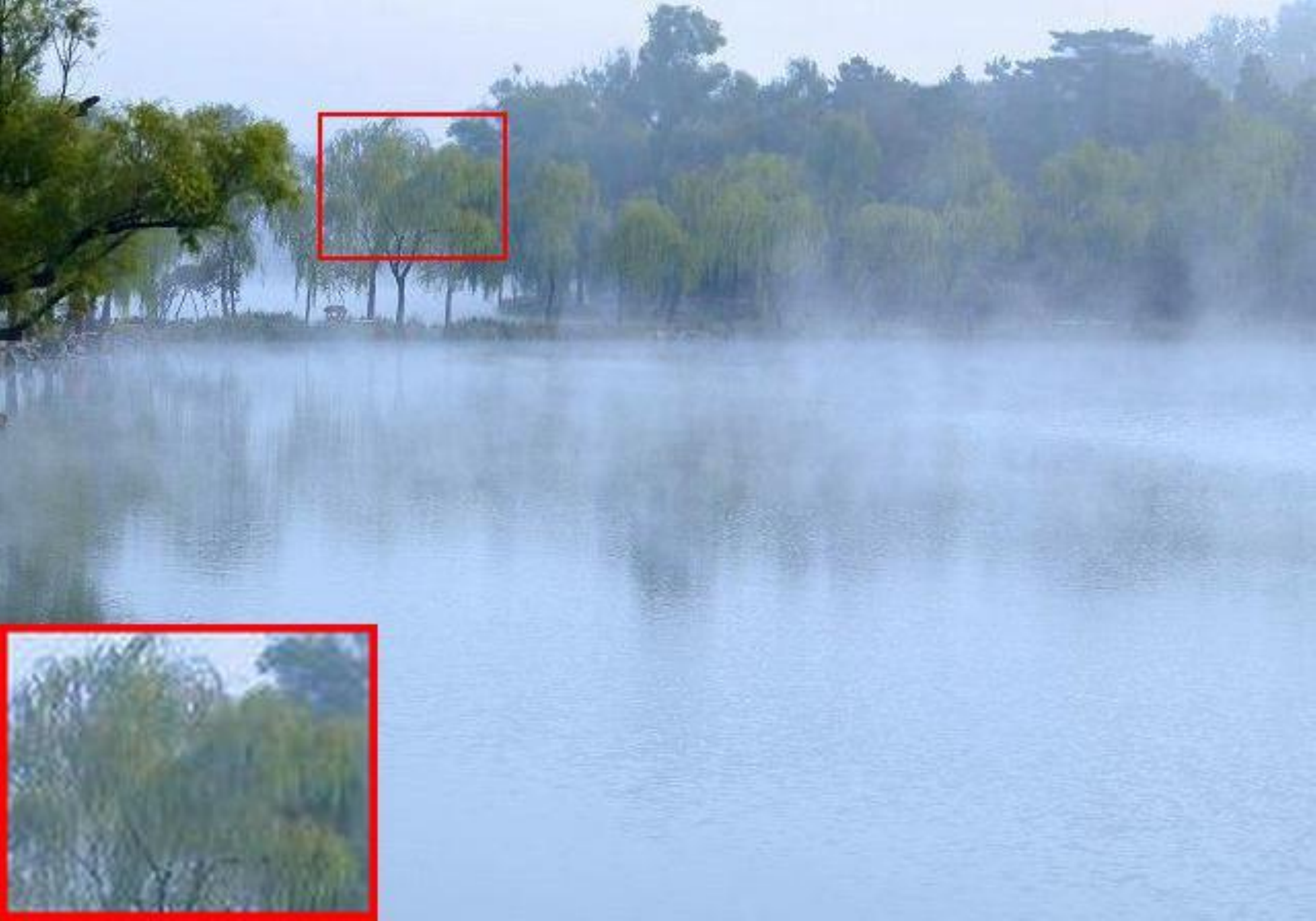}}
		\subfigure{\includegraphics[scale=\m_wid]{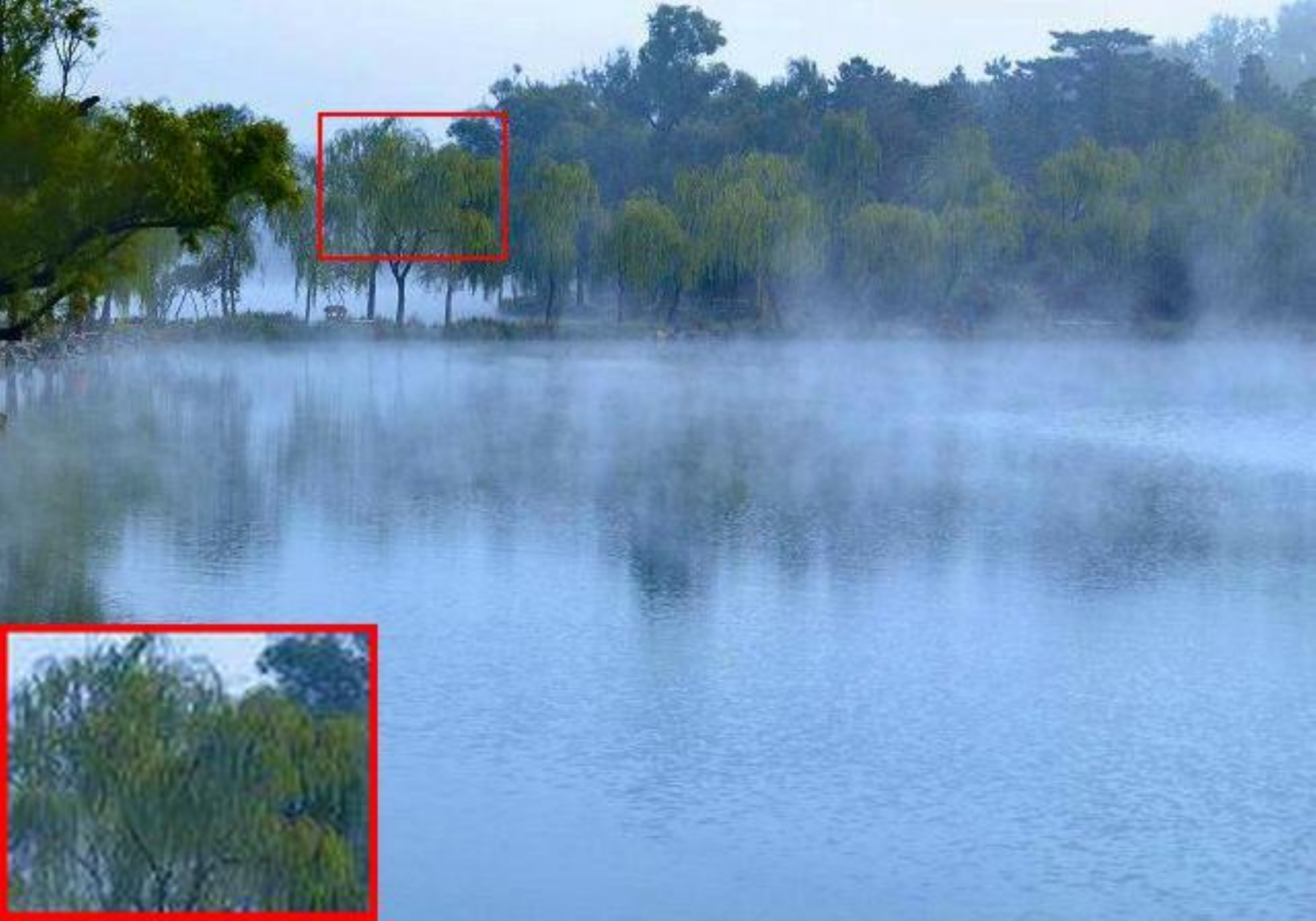}}
		\subfigure{\includegraphics[scale=\m_wid]{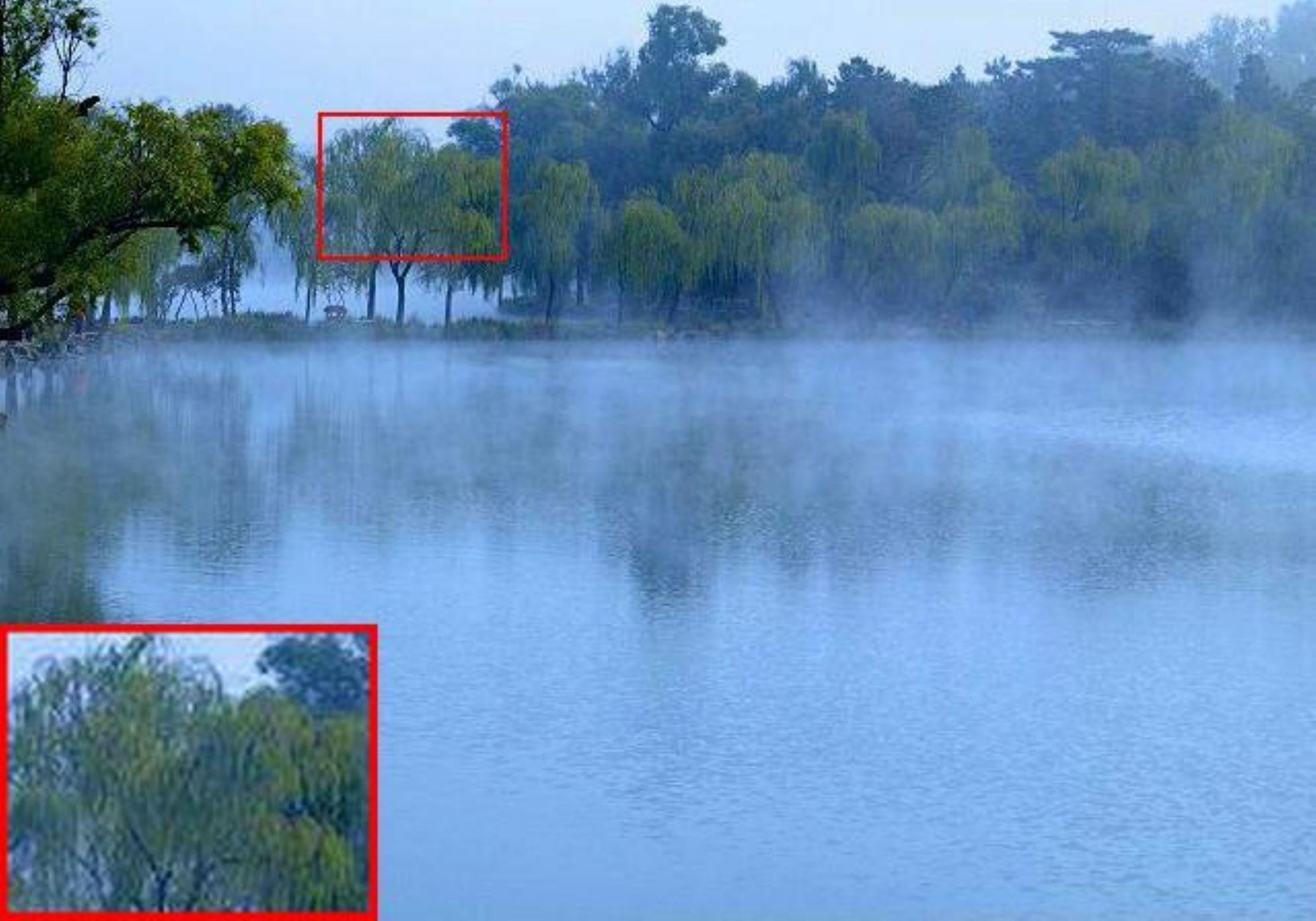}}
		\subfigure{\includegraphics[scale=\m_wid]{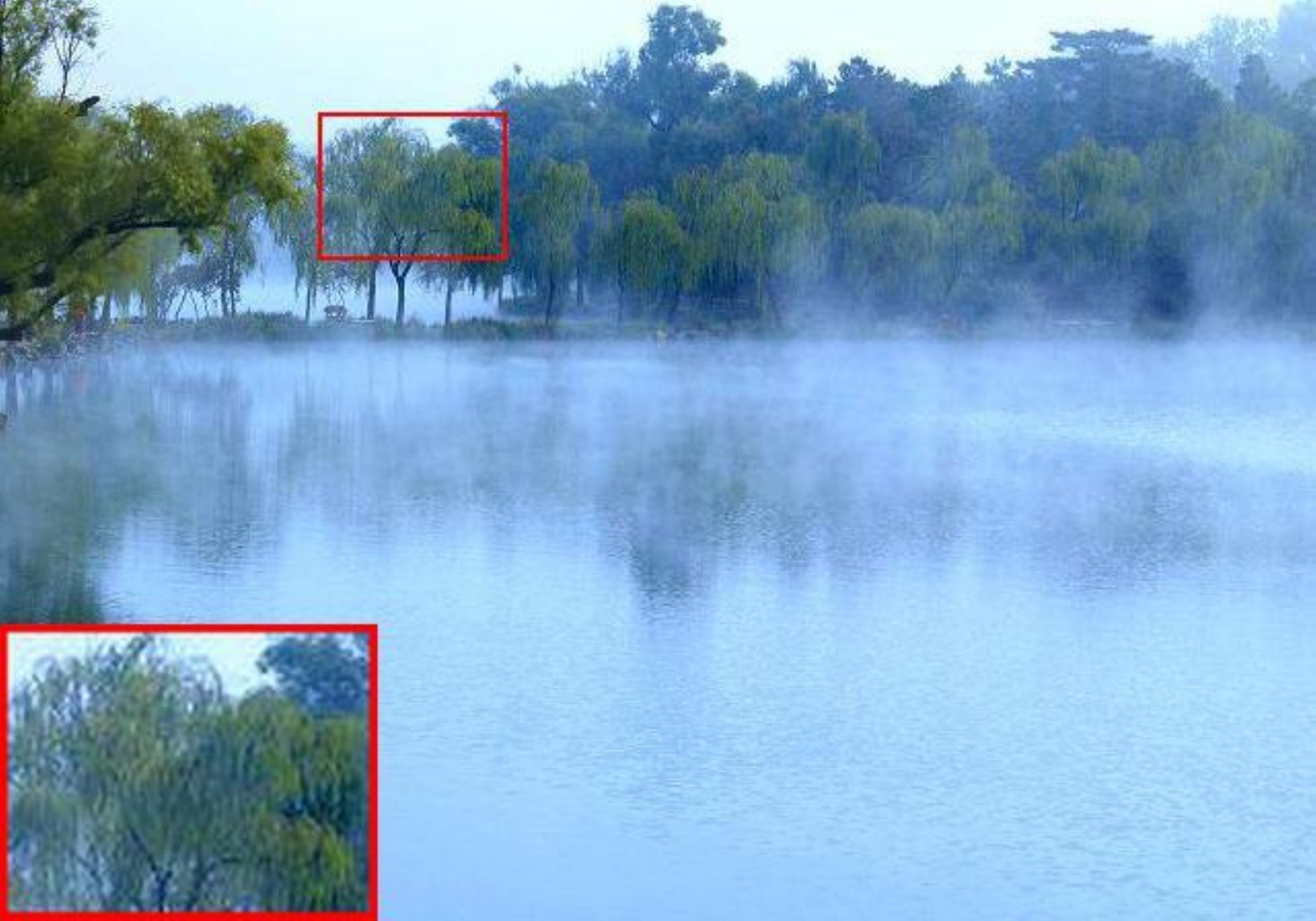}}
		\subfigure{\includegraphics[scale=\m_wid]{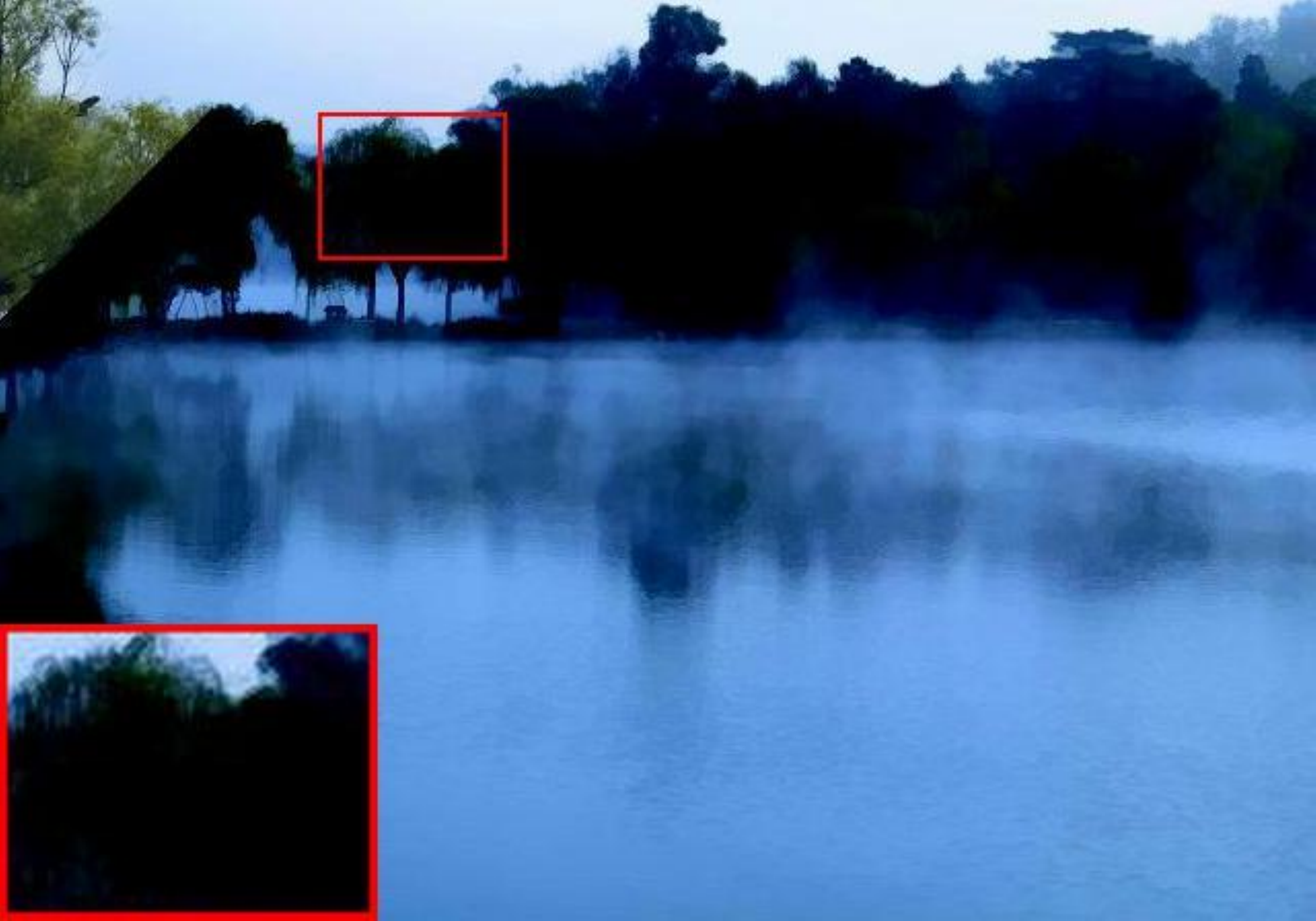}}
		\subfigure{\includegraphics[scale=\m_wid]{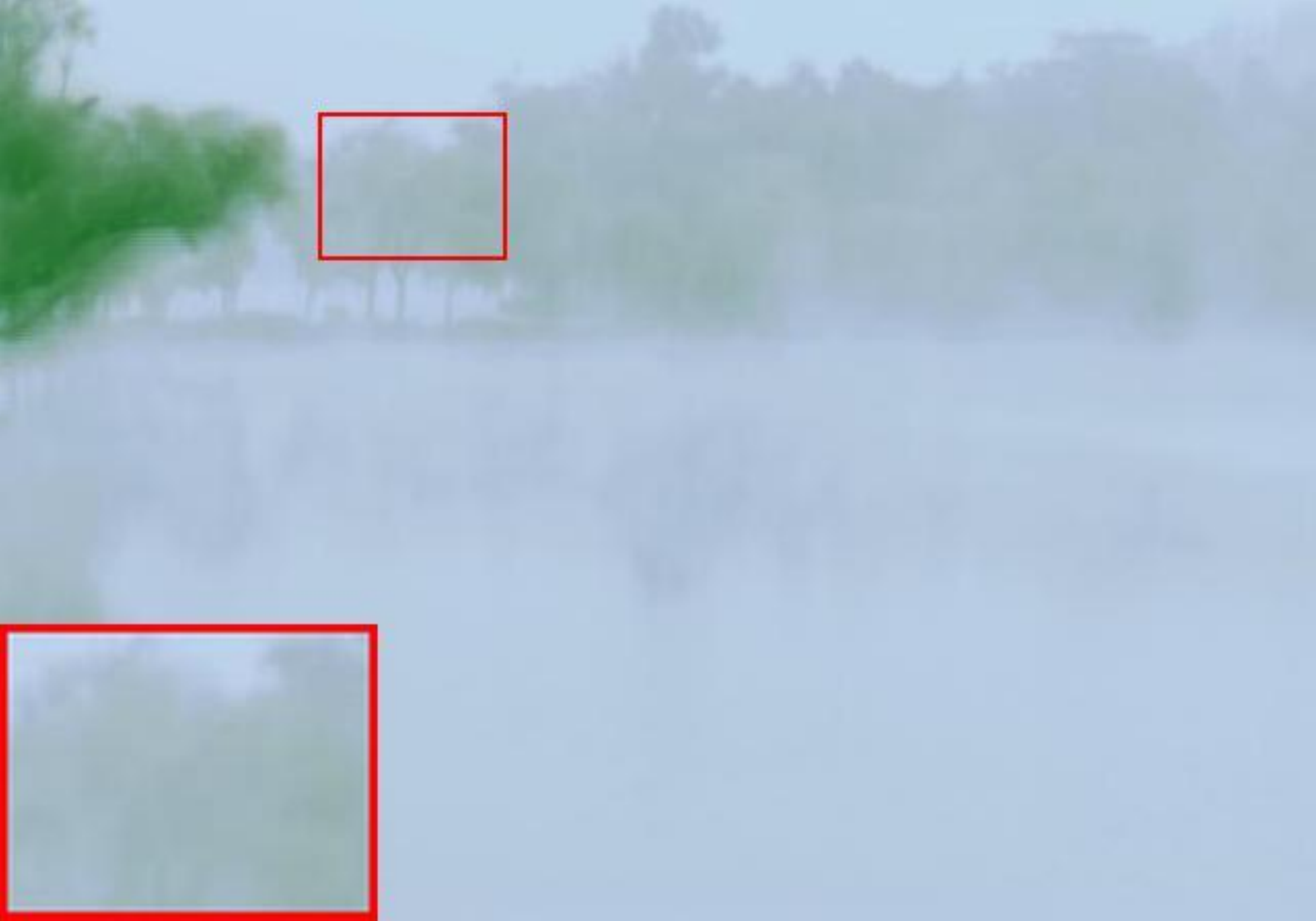}}
		\subfigure{\includegraphics[scale=\m_wid]{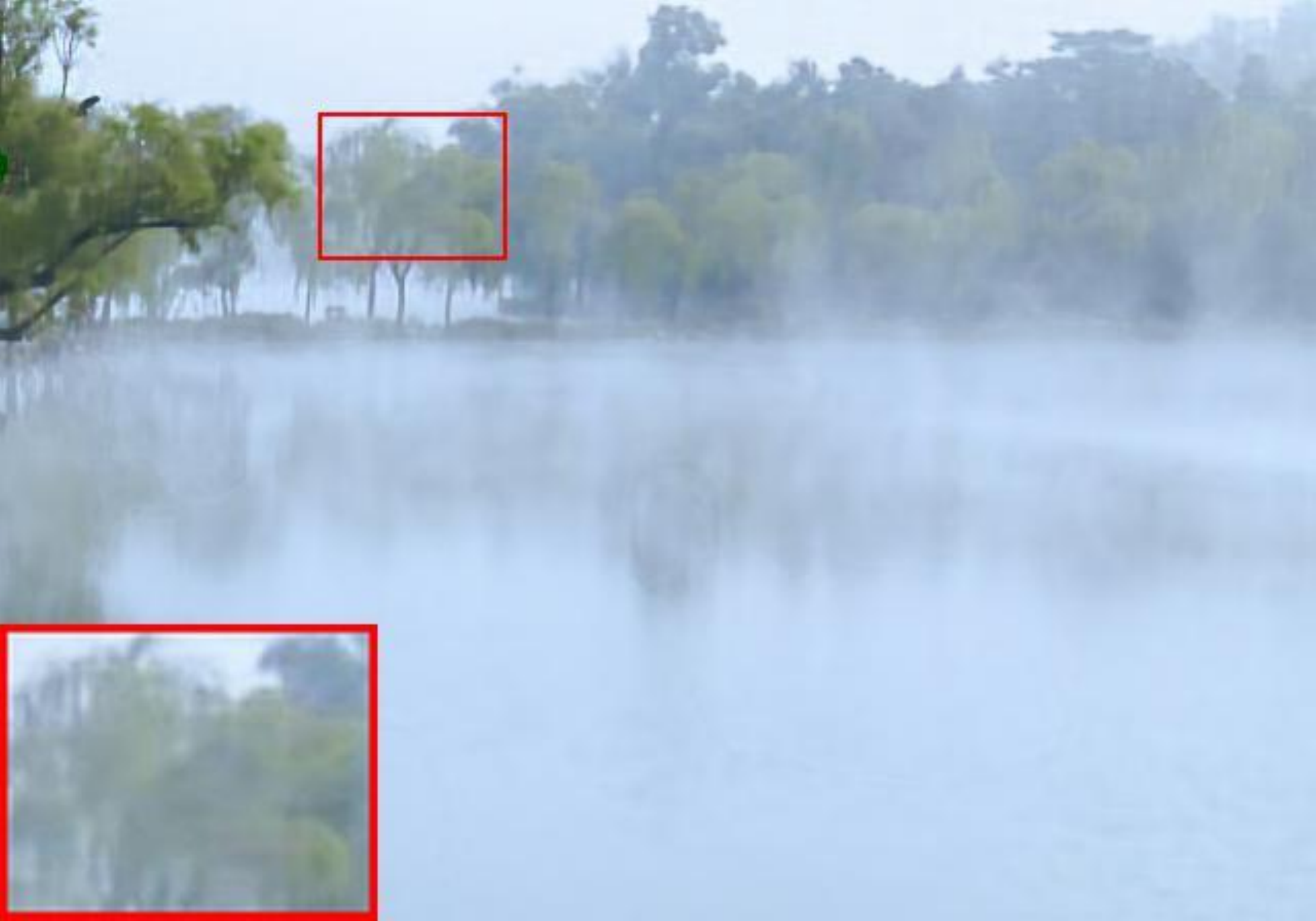}}
		\subfigure{\includegraphics[scale=\m_wid]{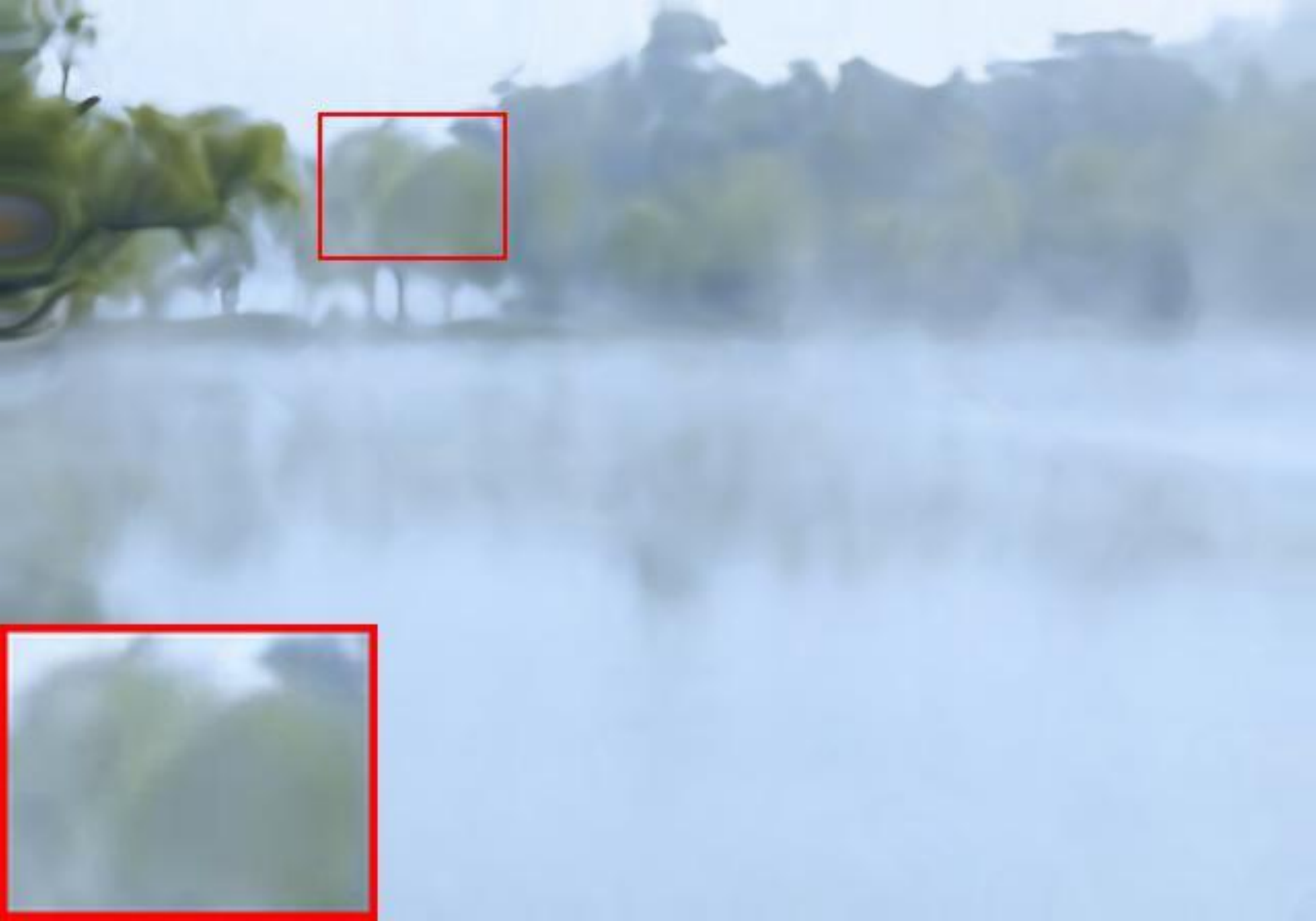}}
		\subfigure{\includegraphics[scale=\m_wid]{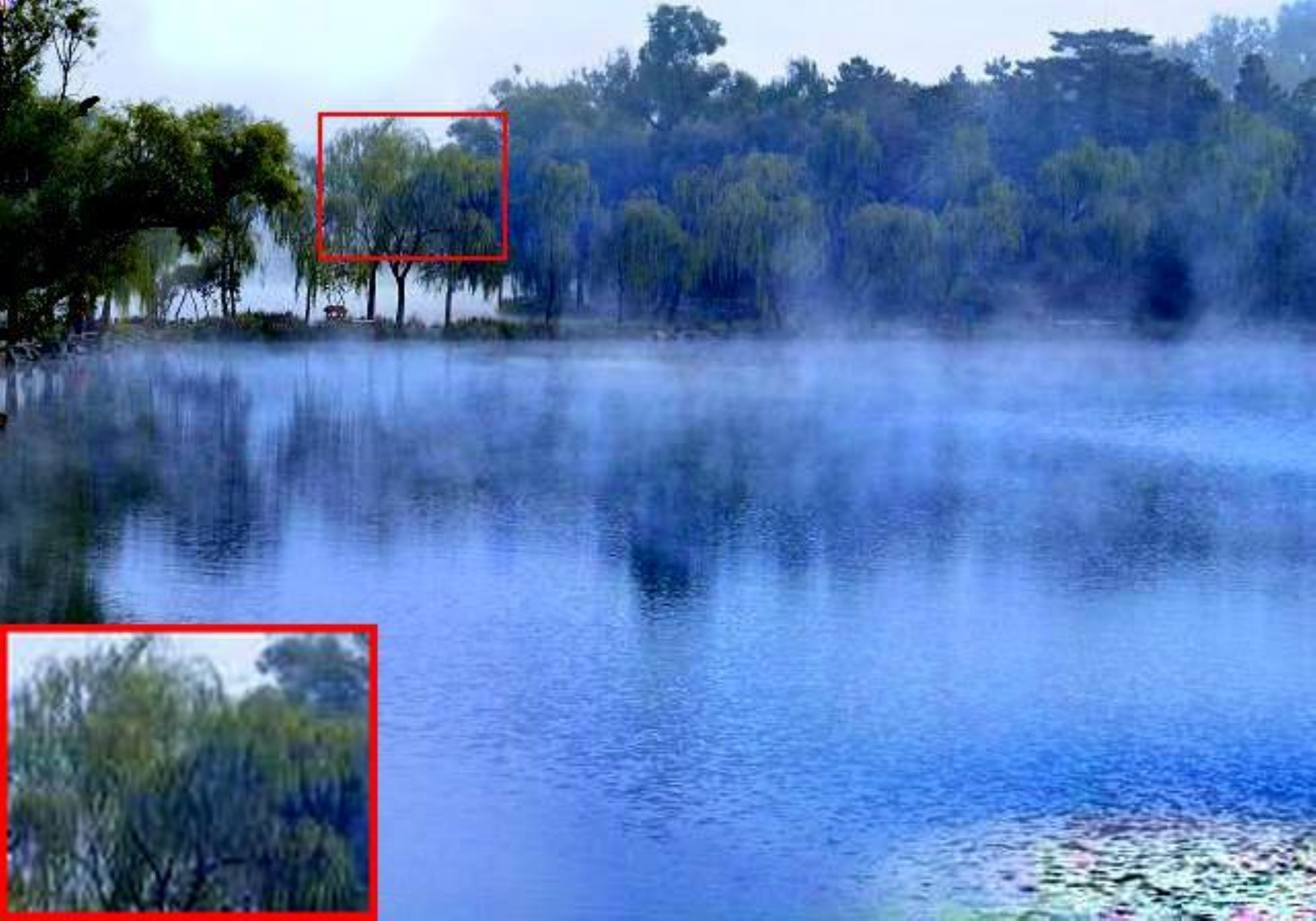}}
		\subfigure{\includegraphics[scale=\m_wid]{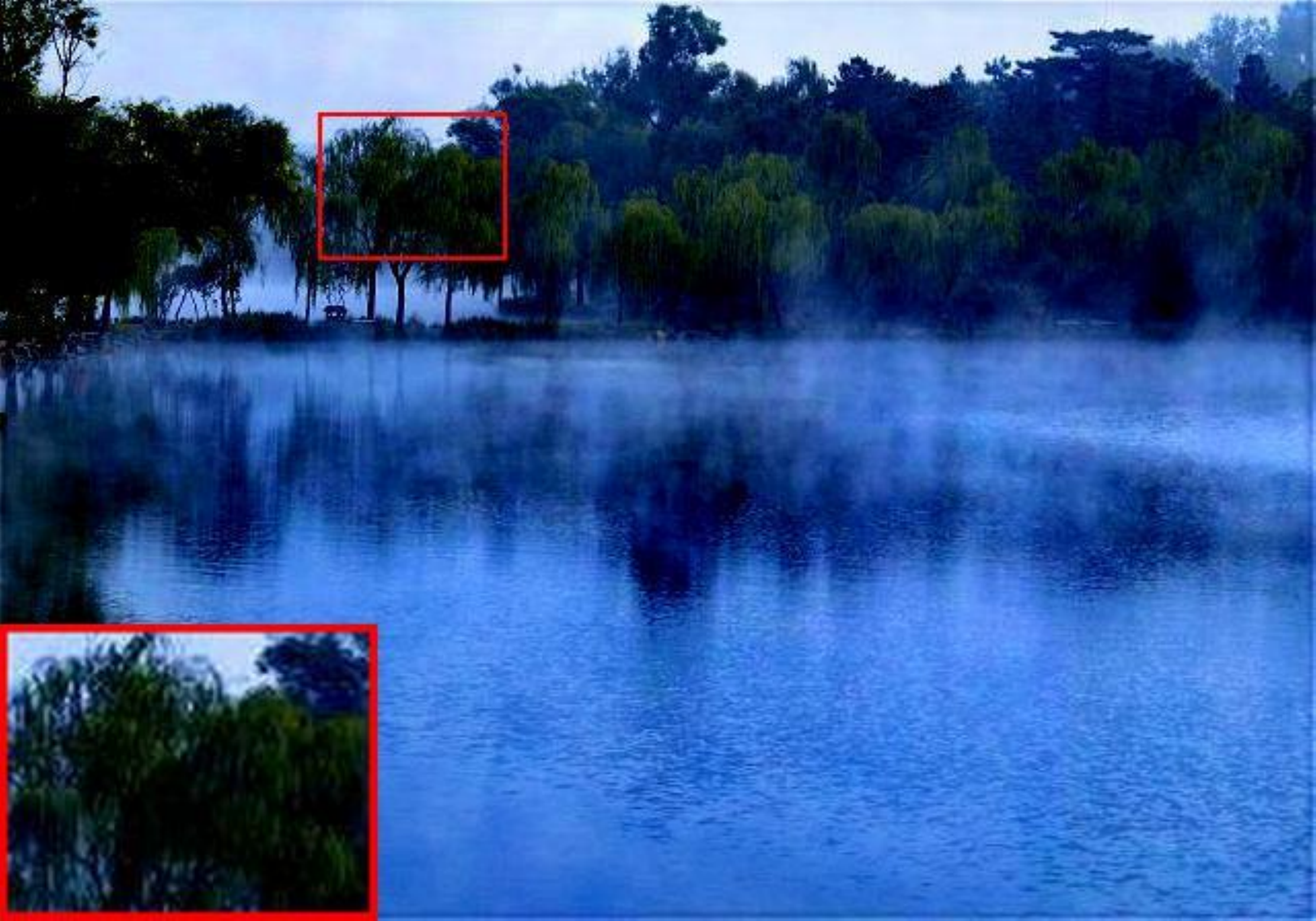}}
		\subfigure{\includegraphics[scale=\m_wid]{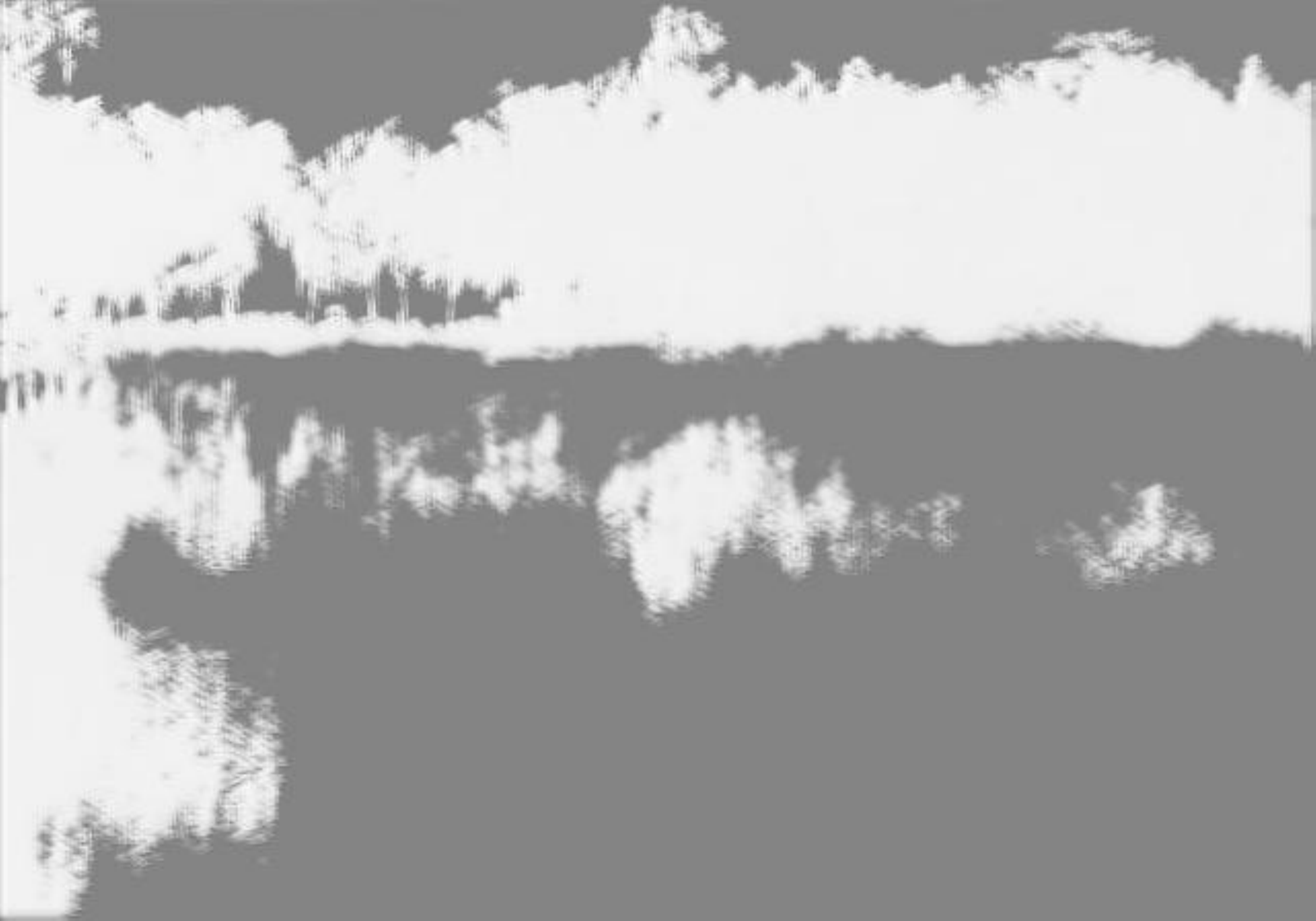}}
	\end{center}
	\vspace{-0.8cm}
	
	\begin{center}
	\def \m_wid{0.0466}
		\subfigure{\includegraphics[scale=\m_wid]{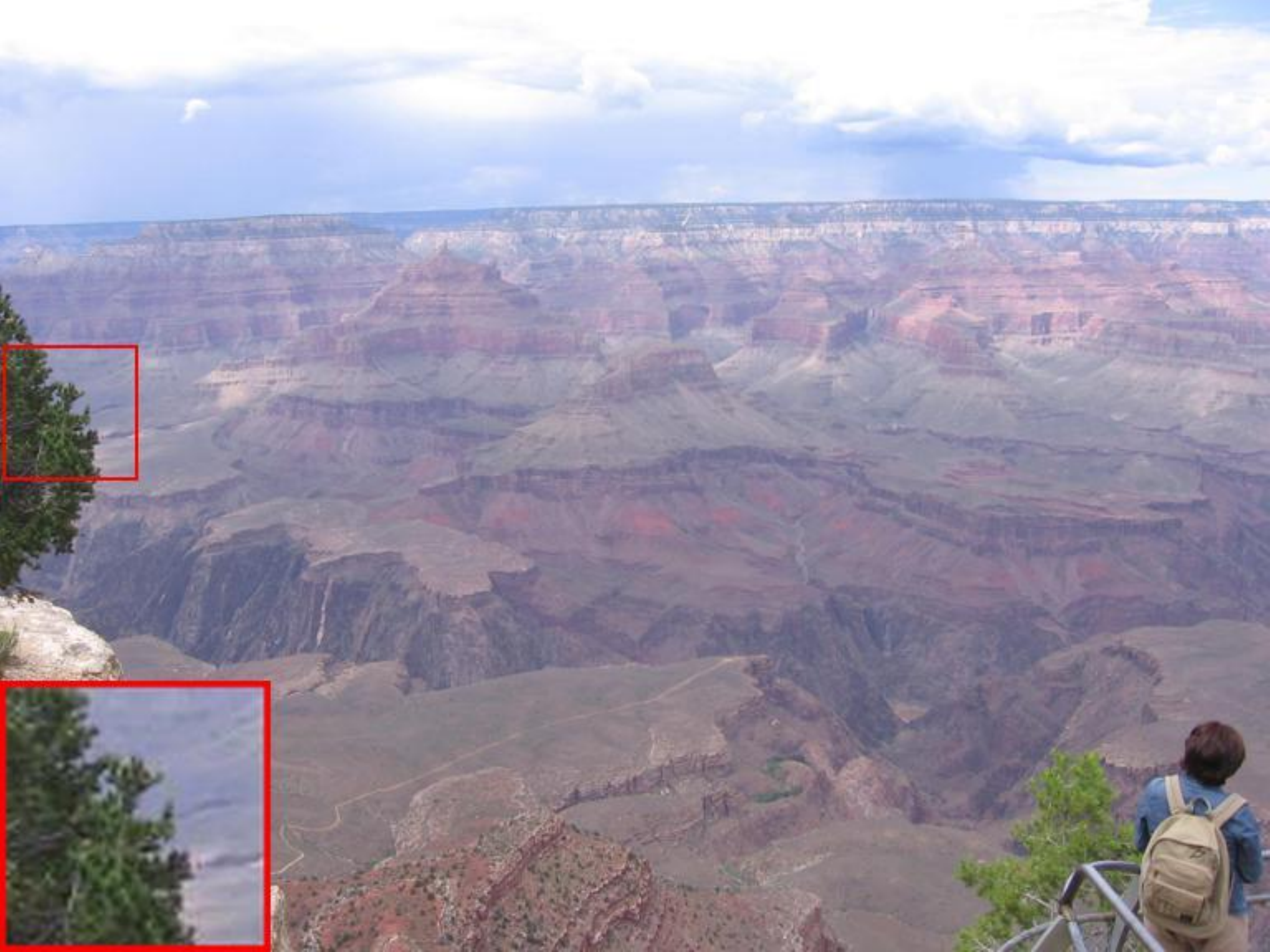}}
		\subfigure{\includegraphics[scale=\m_wid]{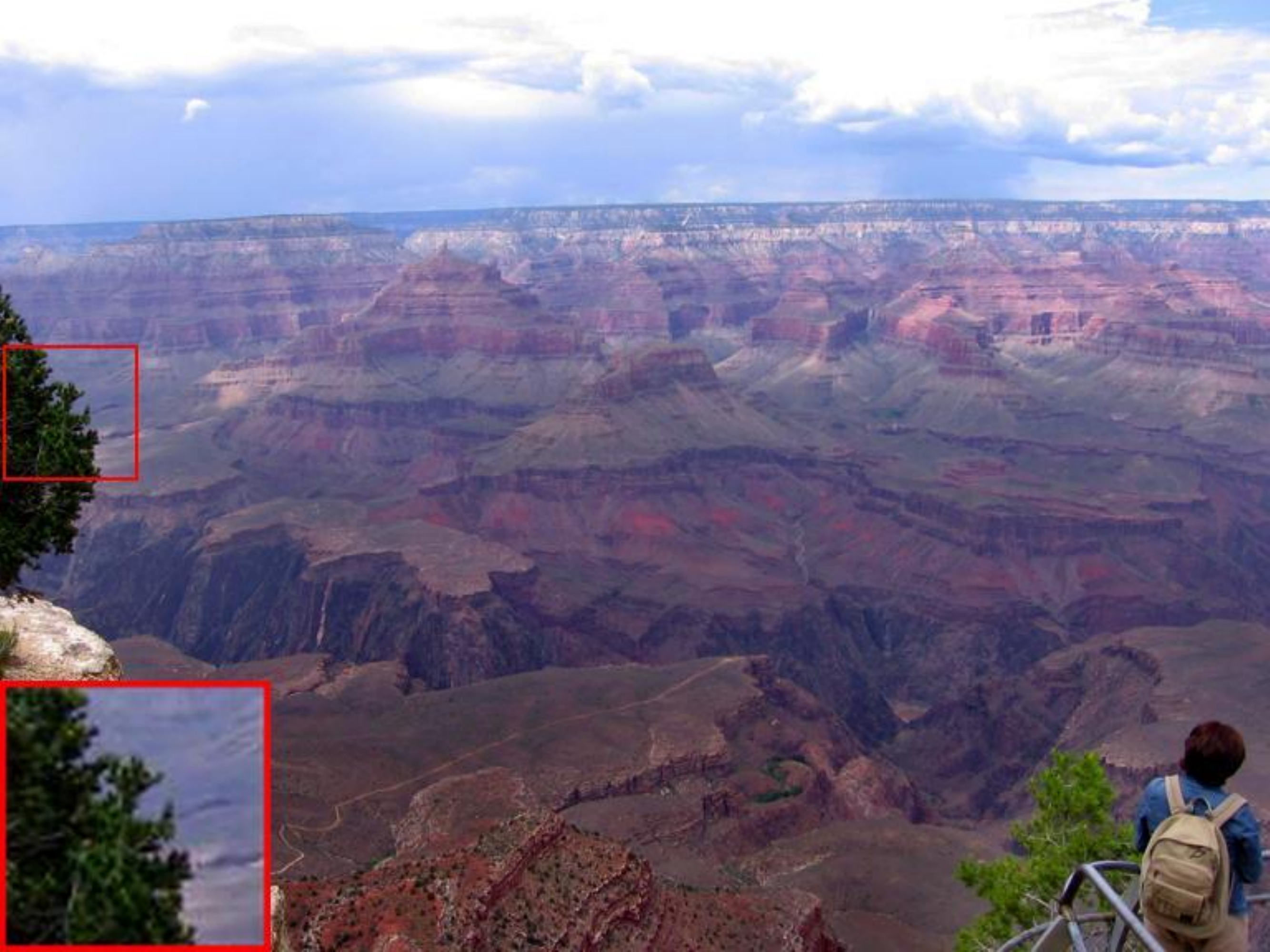}}
		\subfigure{\includegraphics[scale=\m_wid]{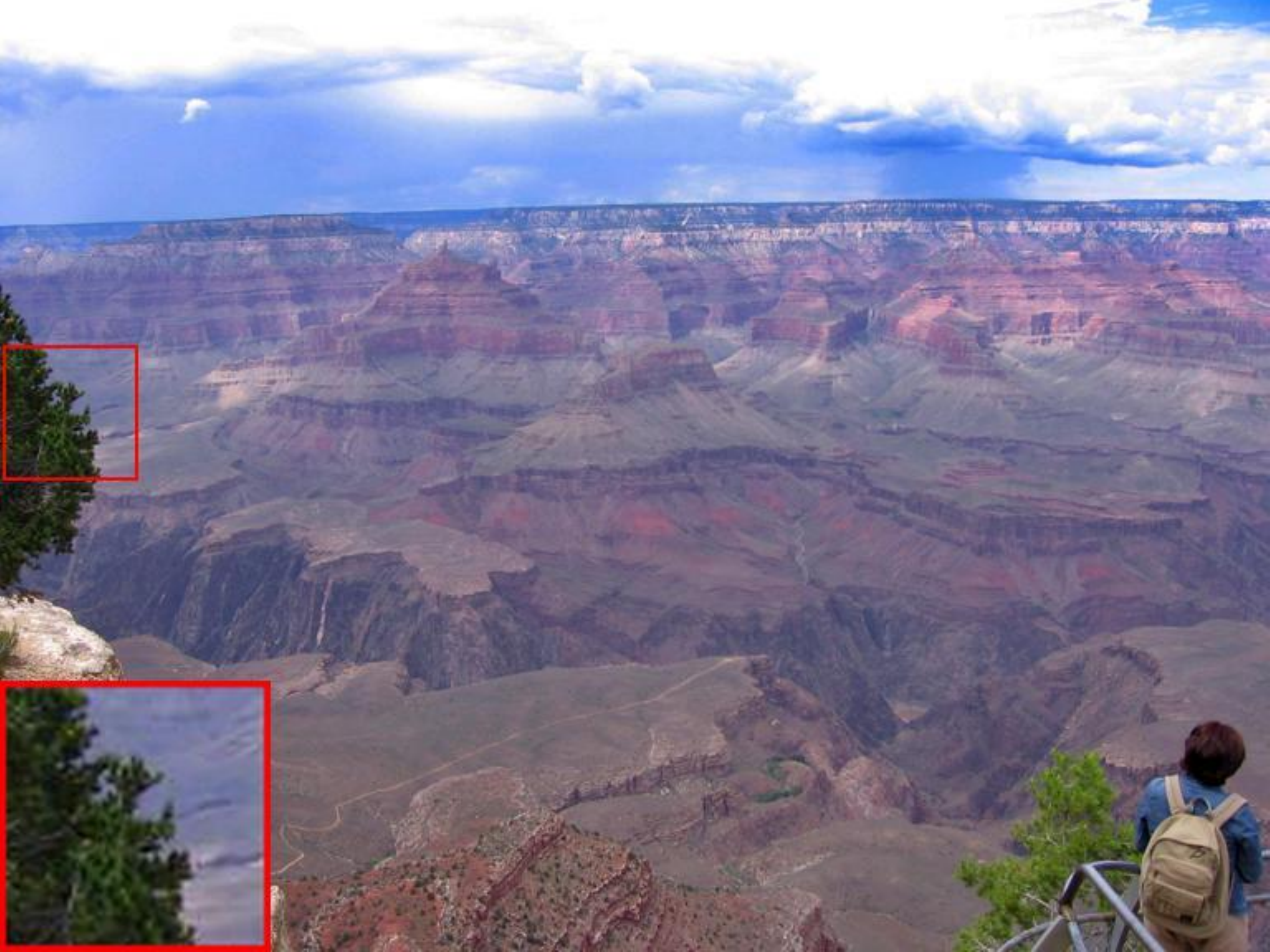}}
		\subfigure{\includegraphics[scale=\m_wid]{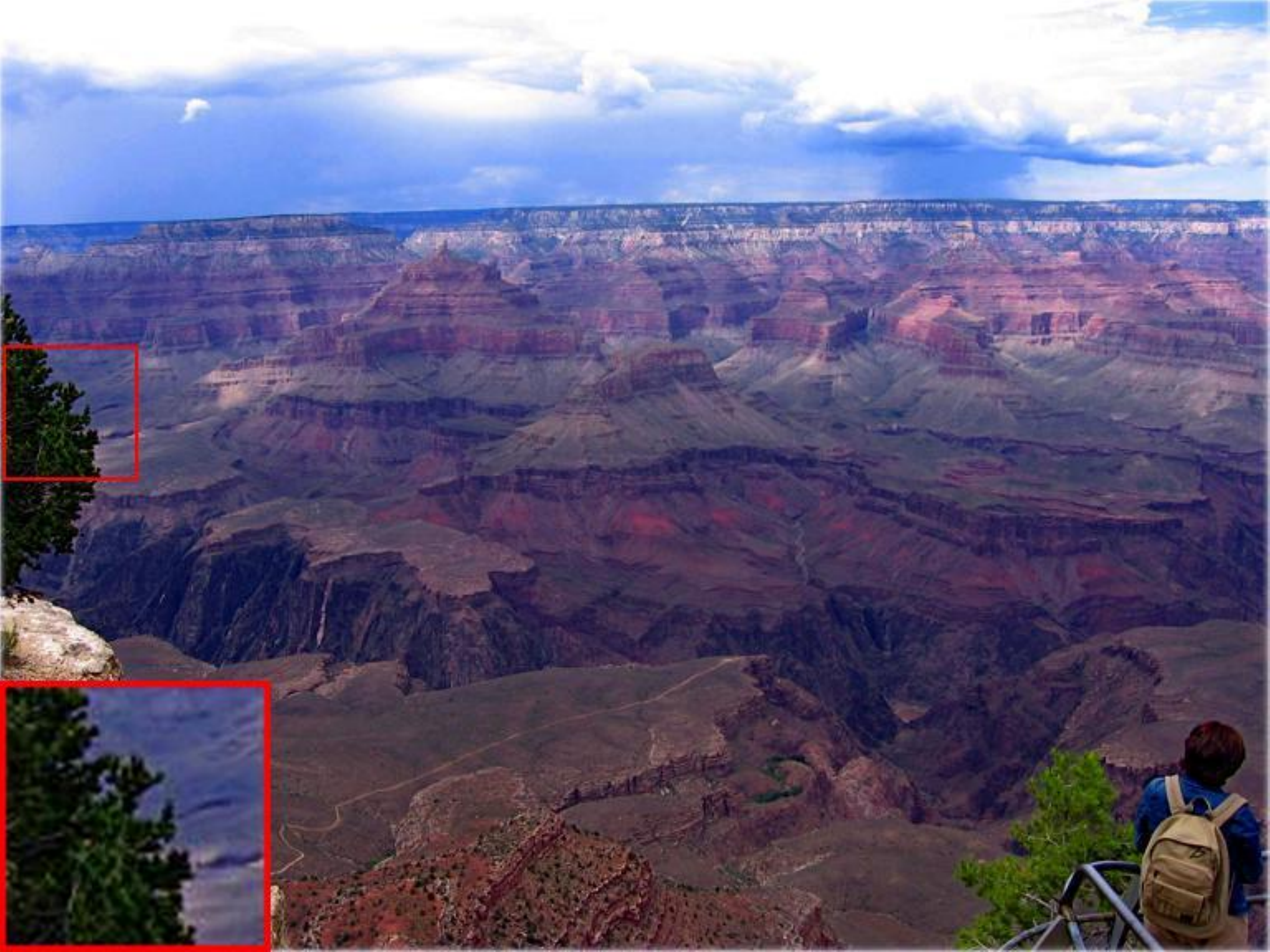}}
		\subfigure{\includegraphics[scale=\m_wid]{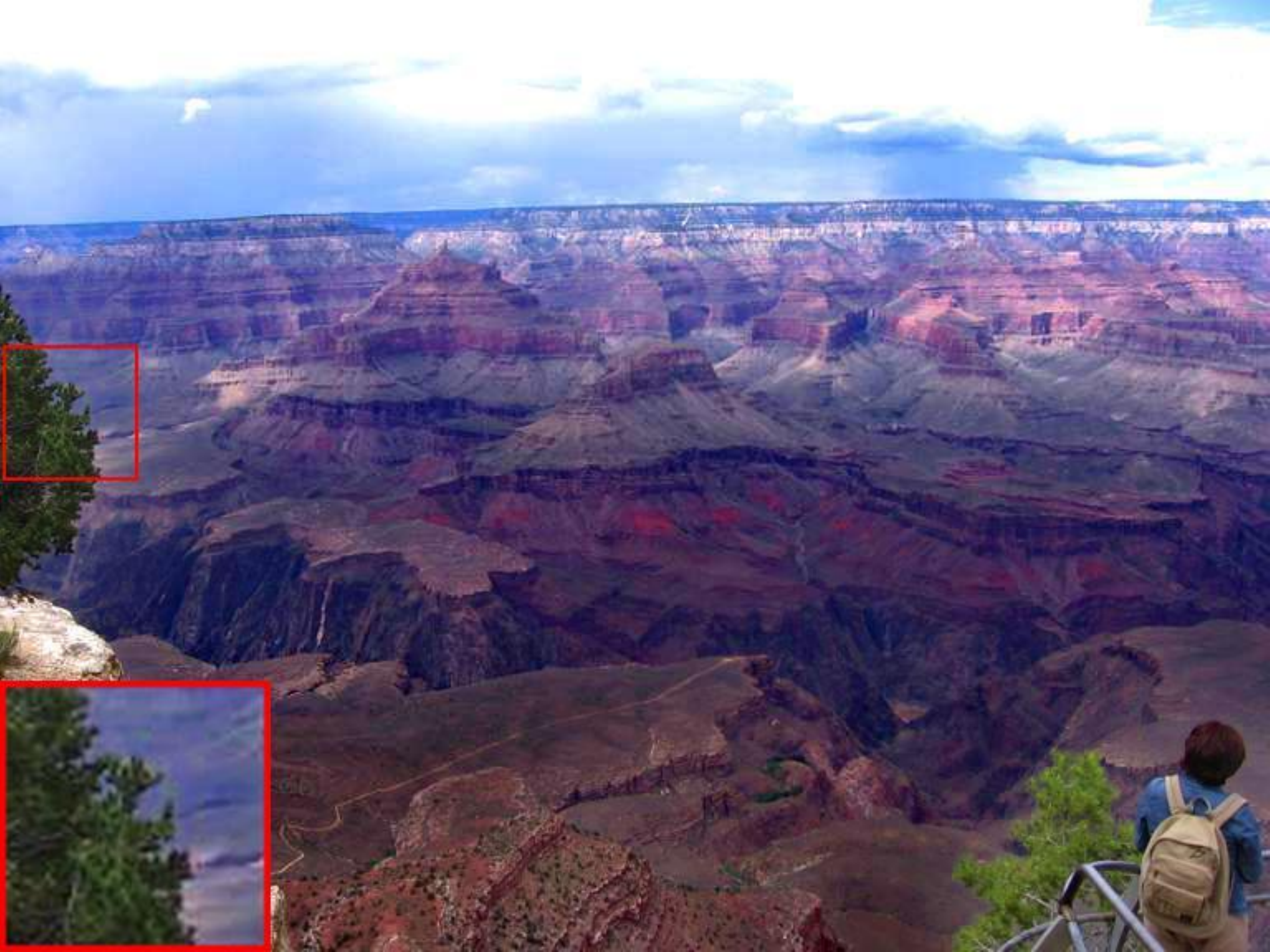}}
		\subfigure{\includegraphics[scale=\m_wid]{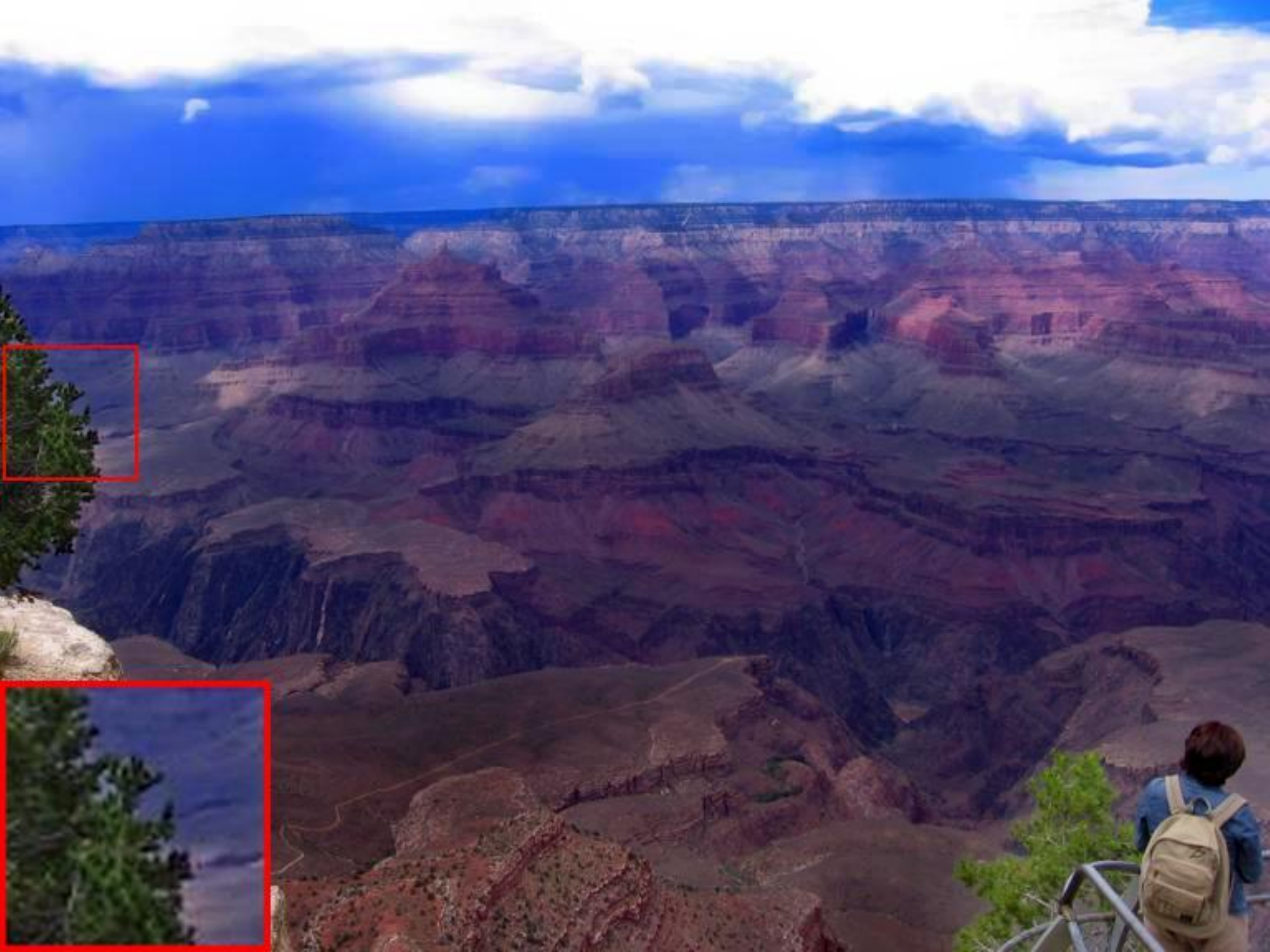}}
		\subfigure{\includegraphics[scale=\m_wid]{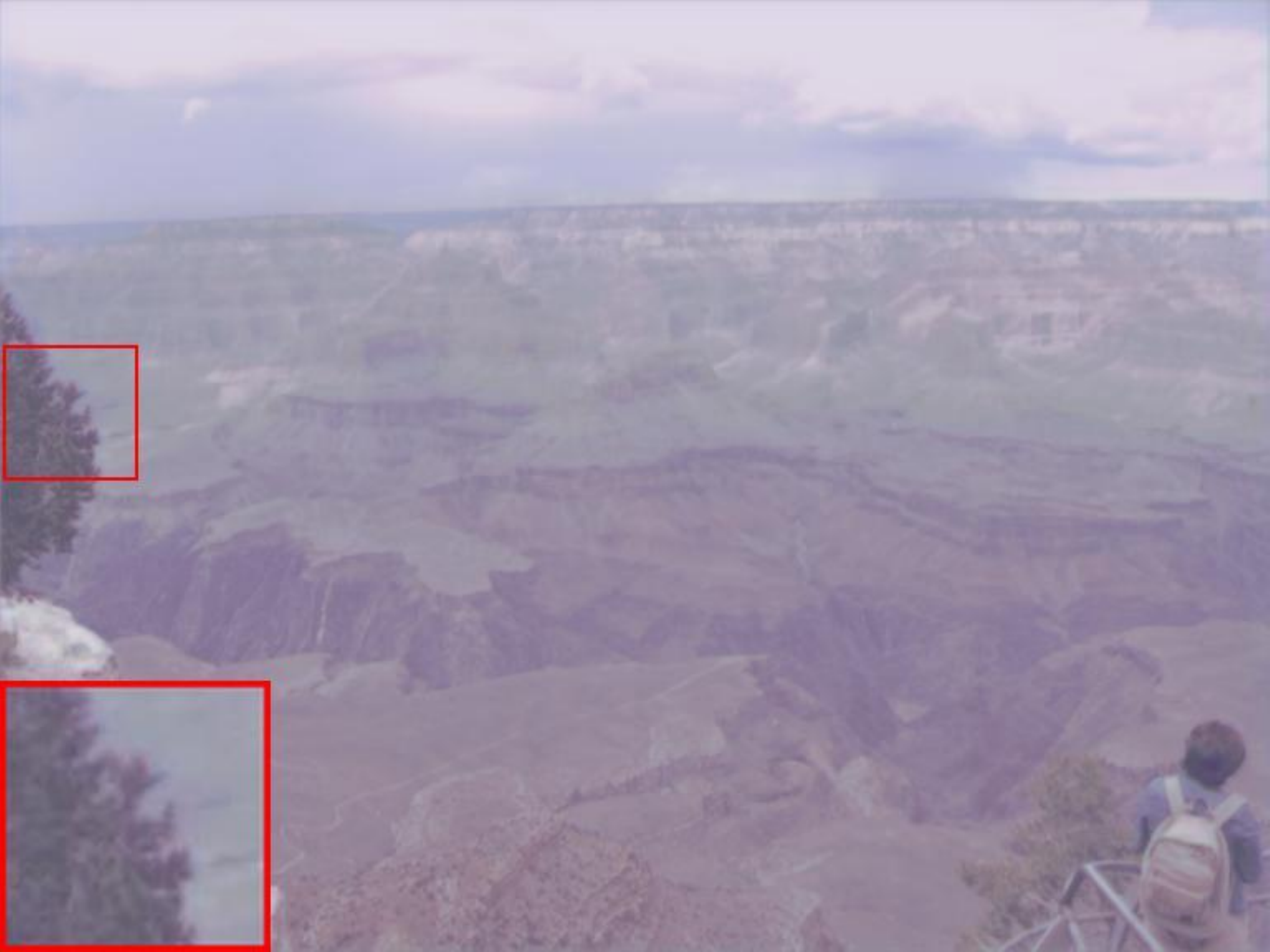}}
		\subfigure{\includegraphics[scale=\m_wid]{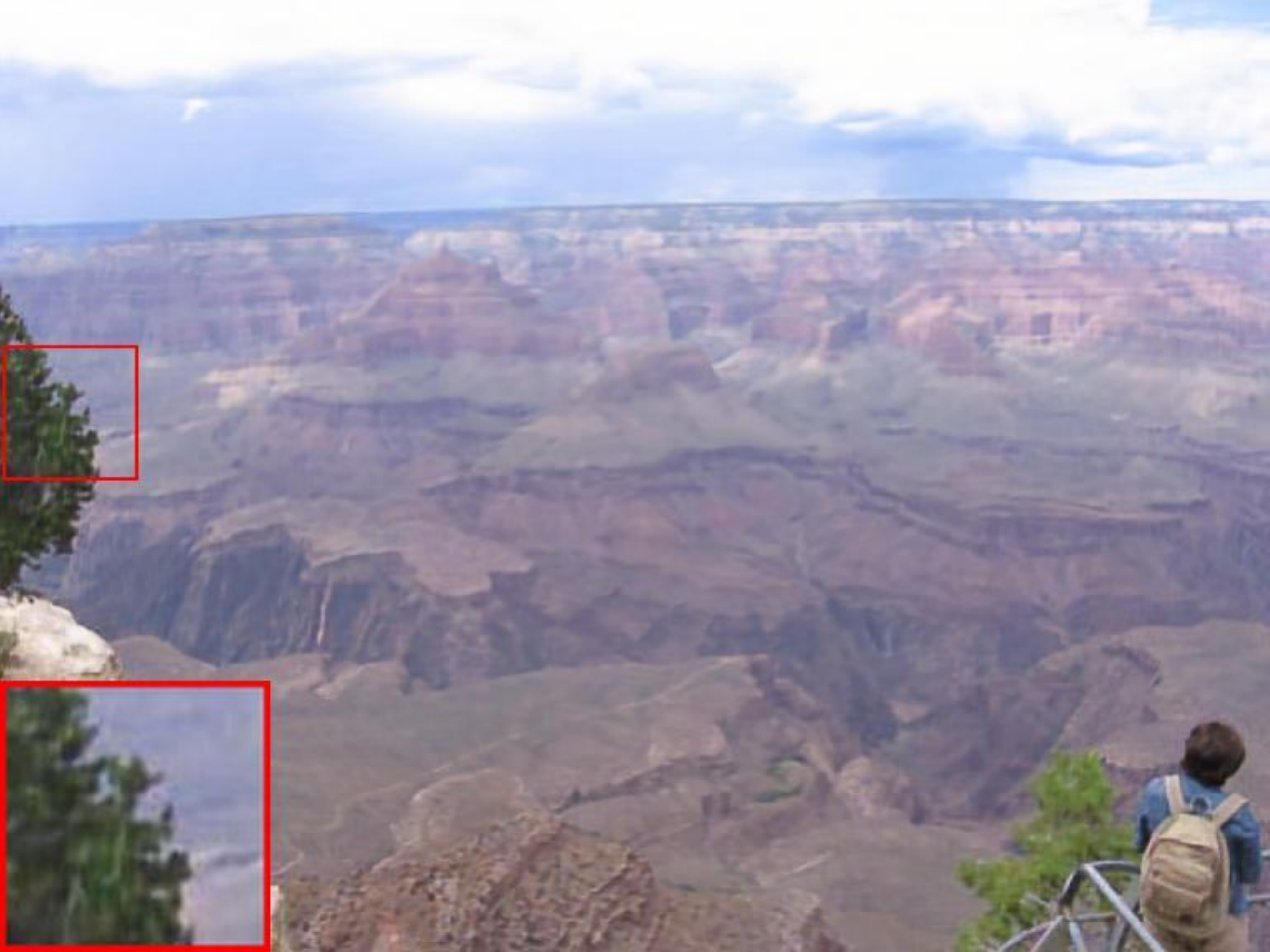}}
		\subfigure{\includegraphics[scale=\m_wid]{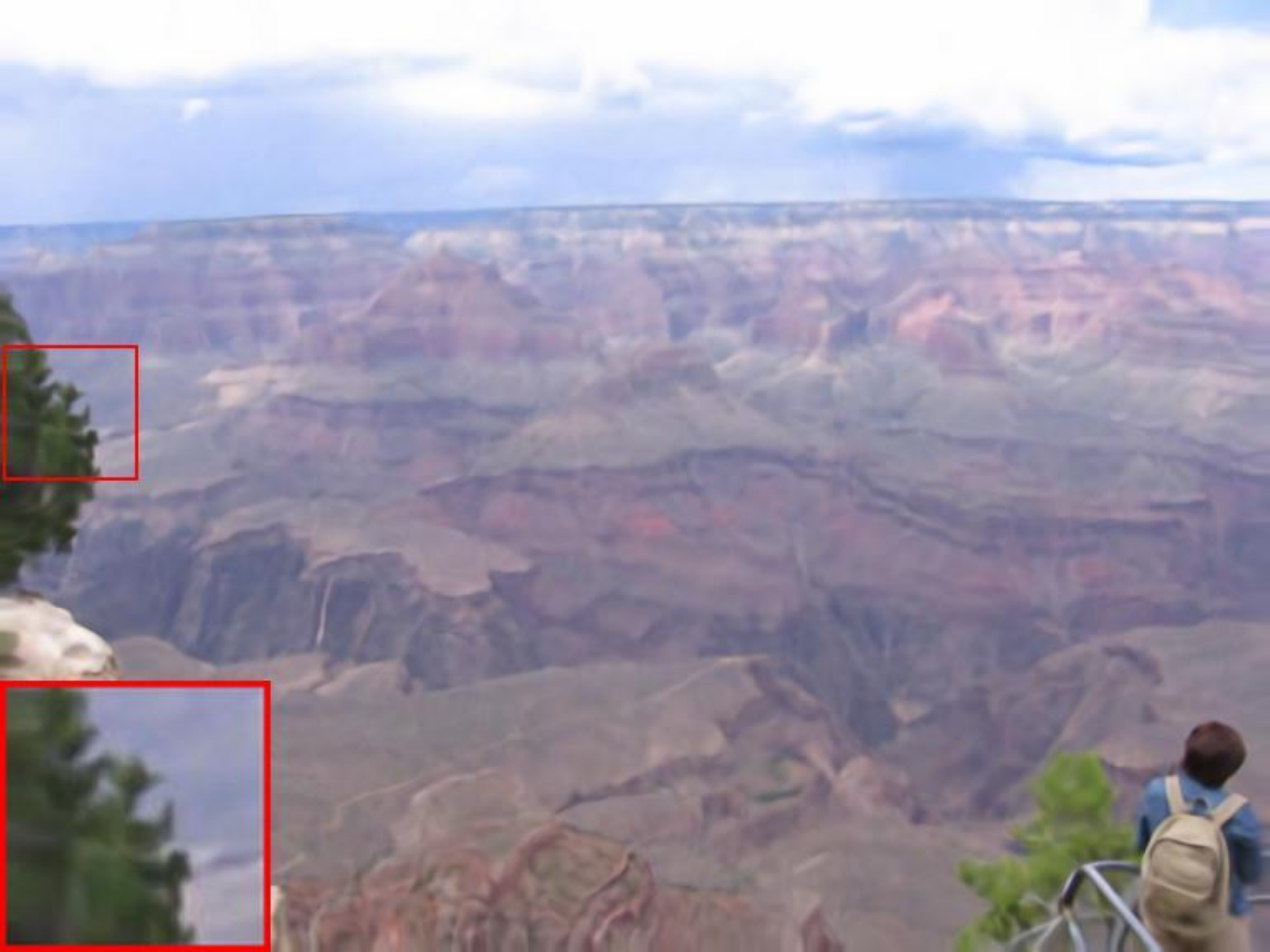}}
		\subfigure{\includegraphics[scale=\m_wid]{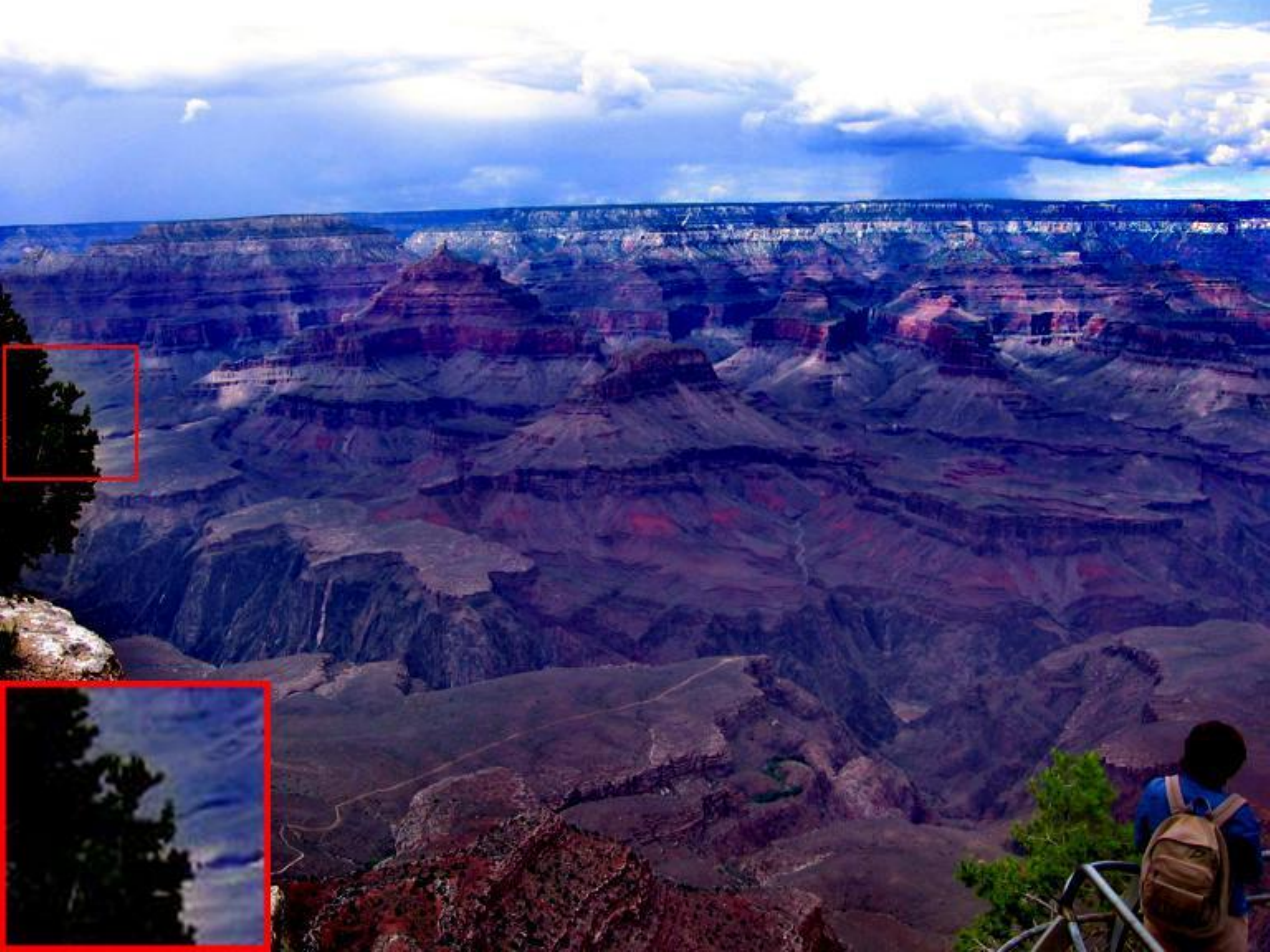}}
		\subfigure{\includegraphics[scale=\m_wid]{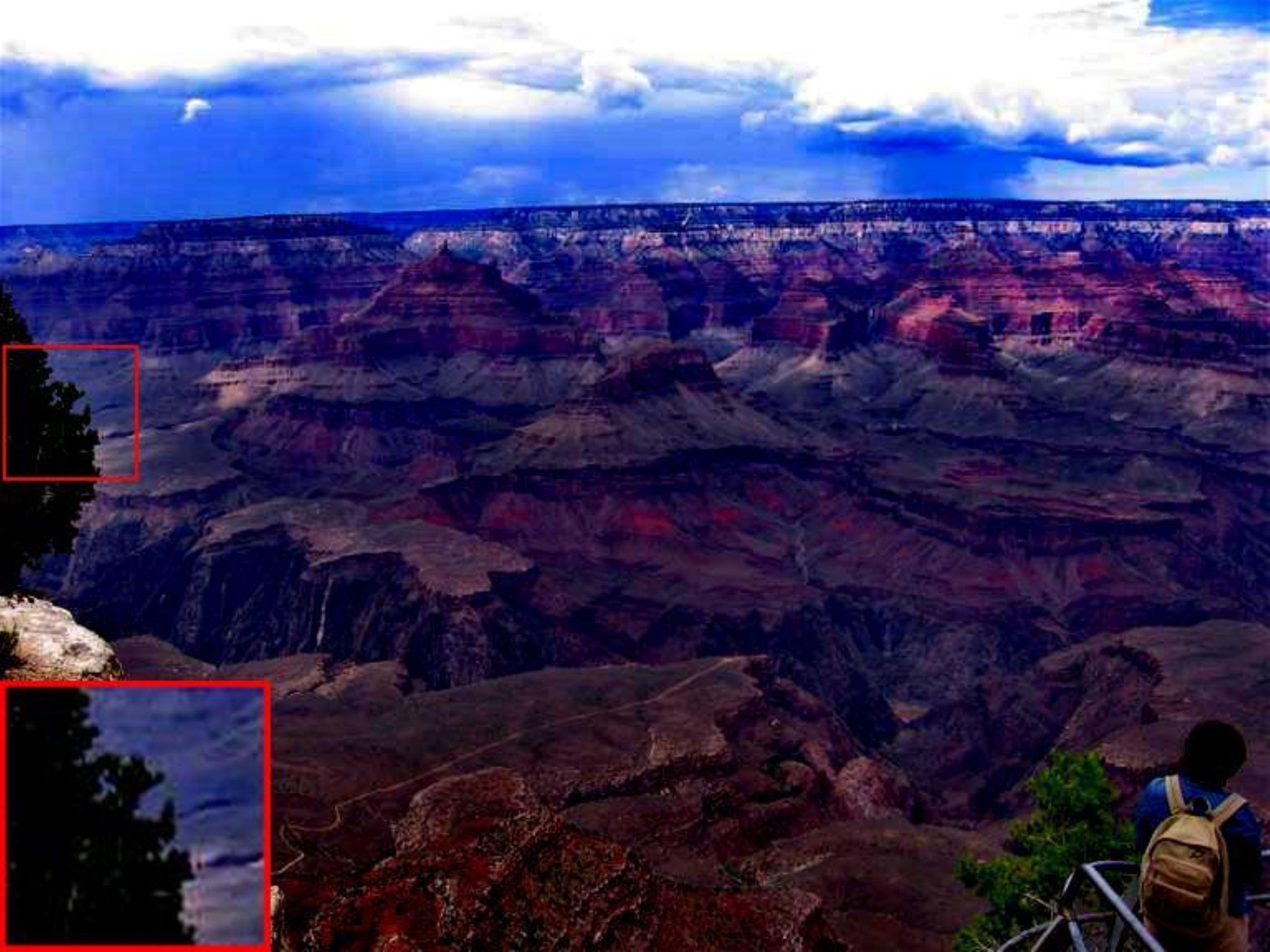}}
		\subfigure{\includegraphics[scale=\m_wid]{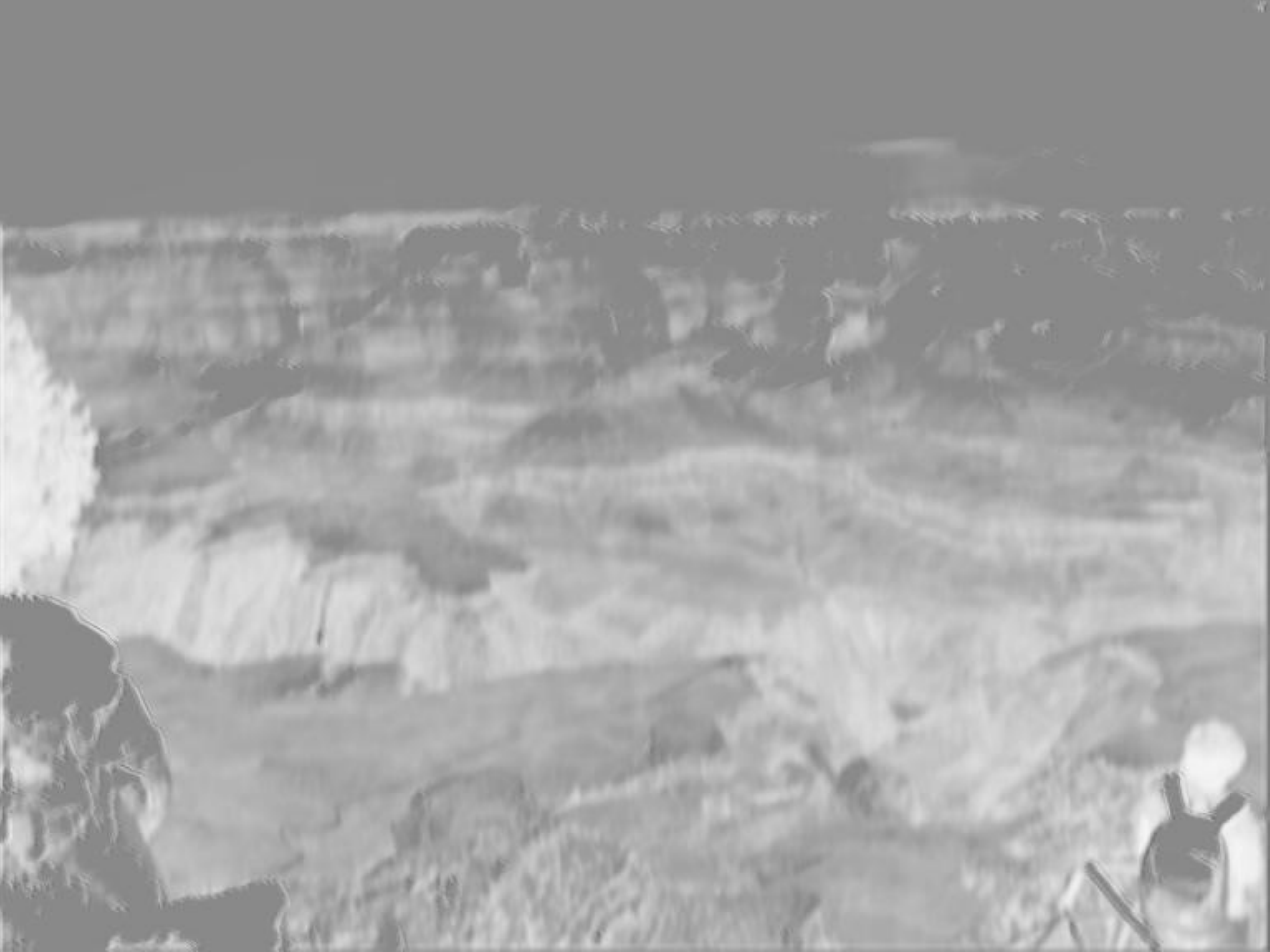}}
	\end{center}
	\vspace{-0.8cm}
	
	\begin{center}
	\def \m_wid{0.0488}
		\subfigure{\includegraphics[scale=\m_wid]{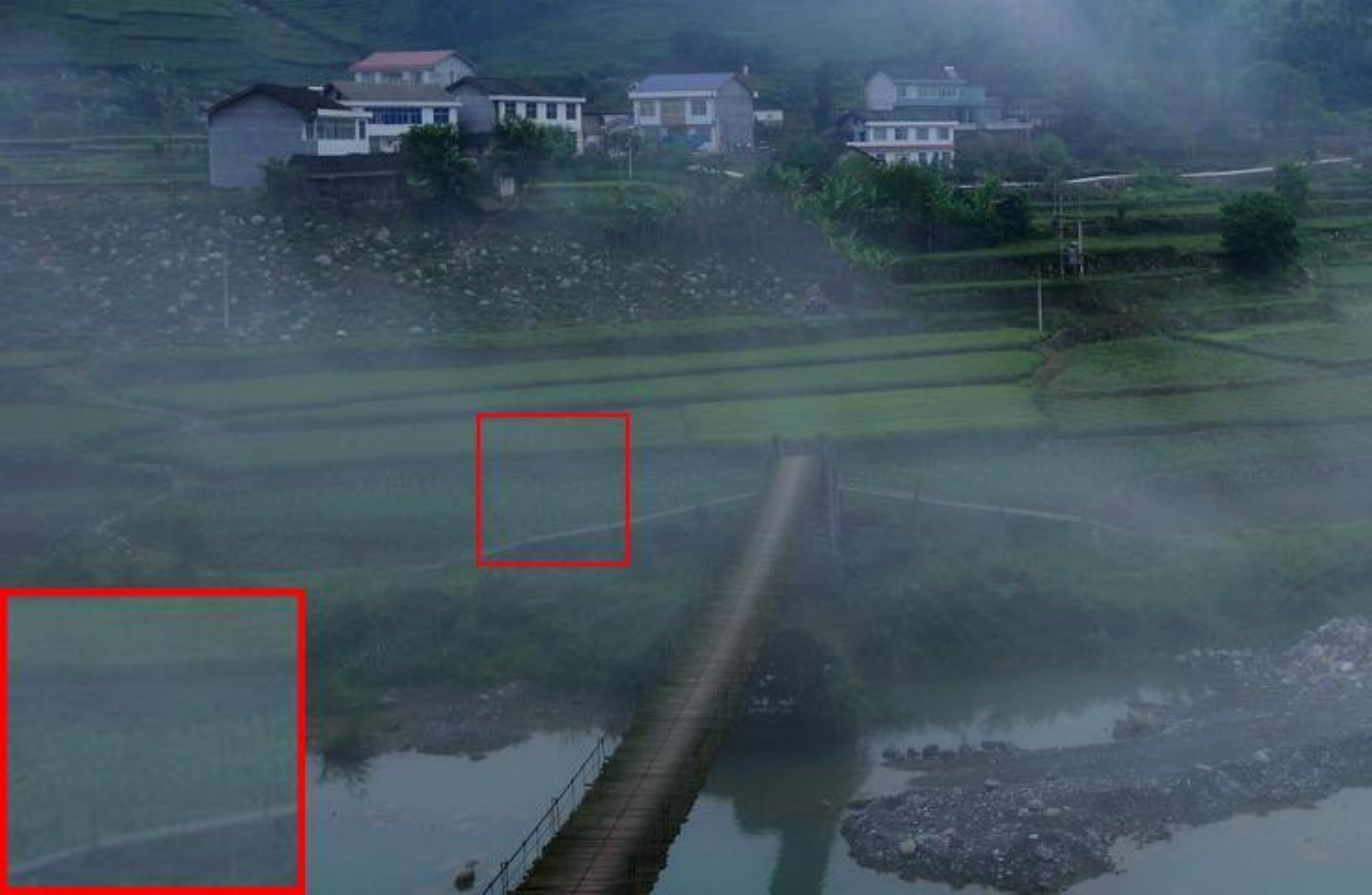}}
		\subfigure{\includegraphics[scale=\m_wid]{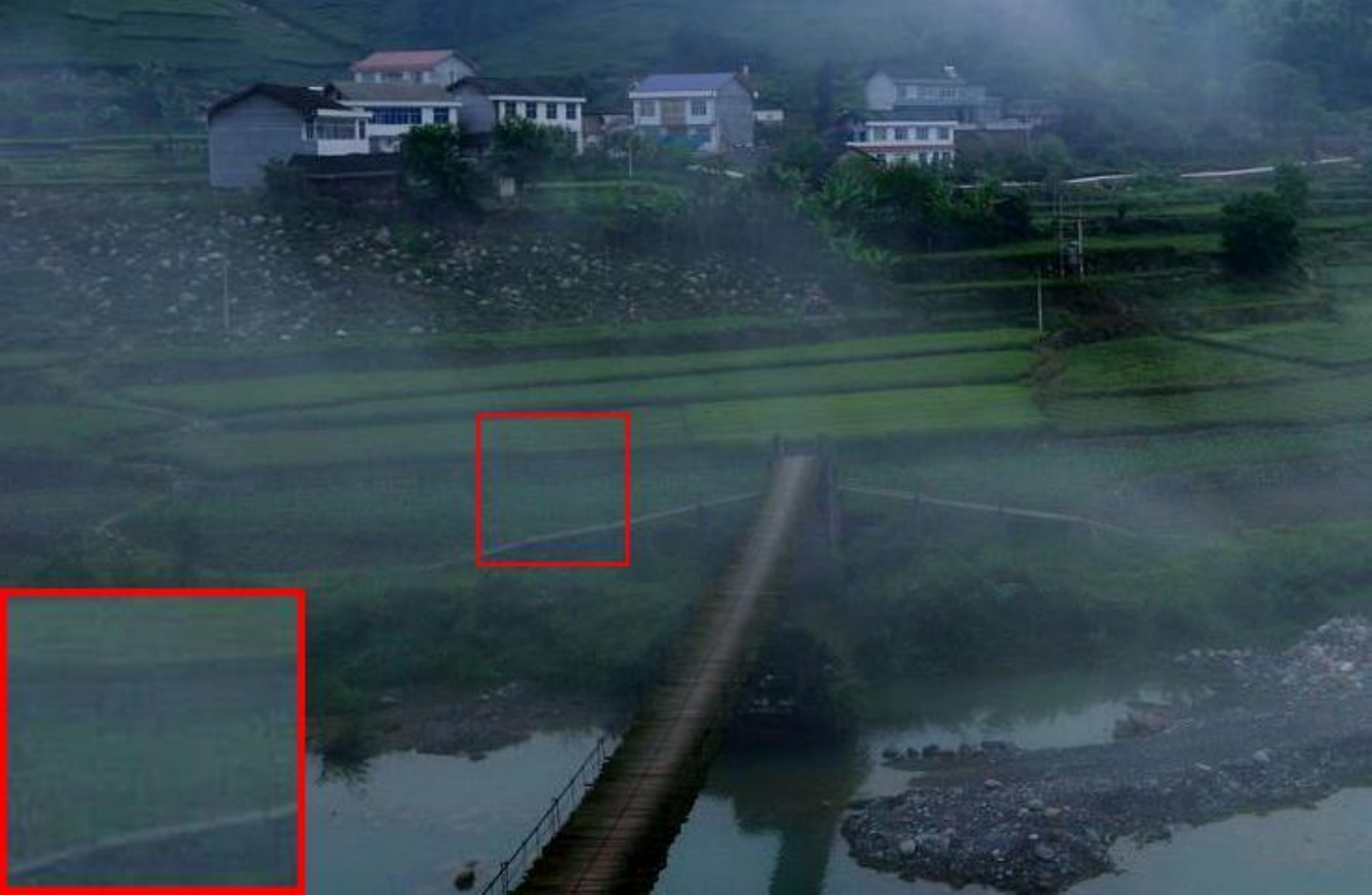}}
		\subfigure{\includegraphics[scale=\m_wid]{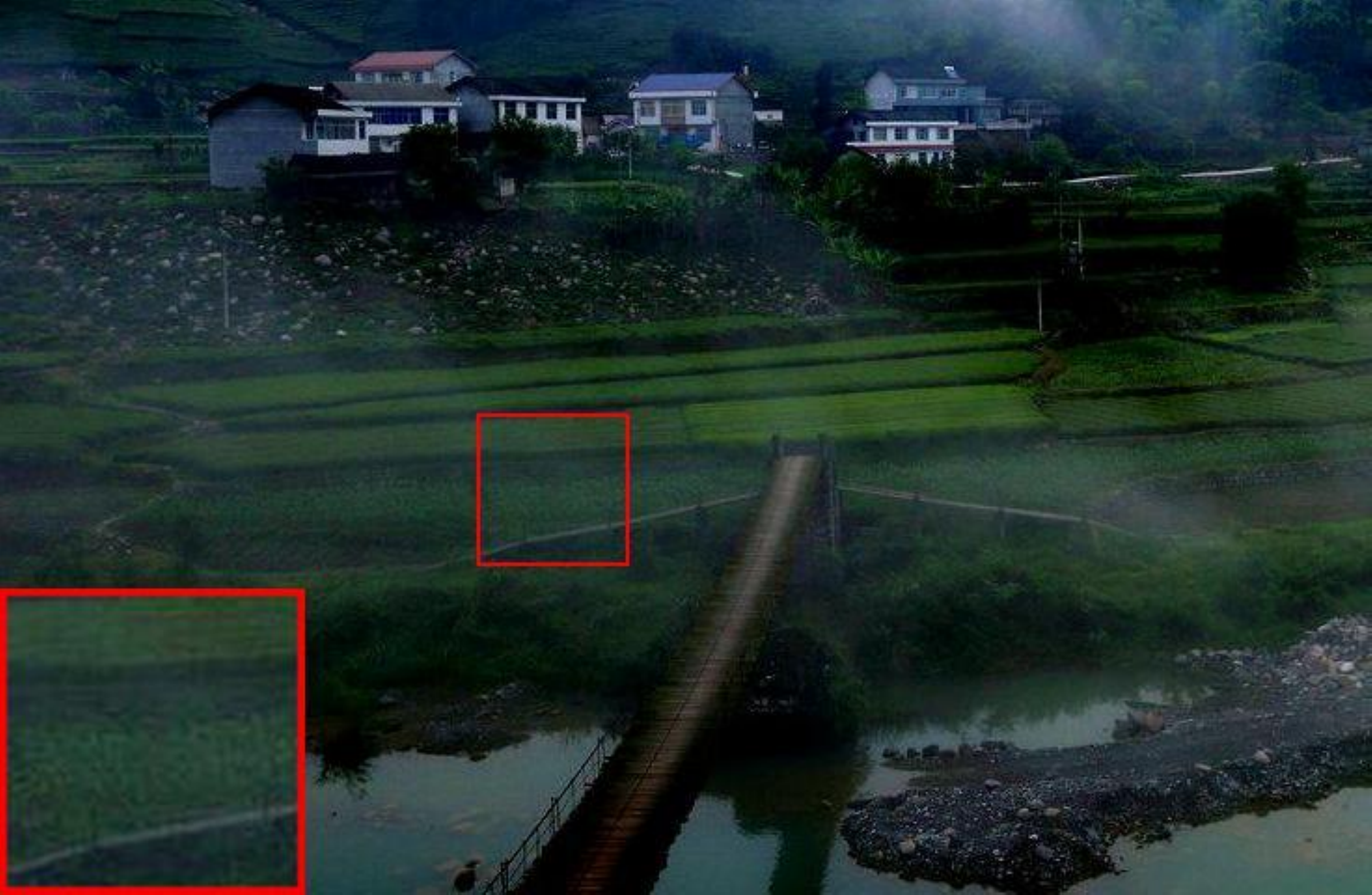}}
		\subfigure{\includegraphics[scale=\m_wid]{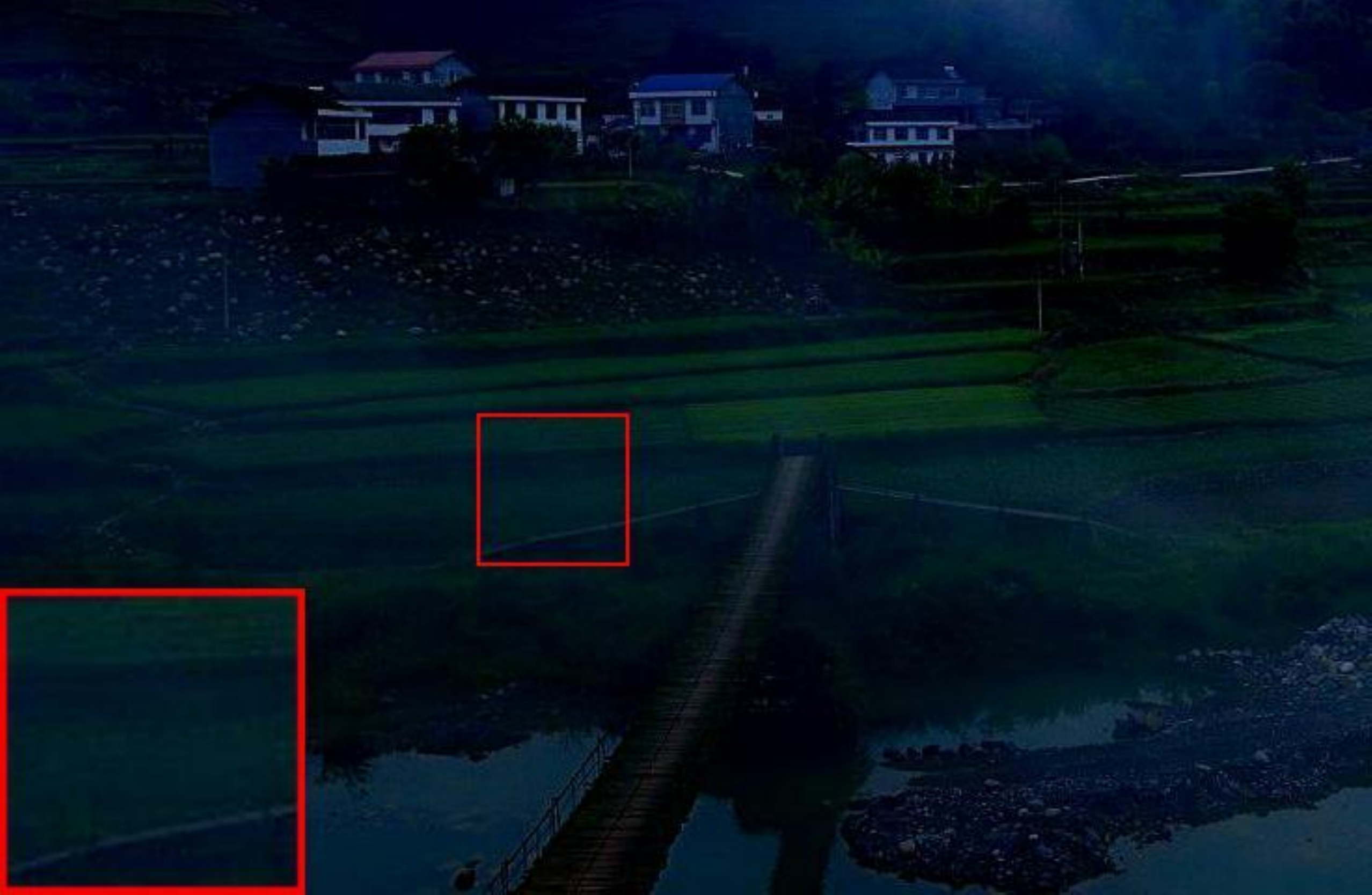}}
		\subfigure{\includegraphics[scale=\m_wid]{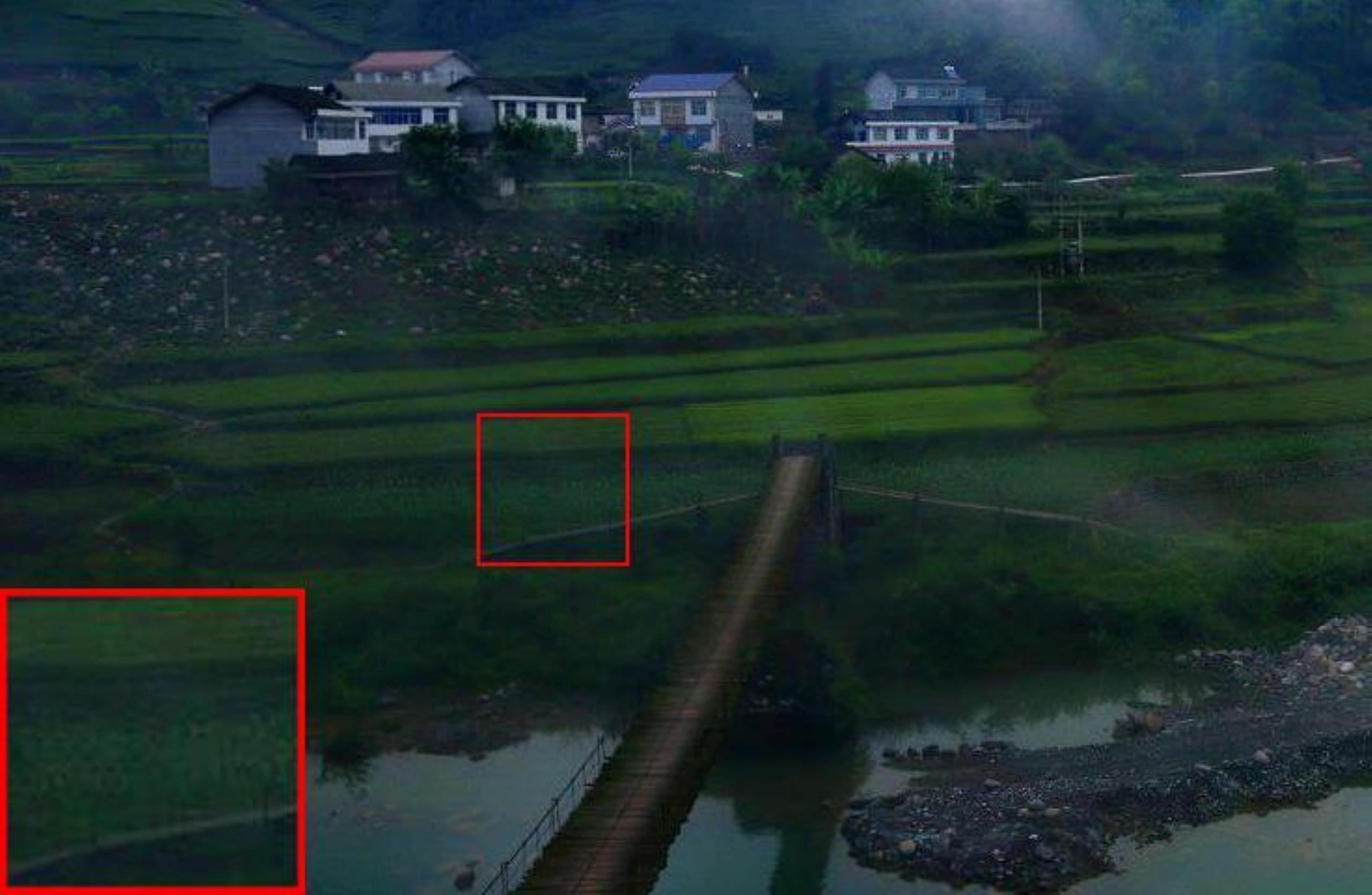}}
		\subfigure{\includegraphics[scale=\m_wid]{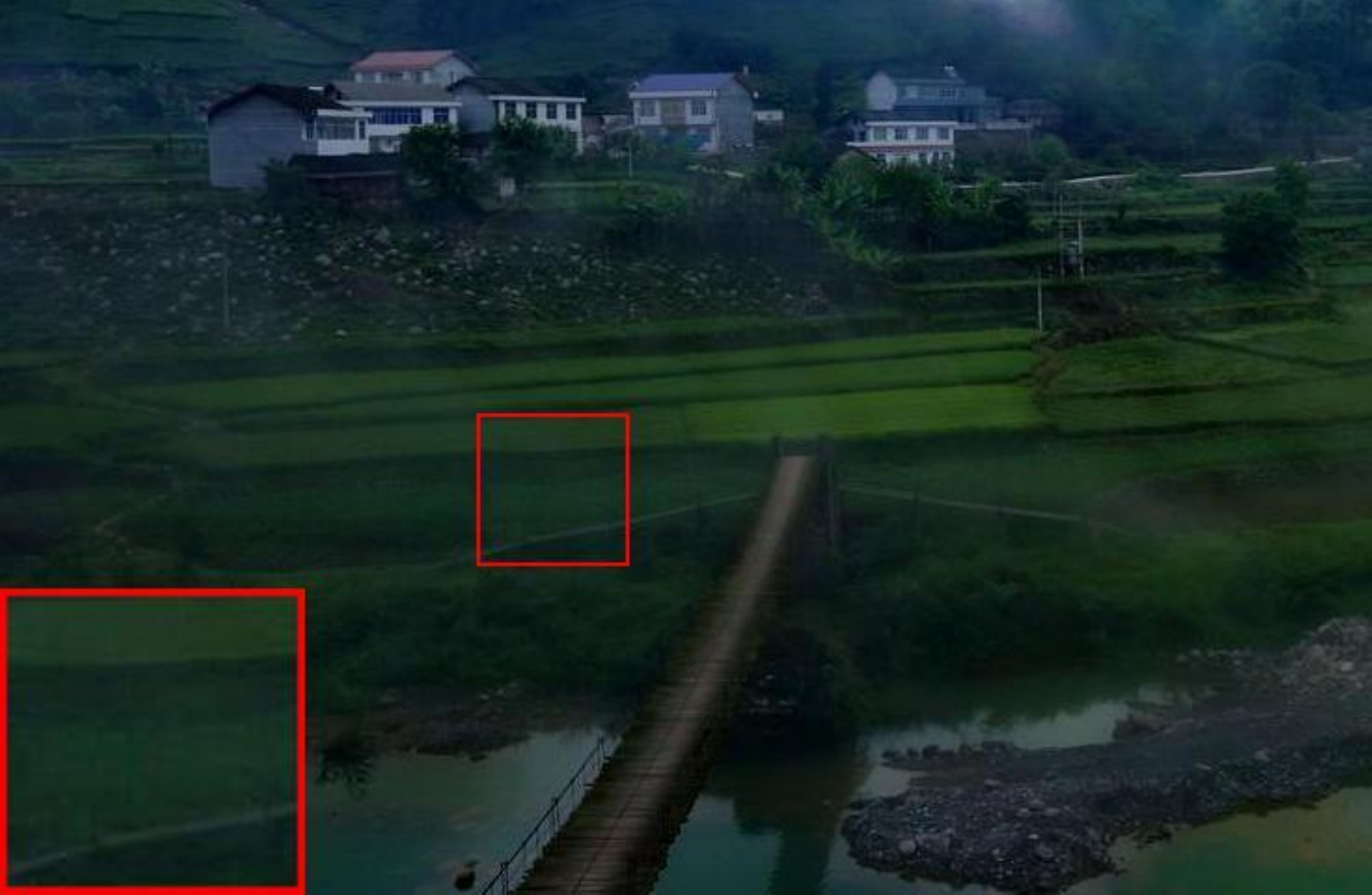}}
		\subfigure{\includegraphics[scale=\m_wid]{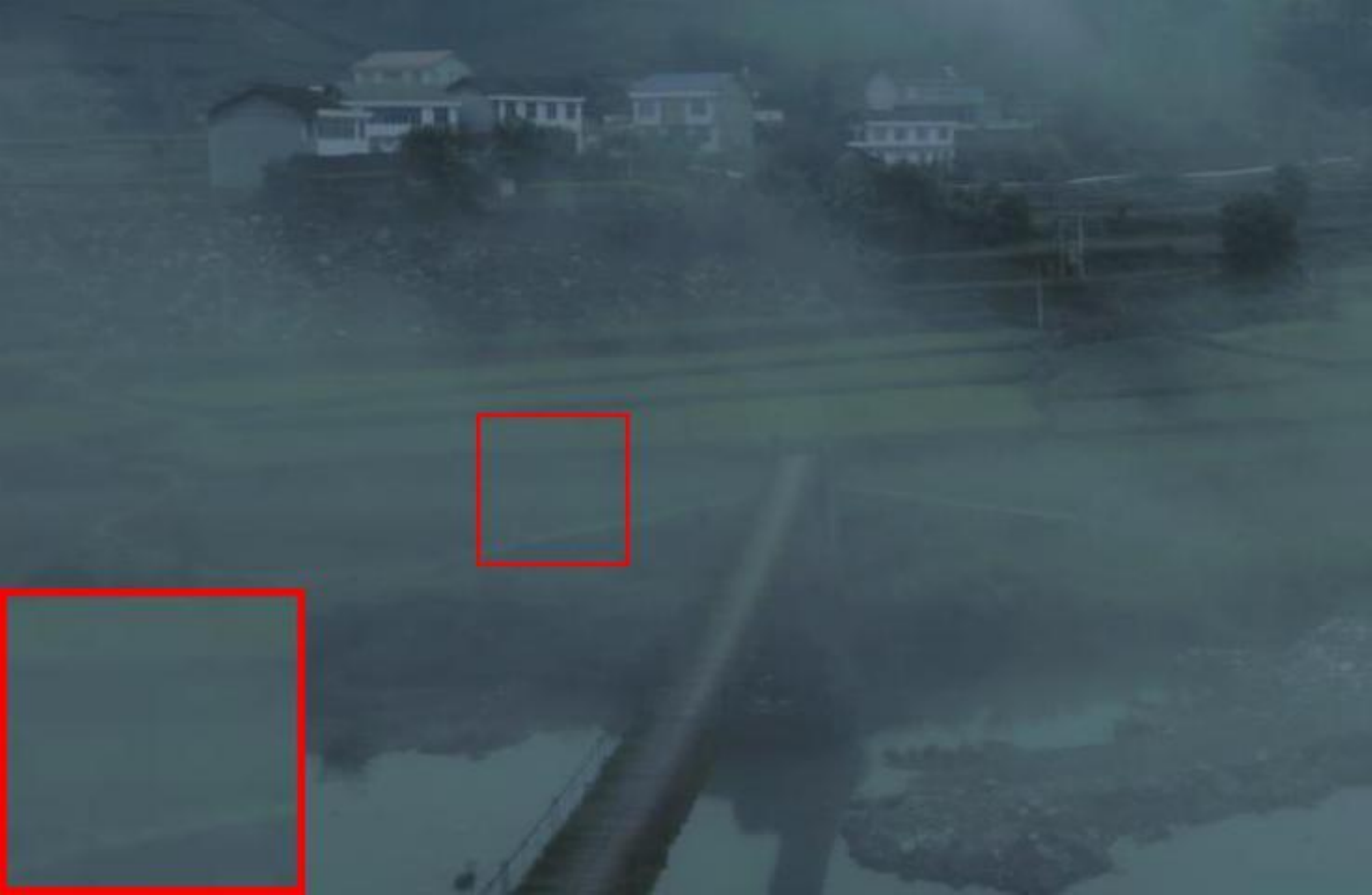}}
		\subfigure{\includegraphics[scale=\m_wid]{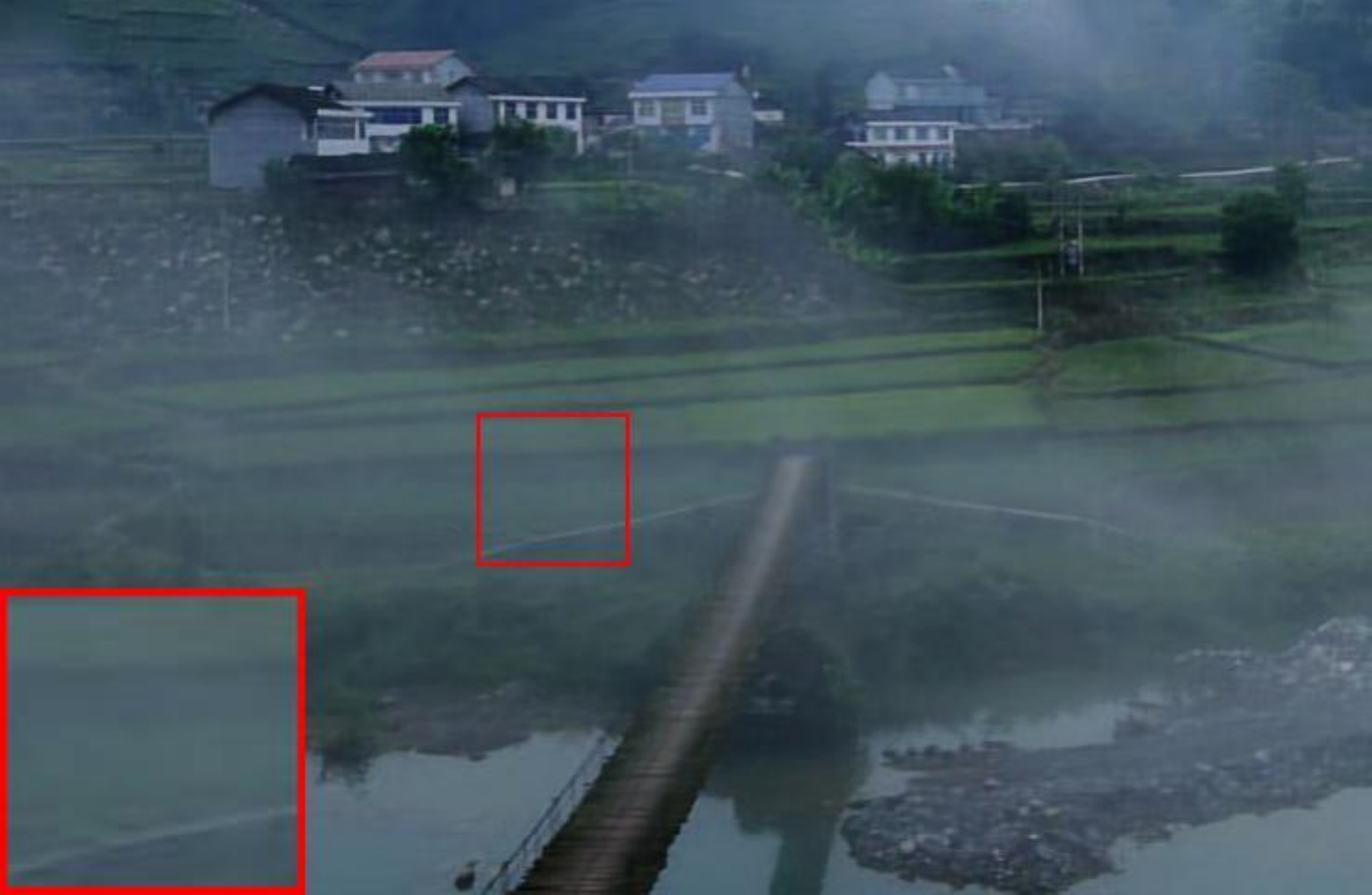}}
		\subfigure{\includegraphics[scale=\m_wid]{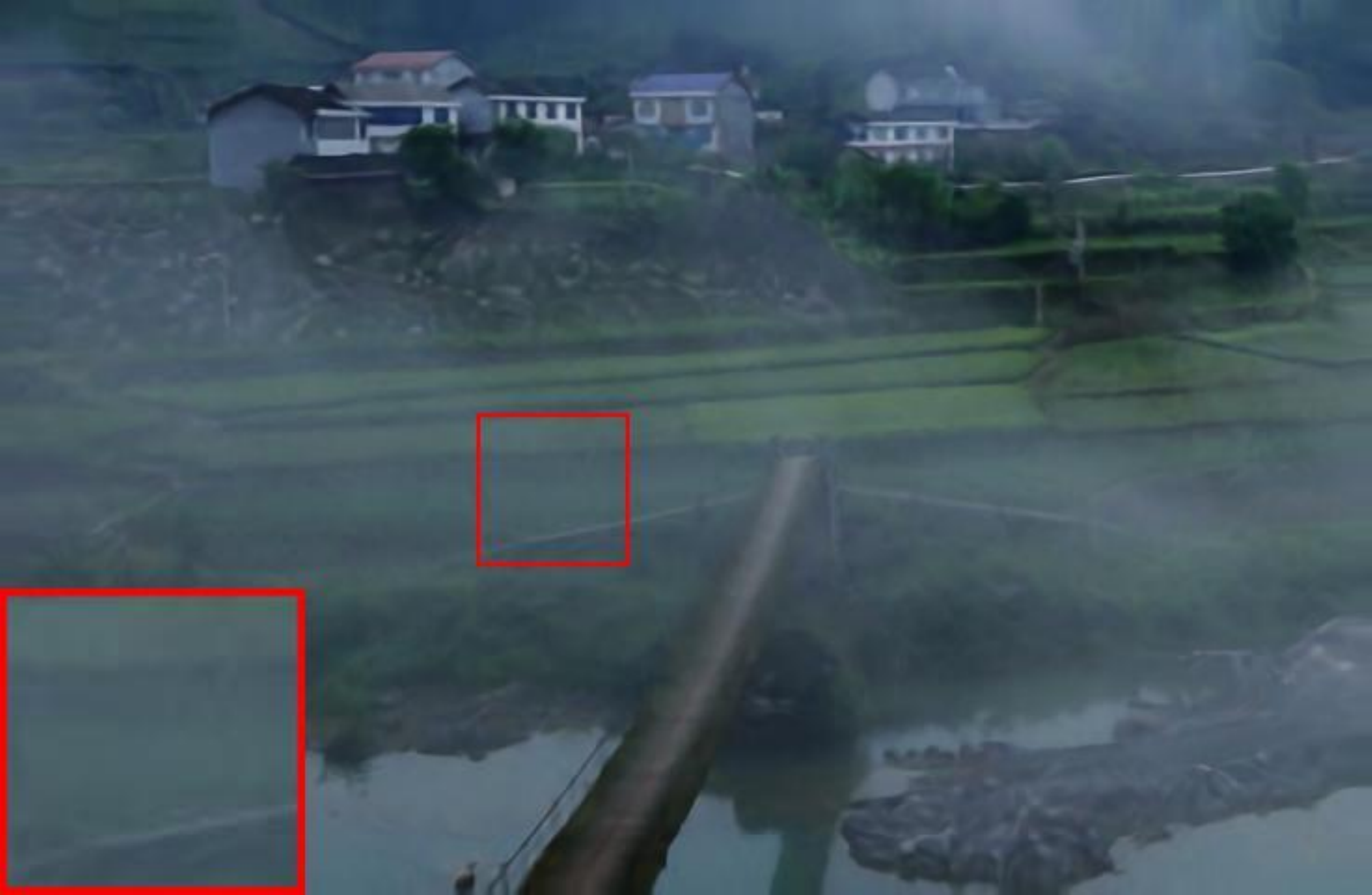}}
		\subfigure{\includegraphics[scale=\m_wid]{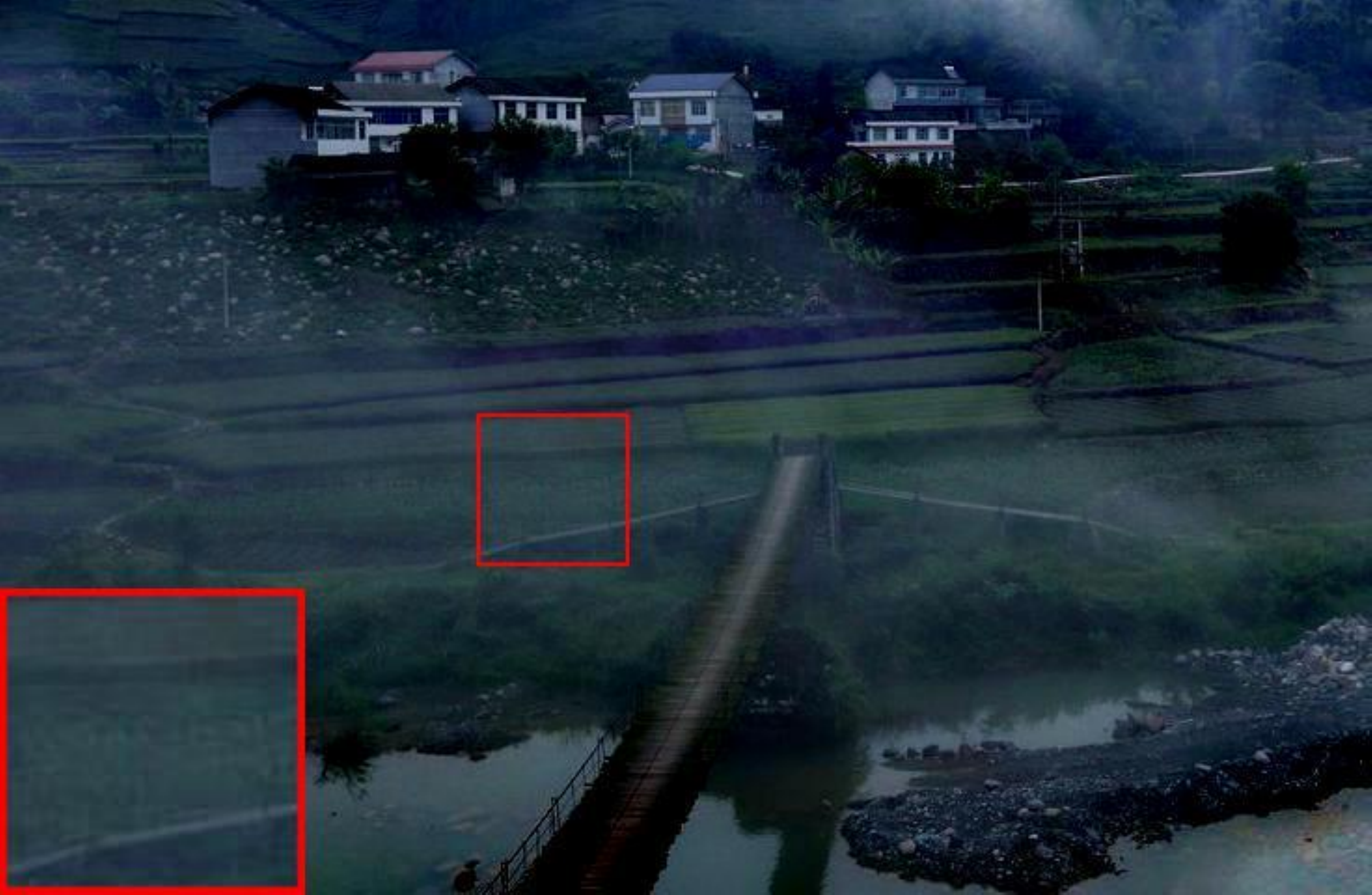}}
		\subfigure{\includegraphics[scale=\m_wid]{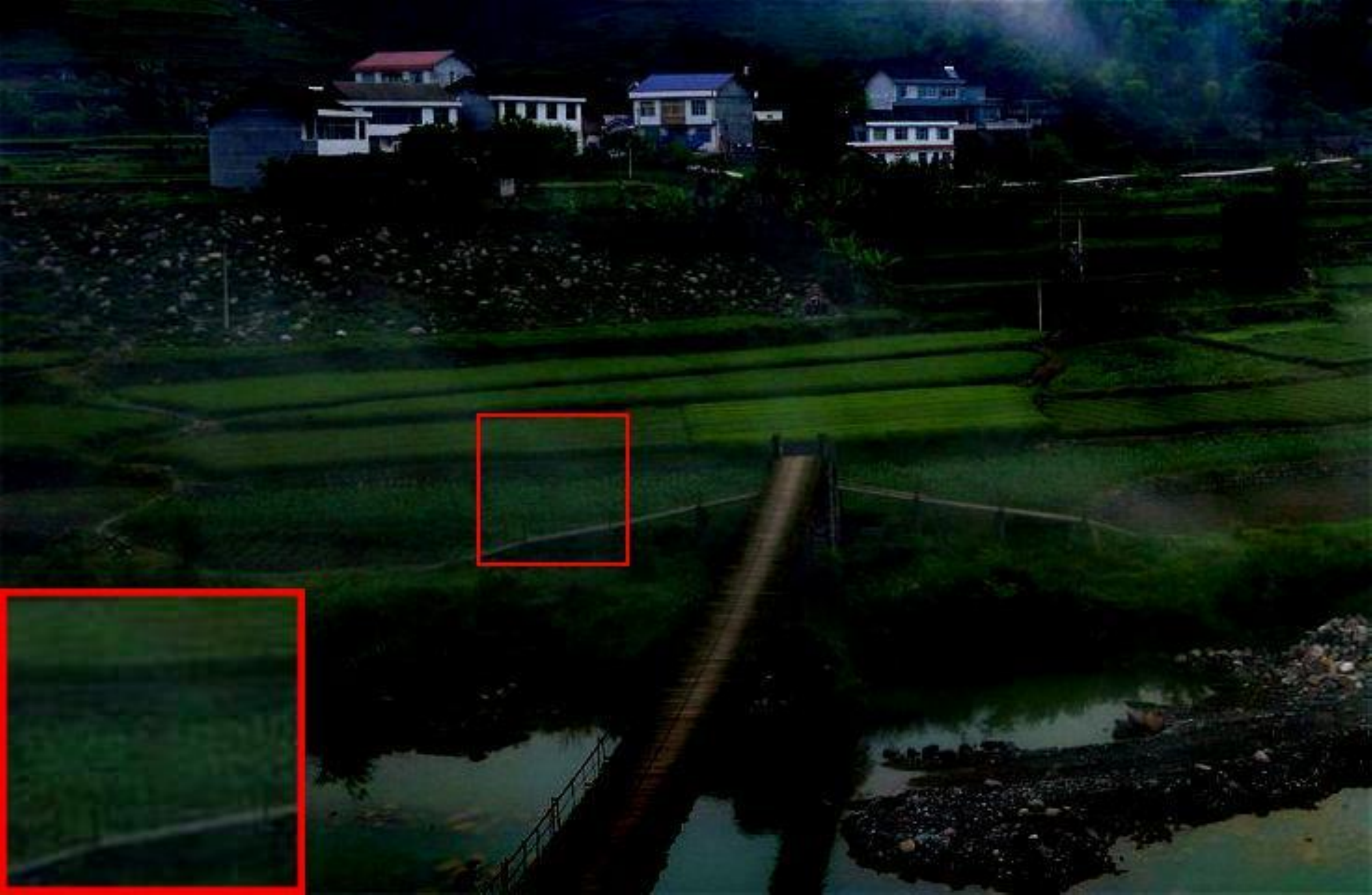}}
		\subfigure{\includegraphics[scale=\m_wid]{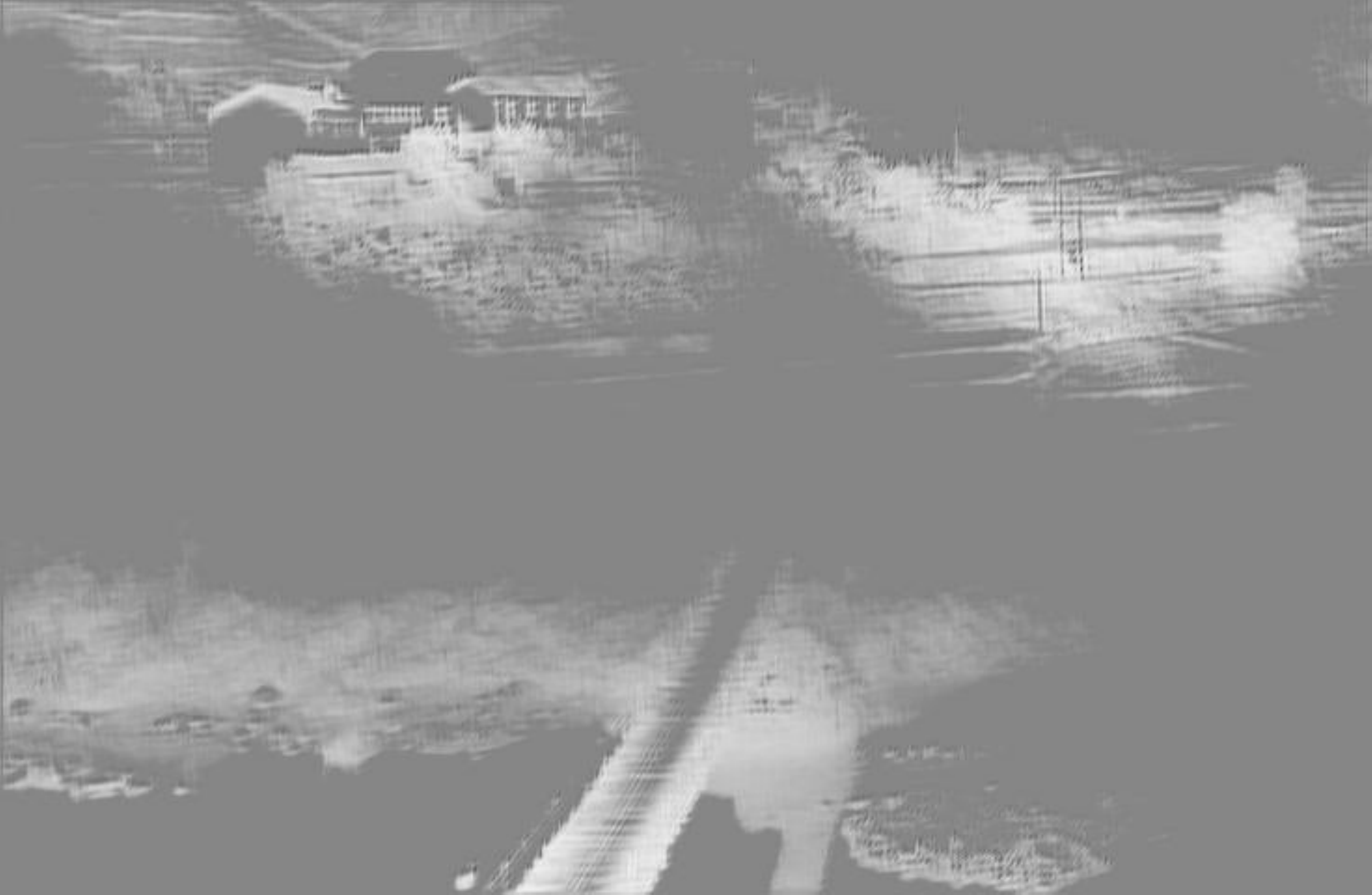}}
	\end{center}
	\vspace{-0.8cm}
		
	\begin{center}
	\def \m_wid{0.0591}
		\subfigure{\includegraphics[scale=\m_wid]{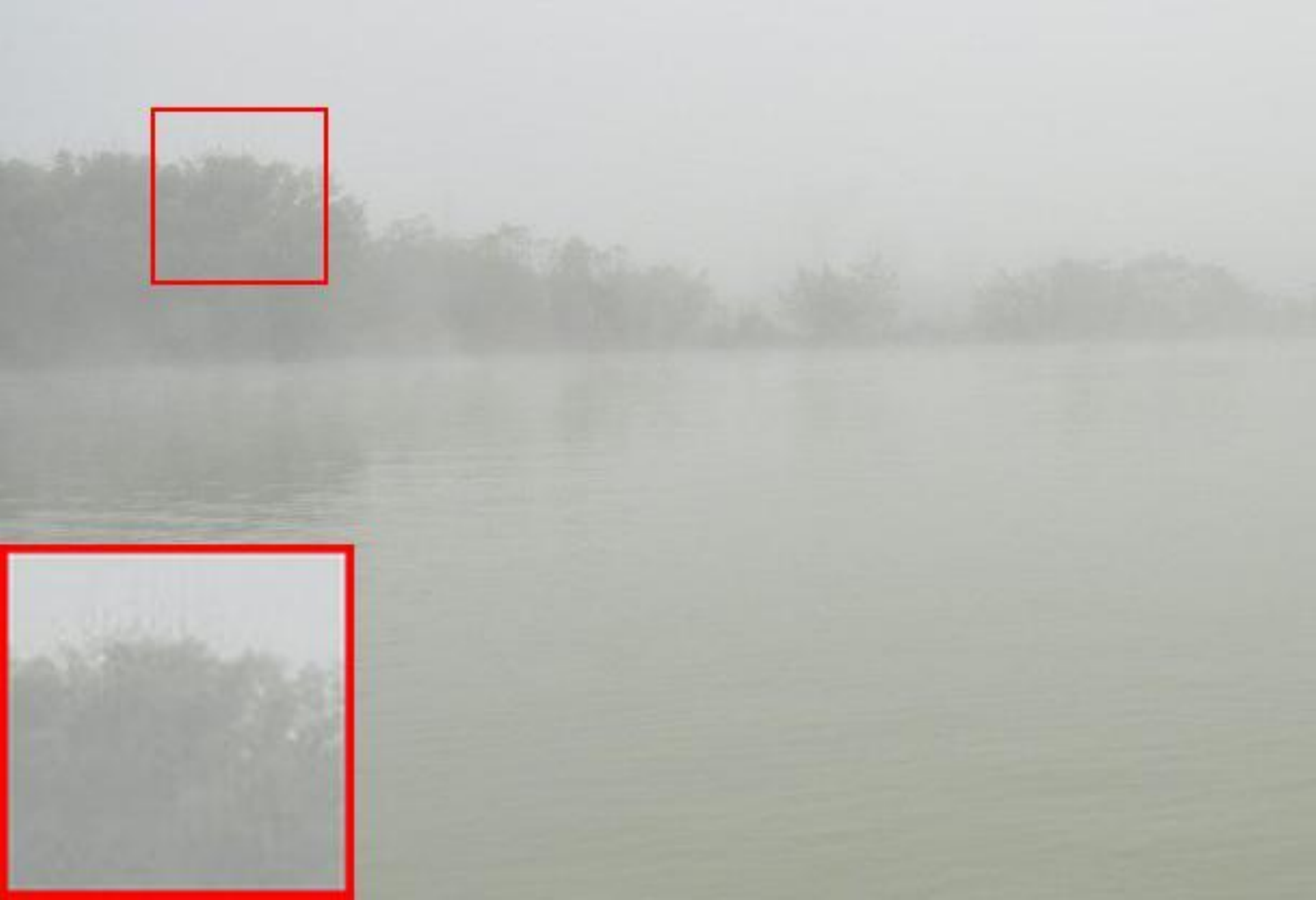}}
		\subfigure{\includegraphics[scale=\m_wid]{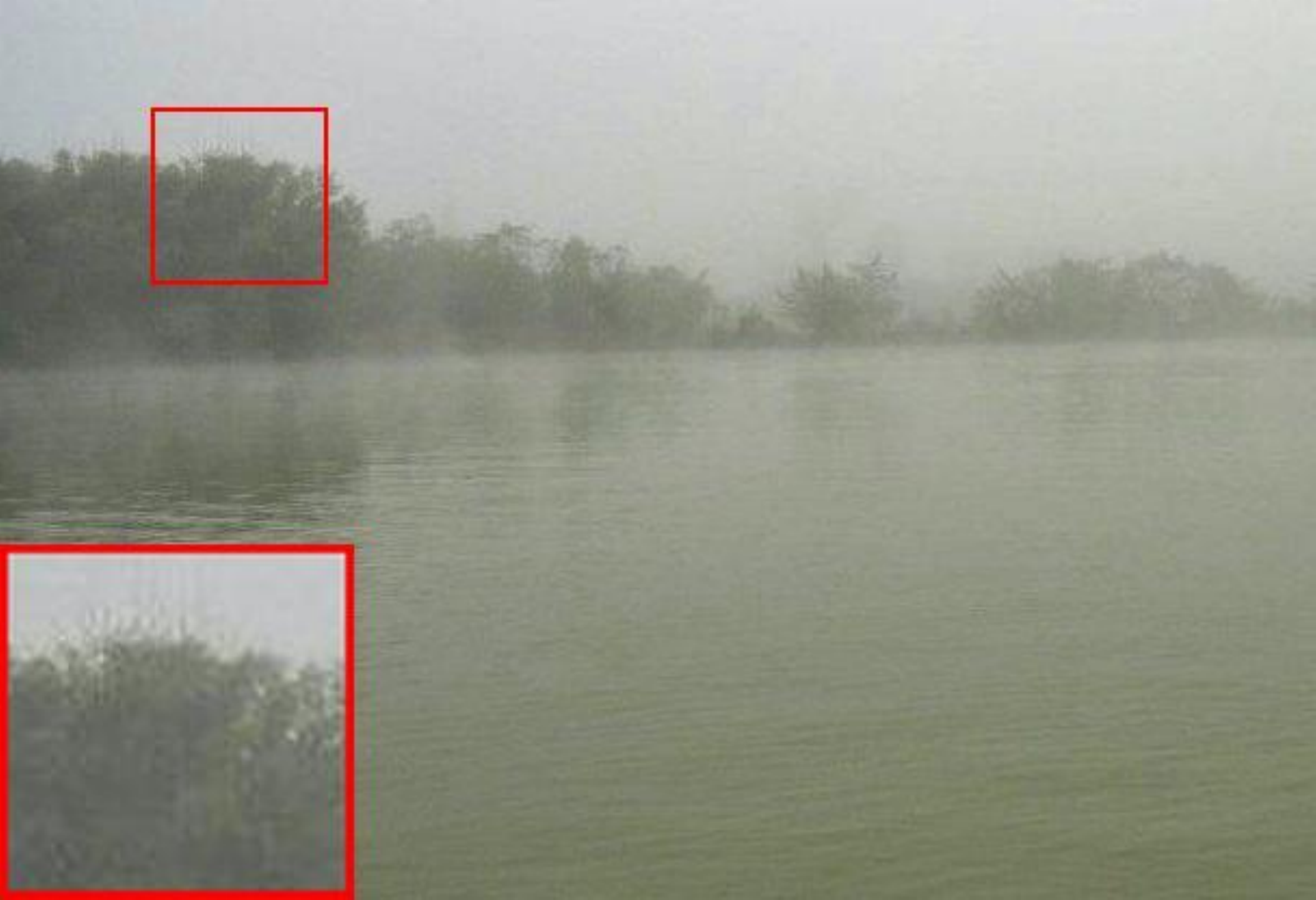}}
		\subfigure{\includegraphics[scale=\m_wid]{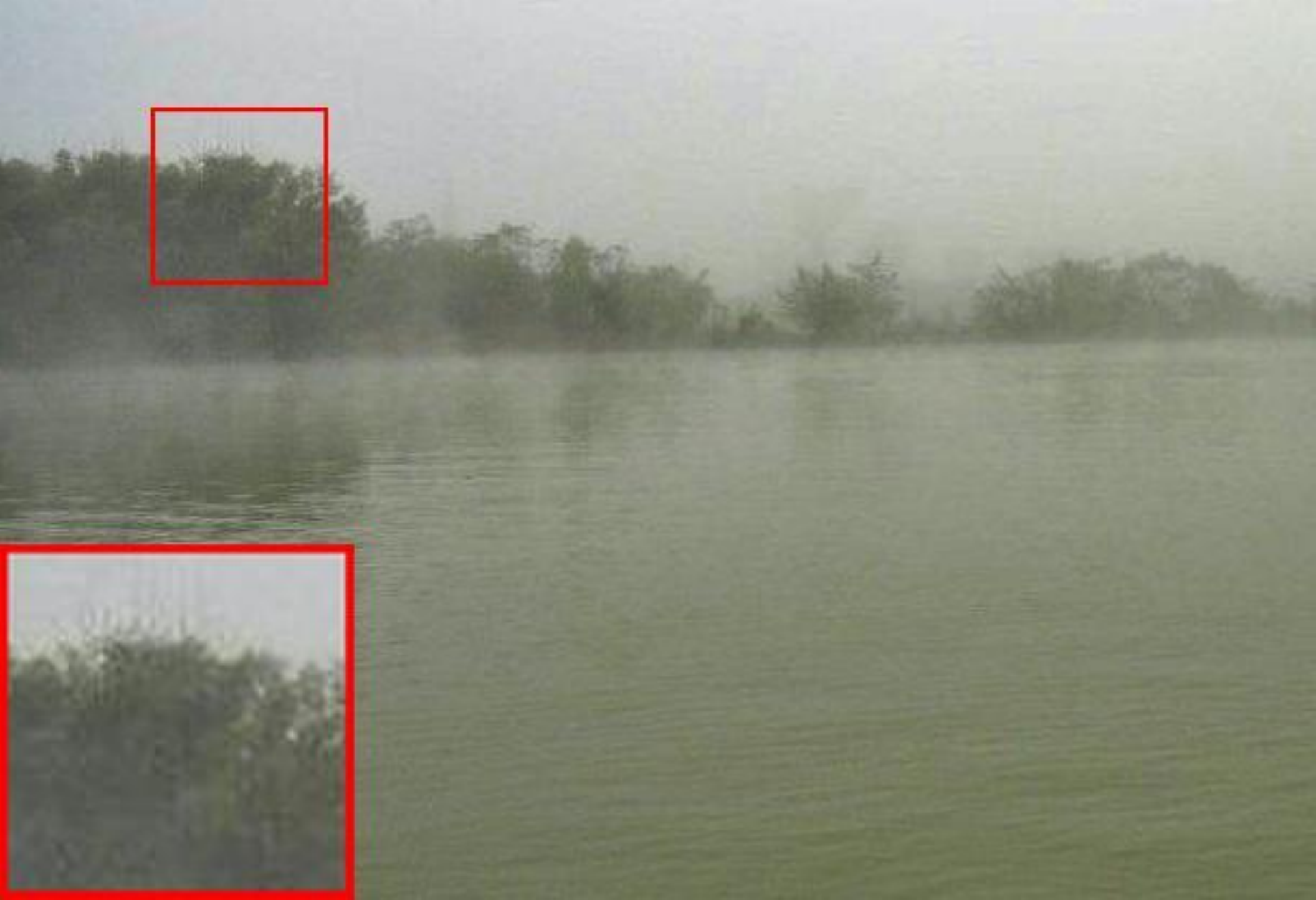}}
		\subfigure{\includegraphics[scale=\m_wid]{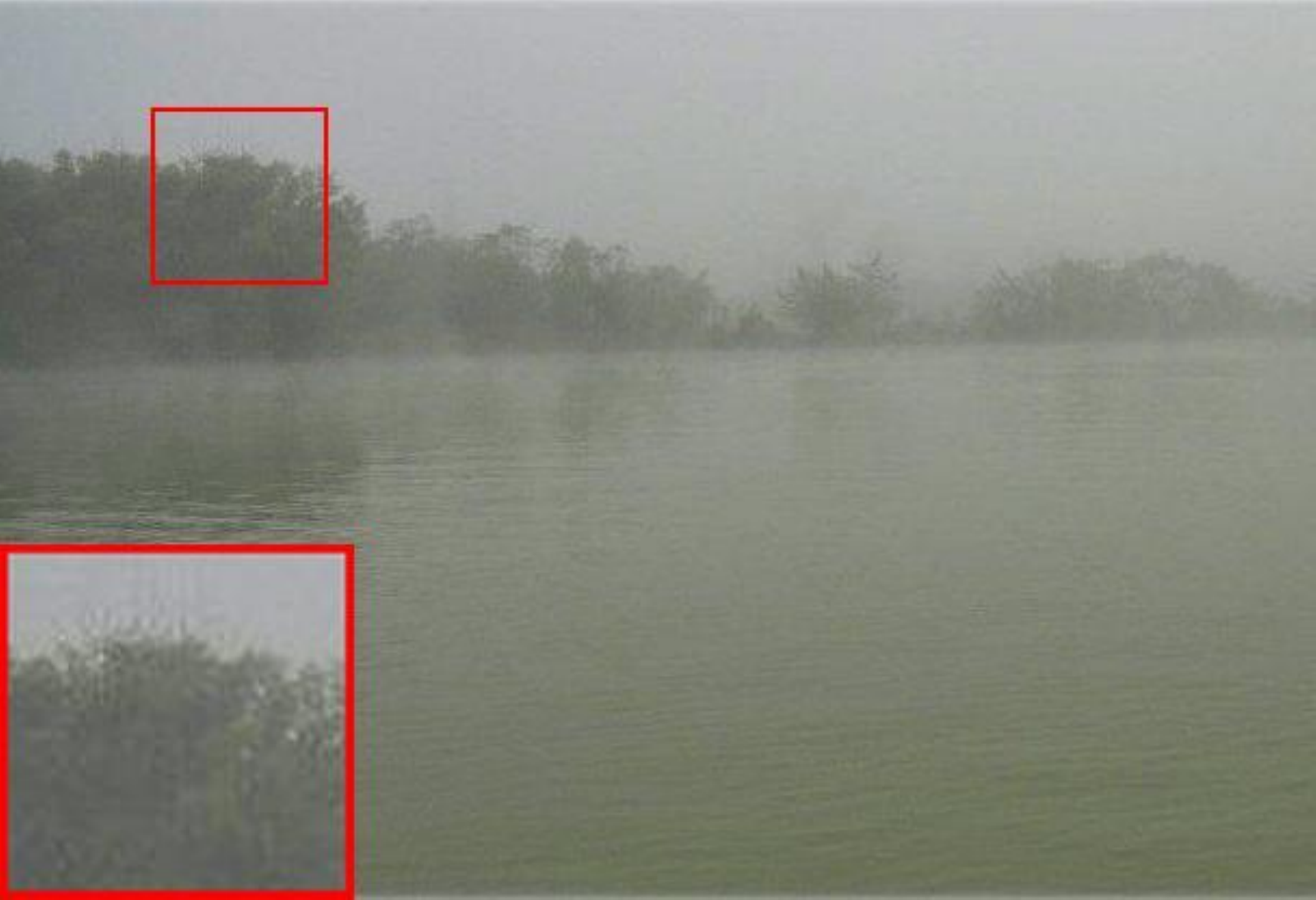}}
		\subfigure{\includegraphics[scale=\m_wid]{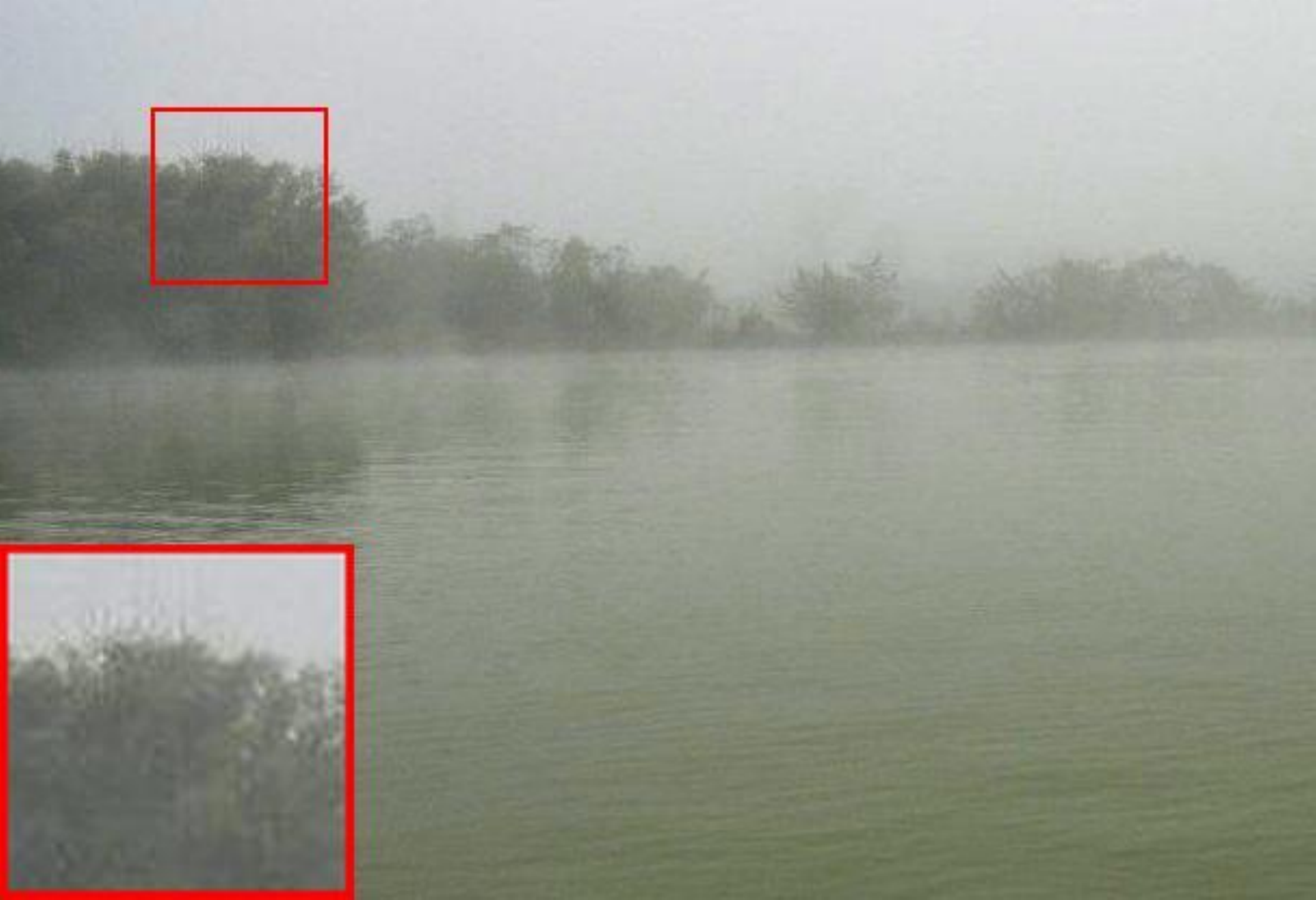}}
		\subfigure{\includegraphics[scale=\m_wid]{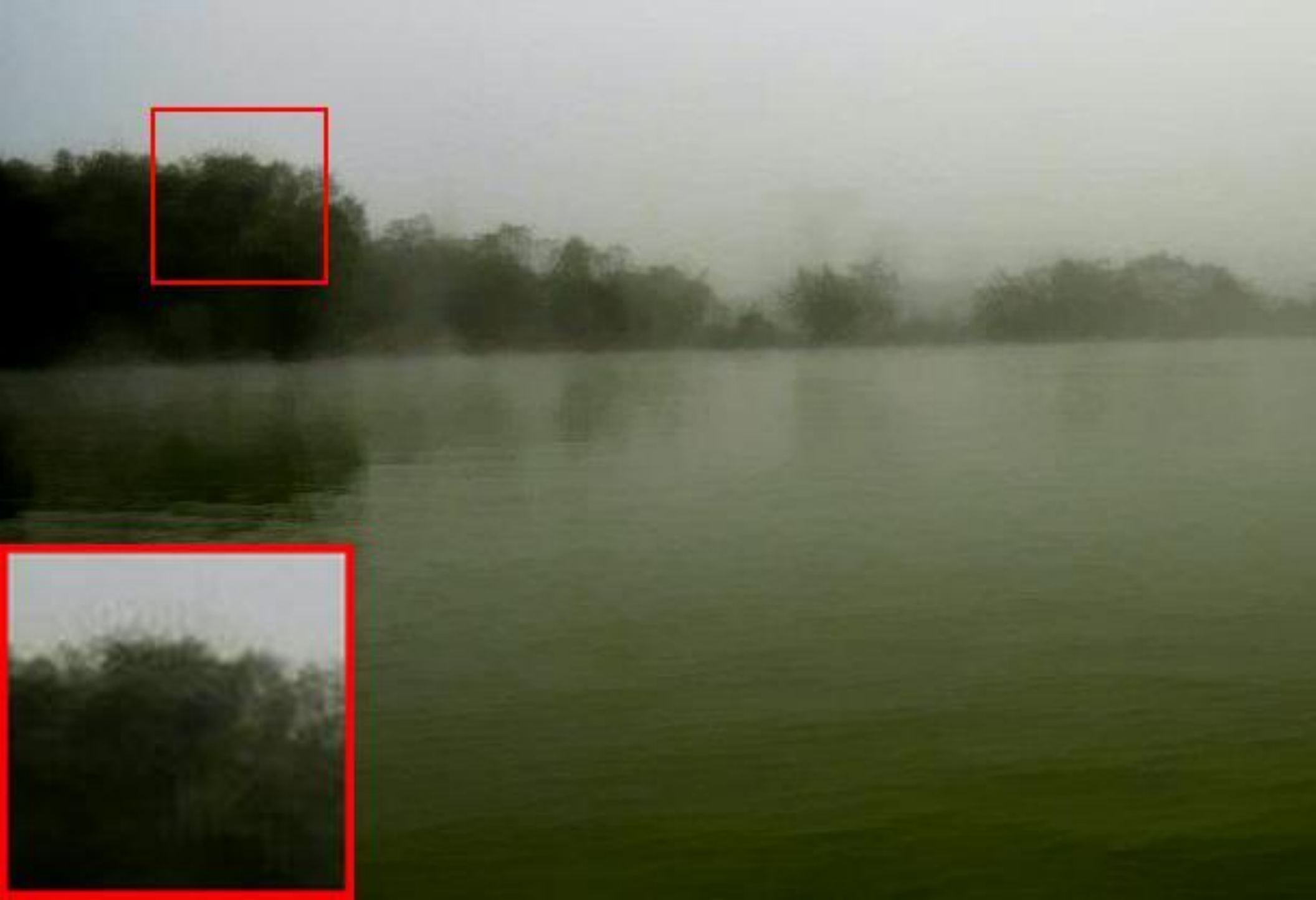}}
		\subfigure{\includegraphics[scale=\m_wid]{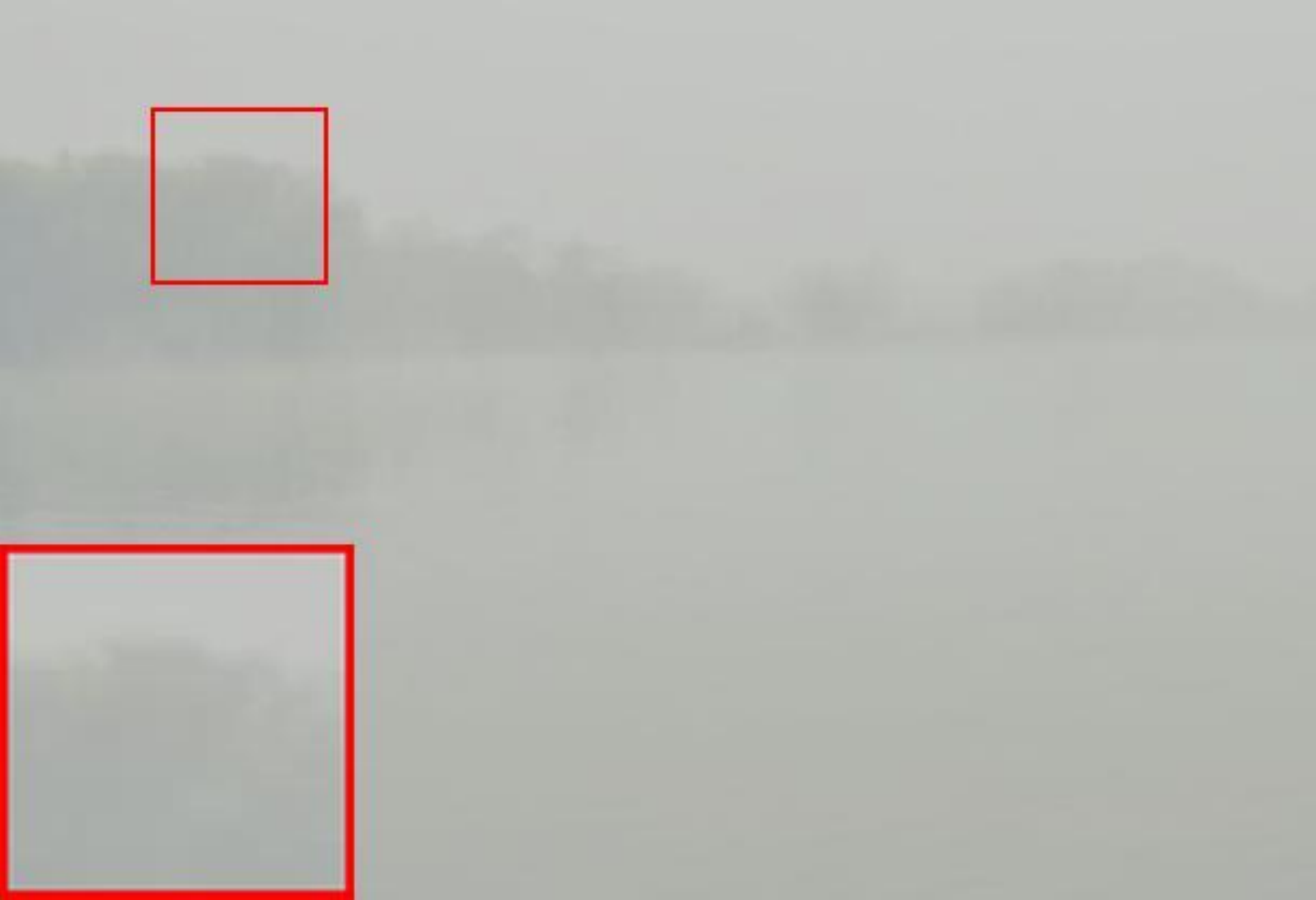}}
		\subfigure{\includegraphics[scale=\m_wid]{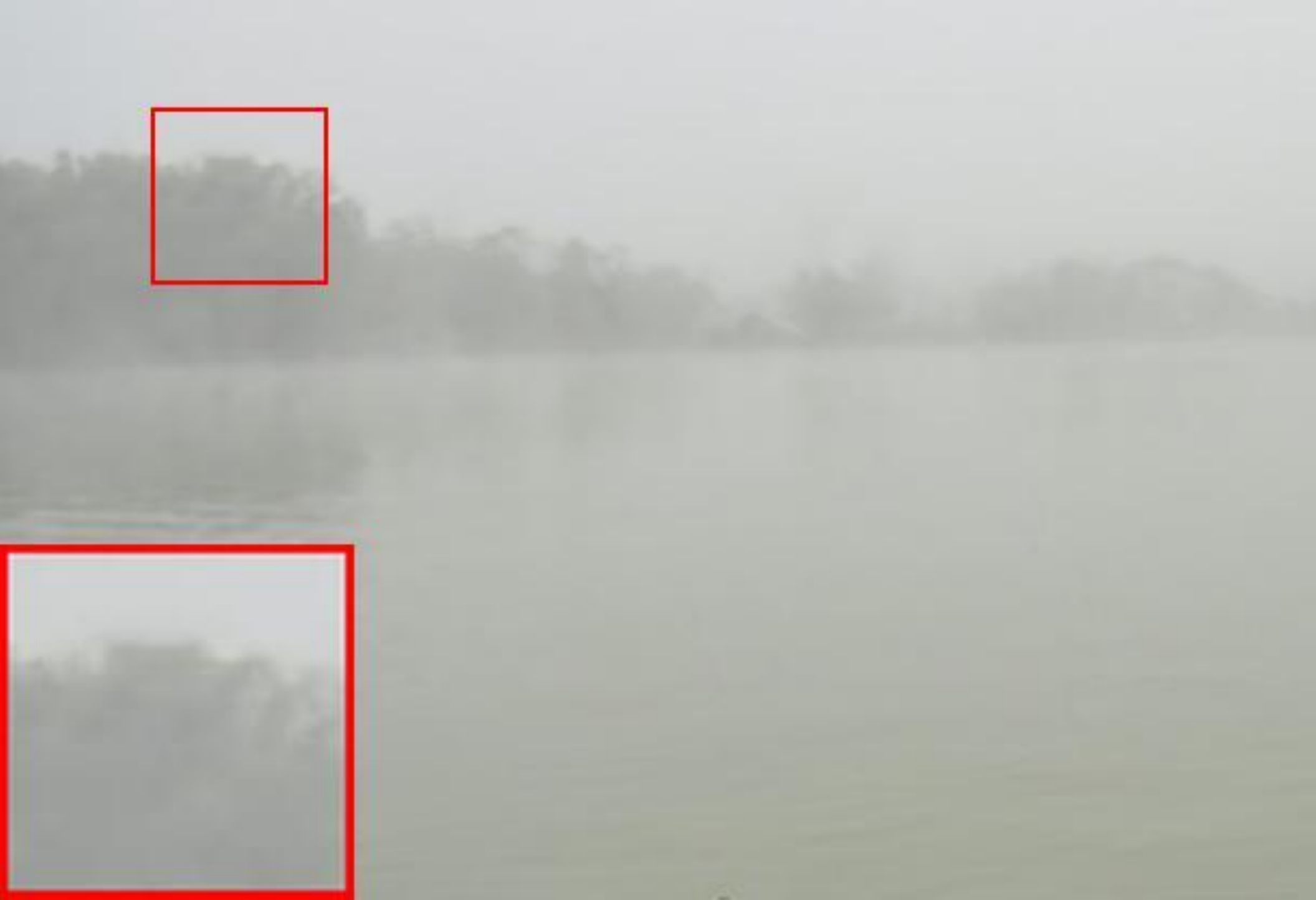}}
		\subfigure{\includegraphics[scale=\m_wid]{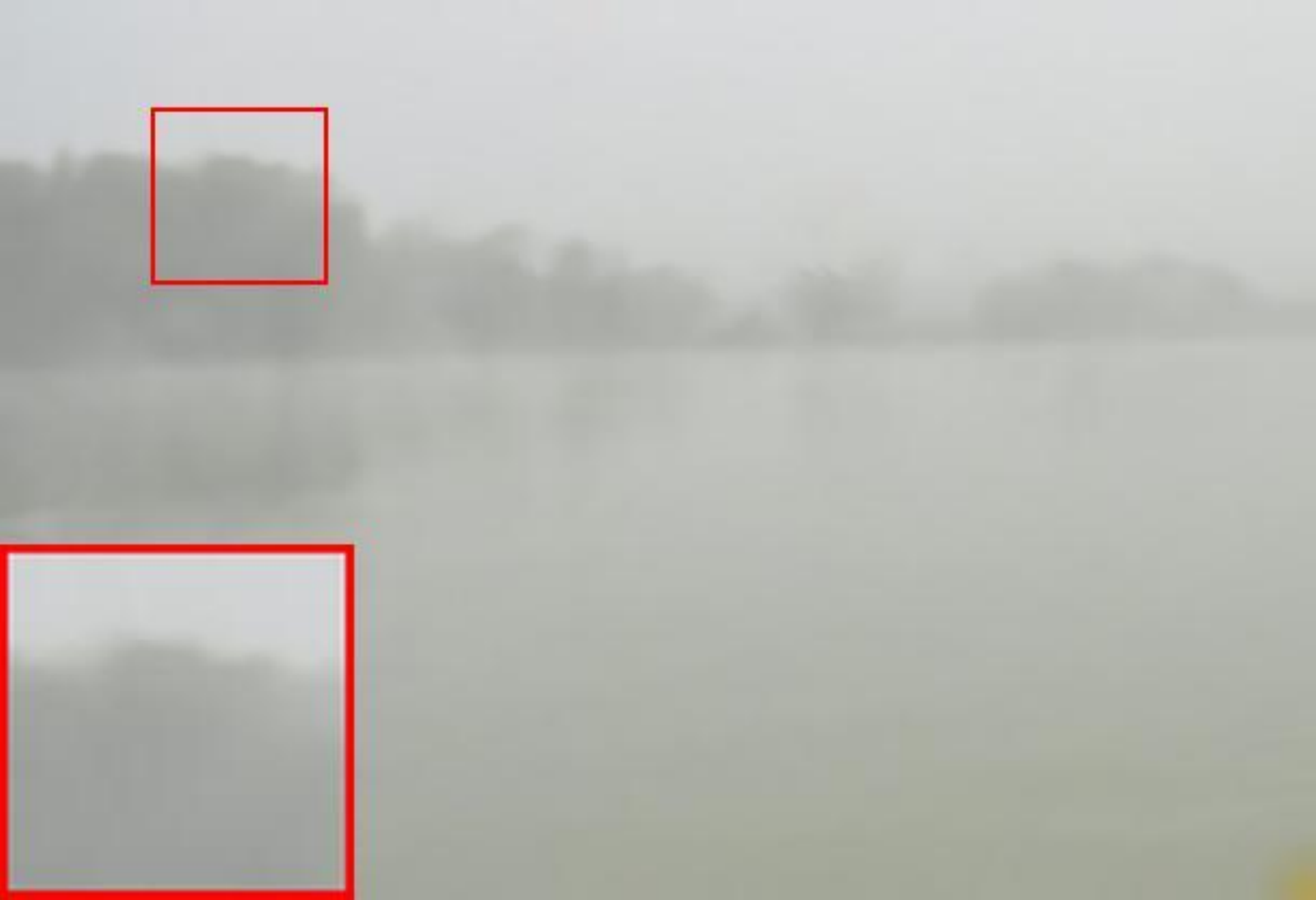}}
		\subfigure{\includegraphics[scale=\m_wid]{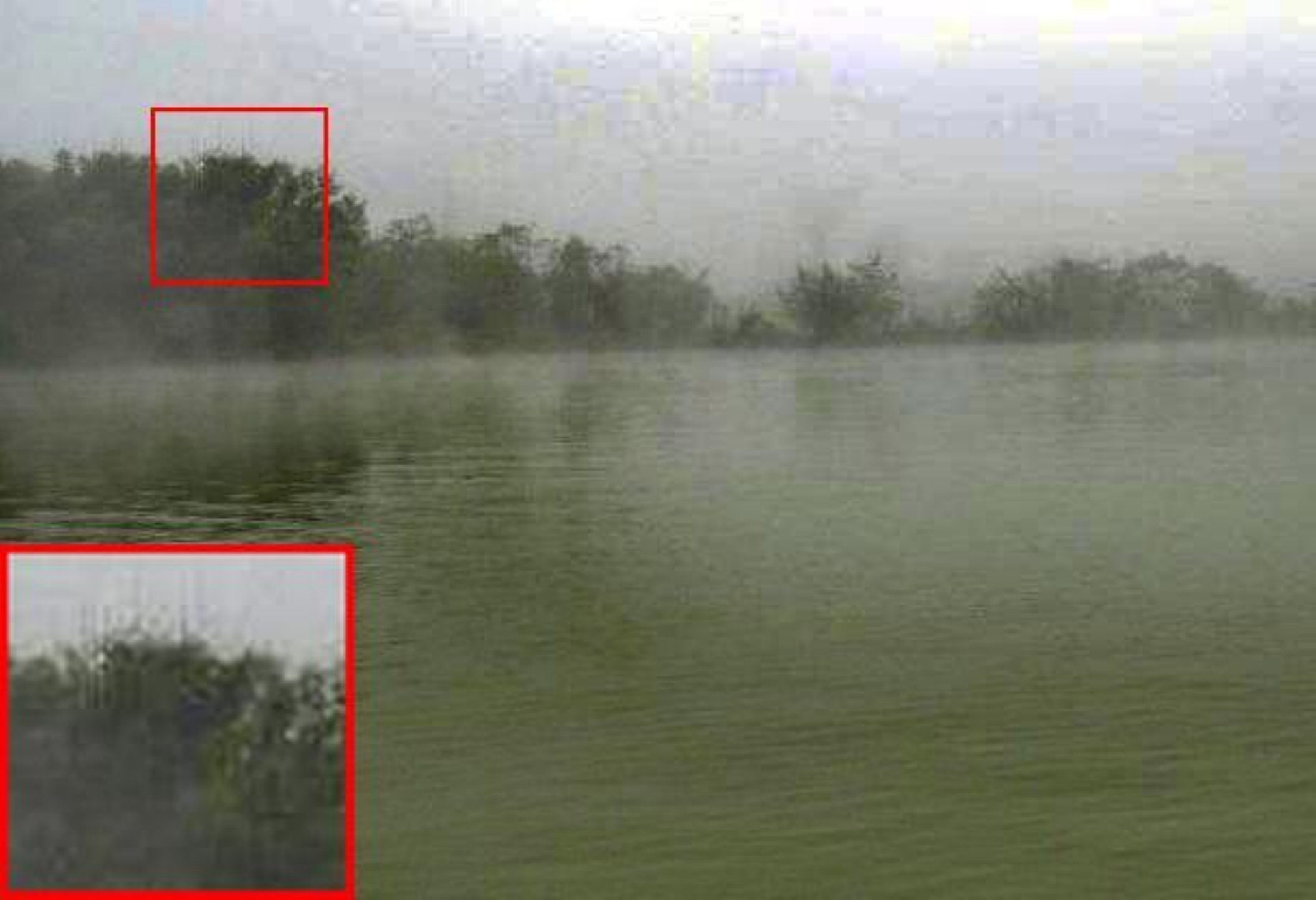}}
		\subfigure{\includegraphics[scale=\m_wid]{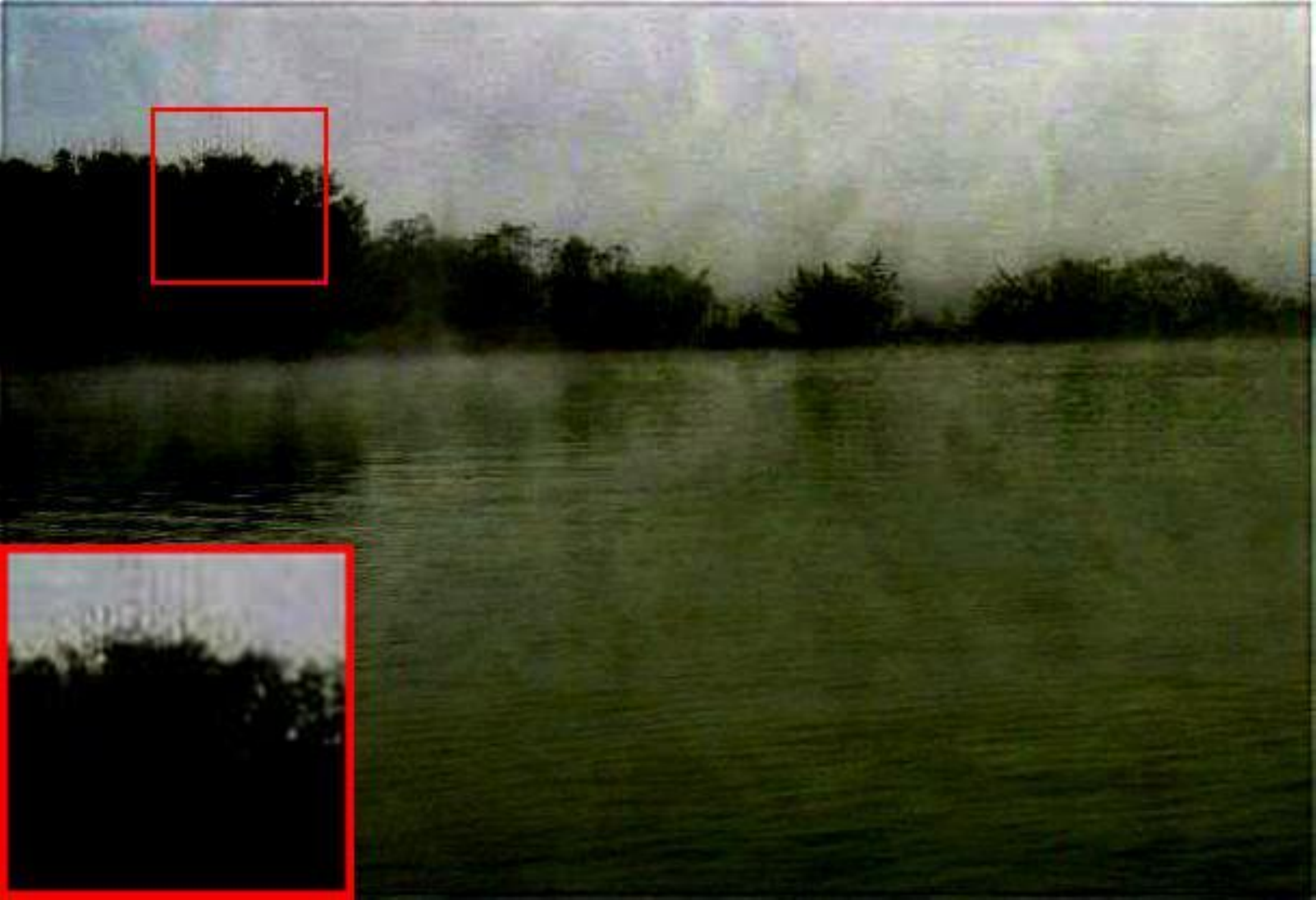}}
		\subfigure{\includegraphics[scale=\m_wid]{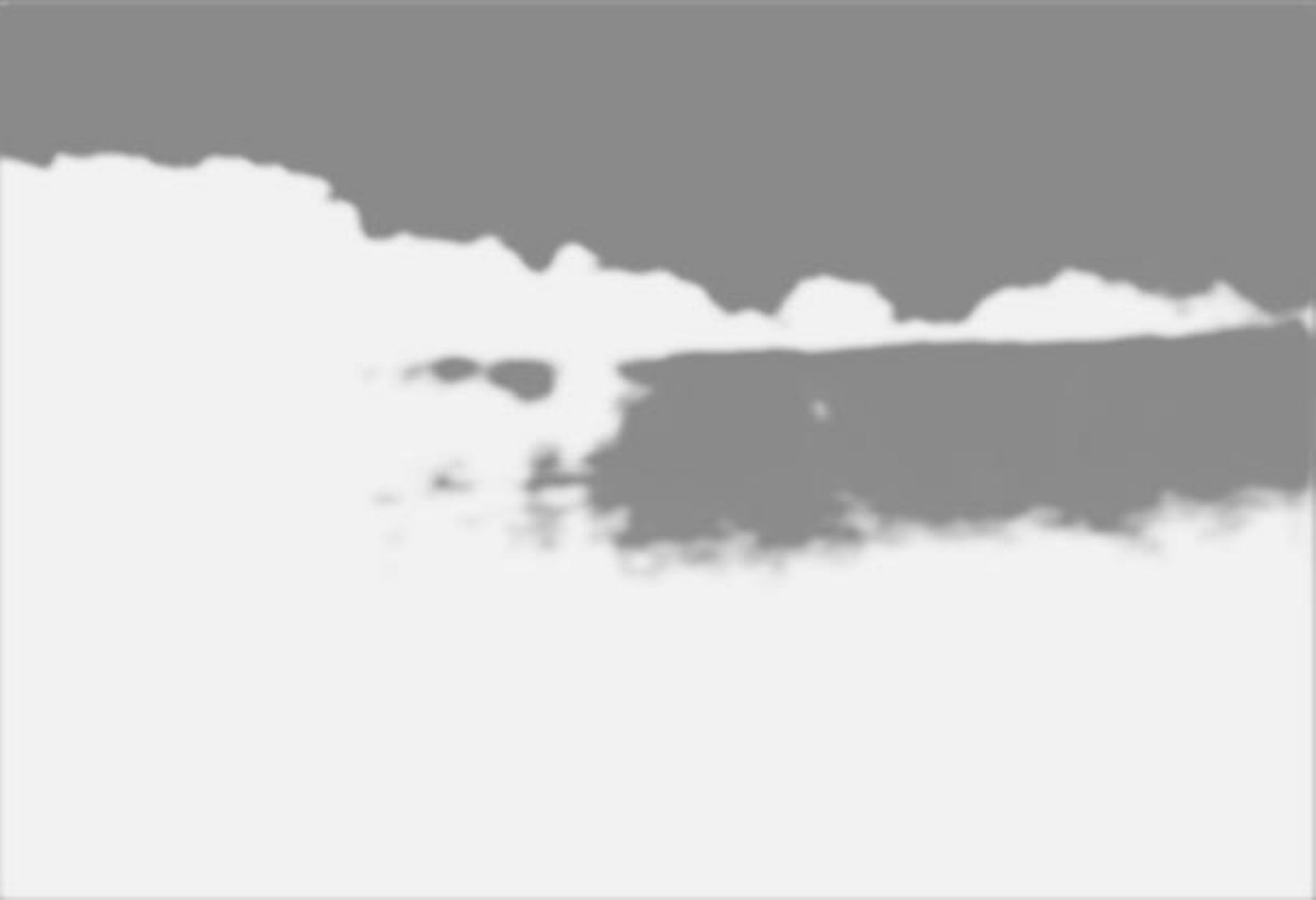}}
	\end{center}
	\vspace{-0.8cm}
	
	\begin{center}
	\def \m_wid{0.0245}
	\def \two_wid{0.0489}
		\subfigure{\includegraphics[scale=\m_wid]{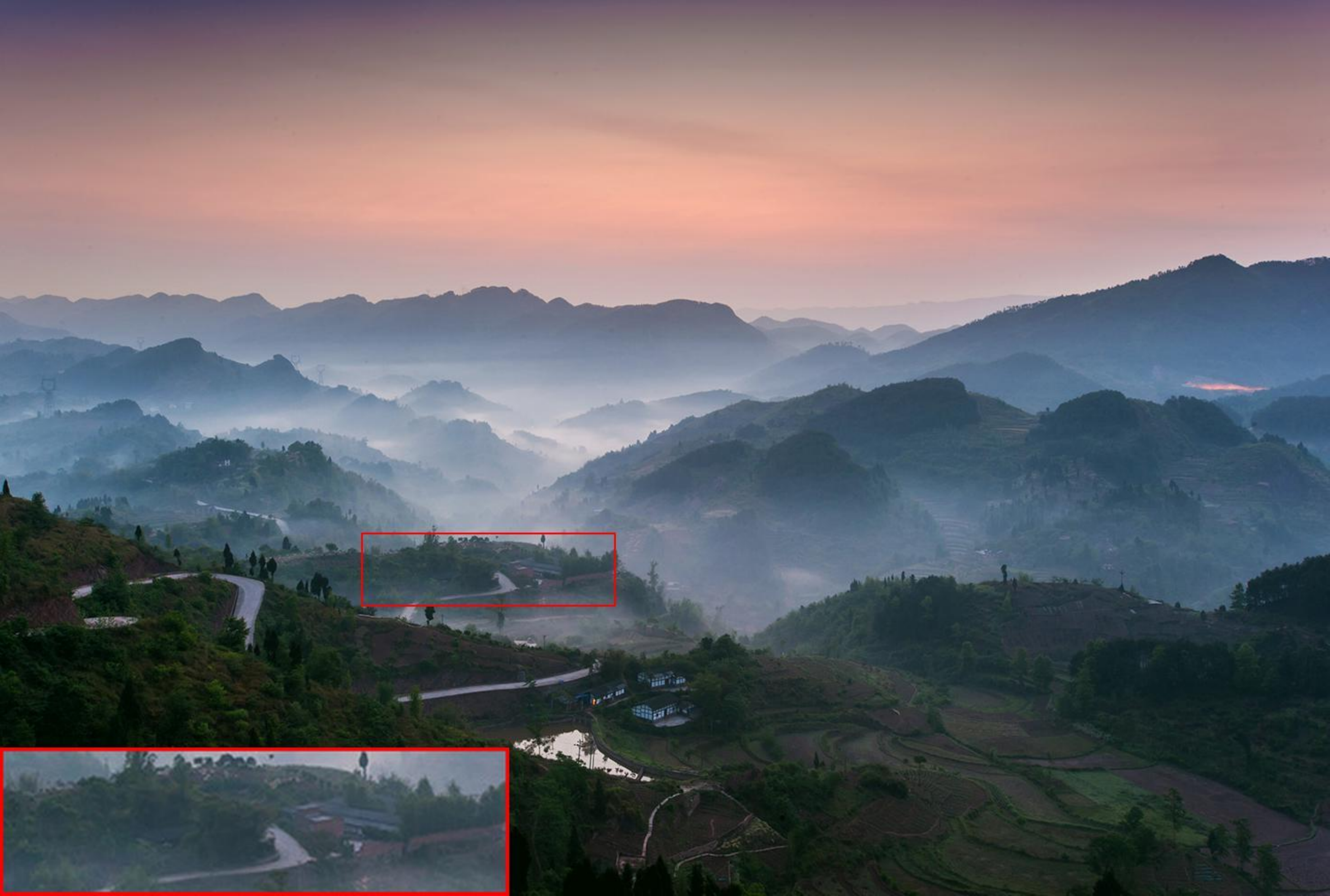}}
		\subfigure{\includegraphics[scale=\m_wid]{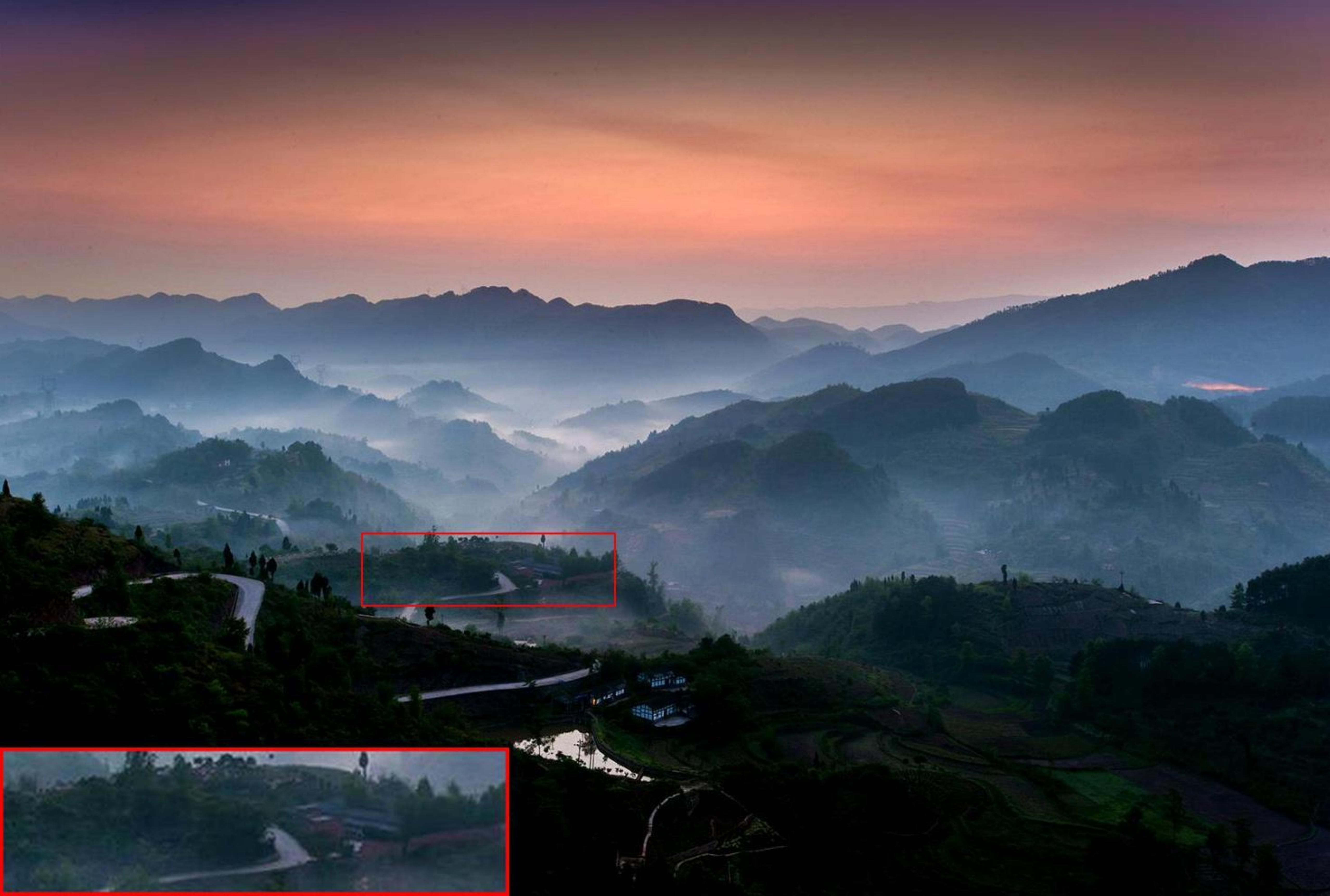}}
		\subfigure{\includegraphics[scale=\m_wid]{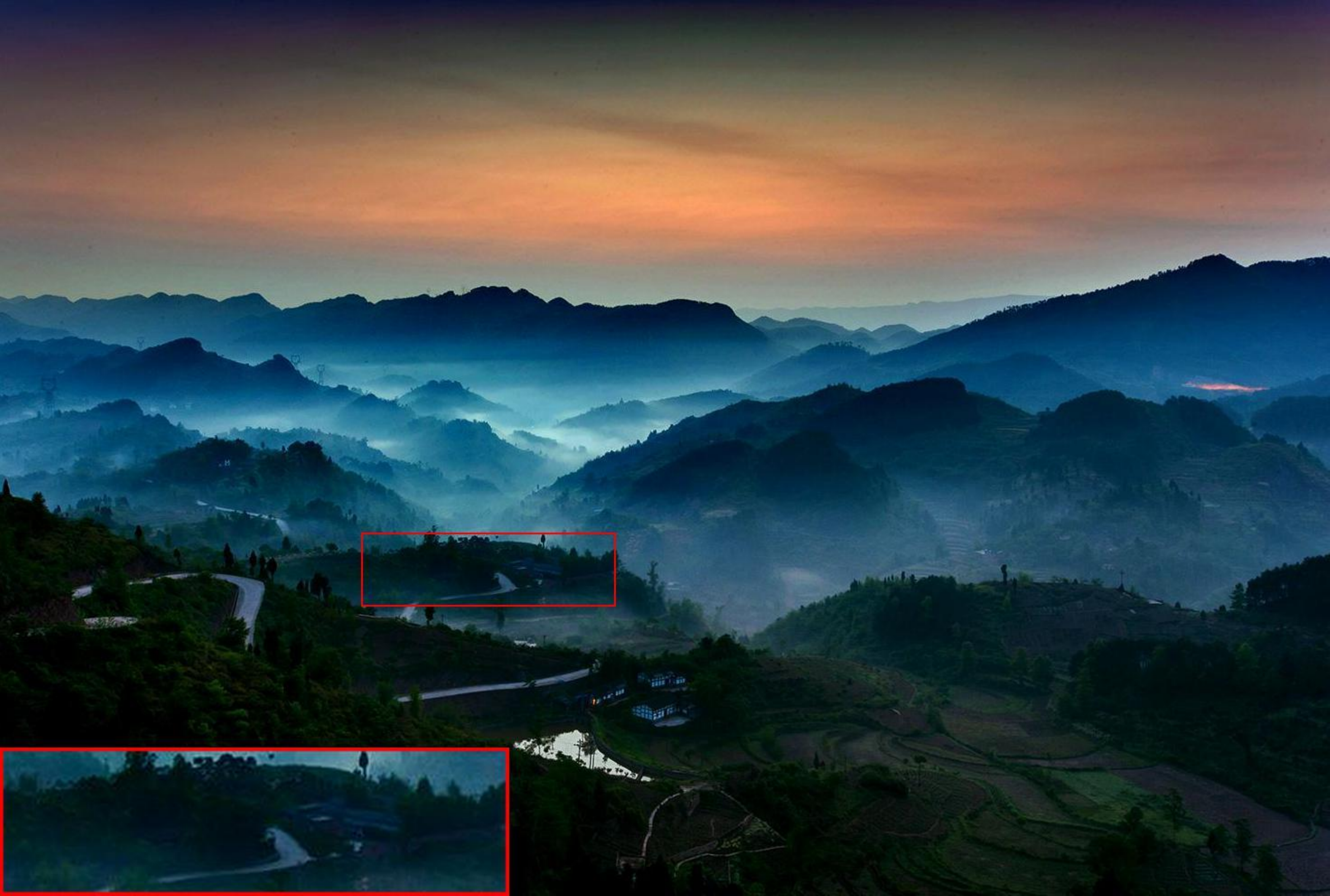}}
		\subfigure{\includegraphics[scale=\m_wid]{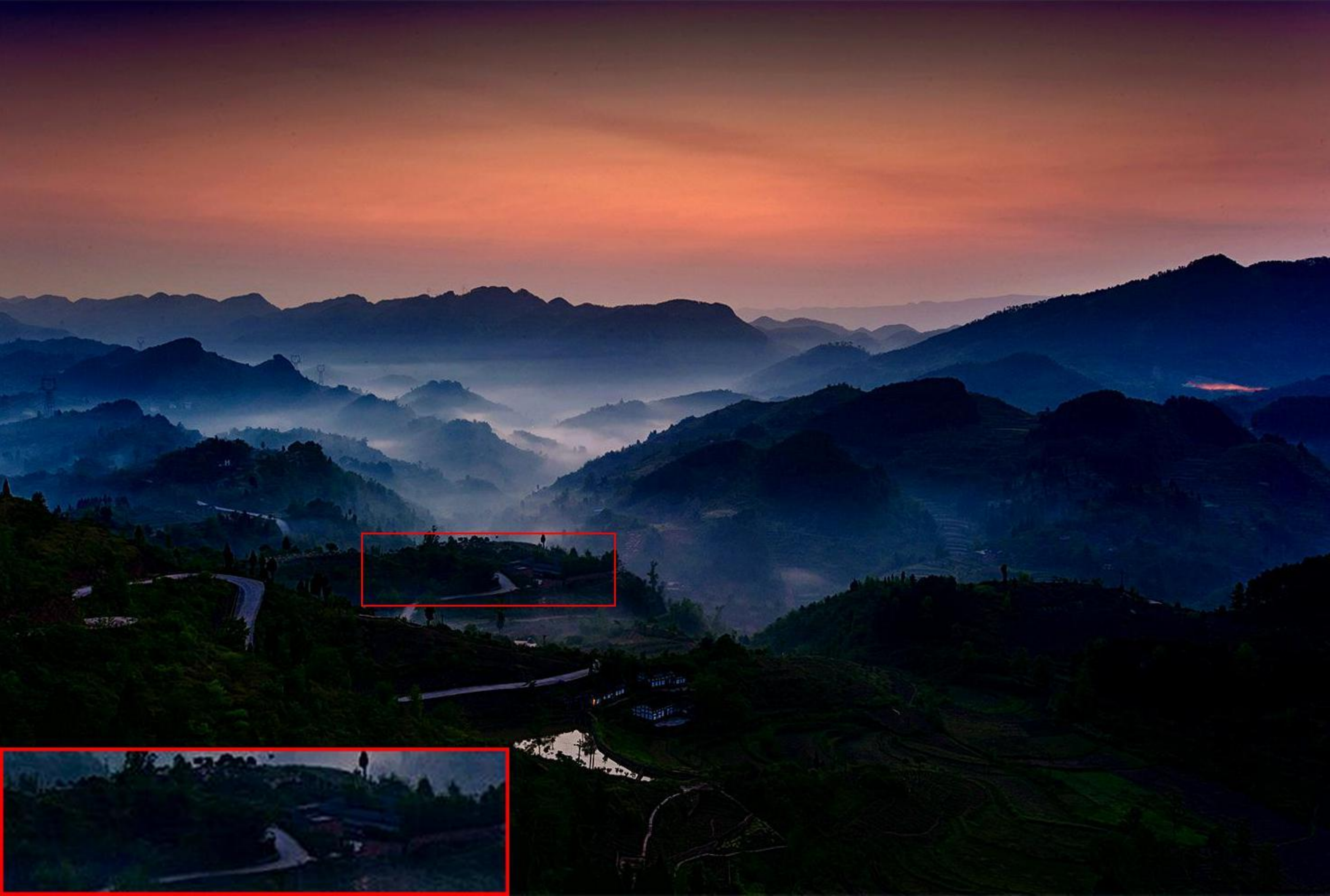}}
		\subfigure{\includegraphics[scale=\m_wid]{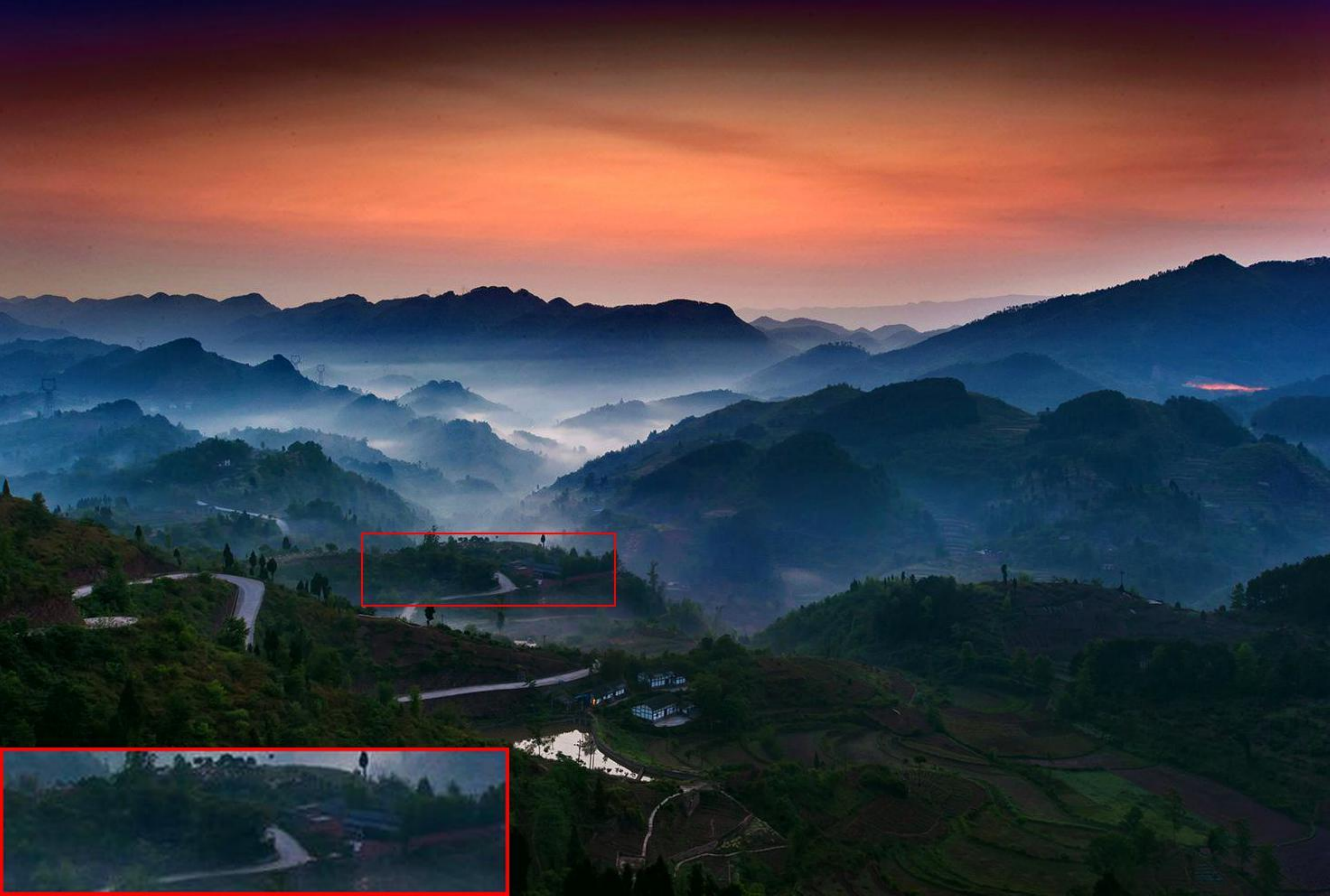}}
		\subfigure{\includegraphics[scale=\m_wid]{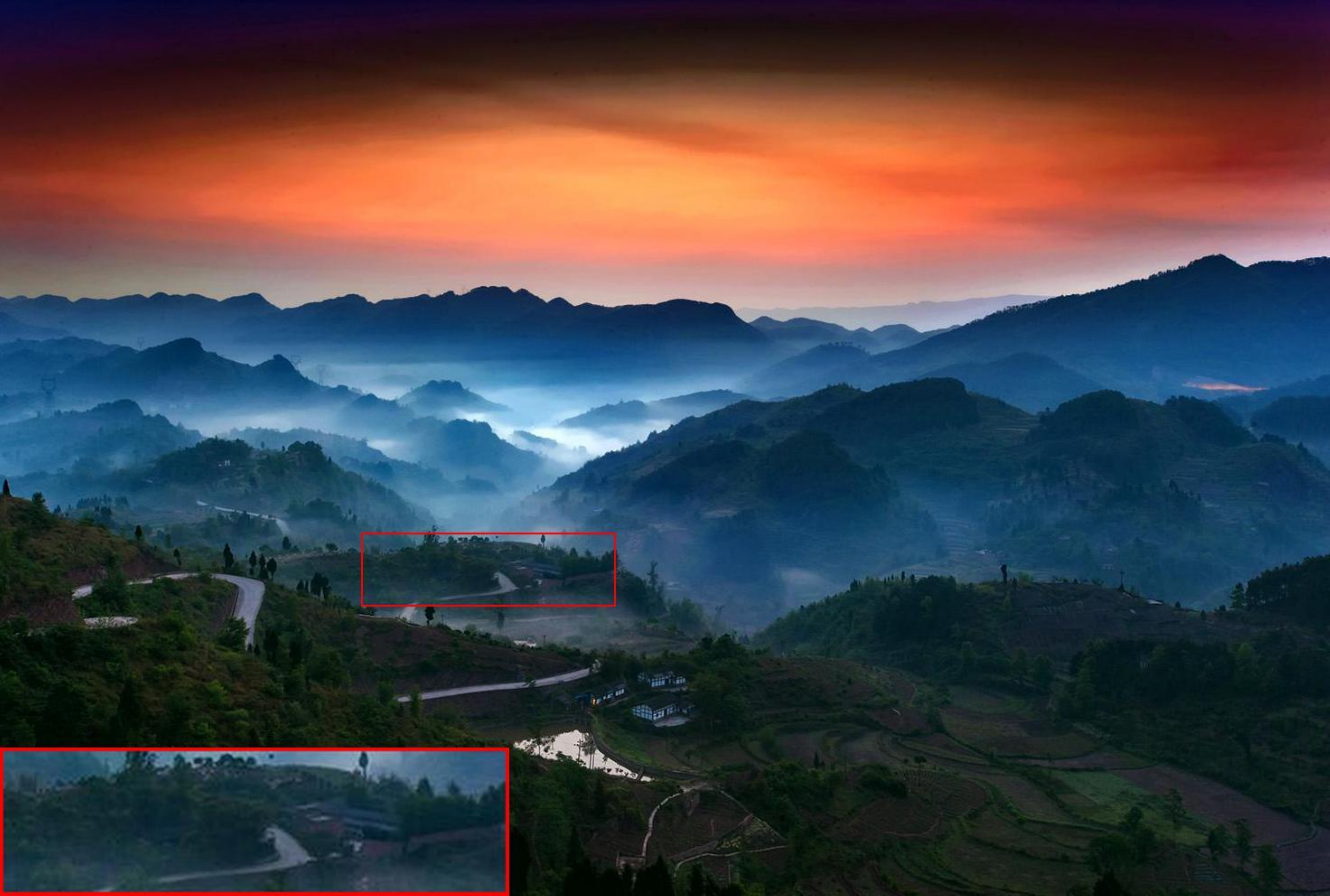}}
		\subfigure{\includegraphics[scale=\m_wid]{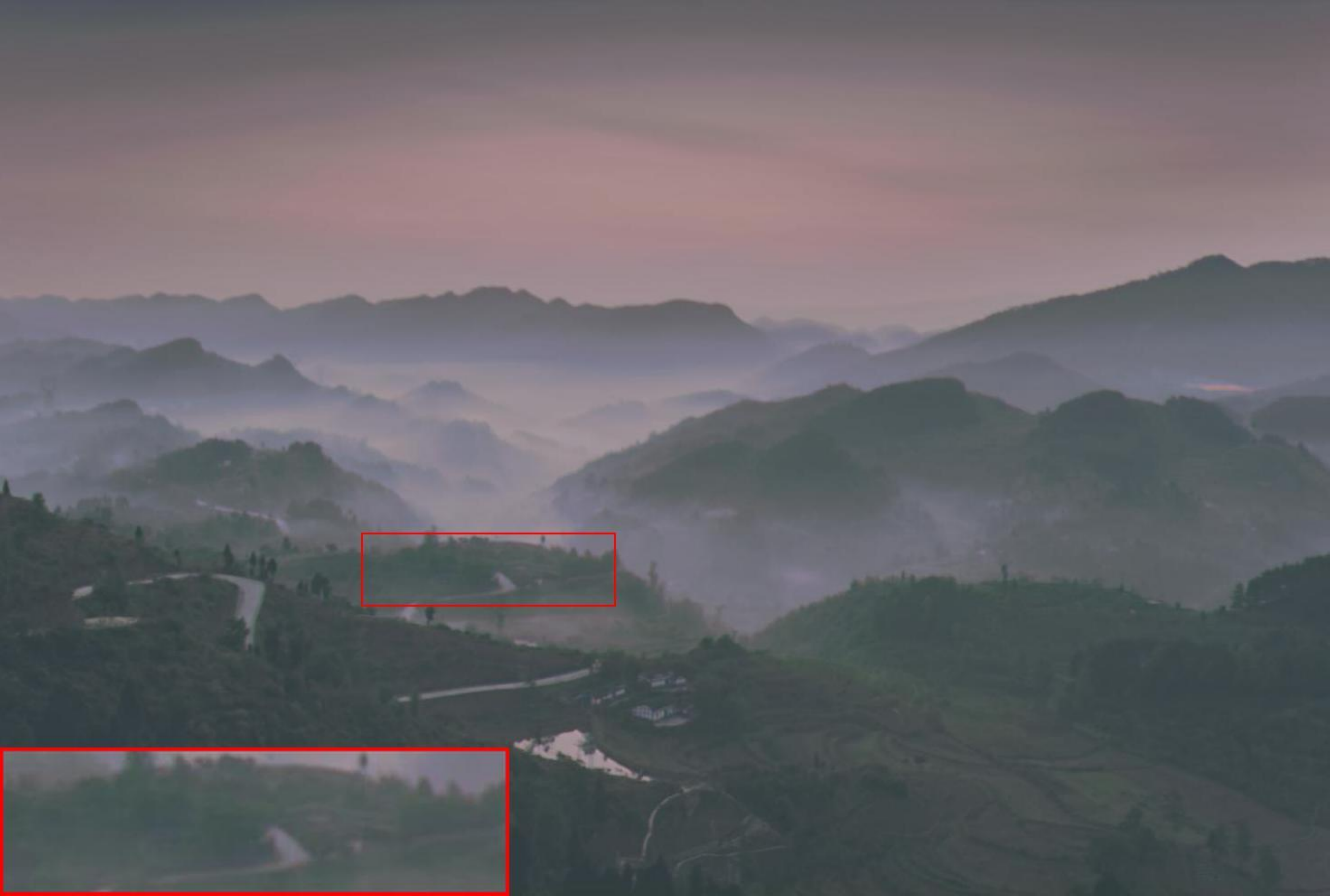}}
		\subfigure{\includegraphics[scale=\m_wid]{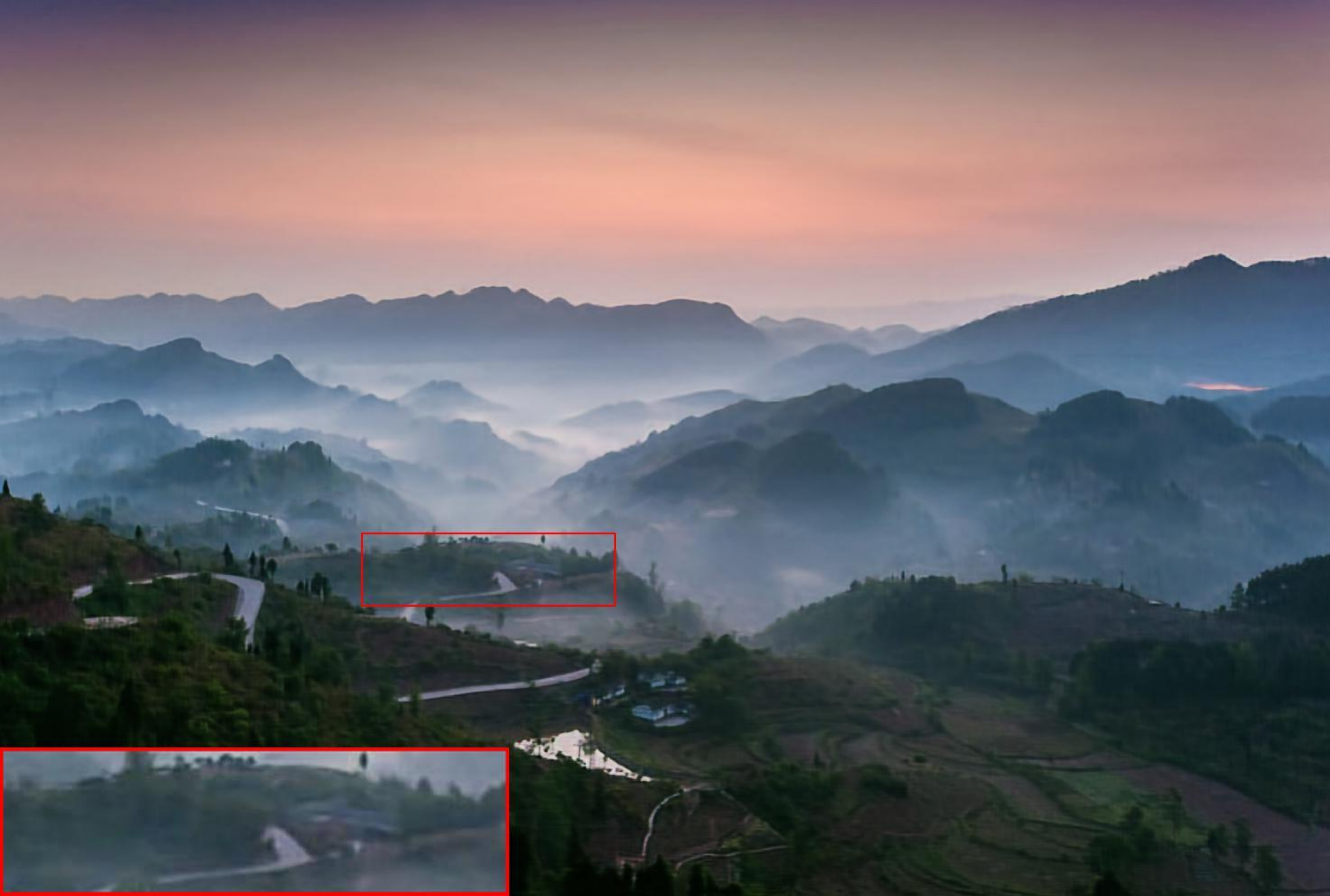}}
		\subfigure{\includegraphics[scale=\two_wid]{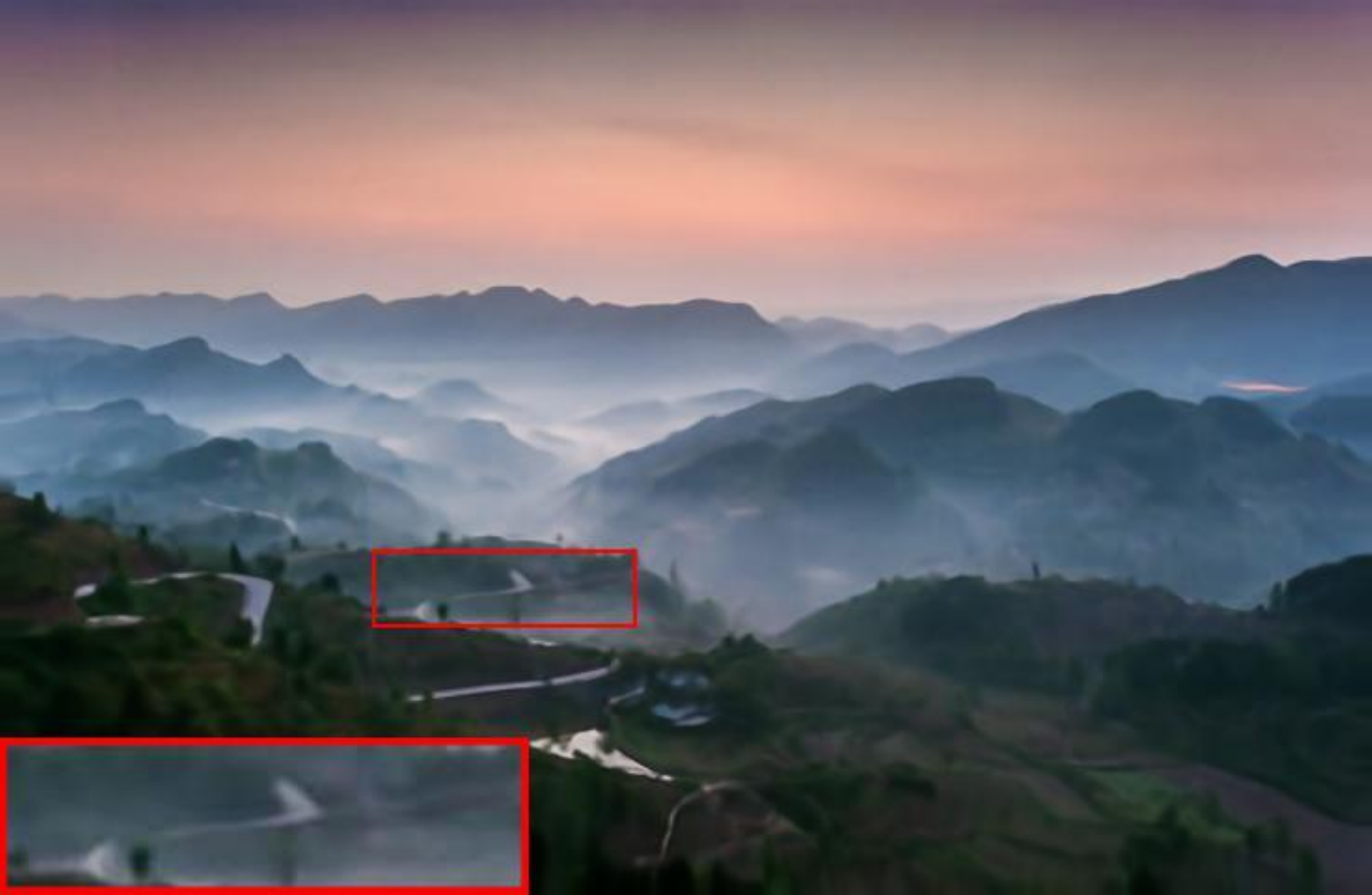}}
		\subfigure{\includegraphics[scale=\m_wid]{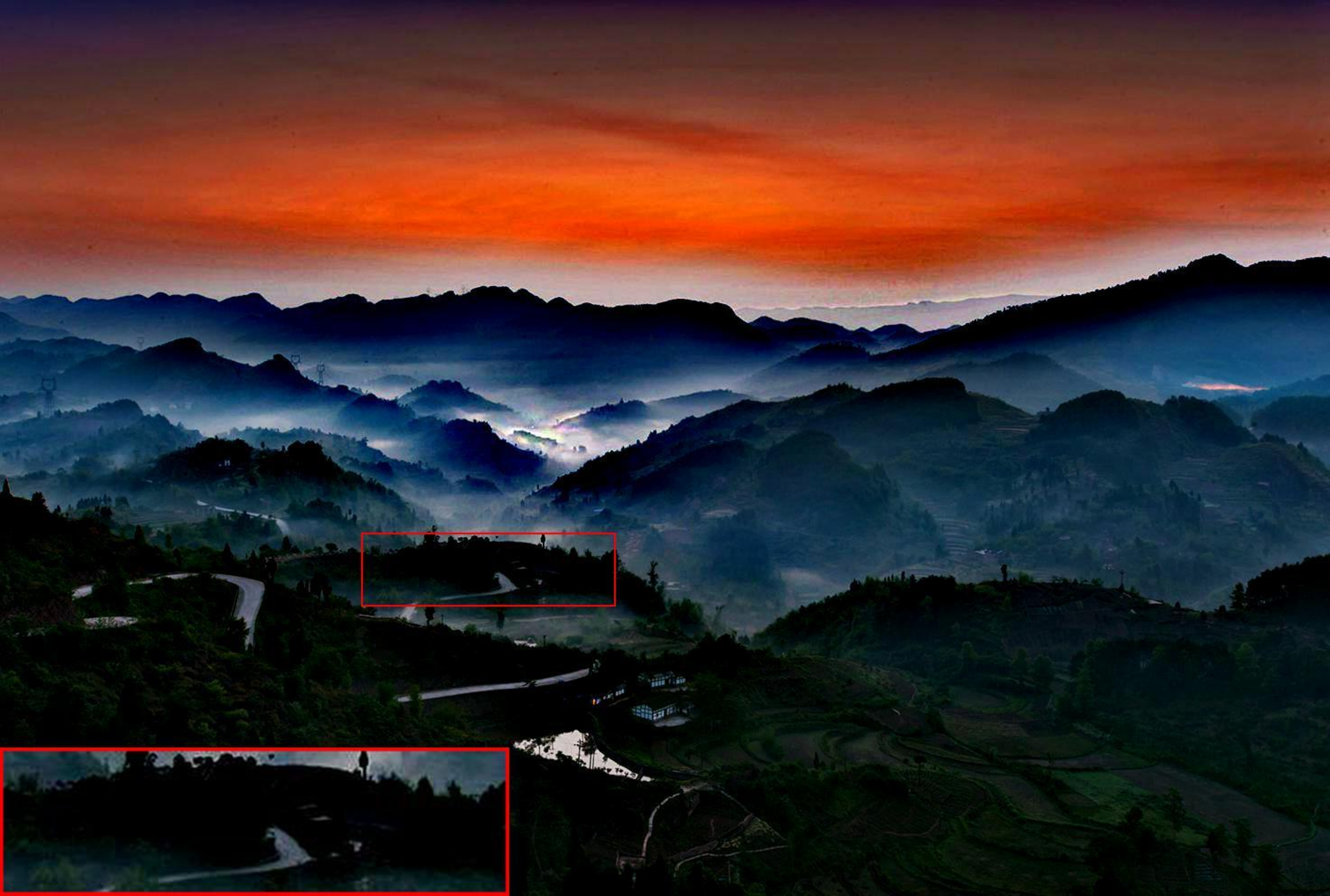}}
		\subfigure{\includegraphics[scale=\two_wid]{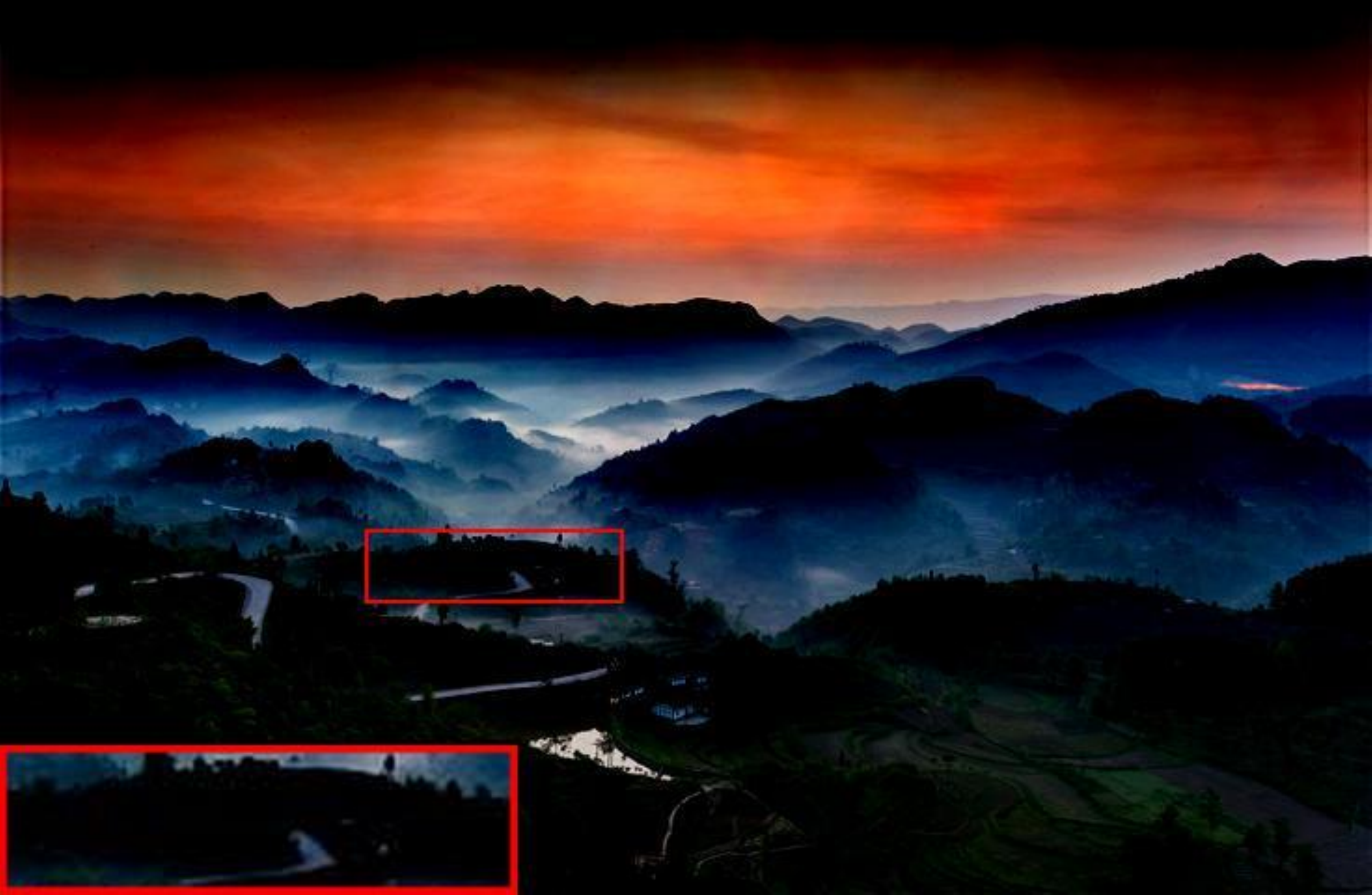}}
		\subfigure{\includegraphics[scale=\two_wid]{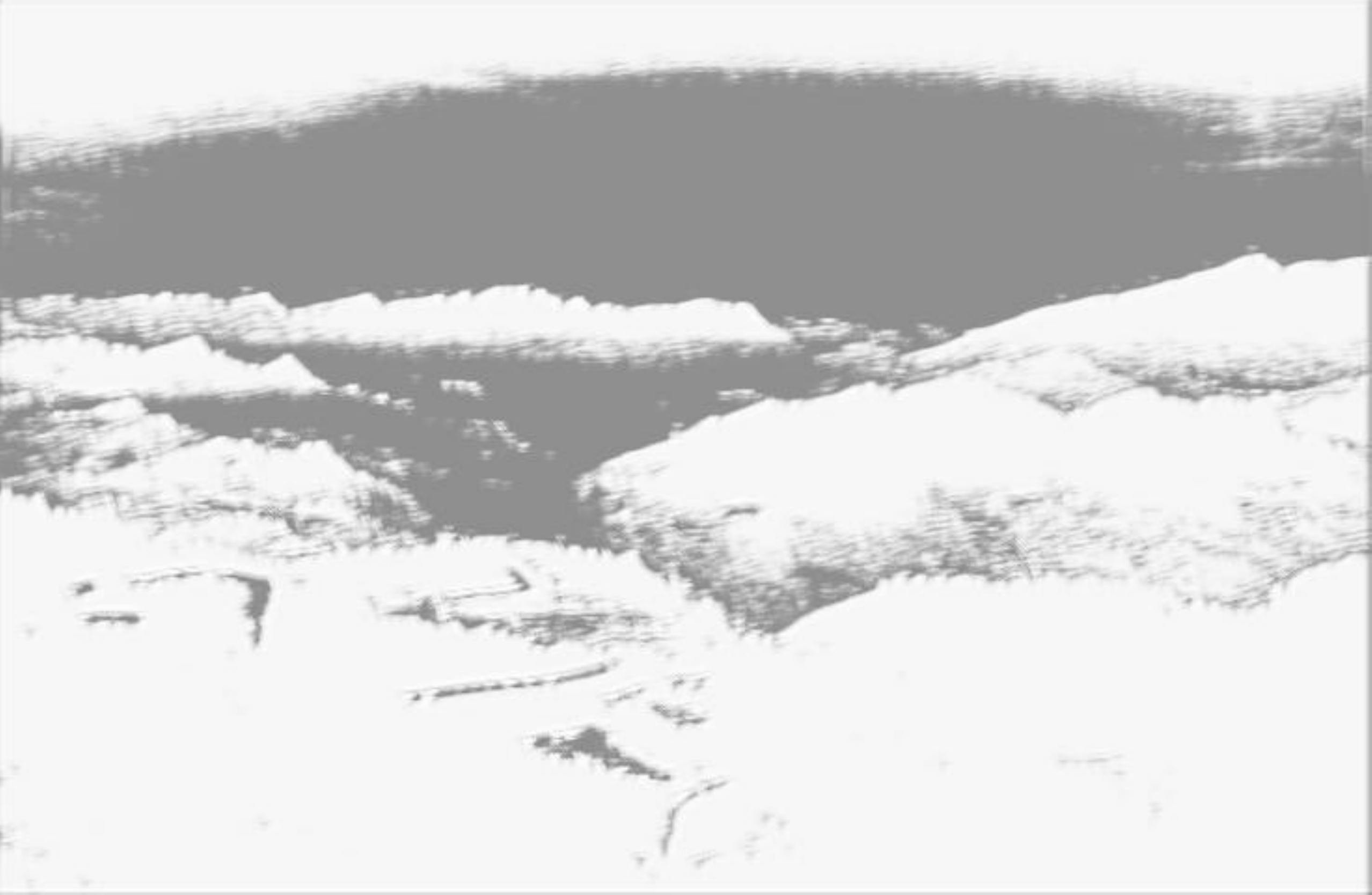}}
	\end{center}
	\vspace{-0.8cm}
	
	\begin{center}
		\def \m_wid{0.0470}
		\subfigure{\includegraphics[scale=\m_wid]{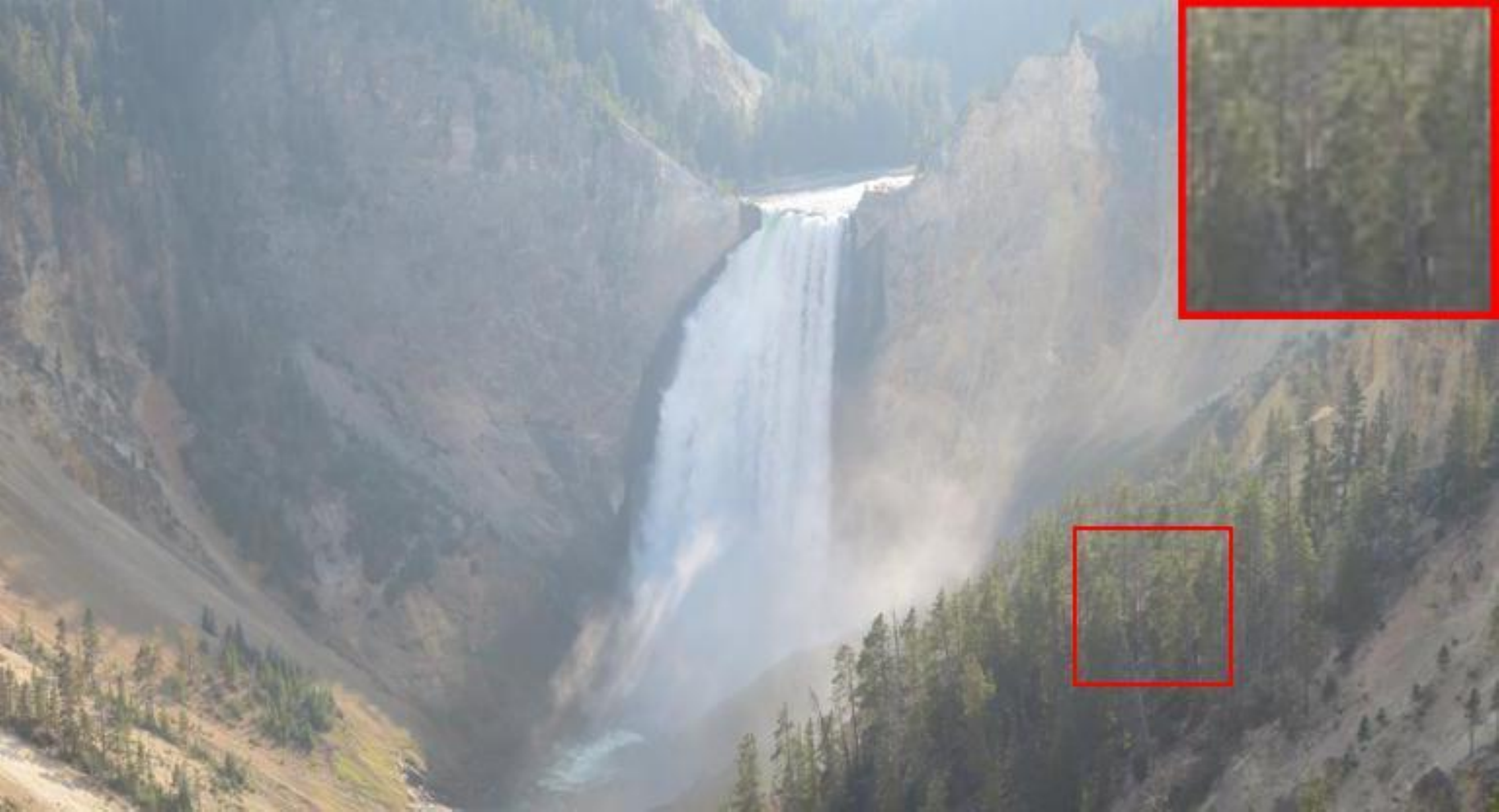}}
		\subfigure{\includegraphics[scale=\m_wid]{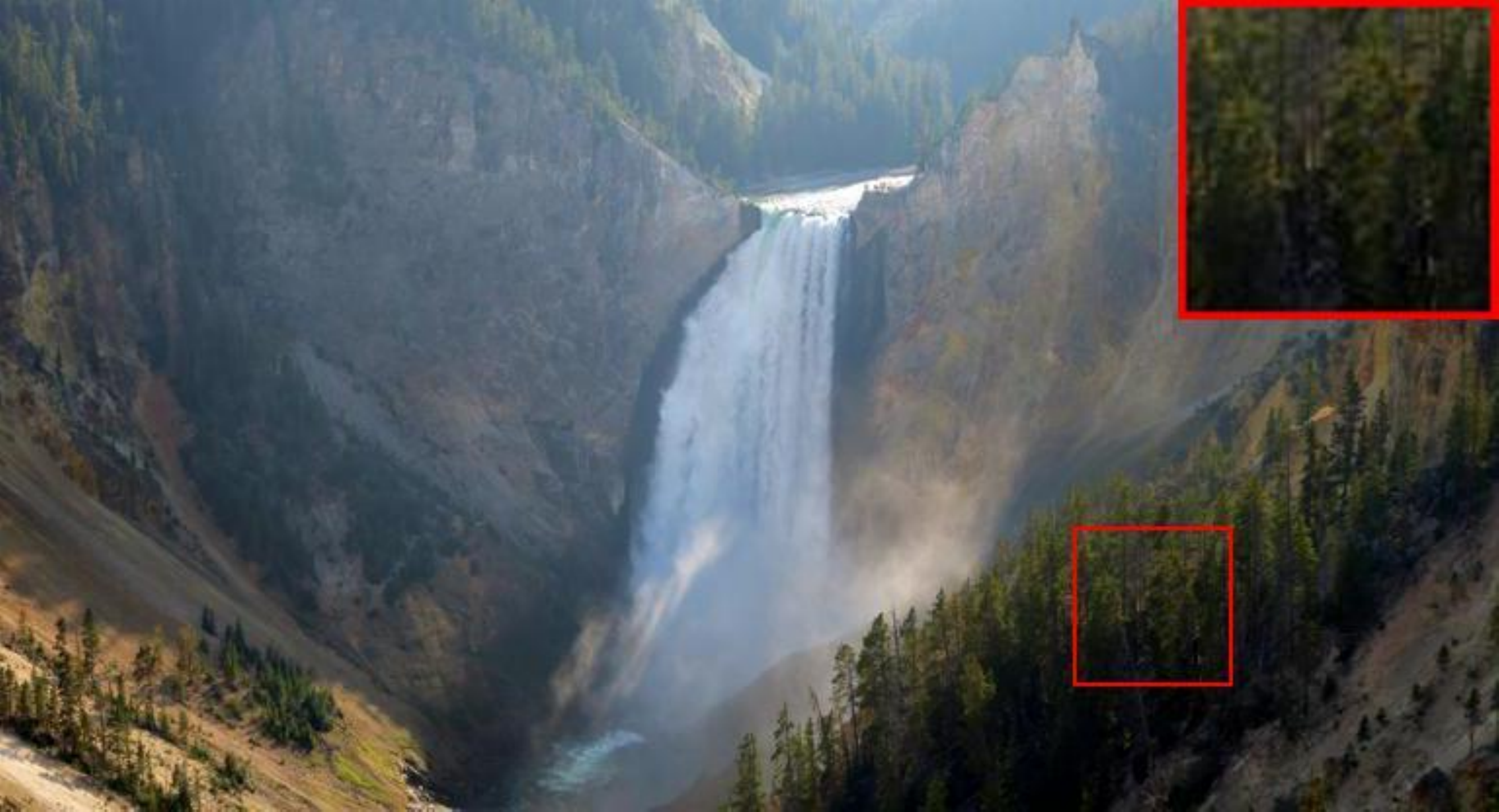}}
		\subfigure{\includegraphics[scale=\m_wid]{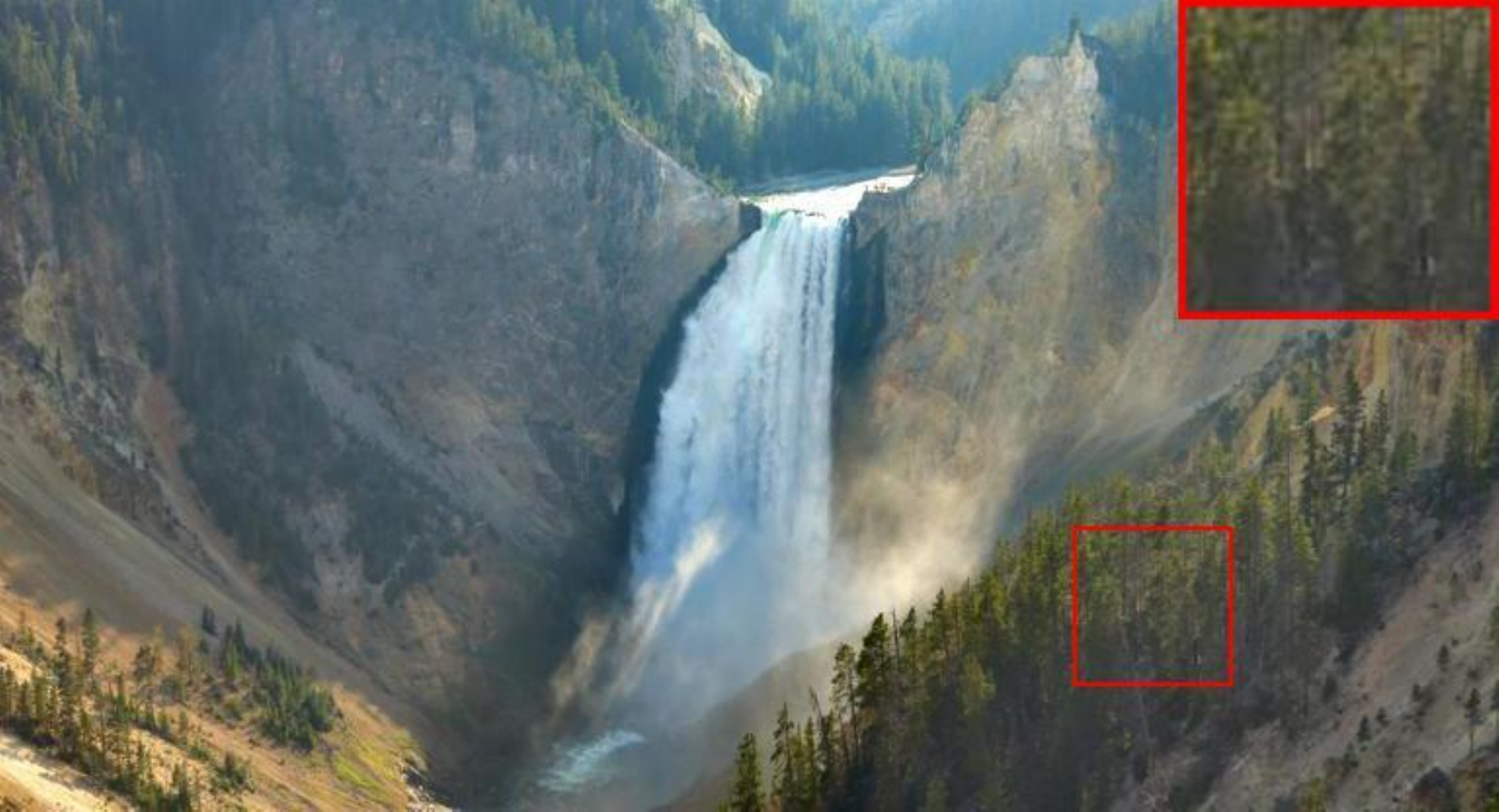}}
		\subfigure{\includegraphics[scale=\m_wid]{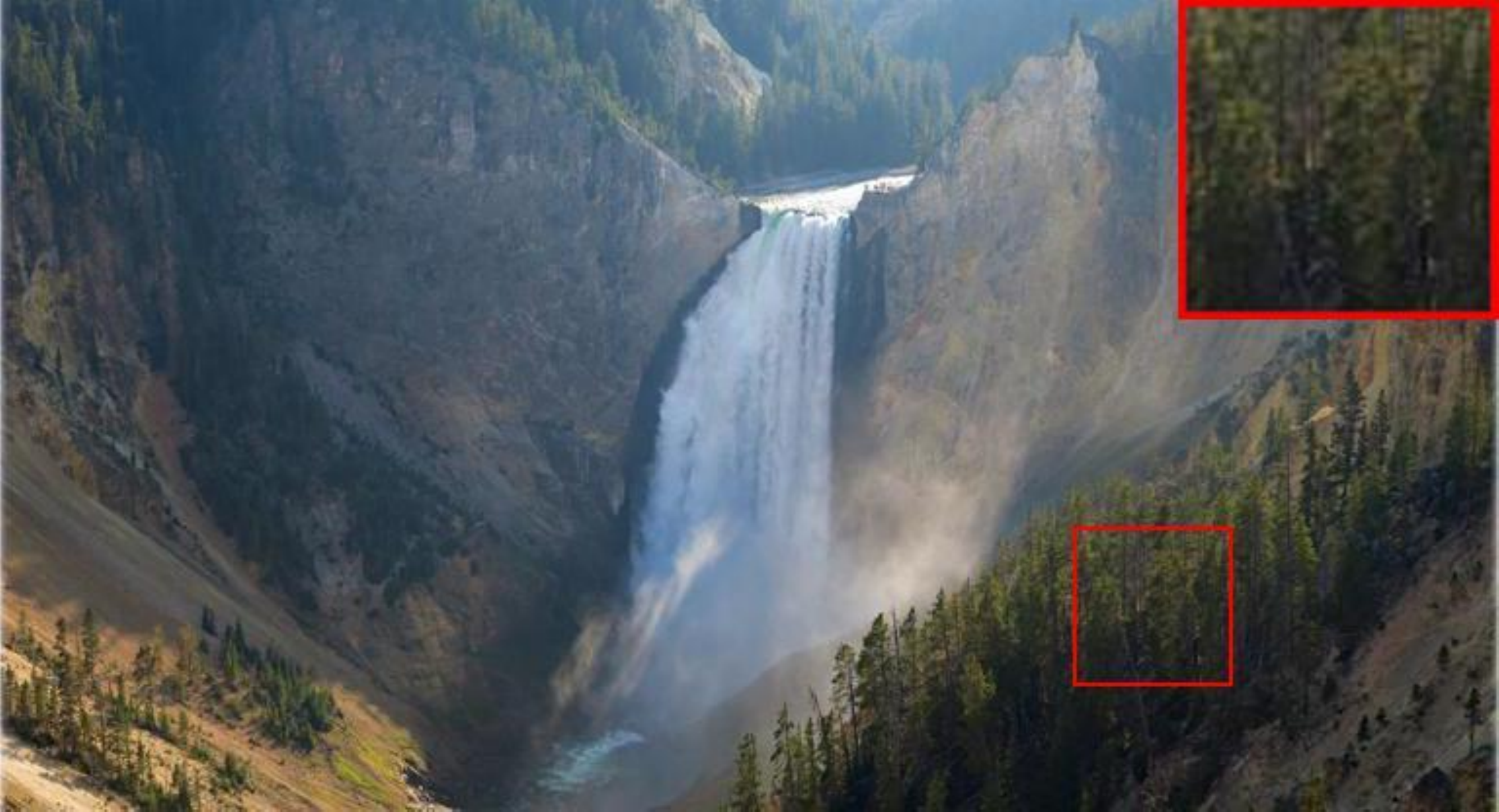}}
		\subfigure{\includegraphics[scale=\m_wid]{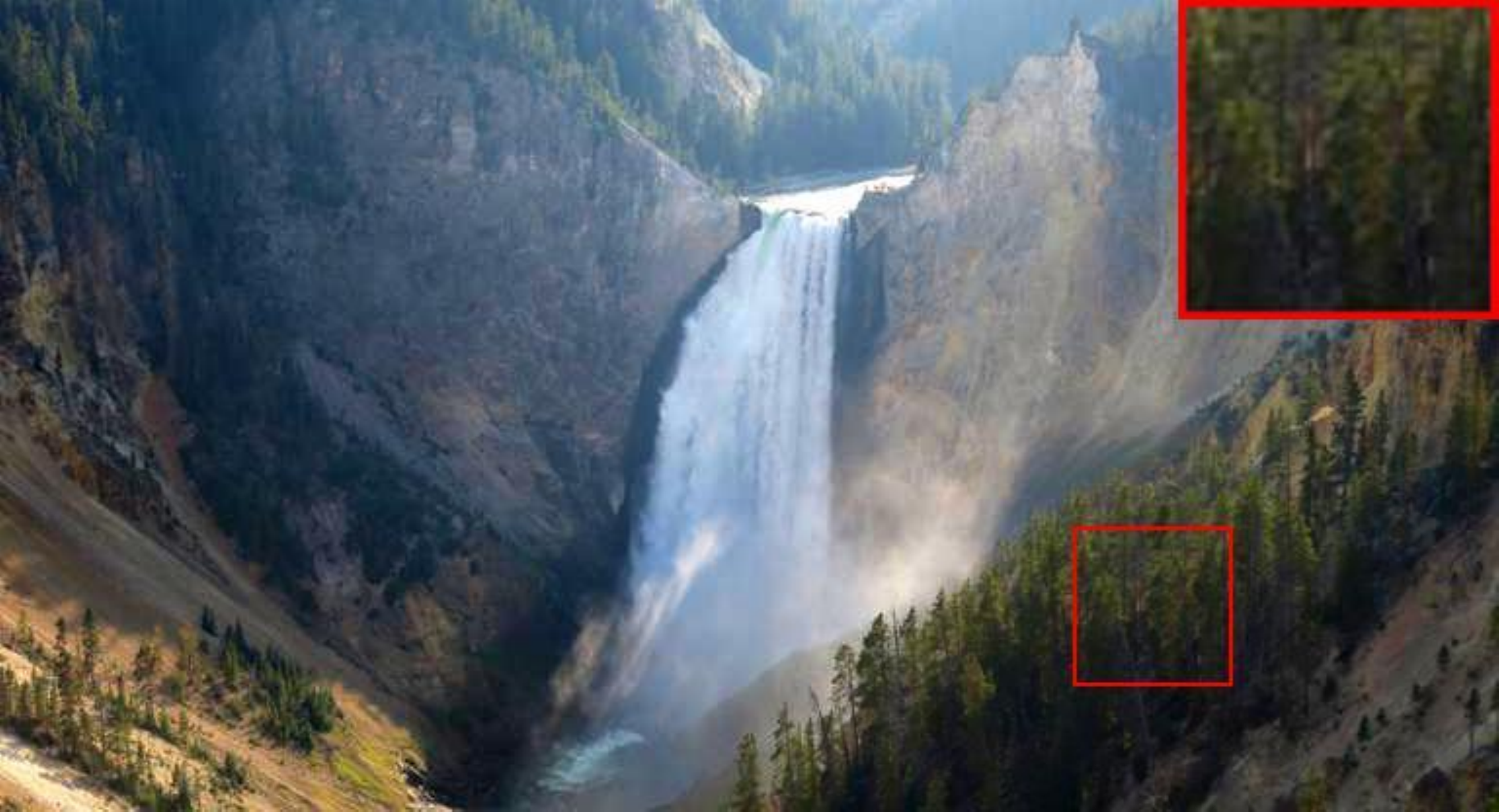}}
		\subfigure{\includegraphics[scale=\m_wid]{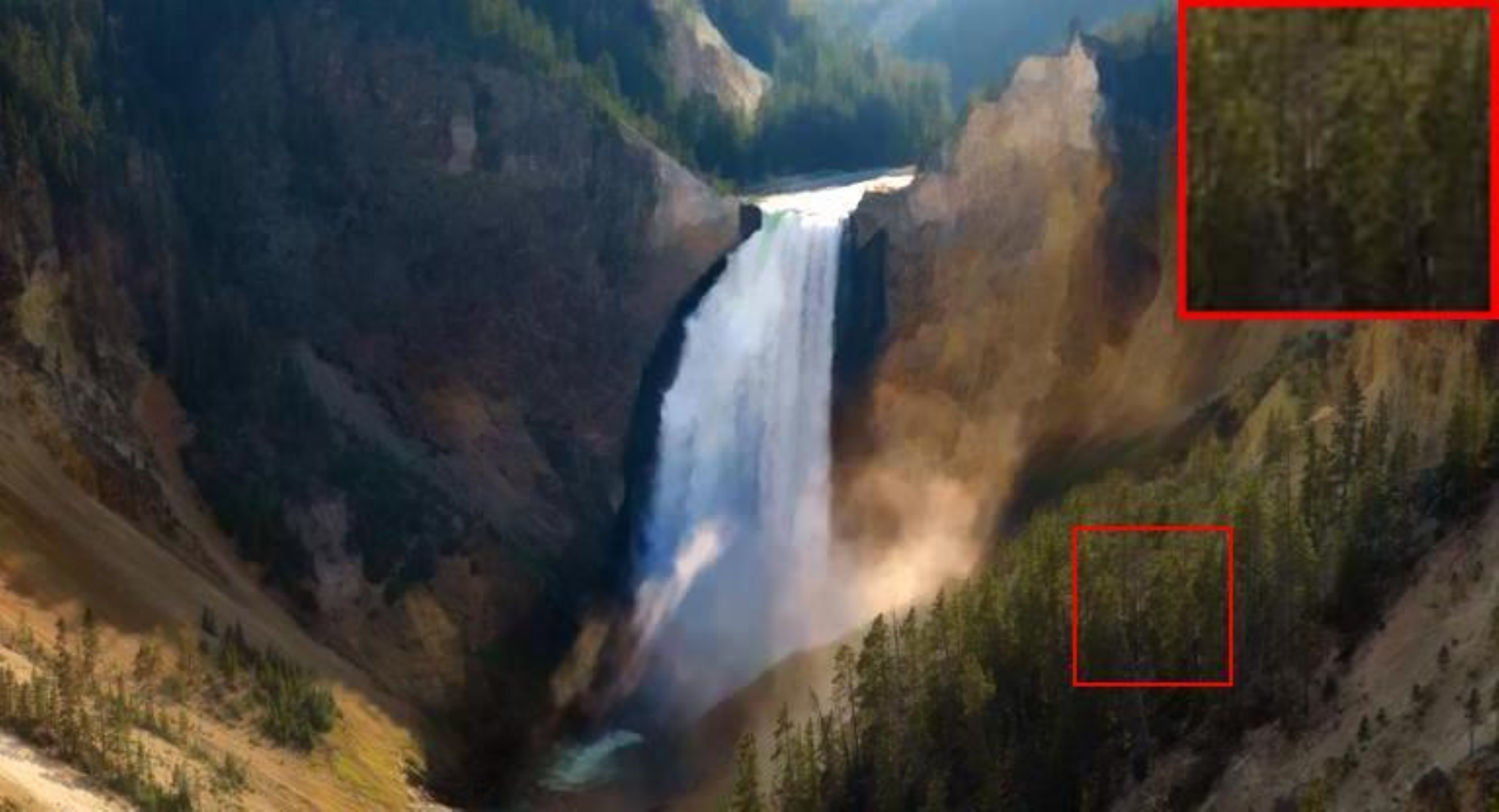}}
		\subfigure{\includegraphics[scale=\m_wid]{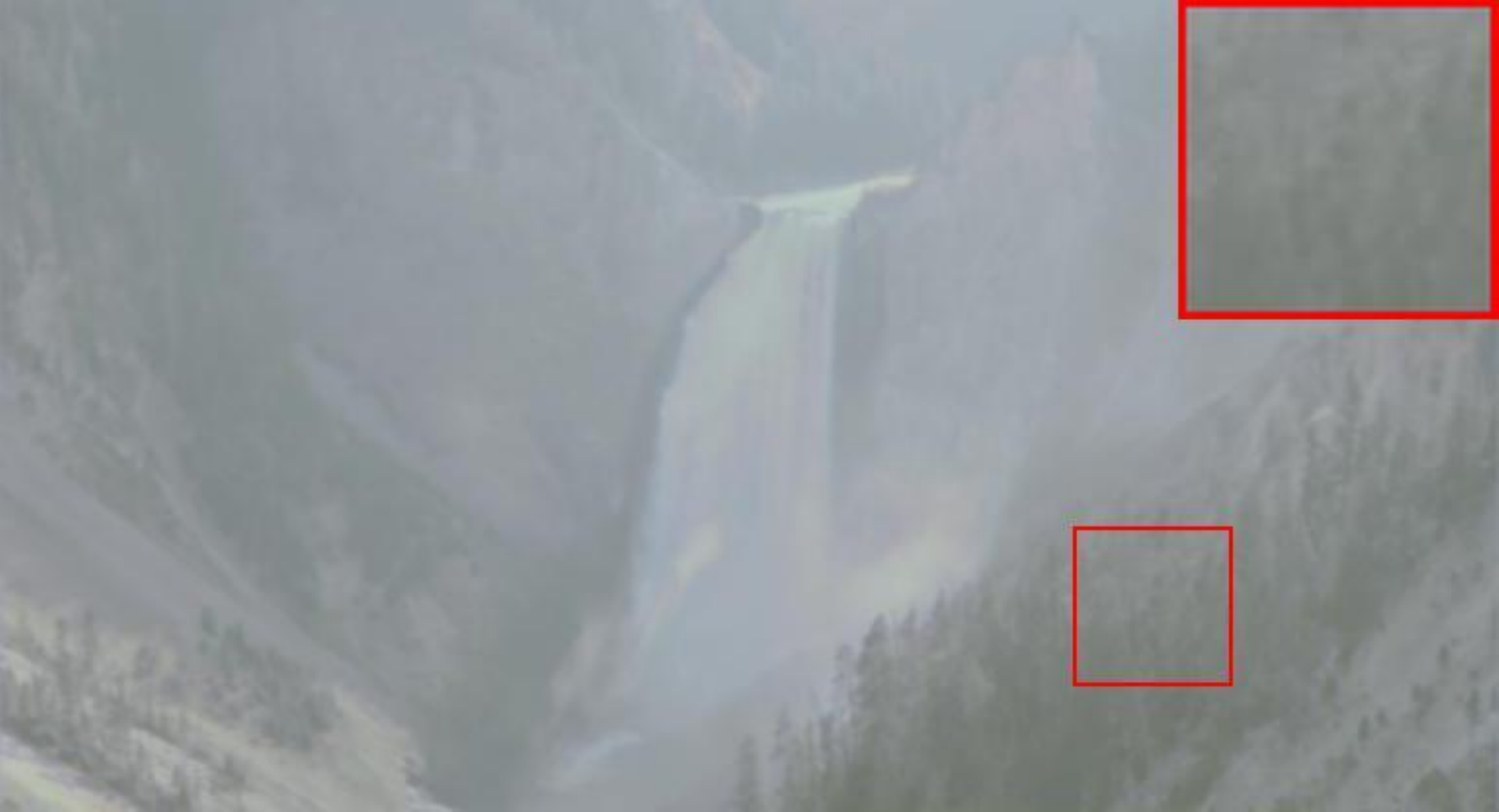}}
		\subfigure{\includegraphics[scale=\m_wid]{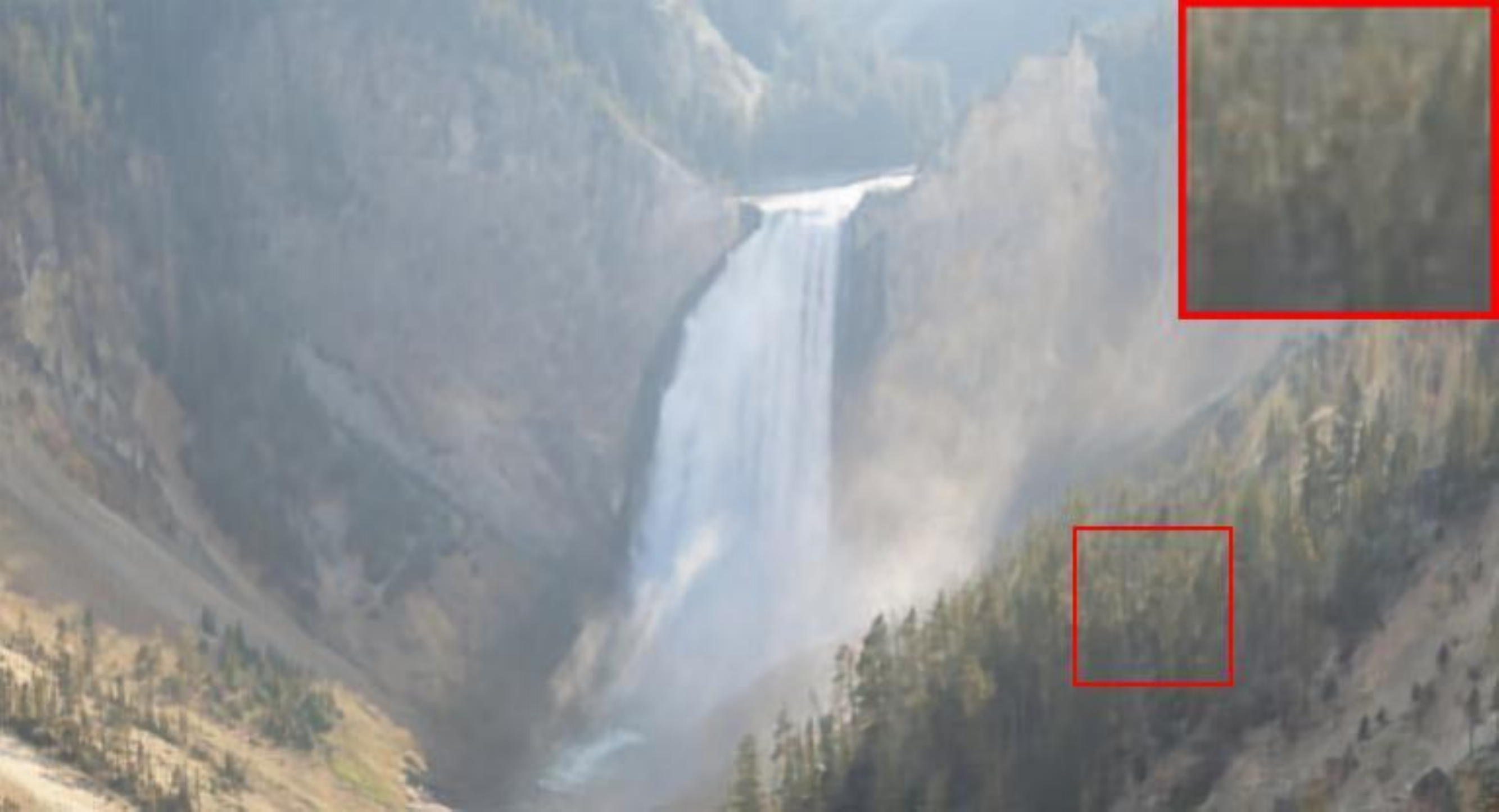}}
		\subfigure{\includegraphics[scale=\m_wid]{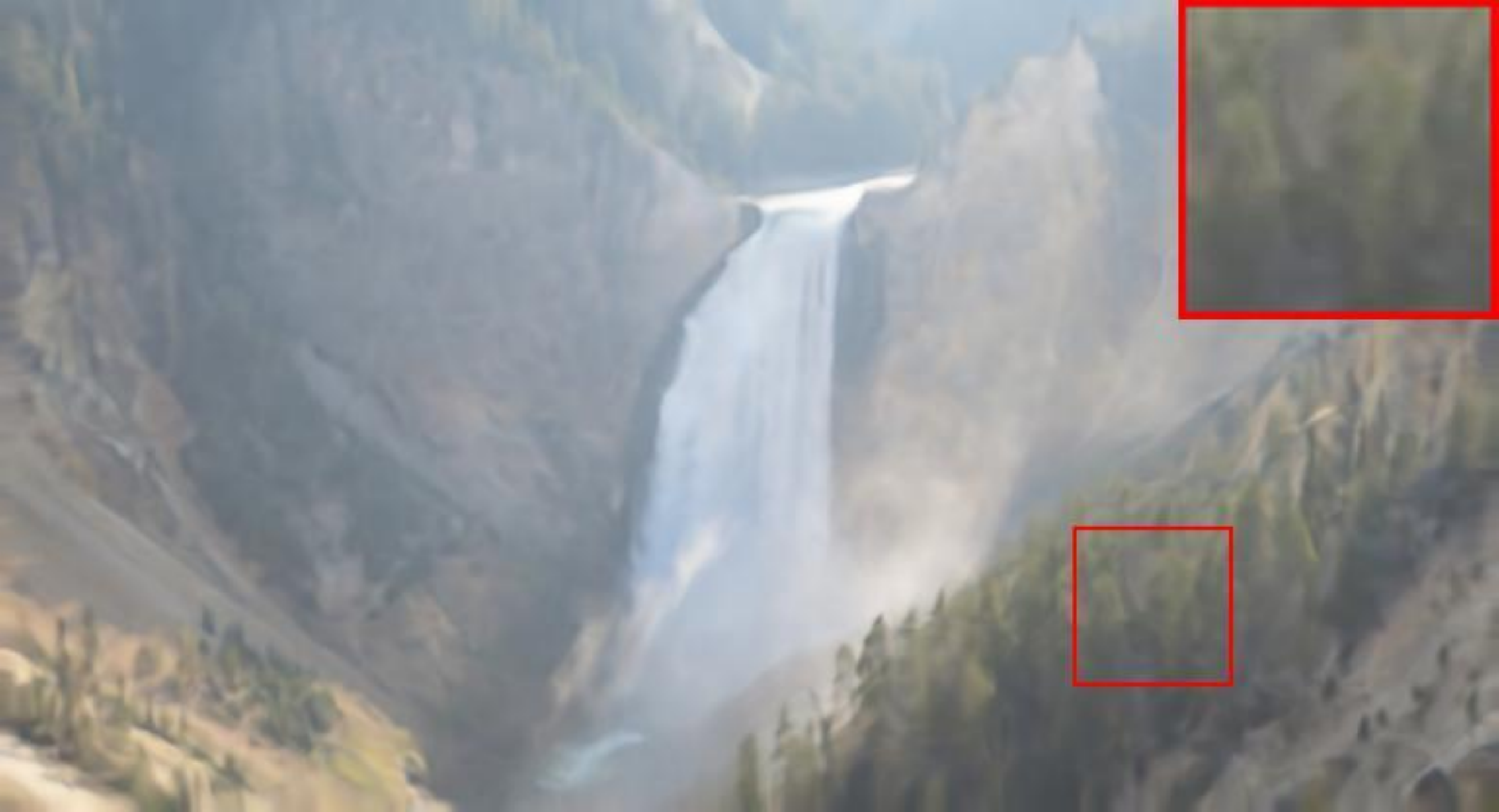}}
		\subfigure{\includegraphics[scale=\m_wid]{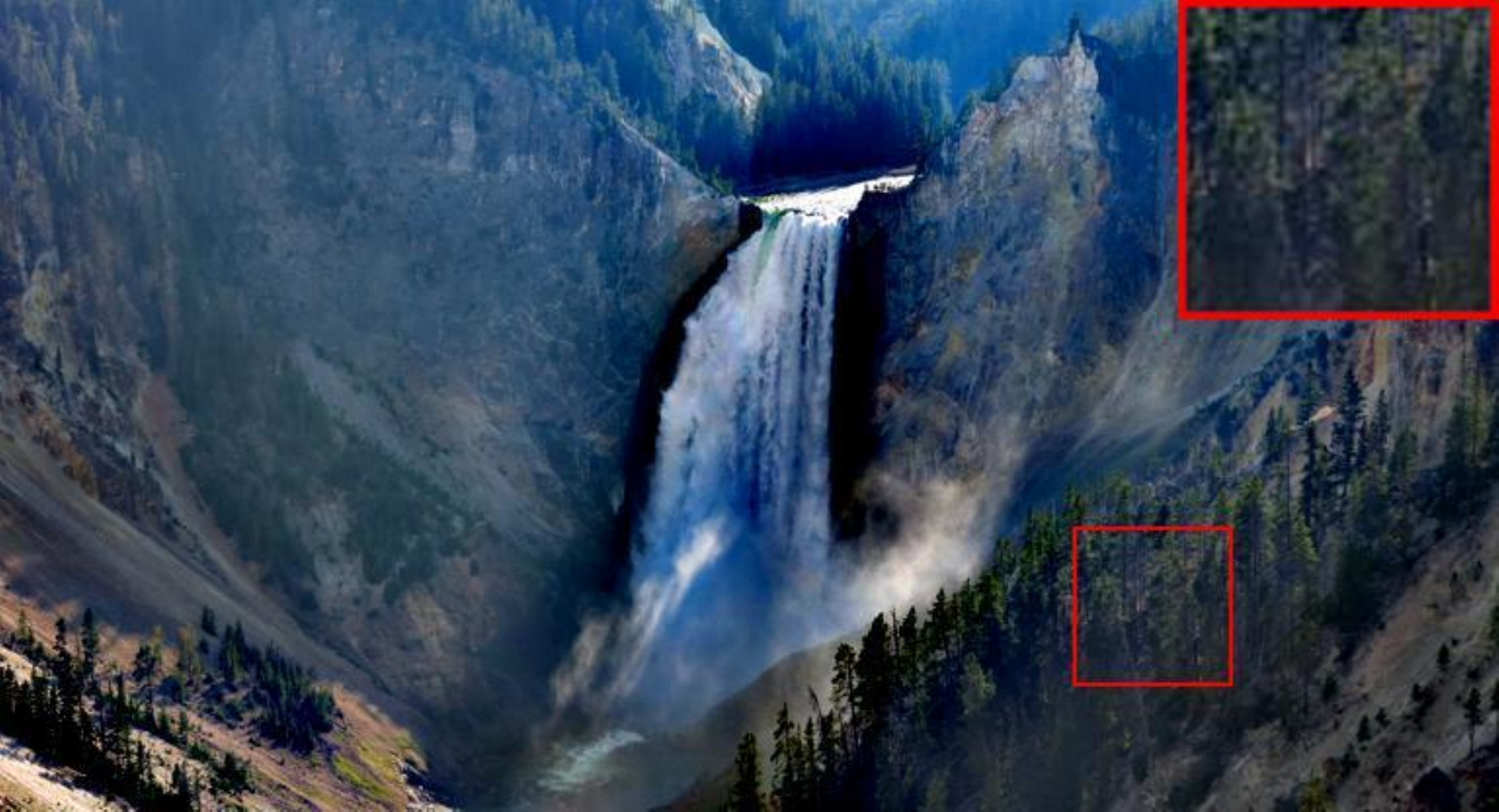}}
		\subfigure{\includegraphics[scale=\m_wid]{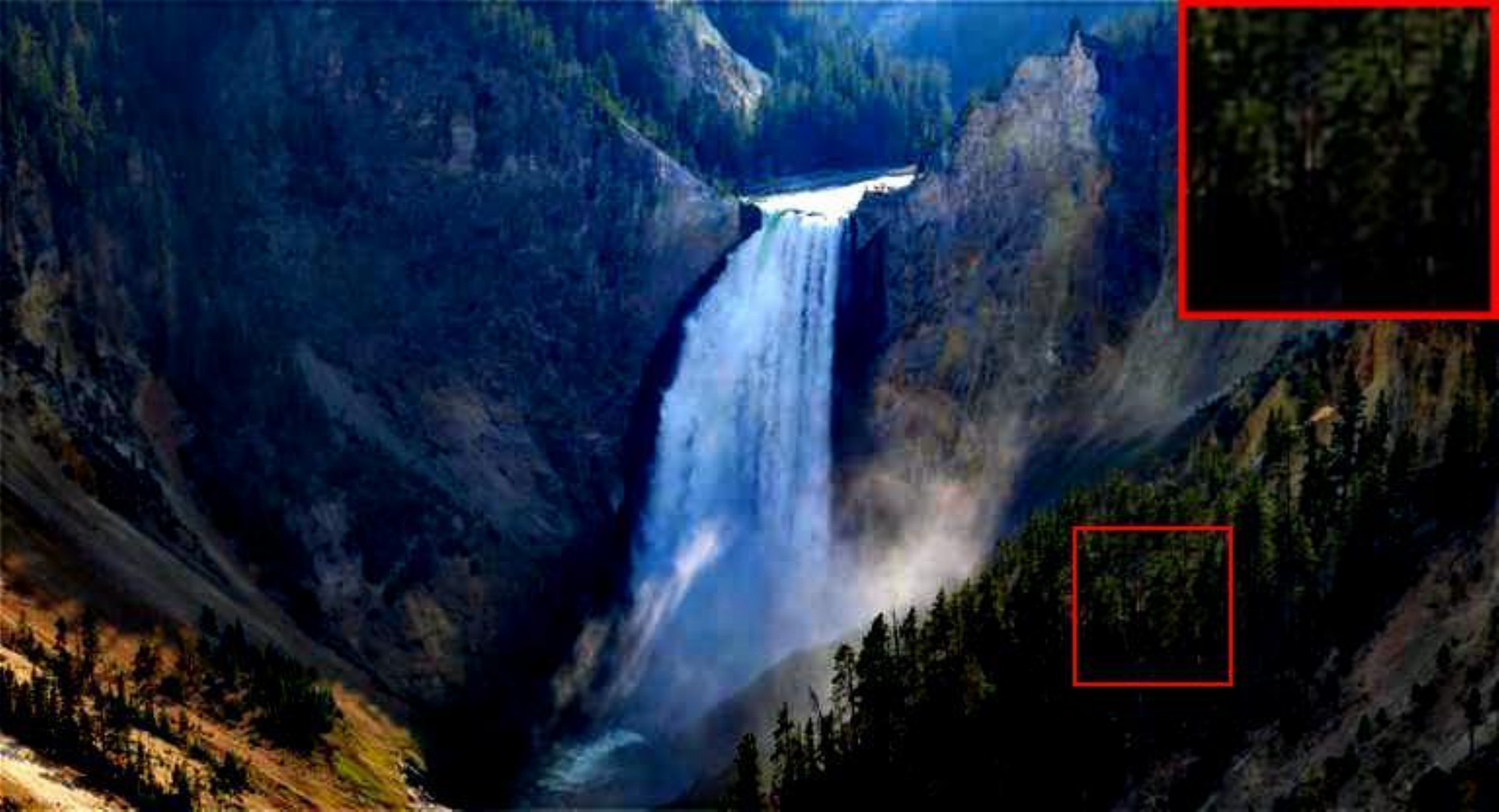}}
		\subfigure{\includegraphics[scale=\m_wid]{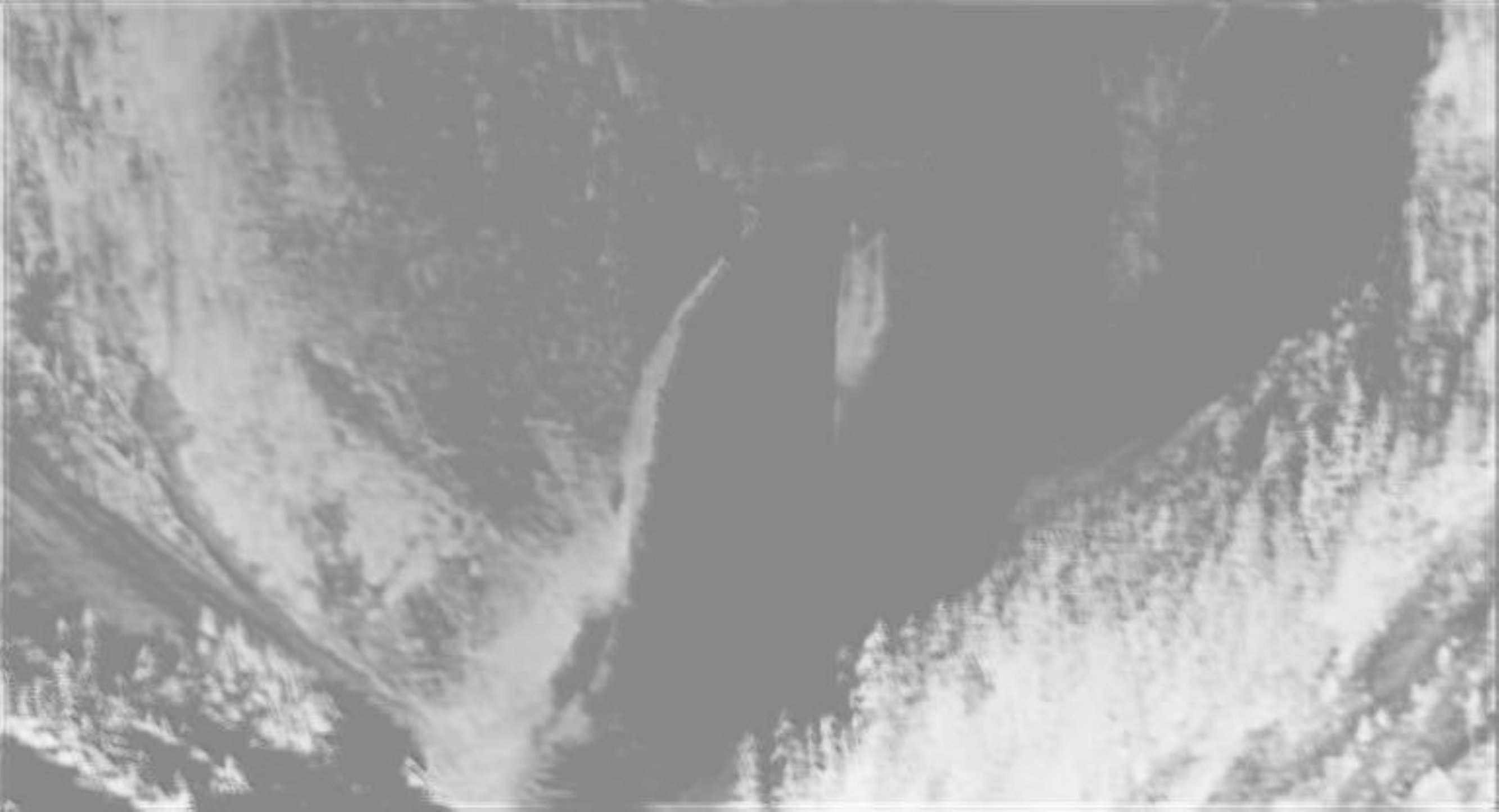}}
	\end{center}
	\vspace{-0.8cm}
	
	\begin{center}
	\def \m_wid{0.047}
		\subfigure{\includegraphics[scale=\m_wid]{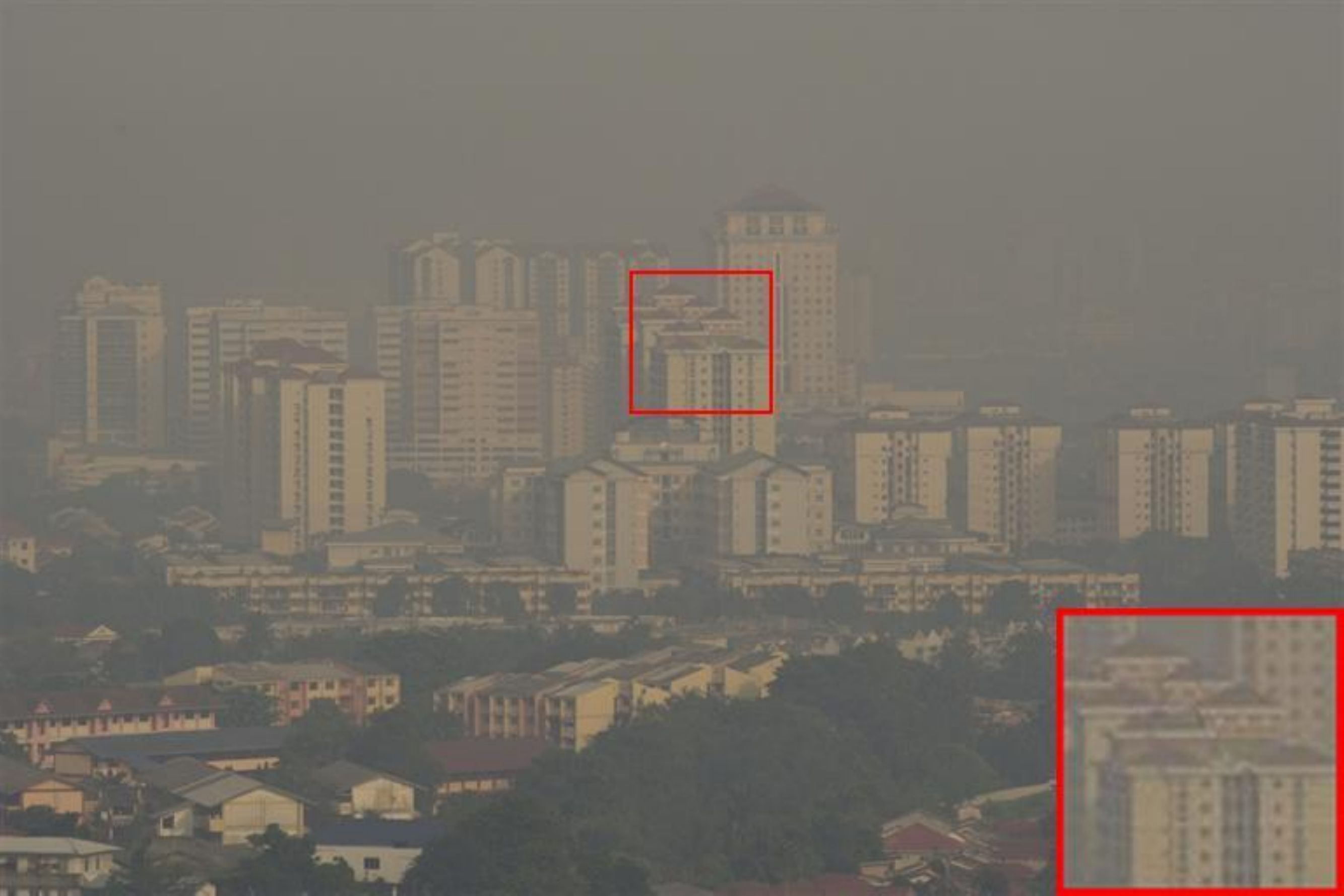}}
		\subfigure{\includegraphics[scale=\m_wid]{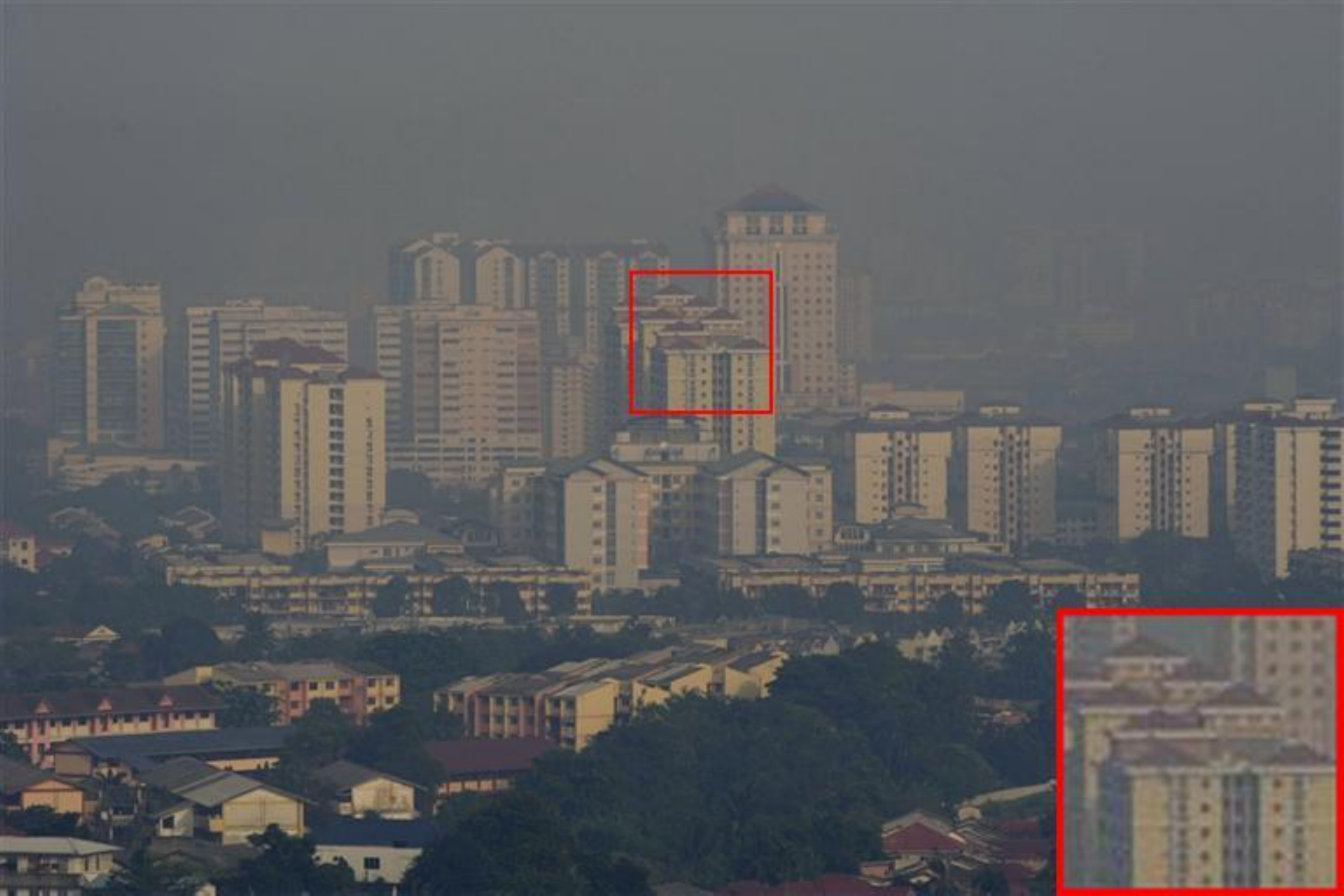}}
		\subfigure{\includegraphics[scale=\m_wid]{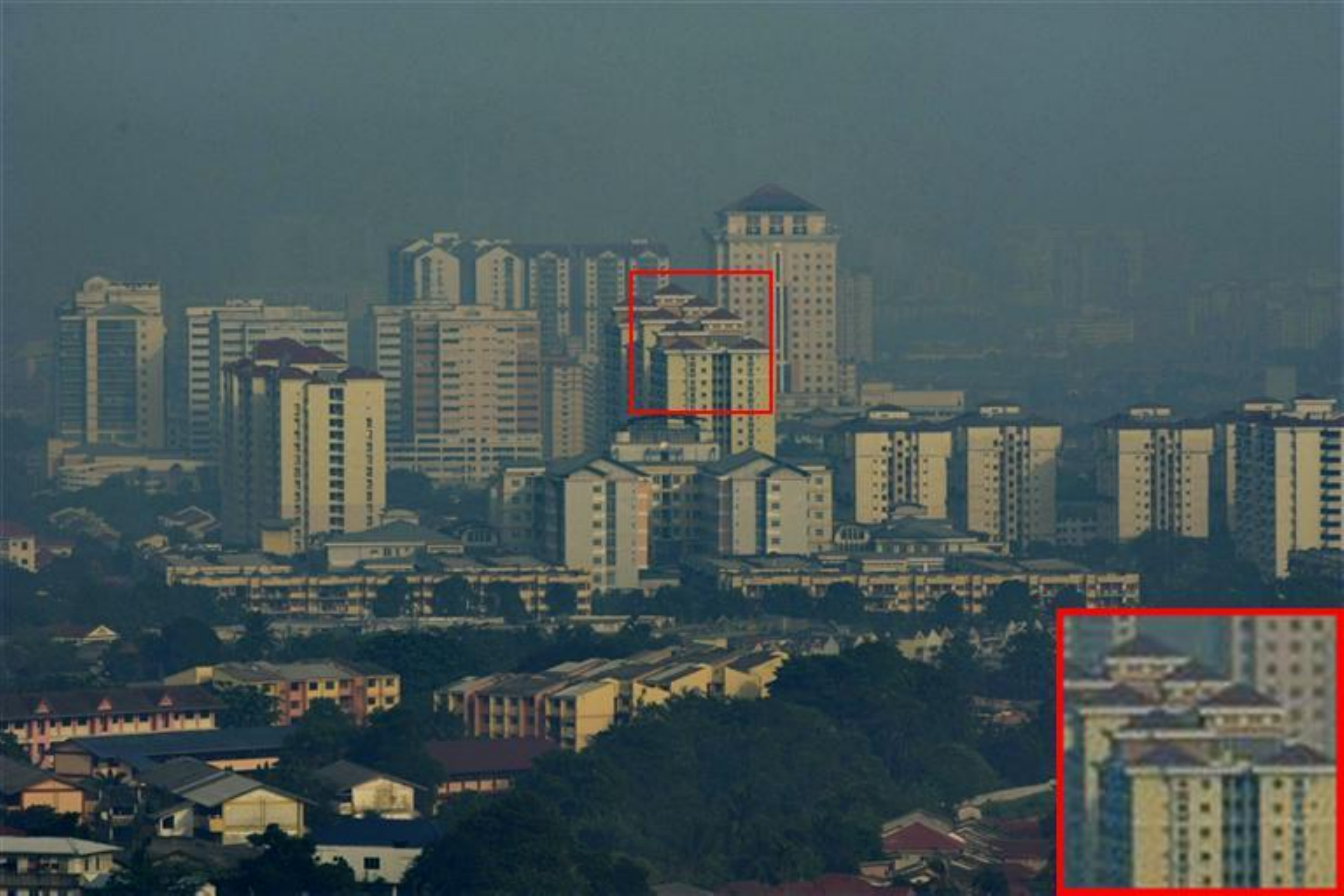}}
		\subfigure{\includegraphics[scale=\m_wid]{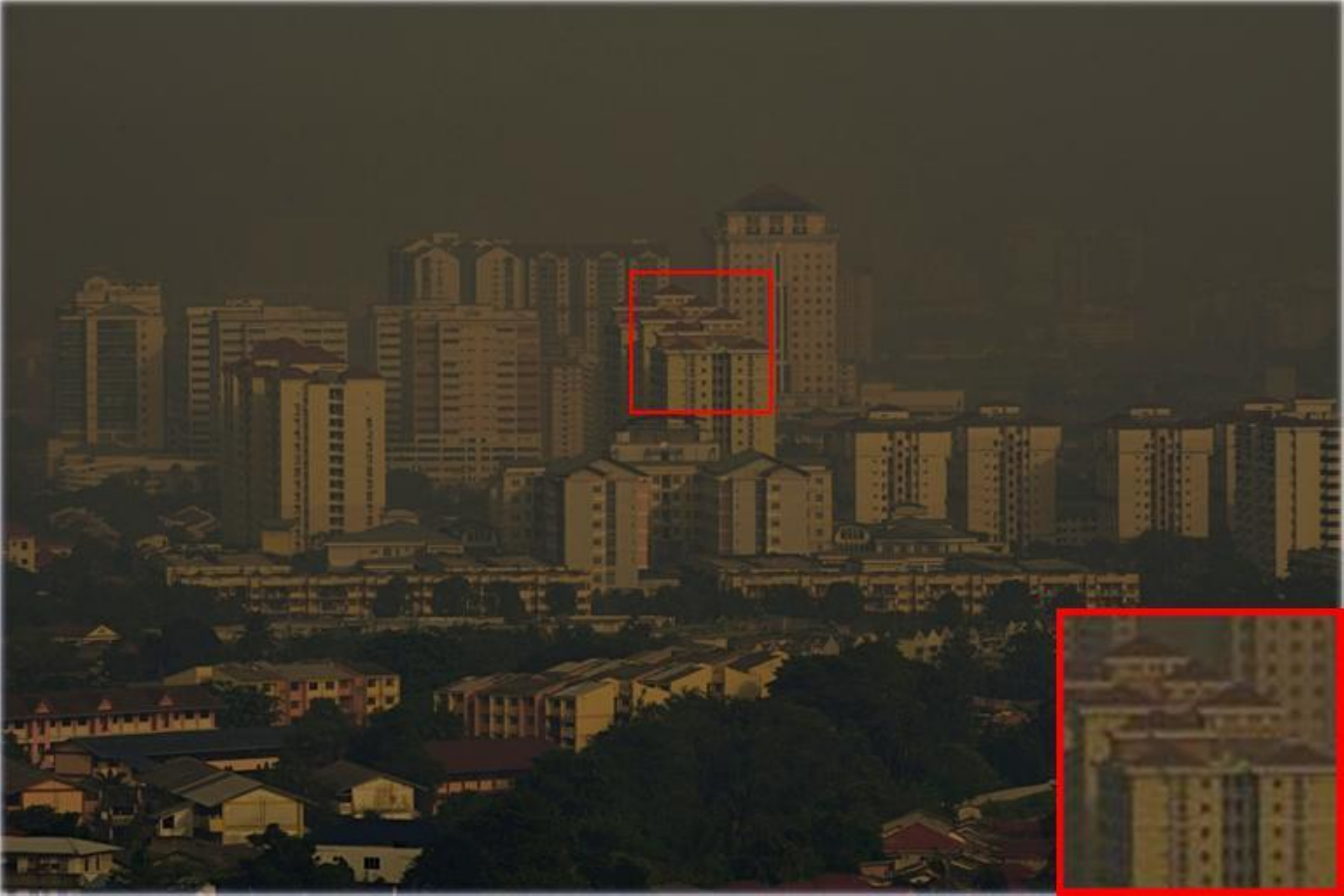}}
		\subfigure{\includegraphics[scale=\m_wid]{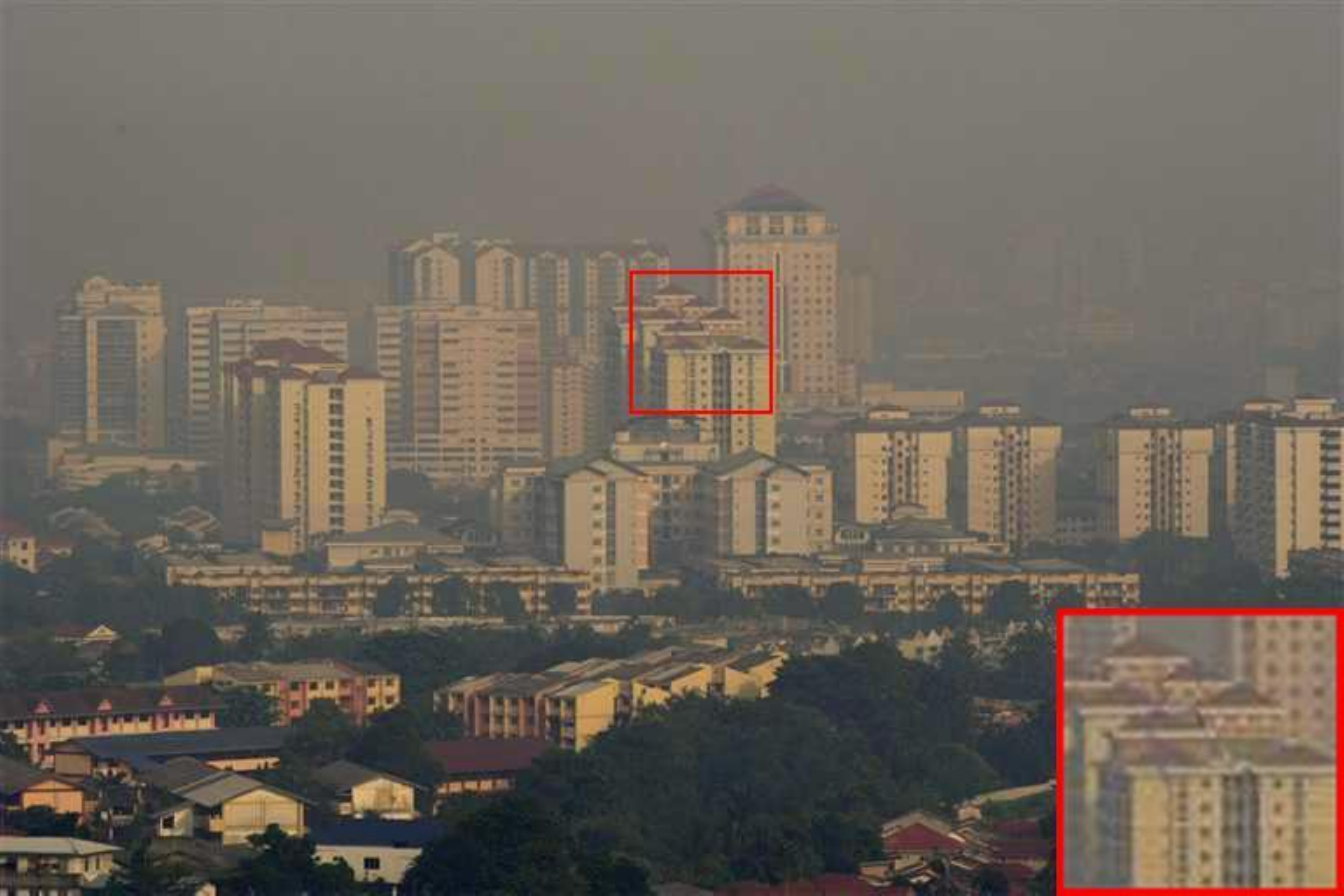}}
		\subfigure{\includegraphics[scale=\m_wid]{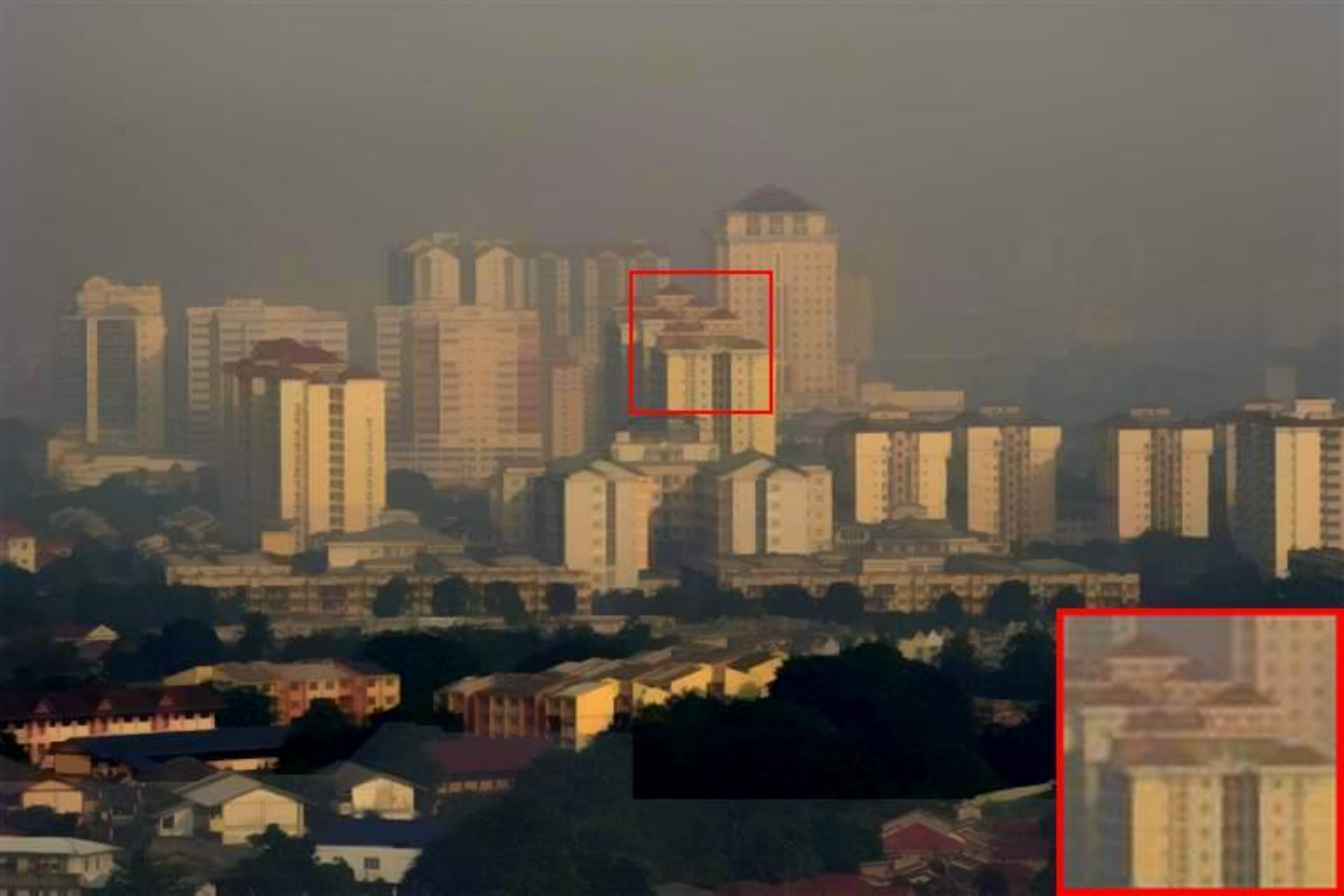}}
		\subfigure{\includegraphics[scale=\m_wid]{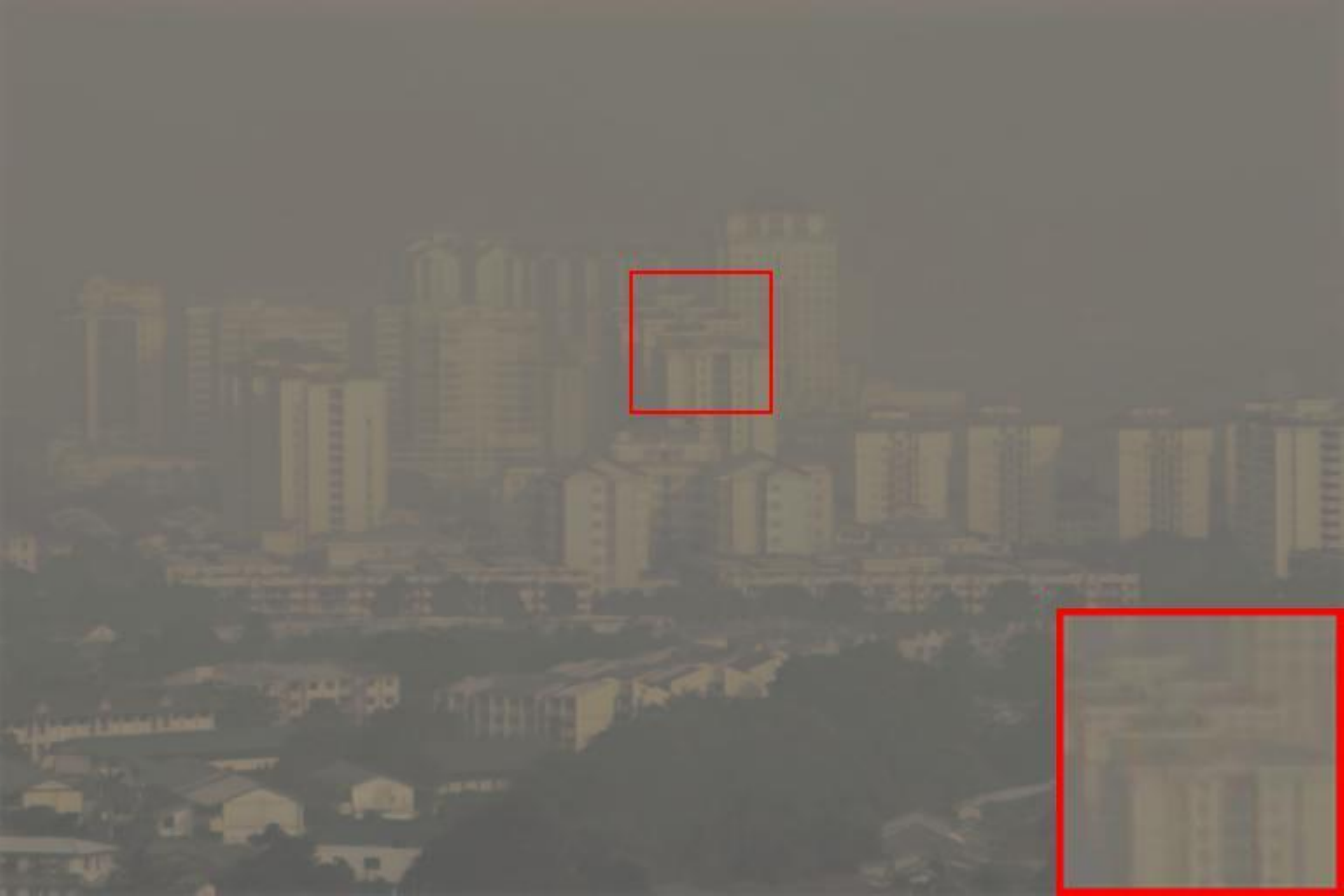}}
		\subfigure{\includegraphics[scale=\m_wid]{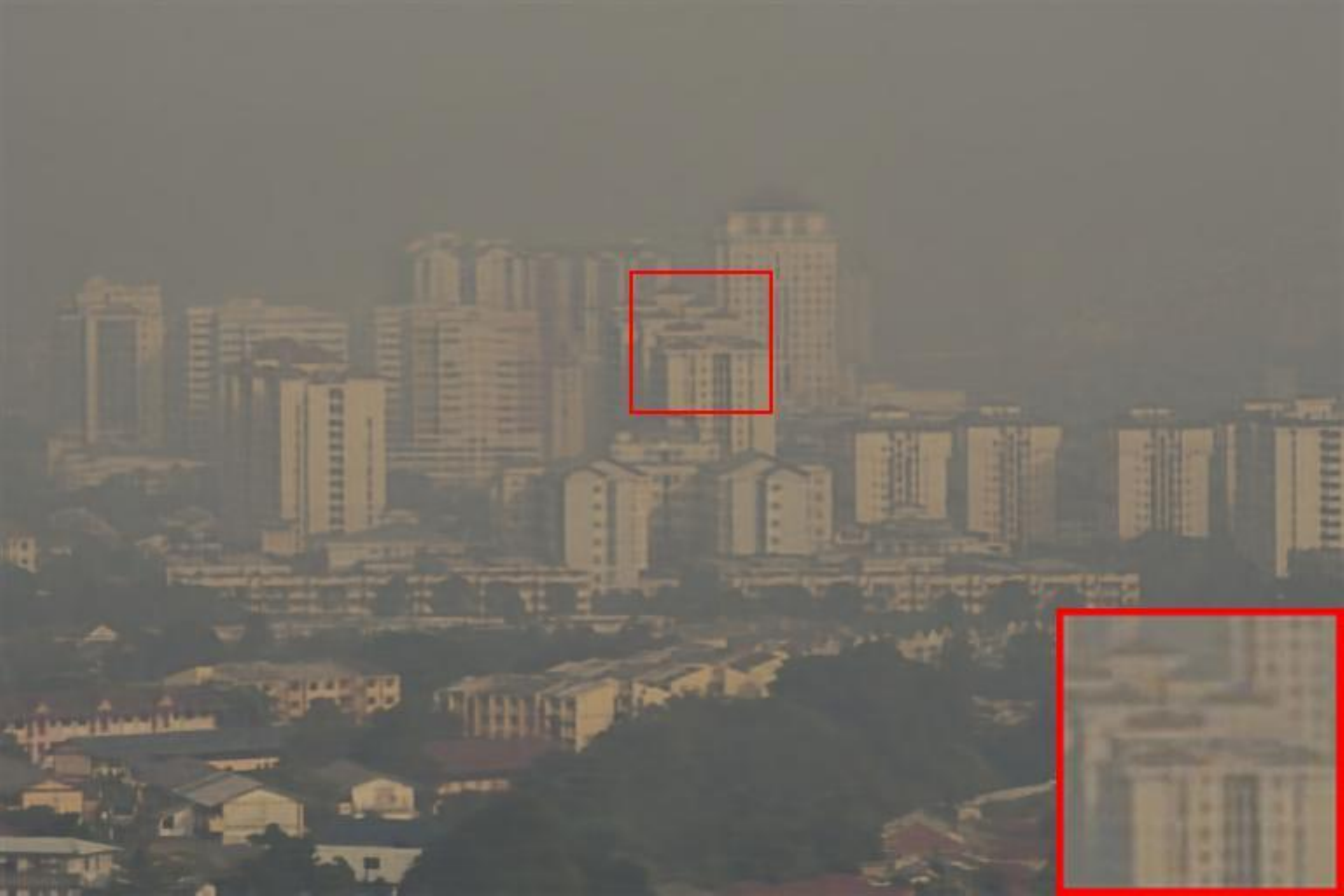}}
		\subfigure{\includegraphics[scale=\m_wid]{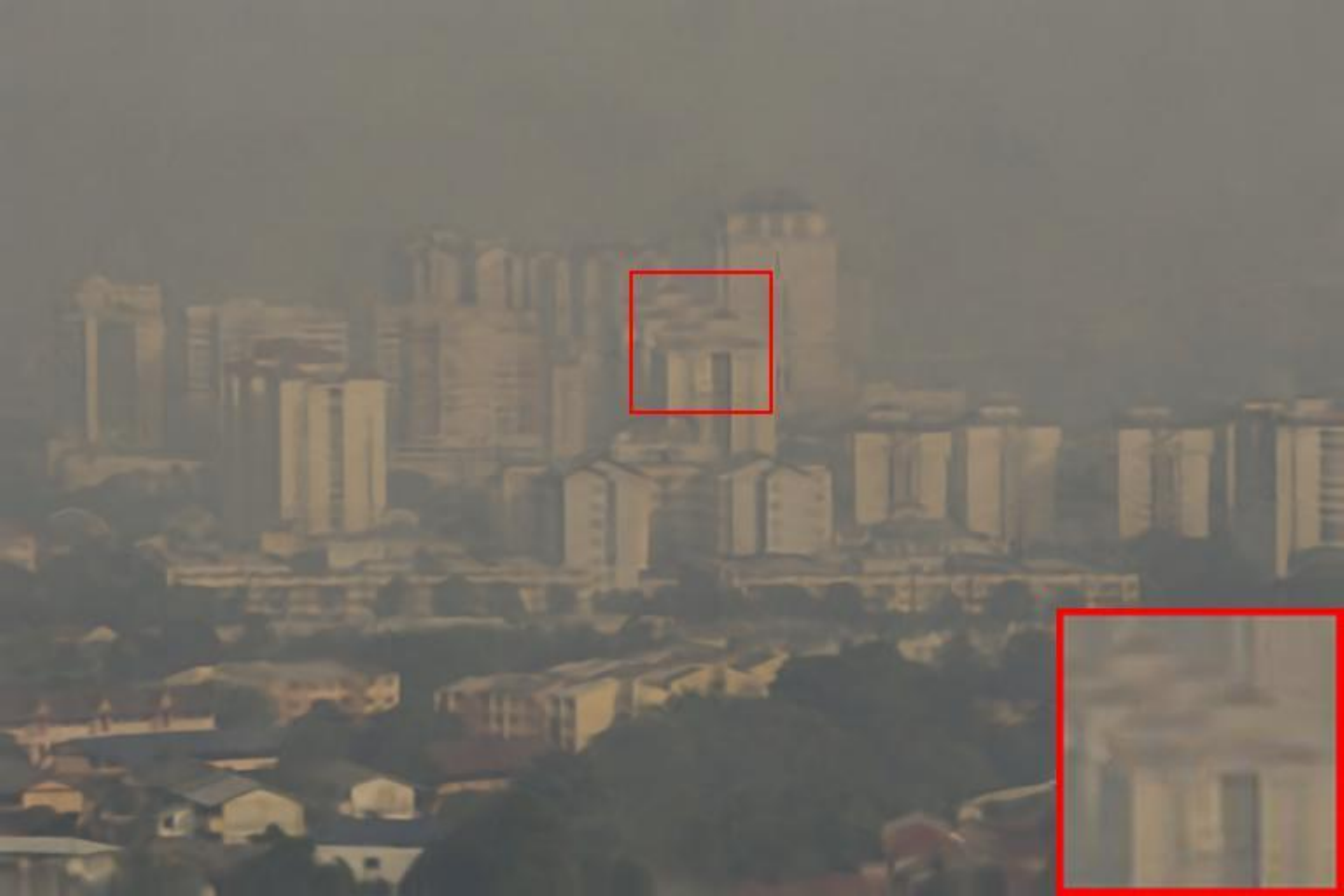}}
		\subfigure{\includegraphics[scale=\m_wid]{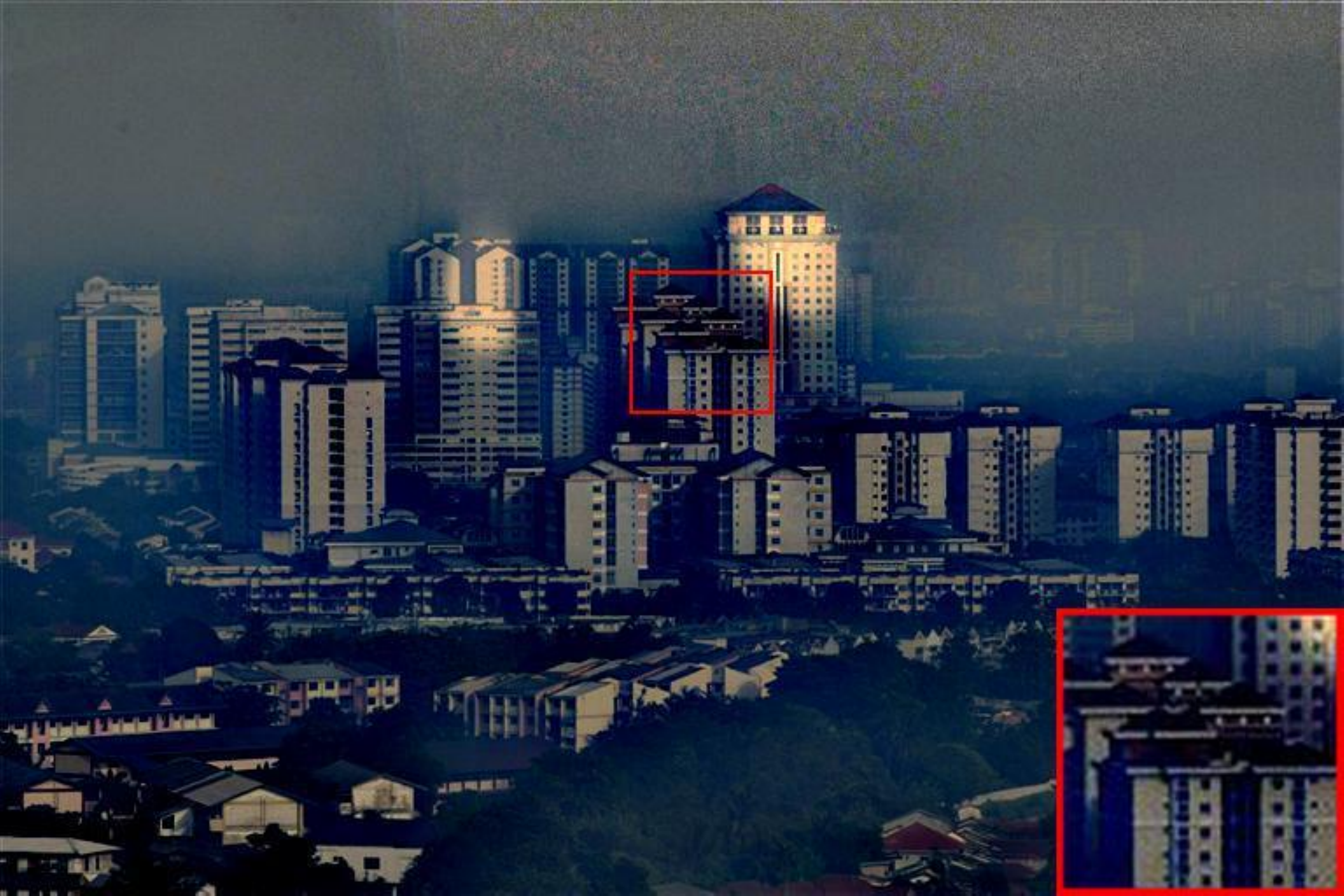}}
		\subfigure{\includegraphics[scale=\m_wid]{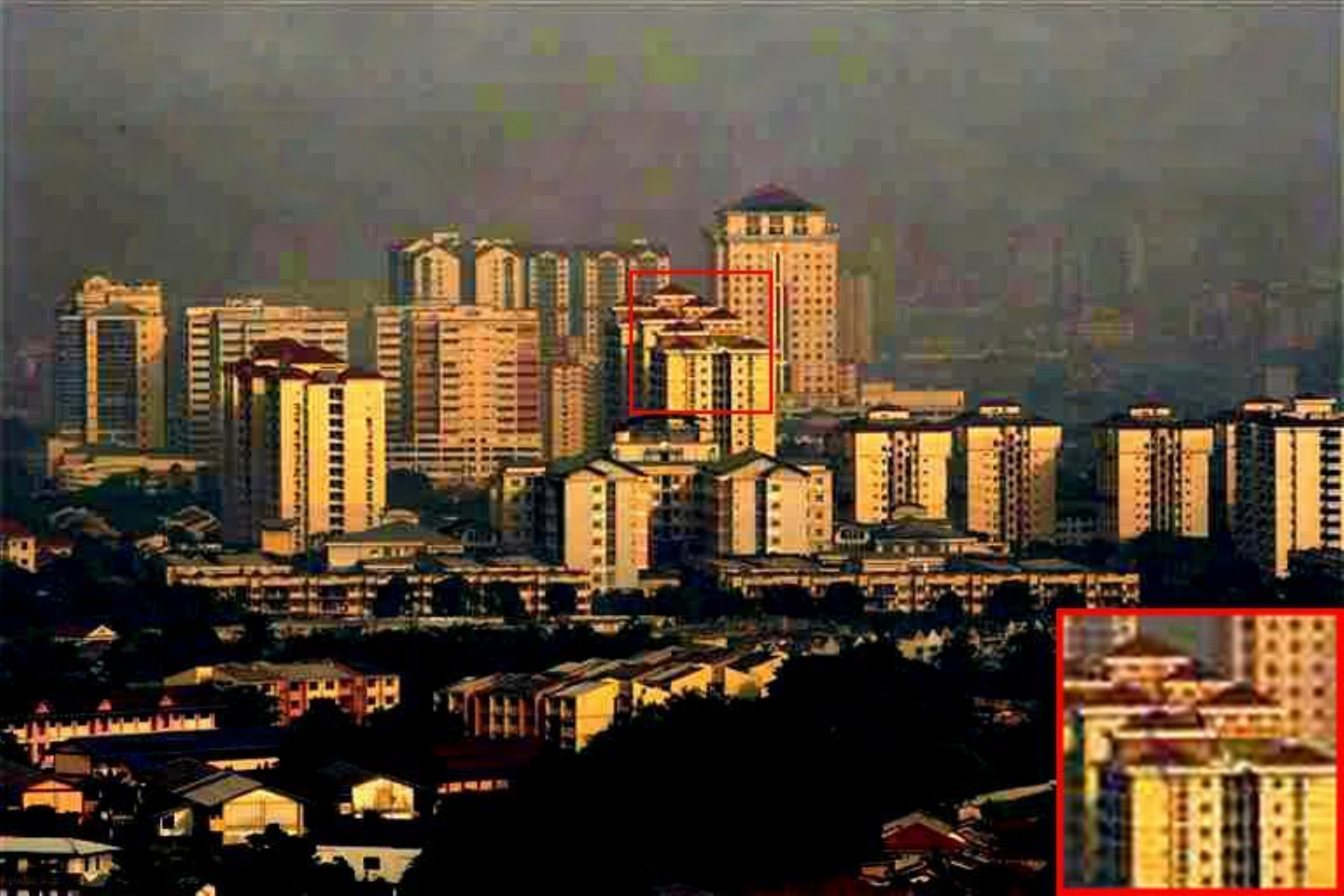}}
		\subfigure{\includegraphics[scale=\m_wid]{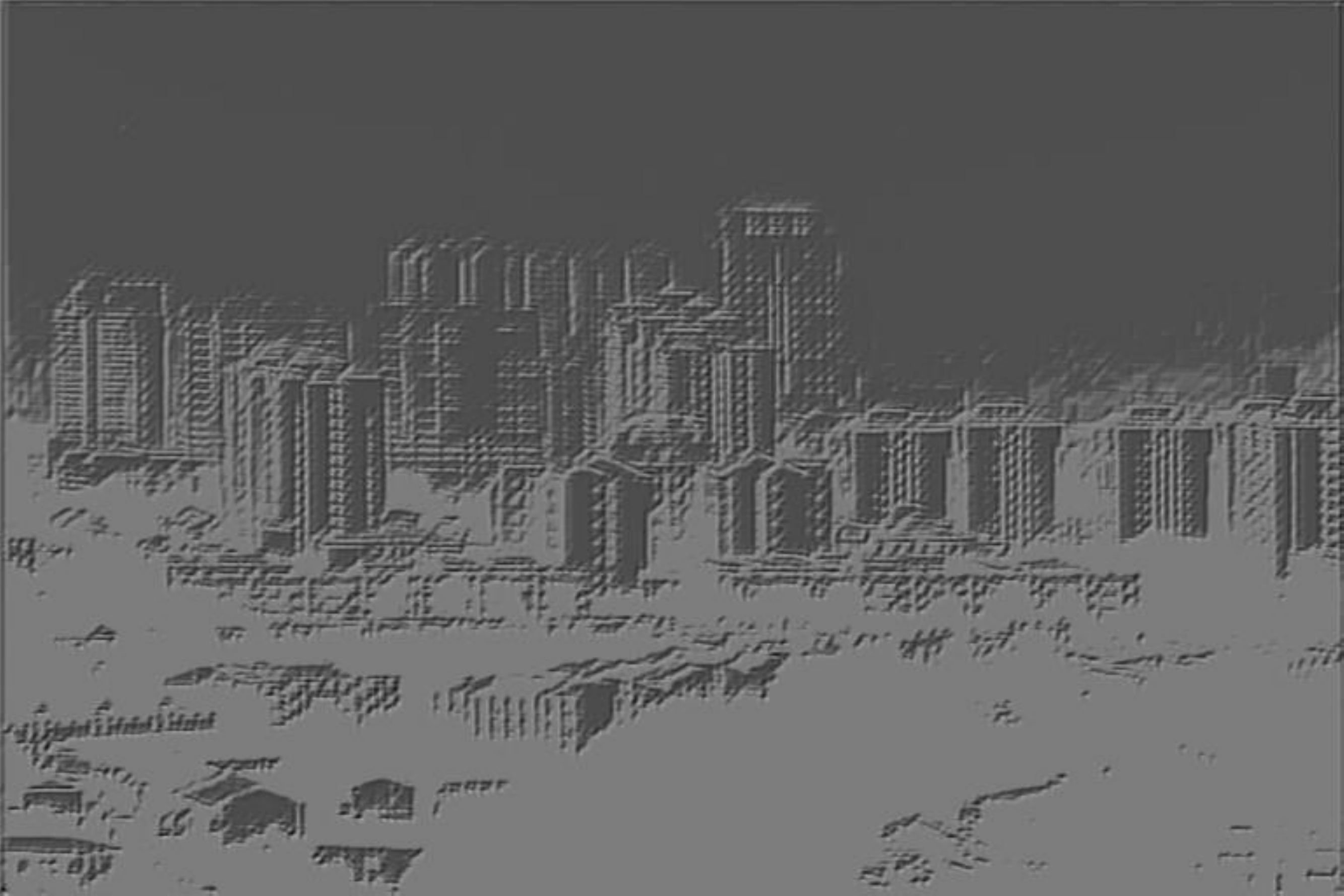}}
	\end{center}
	\vspace{-0.8cm}
	
	\begin{center}
	\def \m_wid{0.0564}
	\def \two_wid{0.1126}
	\setcounter{subfigure}{0}
		\subfigure[]{\label{Figure:RW:Input}\includegraphics[scale=\m_wid]{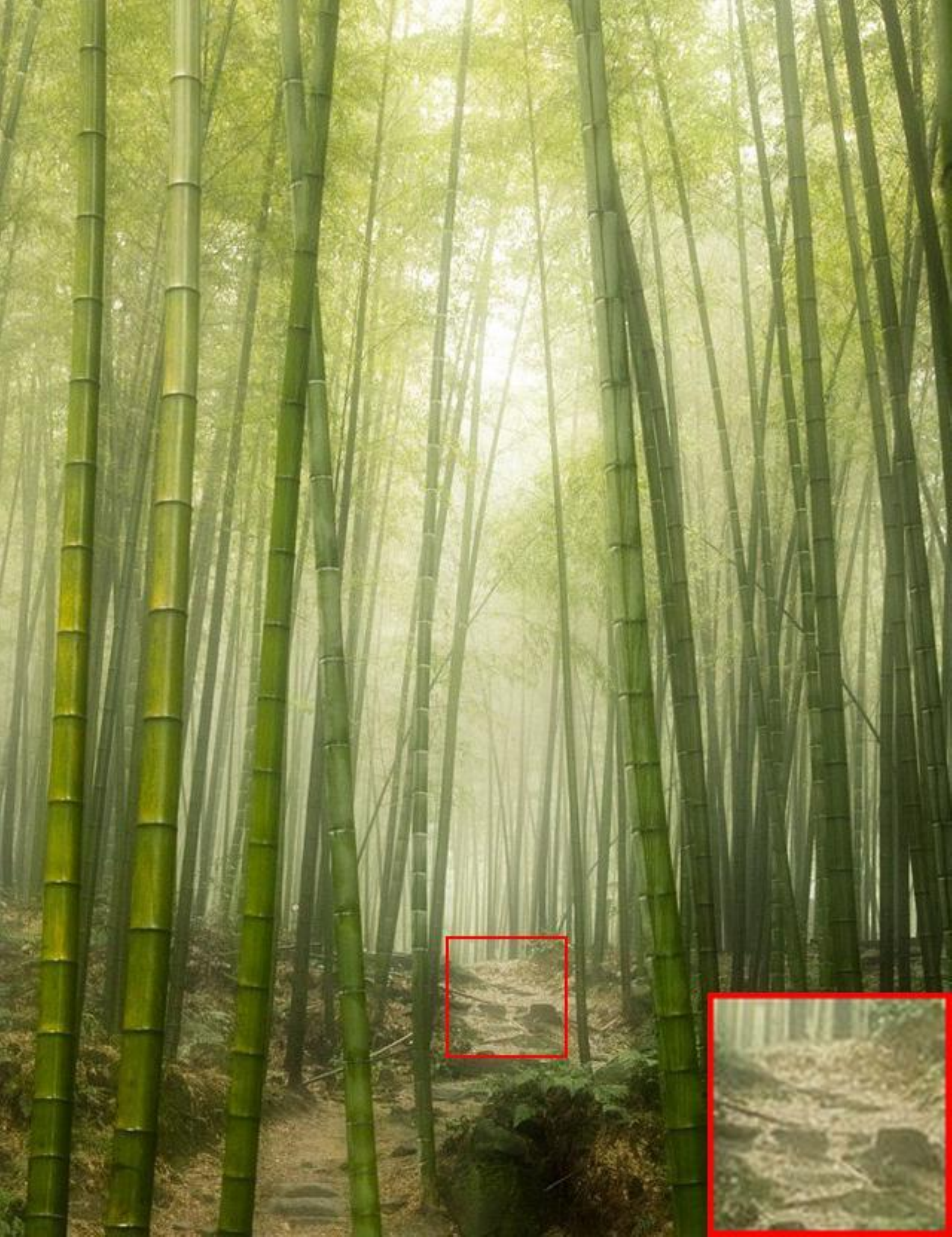}}
		\subfigure[]{\includegraphics[scale=\m_wid]{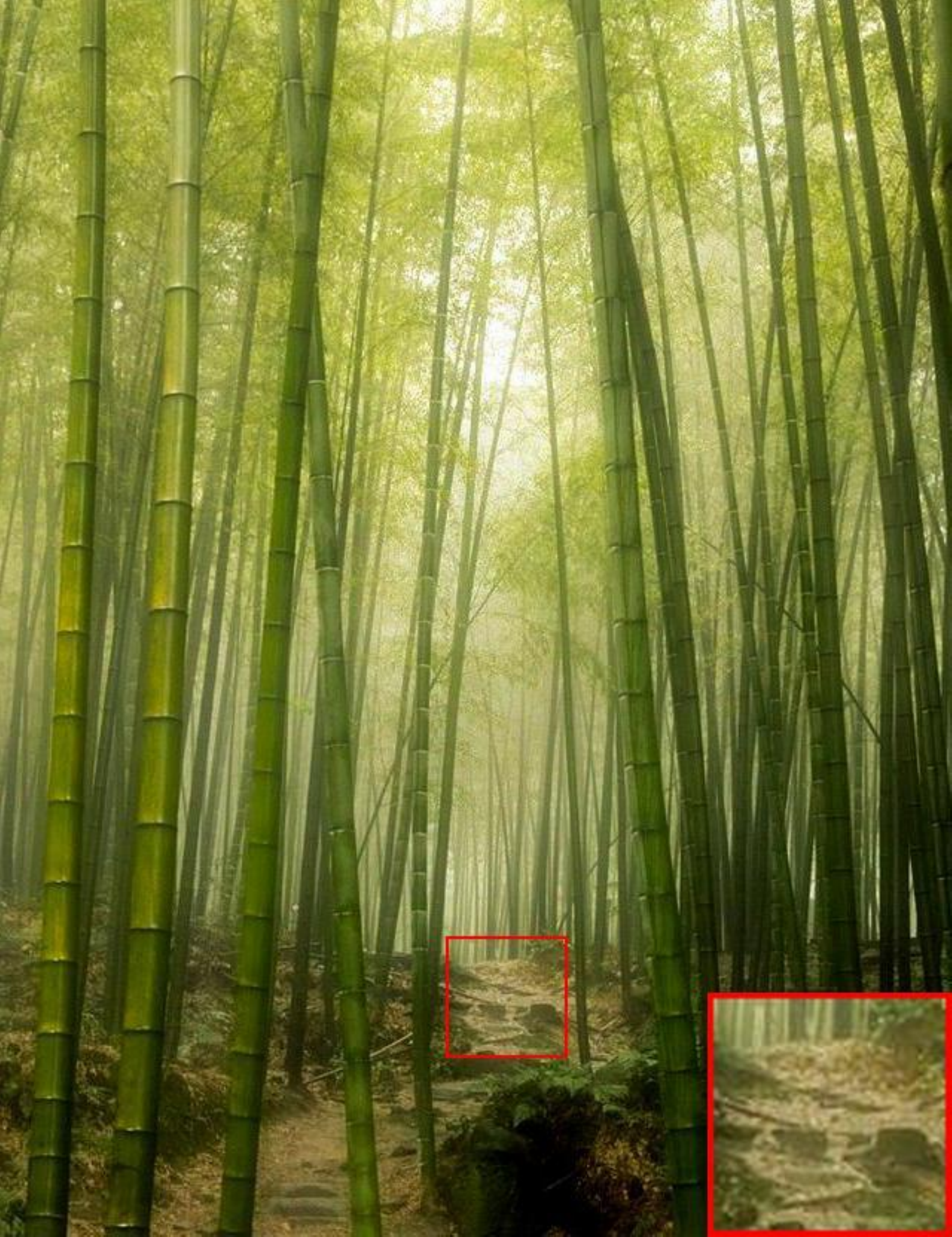}}
		\subfigure[]{\includegraphics[scale=\m_wid]{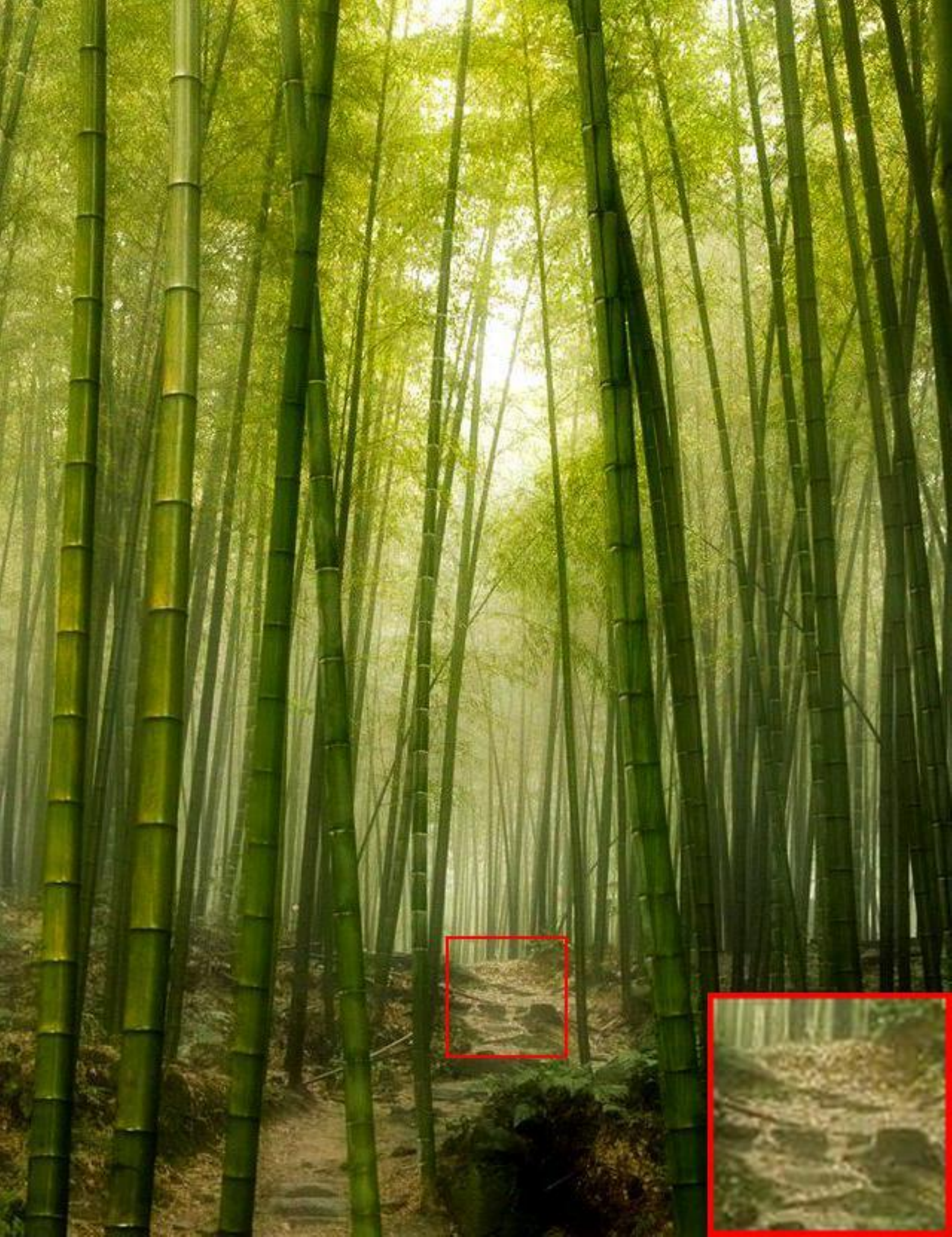}}
		\subfigure[]{\includegraphics[scale=\m_wid]{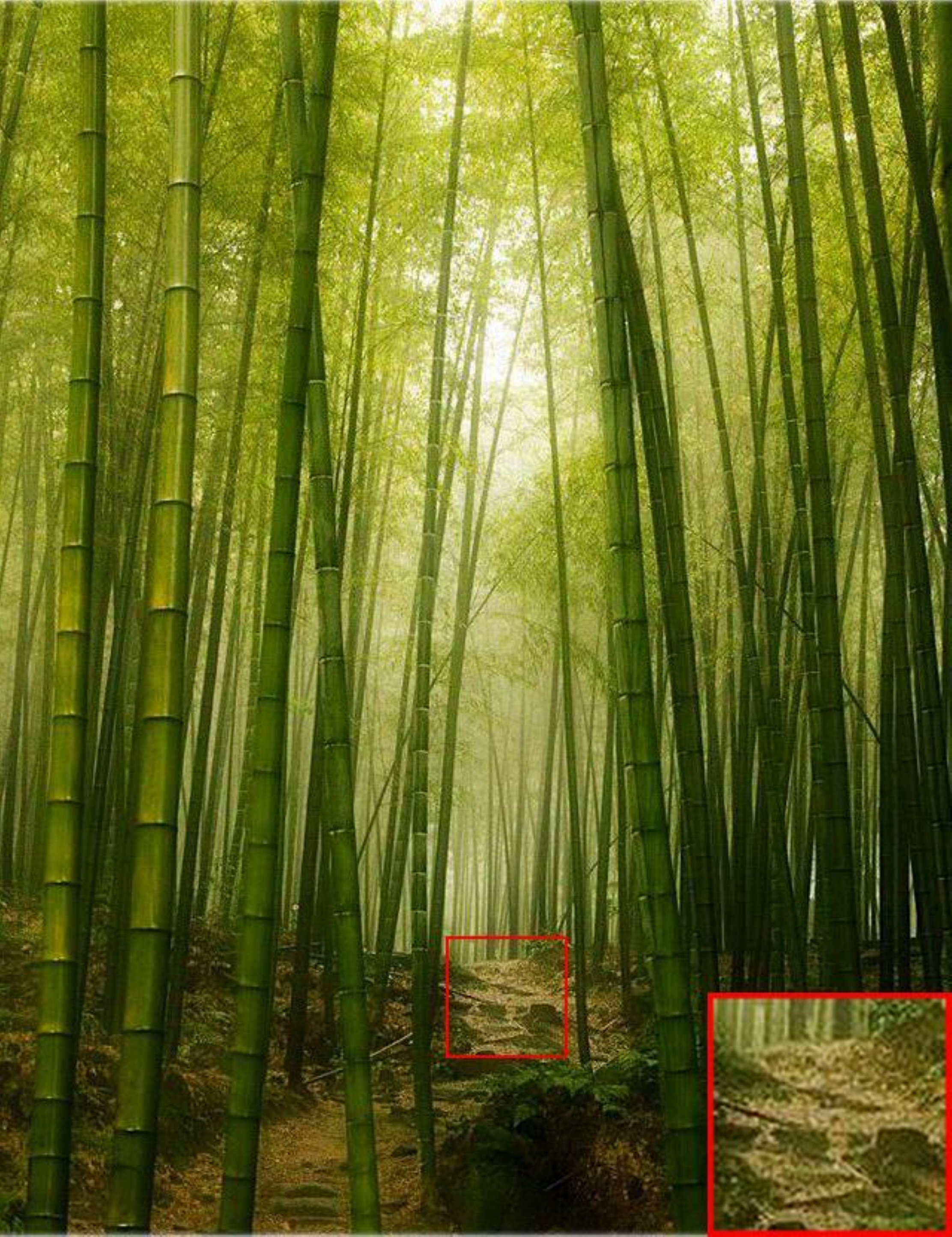}}
		\subfigure[]{\includegraphics[scale=\m_wid]{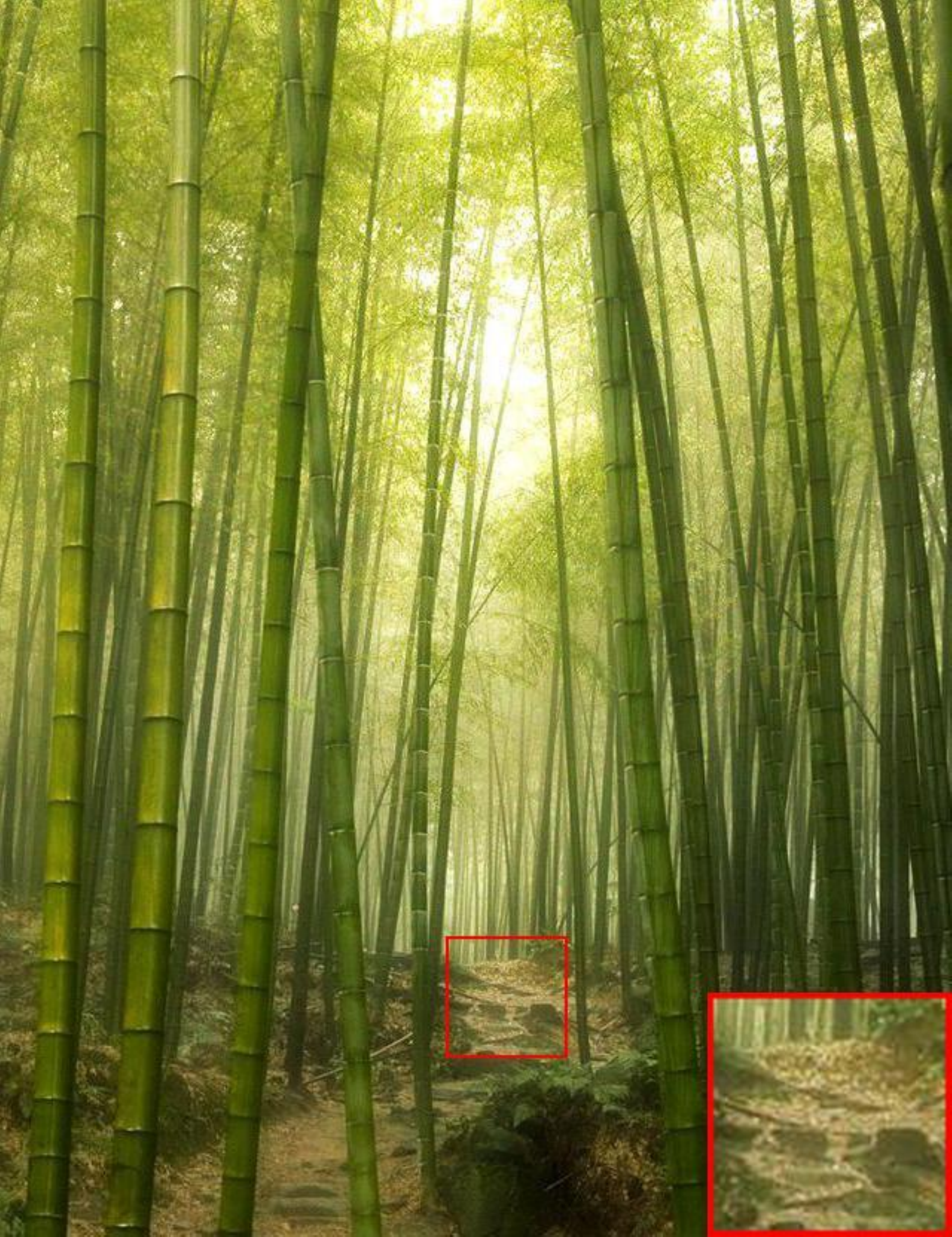}}
		\subfigure[]{\includegraphics[scale=\m_wid]{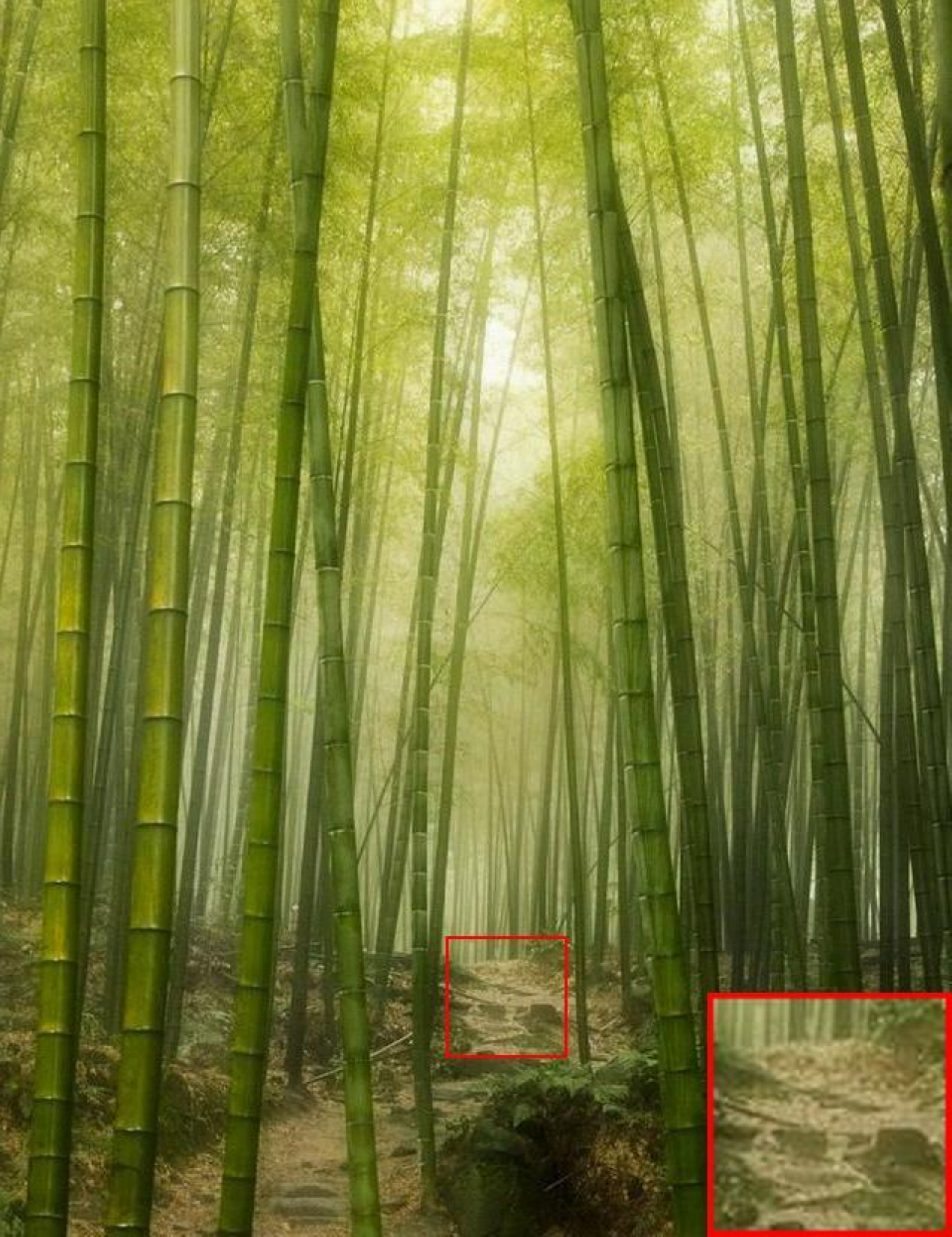}}
		\subfigure[]{\includegraphics[scale=\m_wid]{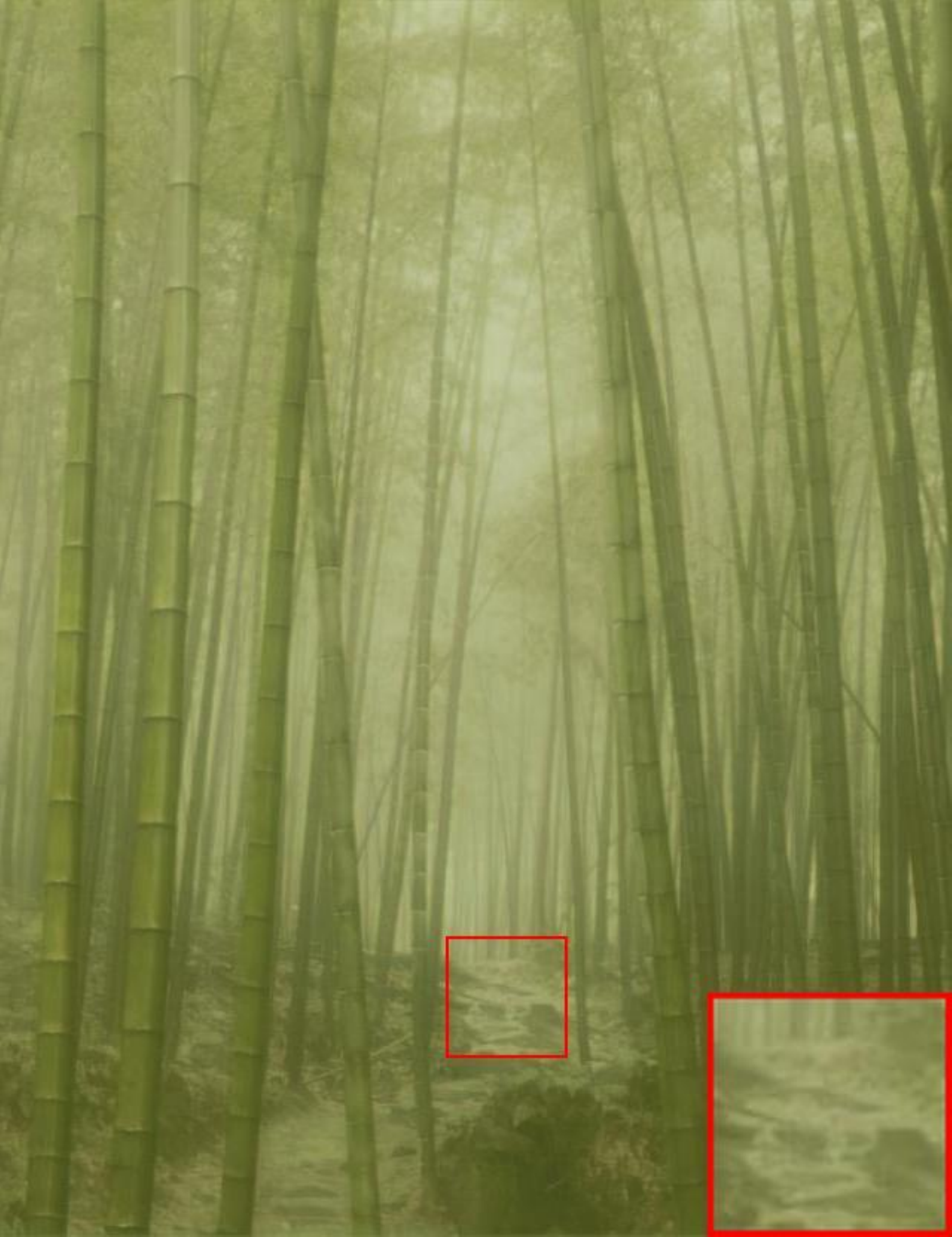}}
		\subfigure[]{\includegraphics[scale=\m_wid]{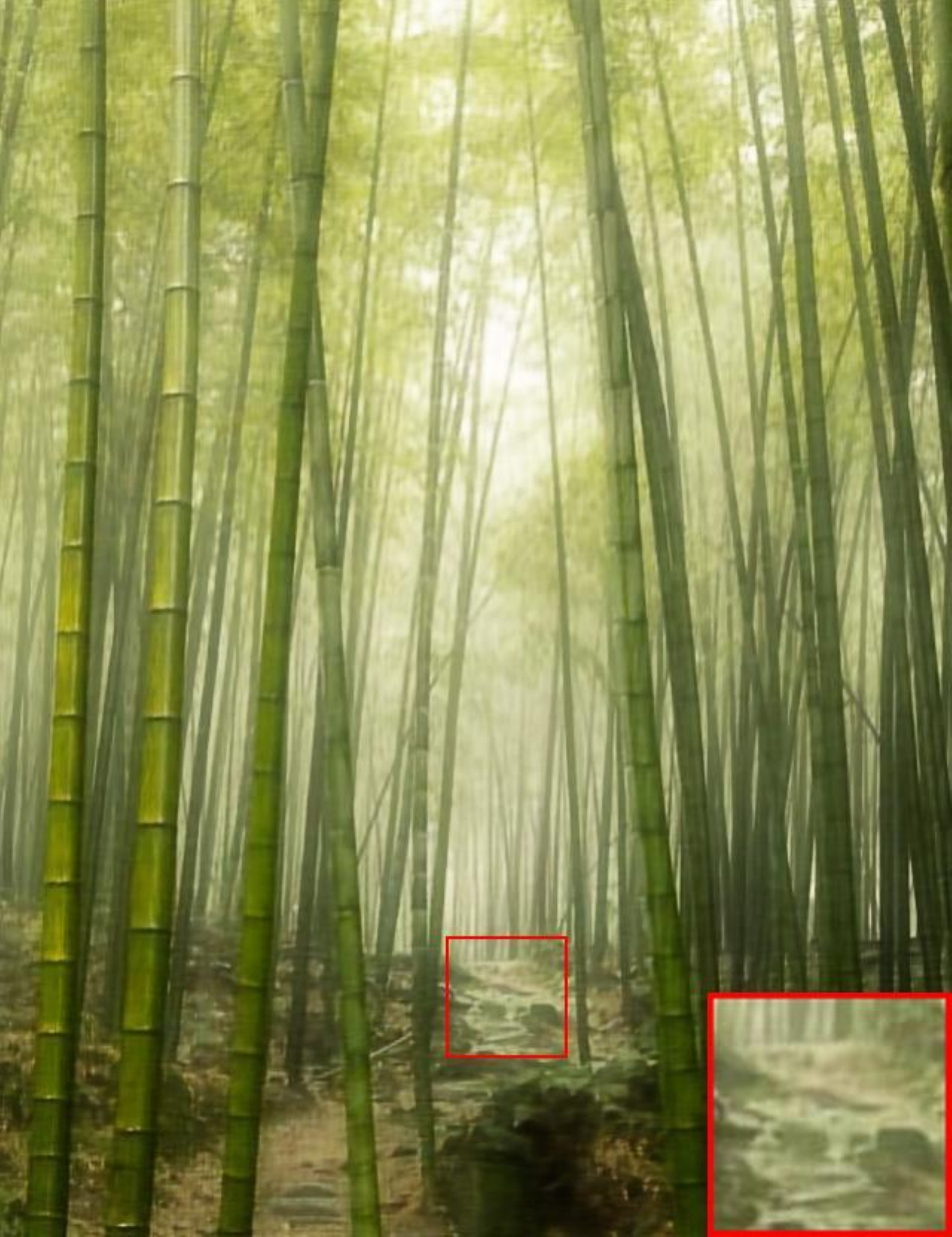}}
		\subfigure[]{\includegraphics[scale=\two_wid]{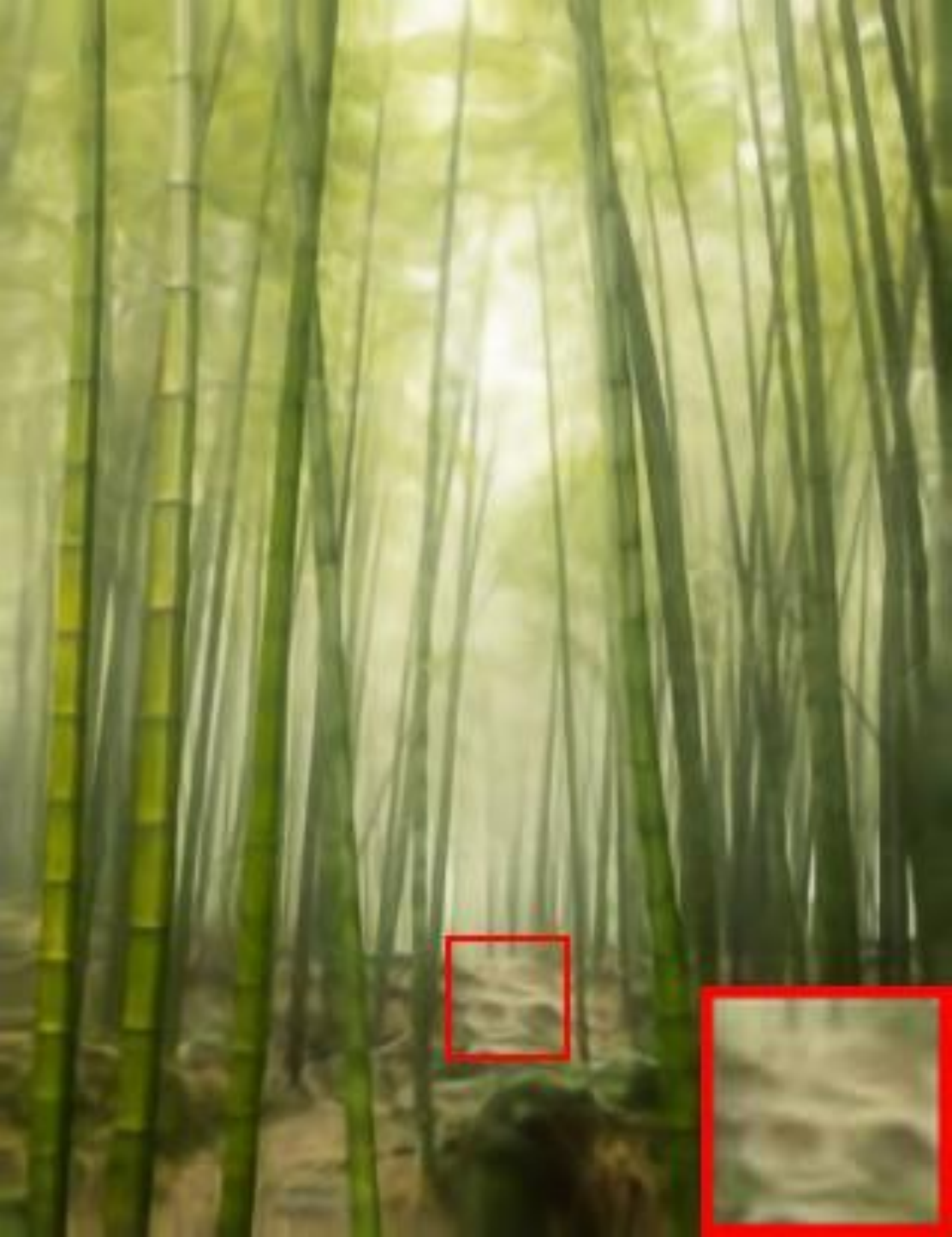}}
		\subfigure[]{\includegraphics[scale=\m_wid]{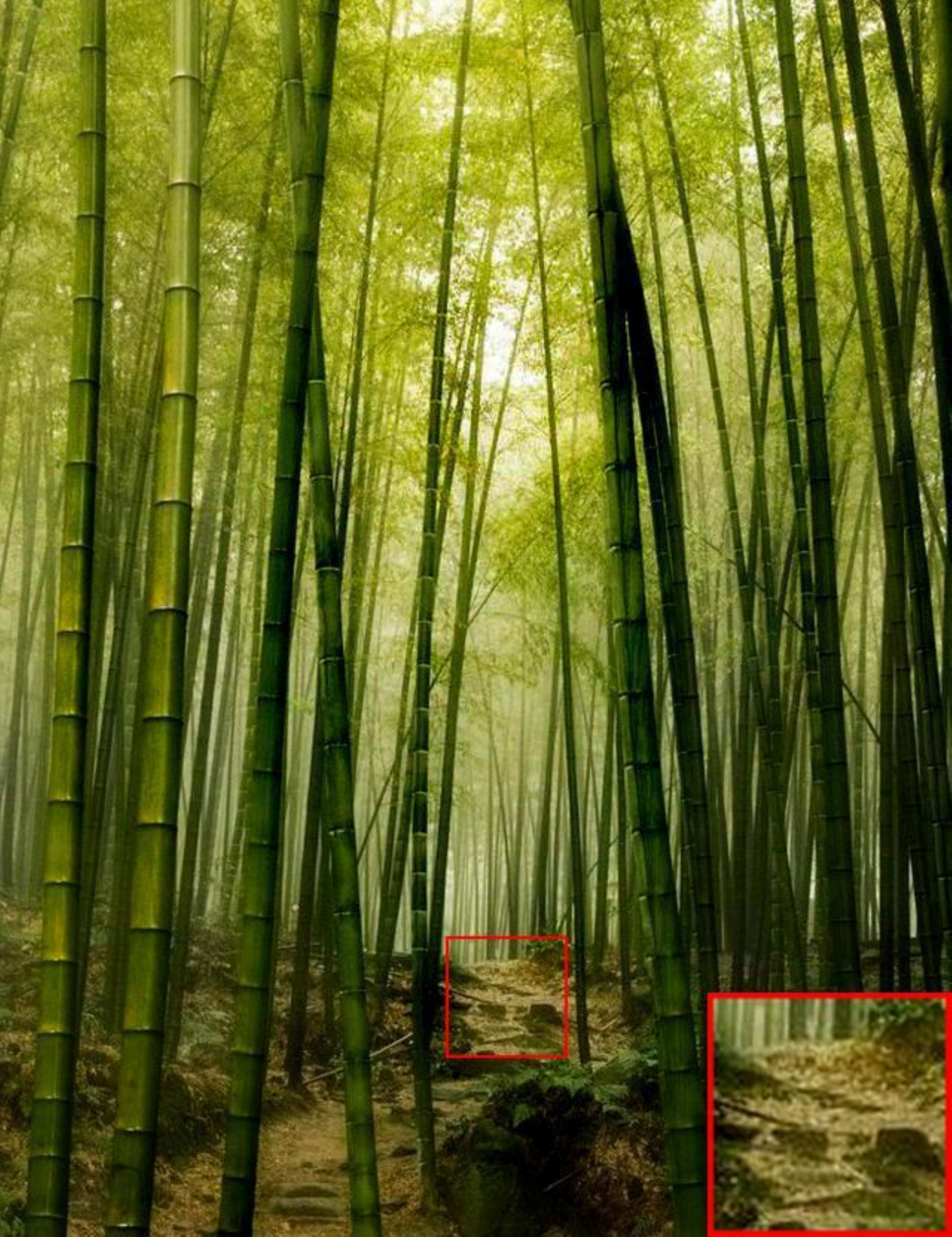}}
		\subfigure[]{\label{Figure:RW:YOLY}\includegraphics[scale=\m_wid]{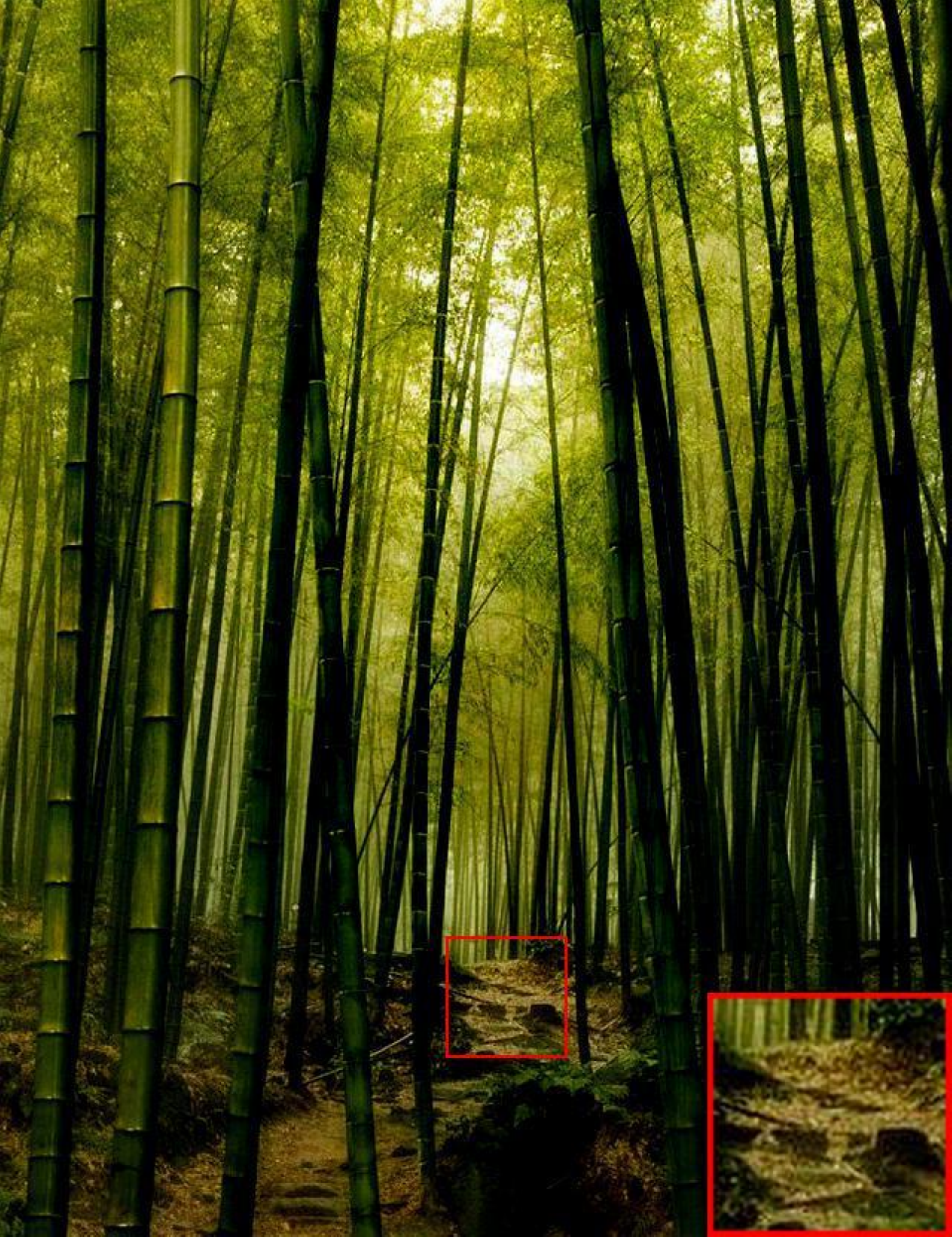}}
		\subfigure[]{\label{Figure:RW:YOLY}\includegraphics[scale=\m_wid]{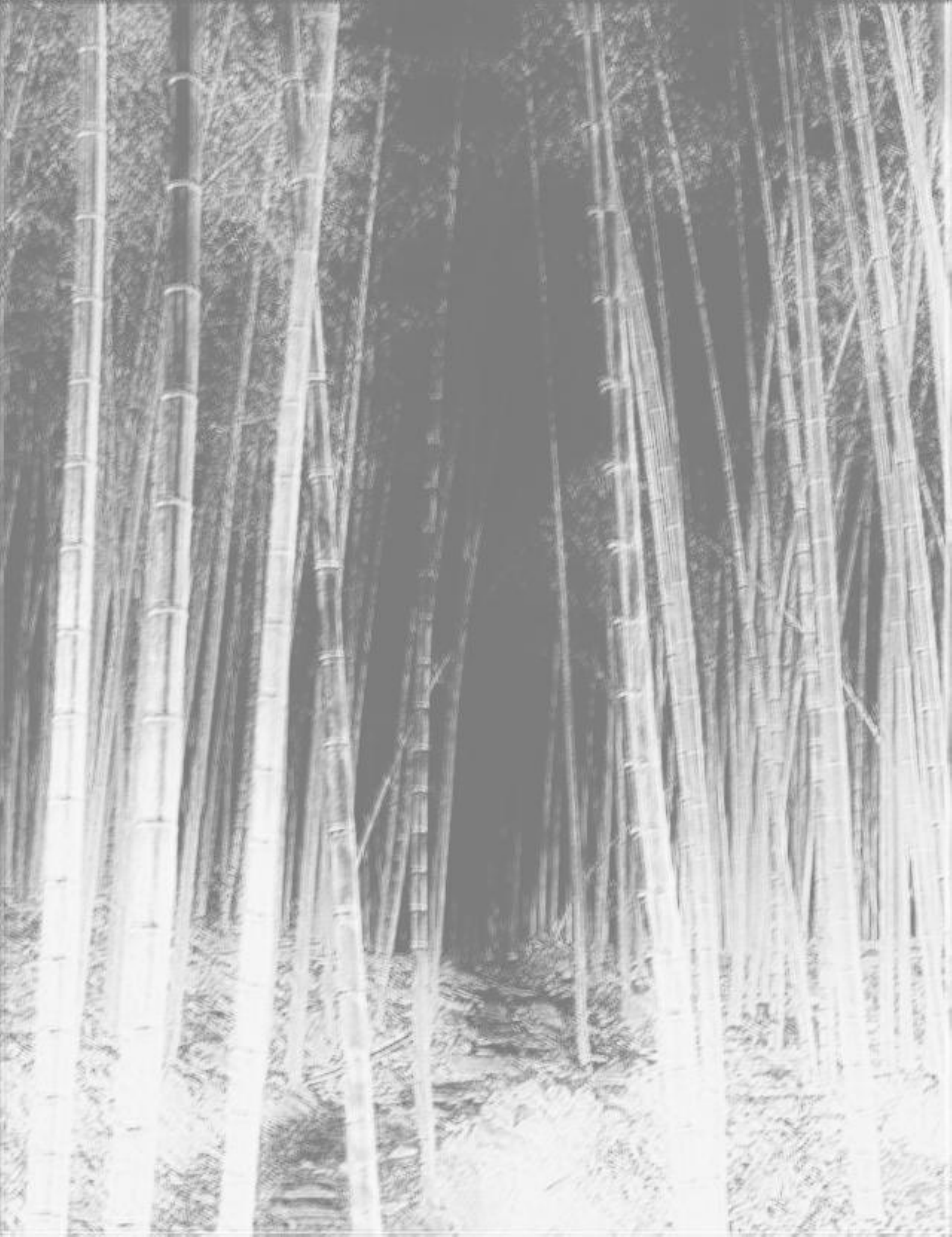}}
	\end{center}
	
	\caption{\label{Figure:RW} Visual Results on the Real-World scenes. From the left to the right column (\textit{i.e.}, Figs.(\ref{Figure:RW:Input}--\ref{Figure:RW:YOLY})), the input hazy image, DehazeNet~\citep{DehazeNet}, MSCNN~\citep{MSCNN}, AOD-Net~\citep{AOD-Net}, DCP~\citep{DCP}, GRM~\citep{GRM}, N2V~\citep{N2V}, DIP~\citep{DCP}, DD~\citep{DD}, DDIP~\citep{Double-DIP}, our method and the estimated transmission map are presented. Some areas are highlighted by red rectangles and zooming-in is recommended for a better visualization and comparison. }
\end{figure*}

\subsection{Comparisons on Real World Dataset}
To demonstrate the effectiveness of the proposed method in real-world hazy scenes, we carry out qualitative experiments on the HSTS real-world image set which is without the ground-truth clean image. 

From Fig.~\ref{Figure:RW}, one could observe that our YOLY demonstrates the best visualization result in almost all scenes. For example, although DehazeNet, MSCNN, AOD-Net, DCP, and DDIP successfully remove most of haze in the pictures, they fail to handle the areas with much more details. Besides, the methods such as DCP and DDIP also suffer from the color distortions in the background. DehazeNet and AOD-Net lose some details after dehazing in the low-light areas. In contrast, our method could be immune from these issues and get a much more favorite recovery. On this dataset, YOLY takes about 59.27s for dehazing each image and each epoch costs about 118.54ms. 

Besides the results on the HSTS real-world image set, we also conduct comparisons on 10 hazy images collected from Internet by us. As shown in Fig.~\ref{Figure:RW}, the proposed YOLY is remarkably superior to the baselines. Moreover, we also illustrate the transmission map learned by YOLY for a more comprehensive investigation in the last column. 


\subsection{Haze Transfer}

Recent works have witnessed the effectiveness of supervised neural networks in image dehazing. To well train the neural networks, a large scale hazy-clean image pairs are required, and most of works resort to the synthetic dataset by manually specifying the parameters of the physical model. Such a handcrafted haze creation solution has suffered from a variety of limitations, \textit{e.g.}, the domain shift when using the synthetic hazy images to address the real-world images. To solve this problem, it is highly expected to develop new haze creation methods which work in a learning manner. However, to the best of our knowledge, there are few efforts have devoted to this problem so far. 

As a by-product of our layer disentanglement idea, our method could transfer haze from a given hazy image to another clean one. In other words, we  provide a novel solution to haze creation. To transfer the haze, our method first disentangles the atmospheric light and the transmission map from the hazy image. After that, the model is further used to generate the new hazy images w.r.t. a given clean image. 

\begin{figure}[!t]
	\def \m_width{0.11}
	\def \a_height{1.5cm}

	\begin{center}
	\subfigure{\includegraphics[width=\m_width\textwidth, height=\a_height]{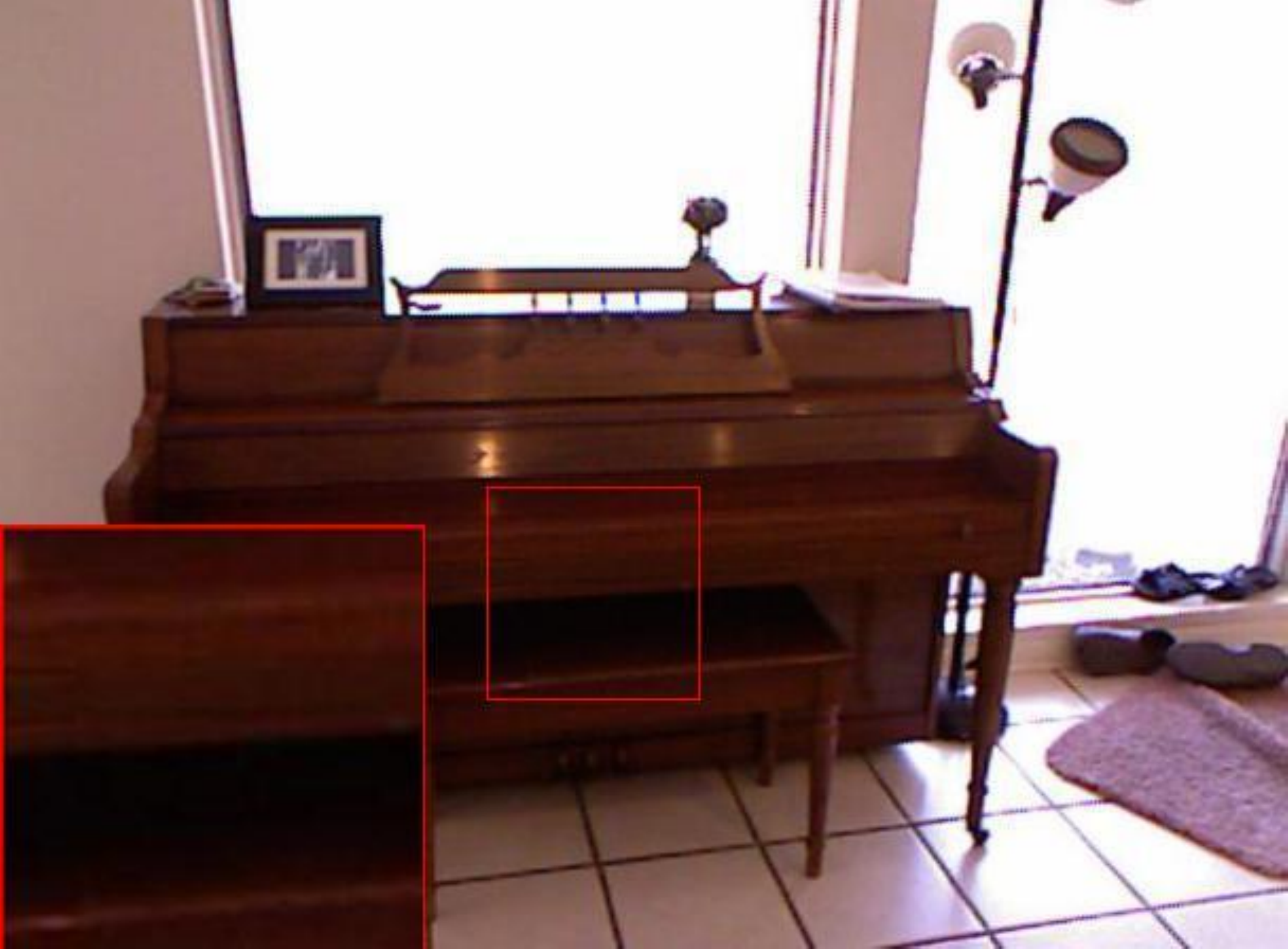}}
	\subfigure{\includegraphics[width=\m_width\textwidth, height=\a_height]{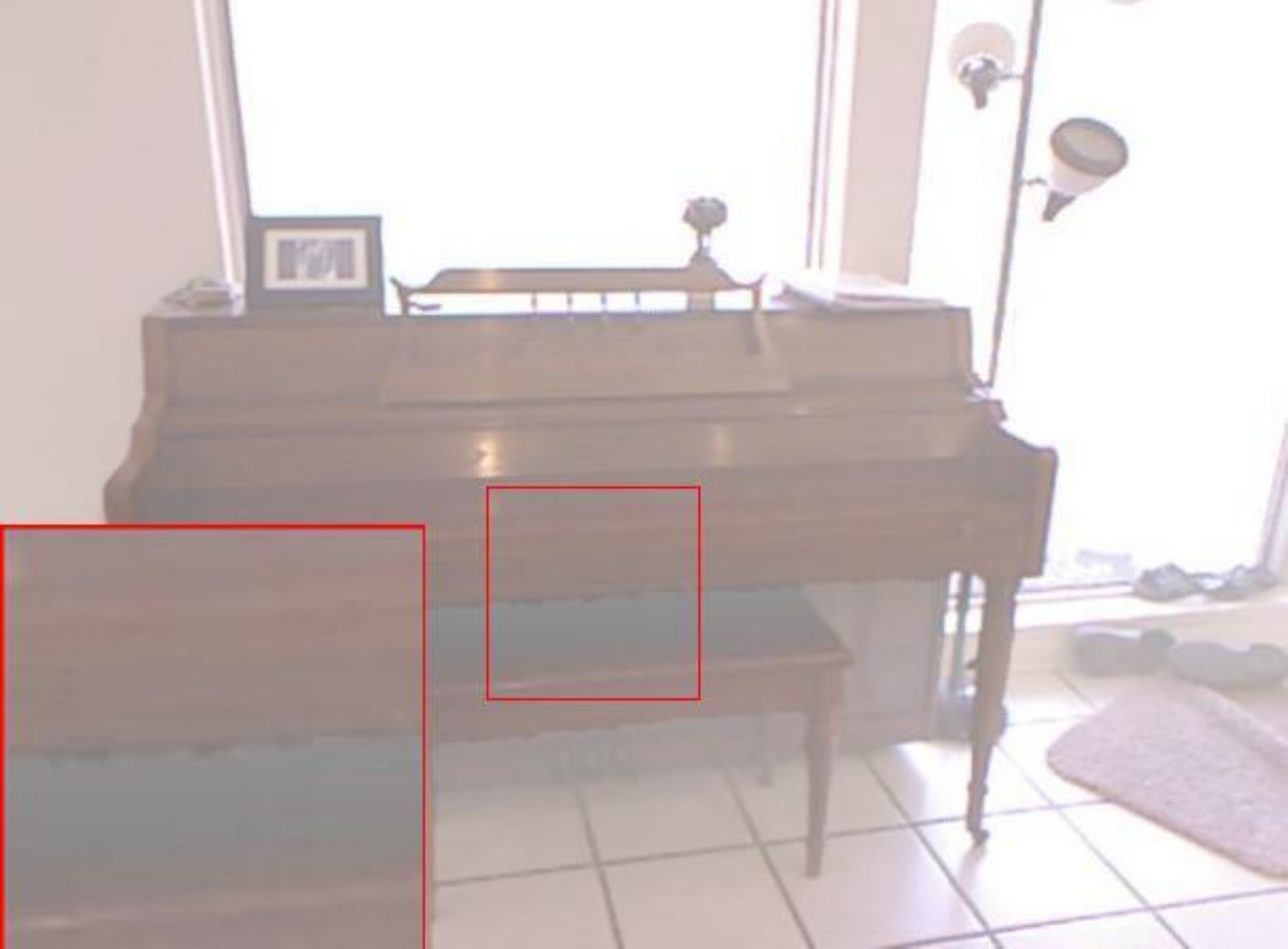}}
	\subfigure{\includegraphics[width=\m_width\textwidth, height=\a_height]{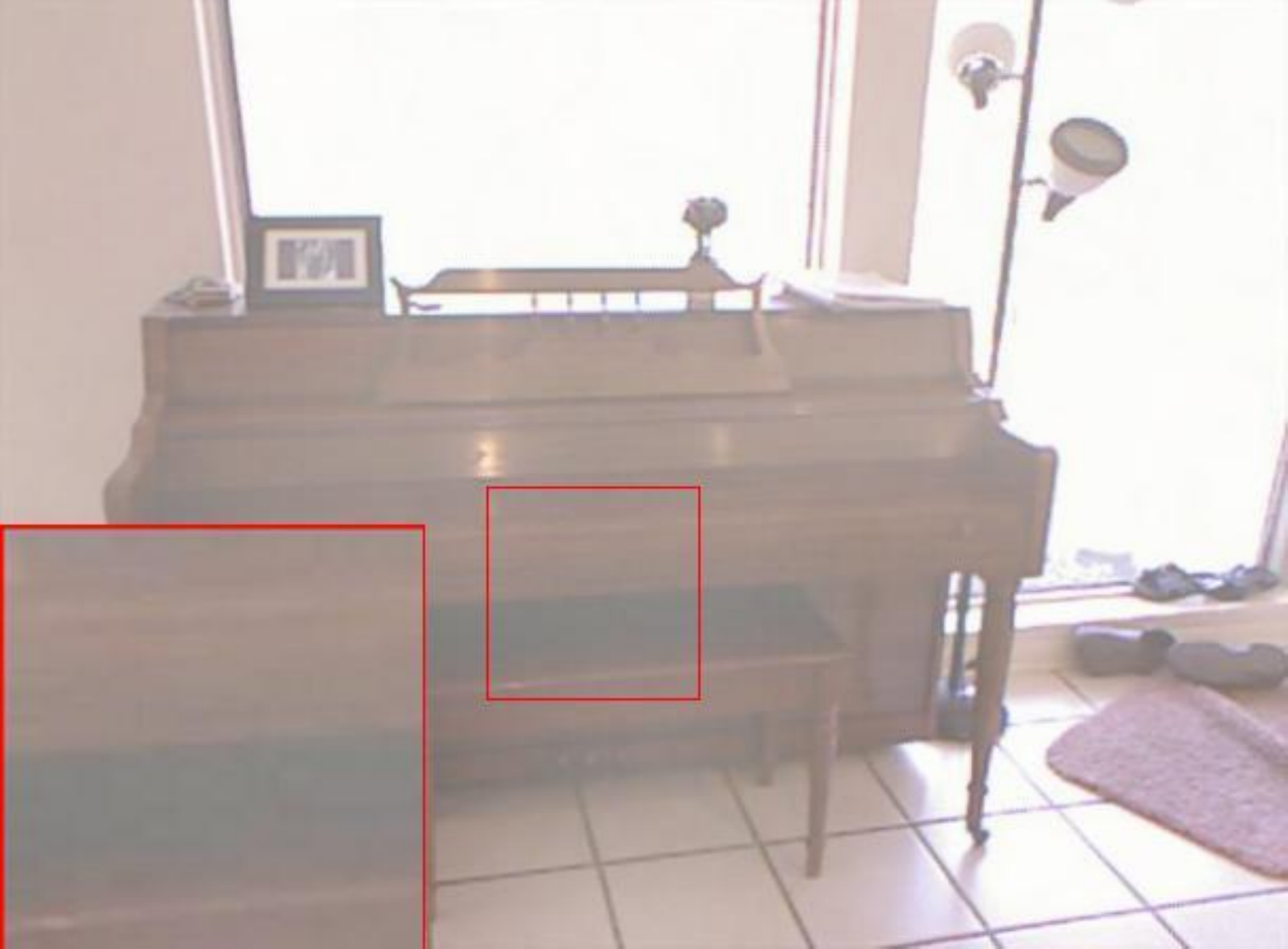}}
	\subfigure{\includegraphics[width=\m_width\textwidth, height=\a_height]{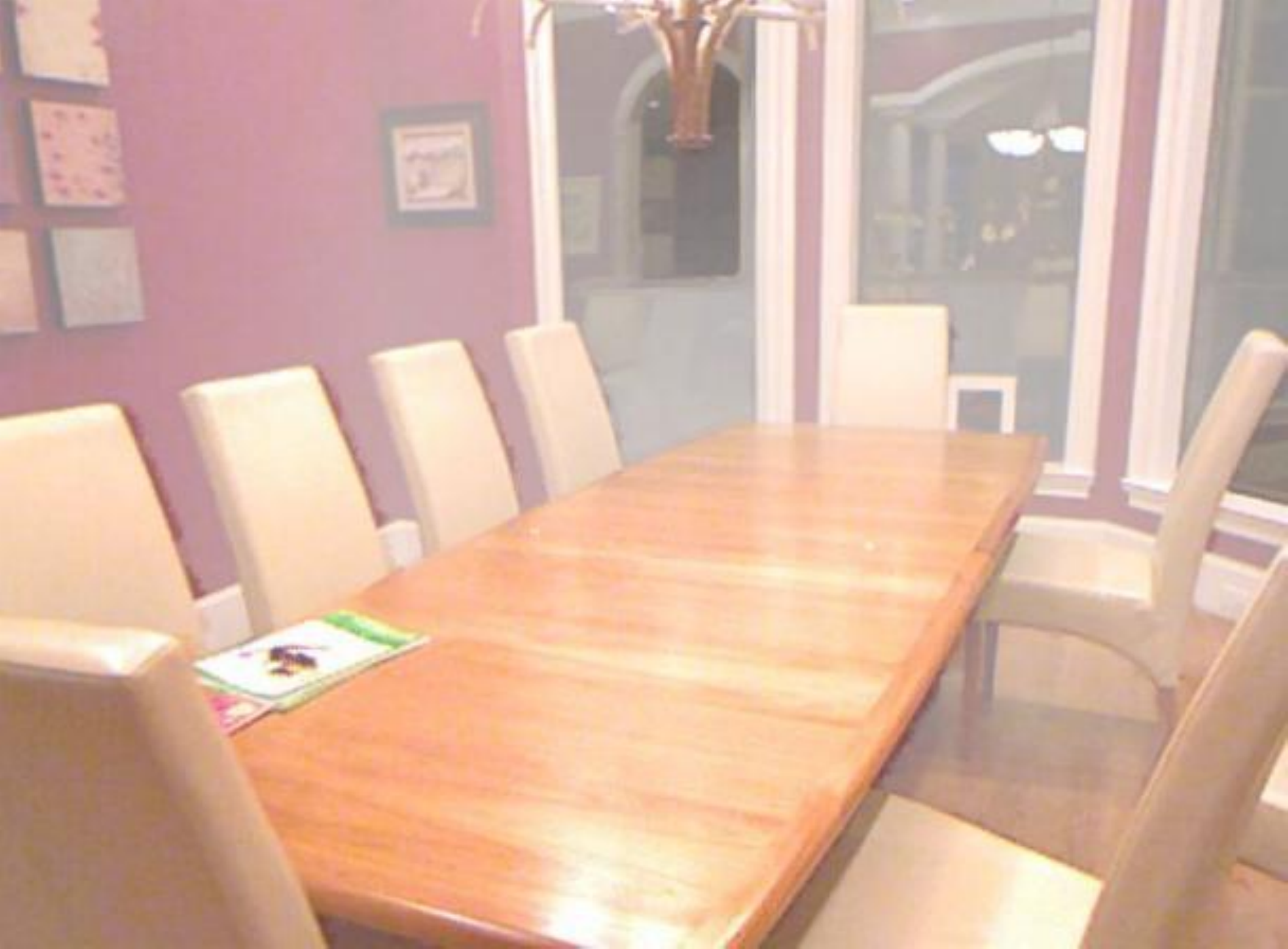}}
	\end{center}
	\vspace{-0.7cm}
	
	\begin{center}
	\subfigure{\includegraphics[width=\m_width\textwidth, height=\a_height]{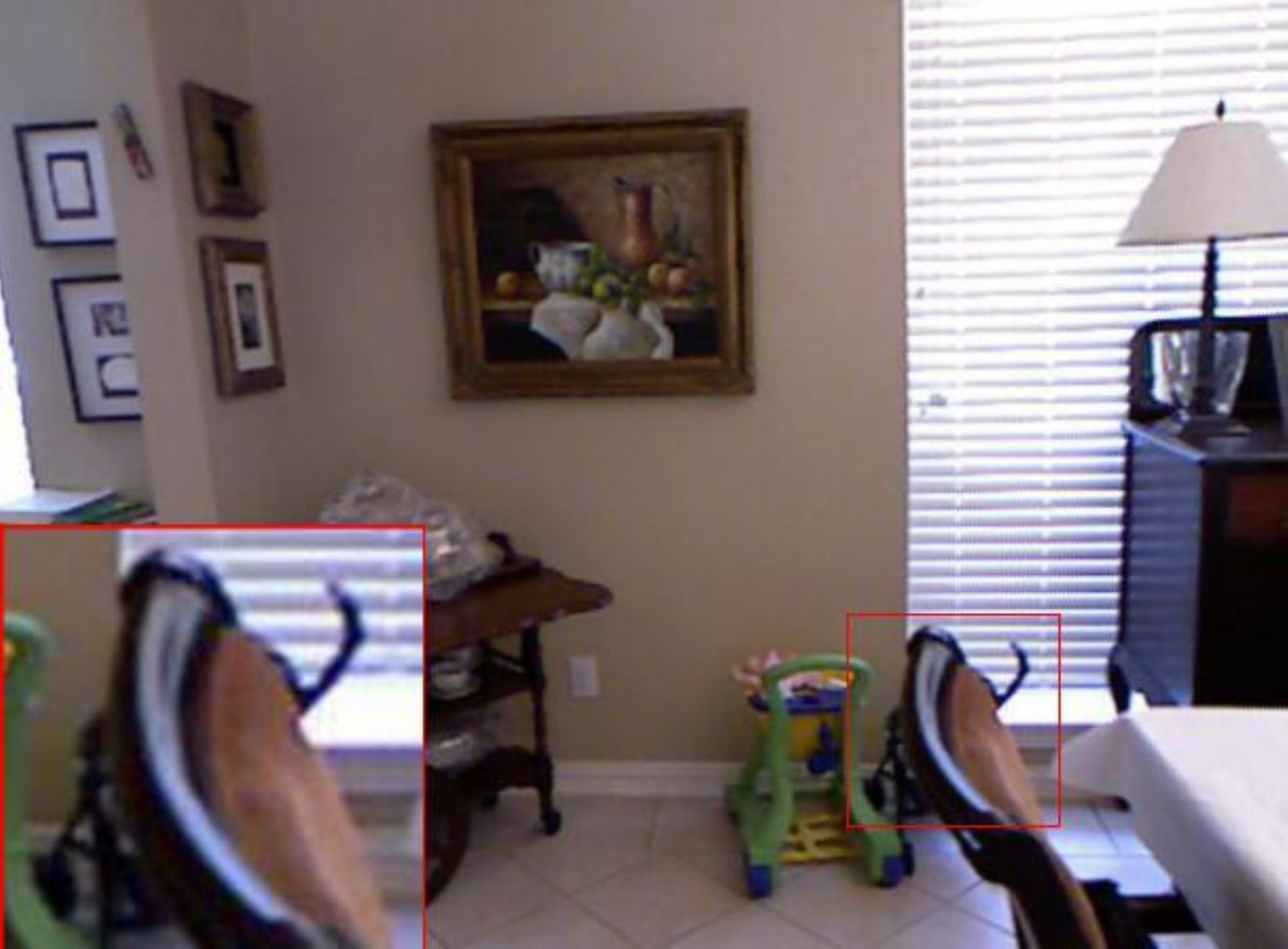}}
	\subfigure{\includegraphics[width=\m_width\textwidth, height=\a_height]{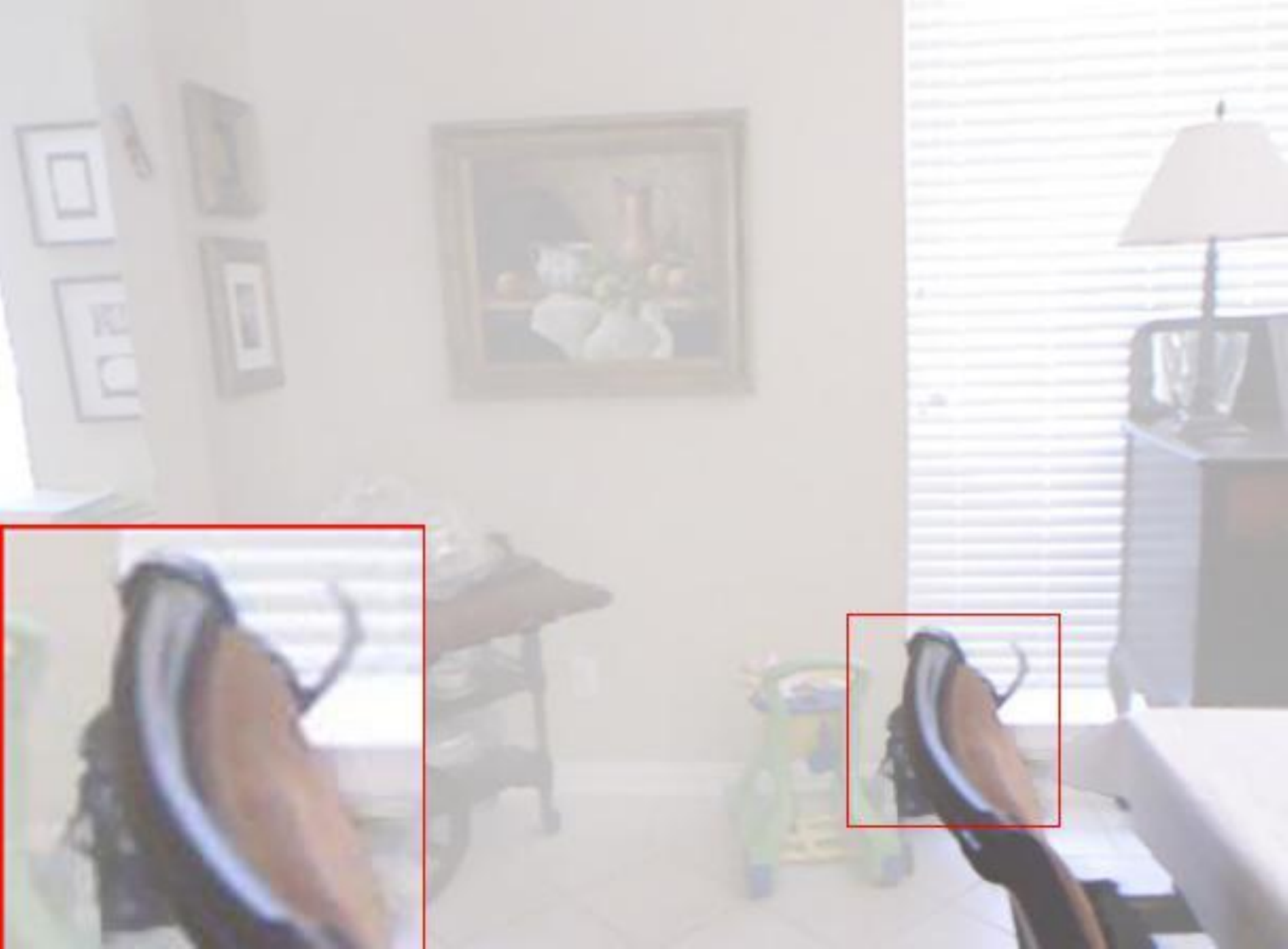}}
	\subfigure{\includegraphics[width=\m_width\textwidth, height=\a_height]{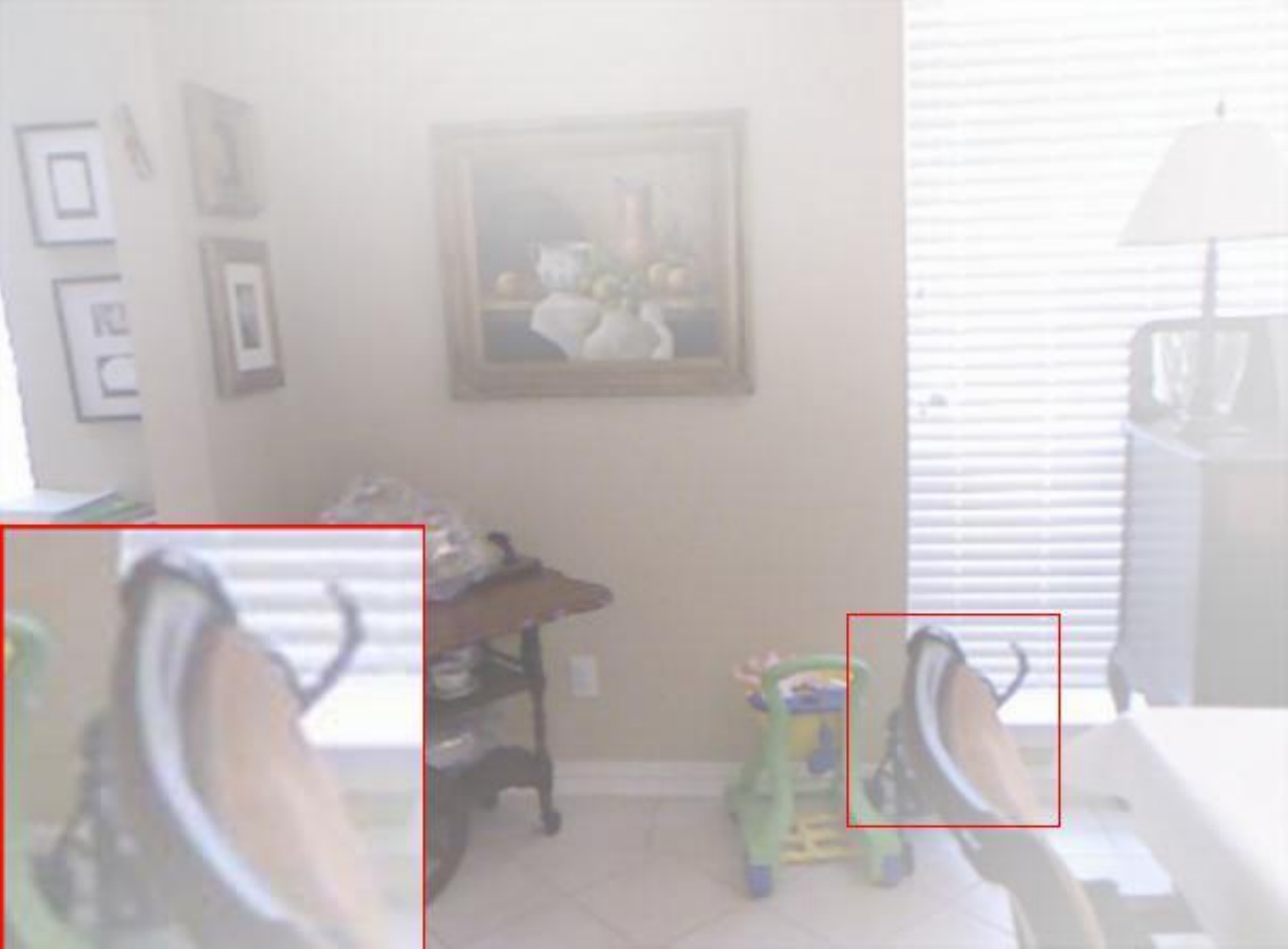}}
	\subfigure{\includegraphics[width=\m_width\textwidth, height=\a_height]{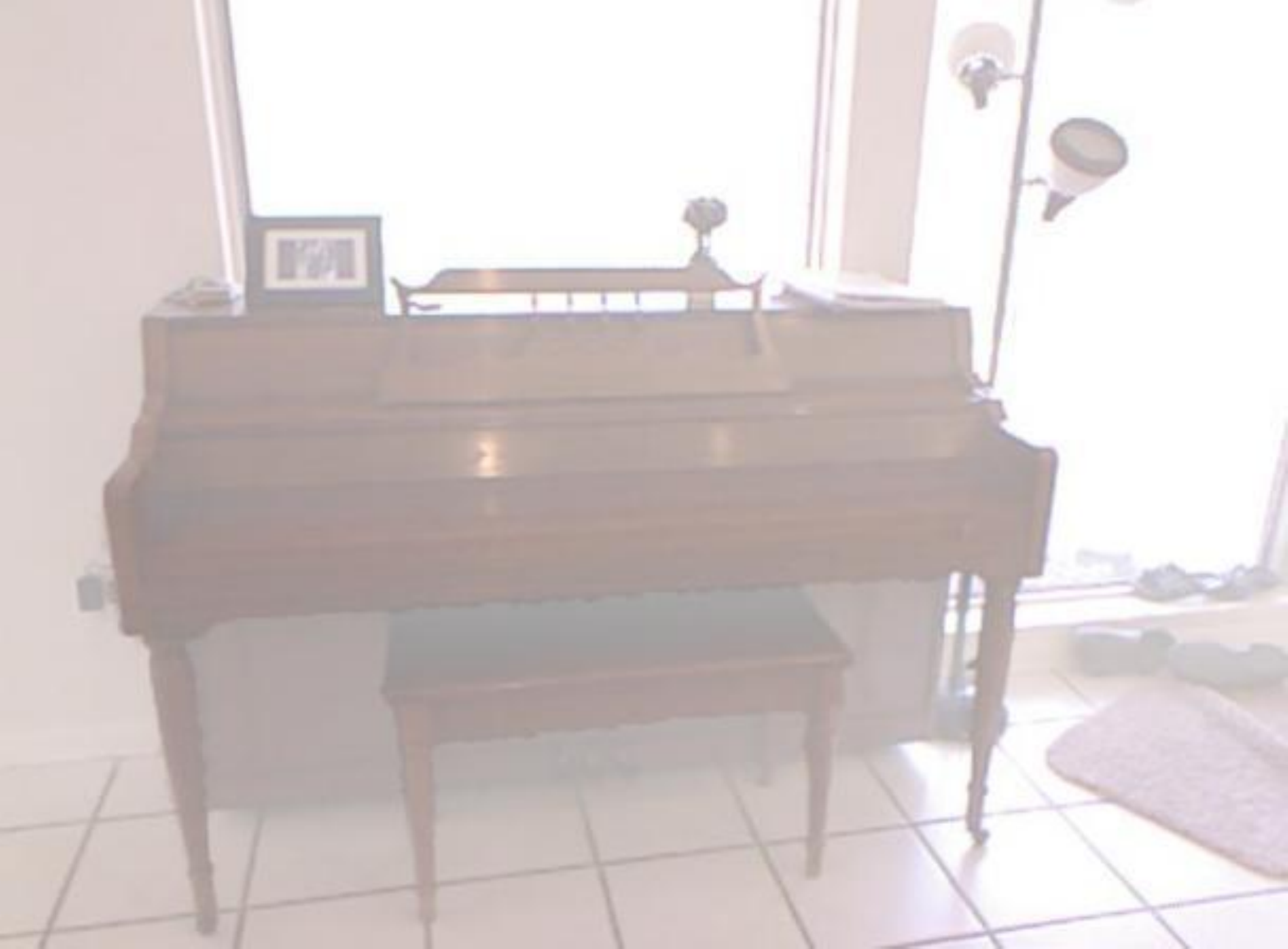}}
	\end{center}
	\vspace{-0.7cm}
	
	\begin{center}
	\subfigure{\includegraphics[width=\m_width\textwidth, height=\a_height]{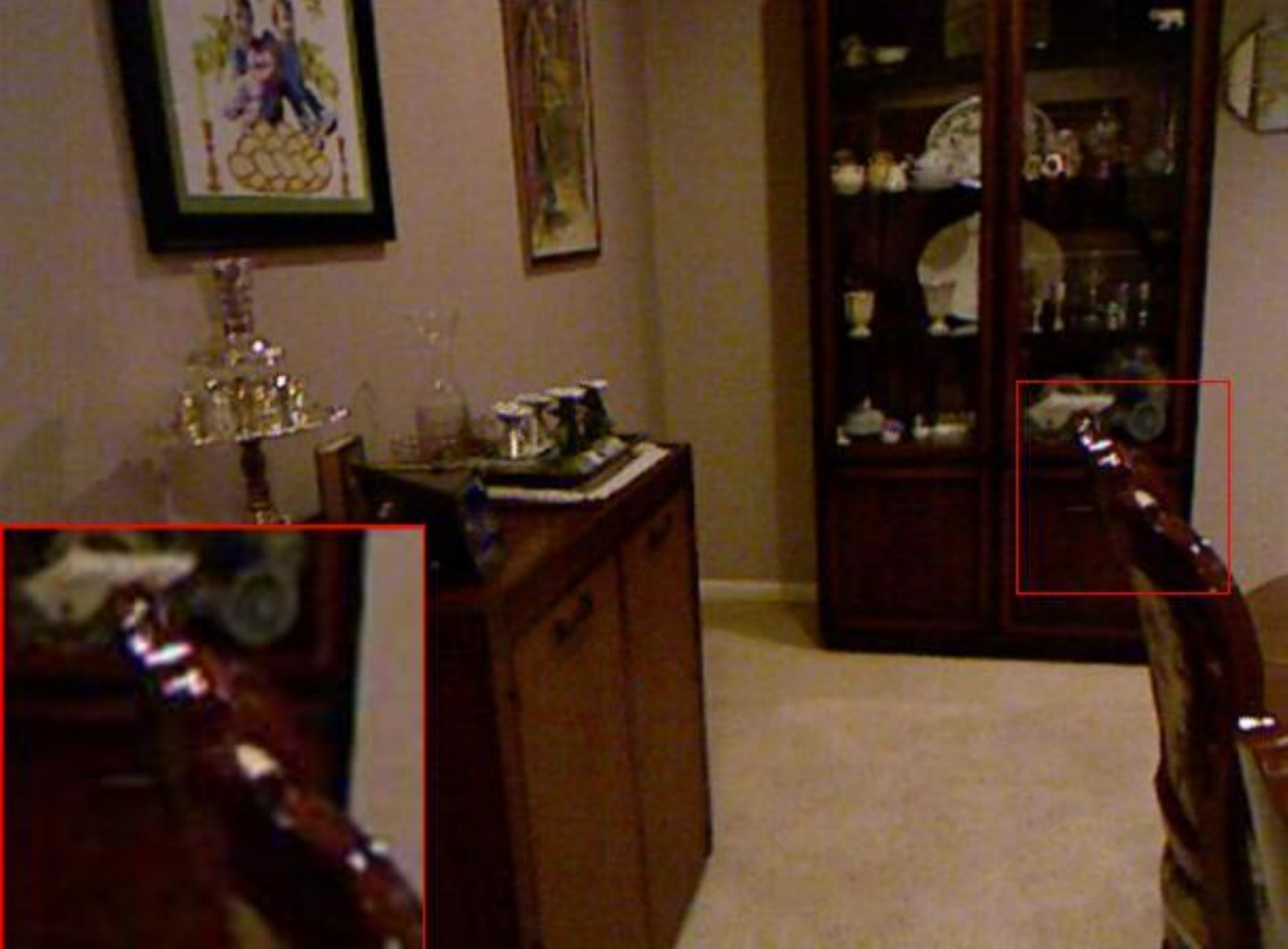}}
	\subfigure{\includegraphics[width=\m_width\textwidth, height=\a_height]{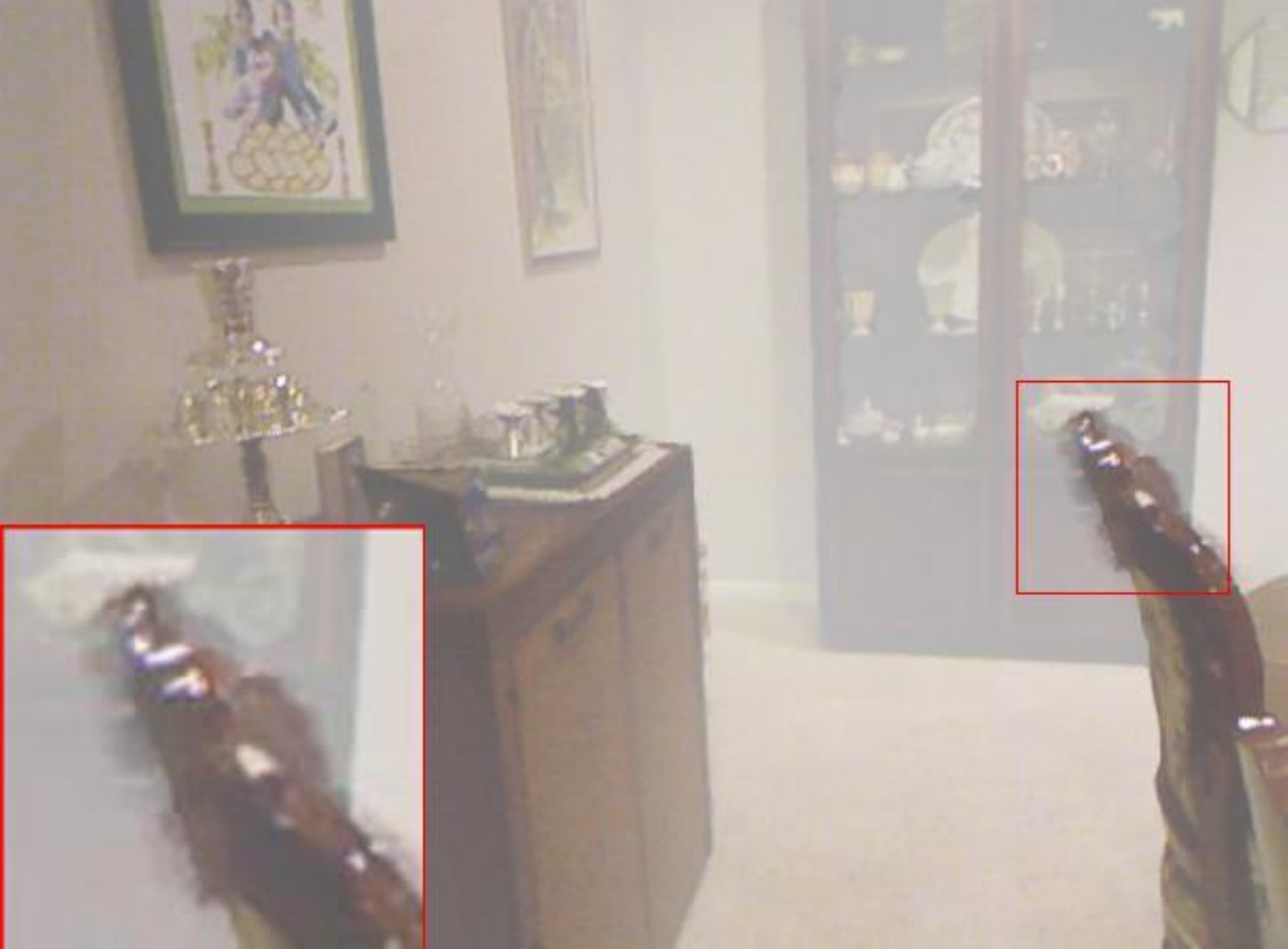}}
	\subfigure{\includegraphics[width=\m_width\textwidth, height=\a_height]{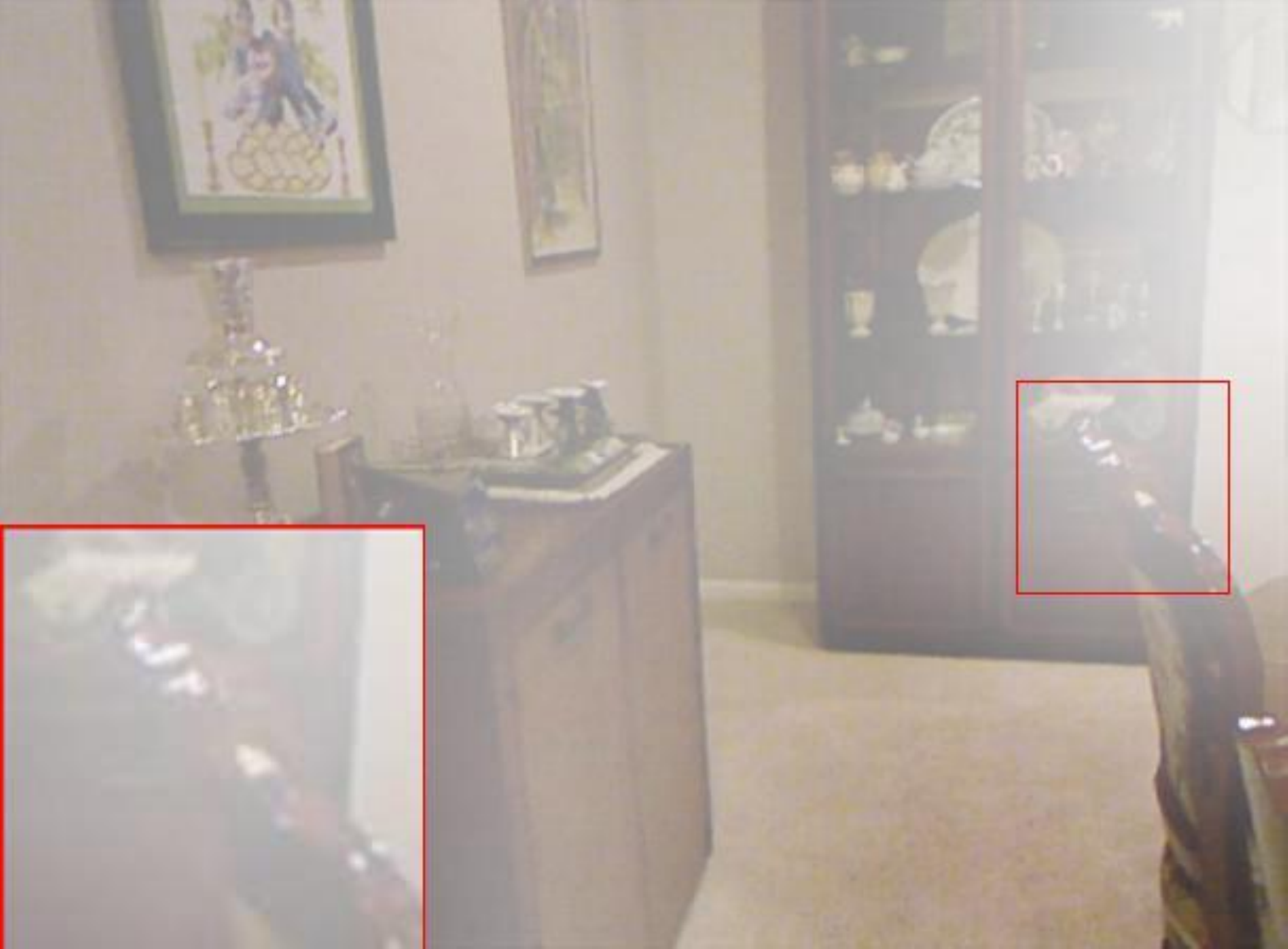}}
	\subfigure{\includegraphics[width=\m_width\textwidth, height=\a_height]{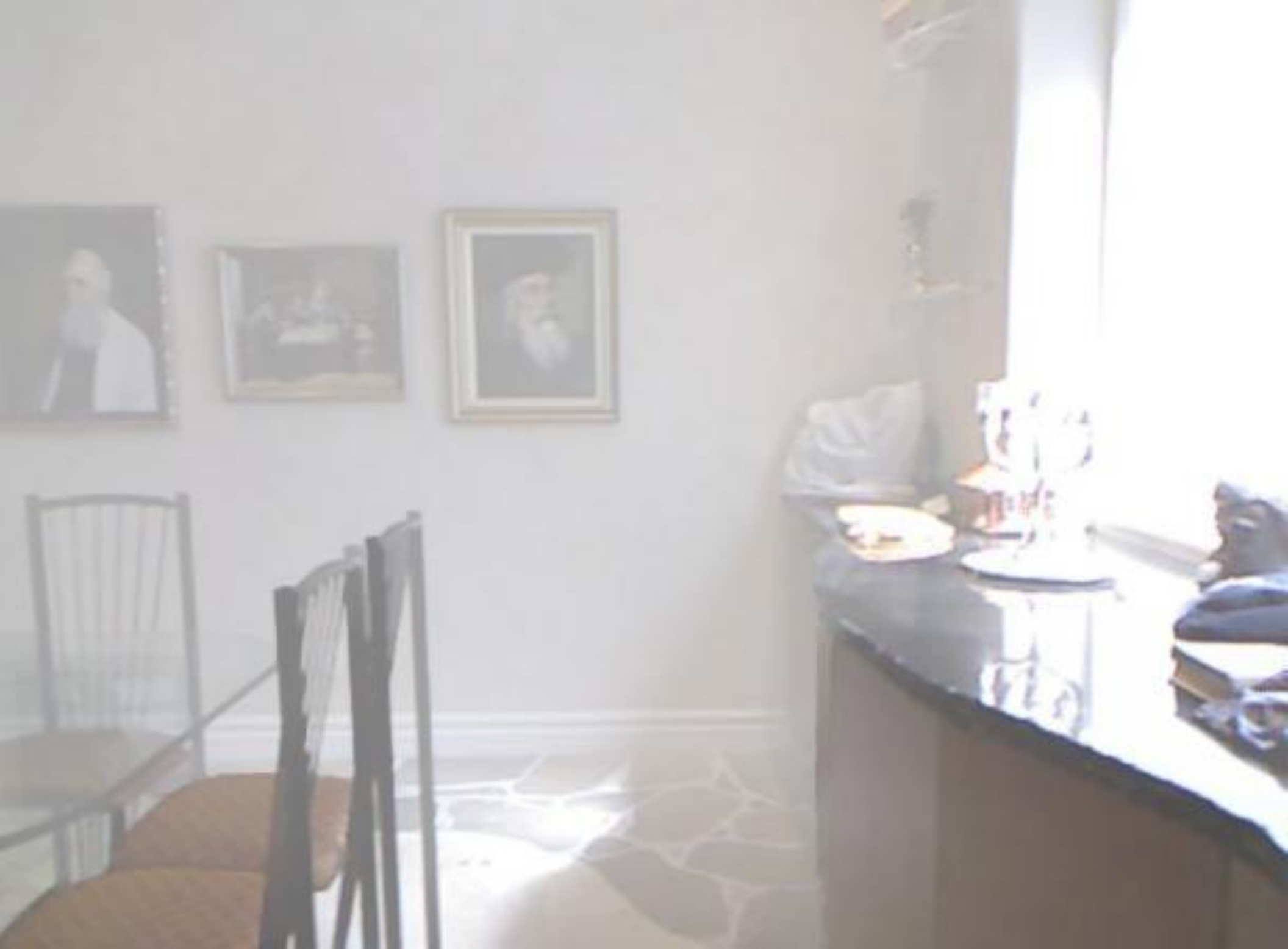}}	
	\end{center}
	\vspace{-0.7cm}
	
	\begin{center}
	\setcounter{subfigure}{0}
	\subfigure[]{\label{Transfer:a_i2i}\includegraphics[width=\m_width\textwidth, height=\a_height]{./Fig/Transfer/i2i/1423/1423_gt.pdf}}
	\subfigure[]{\label{Transfer:b_i2i}\includegraphics[width=\m_width\textwidth, height=\a_height]{./Fig/Transfer/i2i/1423/1423_hazy.pdf}}
	\subfigure[]{\label{Transfer:c_i2i}\includegraphics[width=\m_width\textwidth, height=\a_height]{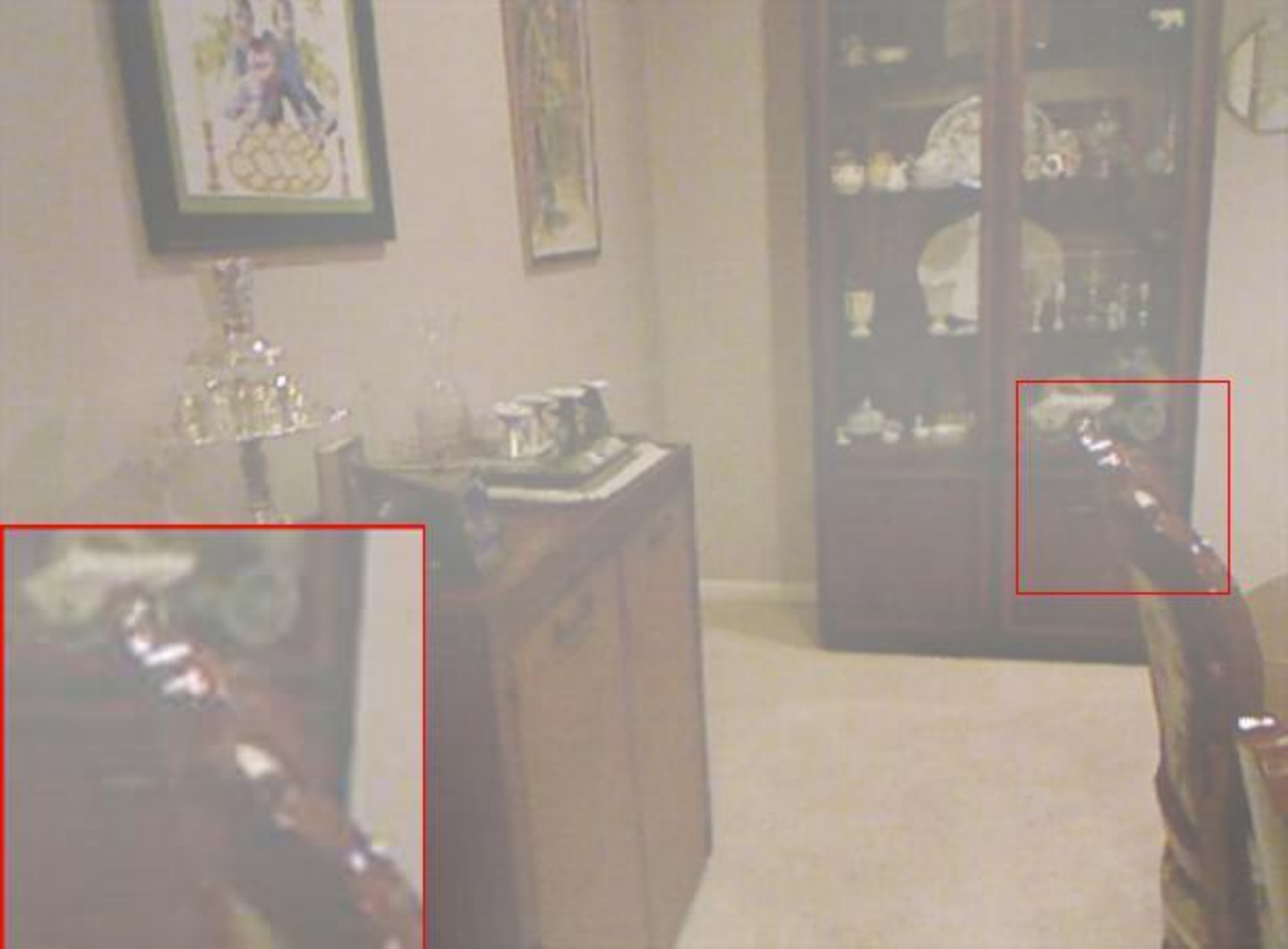}}
	\subfigure[]{\label{Transfer:d_i2i}\includegraphics[width=\m_width\textwidth, height=\a_height]{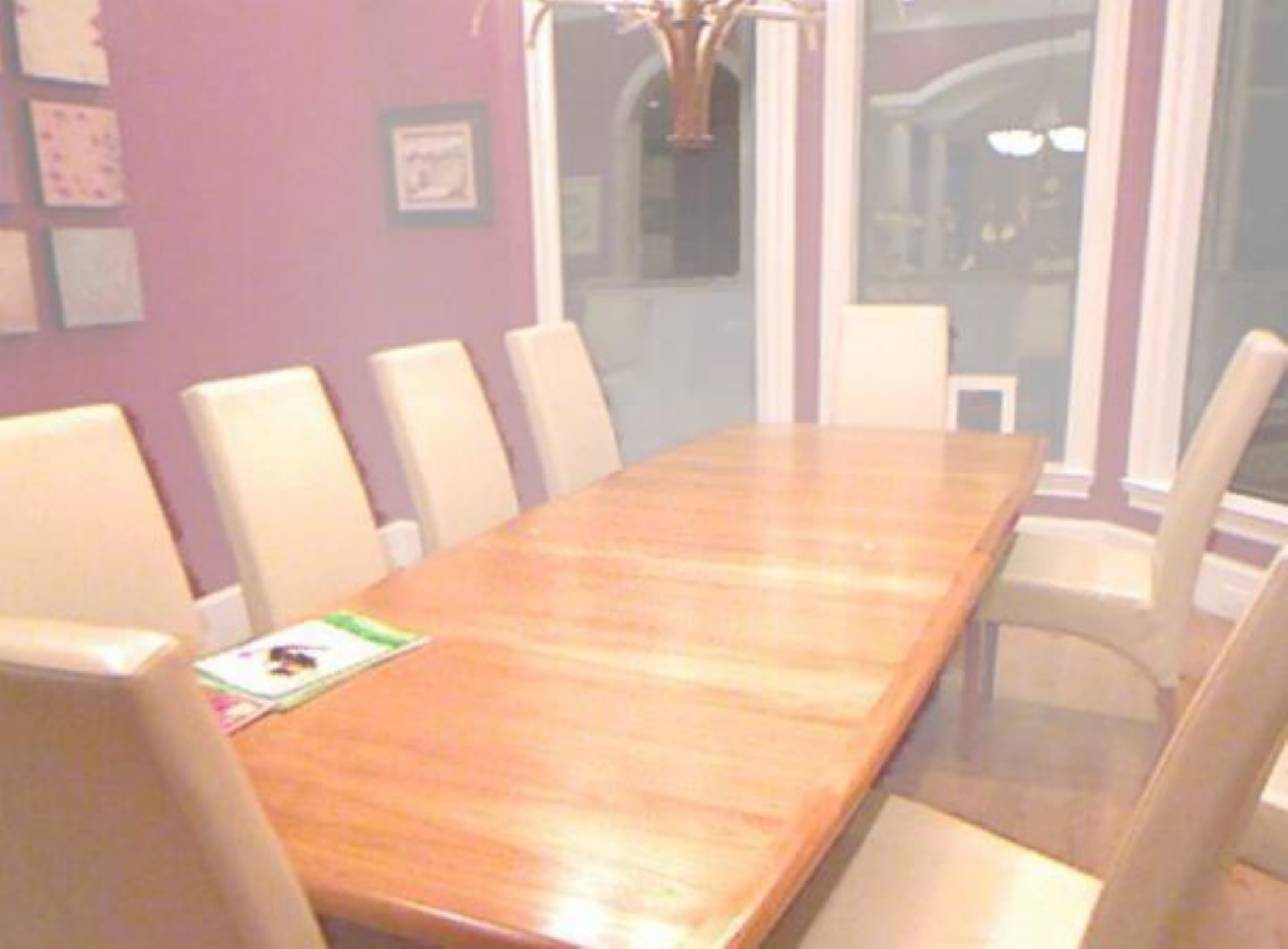}}	
	\end{center}
	
	\caption{\label{Figure:Transfer_i2i} Haze transferring from \textbf{synthetic indoor hazy} images to \textbf{indoor clean} images. Fig.~\ref{Transfer:a_i2i} and \ref{Transfer:b_i2i} are the ground truth clean image and the corresponding handcrafted hazy image. Fig.~\ref{Transfer:c_i2i} is our hazy image synthesized by transferring the haze from Fig.~\ref{Transfer:d_i2i} to Fig.~\ref{Transfer:a_i2i}. From the result, one could find that the handcrafted hazy images show some distortions due to the wrong depth information, \textit{i.e.}, the area near to the border of the piano in the first image and the chair near to the camera in the third and fourth images. There are also some unreal haze distribution in the handcrafted hazy image, \textit{i.e.}, the area near to the camera in the second image. Some areas are highlighted by red rectangles and zooming-in is recommended for a better visualization and comparison. } 

\end{figure}

\begin{figure}[!t]
	\def \m_width{0.11}
	\def \a_height{1.5cm}
	\begin{center}
	\setcounter{subfigure}{0}
	\subfigure[]{\label{Transfer:a_i2o}\includegraphics[width=\m_width\textwidth, height=\a_height]{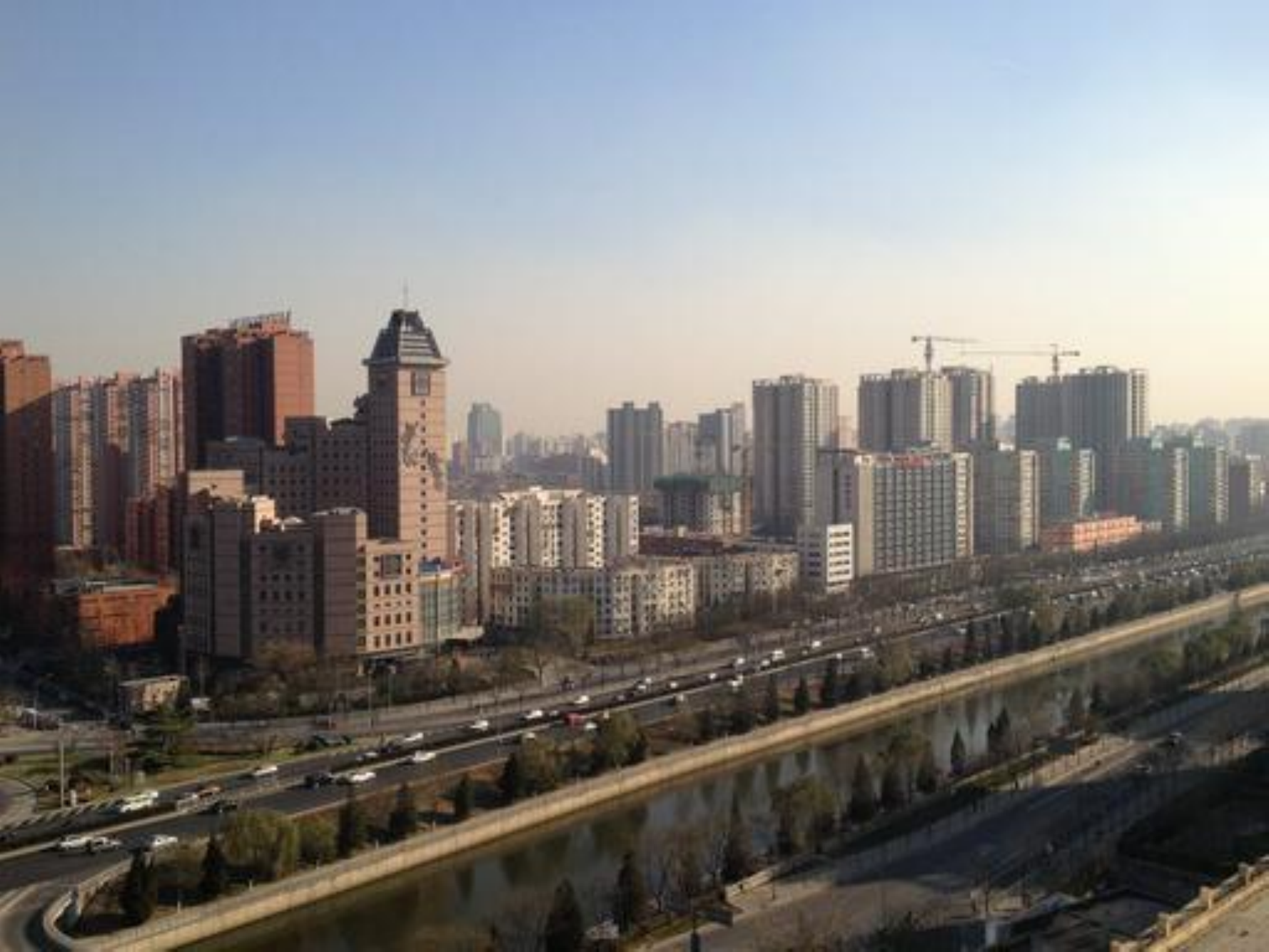}}
	\subfigure[]{\label{Transfer:b_i2o}\includegraphics[width=\m_width\textwidth, height=\a_height]{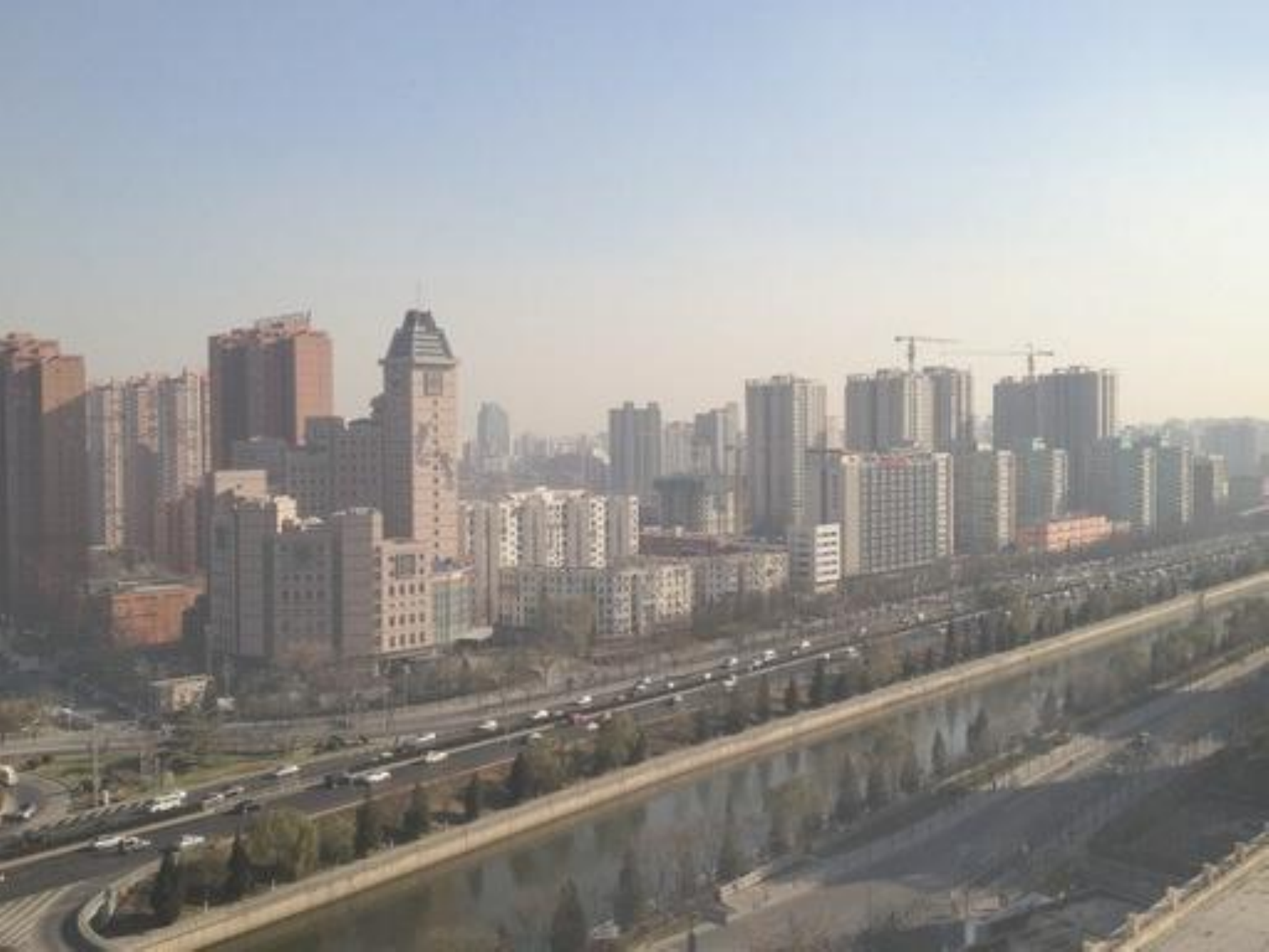}}
	\subfigure[]{\label{Transfer:c_i2o}\includegraphics[width=\m_width\textwidth, height=\a_height]{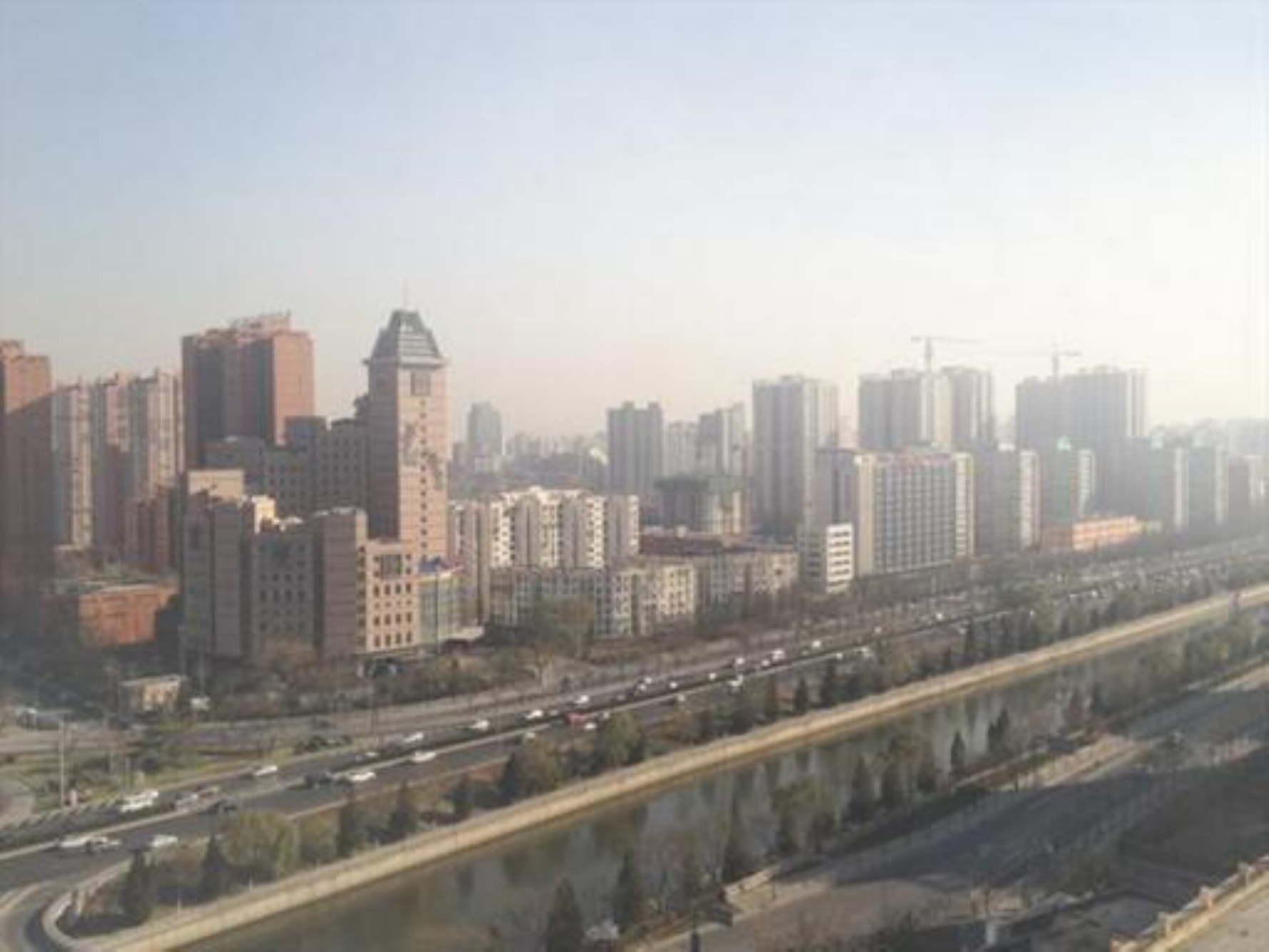}}
	\subfigure[]{\label{Transfer:d_i2o}\includegraphics[width=\m_width\textwidth, height=\a_height]{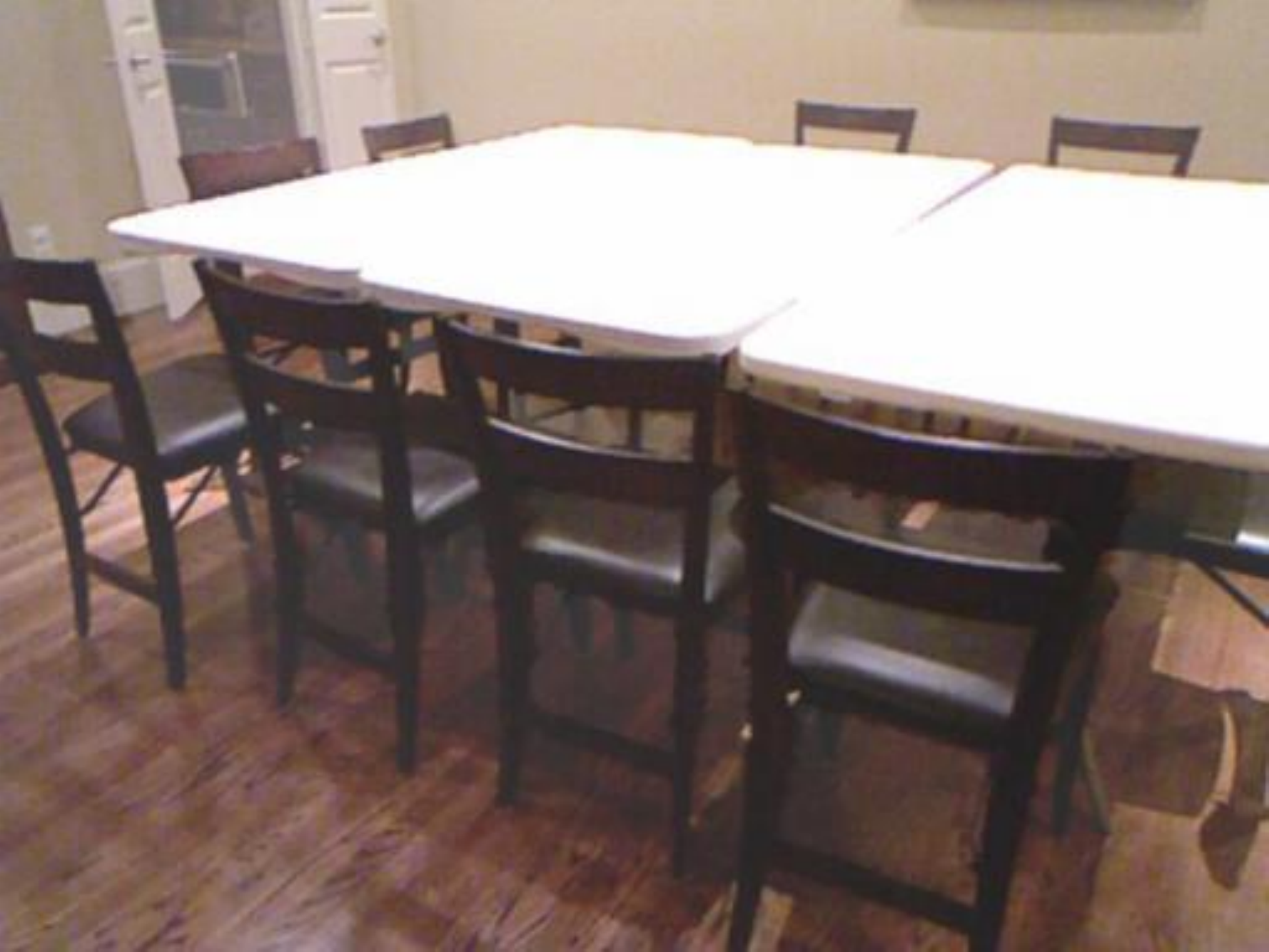}}
	\end{center}
	
	\caption{\label{Figure:Transfer_i2o} Haze transferring from \textbf{synthetic indoor hazy} images to \textbf{outdoor clean} images.  Fig.~\ref{Transfer:a_i2o} and \ref{Transfer:b_i2o} are the ground truth clean image and the corresponding handcrafted hazy image. Fig.~\ref{Transfer:c_i2o} is our hazy image synthesized by transferring the haze from Fig.~\ref{Transfer:d_i2o} to Fig.~\ref{Transfer:a_i2o}. One could find out that the haze is average distributed in the handcrafted hazy image. In contrast, the transferred hazy images show that the haze is light near the camera and heavy in other areas. Namely, our transferred hazy images is more real than the handcrafted ones.
	}

\end{figure}

\begin{figure}[!t]
	\def \m_width{0.11}
	\def \a_height{3cm}

	\begin{center}
	\subfigure{\includegraphics[width=\m_width\textwidth, height=\a_height]{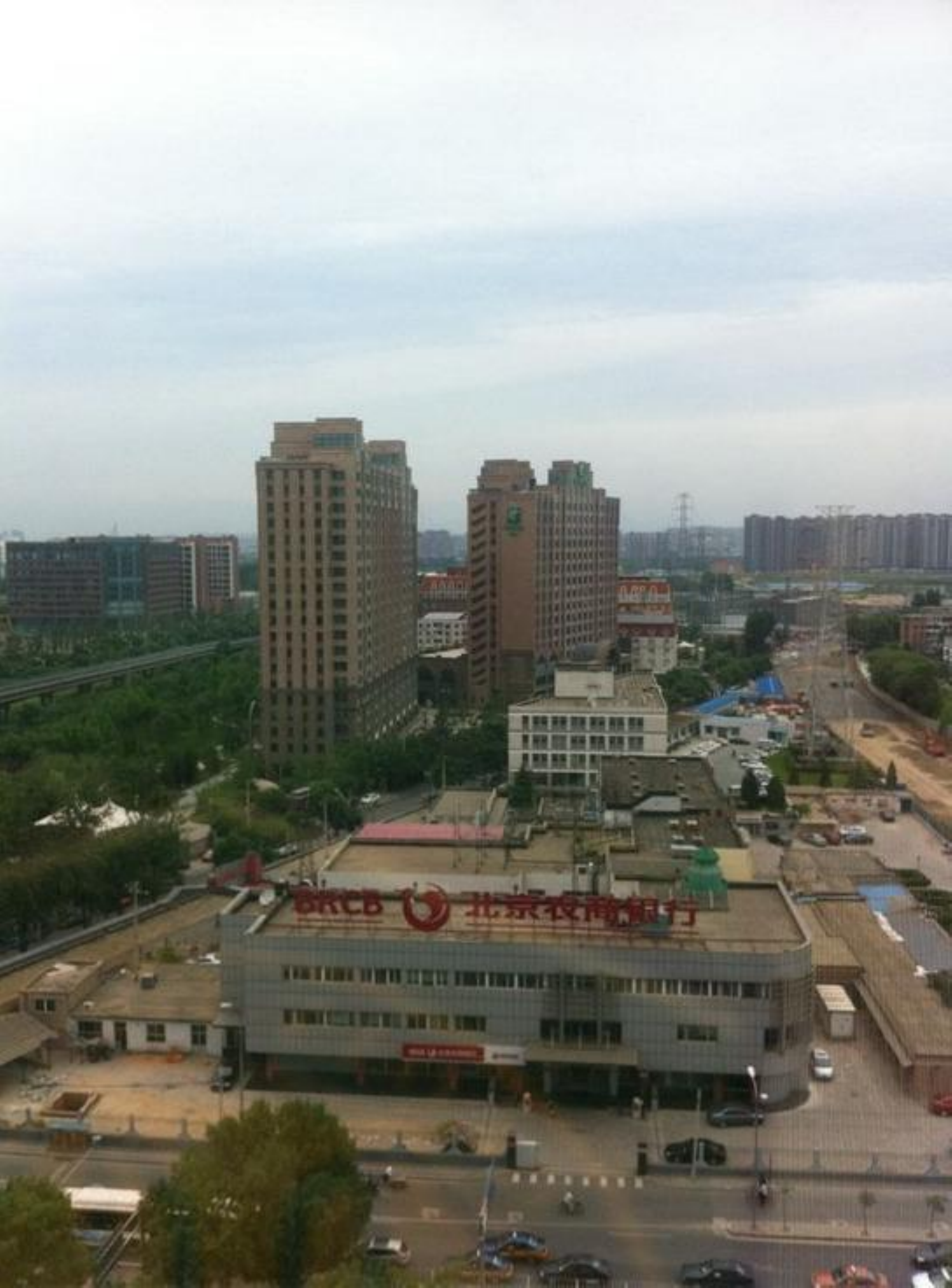}}
	\subfigure{\includegraphics[width=\m_width\textwidth, height=\a_height]{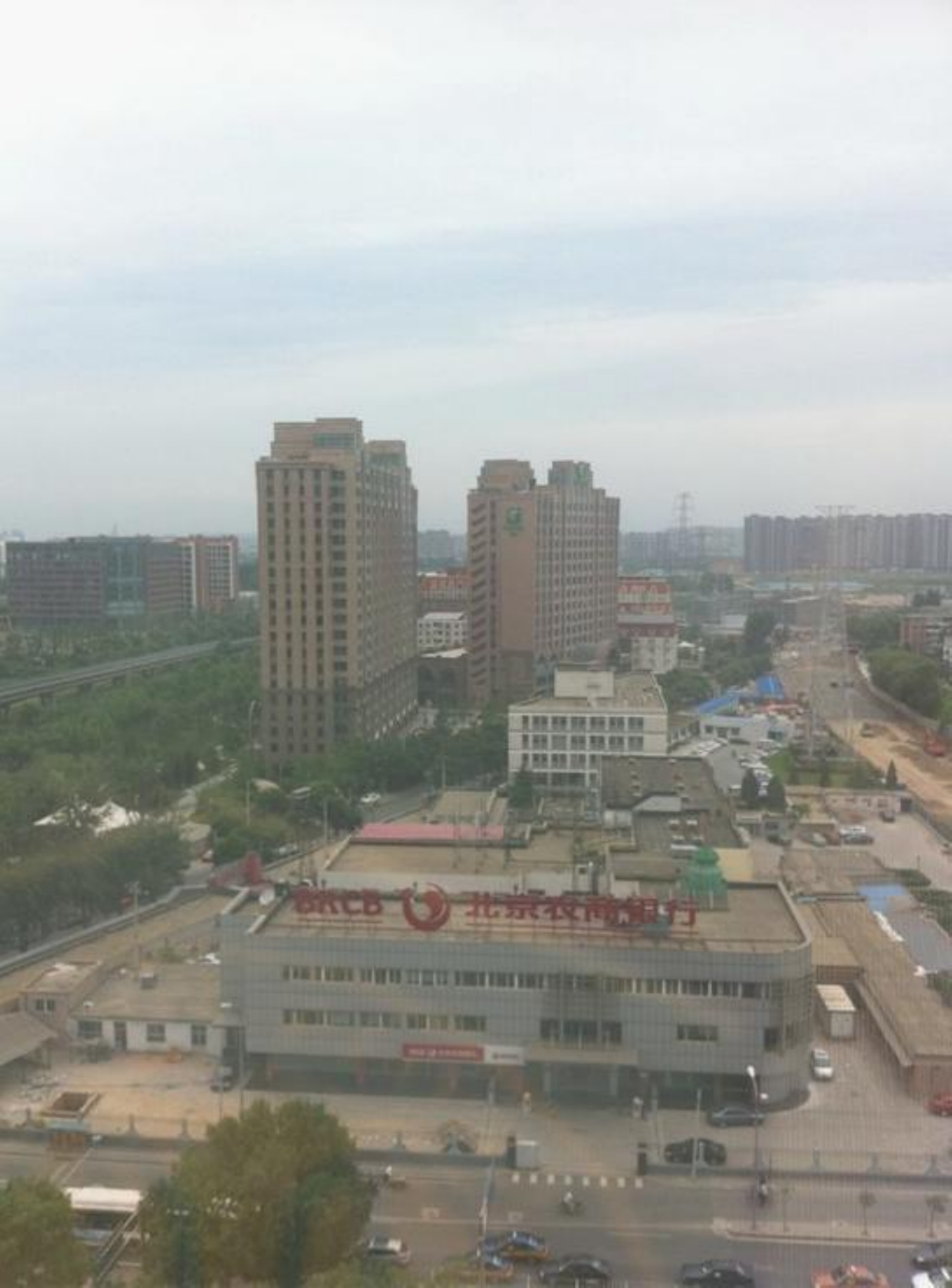}}
	\subfigure{\includegraphics[width=\m_width\textwidth, height=\a_height]{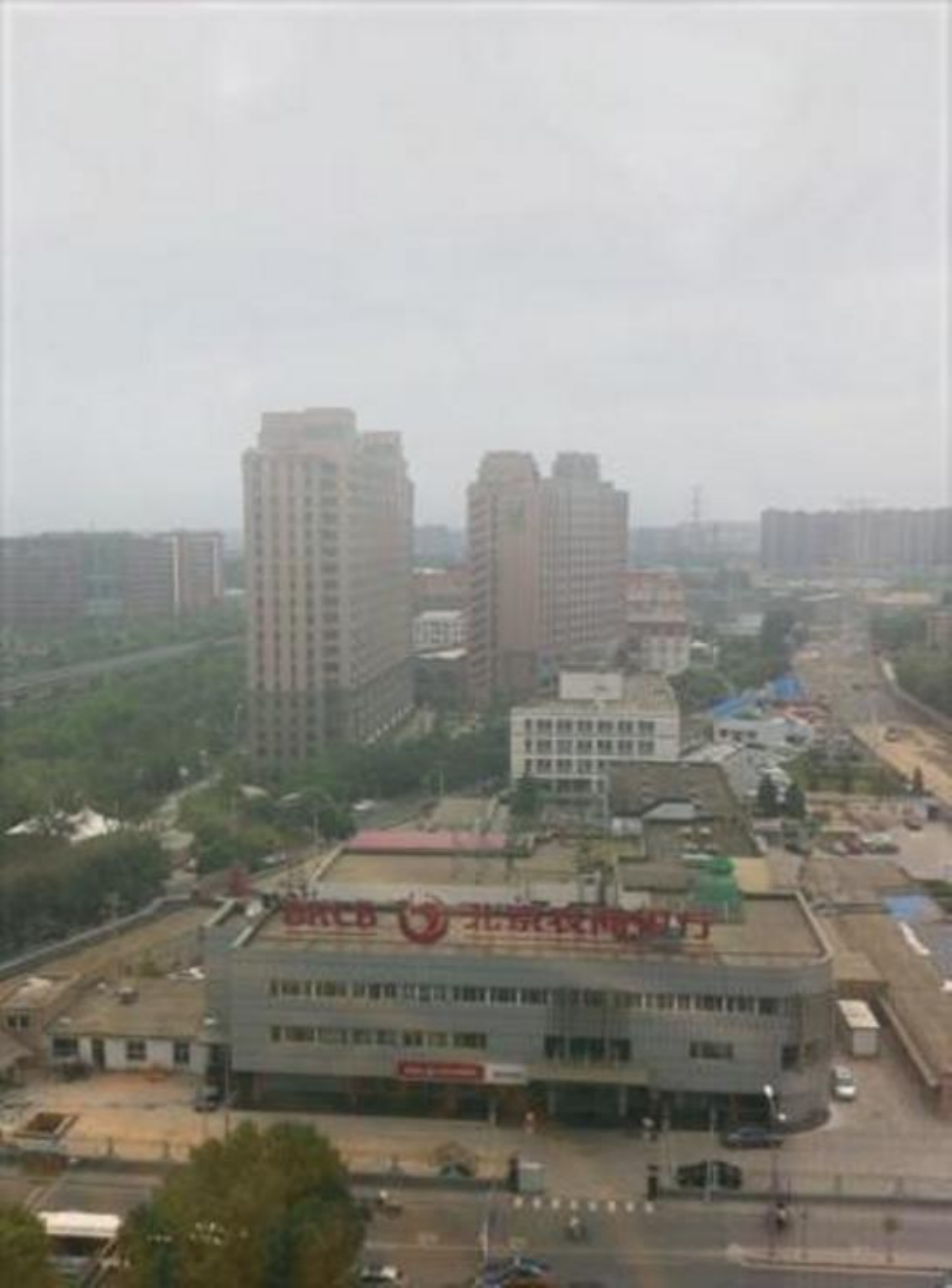}}
	\subfigure{\includegraphics[width=\m_width\textwidth, height=\a_height]{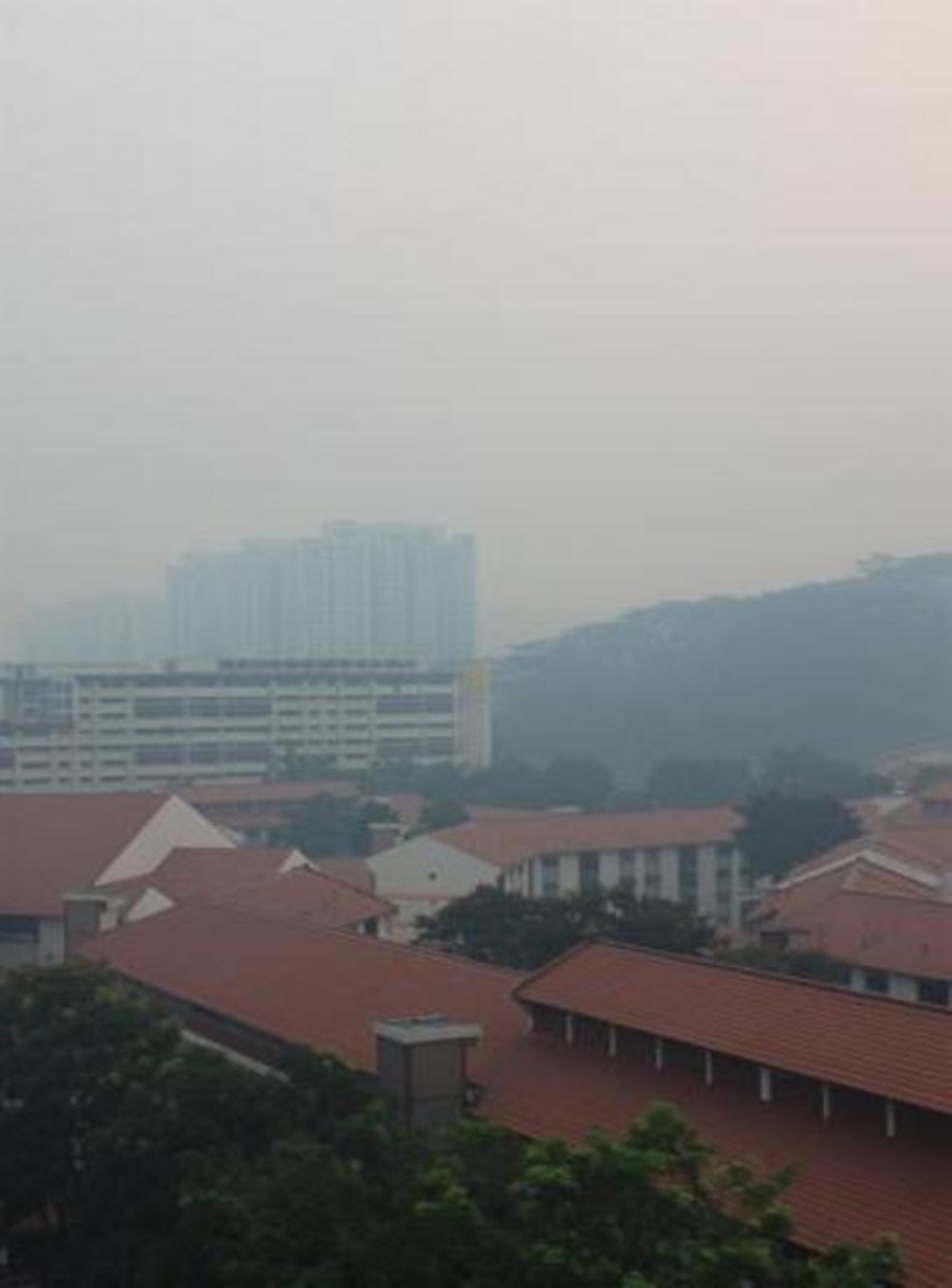}}
	\end{center}
	\vspace{-0.7cm}
	
	\def \m_width{0.11}
	\def \a_height{1.5cm}
	\begin{center}
	\subfigure{\includegraphics[width=\m_width\textwidth, height=\a_height]{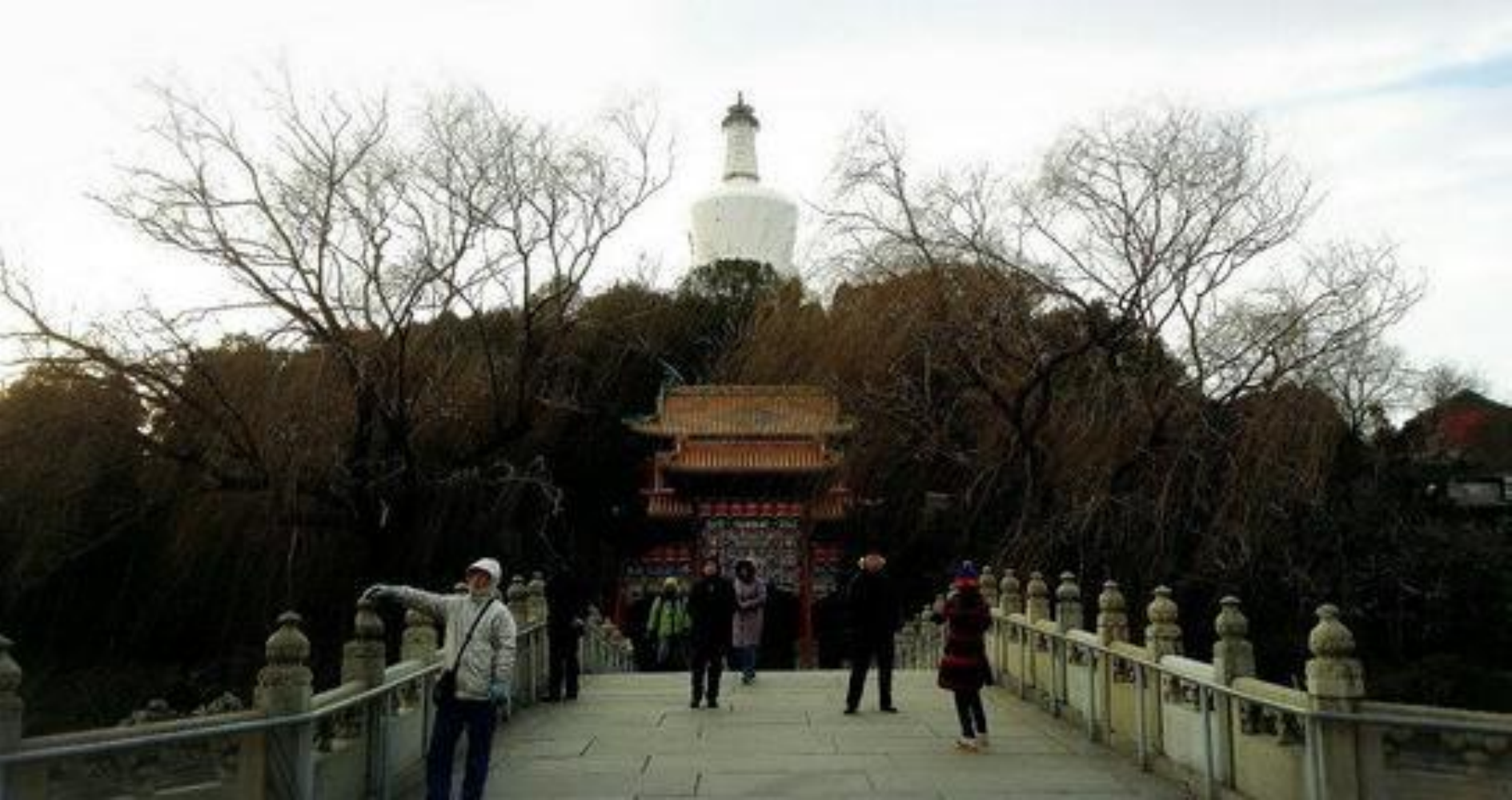}}
	\subfigure{\includegraphics[width=\m_width\textwidth, height=\a_height]{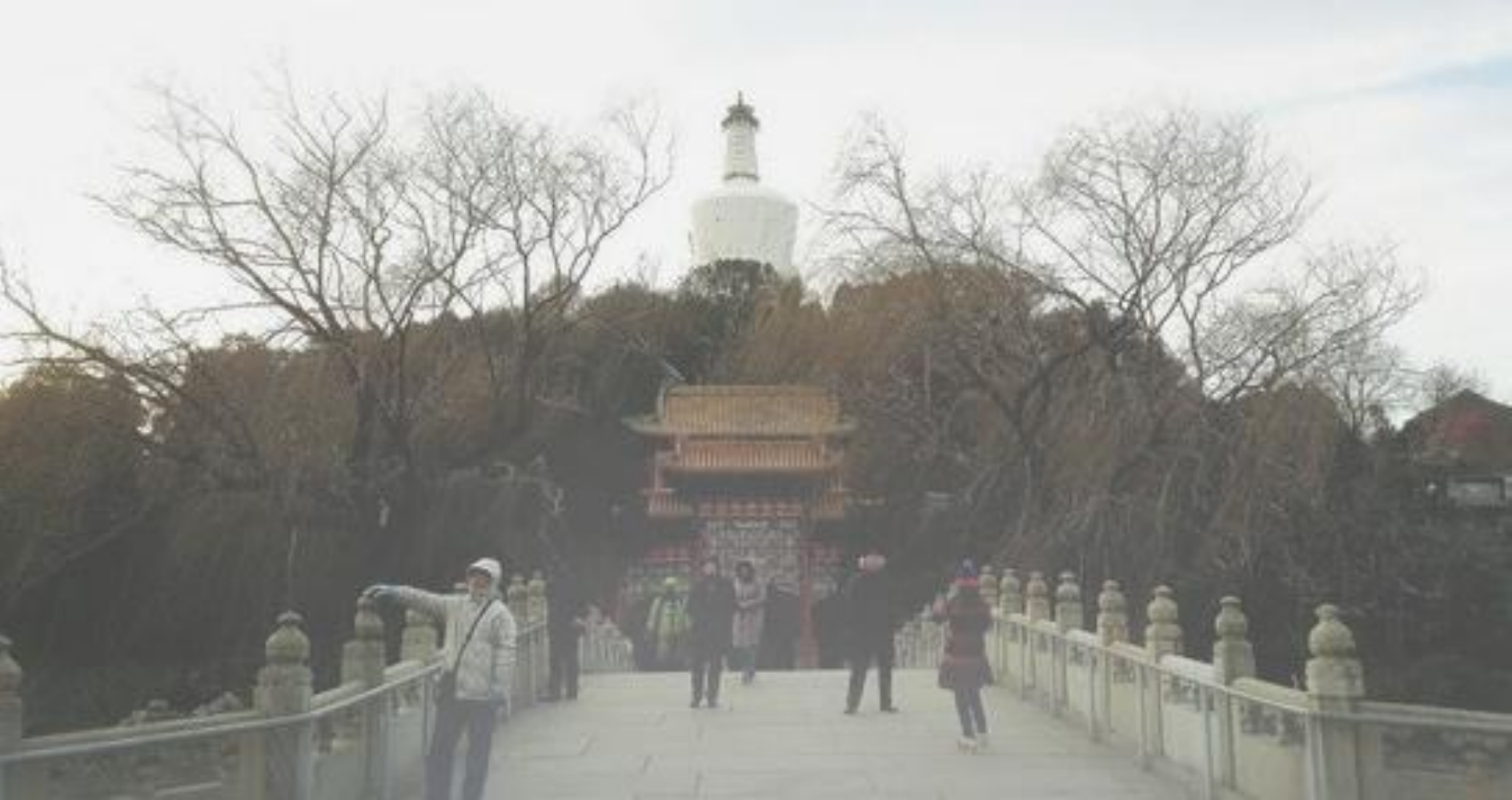}}
	\subfigure{\includegraphics[width=\m_width\textwidth, height=\a_height]{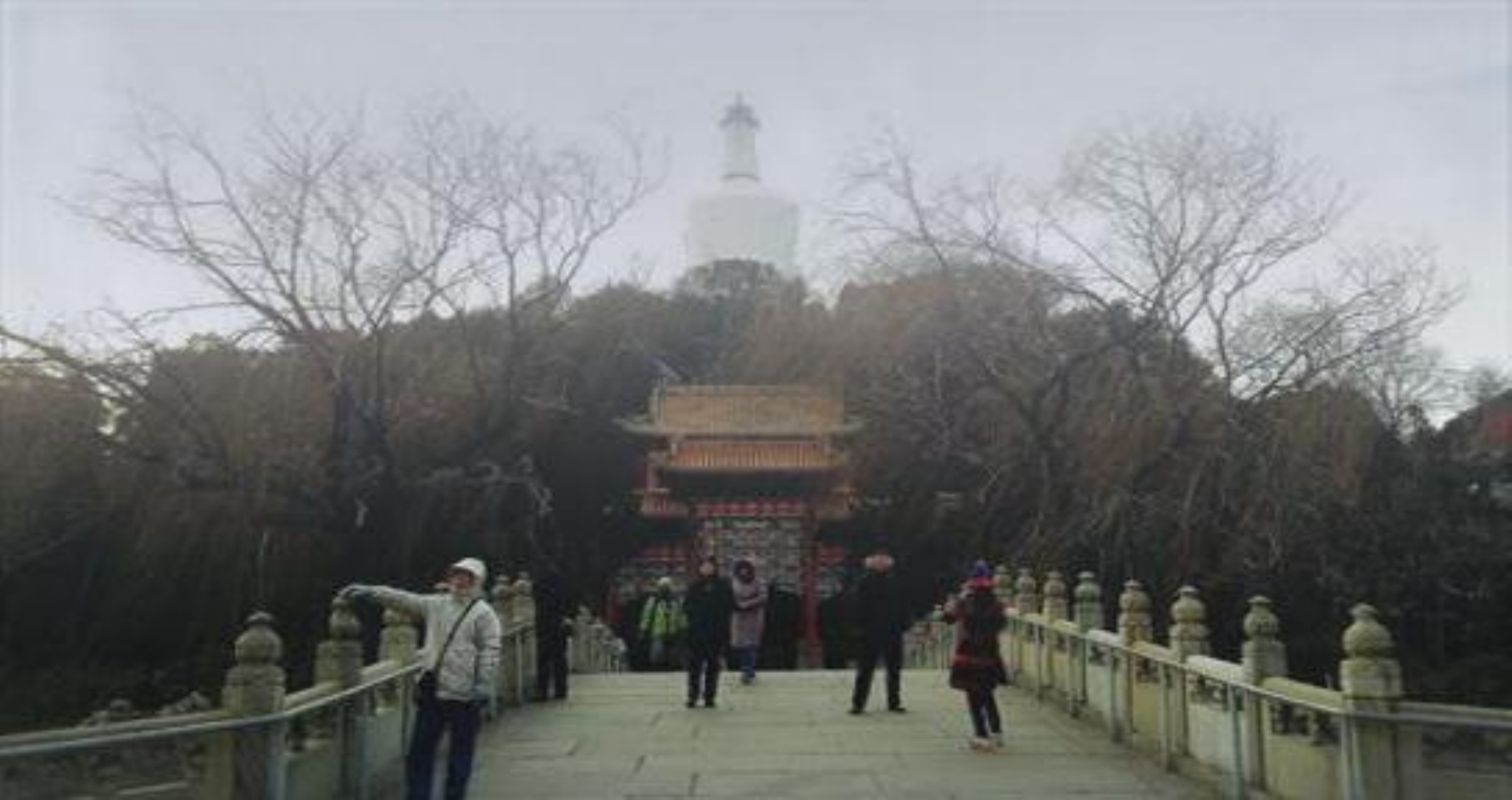}}
	\subfigure{\includegraphics[width=\m_width\textwidth, height=\a_height]{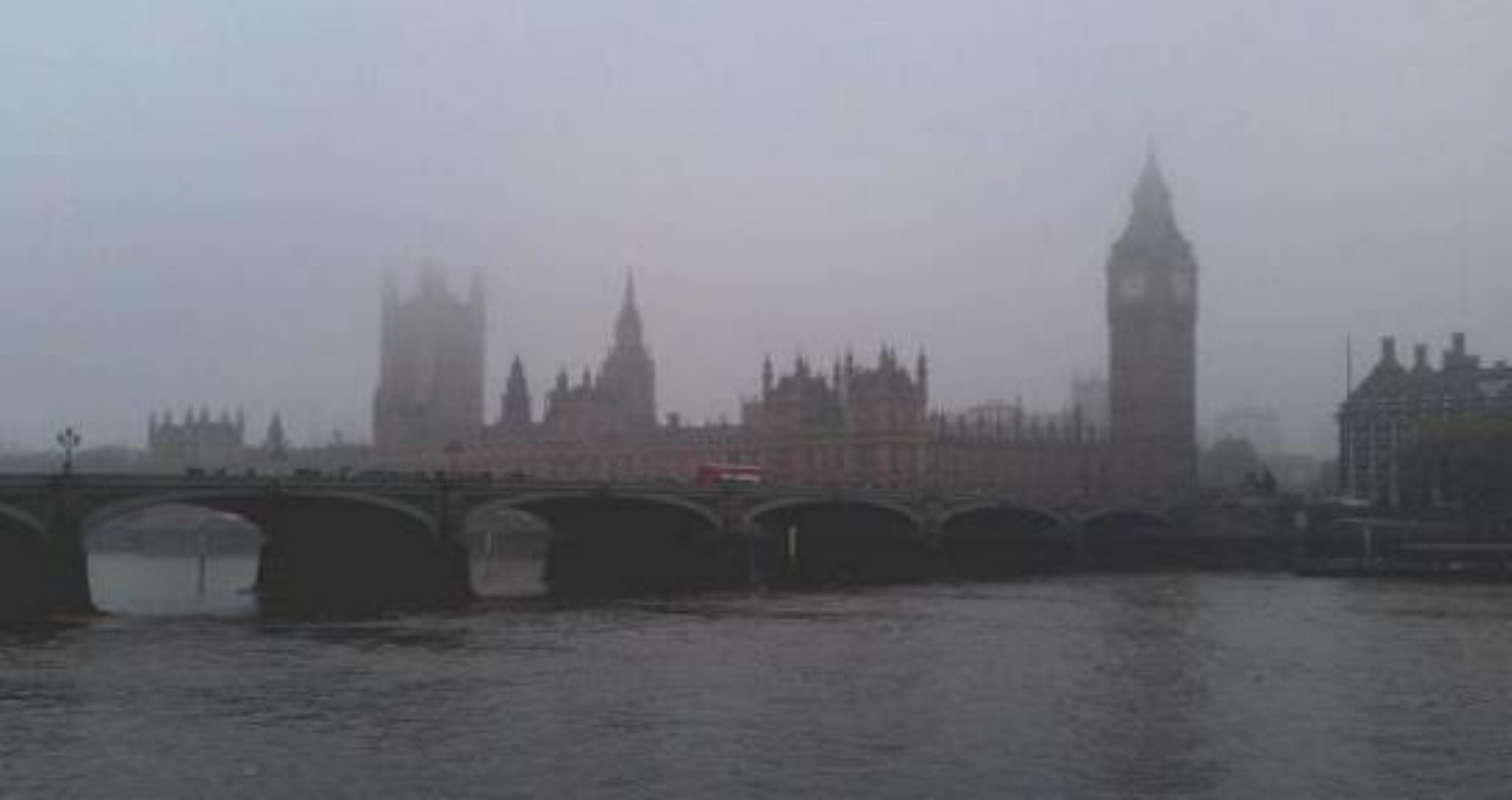}}
	\end{center}
	\vspace{-0.7cm}
	
	\begin{center}
	\subfigure{\includegraphics[width=\m_width\textwidth, height=\a_height]{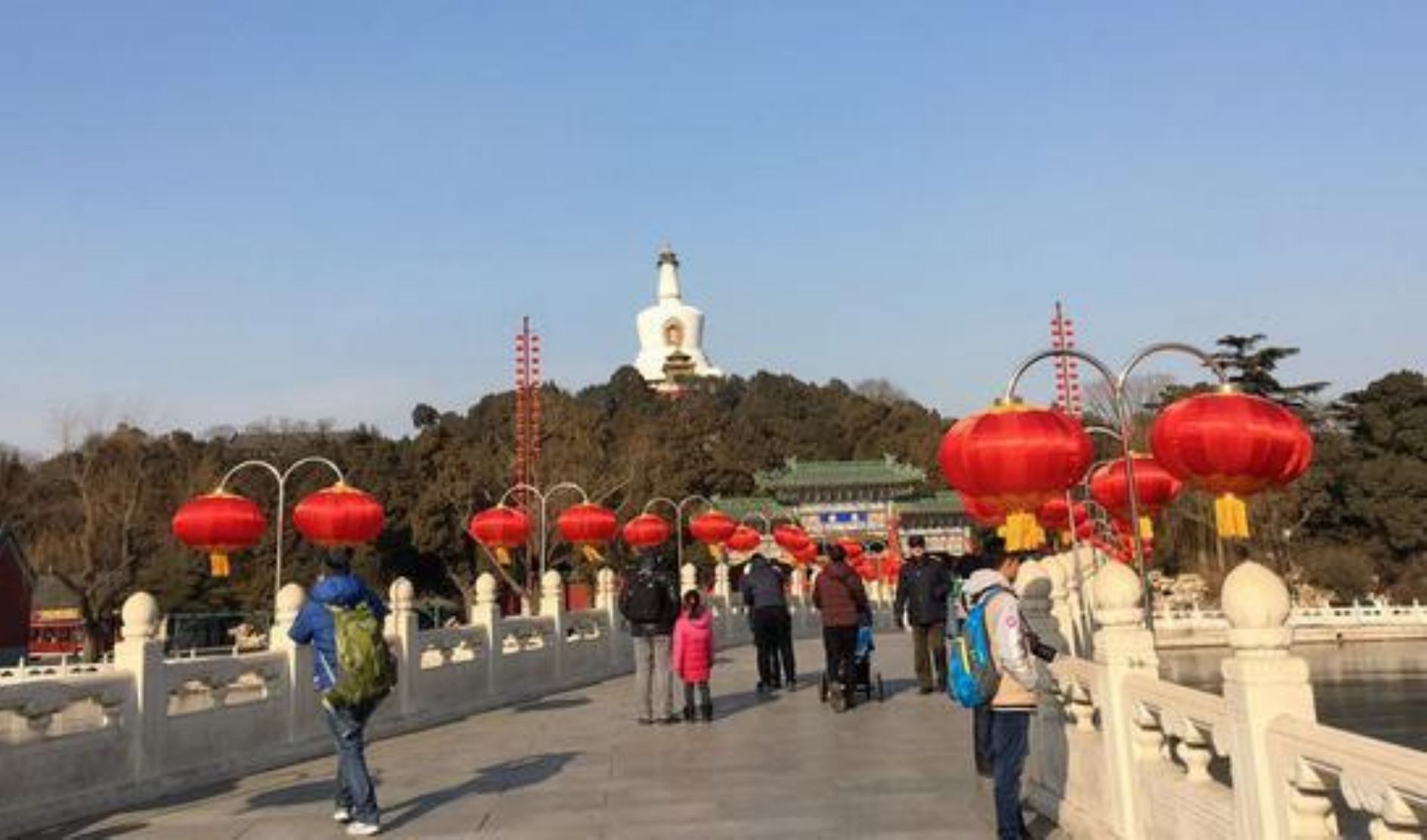}}
	\subfigure{\includegraphics[width=\m_width\textwidth, height=\a_height]{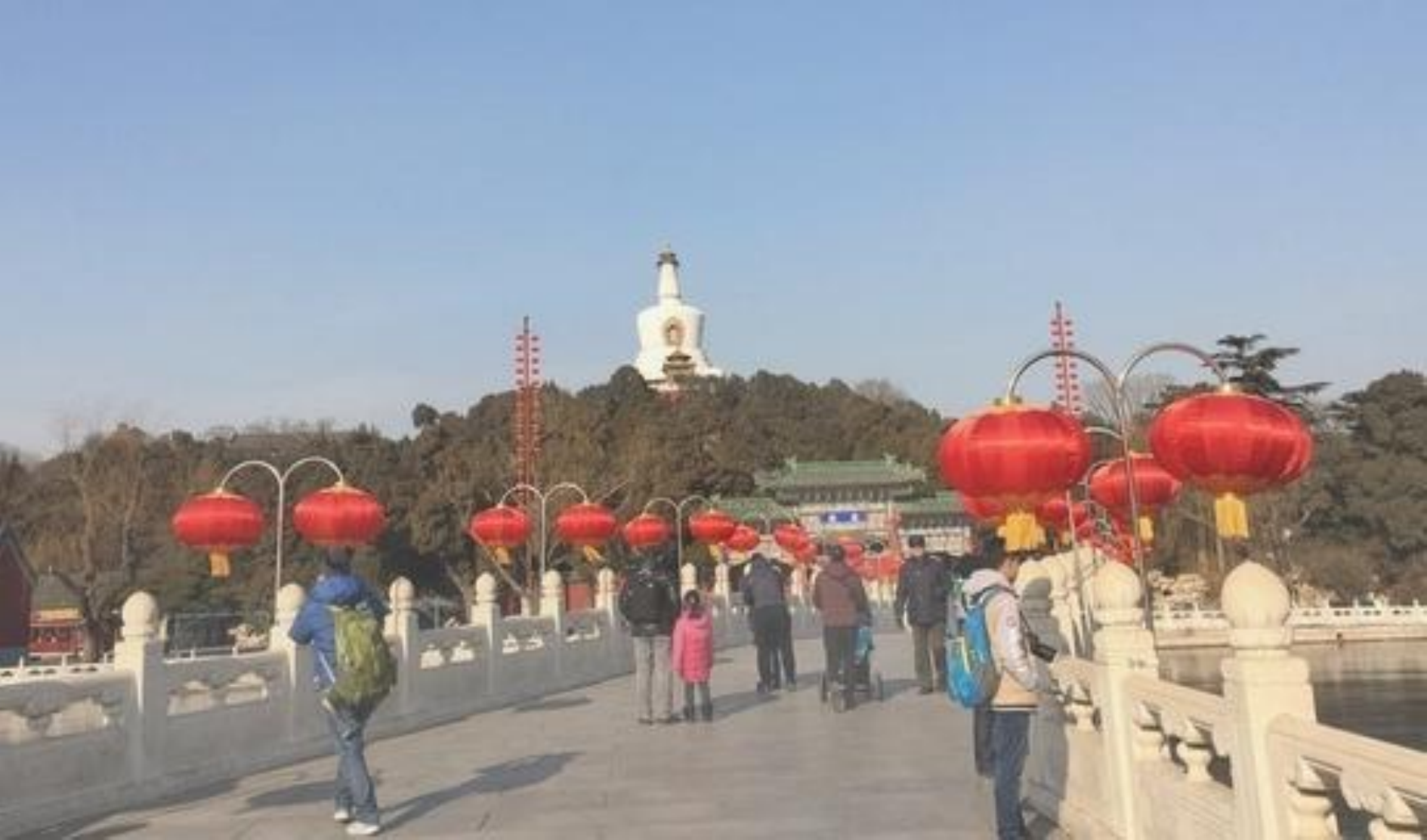}}
	\subfigure{\includegraphics[width=\m_width\textwidth, height=\a_height]{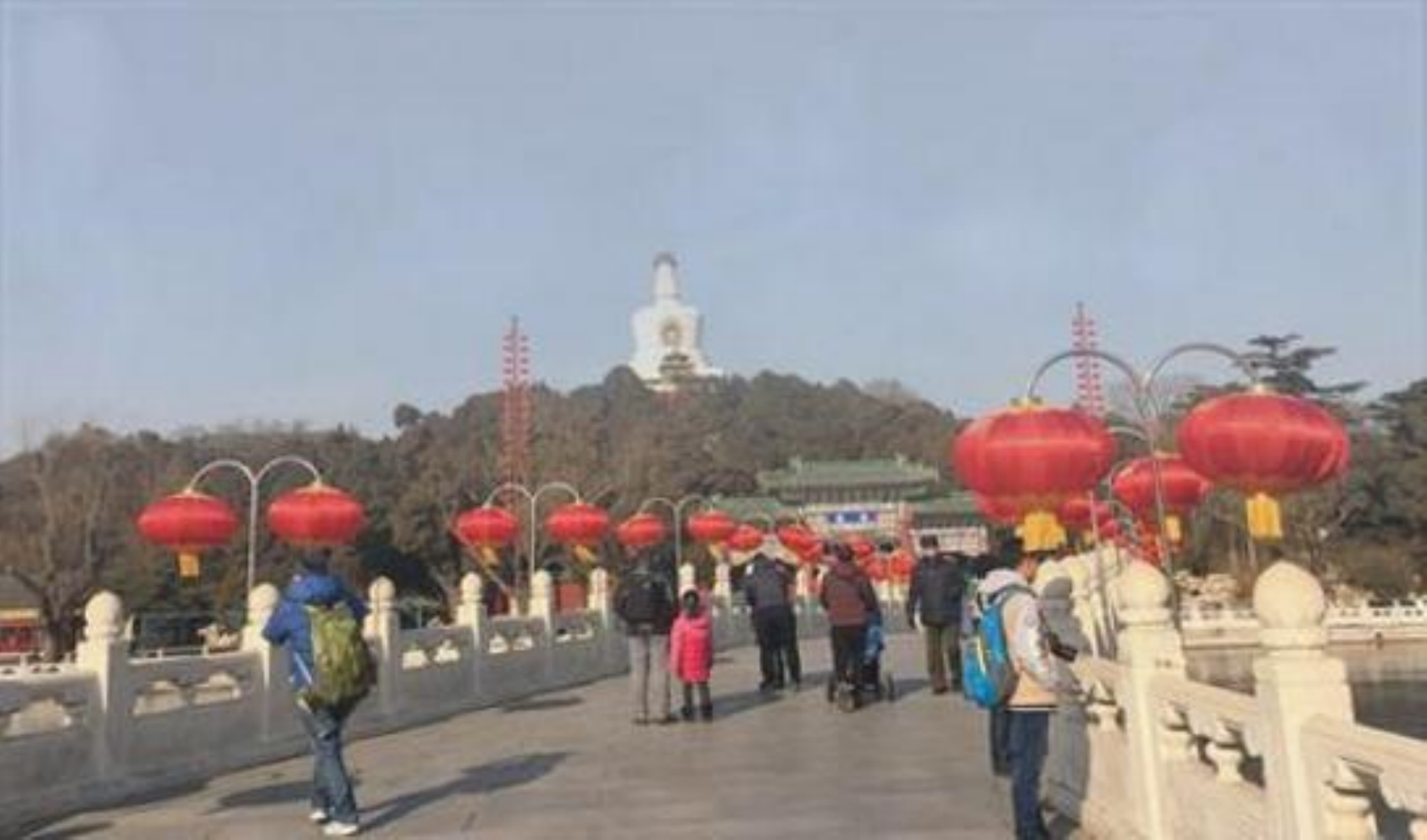}}
	\subfigure{\includegraphics[width=\m_width\textwidth, height=\a_height]{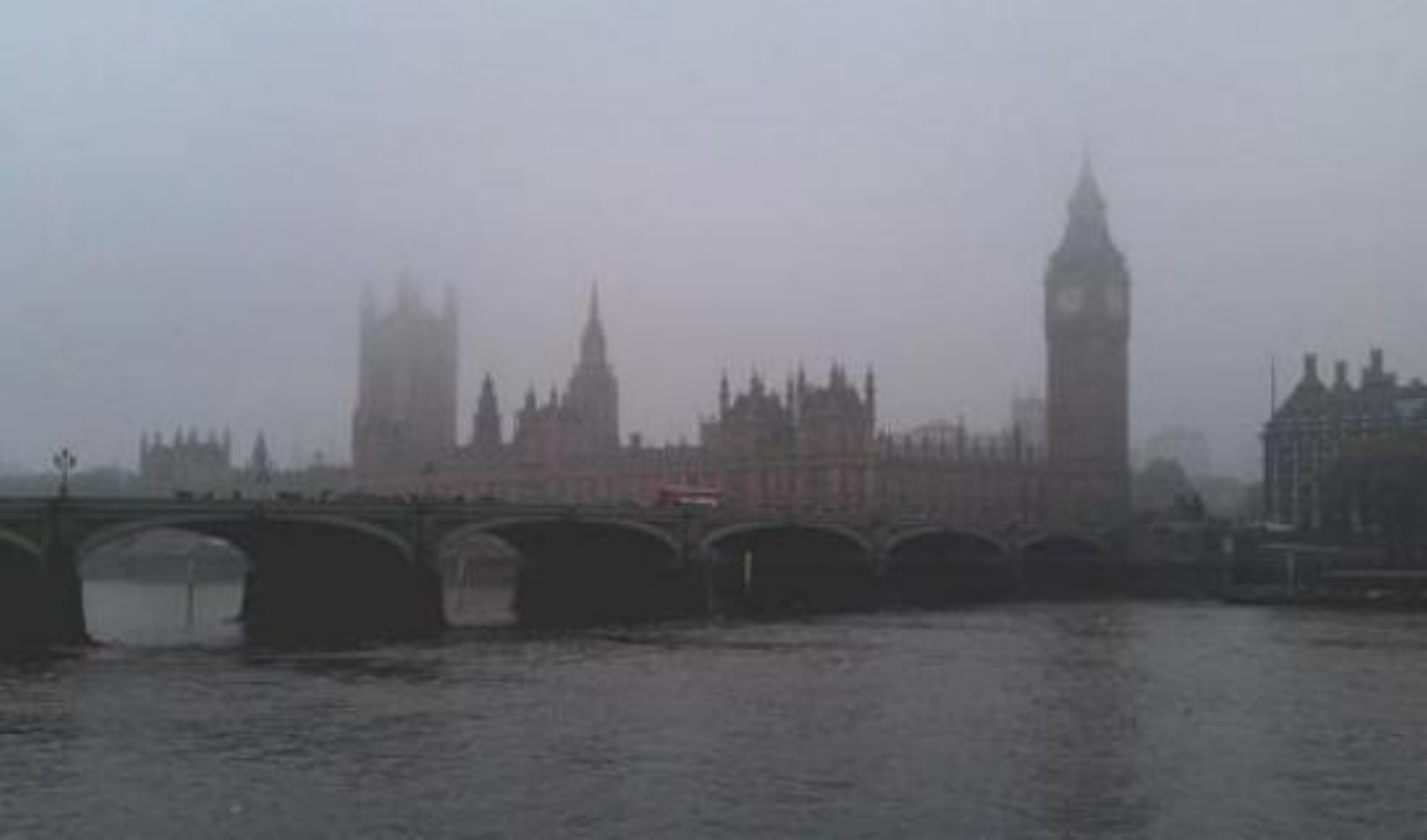}}
	\end{center}
	\vspace{-0.7cm}

	\begin{center}
	\setcounter{subfigure}{0}
	\subfigure[]{\label{Transfer:a_r2o}\includegraphics[width=\m_width\textwidth, height=\a_height]{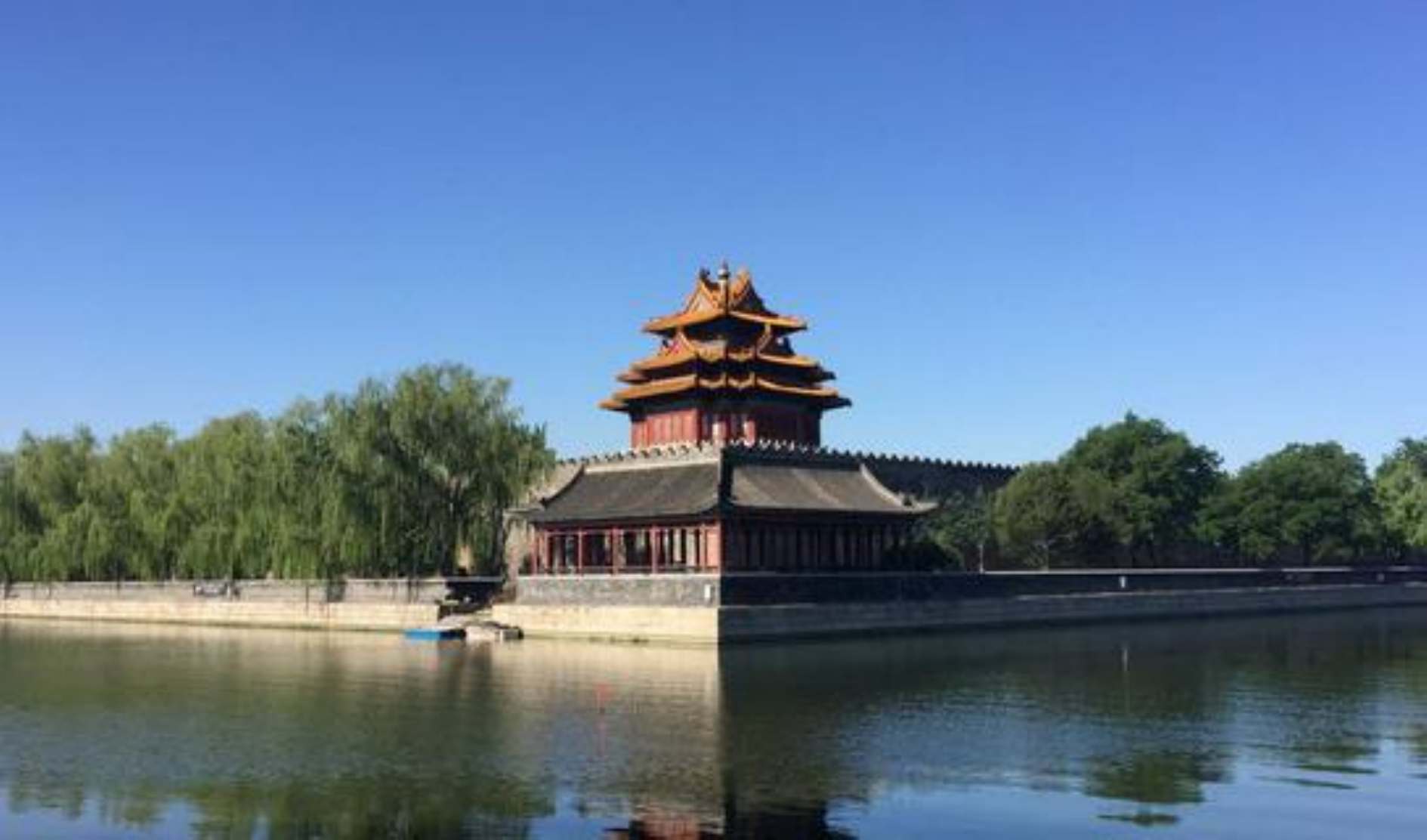}}
	\subfigure[]{\label{Transfer:b_r2o}\includegraphics[width=\m_width\textwidth, height=\a_height]{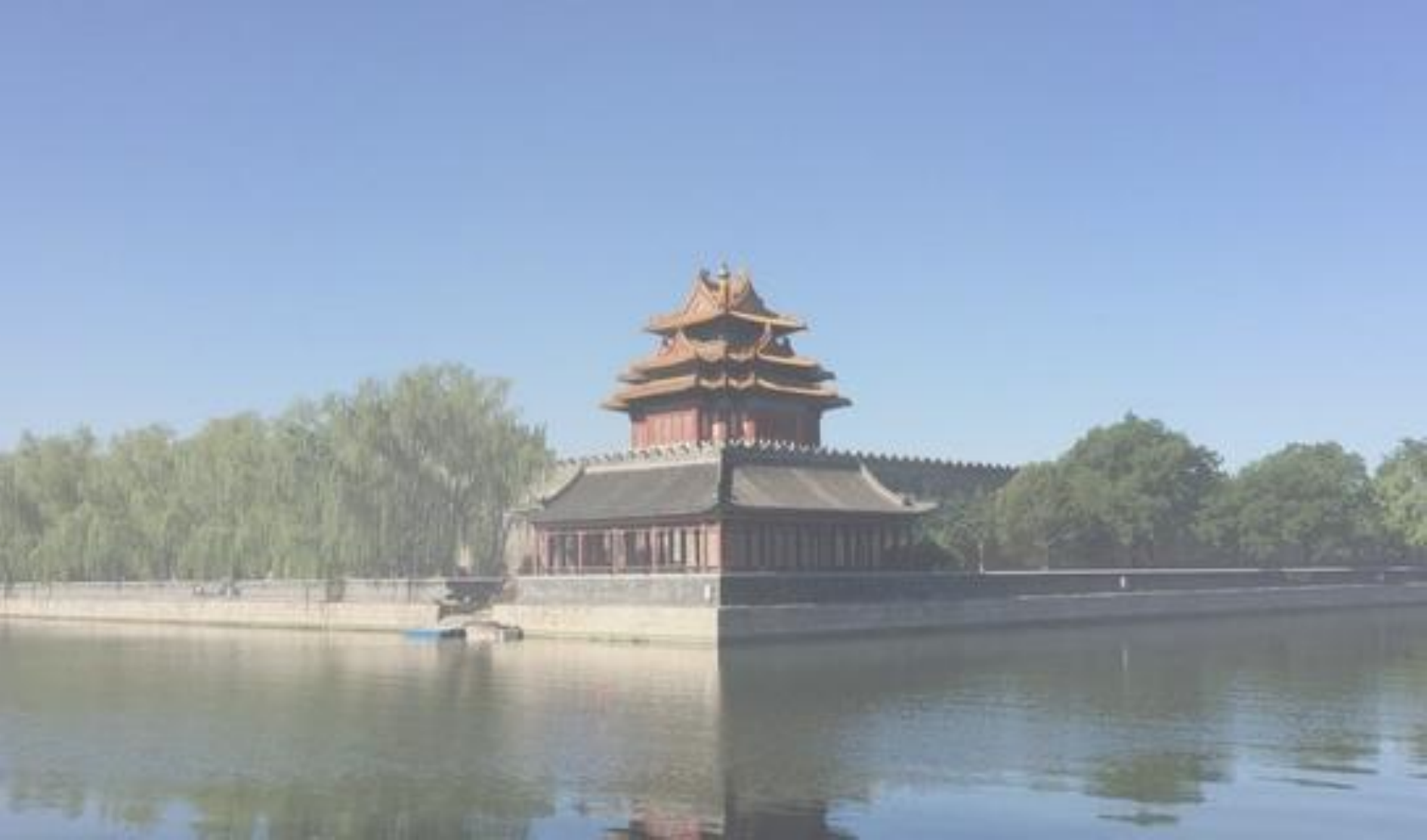}}
	\subfigure[]{\label{Transfer:c_r2o}\includegraphics[width=\m_width\textwidth, height=\a_height]{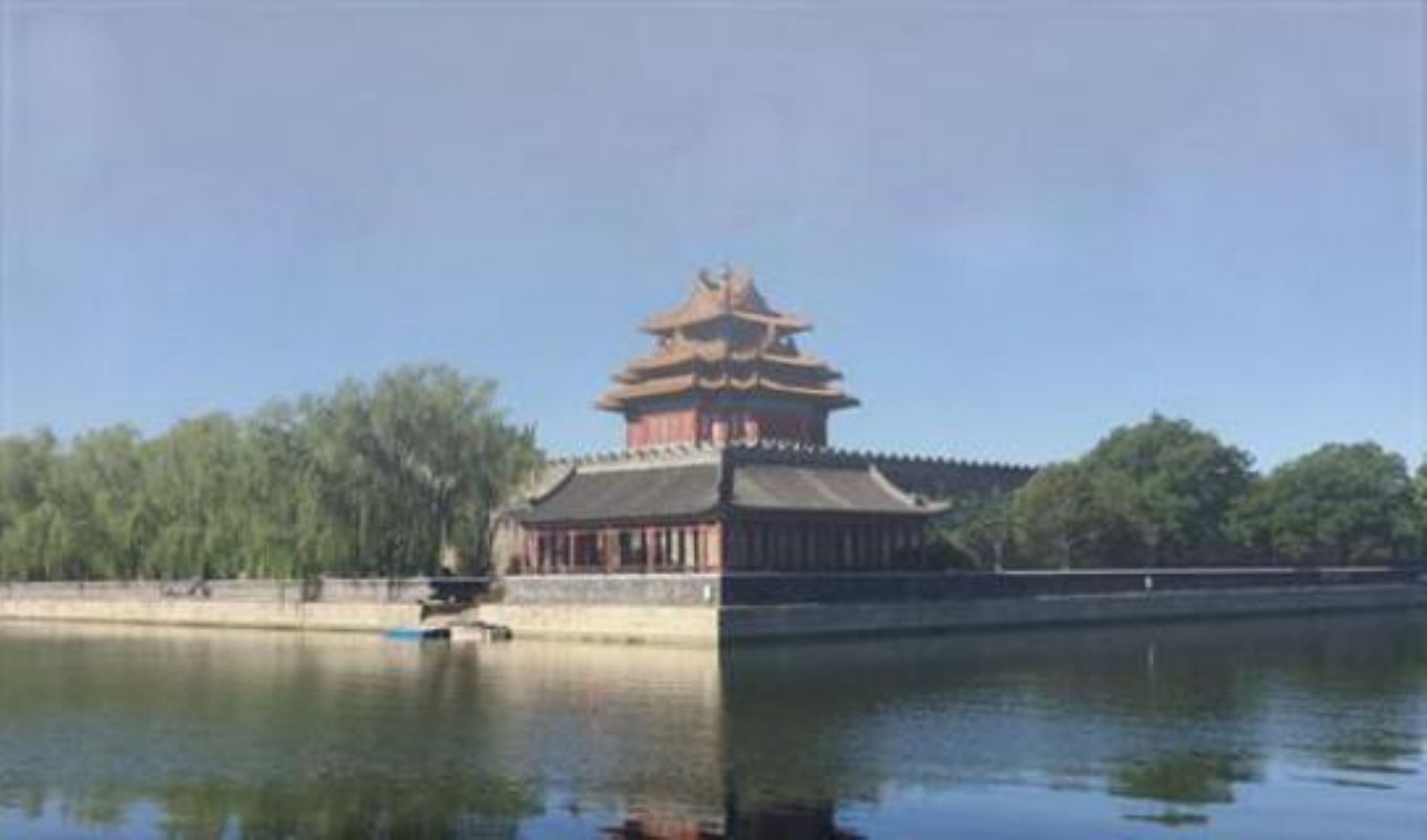}}
	\subfigure[]{\label{Transfer:d_r2o}\includegraphics[width=\m_width\textwidth, height=\a_height]{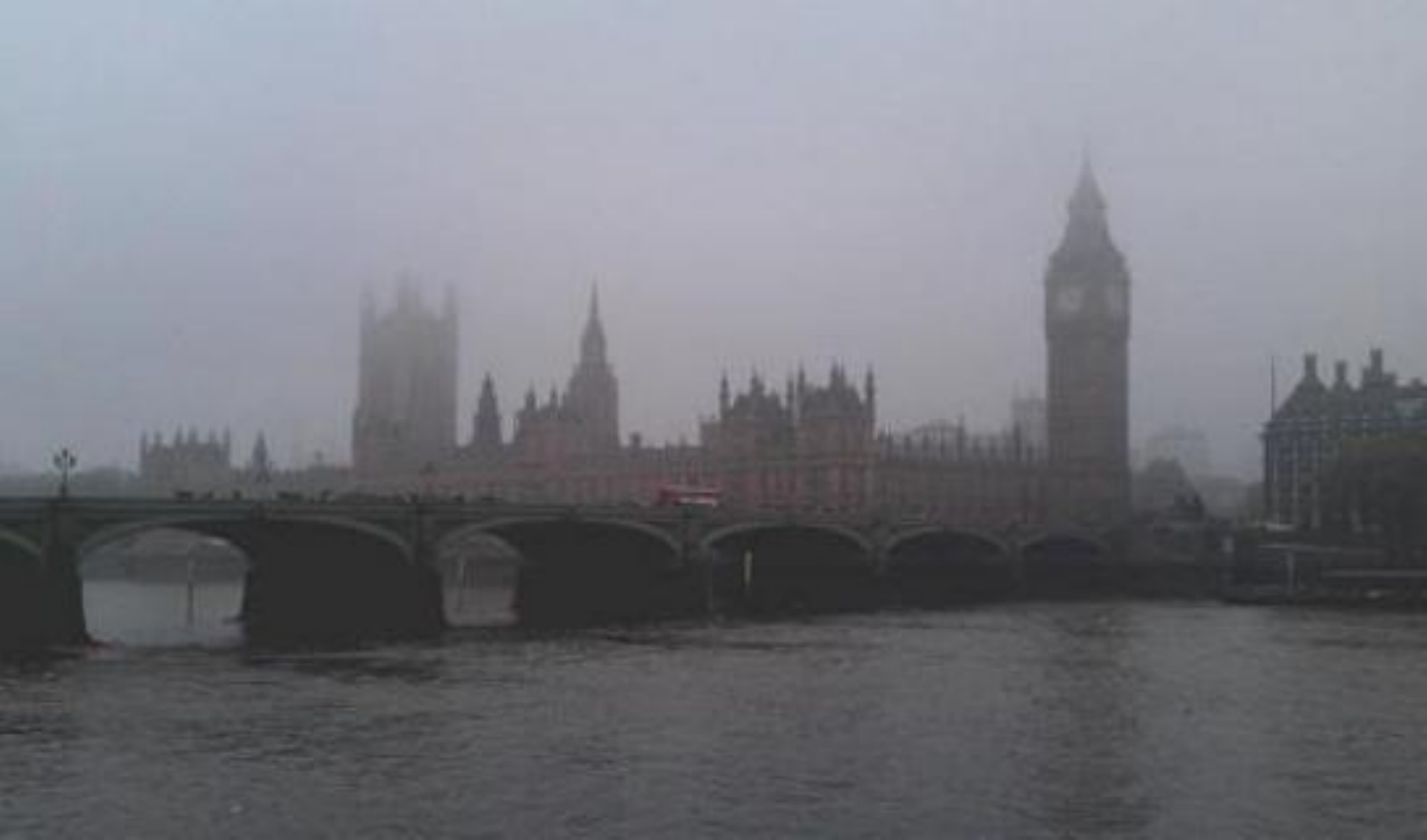}}	
	
	\end{center}

		\caption{\label{Figure:Transfer_r2o} Haze transferring from \textbf{real-world hazy} images to \textbf{outdoor clean} images.  Fig.~\ref{Transfer:a_r2o} and \ref{Transfer:b_r2o} are the ground truth clean image and the corresponding handcrafted hazy image. Fig.~\ref{Transfer:c_r2o} is our hazy image synthesized by transferring the haze from Fig.~\ref{Transfer:d_r2o} to Fig.~\ref{Transfer:a_r2o}. From the result, one  could find out that there are a lot unreal haze distributions in the handcrafted hazy images. For example, the second image shows a heavy haze near the camera and light haze in the sky area, which is inconsistent with the real-world cases. As a contrast, our hazy image is faithful to such common knowledge.
		}
\end{figure}

To demonstrate the effectiveness of our method in haze transfer, we conduct the following experiments, \textit{i.e.}, transferring haze 1) from an indoor image to another indoor clean image, see Fig.~\ref{Figure:Transfer_i2i}; 2) from an indoor image to another outdoor clean image, see Fig.~\ref{Figure:Transfer_i2o}; 3) from an outdoor image to another outdoor clean image. From the results, one could find that our method could successfully transfer the haze as expected. The performance dominance of YOLY is more distinct in Fig.~\ref{Figure:Transfer_r2o}, where the haze source comes from the real world. More specifically, one could observe that the haze nearby the camera is much heavier than other areas in the second image of Fig.~\ref{Transfer:b_r2o}. The major reason might attribute to the difficulty in estimating the depth information of outdoor image. In contrast, the hazy images generated by our method less suffers from this issue, which shows the promising haze creation performance of our method.  

\subsection{Ablation Study}
To verify the effectiveness of our loss function, we conduct an ablation study on the HSTS dataset by removing one of $\mathcal{L}_{H}$, $\mathcal{L}_{KL}$,  $\mathcal{L}_{J}$, and $\mathcal{L}_{Reg}$. From Table~\ref{Table:Ablation}, one could see that: 1) our method leverages the advantages of variational inference in haze removal with the formulation of $\mathcal{L}_{H}$ and $\mathcal{L}_{KL}$; 2) the statistical information used on J-Net could improves the performance of the model; 3) the performance of YOLY is slightly improved with the regularization on the estimation of atmospheric light. 

\begin{table}
\centering
\caption{Ablation Study on the HSTS database.}
\label{Table:Ablation}
\begin{footnotesize}
\begin{tabular}{*{6}{c}}
\toprule
Metrics  & $\mathcal{L}_{H}$ & $\mathcal{L}_{KL}$ & $\mathcal{L}_{J}$ &  $\mathcal{L}_{Reg}$ & Ours \\ 
\midrule
PSNR & 21.16 & 21.68 & 22.53 & 22.28	& 23.82 \\
SSIM & 0.8733 & 0.8690 & 0.9039 & 0.8836 & 0.9125 \\ 
\bottomrule
\end{tabular}
\end{footnotesize}
\end{table}

\section{Conclusion}

In this paper, we propose an unsupervised and untrained image dehazing neural network which separates the observed hazy image into scene radiance layer, transmission map layer and atmospheric light layer. Three advantages of the proposed YOLY are: 1) unsupervised characteristic means that the method does not use the information beyond the image content; 2) untrained characteristic means that the method does not require using an image collection for training like most existing neural networks do; 3) transferable hazy capacity which could synthesize new hazy images by extracting the haze from a given image in a learning- and unsupervised-fashion. Extensive experiments on two synthesis datasets and one real-world dataset demonstrate the promising performance of our method in the quantitative and qualitative comparisons. 

Although our method remarkably outperforms most existing unsupervised shallow and deep methods, it is only comparable to the state-of-the-art supervised image dehazing approaches. Thus, it is promising to continually improve its performance in future. Moreover, it is also valuable to explore how to extend our method to other image/video enhancement tasks such as denoising, inpainting, and so on.

\bibliographystyle{spbasic}      
\bibliographystyle{spmpsci}      
\bibliographystyle{spphys}       
\bibliography{egbib.bib}   


\end{document}